\documentclass[10pt,twocolumn,letterpaper]{article}

\usepackage[pagenumbers]{cvpr} %

\usepackage[export]{adjustbox}
\usepackage{tikz}
\usepackage{pgfplots}
\usepackage{colortbl}
\usepackage[accsupp]{axessibility}

\definecolor{cvprblue}{rgb}{0.21,0.49,0.74}
\usepackage[pagebackref,breaklinks,colorlinks,allcolors=cvprblue]{hyperref}

\title{Stable Flow: Vital Layers for Training-Free Image Editing}

\author
{
    Omri Avrahami$^{1, 2}$ 
    \hspace{3mm} Or Patashnik$^{1, 3}$
    \hspace{3mm} Ohad Fried$^{4}$
    \hspace{3mm} Egor Nemchinov$^{1}$
    \vspace{0.5mm}
    \\
    Kfir Aberman$^{1}$
    \hspace{11mm} Dani Lischinski$^{2}$
    \hspace{11mm} Daniel Cohen-Or$^{1,3}$
    \vspace{1mm}
    \\
    \normalsize{$^{1}$Snap Research}
    \hspace{1mm} \normalsize{$^{2}$The Hebrew University of Jerusalem}
    \hspace{1mm} \normalsize{$^{3}$Tel Aviv University}
    \hspace{1mm} \normalsize{$^{4}$Reichman University}
}

\usepackage{fontawesome}

\newcommand{\ignorethis}[1]{}

\newcommand{\Reals      }     {{\textrm{I\kern-0.18em R}}}

\newcommand{\change     } [1] {\mbox{{\footnotesize $\Delta$} \kern-3pt}#1}

\definecolor{darkred}{rgb}{0.7,0.1,0.1}
\definecolor{darkgreen}{rgb}{0.1,0.6,0.1}
\definecolor{cyan}{rgb}{0.7,0.0,0.7}
\definecolor{otherblue}{rgb}{0.1,0.4,0.8}
\definecolor{maroon}{rgb}{0.76,.13,.28}
\definecolor{burntorange}{rgb}{0.81,.33,0}

\newif\ifdraft
\drafttrue

\ifdraft
  \newcommand{\todo}[1]{{\color{cyan}[\textbf{TODO:} #1]}}
  \newcommand{\omri}[1]{{\color{burntorange}[\textbf{Omri:} #1]}}
  \newcommand{\orpa}[1]{{\color{darkgreen}[\textbf{Or:} #1]}}
  \newcommand{\dc}[1]{{\color{red}\textbf{}#1}}
  \newcommand{\kfir}[1]{{\color{darkred}[\textbf{Kfir:} #1]}}
  \newcommand{\ohad}[1]{{\color{magenta}[\textbf{Ohad:} #1]}}
  \newcommand{\danix}[1]{{\color{blue}[\textbf{Danix:} #1]}}

\else
  \newcommand{\omri}[1]{}
  \newcommand{\orpa}[1]{}
  \newcommand{\dc}[1]{}
  \newcommand{\kfir}[1]{}
  \newcommand{\ohad}[1]{}
  \newcommand{\danix}[1]{}

  \newcommand {\note}[1]{}
  \newcommand {\todo}[1]{}
\fi

\newcommand {\prompt}[1]{{{\textit{``#1''}}}}
\newcommand {\promptstart}[1]{{{\textit{``#1}}}}
\newcommand {\promptend}[1]{{{\textit{#1''}}}}

\definecolor{vitalcolor}{RGB}{242,78,112}
\newcommand {\vital}[1]{{\color{vitalcolor}{#1}}}
\definecolor{nonvitalcolor}{RGB}{41,160,177}
\newcommand {\nonvital}[1]{{\color{nonvitalcolor}{#1}}}
\definecolor{highlightcolor}{RGB}{55, 146, 55}
\newcommand {\highlight}[1]{{\color{highlightcolor}{#1}}}
\newcommand{\clipimg}{$\text{CLIP}_{img}$\xspace}
\newcommand{\cliptxt}{$\text{CLIP}_{txt}$\xspace}
\newcommand{\clipdir}{$\text{CLIP}_{dir}$\xspace}

\newcommand\todosilent[1]{}

\newlength{\ww}

\newcommand\blfootnote[1]{%
  \begingroup
  \renewcommand\thefootnote{}\footnote{#1}%
  \addtocounter{footnote}{-1}%
  \endgroup
}

\begin{document}

\twocolumn[{
    \renewcommand\twocolumn[1][]{#1}
    \maketitle
    \begin{center}
    \vspace{-15px}
    \includegraphics[width=0.99\linewidth]{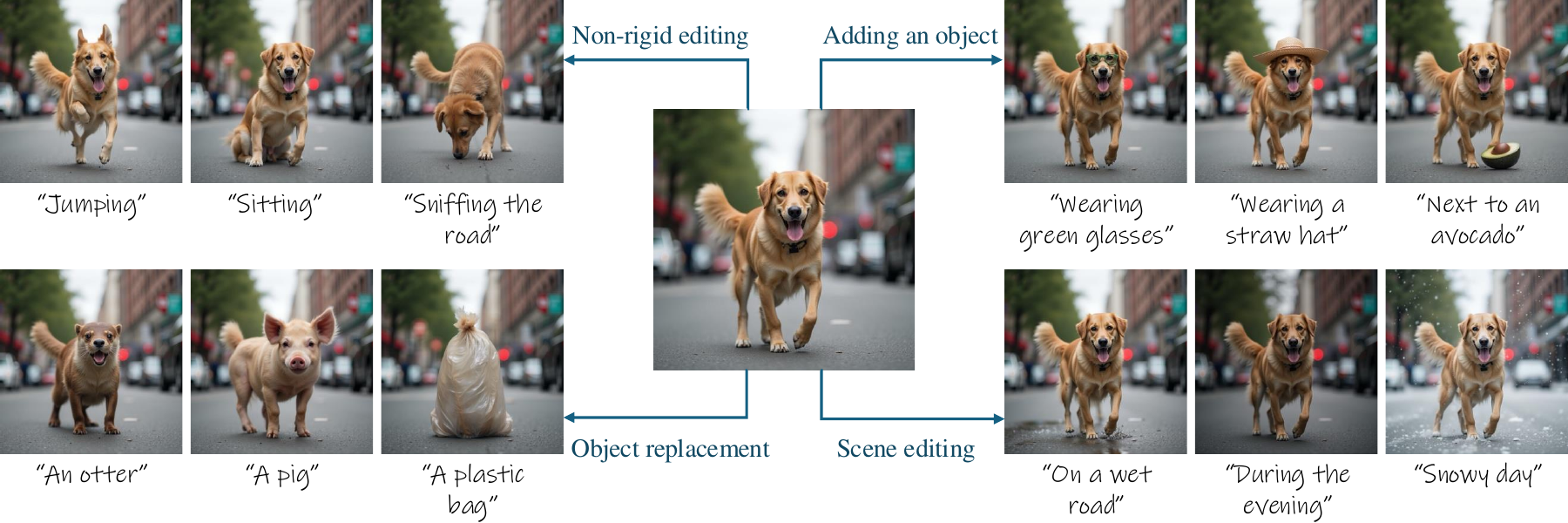}
    \captionsetup{type=figure}
    \caption{\textbf{Stable Flow.} Our training-free editing method is able to perform \emph{various} types of image editing operations, including non-rigid editing, object addition, object removal, and global scene editing. These different edits are done using the \emph{same} mechanism.}
    \label{fig:teaser}
\end{center}

}]

\begin{abstract}
    Diffusion models have revolutionized the field of content synthesis and editing. Recent models have replaced the traditional UNet architecture with the Diffusion Transformer (DiT), and employed flow-matching for improved training and sampling. However, they exhibit limited generation diversity. In this work, we leverage this limitation to perform consistent image edits via selective injection of attention features. The main challenge is that, unlike the UNet-based models, DiT lacks a coarse-to-fine synthesis structure, making it unclear in which layers to perform the injection. Therefore, we propose an automatic method to identify ``vital layers'' within DiT, crucial for image formation, and demonstrate how these layers facilitate a range of controlled \emph{stable} edits, from non-rigid modifications to object addition, using the \emph{same} mechanism. Next, to enable real-image editing, we introduce an improved image inversion method for flow models. Finally, we evaluate our approach through qualitative and quantitative comparisons, along with a user study, and demonstrate its effectiveness across multiple applications.   
\end{abstract}

\blfootnote{Project page is available at: \href{https://omriavrahami.com/stable-flow}{https://omriavrahami.com/stable-flow}
This research was performed while Omri was at Snap.}

\section{Introduction}
\label{sec:introduction}

Over the recent years, we have witnessed an unprecedented explosion in creative applications of generative models, fueled by diffusion-based models~\cite{sohl2015deep,song2019generative,ho2020denoising,song2020denoising}. Recent models, such as FLUX~\cite{flux} and SD3~\cite{Esser2024ScalingRF}, have replaced the traditional UNet architecture~\cite{Ronneberger2015UNetCN} with the Diffusion Transformer (DiT)~\cite{Peebles2022ScalableDM}, and adopted flow matching~\cite{Liu2022FlowSA, Lipman2022FlowMF, Albergo2022BuildingNF} as a superior alternative for training and sampling.

\begin{figure}[tp]
    \centering
    \setlength{\tabcolsep}{0.6pt}
    \renewcommand{\arraystretch}{0.7}
    \setlength{\ww}{0.235\columnwidth}
    \begin{tabular}{ccccc}
        \rotatebox[origin=c]{90}{\scriptsize{(1) SDXL~\cite{Podell2023SDXLIL}}} &
        {\includegraphics[valign=c, width=\ww]{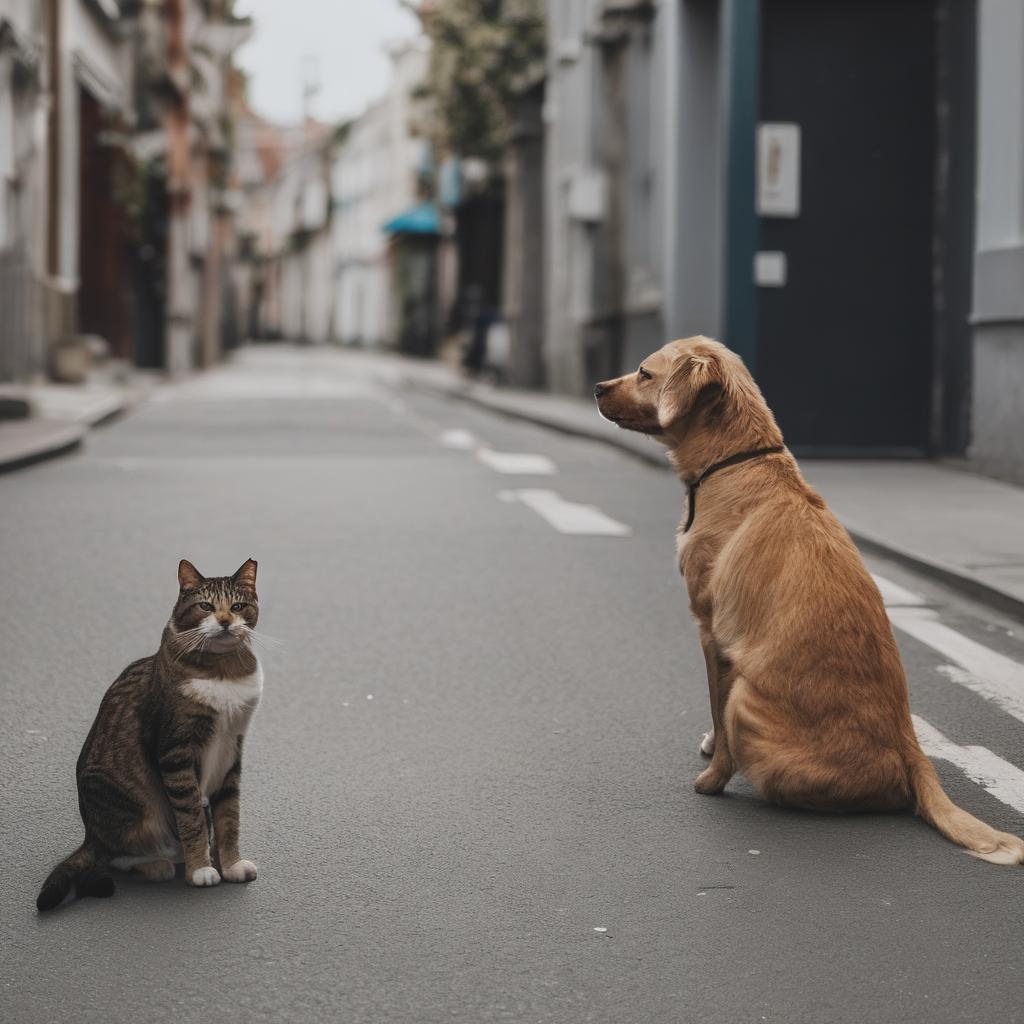}} &
        {\includegraphics[valign=c, width=\ww]{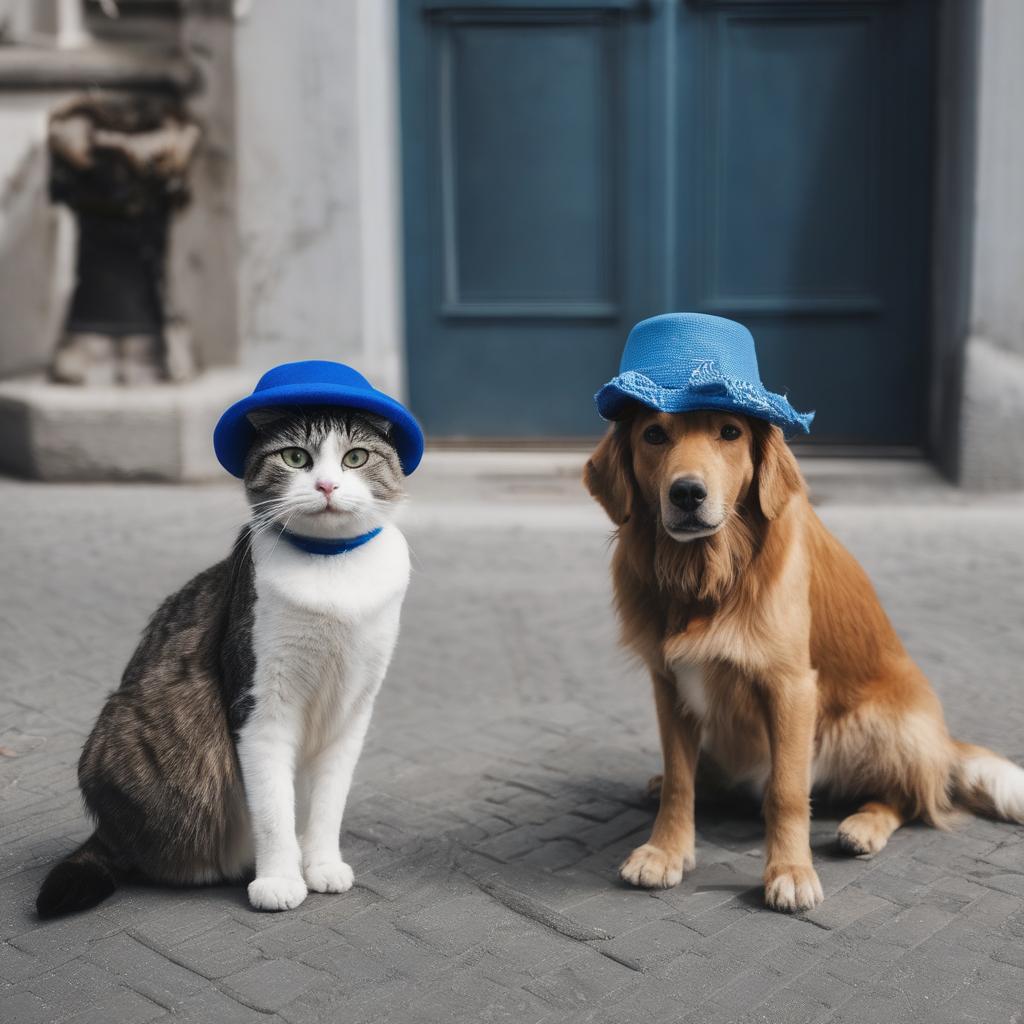}} &
        {\includegraphics[valign=c, width=\ww]{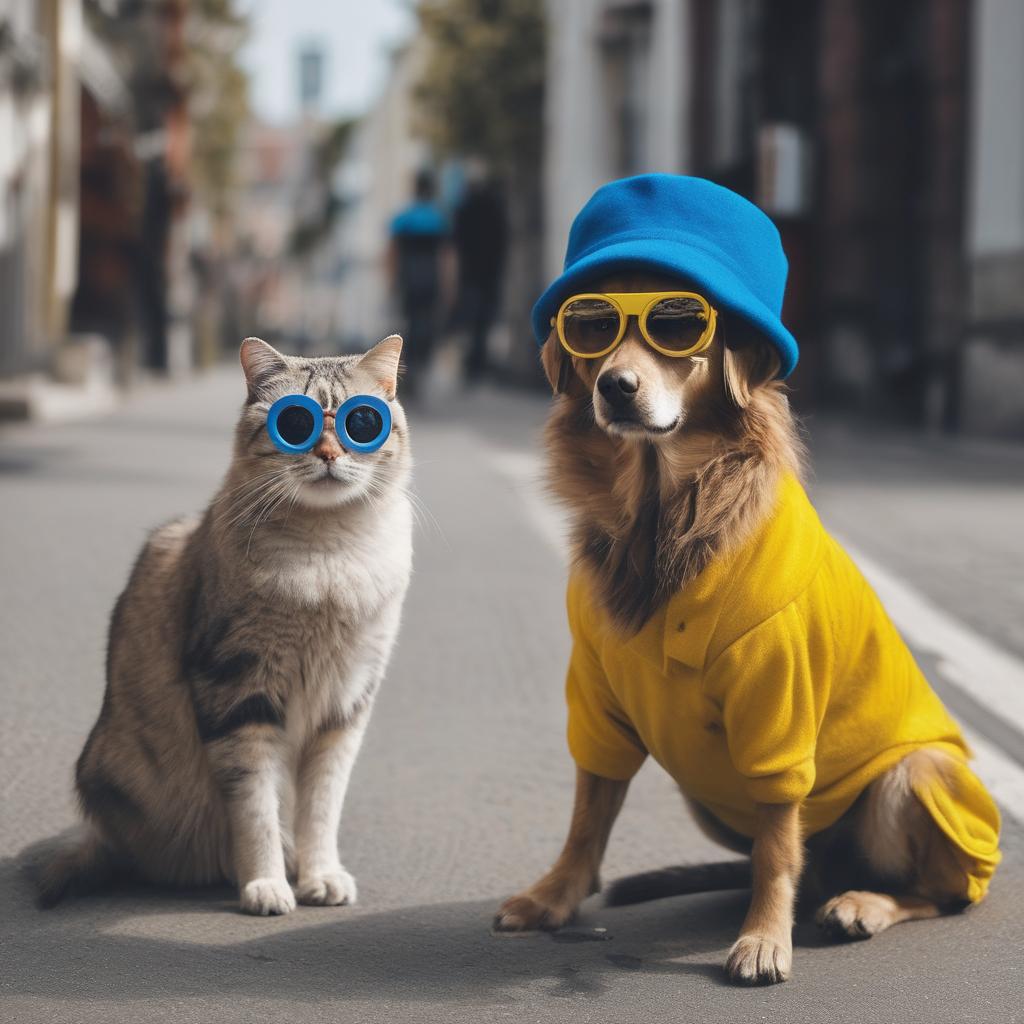}} &
        {\includegraphics[valign=c, width=\ww]{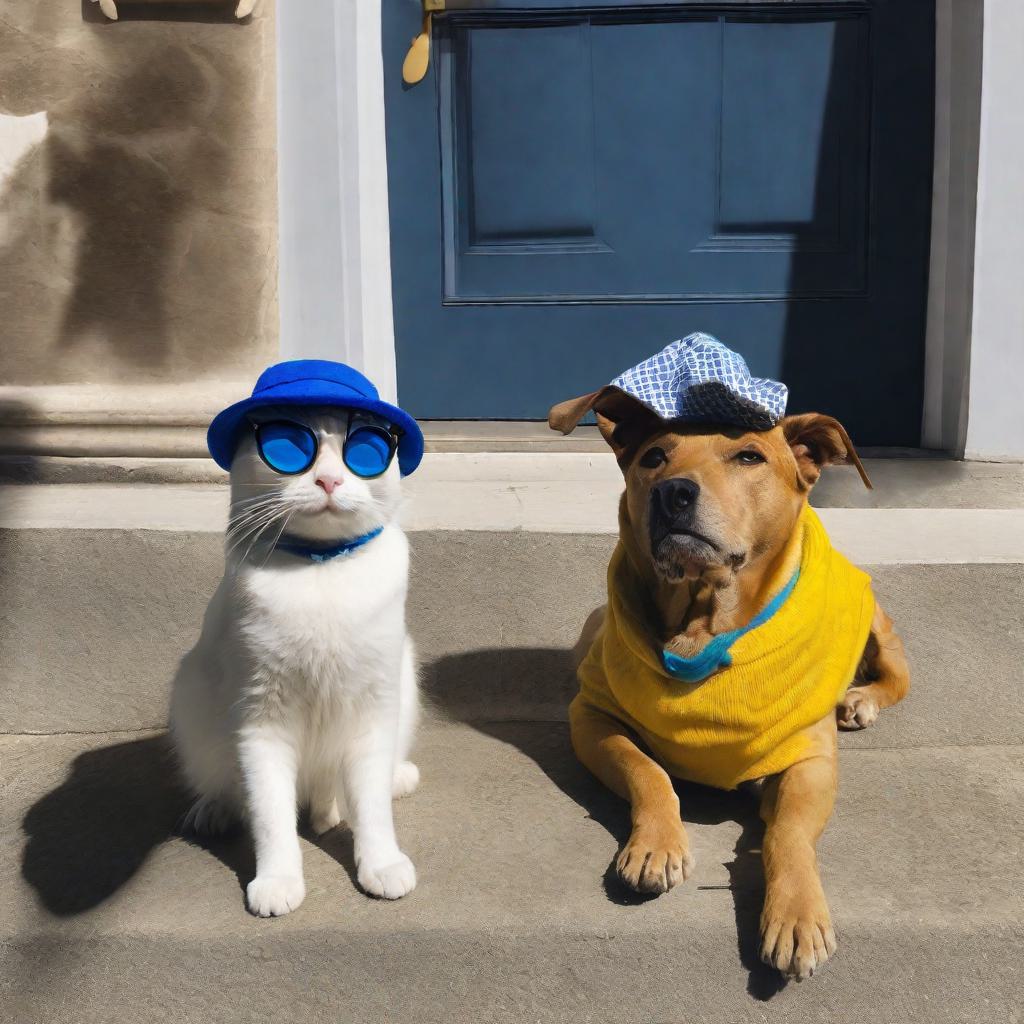}}
        \vspace{3px}
        \\

        \rotatebox[origin=c]{90}{\scriptsize{(2) FLUX~\cite{flux}}} &
        {\includegraphics[valign=c, width=\ww]{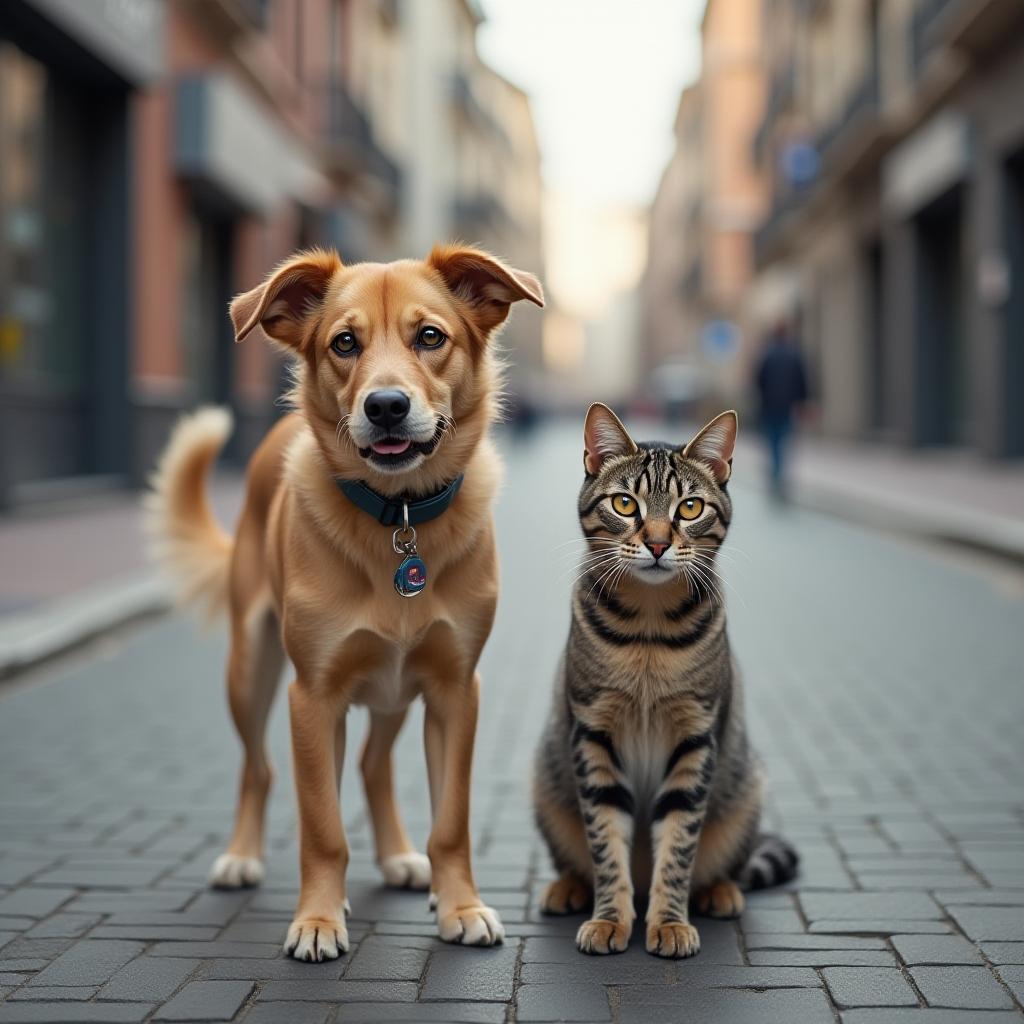}} &
        {\includegraphics[valign=c, width=\ww]{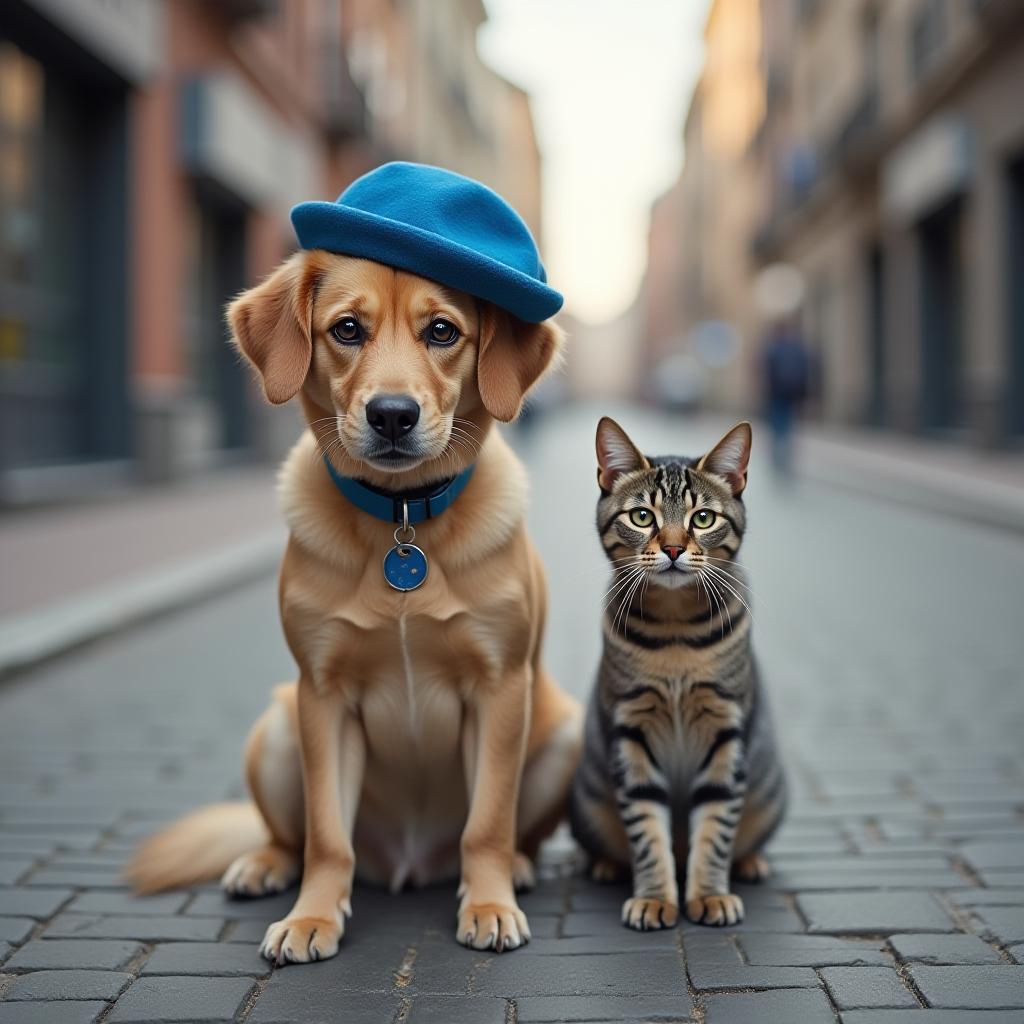}} &
        {\includegraphics[valign=c, width=\ww]{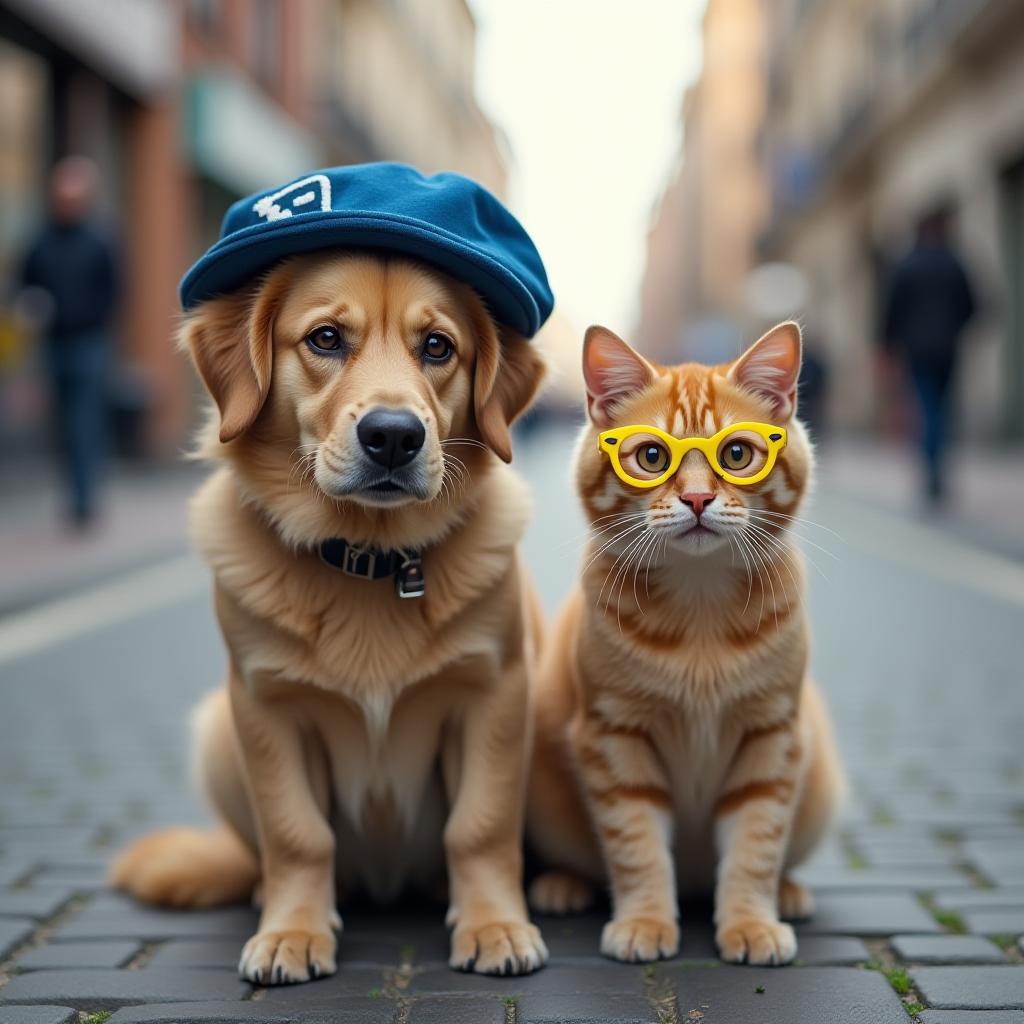}} &
        {\includegraphics[valign=c, width=\ww]{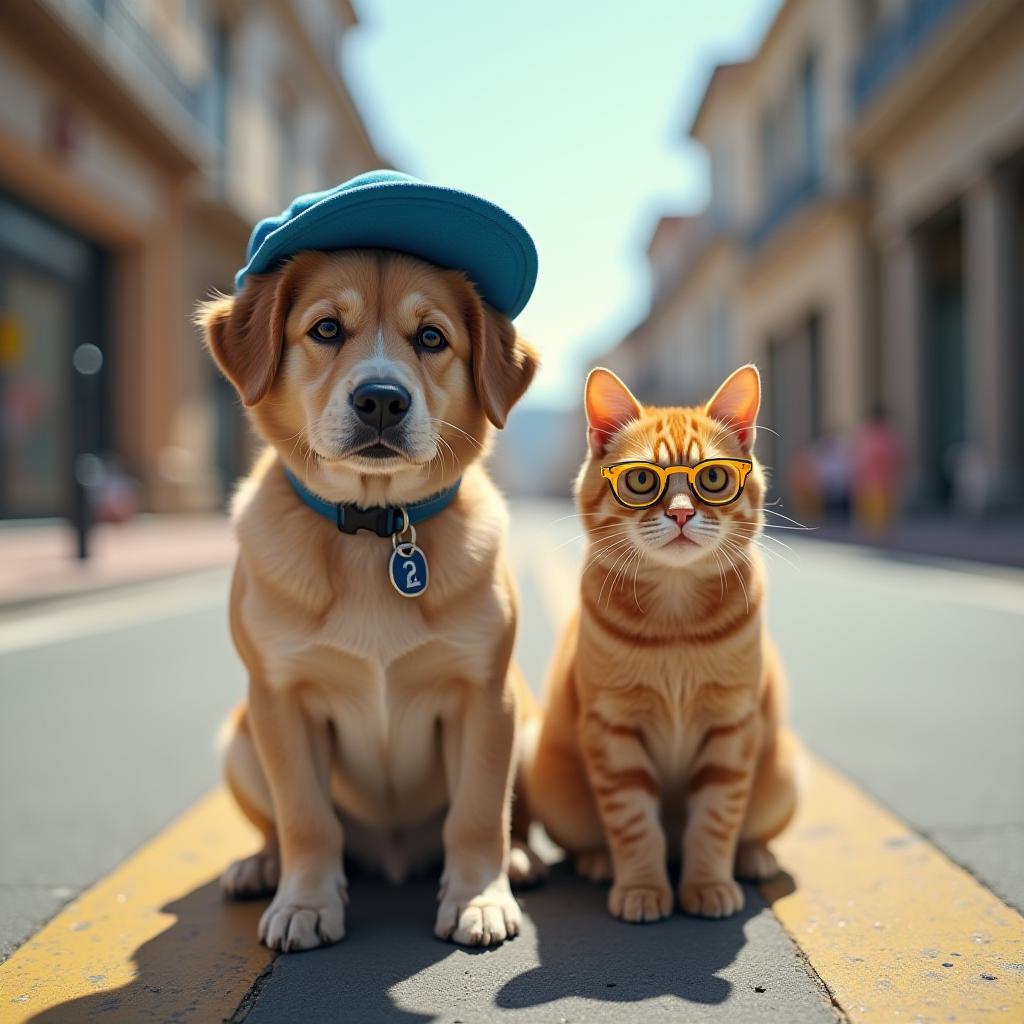}}
        \vspace{3px}
        \\

        \rotatebox[origin=c]{90}{\scriptsize{(3) Stable Flow}} &
        {\includegraphics[valign=c, width=\ww]{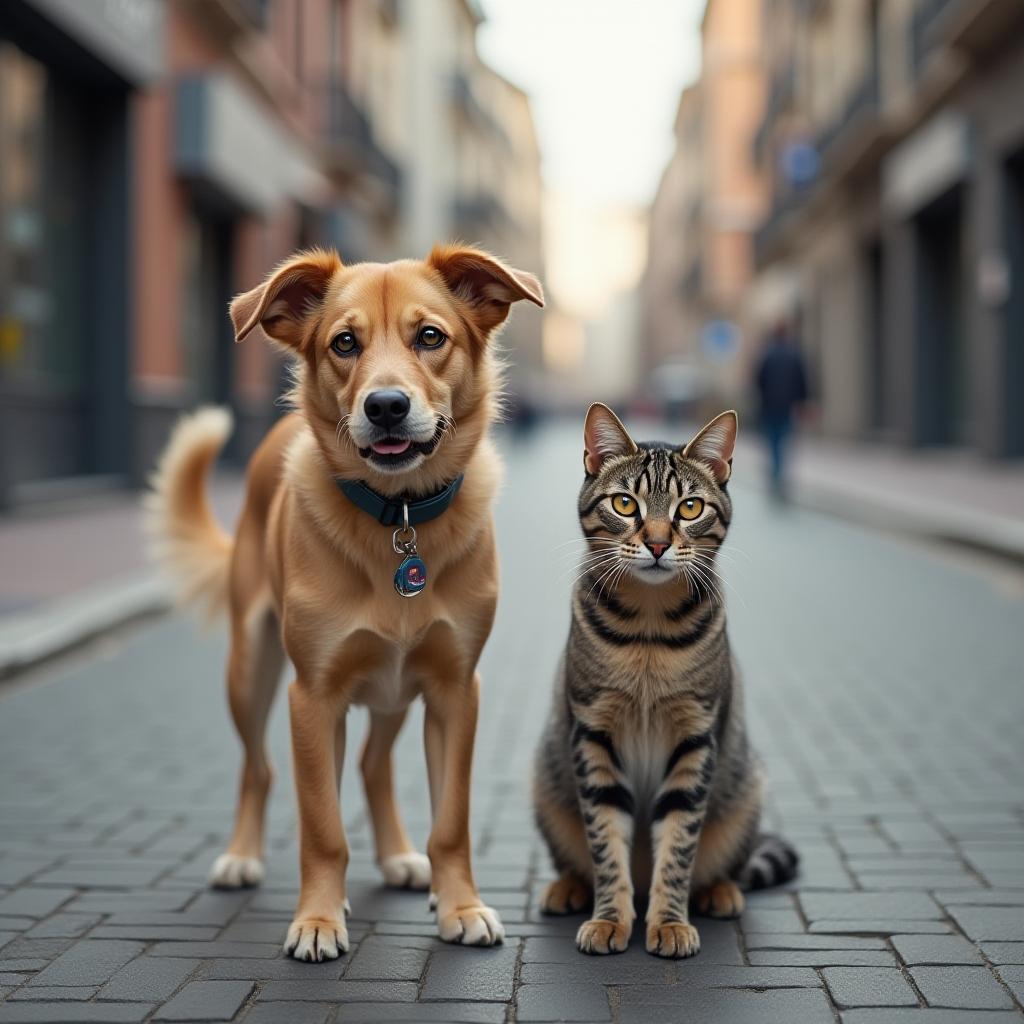}} &
        {\includegraphics[valign=c, width=\ww]{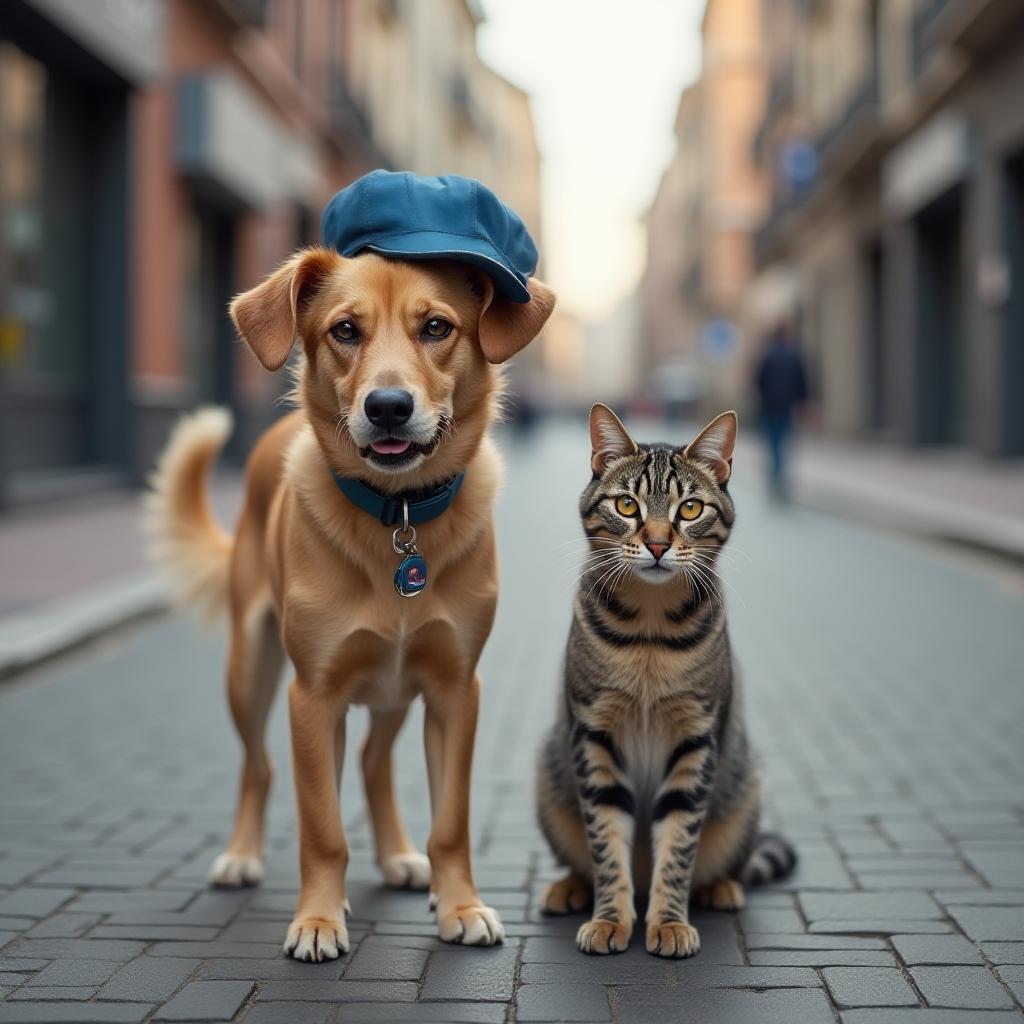}} &
        {\includegraphics[valign=c, width=\ww]{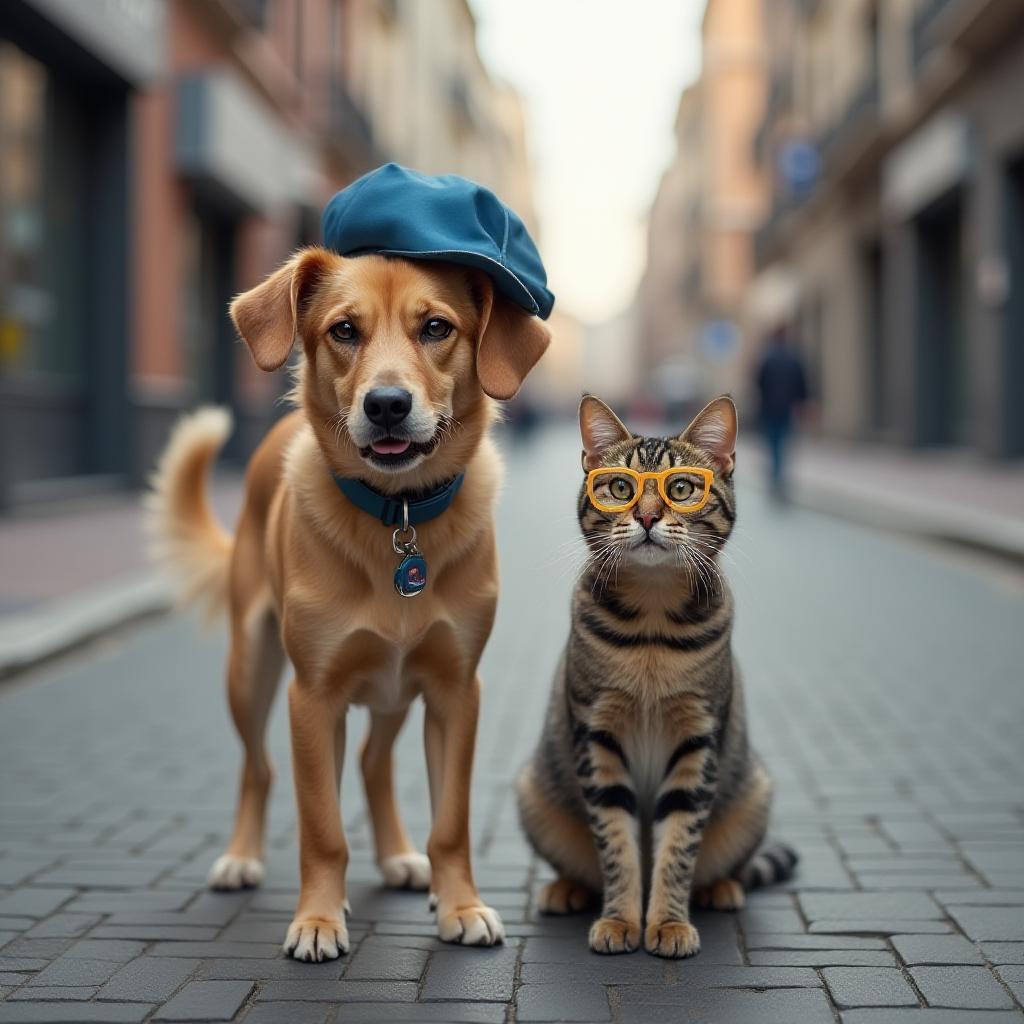}} &
        {\includegraphics[valign=c, width=\ww]{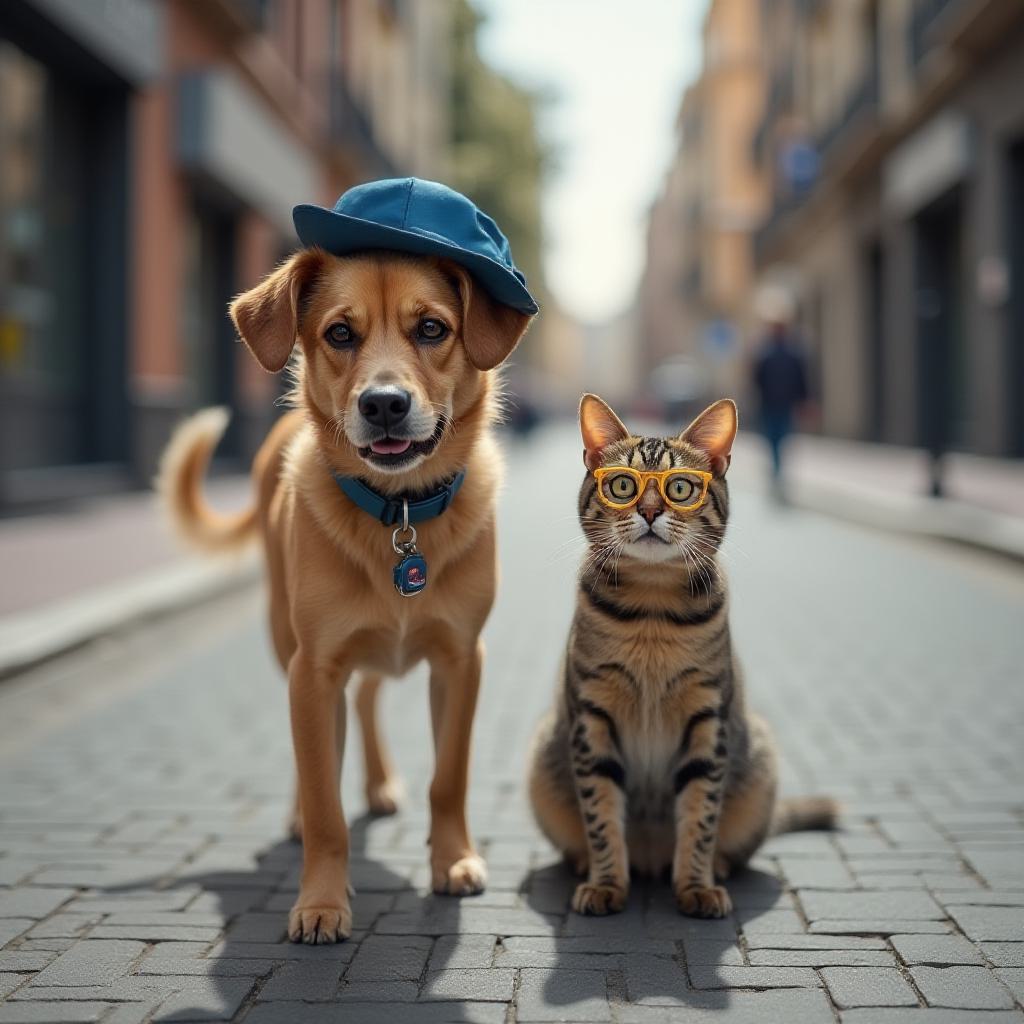}}
        \vspace{1px}
        \\

        &
        \scriptsize{\promptstart{A photo of a}} &
        \scriptsize{+ \promptstart{dog wearing}} &
        \scriptsize{+ \promptstart{cat wearing}} &
        \scriptsize{+ \promptstart{casting}}
        \\

        &
        \scriptsize{\promptend{dog and a cat...}} &
        \scriptsize{\promptend{a blue hat}} &
        \scriptsize{\promptend{yellow glasses}} &
        \scriptsize{\promptend{shadows}}
        \\

    \end{tabular}
    \caption{\textbf{Leveraging Reduced Diversity.} Using the same initial seed with different editing prompts, diffusion models such as (1) SDXL generate diverse results (different identities of the dog and the cat), while (2) FLUX generates a more stable (less diverse) set of results out-of-the-box. However, there are still some unintended differences (the dog is standing in the leftmost column and sitting in the others, the color of the cat is changing, and the road is different on the right). Using our approach, (3) Stable Flow, the edits are stable, maintaining consistency of the unrelated content.}
    \label{fig:motivation}
    \vspace{-10px}
\end{figure}

These flow-based models are based on optimal transport conditional probability paths, resulting in faster training and sampling, compared to diffusion models. This is attributed~\cite{Lipman2022FlowMF} to the fact that they follow straight line trajectories, rather than curved paths. One of the known consequences of this difference, however, is that these models exhibit lower diversity than previous diffusion models~\cite{Fischer2023BoostingLD}, as shown in \Cref{fig:motivation}(1-2). While reduced diversity is generally considered an undesirable characteristic, in this paper, we suggest leveraging it for the task of training-free image editing, as shown in \Cref{fig:motivation}(3) and \Cref{fig:teaser}.

Specifically, we explore image editing via parallel generation~\cite{Wu2022TuneAVideoOT, cao2023masactrl, Alaluf2023CrossImageAF}, where features from the generative trajectory of the source (reference) image are injected into the trajectory of the edited image. Such an approach has been shown effective in the context of convolutional UNet-based diffusion models~\cite{cao2023masactrl}, where the roles of the different attention layers are well understood.
However, such understanding has not yet emerged for DiT~\cite{Peebles2022ScalableDM}. Specifically, DiT does not exhibit the same fine-coarse-fine structure of the UNet~\cite{Peebles2022ScalableDM}, hence it is not clear which layers should be tampered with to achieve the desired editing behavior.

To address this gap, we analyze the importance of the different components in the DiT architecture, in order to determine the subset that should be injected while editing. More specifically, we introduce an \emph{automatic} method for detecting a set of \emph{vital layers} --- layers that are essential for the image formation --- by measuring the deviation in image content resulting from bypassing each layer. We show that there is no simple relationship between the vitality of a layer and its position in the architecture, \ie, the vital layers are spread across the transformer.

A close examination of the vital layers suggests that the injection of features into these layers strikes a good balance in the multimodal attention between the reference image content and the editing prompt. Consequently, limiting the injection of features to \emph{only} the vital layers tends to yield a stable edit, \ie, an edit that changes only the part(s) specified by the text prompt, while leaving the rest of the image intact, as demonstrated in \Cref{fig:motivation}(3). We demonstrate that performing \emph{the same} feature injection, enables performing a \emph{variety} of image edits, including non-rigid editing, addition of objects, and scene changes, as demonstrated in \Cref{fig:teaser}.

In order to support editing real images, it is typically necessary to invert them first \cite{song2020denoising, mokady2022null}. We employ an Inverse Euler Ordinary Differential Equation (ODE) solver to invert real images in the FLUX model~\cite{flux}. However, this method fails to reconstruct the input image in a satisfactory manner. To improve reconstruction accuracy, we introduce an out-of-distribution (OOD) nudging technique, in which we apply a small scalar perturbation to the clean latent before inverting it. We demonstrate that this makes FLUX less prone to undesired changes in the image during the forward pass.

Finally, we compare our method against its baselines qualitatively and quantitatively, and reaffirm the results with a user study. We also demonstrate several applications of our method.

In summary, our contributions are: (1) we propose an automatic method to detect the set of vital layers in DiT models and demonstrate how to use them for image editing; (2) we are the first method to harness the limited diversity of flow-based models to perform \emph{different} image editing tasks using the \emph{same} mechanism; and (3) we present an extension that allows editing real images using the FLUX model.

\begin{figure*}[t]
    \centering
    \includegraphics[width=0.95\linewidth]{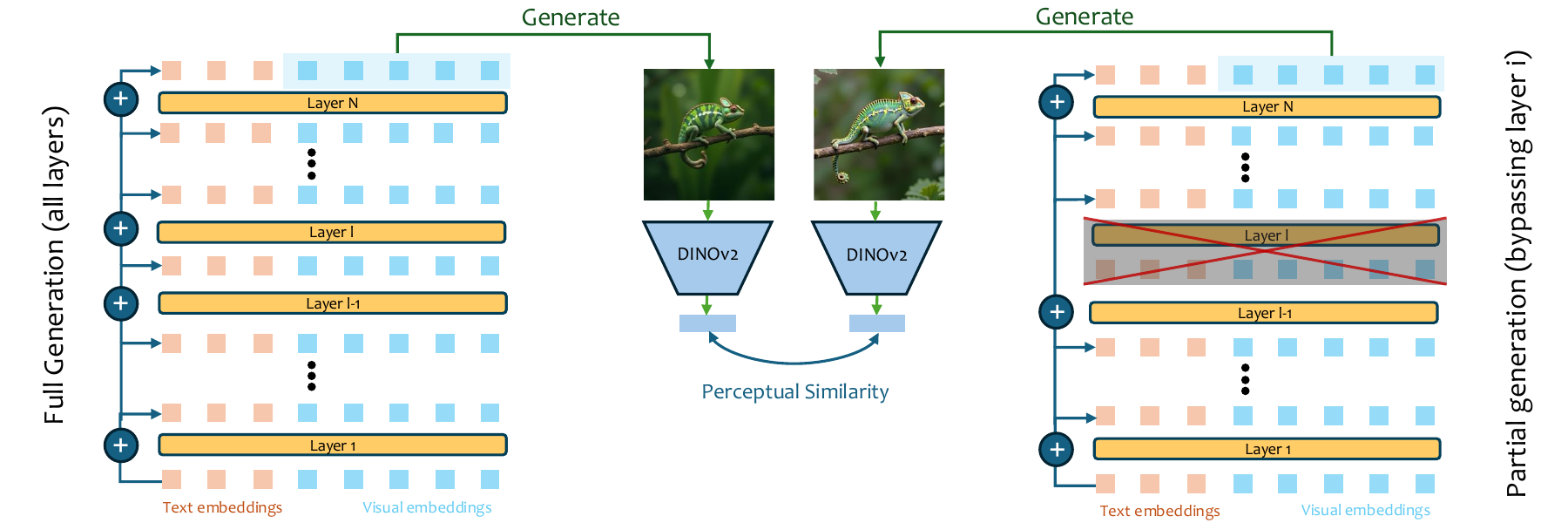}
    \caption{\textbf{Layer Removal.} (Left) Text-to-image DiT models consist of consecutive layers connected through residual connections~\cite{He2015DeepRL}. Each layer implements a multimodal diffusion transformer block~\cite{Esser2024ScalingRF} that processes a combined sequence of text and image embeddings. (Right) For each DiT layer, we perform an ablation by bypassing the layer using its residual connection. Then, we compare the generated result on the ablated model with the complete model using a perceptual similarity metric.}
    \label{fig:method}
    \vspace{-10px}
\end{figure*}

\section{Related Work}
\label{sec:related_work}

\paragraph{Text-Driven Image Editing.} After the emergence of text-to-image models, many works have suggested using them for various applications~\cite{arar2024Palp, Horwitz2024RecoveringTP, Chefer2024StillMovingCV, avrahami2024chosen, Salama2024DatasetSR}, including text-driven image editing tasks~\cite{Huang2024DiffusionMI, Po2023StateOT}. SDEdit~\cite{meng2021sdedit} addressed the image-to-image translation task adding noise and denoising with a new prompt. Blended Diffusion~\cite{blended_2022_CVPR, avrahami2023blendedlatent} suggested performing localized image editing~\cite{Huang2023RegionAwareDF, Li2023ZONEZI, Bau2021PaintBW, Nitzan2024LazyDT, Patashnik2023LocalizingOS} incorporating an input mask into the diffusion process in a training-free manner, while GLIDE~\cite{Nichol2021GLIDETP}, ImagenEditor~\cite{wang2022imagen}, and SmartBrush~\cite{Xie2022SmartBrushTA} offered fine-tuning the model on a given input mask. Other works suggested inferring the mask from the input image~\cite{couairon2022diffedit, Wang2023InstructEditIA} or via an additional click input~\cite{Regev2024Click2MaskLE}.
Other works~\cite{Hertz2022PrompttoPromptIE, pnpDiffusion2022, Parmar2023ZeroshotIT, cao2023masactrl} offered to inject information from the input image using parallel generation. While some methods~\cite{BarTal2022Text2LIVETL, Kawar2022ImagicTR, valevski2022unitune, Zhang2023ForgeditTG, zhang2024fast} suggested fine-tuning the model per-image, a recent line of work suggested training a designated model on a large synthetic dataset for instruction-based edits~\cite{brooks2022instructpix2pix, Zhang2023MagicBrush, Sheynin2023EmuEP}. However, none of the above methods is a training-free method that supports non-rigid editing, object adding/replacement and scene editing altogether. 

\paragraph{Image Inversion.} In the realm of generative models, \emph{inversion} \cite{Xia2021GANIA} is the task of finding a code within the latent space of a generator \cite{goodfellow2014generative,karras2019style, karras2020analyzing} that faithfully reconstructs a given image. Initial methods were developed for GAN models~\cite{abdal2019image2stylegan, abdal2020image2stylegan++, Zhu2020ImprovedSE, Richardson2020EncodingIS,Tov2021DesigningAE, Pidhorskyi2020AdversarialLA, zhu2020domain, alaluf2021hyperstyle, Roich2021PivotalTF, Bau2019SemanticPM, Dinh2021HyperInverterIS, Parmar2022SpatiallyAdaptiveMS, Yang2023ObjectawareIA}, and more recently for diffusion-based models~\cite{song2020denoising, mokady2022null, Wallace2022EDICTED, Pan2023EffectiveRI, huberman2023edit, Deutch2024TurboEditTI, Garibi2024ReNoiseRI, Han2023ProxEditIT, meiri2023fixed, Brack2023LEDITSLI}. In this work, we suggest inverting a real image in flow models using latent nudging, as we found that the standard inverse ODE solver is insufficient.

\section{Method}
\label{sec:method}

Our goal is to edit images based on text prompts while faithfully preserving the unedited regions of the source image. Given an input image $x$ and an editing prompt $p$, we aim to generate a modified image $\hat{x}$ that exhibits the desired changes specified by $p$ while maintaining the original content elsewhere. We leverage the limited diversity of the FLUX model and further constrain it to enable such \emph{stable} image edits through selective injection of attention features of the source image into the process that generates $\hat{x}$. Our approach is described in more detail below. In \Cref{sec:layers_importance}, we evaluate layer importance in the DiT model by analyzing the perceptual impact of layer removal (\Cref{fig:method}). Next, in \Cref{sec:image_editing}, we employ the most influential layers (termed \emph{vital} layers) for image editing through attention injection. Finally, in \Cref{sec:latent_nudging}, we extend our method to real image editing by inverting images into the latent space using the Euler inverse ODE solver, enhanced by \emph{latent nudging}.

\subsection{Measuring the Importance of DiT Layers}
\label{sec:layers_importance}

Recent text-to-image diffusion models~\cite{ho2020denoising, ramesh2022hierarchical, Rombach2021HighResolutionIS, Saharia2022PhotorealisticTD, Podell2023SDXLIL} predominantly use CNN-based UNets, which exhibit well-understood layer roles. In discriminative tasks, early layers detect simple features like edges, while deeper layers capture higher-level semantic concepts~\cite{Zeiler2013VisualizingAU, Simonyan2013DeepIC}. Similarly, in generative models, early-middle layers determine shape and color, while deeper layers control finer details~\cite{karras2019style}. This structure has been successfully exploited in text-driven editing~\cite{pnpDiffusion2022, cao2023masactrl, geyer2023tokenflow} through targeted manipulation of UNet decoder layers~\cite{Ronneberger2015UNetCN}.

In contrast, state-of-the-art text-to-image DiT~\cite{Peebles2022ScalableDM} models (FLUX \cite{flux} and SD3~\cite{Esser2024ScalingRF}) employ a fundamentally different architecture, as shown in \Cref{fig:method}(left). These models consist of consecutive layers connected through residual connections~\cite{He2015DeepRL}, without convolutions. Each layer implements a multimodal diffusion transformer block~\cite{Esser2024ScalingRF} (MM-DiT-Block) that processes a combined sequence of text and image embeddings. Unlike in UNets, the roles of the different layers are not yet intuitively clear, making it challenging to determine which layers are best suited for image editing.

To quantify layer importance in the FLUX model, we devised a systematic evaluation approach. Using ChatGPT~\cite{chatgpt}, we automatically generated a set $P$ of $k=64$ diverse text prompts, and draw a set $S$ of random seeds. Each of these prompts was used to generate a reference image, yielding in a set $G_{\textit{ref}}$. For each DiT layer $\ell \in \mathbb{L}$, we performed an ablation by bypassing the layer using its residual connection, as illustrated in \Cref{fig:method}(right). This process generated a set of images $G_{\ell}$ from the same prompts and seeds. See the supplementary material for more details.

To assess the impact of each layer, we measured the perceptual similarity between $G_{\textit{ref}}$ and $G_{\ell}$ and using DINOv2~\cite{Oquab2023DINOv2LR} (see \Cref{fig:method}). The results, plotted in \Cref{fig:layer_removal_quantiative}, show that removing certain layers significantly affects the generated images, while others have minimal impact. Importantly, influential layers are distributed across the transformer rather than concentrated in specific regions. We formally define the \emph{vitality} of layer $\ell$ as:
\begin{equation}
    vitality(\ell) = 1 - \frac{1}{k} \sum_{s \in S, p \in P} d(M_{\text{full}}(s, p), M_{\text{-}\ell}(s, p)),
    \label{eqn:vitality}
\end{equation}
where $M_{\text{full}}$ represents the complete model, $M_{\text{-}\ell}$ denotes the model with layer $\ell$ omitted, and $d(\cdot,\cdot)$ is the perceptual similarity metric. The set of vital layers $V$ is then defined as:
\vspace{-5px}
\begin{equation}
    V = \left\{ \ell \in \mathbb{L} \mid vitality(\ell) \ge \tau_{vit} \right\},
    \label{eqn:vital_set}
\end{equation}
where $\tau_{vit}$ is the vitality threshold.

\begin{figure}[t]
    \centering
    \setlength{\tabcolsep}{0.6pt}
    \renewcommand{\arraystretch}{0.8}
    \setlength{\ww}{0.213\columnwidth}
    \begin{tabular}{ccccc}
        \rotatebox[origin=c]{90}{\footnotesize{$G_{\textit{ref}}$}} &
        {\includegraphics[valign=c, width=\ww]{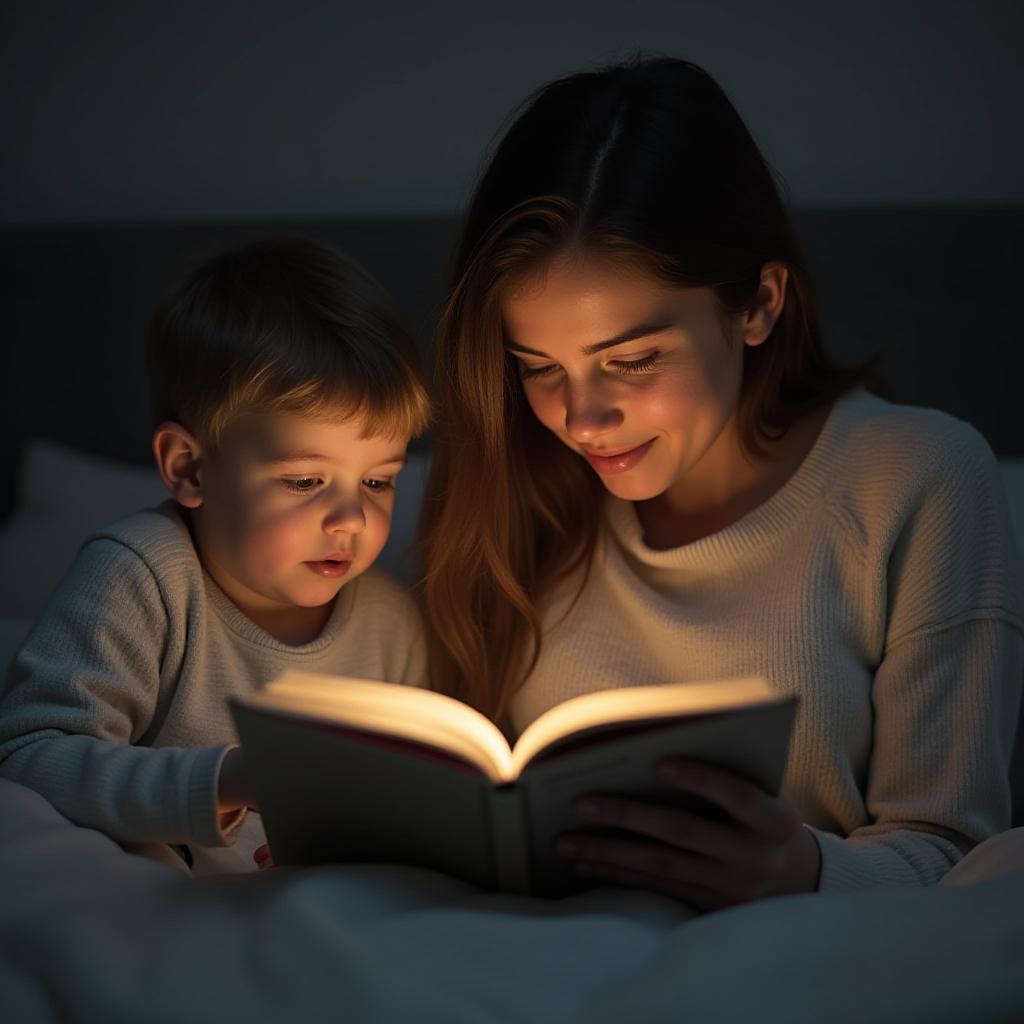}} &
        {\includegraphics[valign=c, width=\ww]{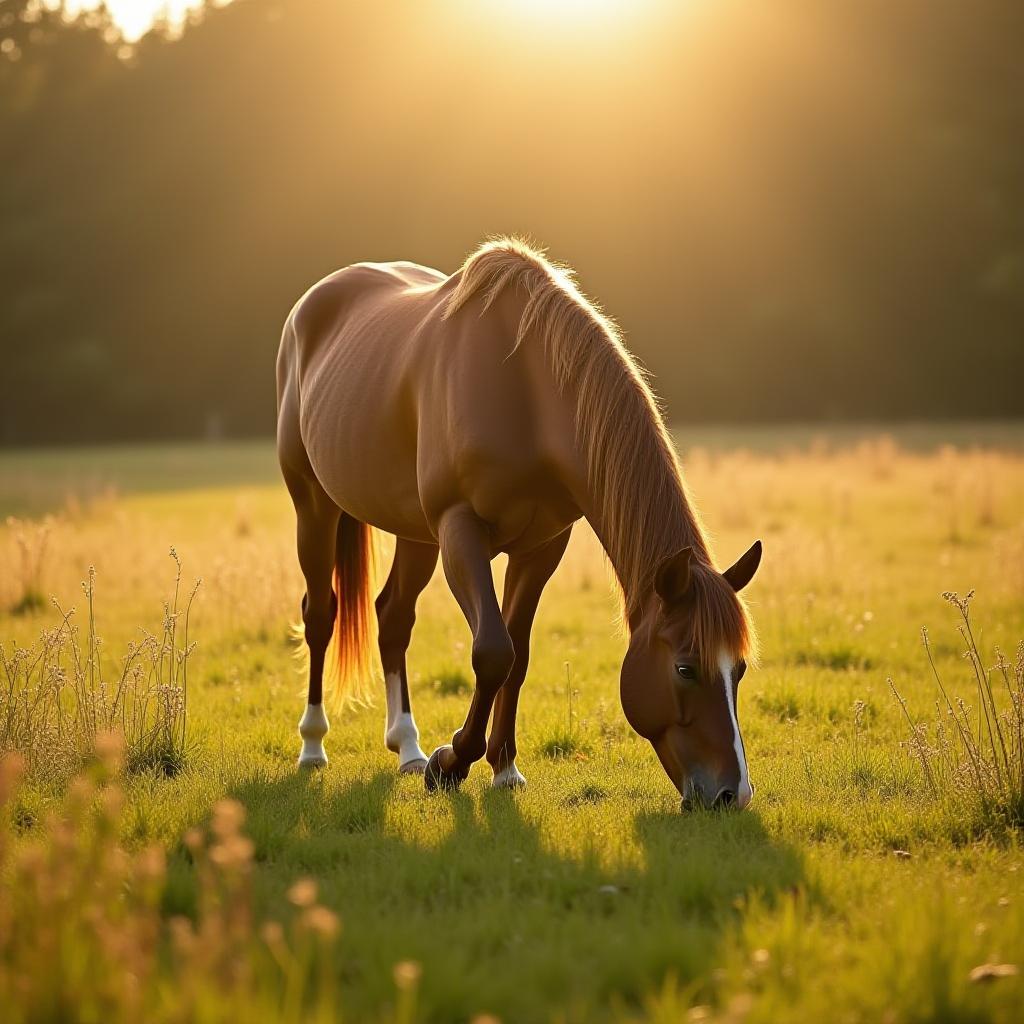}} &
        {\includegraphics[valign=c, width=\ww]{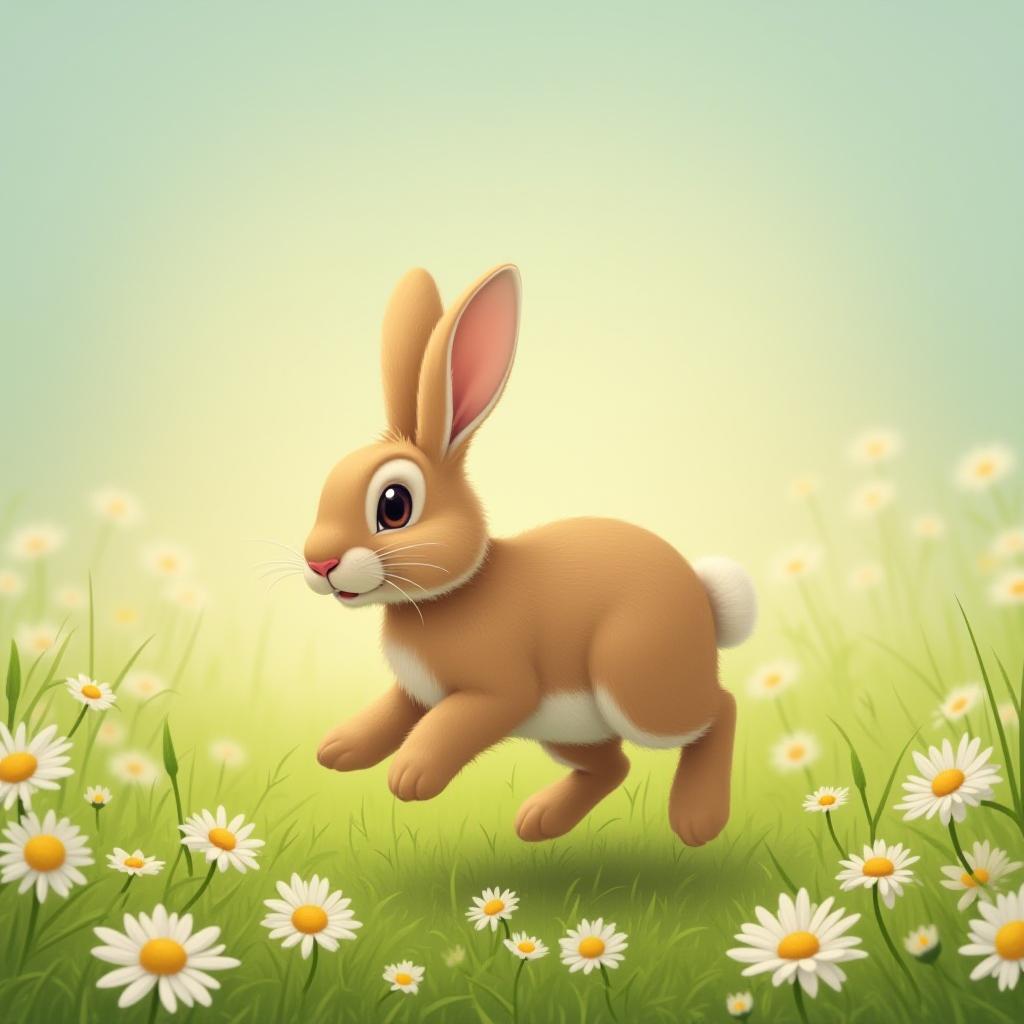}} &
        {\includegraphics[valign=c, width=\ww]{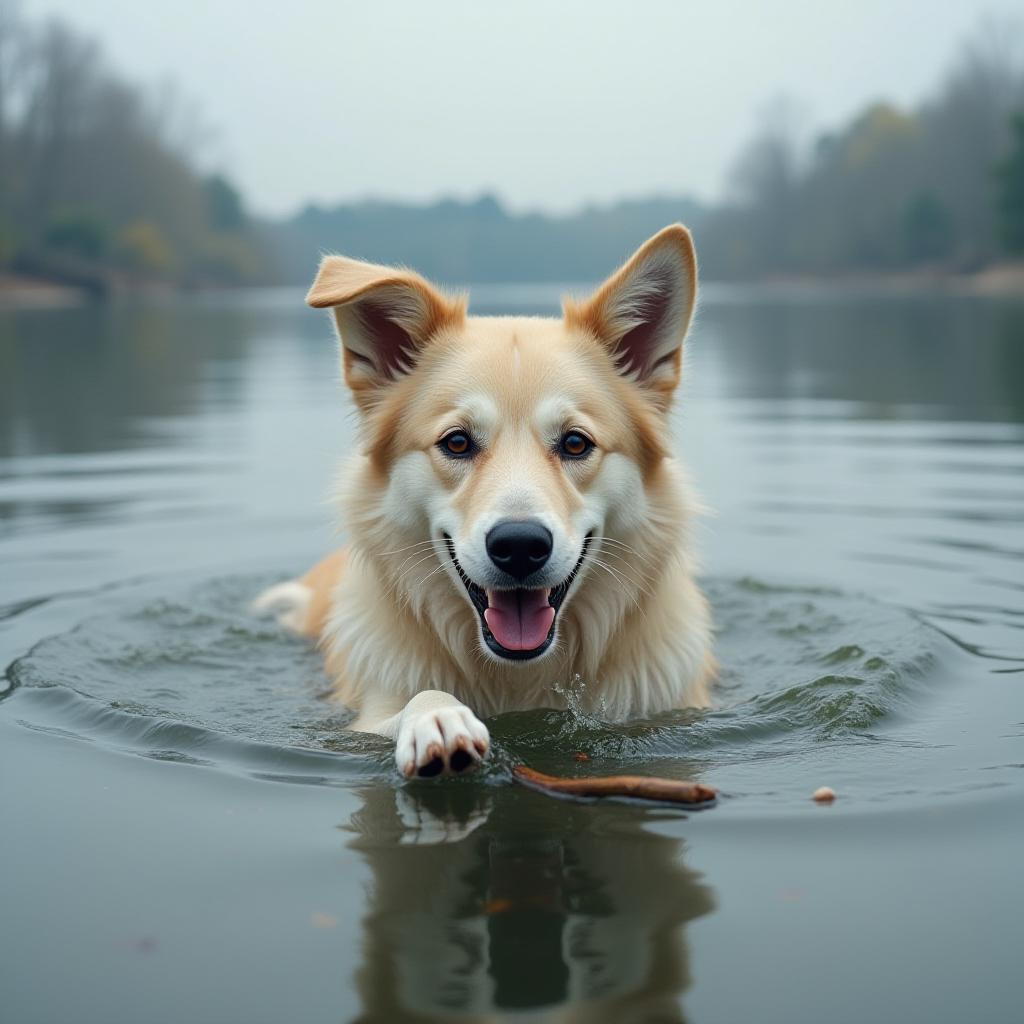}}
        \vspace{5px}
        \\

        \rotatebox[origin=c]{90}{\vital{\footnotesize{$G_{0}$}}} &
        {\includegraphics[valign=c, width=\ww]{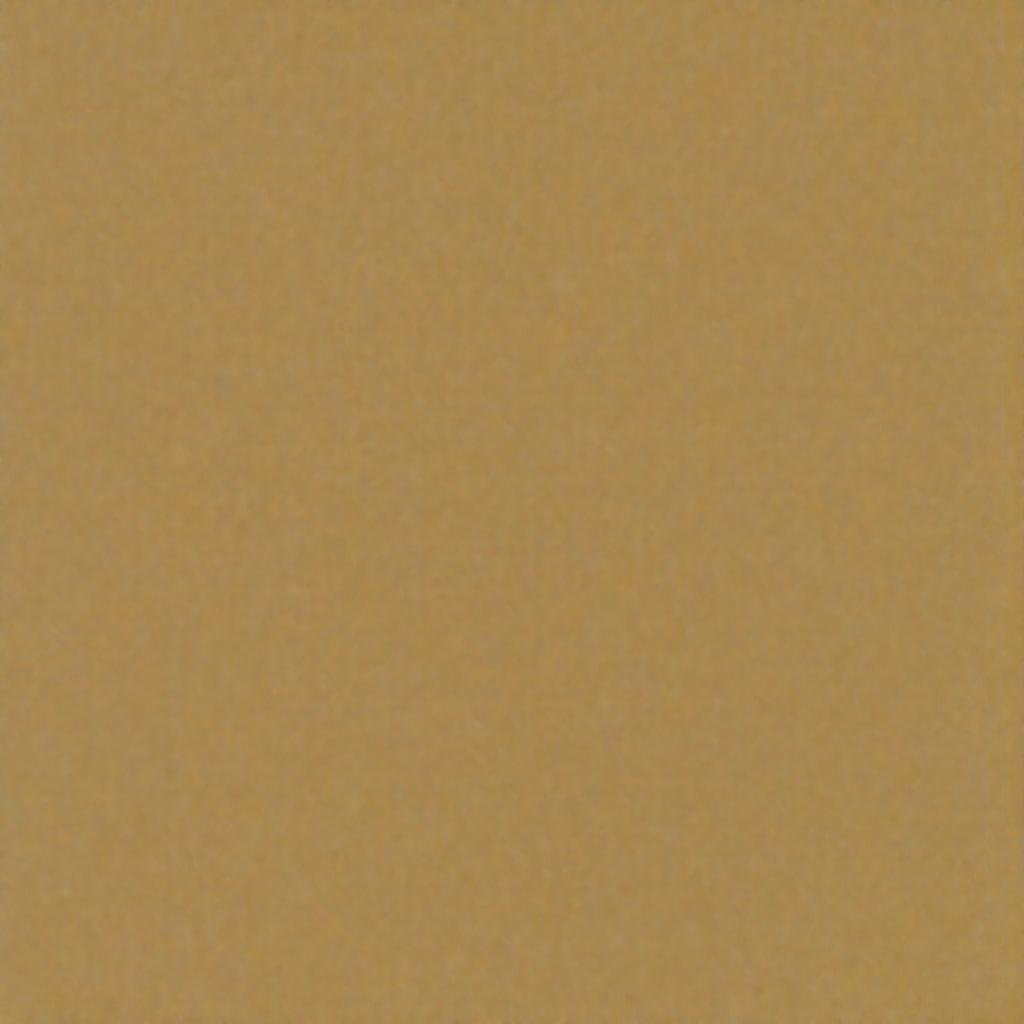}} &
        {\includegraphics[valign=c, width=\ww]{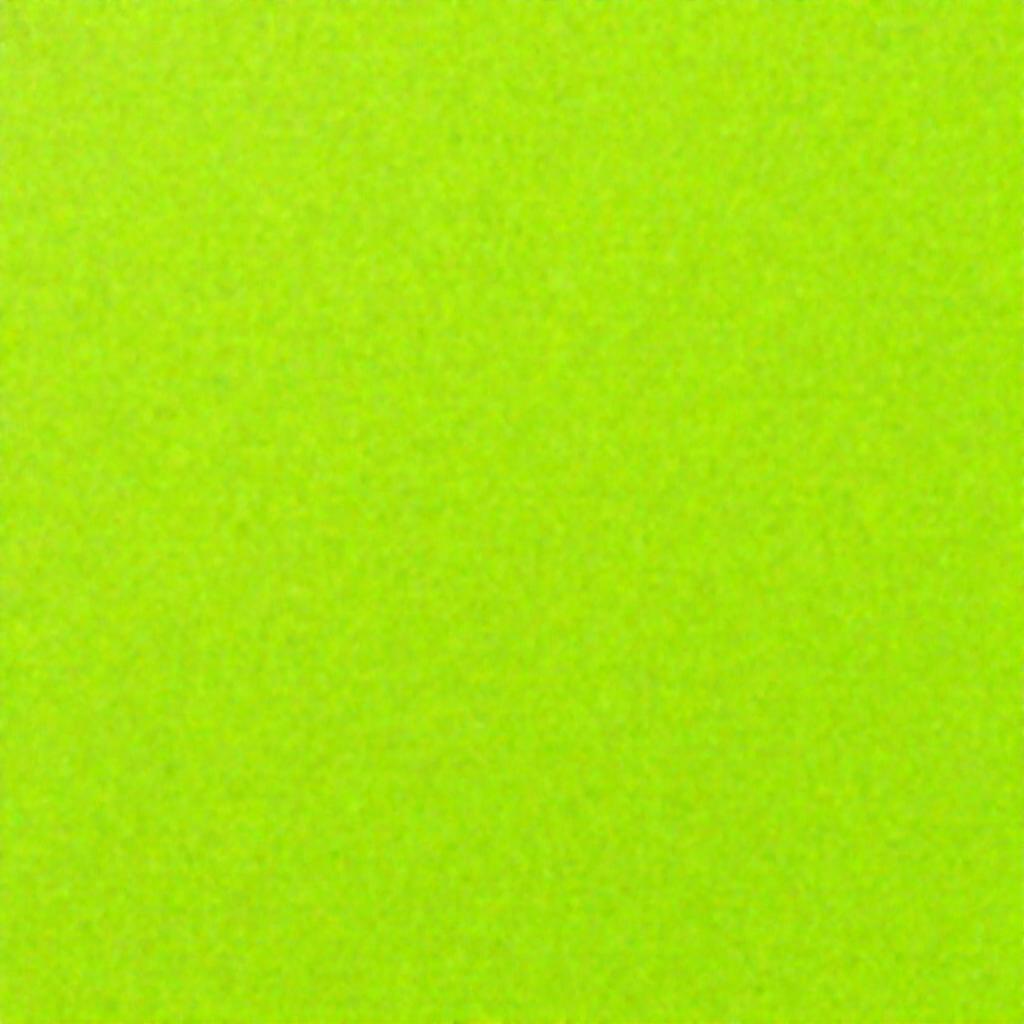}} &
        {\includegraphics[valign=c, width=\ww]{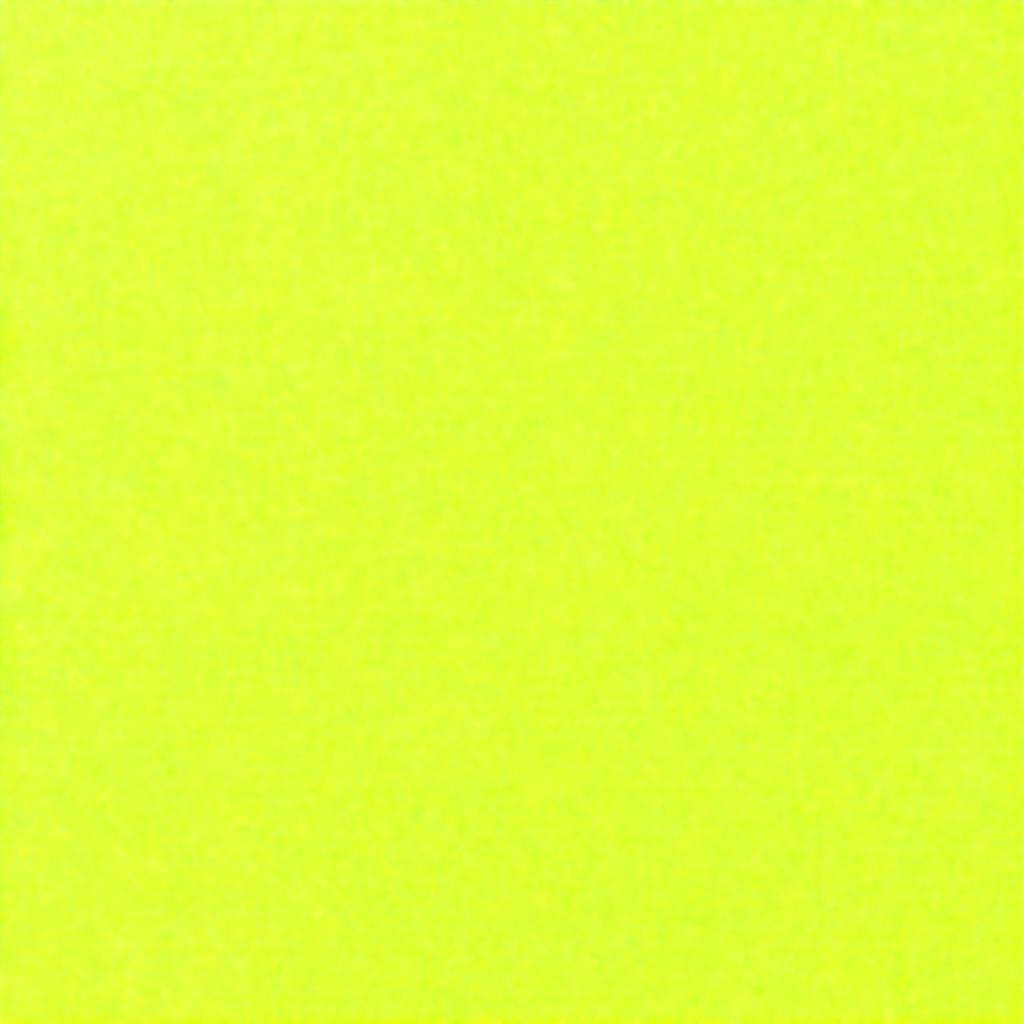}} &
        {\includegraphics[valign=c, width=\ww]{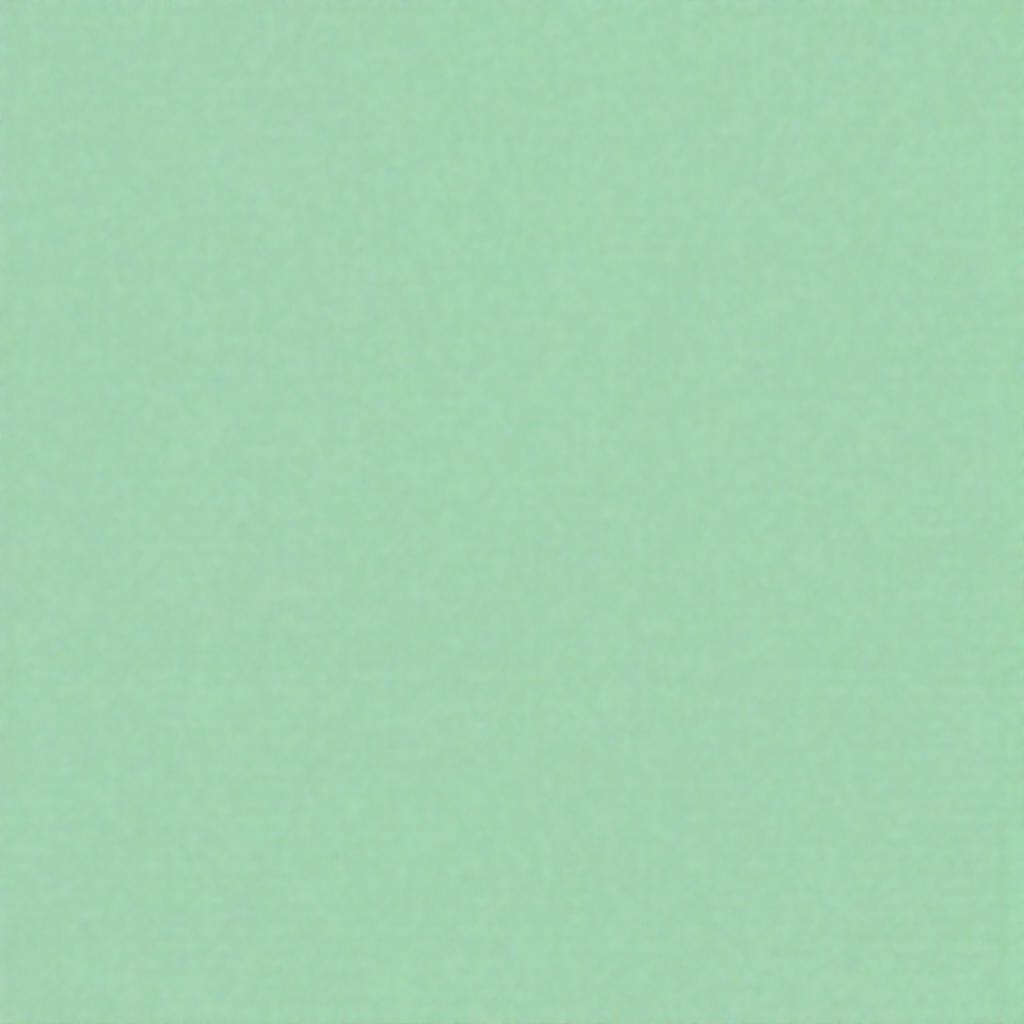}}
        \vspace{1px}
        \\

        \rotatebox[origin=c]{90}{\nonvital{\footnotesize{$G_{5}$}}} &
        {\includegraphics[valign=c, width=\ww]{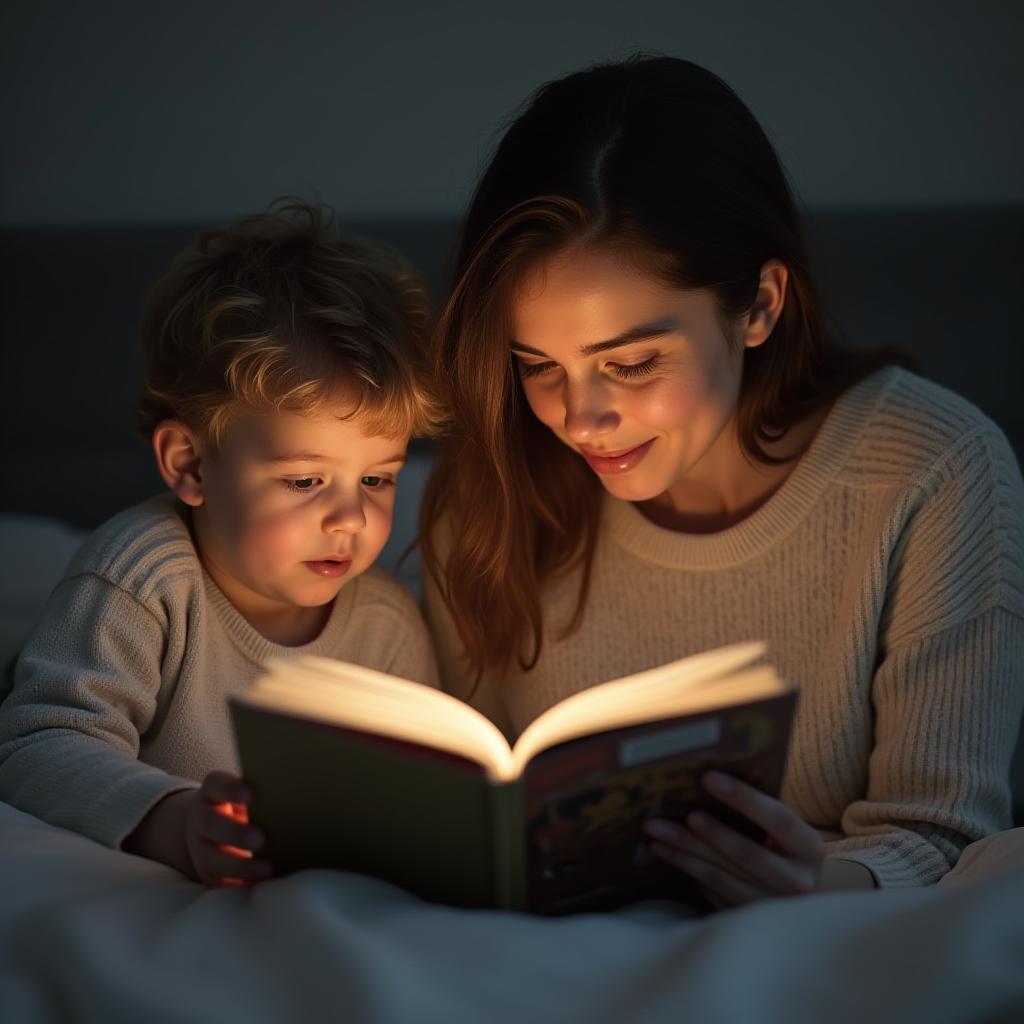}} &
        {\includegraphics[valign=c, width=\ww]{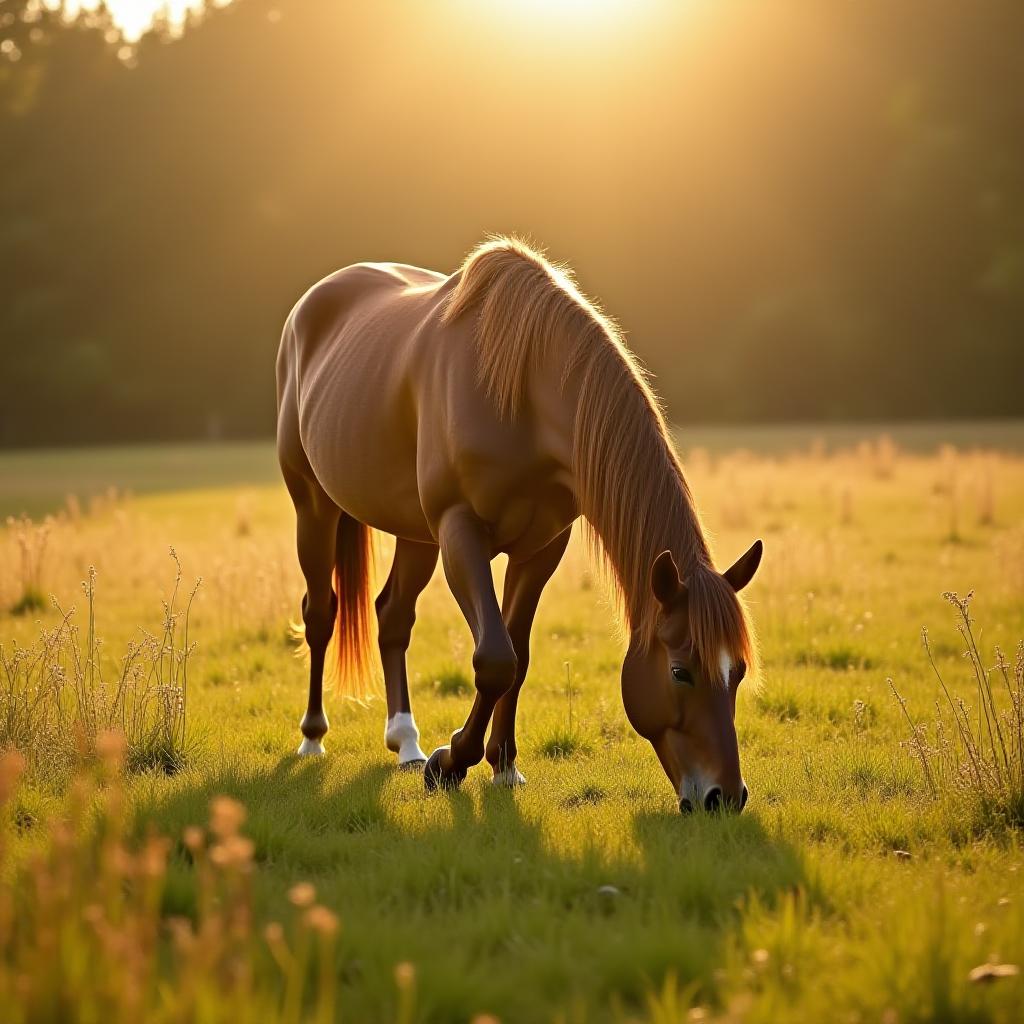}} &
        {\includegraphics[valign=c, width=\ww]{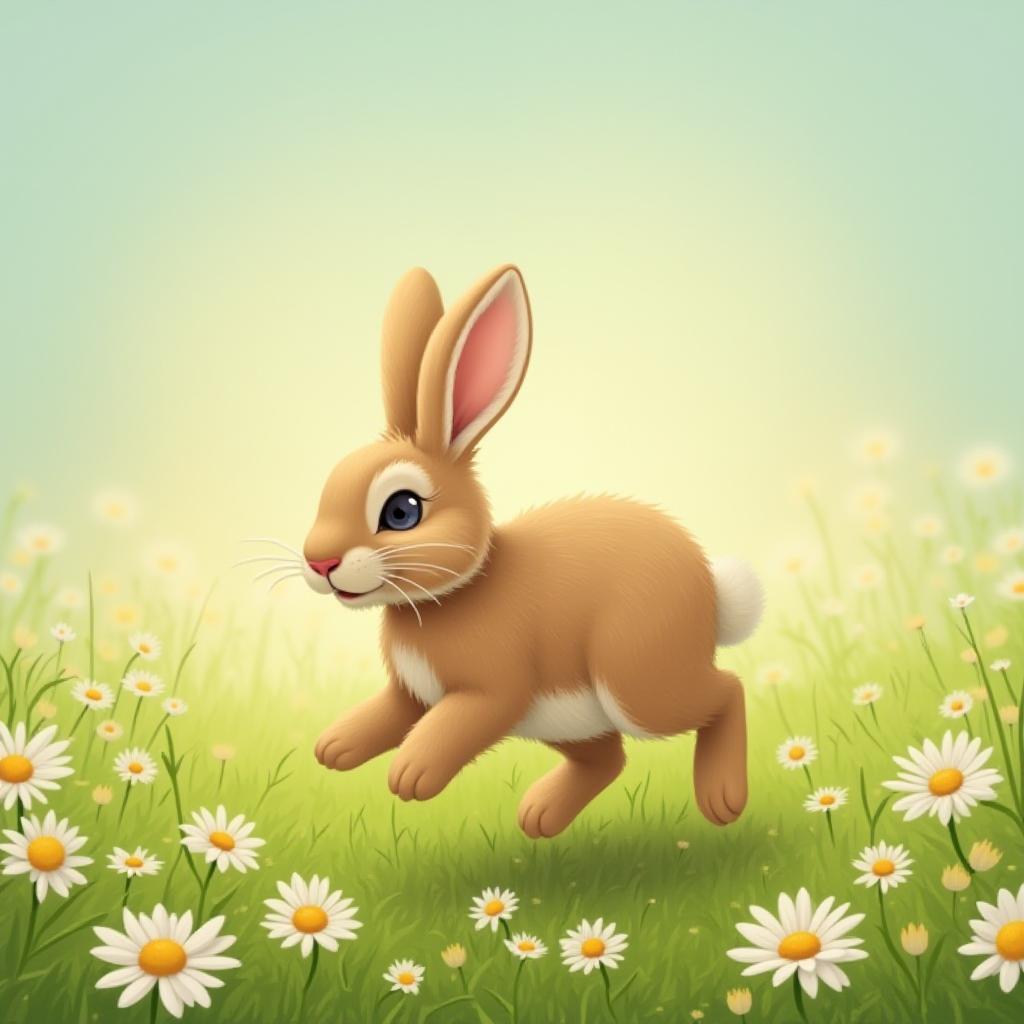}} &
        {\includegraphics[valign=c, width=\ww]{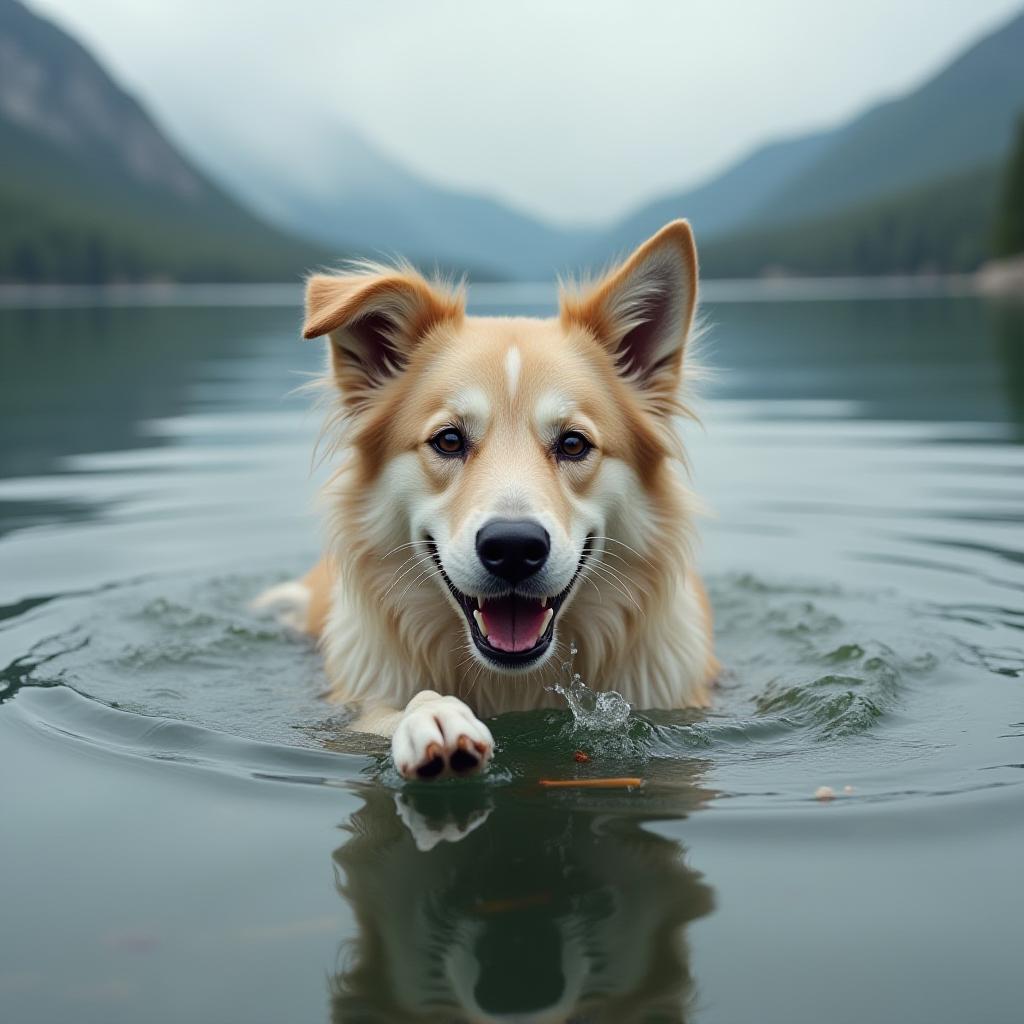}}
        \vspace{1px}
        \\

        \rotatebox[origin=c]{90}{\vital{\footnotesize{$G_{18}$}}} &
        {\includegraphics[valign=c, width=\ww]{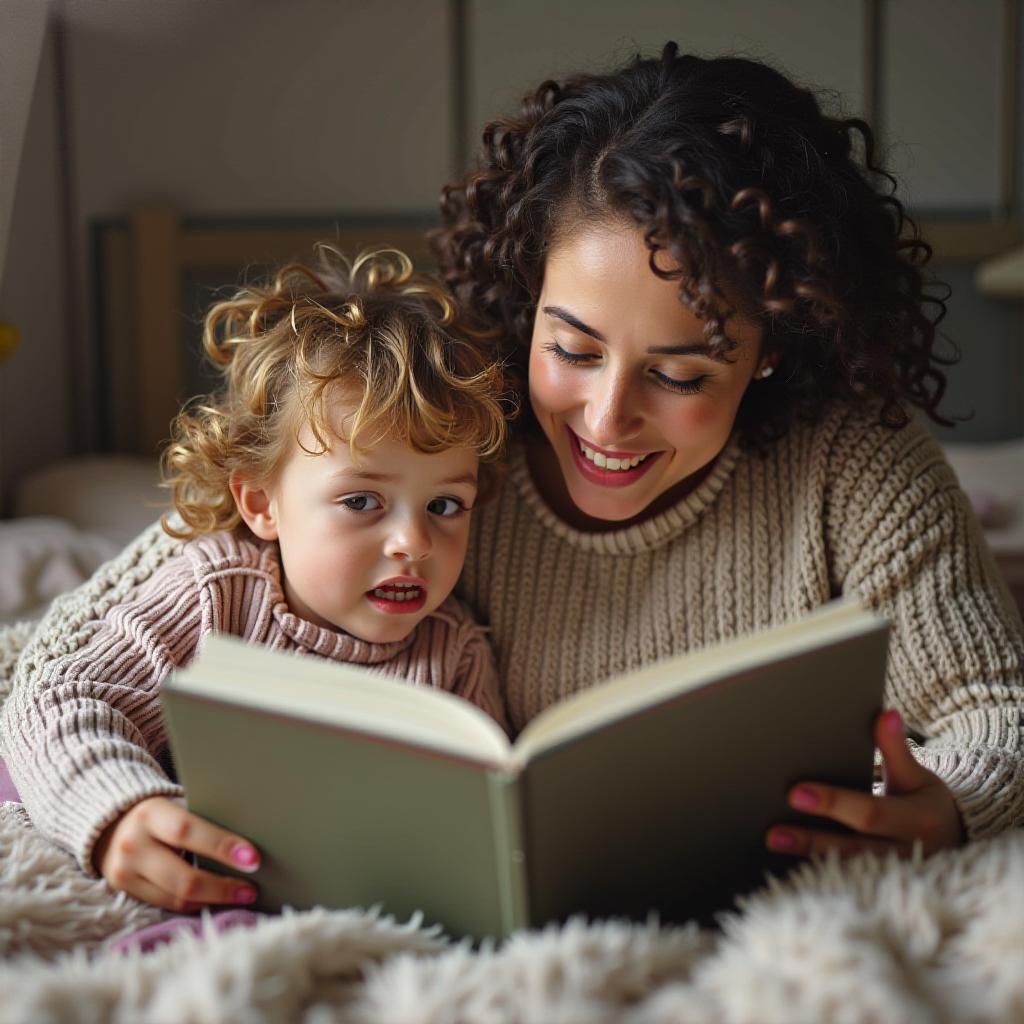}} &
        {\includegraphics[valign=c, width=\ww]{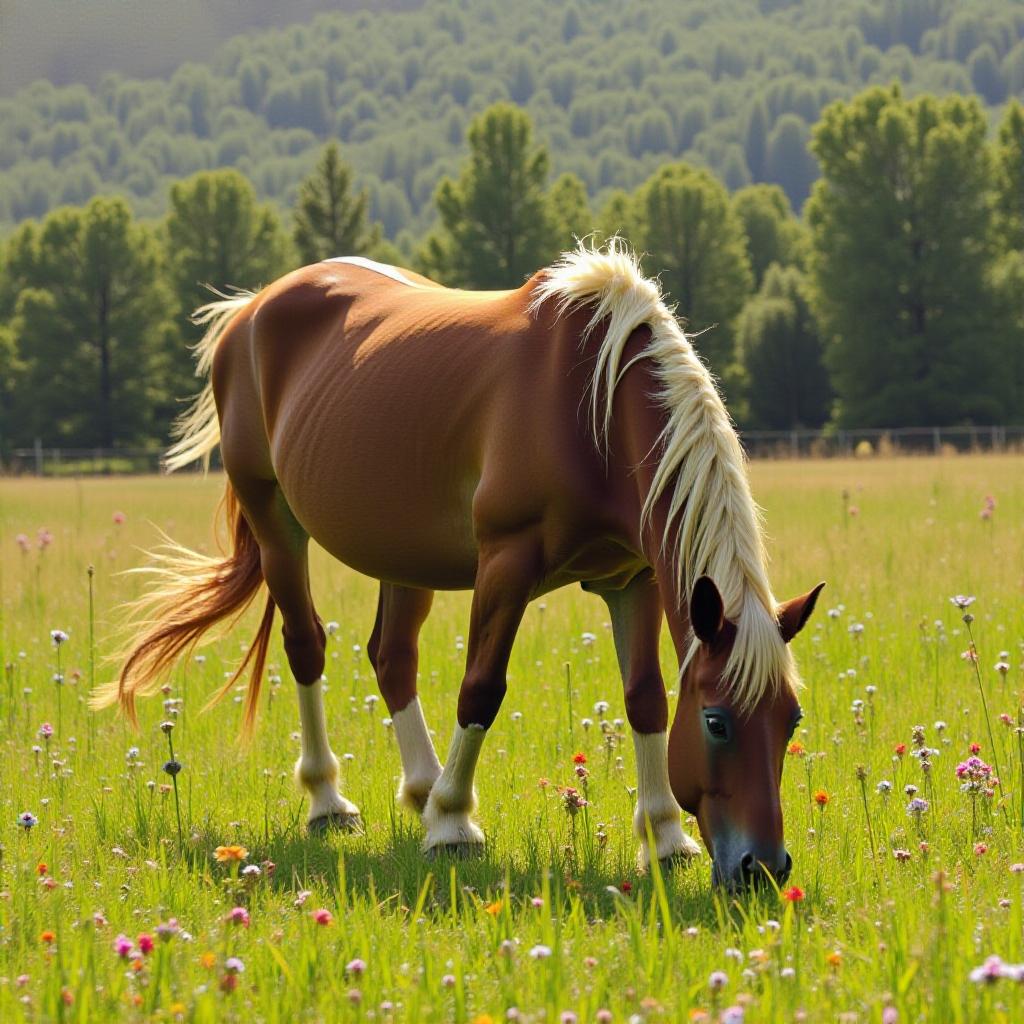}} &
        {\includegraphics[valign=c, width=\ww]{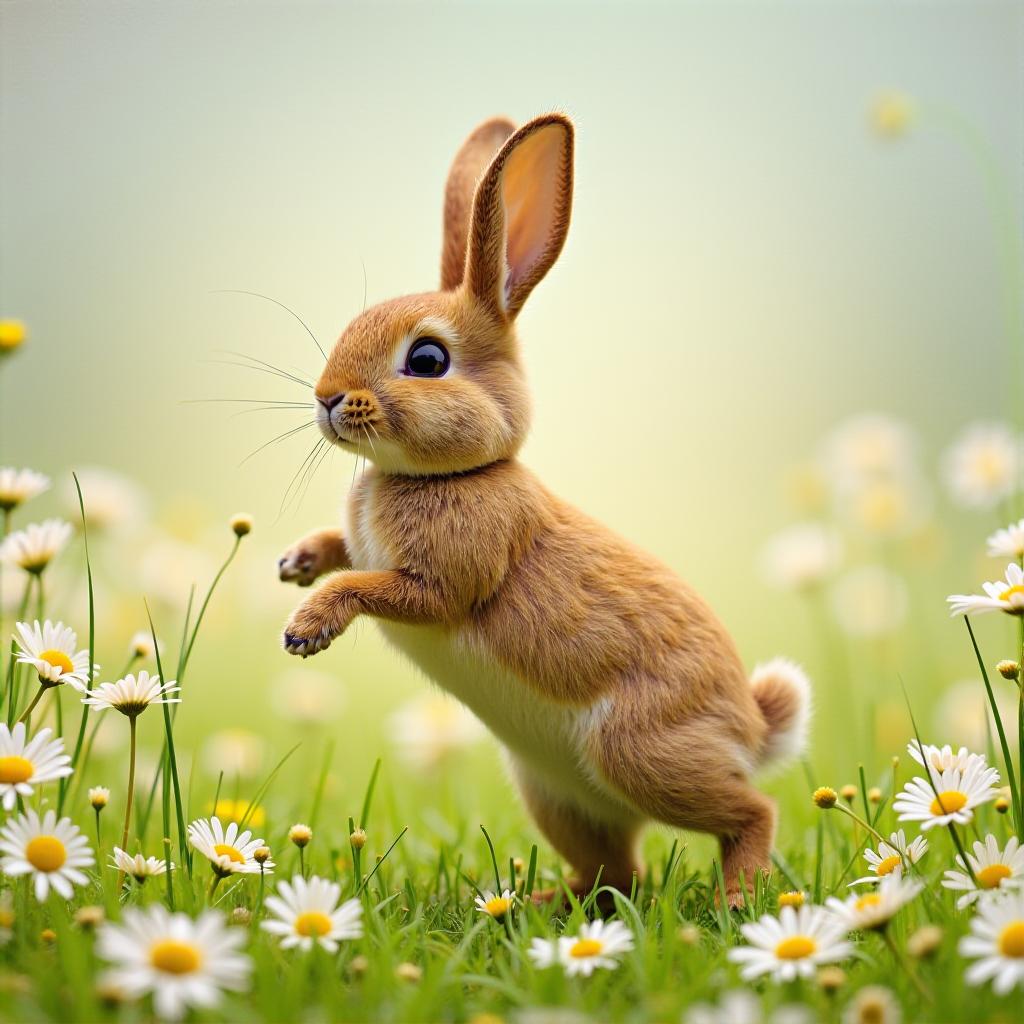}} &
        {\includegraphics[valign=c, width=\ww]{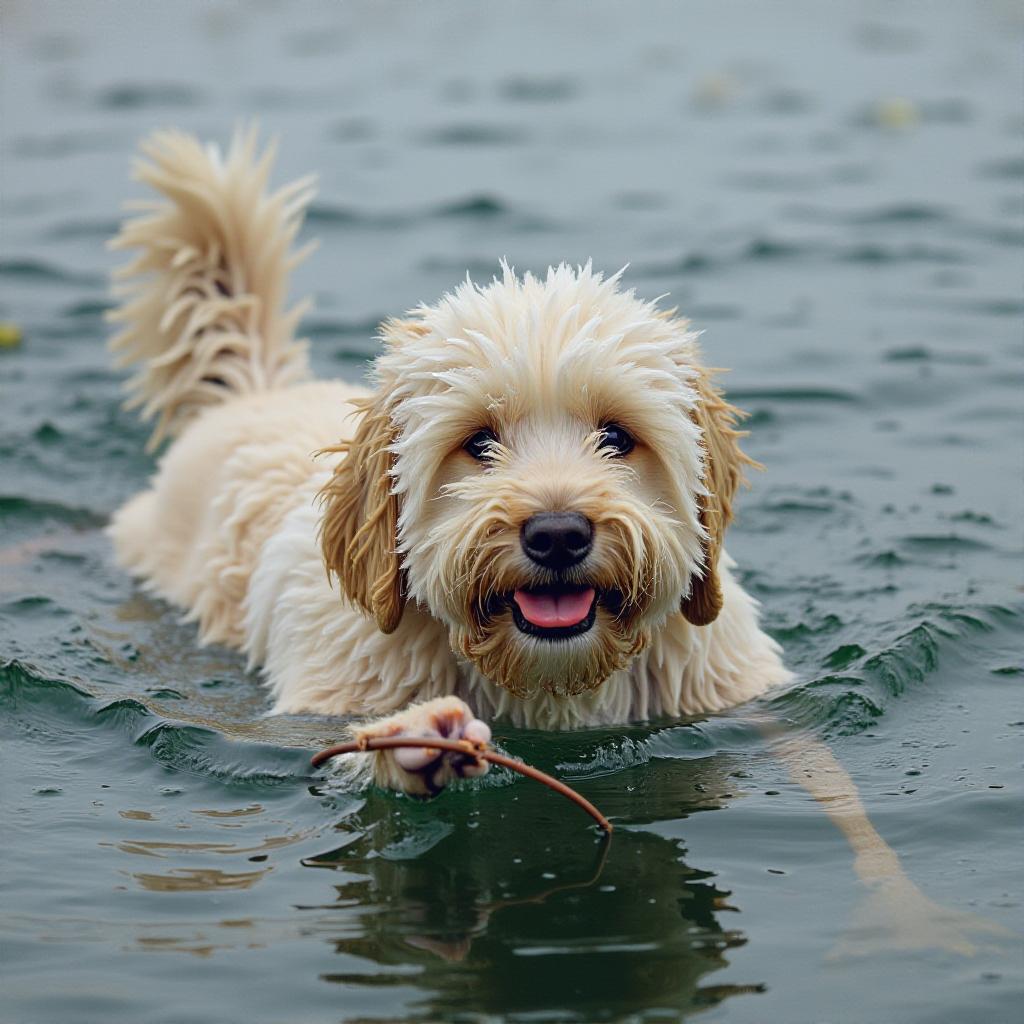}}
        \vspace{1px}
        \\

        \rotatebox[origin=c]{90}{\nonvital{\footnotesize{$G_{52}$}}} &
        {\includegraphics[valign=c, width=\ww]{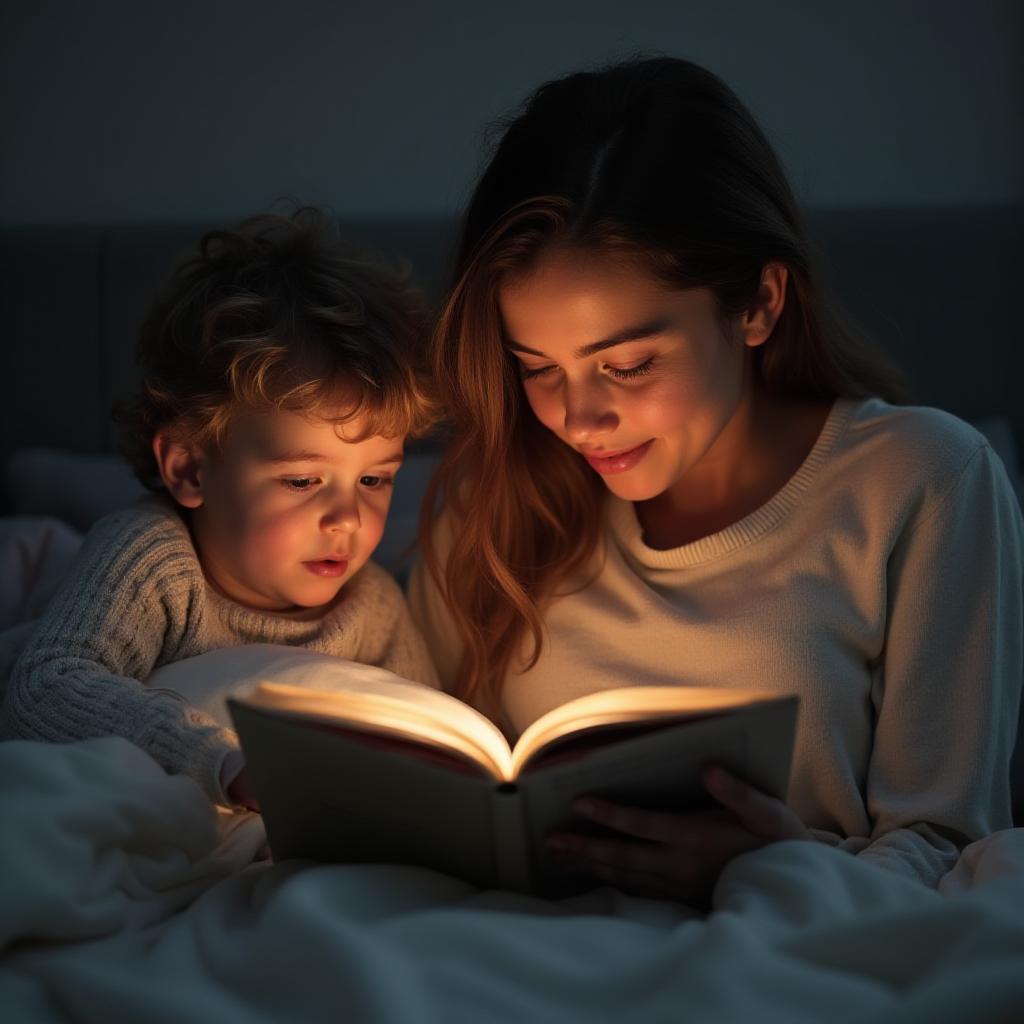}} &
        {\includegraphics[valign=c, width=\ww]{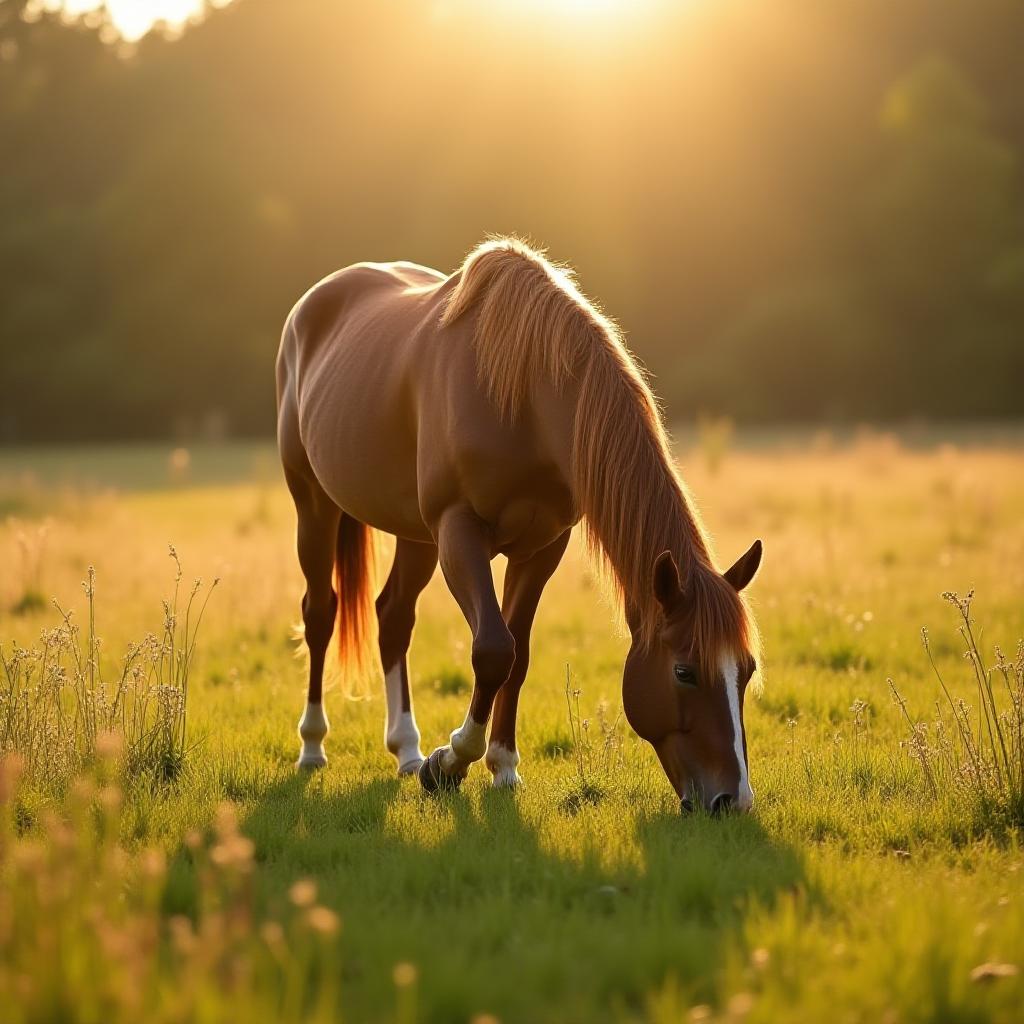}} &
        {\includegraphics[valign=c, width=\ww]{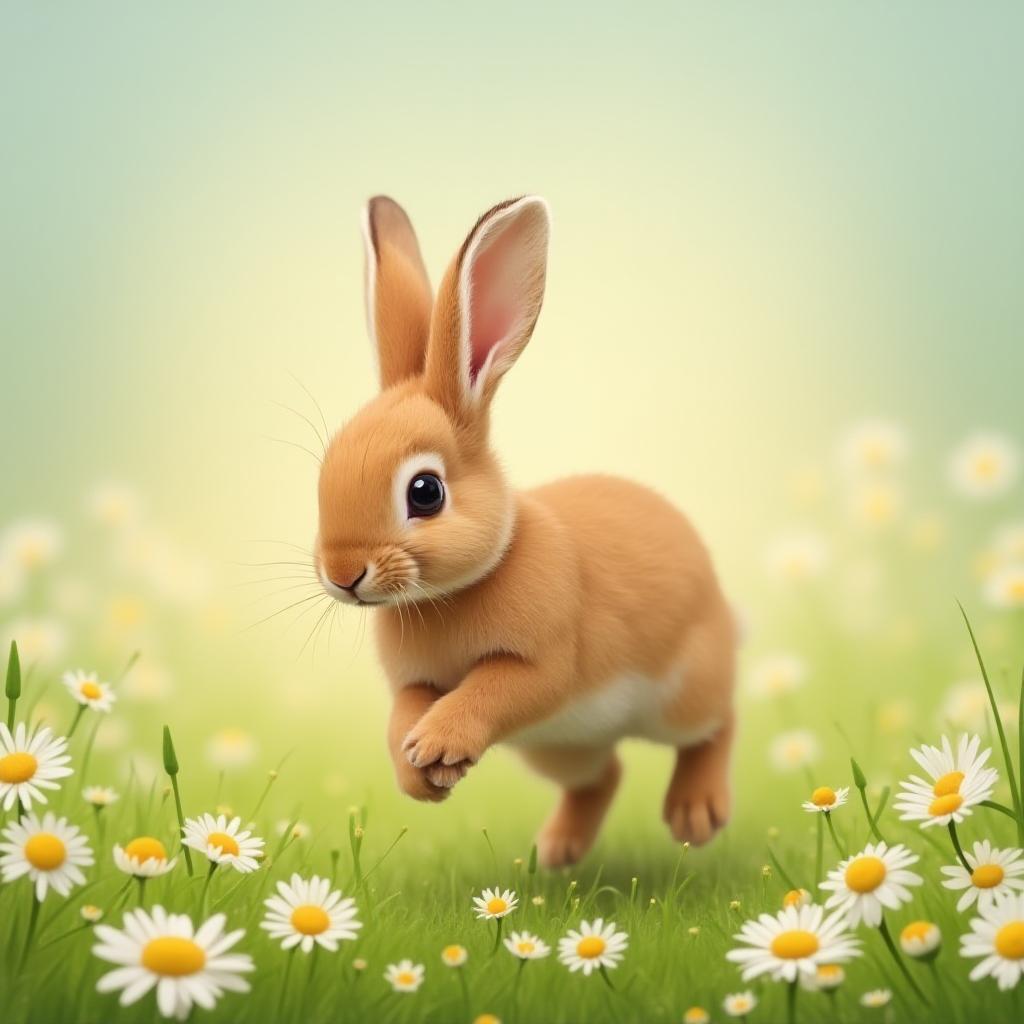}} &
        {\includegraphics[valign=c, width=\ww]{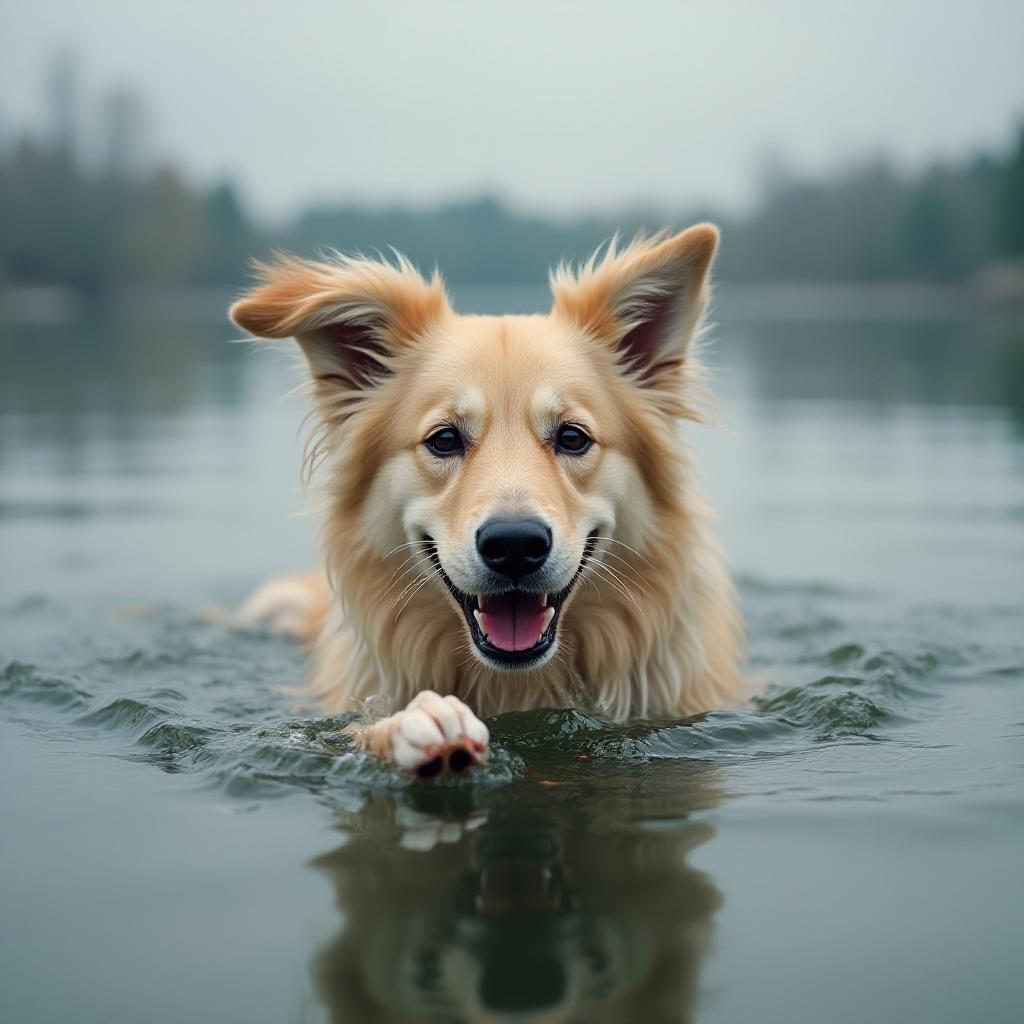}}
        \vspace{1px}
        \\

        \rotatebox[origin=c]{90}{\vital{\footnotesize{$G_{56}$}}} &
        {\includegraphics[valign=c, width=\ww]{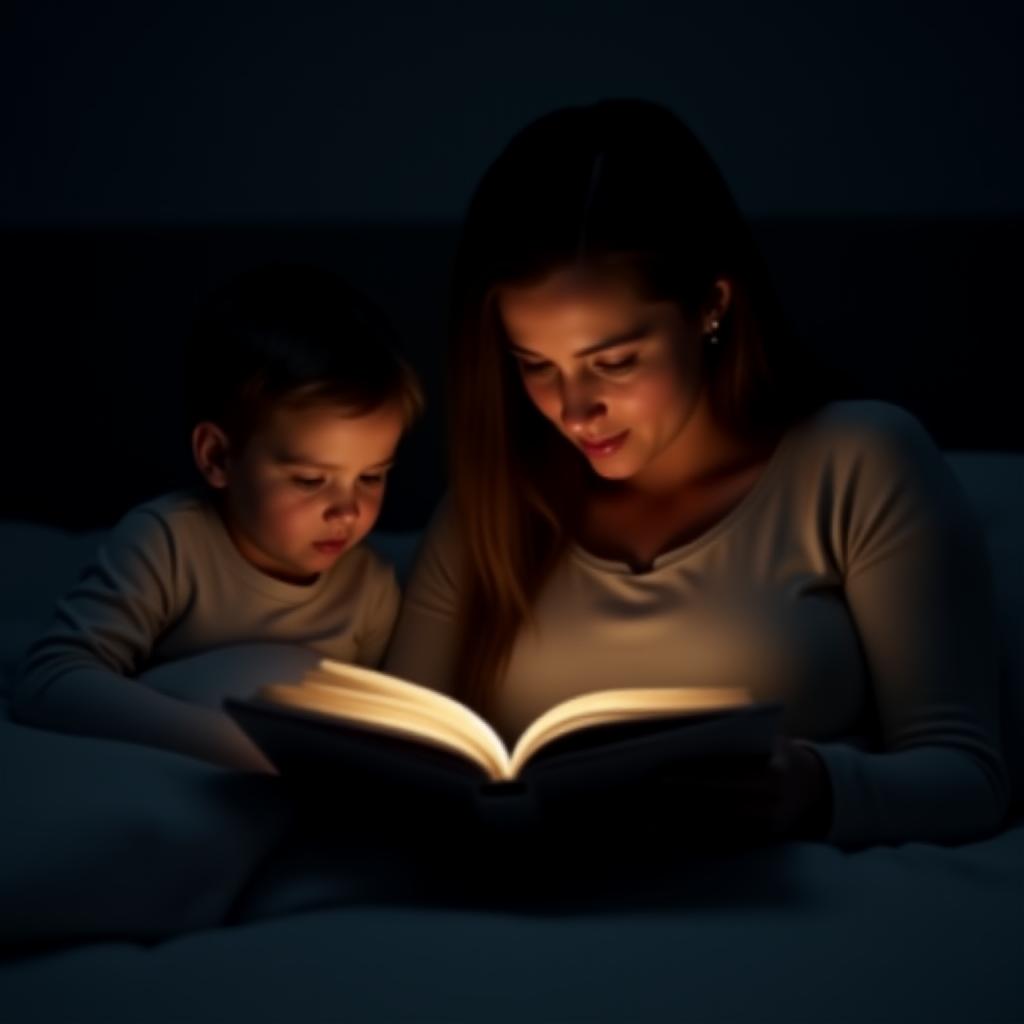}} &
        {\includegraphics[valign=c, width=\ww]{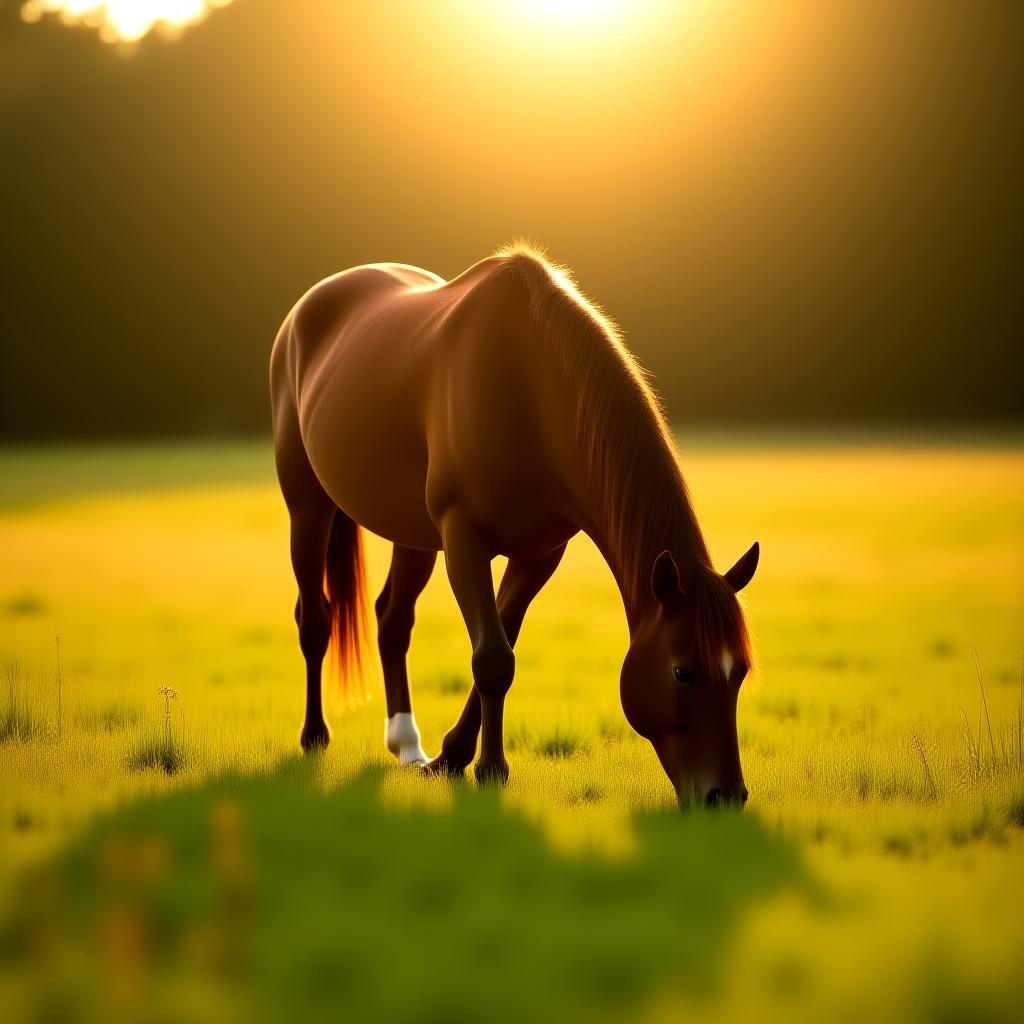}} &
        {\includegraphics[valign=c, width=\ww]{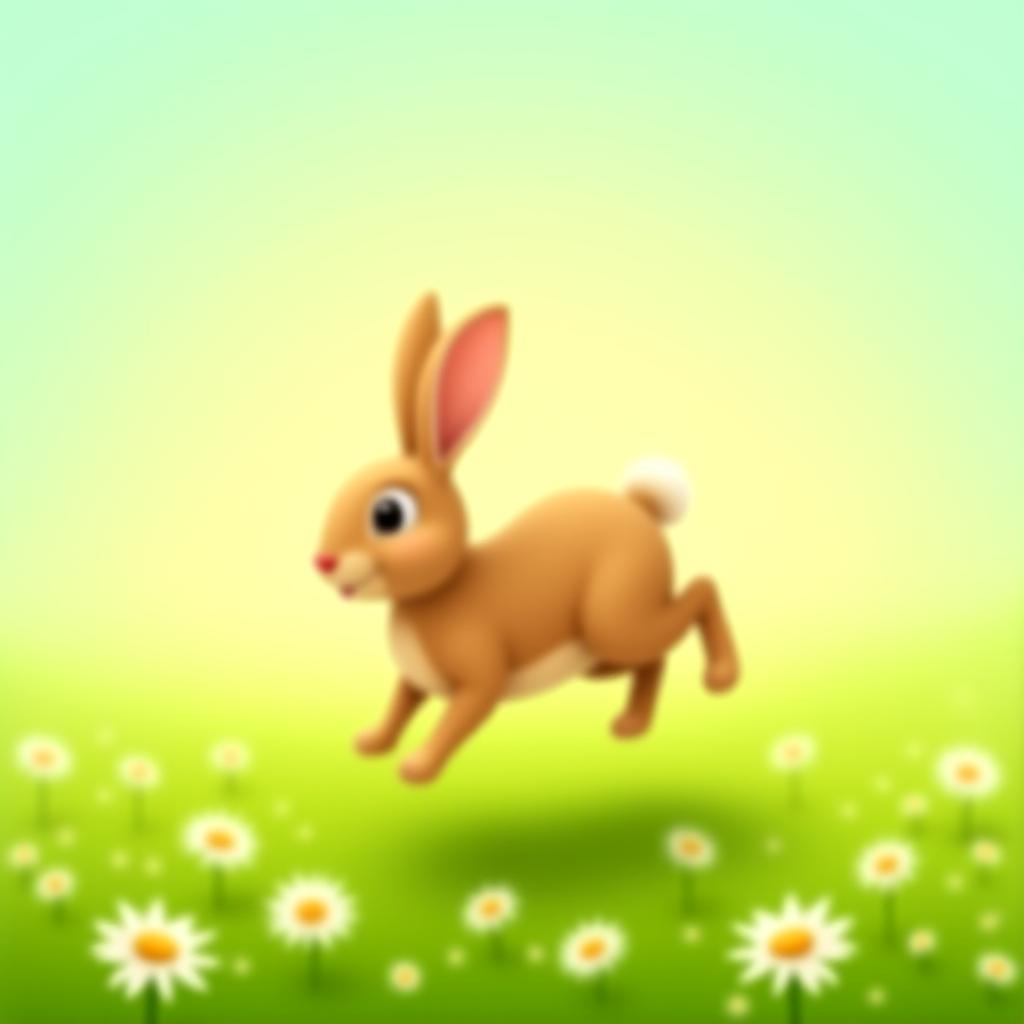}} &
        {\includegraphics[valign=c, width=\ww]{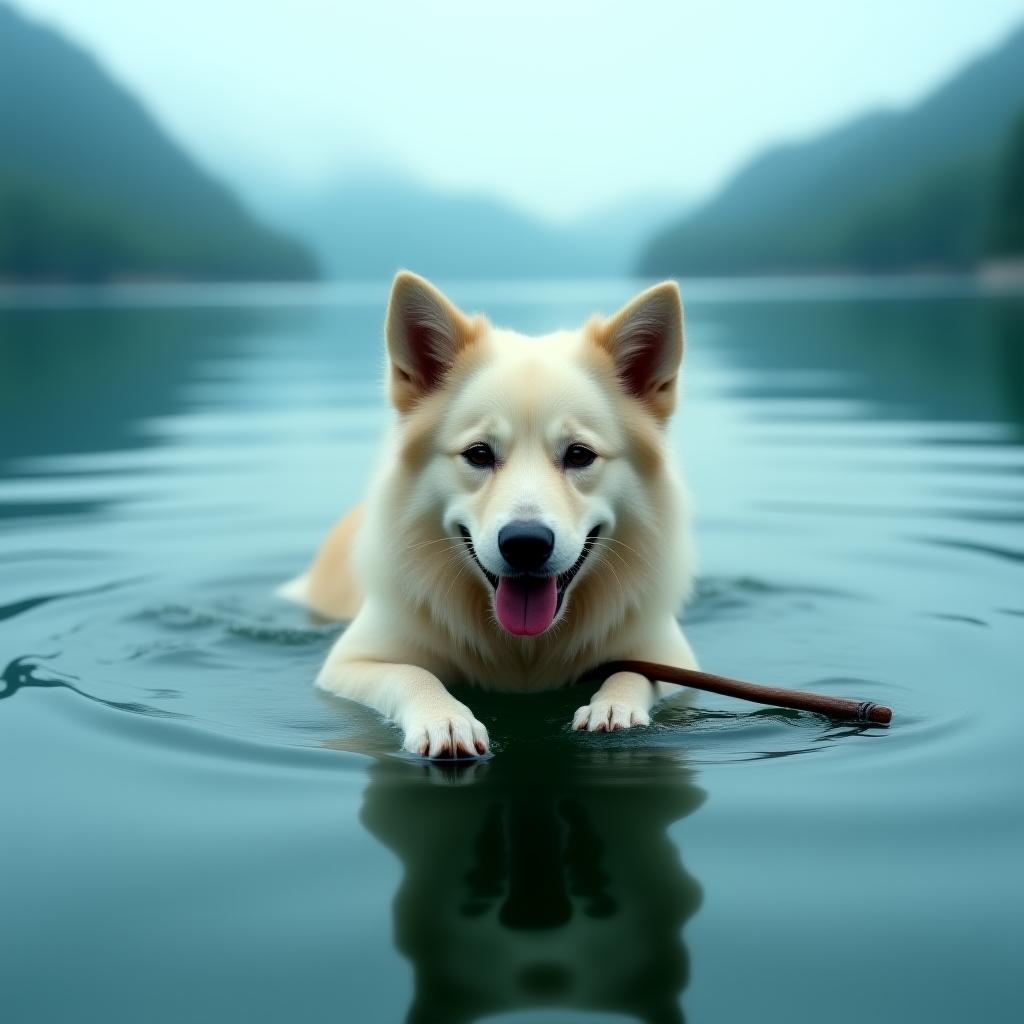}}
        \vspace{0.01px}
        \\

    \end{tabular}
    \caption{\textbf{Layer Removal Qualitative Comparison.} As explained in \Cref{sec:layers_importance}, we illustrate the qualitative differences between \vital{vital} and \nonvital{non-vital} layers. While bypassing  \nonvital{non-vital} layers (\nonvital{$G_{5}$} and \nonvital{$G_{52}$}) results in minor alterations, bypassing \vital{vital} layers leads to significant changes: complete noise generation (\vital{$G_{0}$}), global structure and identity changes (\vital{$G_{18}$}), and alterations in texture and fine details (\vital{$G_{56}$}).}
    \label{fig:layer_removal_qualitative}
    \vspace{-15px}
\end{figure}

\Cref{fig:layer_removal_qualitative} illustrates the qualitative differences between vital and non-vital layers. While bypassing non-vital layers results in minor alterations, removing vital layers leads to significant changes: complete noise generation ($G_{0}$), global structure and identity changes ($G_{18}$), and alterations in texture and fine details ($G_{56}$).

\begin{figure}[t]
    \begin{tikzpicture} [thick,scale=0.9, every node/.style={scale=1}]
        \begin{axis}[
            xlabel={\small{Layer Index}},
            ylabel={\small{Perceptual Similarity}},
            compat=newest,
            xtick distance=5,
            width=9.5cm,
            height=6.5cm,
            scatter/classes={
                a={mark=*, fill=nonvitalcolor},
                b={mark=*, fill=vitalcolor}
            }
            ]
            \addplot[scatter, only marks, scatter src=explicit symbolic]
            table[meta=label] {
                x y label
                3 0.8879982 a
                4 0.8886729 a
                5 0.92280376 a
                6 0.8878546 a
                7 0.9249694 a
                8 0.8918034 a
                9 0.9145046 a
                10 0.9319223 a
                11 0.89892554 a
                12 0.9230972 a
                13 0.9419161 a
                14 0.9193646 a
                15 0.903646 a
                16 0.8901205 a
                17 0.8663492 b
                18 0.7881612 b
                19 0.90022004 a
                20 0.9195701 a
                21 0.93238455 a
                22 0.9055613 a
                23 0.89533365 a
                24 0.8947287 a
                25 0.87295866 b
                26 0.88893557 a
                27 0.8900309 a
                28 0.8617277 b
                29 0.91026425 a
                30 0.8965948 a
                31 0.87639654 b
                32 0.89346516 a
                33 0.9110552 a
                34 0.9280884 a
                35 0.94198036 a
                36 0.934934 a
                37 0.93723893 a
                38 0.9468137 a
                39 0.9504659 a
                40 0.94309264 a
                41 0.95357776 a
                42 0.9421407 a
                43 0.9493426 a
                44 0.9300462 a
                45 0.9336098 a
                46 0.93461245 a
                47 0.9457144 a
                48 0.93769985 a
                49 0.93422663 a
                50 0.92652655 a
                51 0.91905427 a
                52 0.92307425 a
                53 0.8525807 b
                54 0.8624263 b
                55 0.91839397 a
                56 0.86762947 b
            };
            \end{axis}
    \end{tikzpicture}

    \vspace{-5px}
    \caption{\textbf{Layer Removal Quantitative Comparison.} As explained in \Cref{sec:layers_importance}, we measured the effect of removing each layer of the model by calculating the perceptual similarity between the generated images with and without this layer. Lower perceptual similarity indicates significant changes in the generated images (\Cref{fig:layer_removal_qualitative}). As can be seen, removing certain layers significantly affects the generated images, while others have minimal impact. Importantly, influential layers are distributed across the transformer rather than concentrated in specific regions. Note that the first vital layers were omitted for clarity (as their perceptual similarity approached zero).}
    \label{fig:layer_removal_quantiative}
    \vspace{-10px}
\end{figure}

\begin{figure*}[t]
    \centering
    \includegraphics[width=1\linewidth]{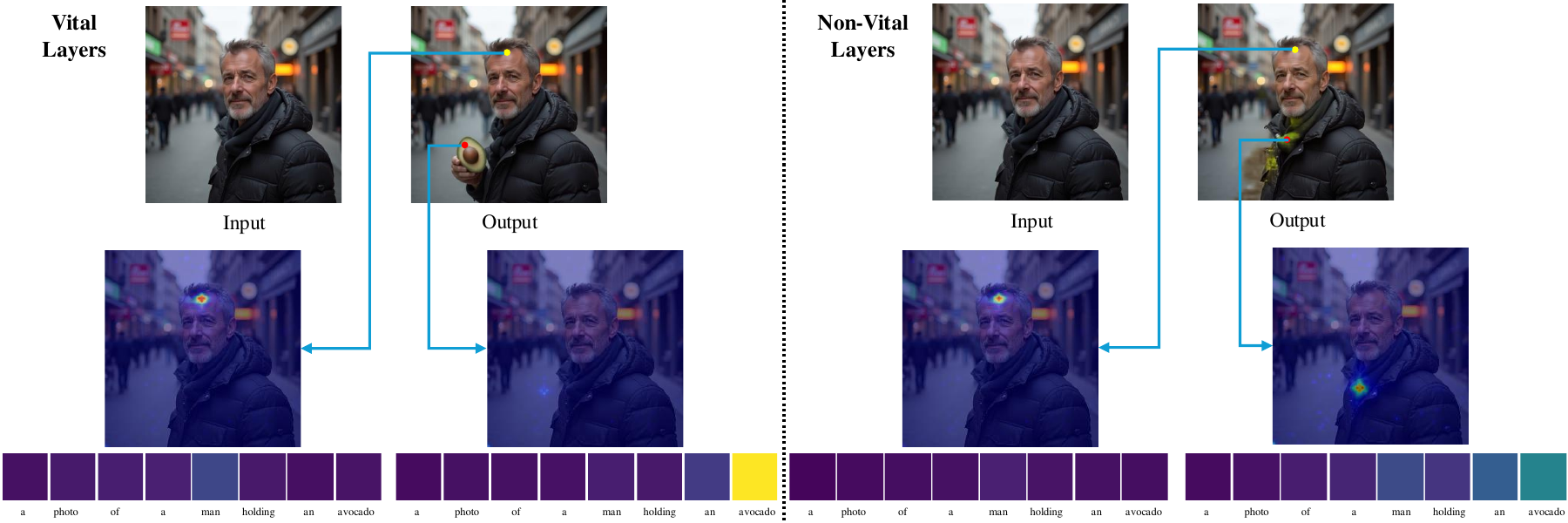}
    \caption{\textbf{Multi-Modal Attention Distribution.} Given an input image of a man, we edit it to hold an avocado by injecting the reference image attention activations in the vital layers only (left), or in the non-vital layers (right), and visualize the multimodal attention of two points: a yellow point in a region that should remain unchanged (requiring copying from the reference image), and a red point in an area targeted for editing (requiring generation based on the text prompt). As can be seen, in vital layers (left), points meant to remain unchanged show dominant attention to visual features, while points targeted for editing exhibit stronger attention to relevant text tokens (e.g., \prompt{avocado}). Conversely, non-vital layers (right) show predominantly image-based attention even in regions marked for editing. This suggests that injecting features into vital layers strikes a good multimodal attention balance between preserving source content and incorporating text-guided modifications.}
    \label{fig:attention_visualization}
    \vspace{-10px}
\end{figure*}

\subsection{Image Editing using Vital Layers}
\label{sec:image_editing}

Given a source image $x$ generated with a known seed $s$ and prompt $p$, we aim to modify $p$ and generate an edited image $\hat{x}$ that exhibits the desired changes, while otherwise preserving the source content. We adapt the self-attention injection mechanism, previously shown effective for image and video editing~\cite{Wu2022TuneAVideoOT, cao2023masactrl} in UNet-based diffusion models, to the DiT-based FLUX architecture. Since each DiT layer processes a sequence of image and text embeddings, we propose generating both $x$ and $\hat{x}$ in parallel while \emph{selectively replacing} the image embeddings of $\hat{x}$ with those of $x$, but only within the vital layers set $V$.

Remarkably, as shown in \Cref{fig:teaser}, this training-free approach successfully performs diverse editing tasks, including non-rigid deformations, object addition, object replacement, and global modifications, all using the \emph{same} set of vital layers $V$.

To understand this effectiveness, we analyze the multimodal attention patterns in FLUX. Each visual token simultaneously attends to all visual and text tokens, with attention weights normalized across both modalities. \Cref{fig:attention_visualization} contrasts the attention patterns in vital versus non-vital layers at two key points: a yellow point in a region that should remain unchanged (requiring copying from the reference image), and a red point in an area targeted for editing (requiring generation based on the text prompt). In vital layers (left), points meant to remain unchanged show dominant attention to visual features, while points targeted for editing exhibit stronger attention to relevant text tokens (e.g., \prompt{avocado}). Conversely, non-vital layers (right) show predominantly image-based attention even in regions marked for editing. This suggests that injecting features into vital layers strikes a good multimodal attention balance between preserving source content and incorporating text edits.

\subsection{Latent Nudging for Real Image Editing}
\label{sec:latent_nudging}

\begin{figure}[tp]
    \centering
    \setlength{\tabcolsep}{0.6pt}
    \renewcommand{\arraystretch}{0.8}
    \setlength{\ww}{0.29\columnwidth}
    \begin{tabular}{cccc}
        \rotatebox[origin=c]{90}{\footnotesize{(a) w/o nudging}} &
        {\includegraphics[valign=c, width=\ww]{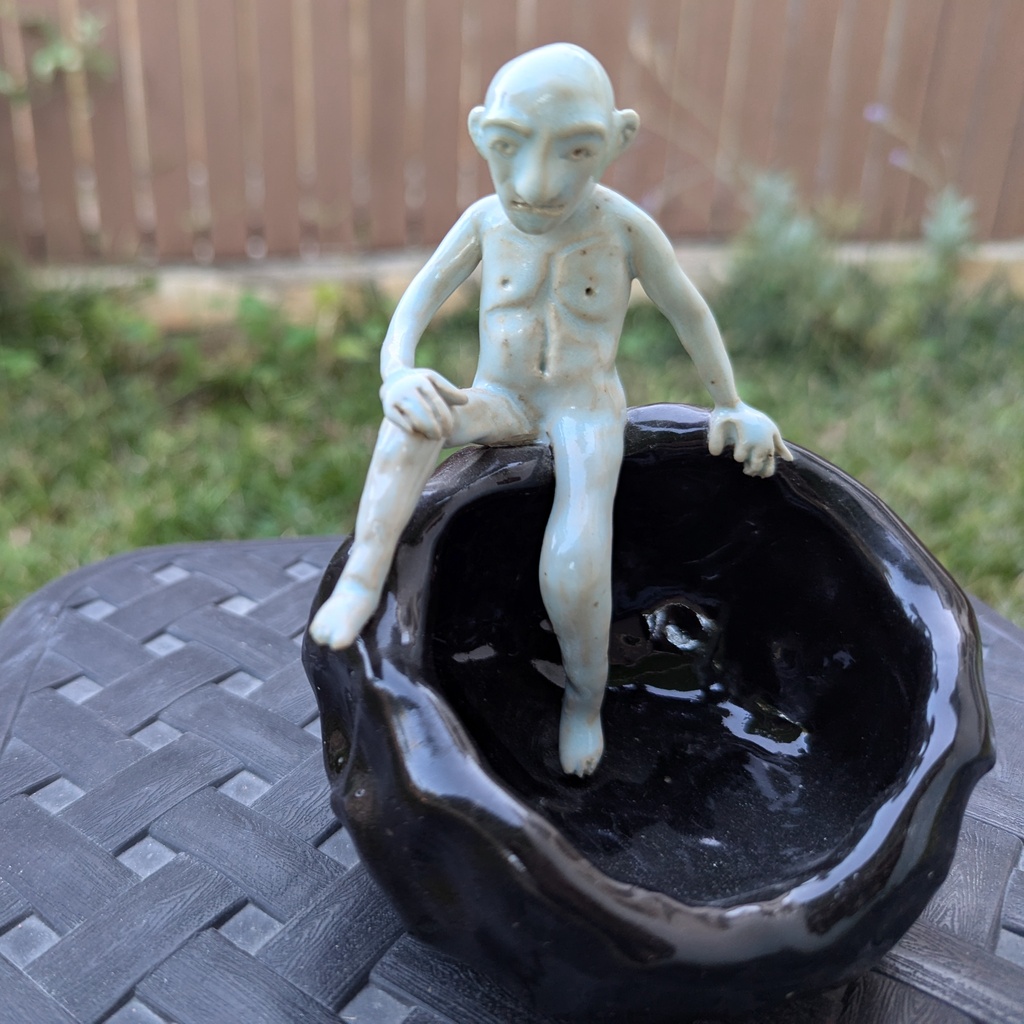}} &
        {\includegraphics[valign=c, width=\ww]{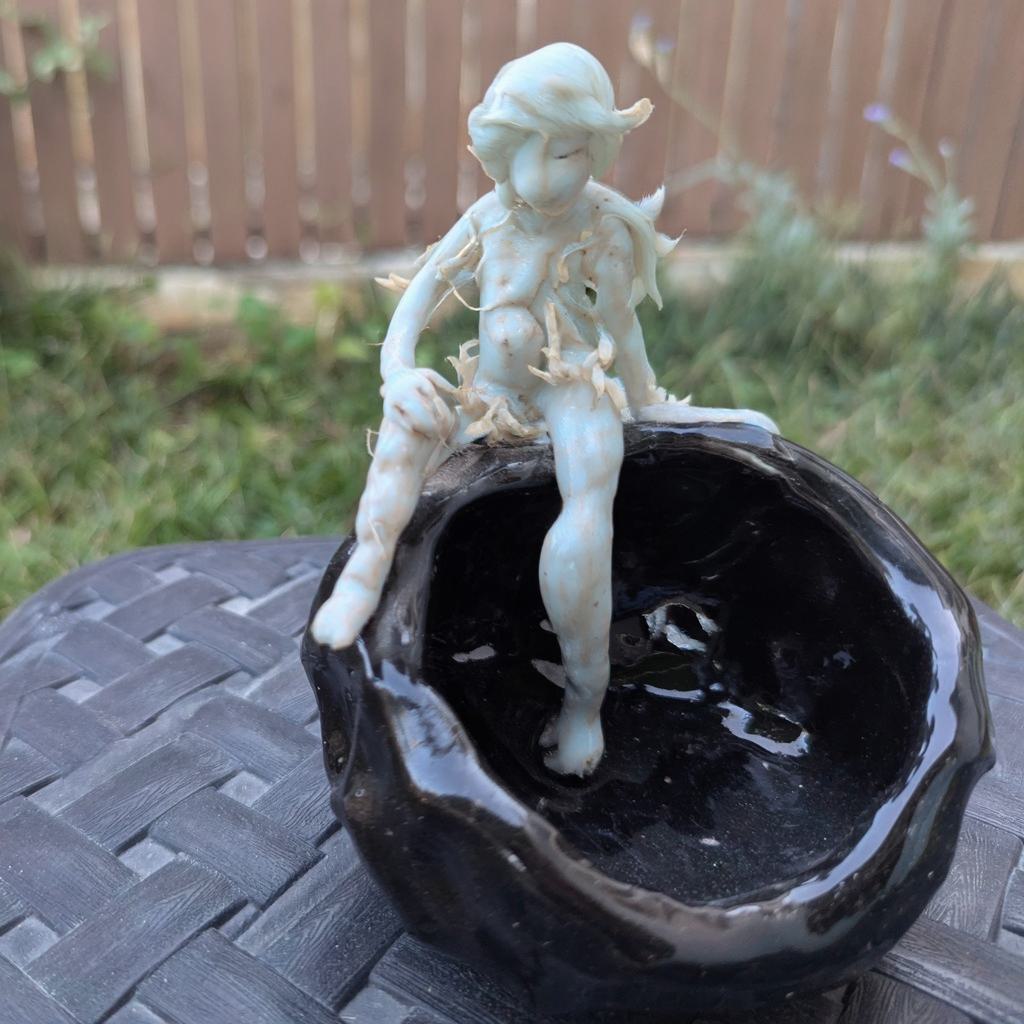}} &
        {\includegraphics[valign=c, width=\ww]{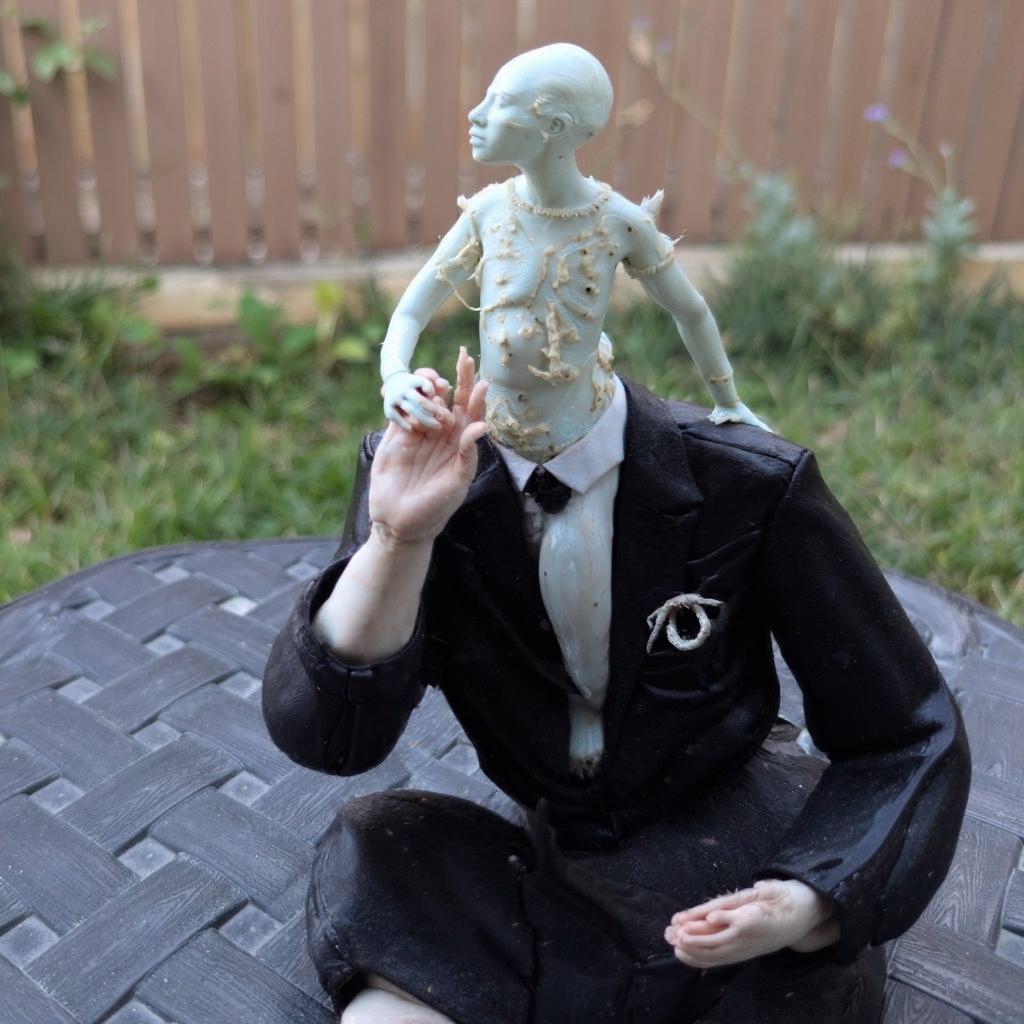}}
        \vspace{3px}
        \\

        \rotatebox[origin=c]{90}{\footnotesize{(b) w nudging}} &
        {\includegraphics[valign=c, width=\ww]{figures/latent_nudging/assets/inp.jpg}} &
        {\includegraphics[valign=c, width=\ww]{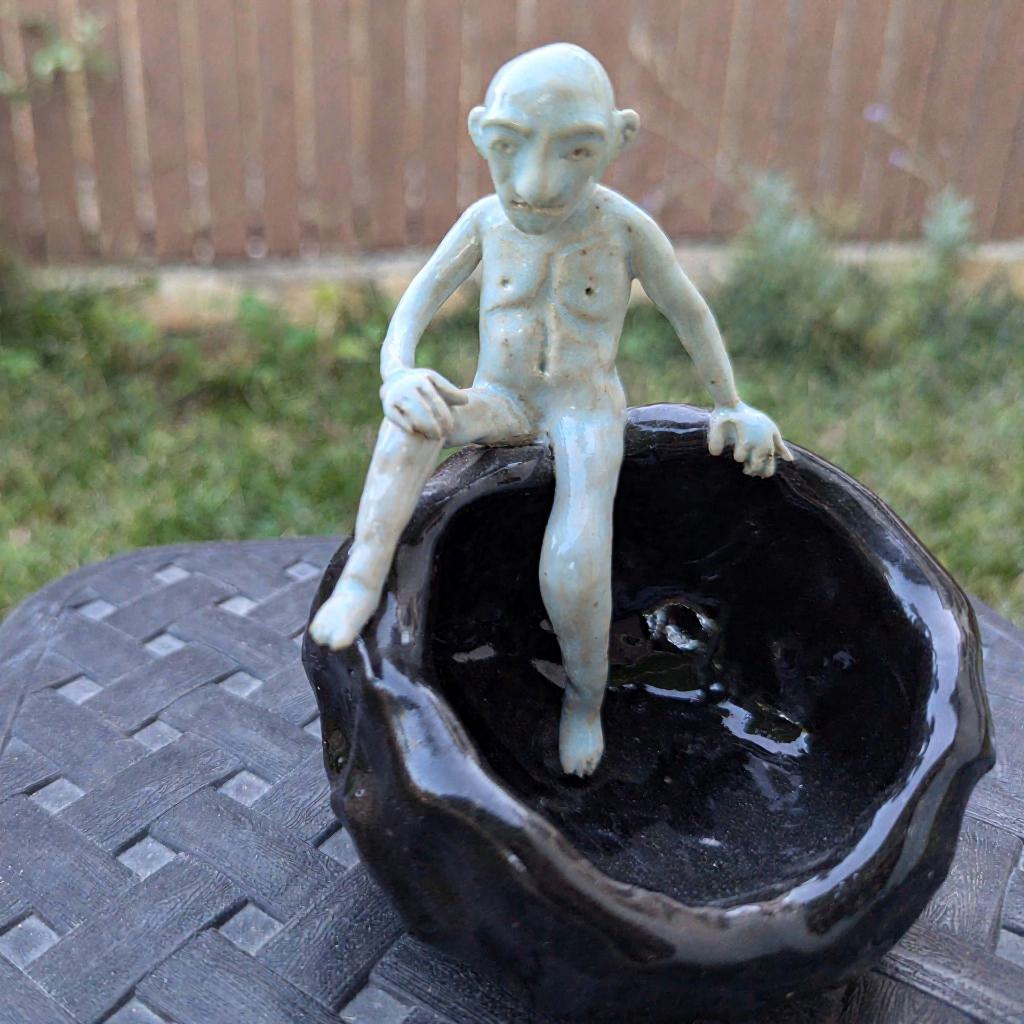}} &
        {\includegraphics[valign=c, width=\ww]{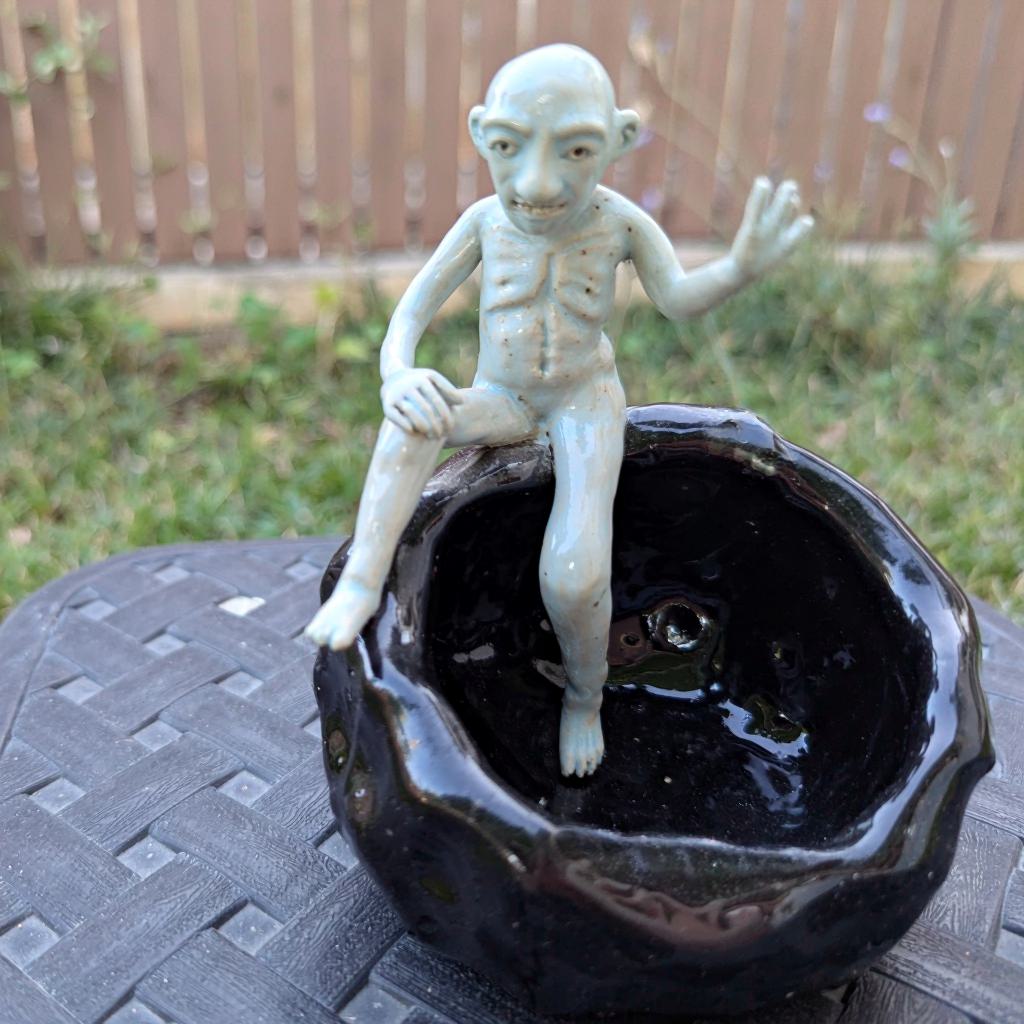}}
        \vspace{1px}
        \\

        &
        \footnotesize{Input image} &
        \footnotesize{Reconstruction} &
        \footnotesize{\prompt{Raising its hand}}

    \end{tabular}
    \vspace{-5px}
    \caption{\textbf{Latent Nudging.} As described in \Cref{sec:latent_nudging}, when inverting a real image, (a) a simple inverse Euler ODE solver leads to corrupted image reconstructions and unintended modifications during editing. On the other hand, (b) using our latent nudging technique significantly reduces reconstruction errors and better constrains edits to the intended regions.}
    \label{fig:latent_nudging}
    \vspace{-11px}
\end{figure}

Flow models generate samples by matching a prior distribution $p_0$ (Gaussian noise) to a data distribution $p_1$ (the image manifold). In the space $\mathbb{R}^d$, we define two key components: a probability density path $p_t: [0, 1] \times \mathbb{R}^d \rightarrow \mathbb{R}_{>0}$, which specifies time-dependent probability density functions ($\int p_t(x)dx = 1$), and a vector field $u_t:[0,1]\times \mathbb{R}^d \rightarrow \mathbb{R}^d$. This vector field generates a flow $\phi:[0,1]\times \mathbb{R}^d \rightarrow \mathbb{R}^d$ through the ordinary differential equation (ODE): $\frac{d}{dt}\phi_t(x) = u_t(\phi_t(x)); \phi_0(x)= x$. Transforming a sample from $p_0$ to a sample in $p_1$ is 
achieved using ODE solvers such as Euler.

To edit real images, we must first \emph{invert} them into the latent space, transforming samples from $p_1$ to $p_0$. We initially implemented an inverse Euler ODE solver for FLUX by reversing the vector field prediction. Given the forward Euler step:
\vspace{-5px}
\begin{equation}
z_{t-1} = z_{t} + (\sigma_{t+1} - \sigma_{t}) * u_t(z_t)
\end{equation}
where $z_t$ represents the latent at timestep $t$, $\sigma_{t}$ is the optimal transport standard deviation at time $t$, and $u_t$ is the learned vector field, we proposed the inverse step:
\begin{equation}
z_{t} = z_{t-1} + (\sigma_{t} - \sigma_{t+1}) * u_t(z_{t-1})
\end{equation}
assuming $u_t(z_{t}) \approx u_t(z_{t-1})$ for small steps.

However, as \Cref{fig:latent_nudging}(a) demonstrates, this approach proves insufficient for FLUX, resulting in corrupted image reconstructions and unintended modifications during editing. We hypothesize that the assumption $u(z_{t}) \approx u(z_{t-1})$ does not hold, which causes the model to significantly alter the image during the forward process. To address this, we introduce \emph{latent nudging}: multiplying the initial latent $z_0$ by a small scalar $\lambda=1.15$ to slightly offset it from the training distribution. While this modification is visually imperceptible (\Cref{fig:latent_nudging}(b)), it significantly reduces reconstruction errors and constrains edits to the intended regions. See the supplementary material for more details and ablations.

\section{Experiments}
\label{sec:experiments}

In \Cref{sec:comparisons} we compare our method against its baselines, both qualitatively and quantitatively. Next, in \Cref{sec:user_study} we conduct a user study and report results. Furthermore, in \Cref{sec:ablation_study} we present the ablation study results. Finally, in \Cref{sec:applications} we demonstrate several applications.

\subsection{Qualitative and Quantitative Comparison}
\label{sec:comparisons}

\begin{figure*}[tp]
    \centering
    \setlength{\tabcolsep}{0.6pt}
    \renewcommand{\arraystretch}{0.8}
    \setlength{\ww}{0.13\linewidth}
    \begin{tabular}{c @{\hspace{10\tabcolsep}} cccccc}

        \footnotesize{Input} &
        \footnotesize{SDEdit}~\cite{meng2021sdedit} &
        \footnotesize{P2P+NTI}~\cite{Hertz2022PrompttoPromptIE, mokady2022null} &
        \footnotesize{Instruct-P2P}~\cite{brooks2022instructpix2pix} &
        \footnotesize{MagicBrush}~\cite{Zhang2023MagicBrush} &
        \footnotesize{MasaCTRL}~\cite{cao2023masactrl} &
        \footnotesize{Stable Flow (ours)}
        \vspace{2px}
        \\

        {\includegraphics[valign=c, width=\ww]{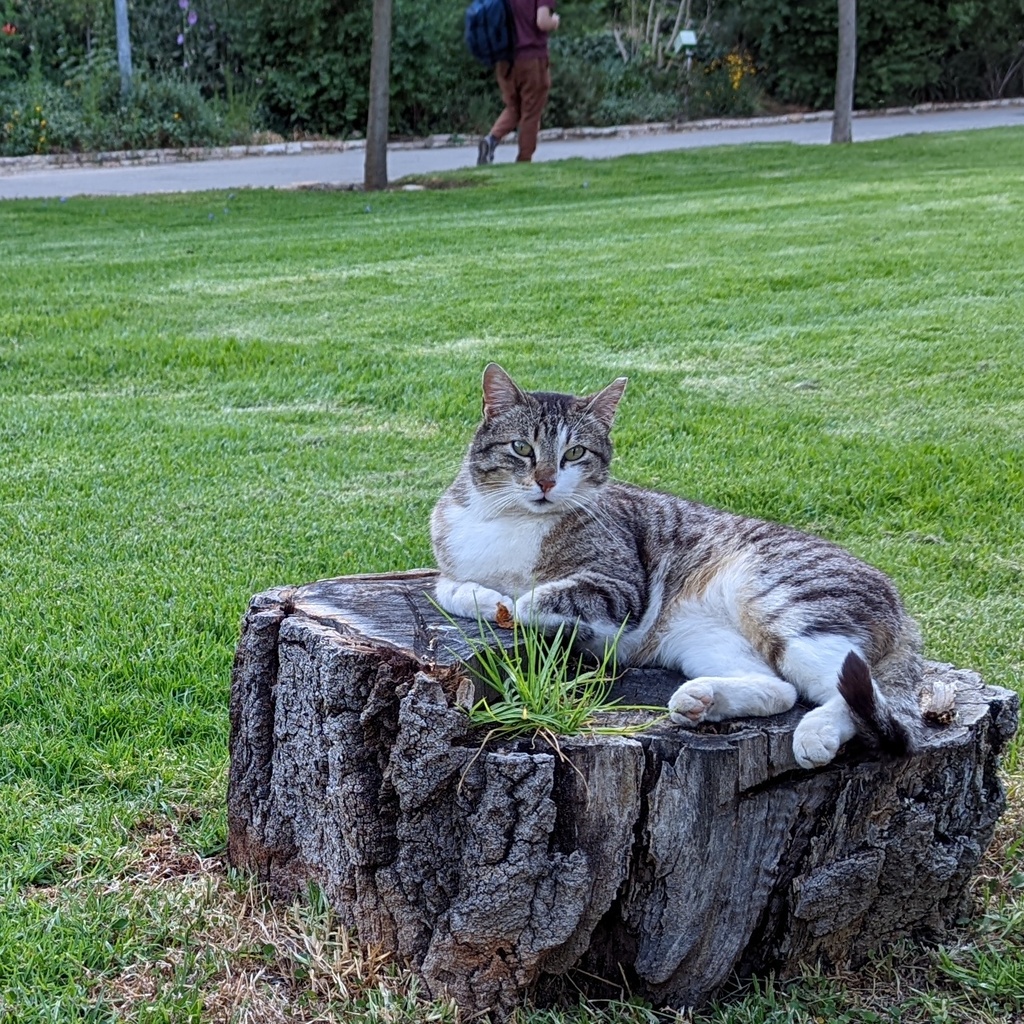}} &
        {\includegraphics[valign=c, width=\ww]{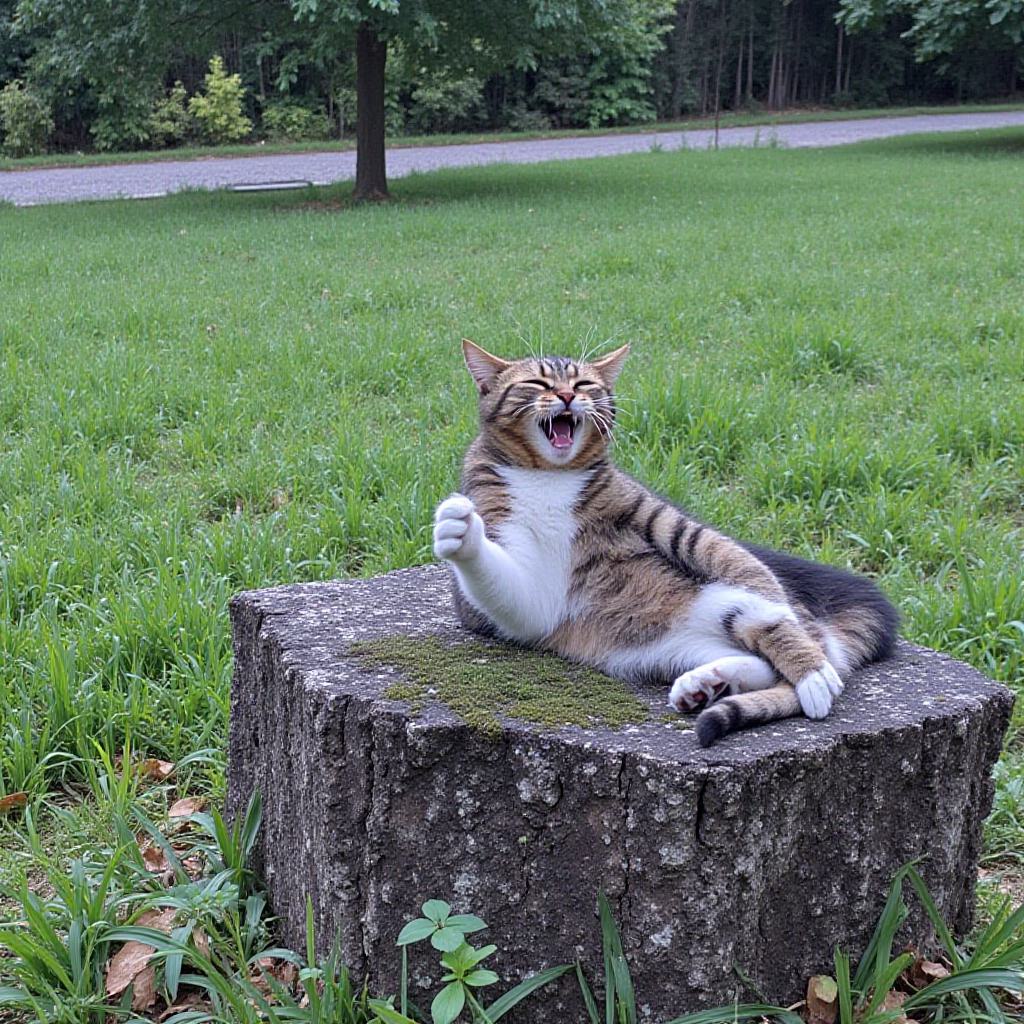}} &
        {\includegraphics[valign=c, width=\ww]{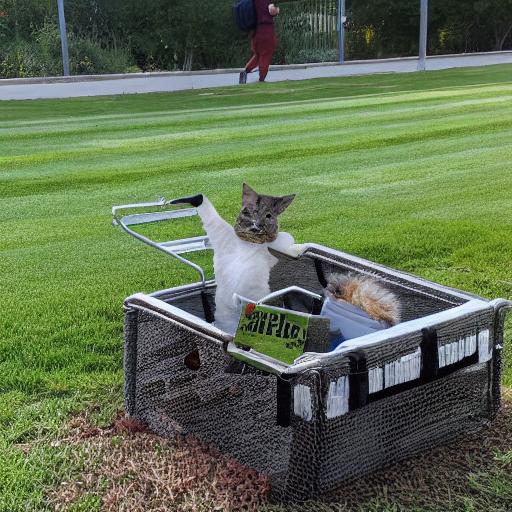}} &
        {\includegraphics[valign=c, width=\ww]{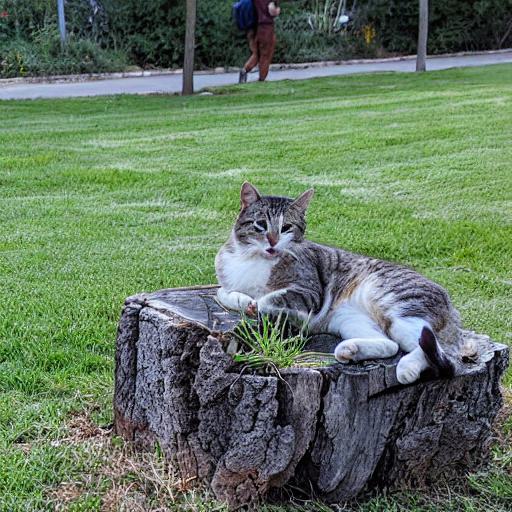}} &
        {\includegraphics[valign=c, width=\ww]{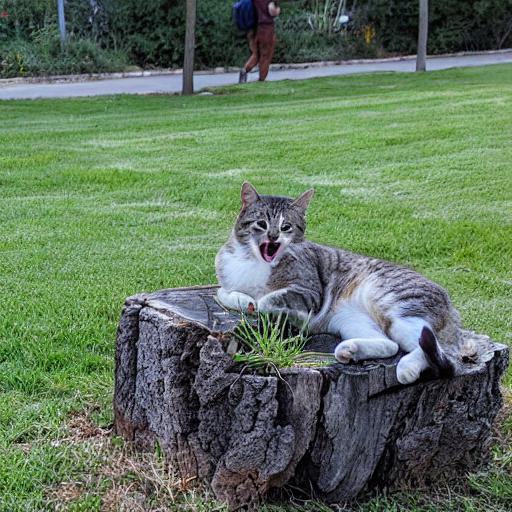}} &
        {\includegraphics[valign=c, width=\ww]{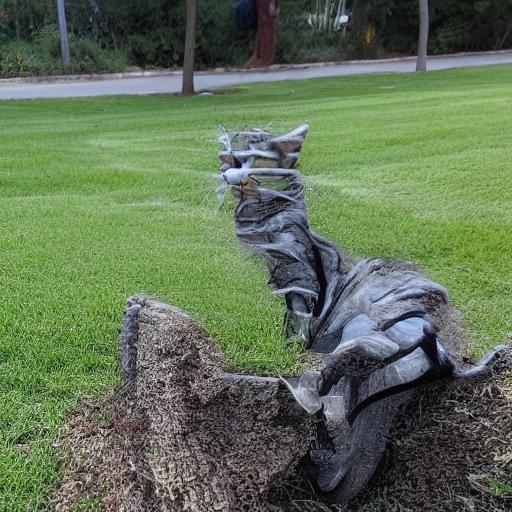}} &
        {\includegraphics[valign=c, width=\ww]{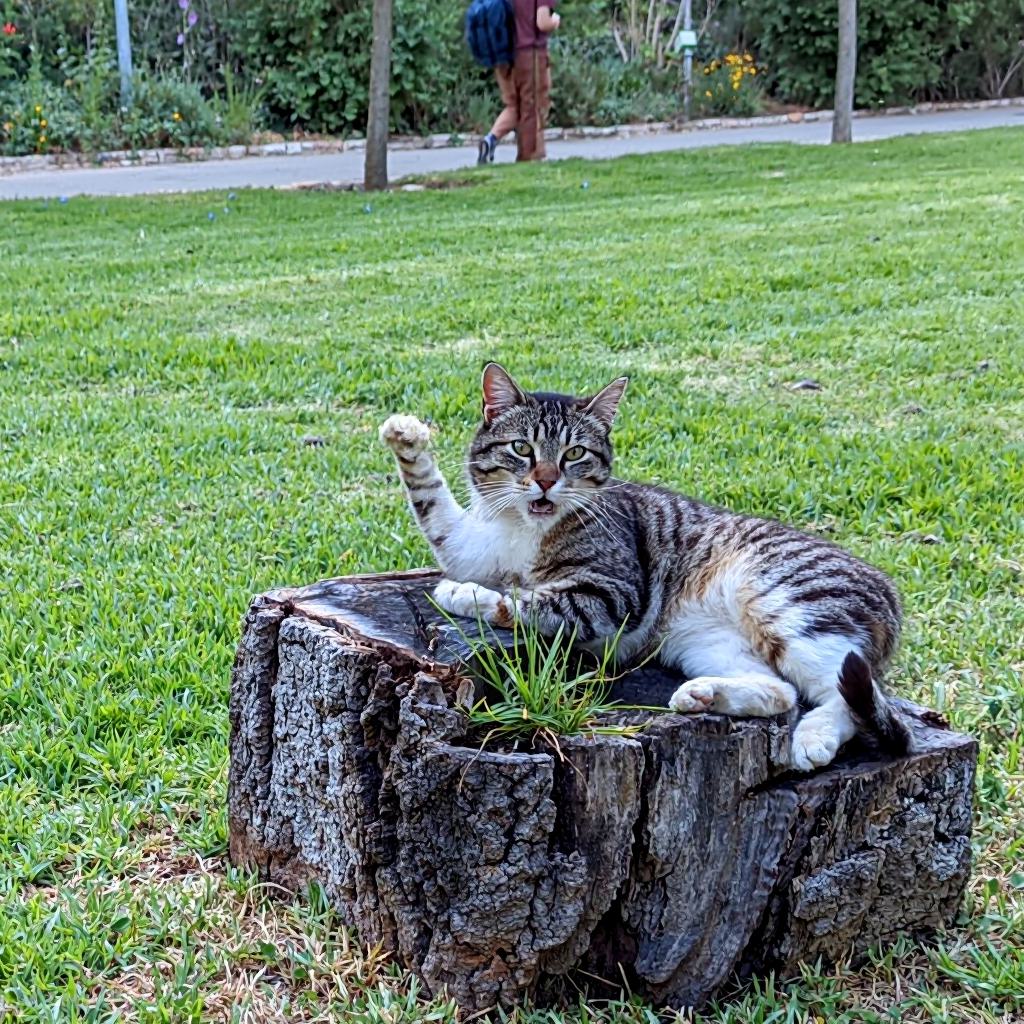}}
        \vspace{1px}
        \\

        &
        \multicolumn{6}{c}{\small{\prompt{The cat is \textbf{yelling} and \textbf{raising} its paw}}}
        \vspace{5px}
        \\

        {\includegraphics[valign=c, width=\ww]{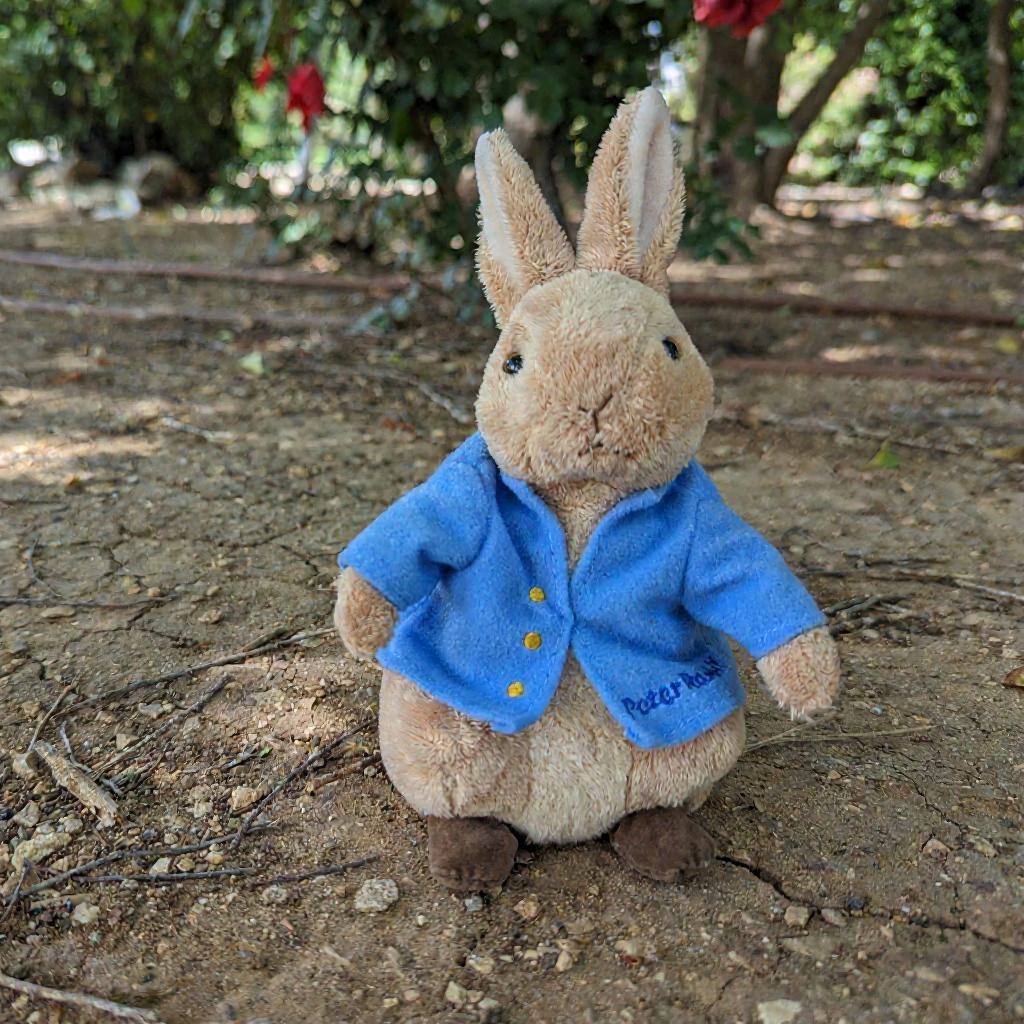}} &
        {\includegraphics[valign=c, width=\ww]{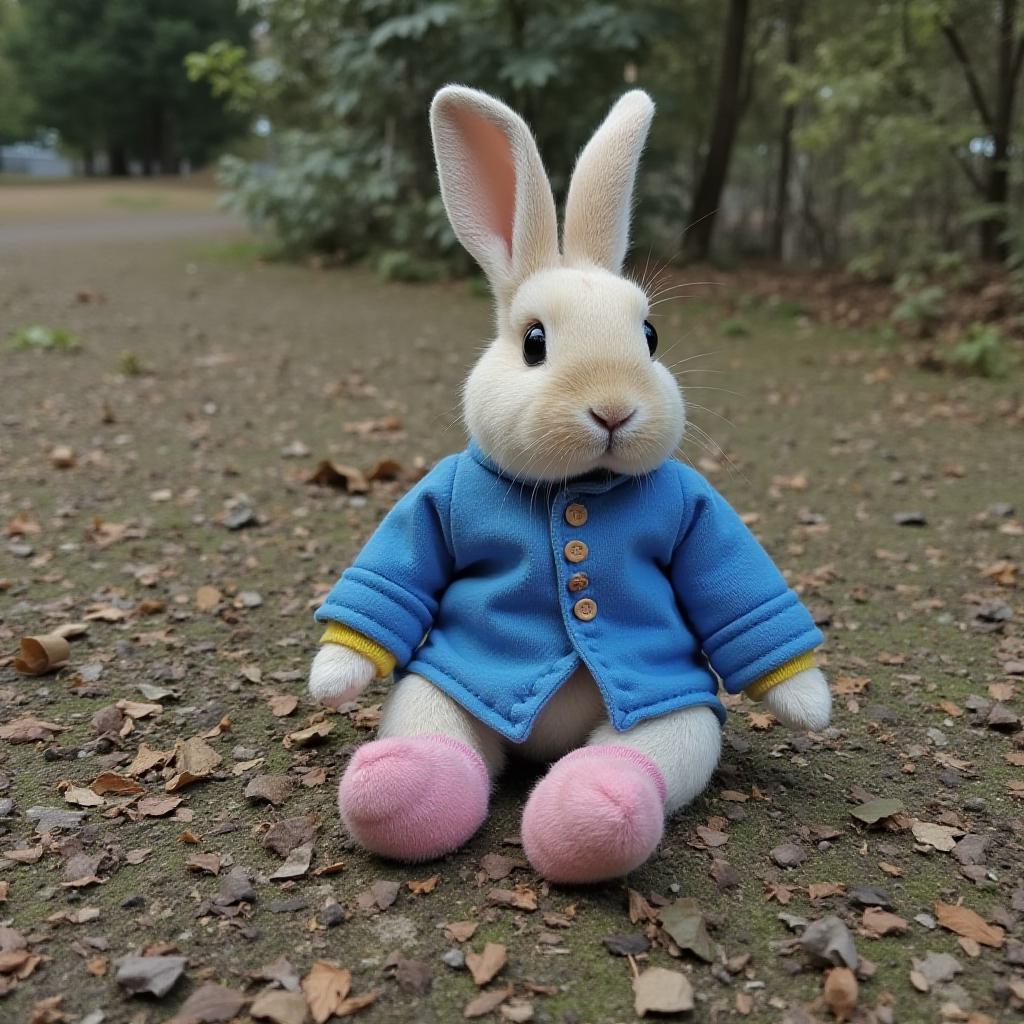}} &
        {\includegraphics[valign=c, width=\ww]{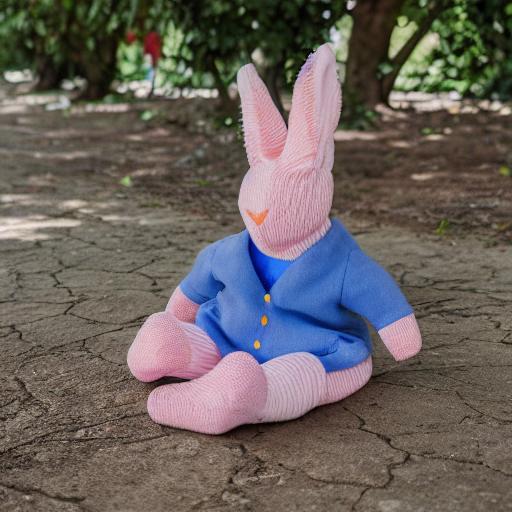}} &
        {\includegraphics[valign=c, width=\ww]{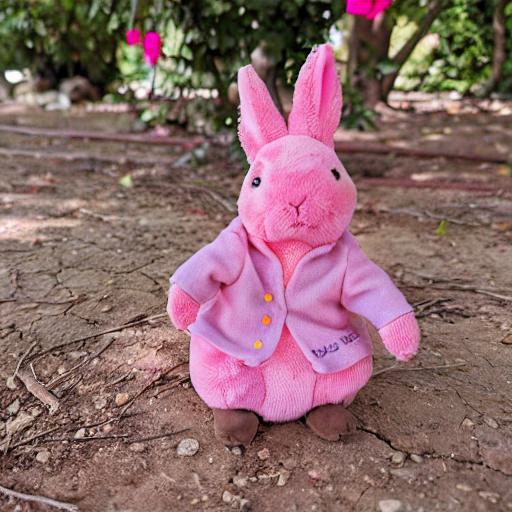}} &
        {\includegraphics[valign=c, width=\ww]{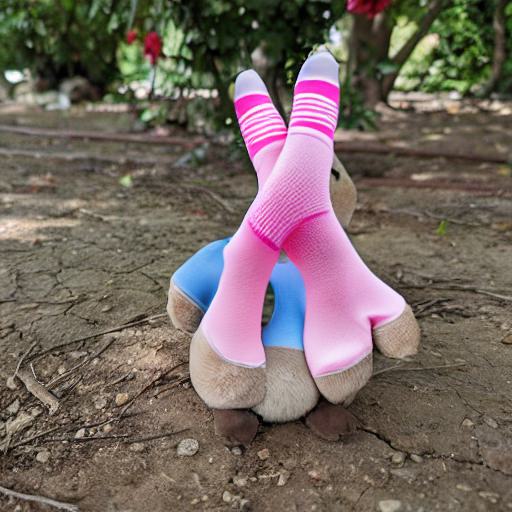}} &
        {\includegraphics[valign=c, width=\ww]{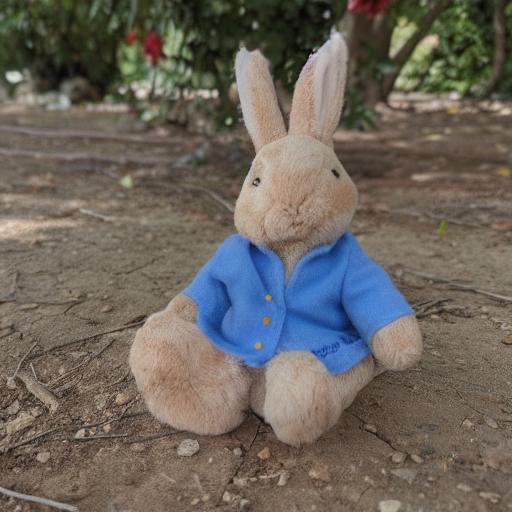}} &
        {\includegraphics[valign=c, width=\ww]{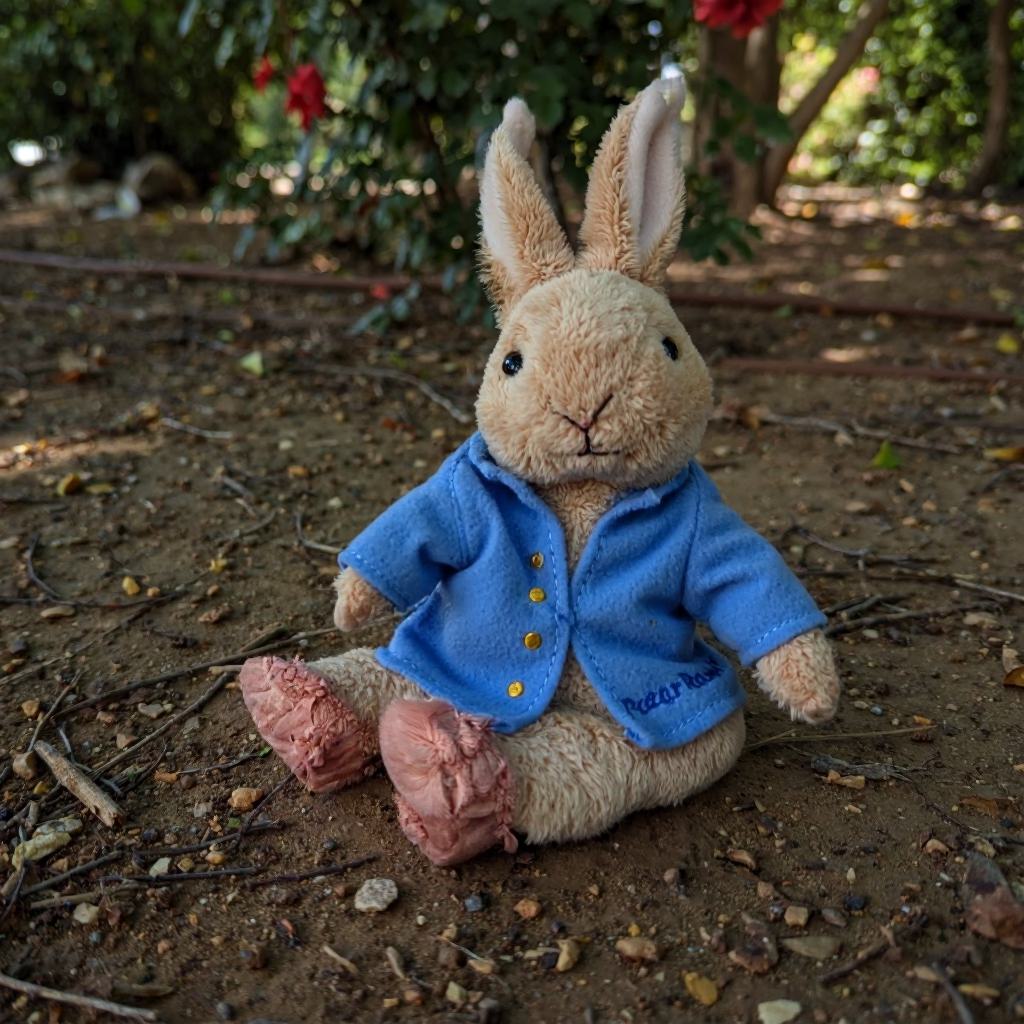}}
        \vspace{1px}
        \\

        &
        \multicolumn{6}{c}{\small{\prompt{A rabbit toy \textbf{sitting} and wearing \textbf{pink socks} during the \textbf{late afternoon}}}}
        \vspace{5px}
        \\

        {\includegraphics[valign=c, width=\ww]{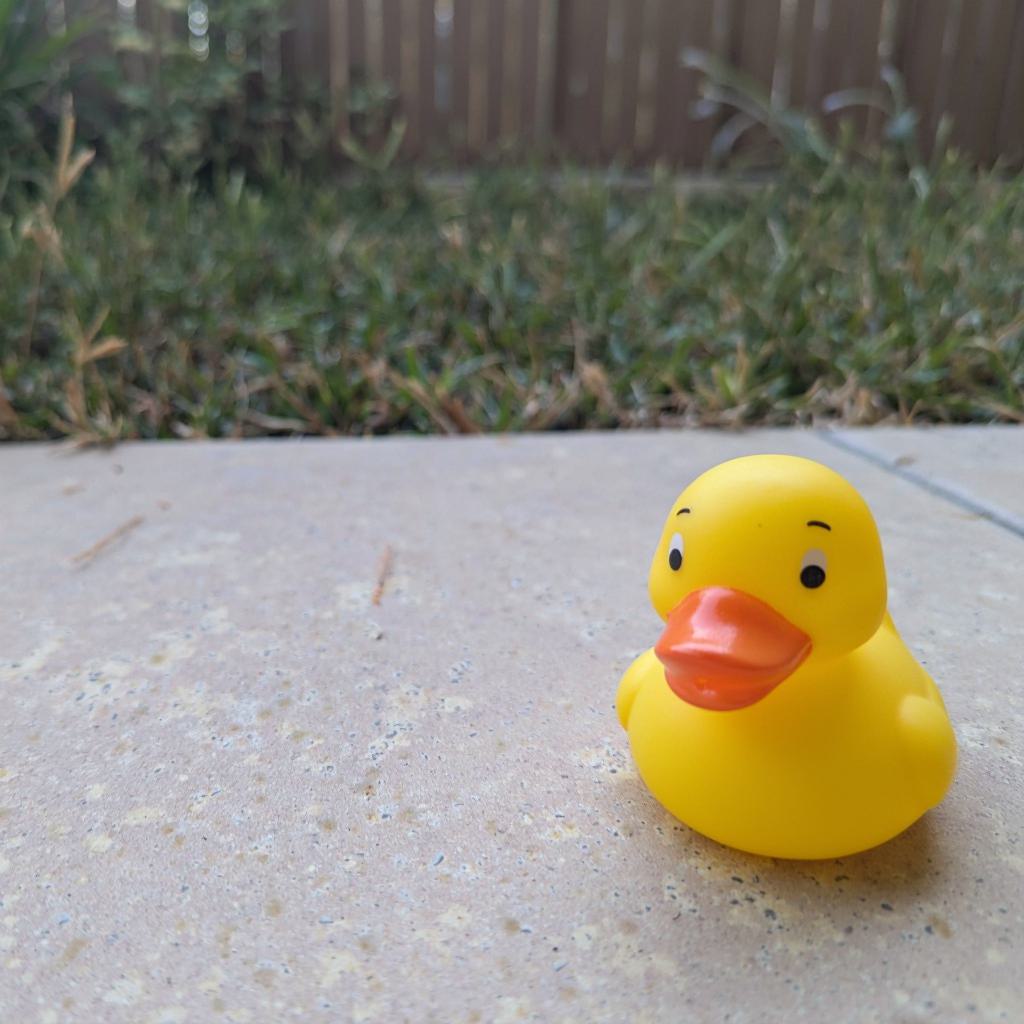}} &
        {\includegraphics[valign=c, width=\ww]{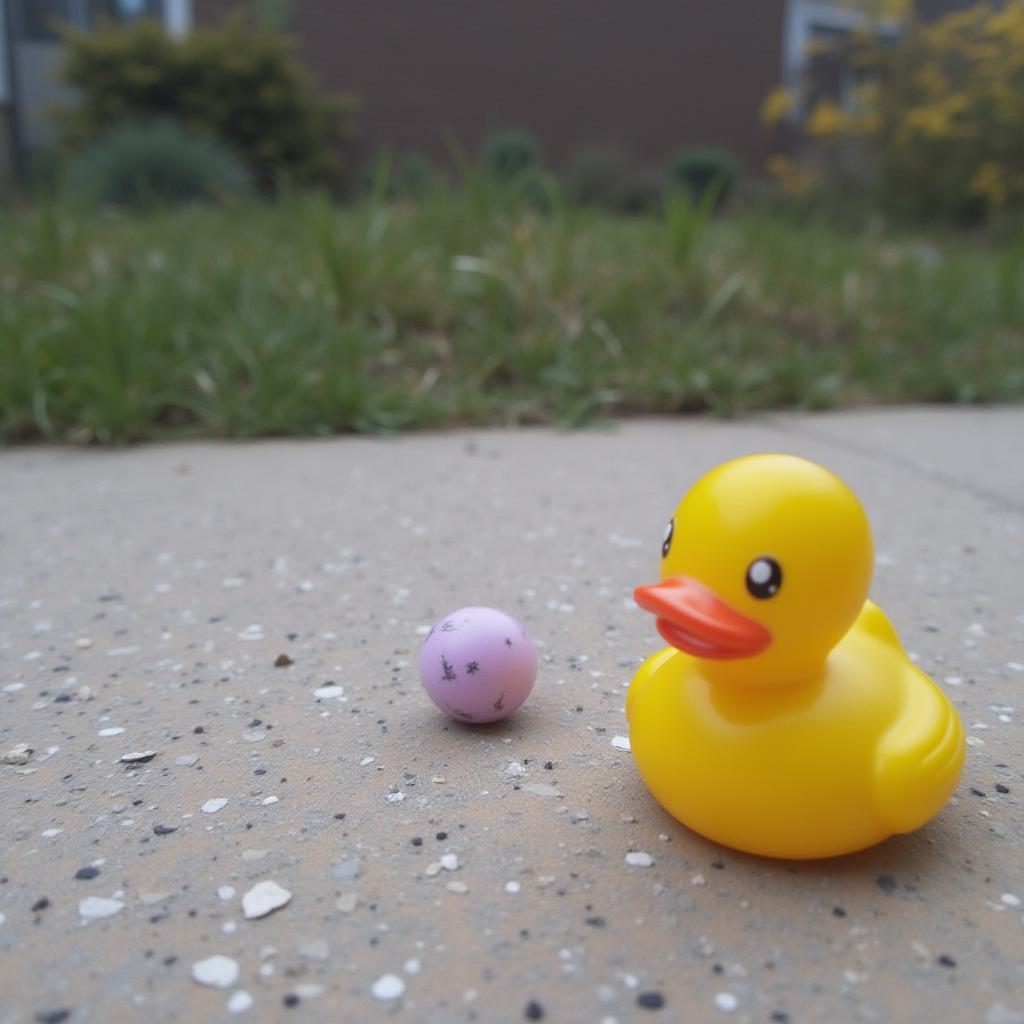}} &
        {\includegraphics[valign=c, width=\ww]{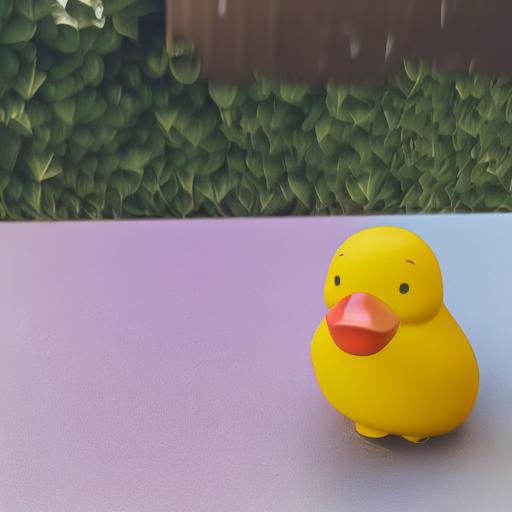}} &
        {\includegraphics[valign=c, width=\ww]{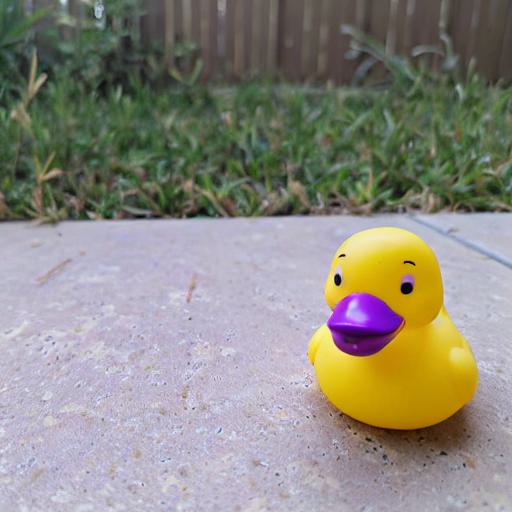}} &
        {\includegraphics[valign=c, width=\ww]{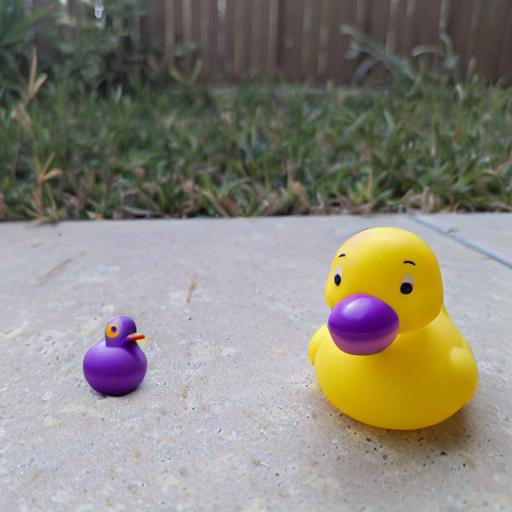}} &
        {\includegraphics[valign=c, width=\ww]{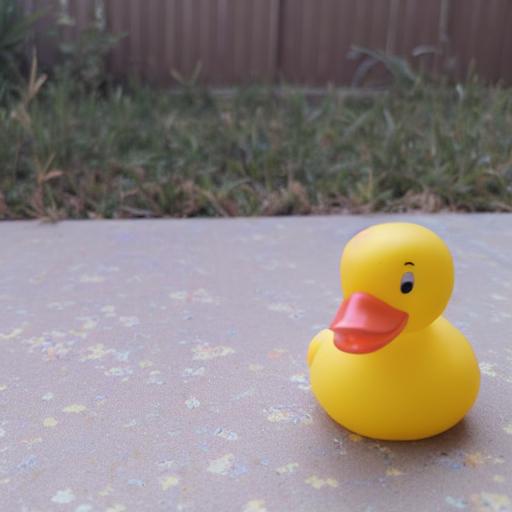}} &
        {\includegraphics[valign=c, width=\ww]{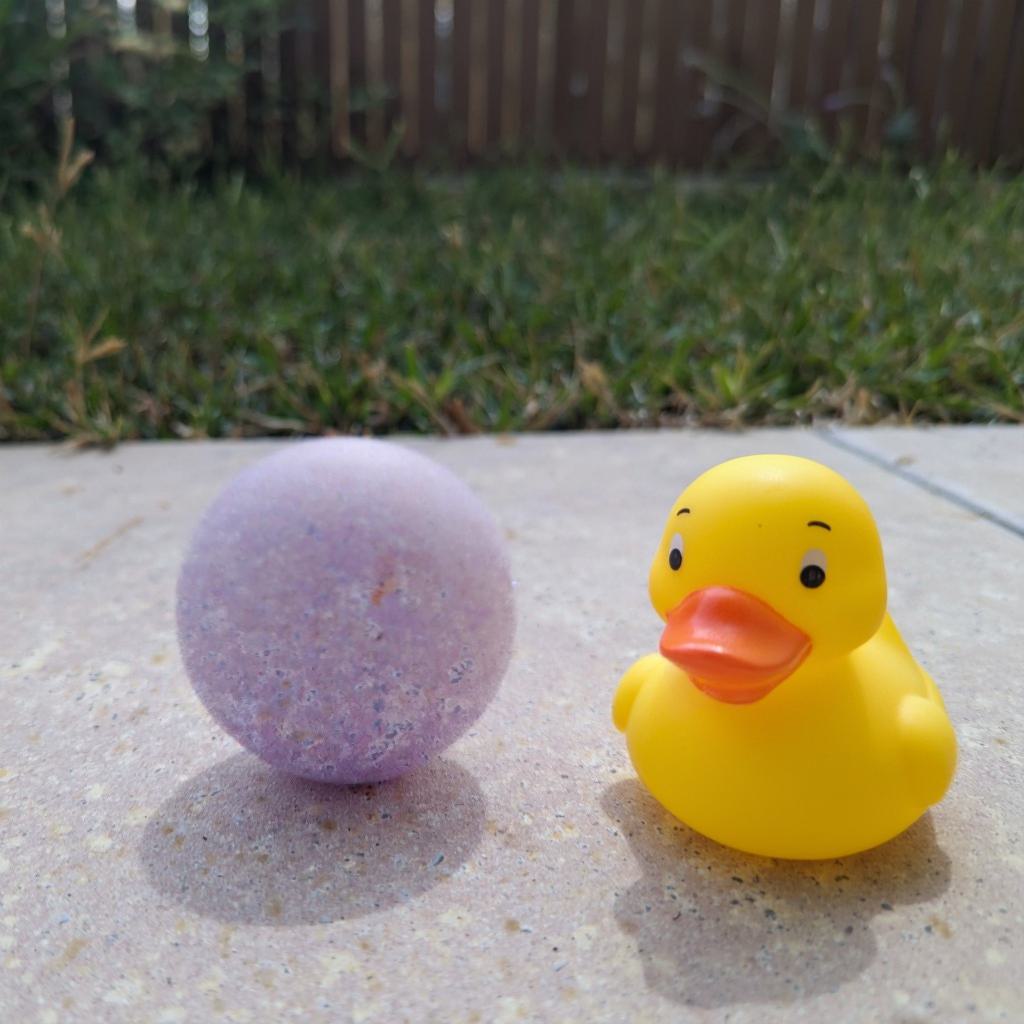}}
        \vspace{1px}
        \\

        &
        \multicolumn{6}{c}{\small{\prompt{A rubber duck next to a \textbf{purple ball} during a \textbf{sunny} day}}}
        \vspace{5px}
        \\

        {\includegraphics[valign=c, width=\ww]{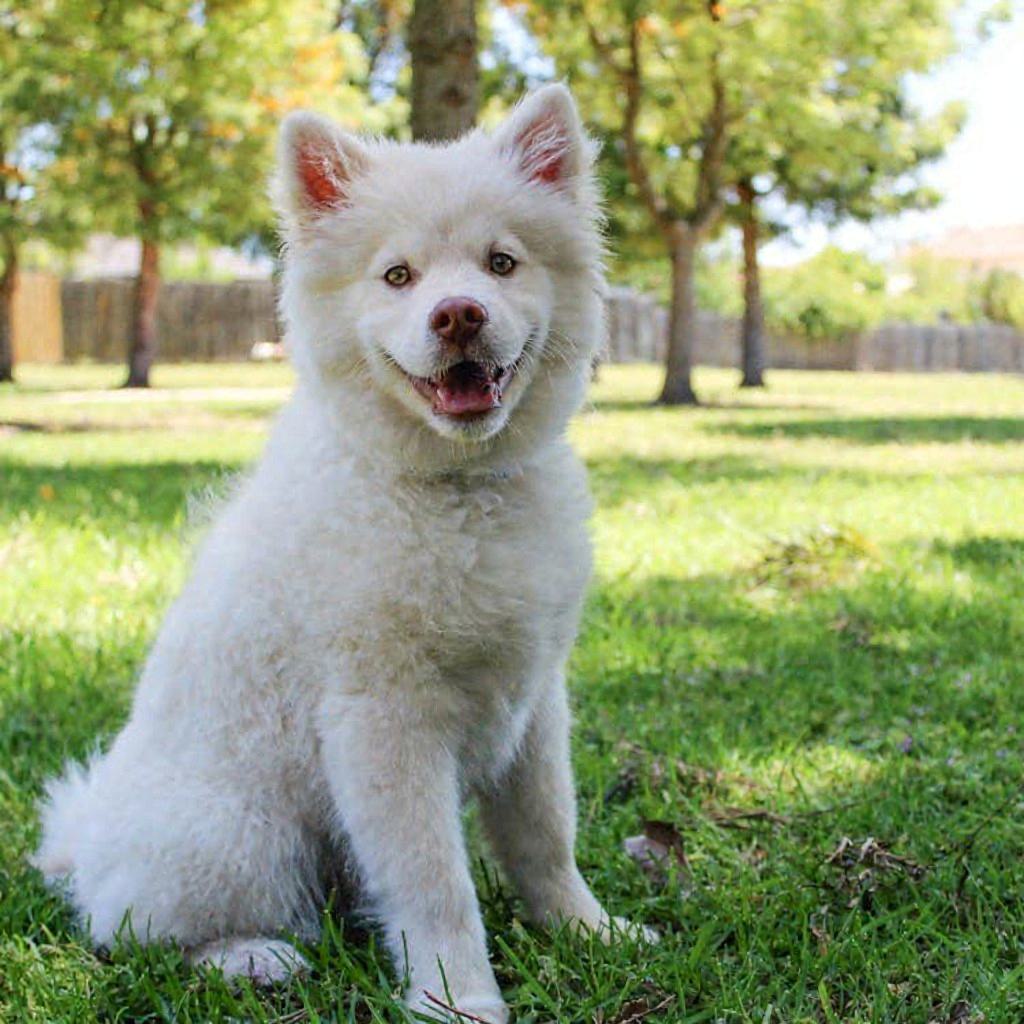}} &
        {\includegraphics[valign=c, width=\ww]{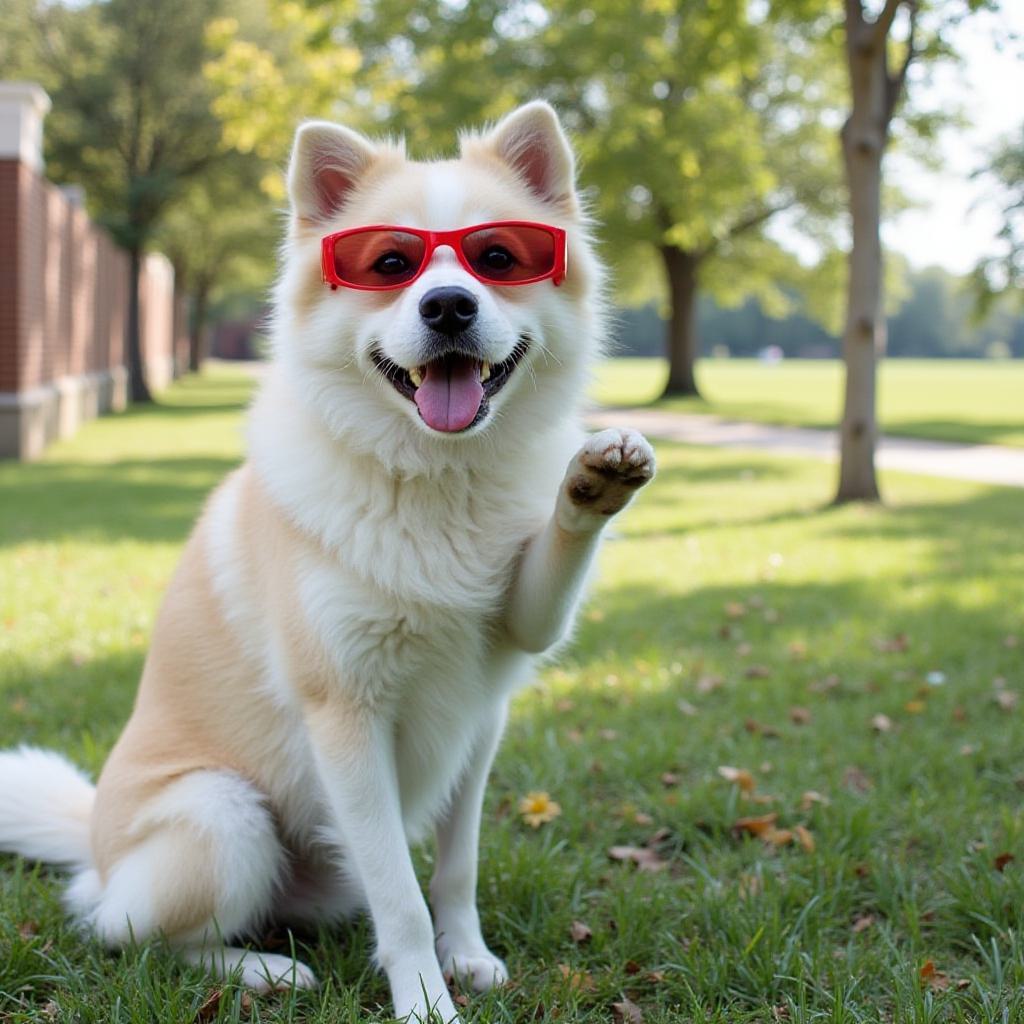}} &
        {\includegraphics[valign=c, width=\ww]{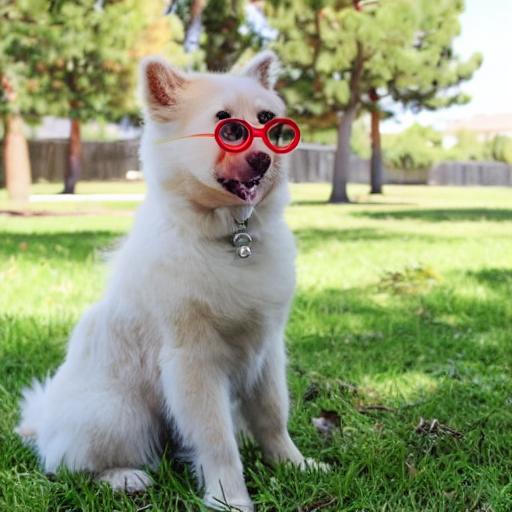}} &
        {\includegraphics[valign=c, width=\ww]{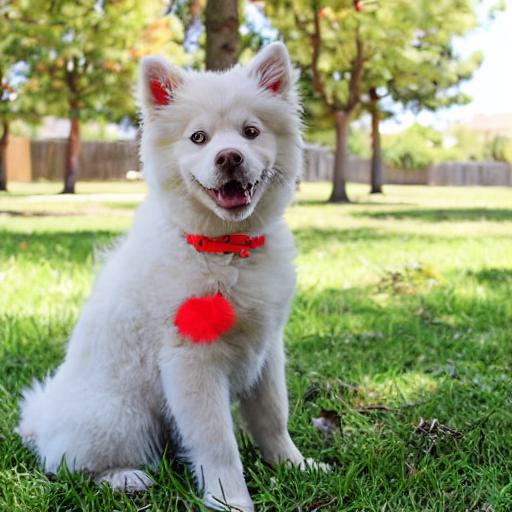}} &
        {\includegraphics[valign=c, width=\ww]{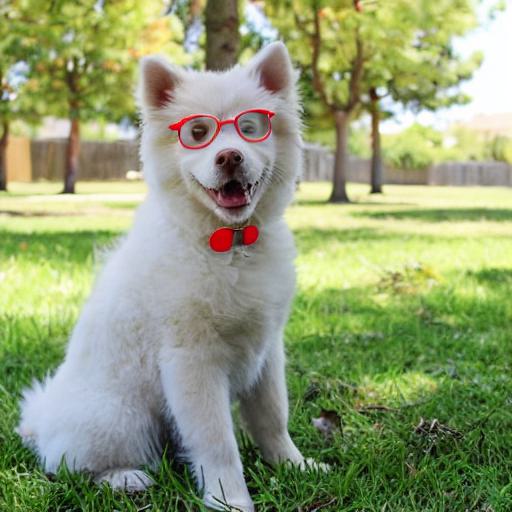}} &
        {\includegraphics[valign=c, width=\ww]{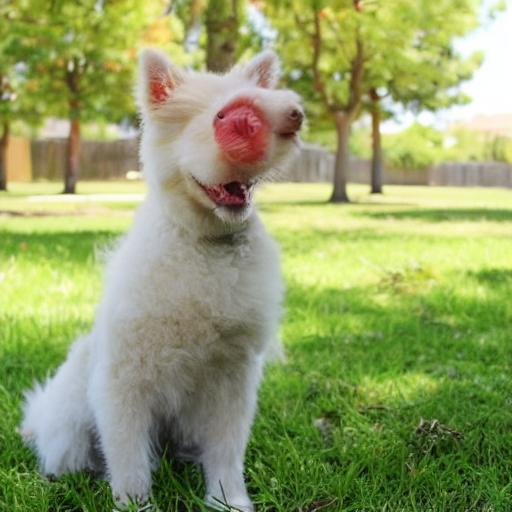}} &
        {\includegraphics[valign=c, width=\ww]{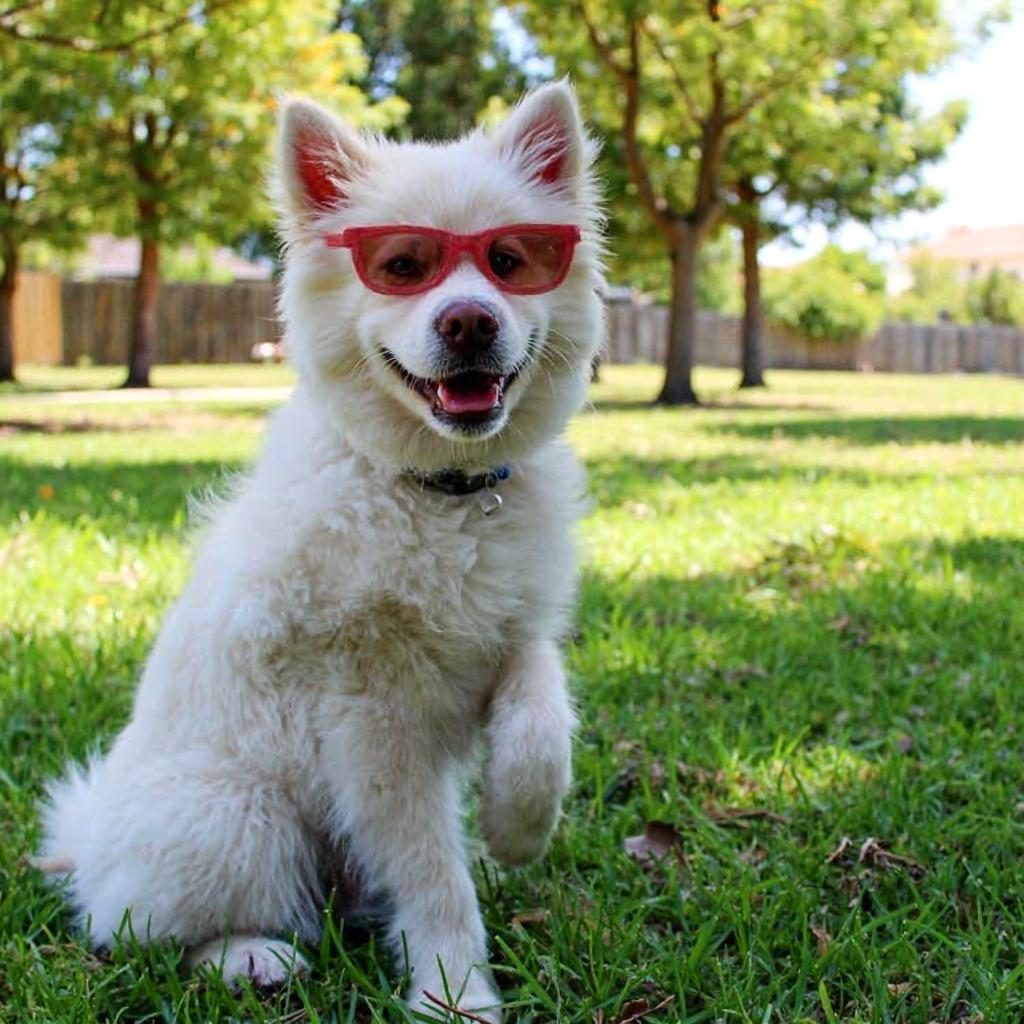}}
        \vspace{1px}
        \\

        &
        \multicolumn{6}{c}{\small{\prompt{A dog with a small collar \textbf{lifting} its paw while wearing \textbf{red glasses}}}}
        \vspace{5px}
        \\        

        {\includegraphics[valign=c, width=\ww]{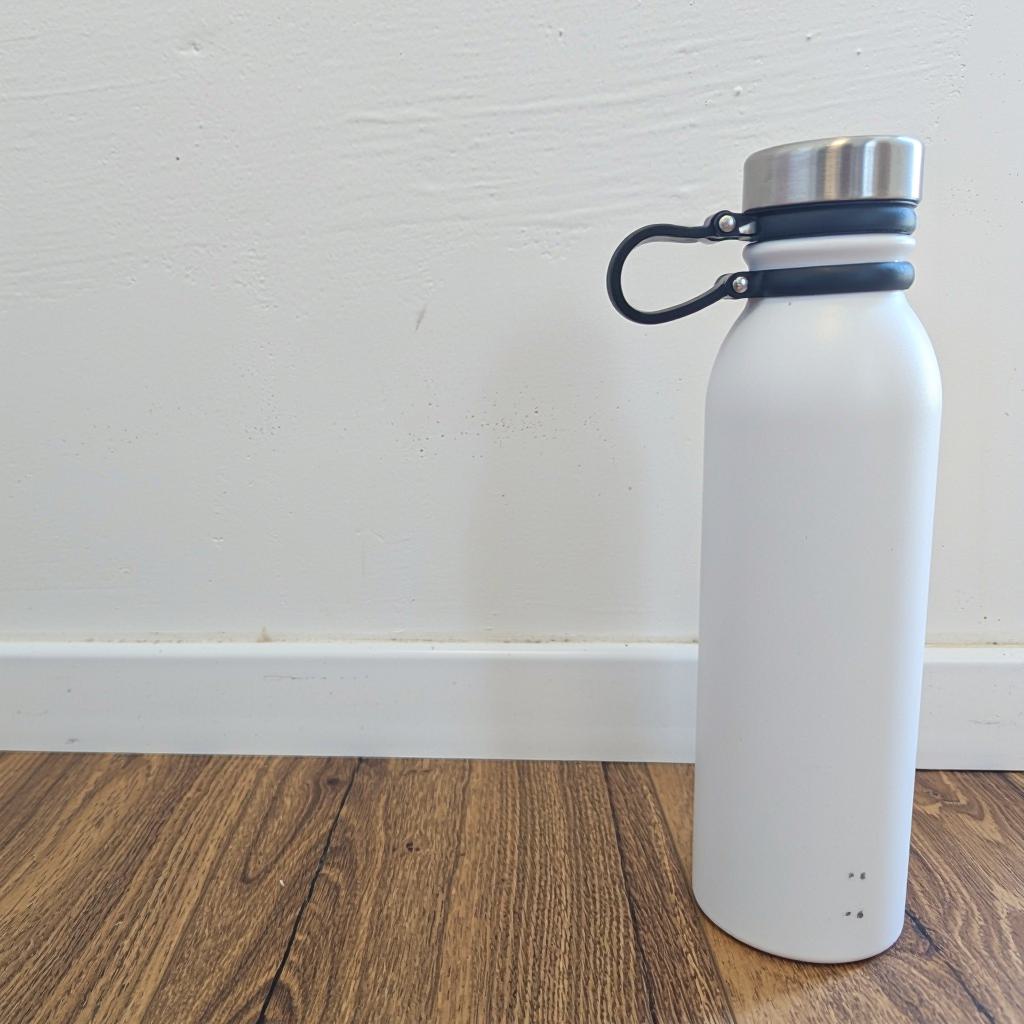}} &
        {\includegraphics[valign=c, width=\ww]{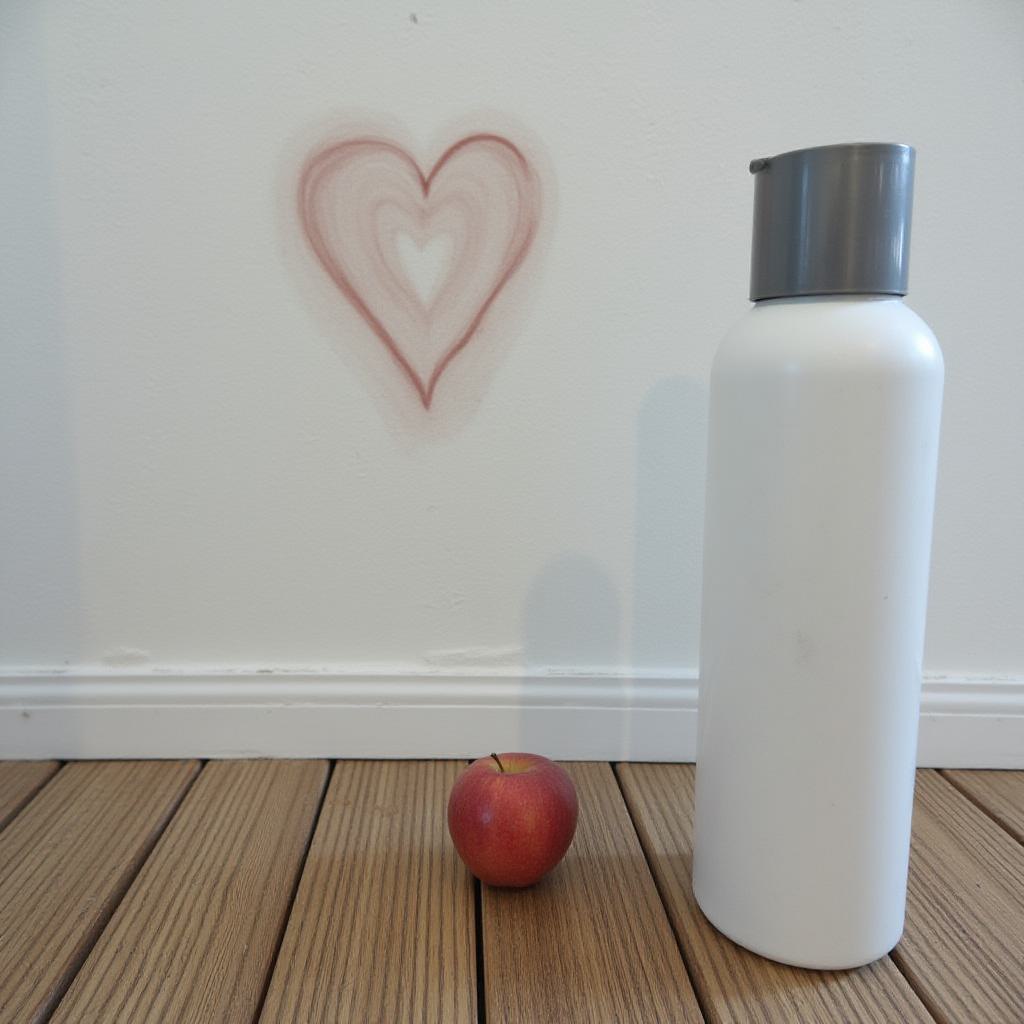}} &
        {\includegraphics[valign=c, width=\ww]{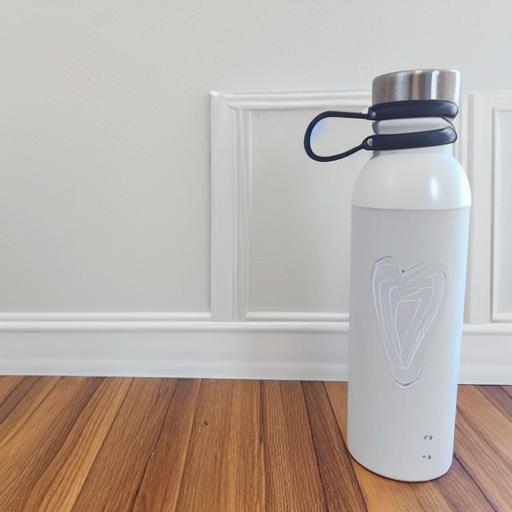}} &
        {\includegraphics[valign=c, width=\ww]{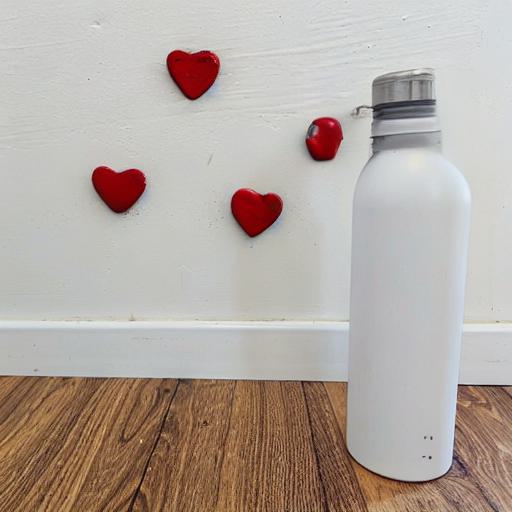}} &
        {\includegraphics[valign=c, width=\ww]{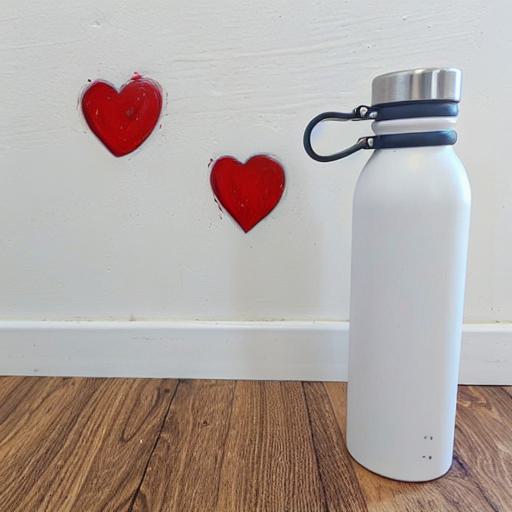}} &
        {\includegraphics[valign=c, width=\ww]{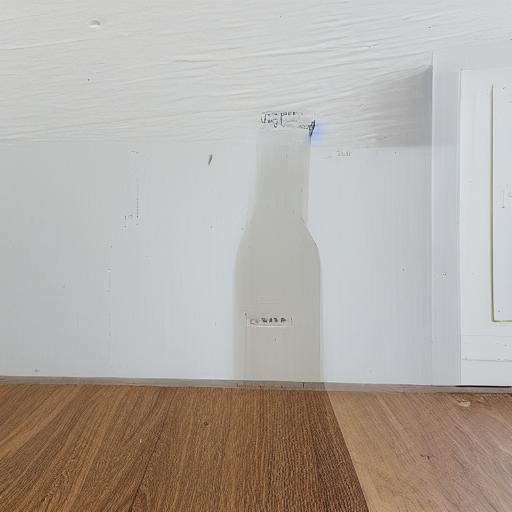}} &
        {\includegraphics[valign=c, width=\ww]{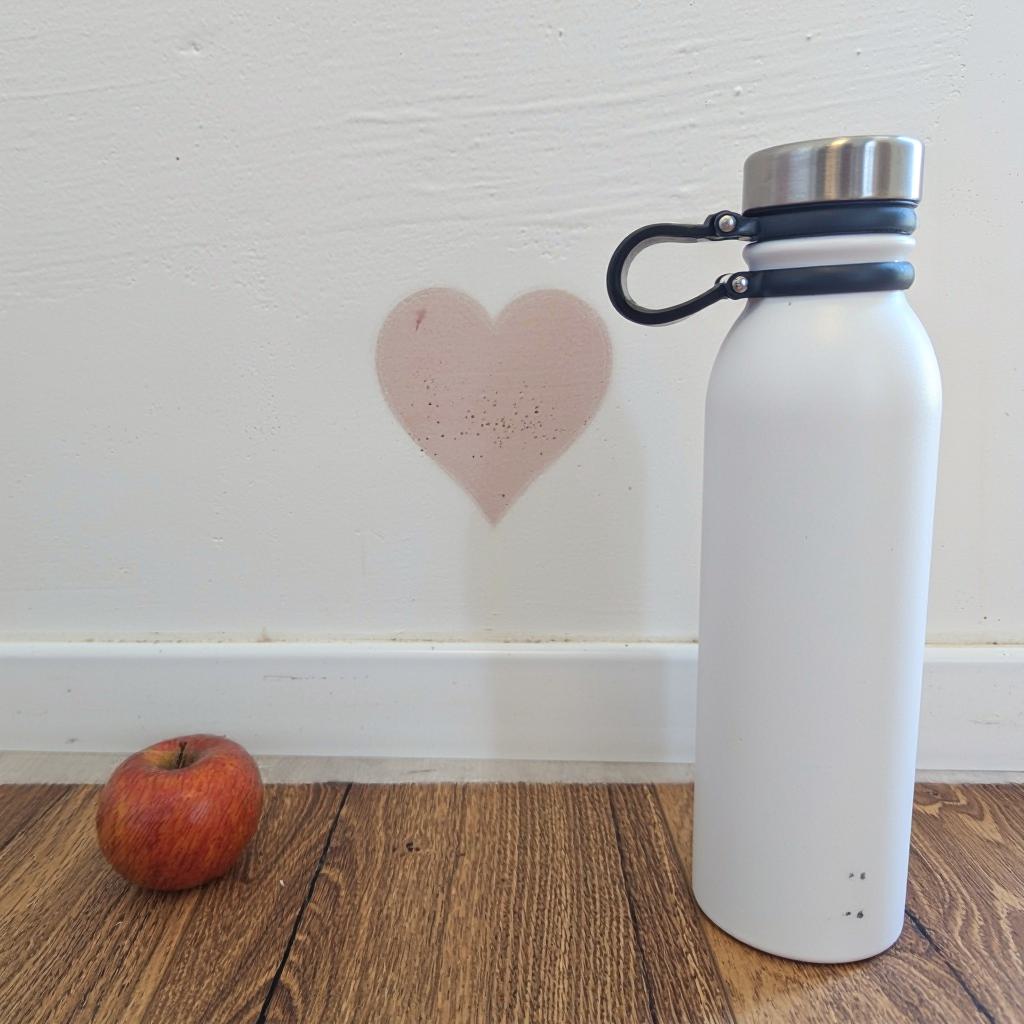}}
        \vspace{1px}
        \\

        &
        \multicolumn{6}{c}{\small{\prompt{A bottle next to an \textbf{apple}. There is a \textbf{heart} painting on the wall.}}}
        \vspace{5px}
        \\

        {\includegraphics[valign=c, width=\ww]{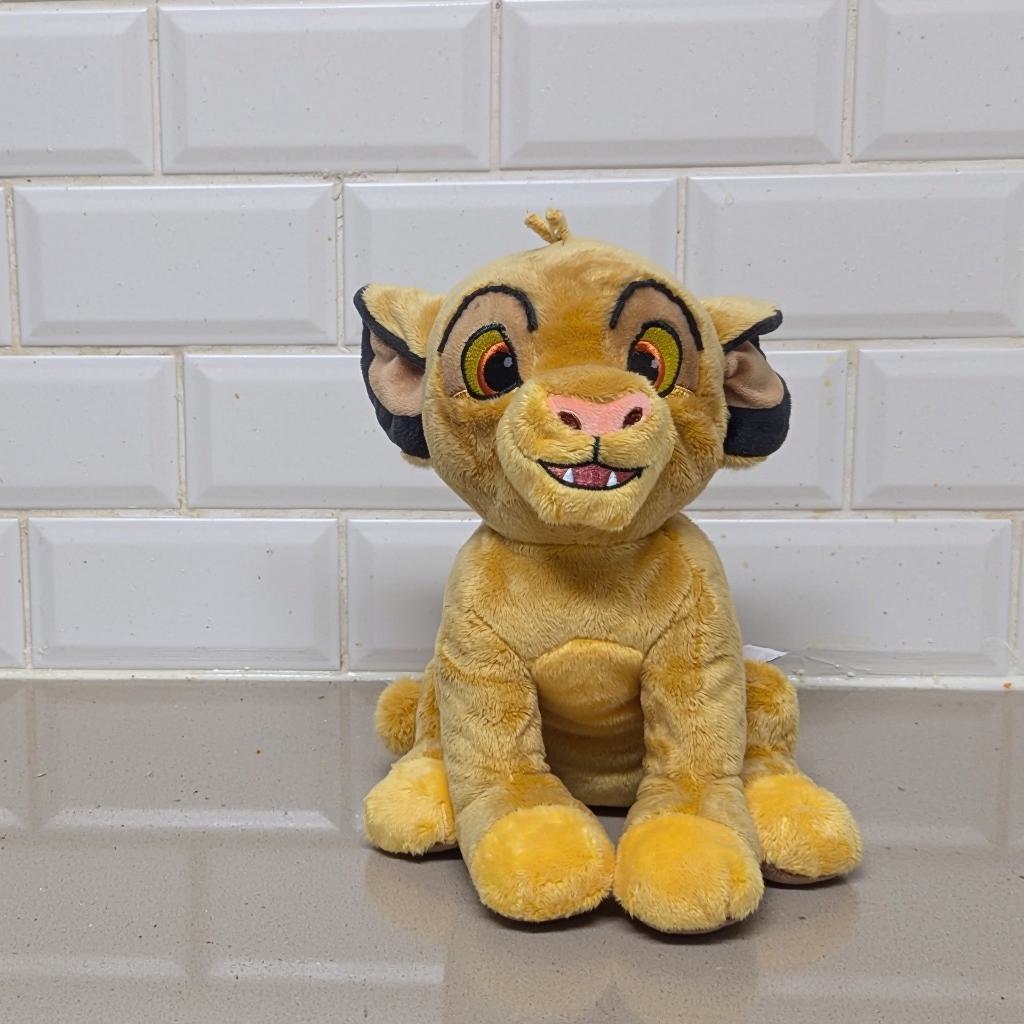}} &
        {\includegraphics[valign=c, width=\ww]{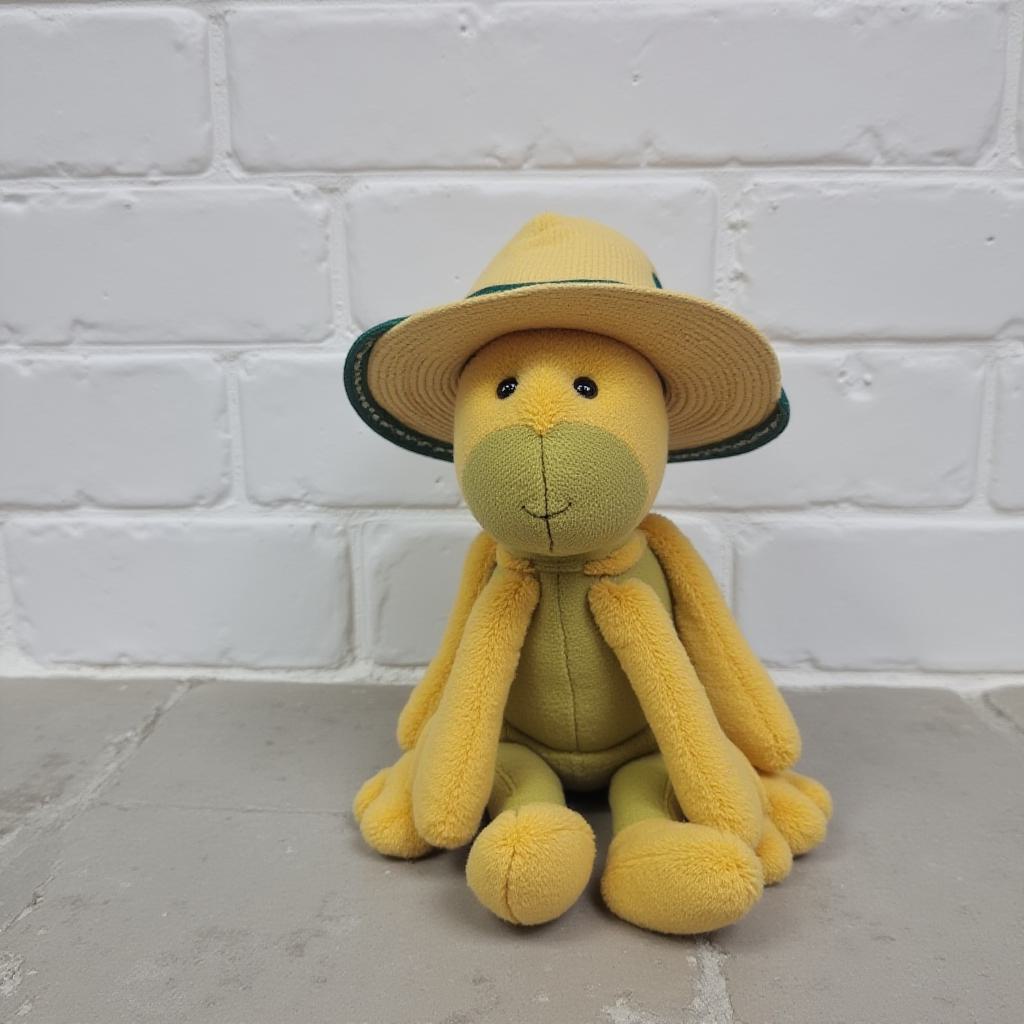}} &
        {\includegraphics[valign=c, width=\ww]{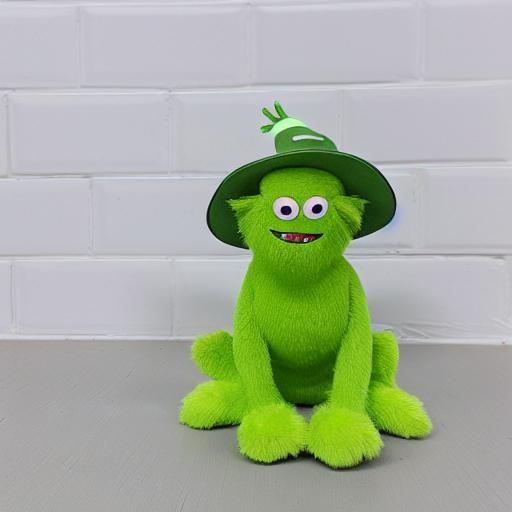}} &
        {\includegraphics[valign=c, width=\ww]{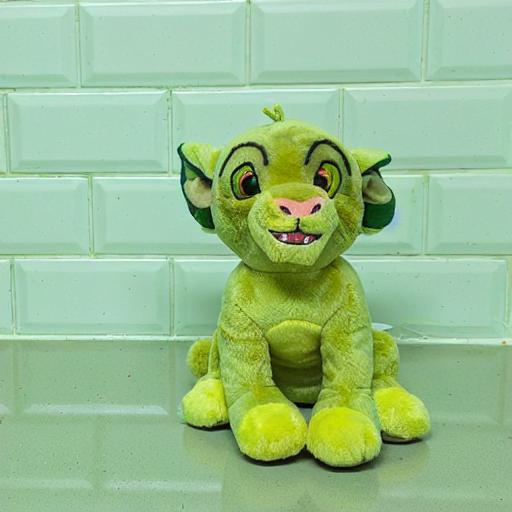}} &
        {\includegraphics[valign=c, width=\ww]{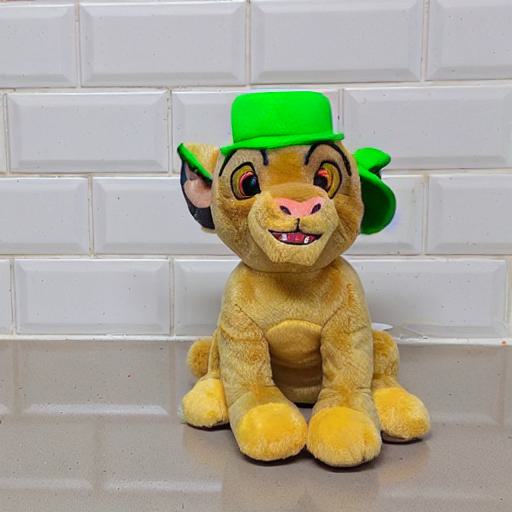}} &
        {\includegraphics[valign=c, width=\ww]{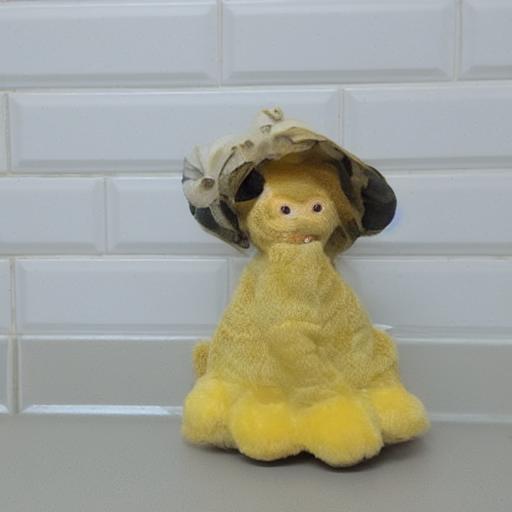}} &
        {\includegraphics[valign=c, width=\ww]{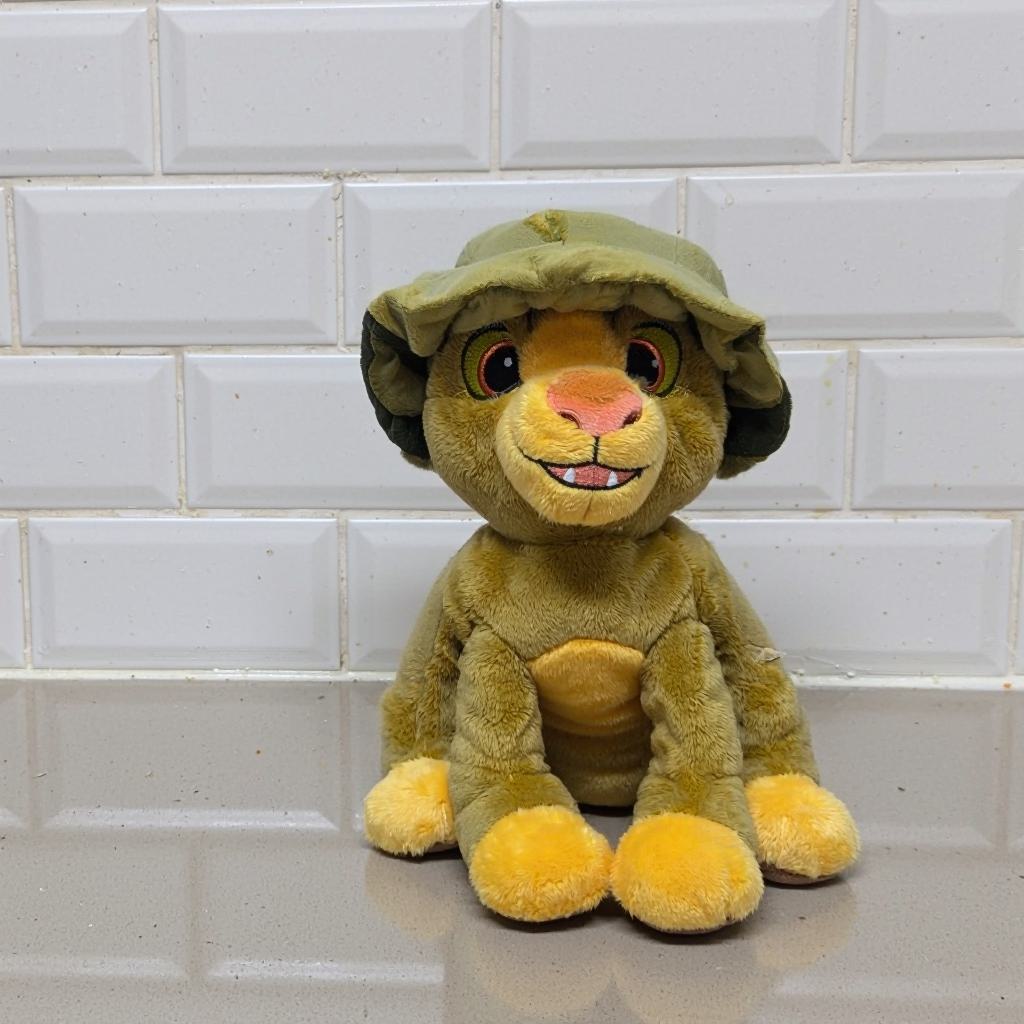}}
        \vspace{1px}
        \\

        &
        \multicolumn{6}{c}{\small{\prompt{A doll with a \textbf{green body} wearing a \textbf{hat}}}}
        \vspace{5px}
        \\

        {\includegraphics[valign=c, width=\ww]{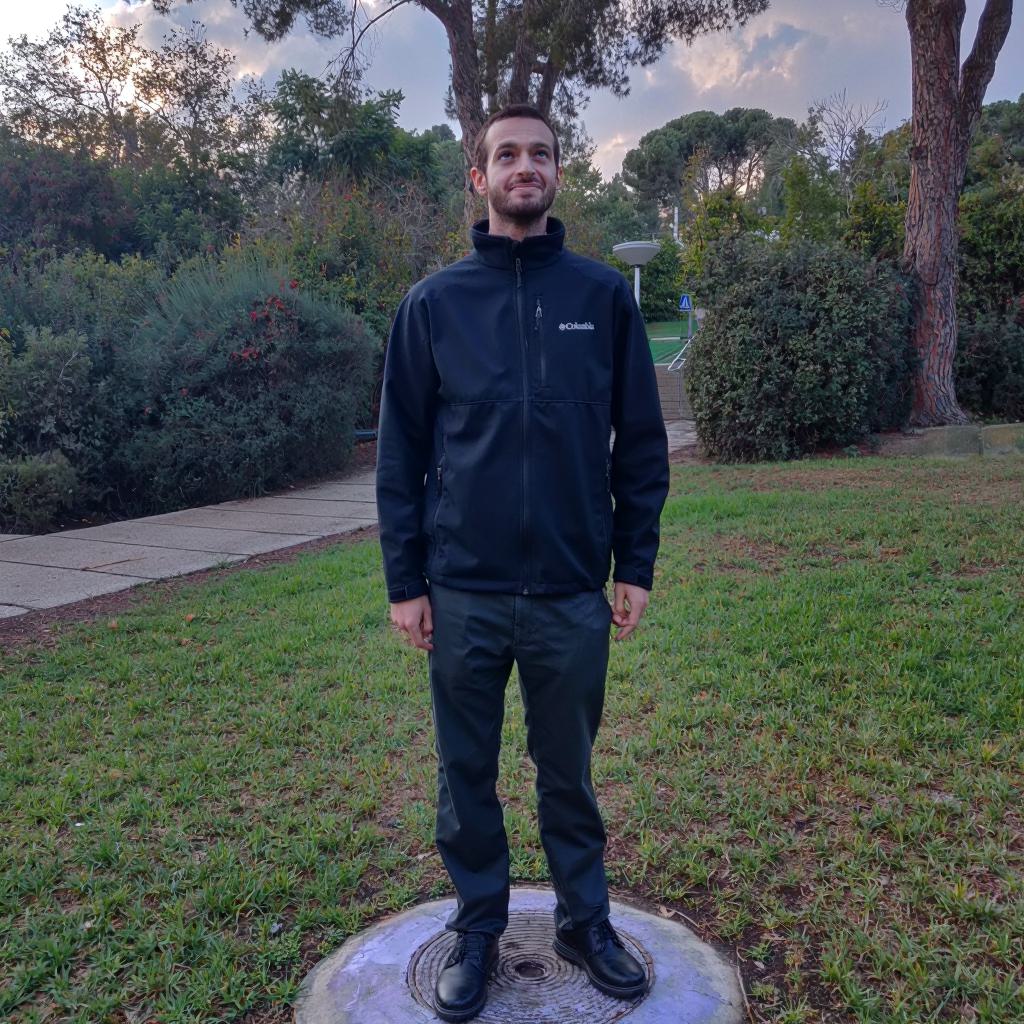}} &
        {\includegraphics[valign=c, width=\ww]{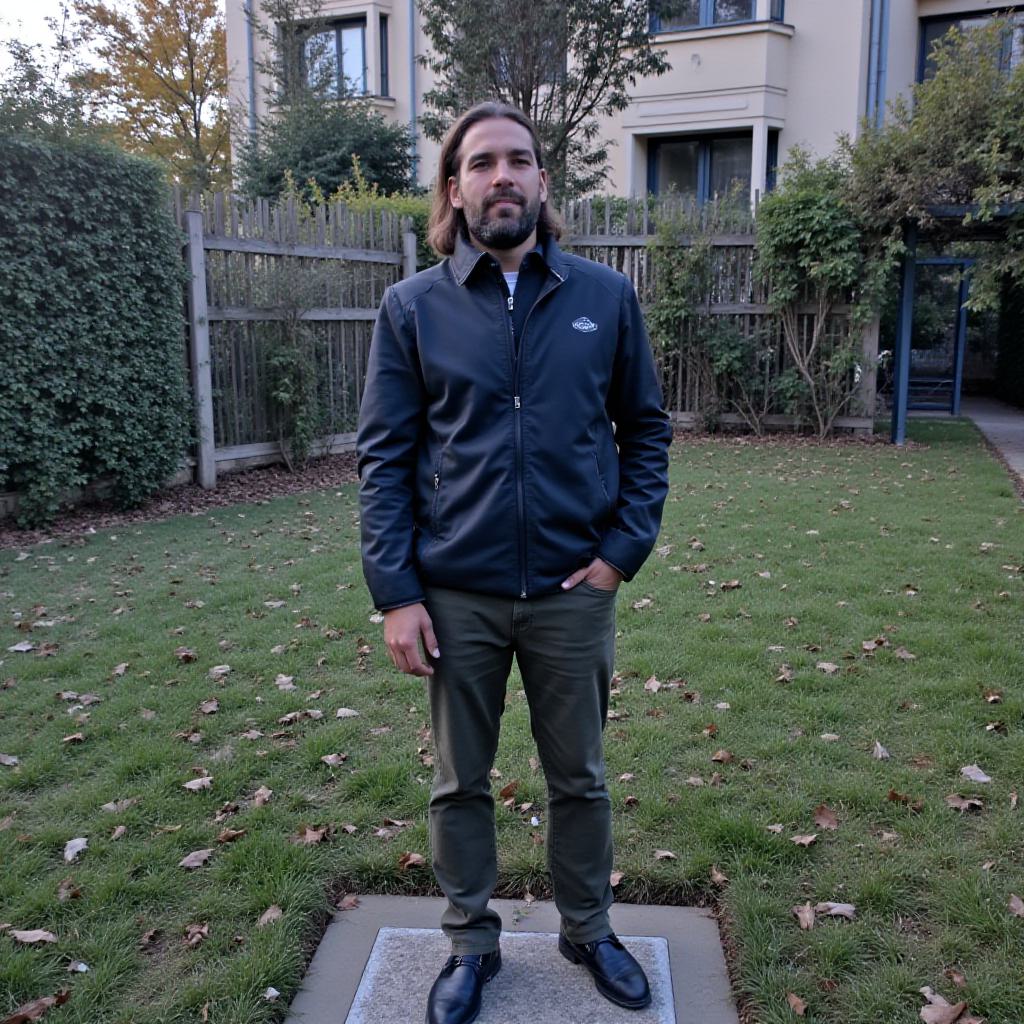}} &
        {\includegraphics[valign=c, width=\ww]{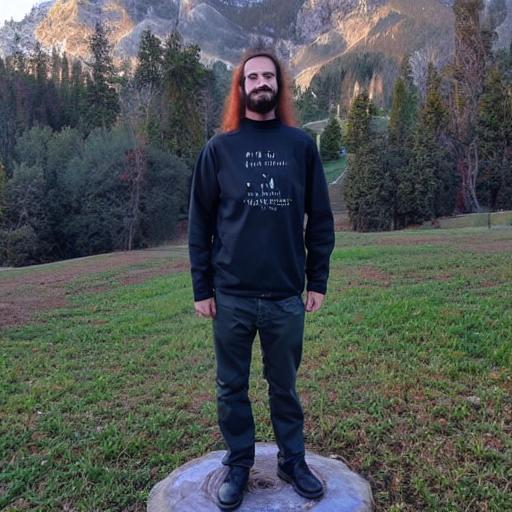}} &
        {\includegraphics[valign=c, width=\ww]{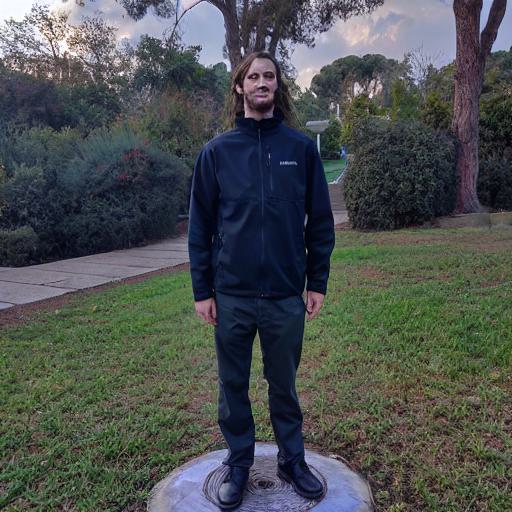}} &
        {\includegraphics[valign=c, width=\ww]{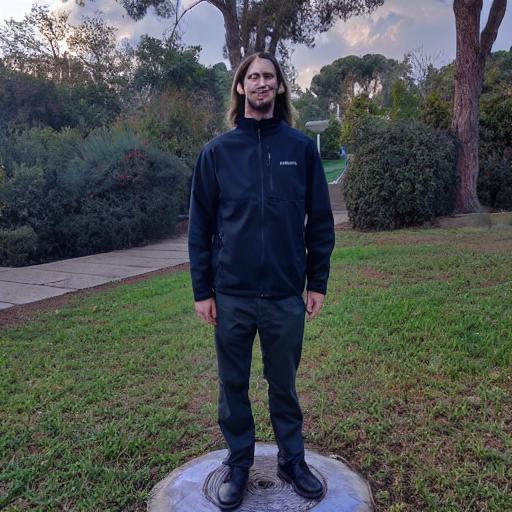}} &
        {\includegraphics[valign=c, width=\ww]{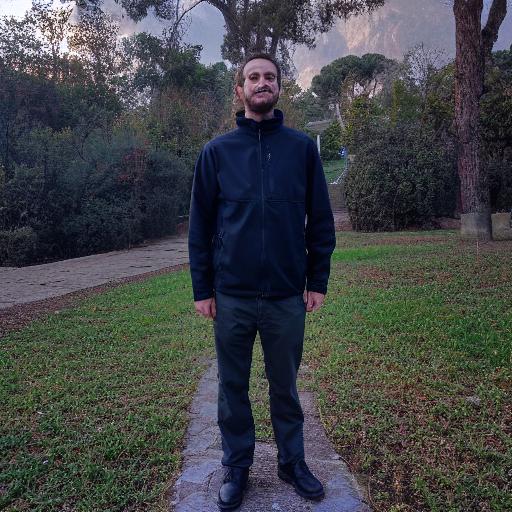}} &
        {\includegraphics[valign=c, width=\ww]{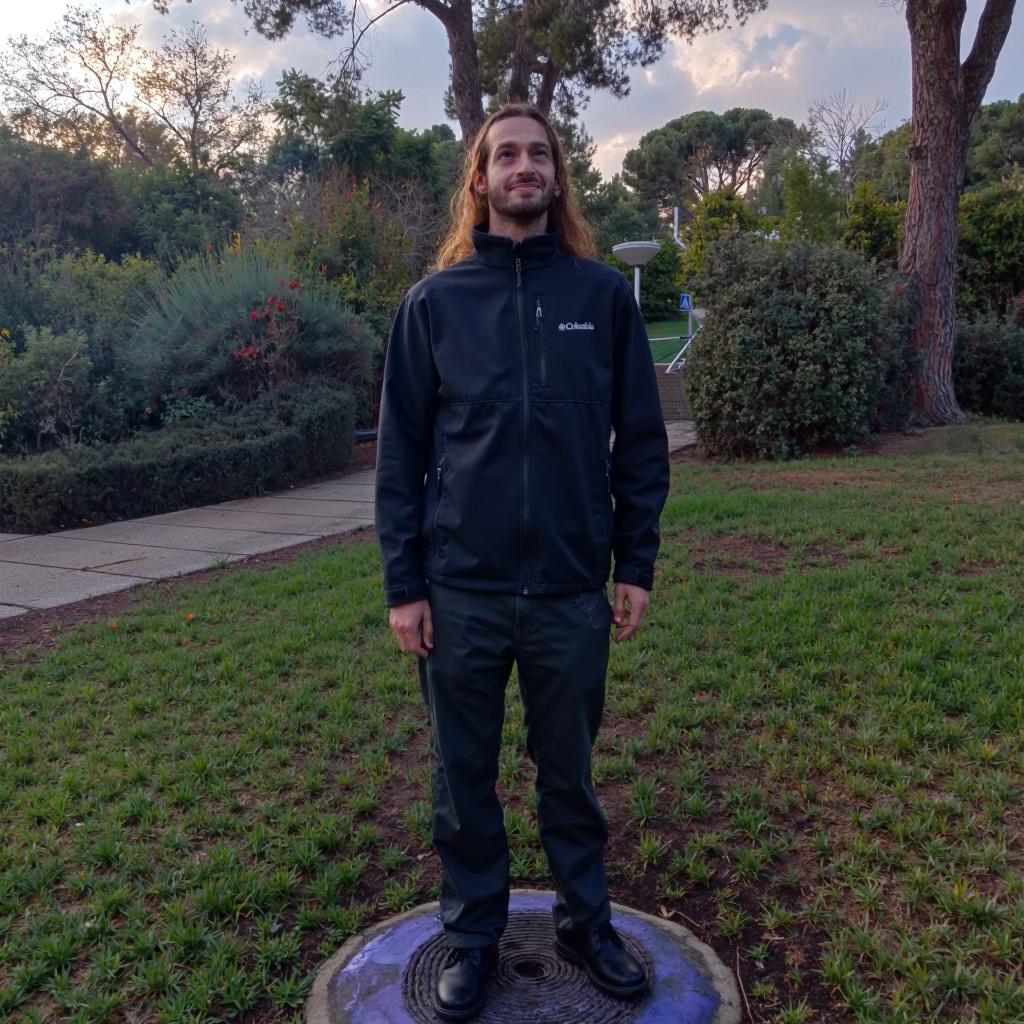}}
        \vspace{1px}
        \\

        &
        \multicolumn{6}{c}{\small{\prompt{A man with a \textbf{long hair}}}}
        \\

    \end{tabular}
    \vspace{-3px}
    \caption{\textbf{Qualitative Comparison.} We compare our method on real images against the baselines. SDEdit~\cite{meng2021sdedit} faces challenges with preserving object identities and backgrounds (\eg, rabbit and cat examples). P2P+NTI ~\cite{Hertz2022PrompttoPromptIE, mokady2022null} struggles with both preserving object identities (\eg, rabbit and lion dolls examples) and adding new objects (\eg, missing ball in the duck example and missing heart in the bottle example). Instruct-P2P~\cite{brooks2022instructpix2pix} and MagicBrush~\cite{Zhang2023MagicBrush} struggle with performing non-rigid editing (\eg, raising of the paws in dog and cat examples, and the sitting of the rabbit in its example). MasaCTRL~\cite{cao2023masactrl} has difficulty with preserving object identities (\eg, cat, dog and lion doll examples) and adding new objects (\eg, missing ball in the duck example and missing socks in the rabbit example). Our method, on the other hand, is able to adhere to the editing prompt while preserving the identities.}
    \label{fig:qualitative_comparison}
\end{figure*}

\begin{table}
    \centering
    \caption{\textbf{Quantitative Comparison.} We compare our method against the baselines in terms of text similarity (\cliptxt), image similarity (\clipimg) and image-text direction similarity (\clipdir). As can be seen, P2P+NTI~\cite{Hertz2022PrompttoPromptIE, mokady2022null}, Instruct-P2P~\cite{brooks2022instructpix2pix}, and MasaCTRL~\cite{cao2023masactrl} suffer from low similarity to the text prompt. SDEdit~\cite{Yang2022PaintBE} and MagicBrush~\cite{Zhang2023MagicBrush} adhere more to the text prompt, but they struggle with image similarity and image-text direction similarity. Our method, on the other hand, achieves better image and image-text direction similarity.}
    \begin{adjustbox}{width=1\columnwidth}
        \begin{tabular}{>{\columncolor[gray]{0.95}}lccc}
            \toprule

            \textbf{Method} & 
            \cliptxt $(\uparrow)$ &
            \clipimg $(\uparrow)$ &
            \clipdir $(\uparrow)$
            \\
            
            \midrule

            SDEdit~\cite{Yang2022PaintBE} &
            \textbf{0.24} &
            0.71 &
            0.07
            \\

            P2P+NTI~\cite{Hertz2022PrompttoPromptIE, mokady2022null} &
            0.21 &
            0.76 &
            0.08
            \\

            Instruct-P2P~\cite{brooks2022instructpix2pix} &
            0.22 &
            0.87 &
            0.07
            \\

            MagicBrush~\cite{Zhang2023MagicBrush} &
            \textbf{0.24} &
            0.88 &
            0.11
            \\

            MasaCTRL~\cite{cao2023masactrl} &
            0.20 &
            0.76 &
            0.03
            \\

            \midrule

            Stable Flow (ours) &
            0.23 &
            \textbf{0.92} &
            \textbf{0.14}
            \\
            
            \bottomrule
        \end{tabular}
    \end{adjustbox}
    \label{tab:quantitative_comparison}
\end{table}

\begin{table}
    \centering
    \caption{\textbf{Ablation Study.} We conduct an ablation study and find that performing attention injection in all the layers or performing an attention extension in all the layers significantly reduces the text similarity. Furthermore, performing an attention injection in the non-vital layers or removing the latent nudging reduces the image similarity.}
    \begin{adjustbox}{width=1\columnwidth}
        \begin{tabular}{>{\columncolor[gray]{0.95}}lccc}
            \toprule

            \textbf{Method} & 
            \cliptxt $(\uparrow)$ &
            \clipimg $(\uparrow)$ &
            \clipdir $(\uparrow)$
            \\
            
            \midrule

            Stable Flow (ours) &
            0.23 &
            0.92 &
            \textbf{0.14}
            \\

            \midrule

            Injection all layers &
            0.17 &
            \textbf{0.98} &
            0.00
            \\

            Injection non-vital layers &
            \textbf{0.25} &
            0.72 &
            0.09
            \\

            Extension all layers &
            0.18 &
            \textbf{0.98} &
            0.01
            \\

            w/o latent nudging &
            0.22 &
            0.62 &
            0.05
            \\
            
            \bottomrule
        \end{tabular}
    \end{adjustbox}
    \label{tab:ablation_study}
\end{table}

\begin{table}
    \centering
    \caption{\textbf{User Study.} We compare our method against the baselines using the standard two-alternative forced-choice format. Users were asked to rate which editing result is better (Ours vs. the baseline) in terms of: (1) target prompt adherence, (2) input image preservation, (3) realism and (4) overall edit quality. We report the win rate of our method compared to each baseline. As shown, our method outperforms the baselines across all categories, achieving a win rate higher than the random chance of 50\%.}
    \begin{adjustbox}{width=1\columnwidth}
        \begin{tabular}{>{\columncolor[gray]{0.95}}lcccc}
            \toprule

            \textbf{Ours vs} & 
            Prompt Adher. $(\uparrow)$ &
            Image Pres. $(\uparrow)$ &
            Realism $(\uparrow)$ &
            Overall $(\uparrow)$
            \\
            
            \midrule

            SDEdit~\cite{meng2021sdedit} &
            69.00\% &
            68.00\% &
            63.66\% &
            70.66\%
            \\

            P2P+NTI~\cite{Hertz2022PrompttoPromptIE, mokady2022null} &
            76.00\% &
            71.00\% &
            72.66\% &
            65.33\%
            \\

            Instruct-P2P~\cite{brooks2022instructpix2pix} &
            76.33\% &
            75.66\% &
            68.00\% &
            60.33\%
            \\

            MagicBrush~\cite{Zhang2023MagicBrush} &
            61.33\% &
            67.33\% &
            76.66\% &
            74.00\%
            \\

            MasaCTRL~\cite{cao2023masactrl} &
            82.33\% &
            80.00\% &
            80.33\% &
            72.00\%
            \\
            
            \bottomrule
        \end{tabular}
    \end{adjustbox}
    \label{tab:user_study}
    \vspace{-10px}
\end{table}

We compare our method against the most relevant text-driven image editing methods. We re-implement SDEdit~\cite{meng2021sdedit} using the FLUX.1-dev~\cite{flux} model, and use the official public implementations of P2P+NTI~\cite{Hertz2022PrompttoPromptIE, mokady2022null}, Instruct-P2P~\cite{brooks2022instructpix2pix}, MagicBrush~\cite{Zhang2023MagicBrush} and MasaCTRL~\cite{cao2023masactrl}. See the supplementary material for more details.

In \Cref{fig:qualitative_comparison} we compare our method against the baselines qualitatively on real images. As can be seen, SDEdit~\cite{meng2021sdedit} has difficulty maintaining object identities and backgrounds. P2P+NTI ~\cite{Hertz2022PrompttoPromptIE, mokady2022null} struggles with preserving object identities and with adding new objects. Instruct-P2P~\cite{brooks2022instructpix2pix} and MagicBrush~\cite{Zhang2023MagicBrush} face challenges with non-rigid editing. MasaCTRL~\cite{cao2023masactrl} struggles with preserving object identities and adding new objects. Our method, on the other hand, is adhering to the editing prompt while preserving the identities.

To quantify the performance of our method and the baselines, we prepared an evaluation dataset based on COCO~\cite{Lin2014MicrosoftCC}, that in contrast to previous benchmarks~\cite{Sheynin2023EmuEP, Zhang2023MagicBrush}, also contains non-rigid editing tasks. We start by filtering the dataset automatically to contain at least one prominent non-rigid body. Next, for each image, we use various image editing tasks (non-rigid editing, object addition, object replacement and scene editing) that take into account the prominent object, resulting in a total dataset of 3,200 samples. See the supplementary material for more details.

We evaluated the editing results using three metrics: (1) \clipimg that measures the similarity between the input image and the edited image, (2) \cliptxt that measure the similarity between the edited image and the target editing prompt, and (3) \clipdir~\cite{Patashnik2021StyleCLIPTM, Gal2021StyleGANNADA} that measures the similarity between the direction of the prompt change and the direction of the image change.

As can be seen in \Cref{tab:quantitative_comparison}, P2P+NTI~\cite{Hertz2022PrompttoPromptIE, mokady2022null}, Instruct-P2P~\cite{brooks2022instructpix2pix}, and MasaCTRL~\cite{cao2023masactrl} suffer from low similarity to the text prompt. SDEdit~\cite{Yang2022PaintBE} and MagicBrush~\cite{Zhang2023MagicBrush} adhere more to the text prompt, but they struggle with image similarity and image-text direction similarity. Our method, on the other hand, is able to achieve better image and image-text direction similarity.

\subsection{User Study}
\label{sec:user_study}
We conduct an extensive user study using the Amazon Mechanical Turk (AMT)~\cite{amt} platform, with the automatically generated test examples from \Cref{sec:comparisons}. We compare all the baselines against our method using standard two-alternative forced-choice format. Users were given the input image, the edit text prompt, and two editing results (one from our method and one from the baseline). For each comparison, the users were asked to rate which editing result is better in terms of: (1) target prompt adherence, (2) input image preservation, (3) realism and (4) overall edit quality (\ie, when taking all factors into account). As can be seen in \Cref{tab:user_study}, our method is preferred over the baselines in overall terms as well as the other terms. See the supplementary material for more details and statistical analysis.

\subsection{Ablation Study}
\label{sec:ablation_study}

We conduct an ablation study for the following cases: (1) \textit{Attention injection in all layers} --- we perform the attention injection that is described in \Cref{sec:image_editing} in all the layers (instead on the vital layers only). (2) \textit{Attention injection non-vital layers} --- we performed the attention injection in some non-vital layers (same amount of layers as vital layers). (3) \textit{Attention extension} --- instead of performing attention injection as described in \Cref{sec:image_editing}, we extended~\cite{Hertz2023StyleAI, Tewel2024TrainingFreeCT} the attention s.t. the generated images can attend to the reference image, as well as themselves. (4) \textit{w/o latent nudging} --- we omitted the latent nudging component (\Cref{sec:latent_nudging}).

As can be seen in \Cref{tab:ablation_study}, we found that (1) performing attention injection in all the layers or performing (3) an attention extension in all the layers, significantly harms the text similarity. In addition, (2) performing an attention extension in the non-vital layers or (4) removing the latent nudging reduces the image similarity.

\subsection{Applications}
\label{sec:applications}

\begin{figure}[tp]
    \centering
    \setlength{\tabcolsep}{0.6pt}
    \renewcommand{\arraystretch}{0.7}
    \setlength{\ww}{0.235\columnwidth}
    \begin{tabular}{cc @{\hspace{7\tabcolsep}} ccc}
        \rotatebox[origin=c]{90}{\scriptsize{(1) Inc. Editing}} &
        {\includegraphics[valign=c, width=\ww]{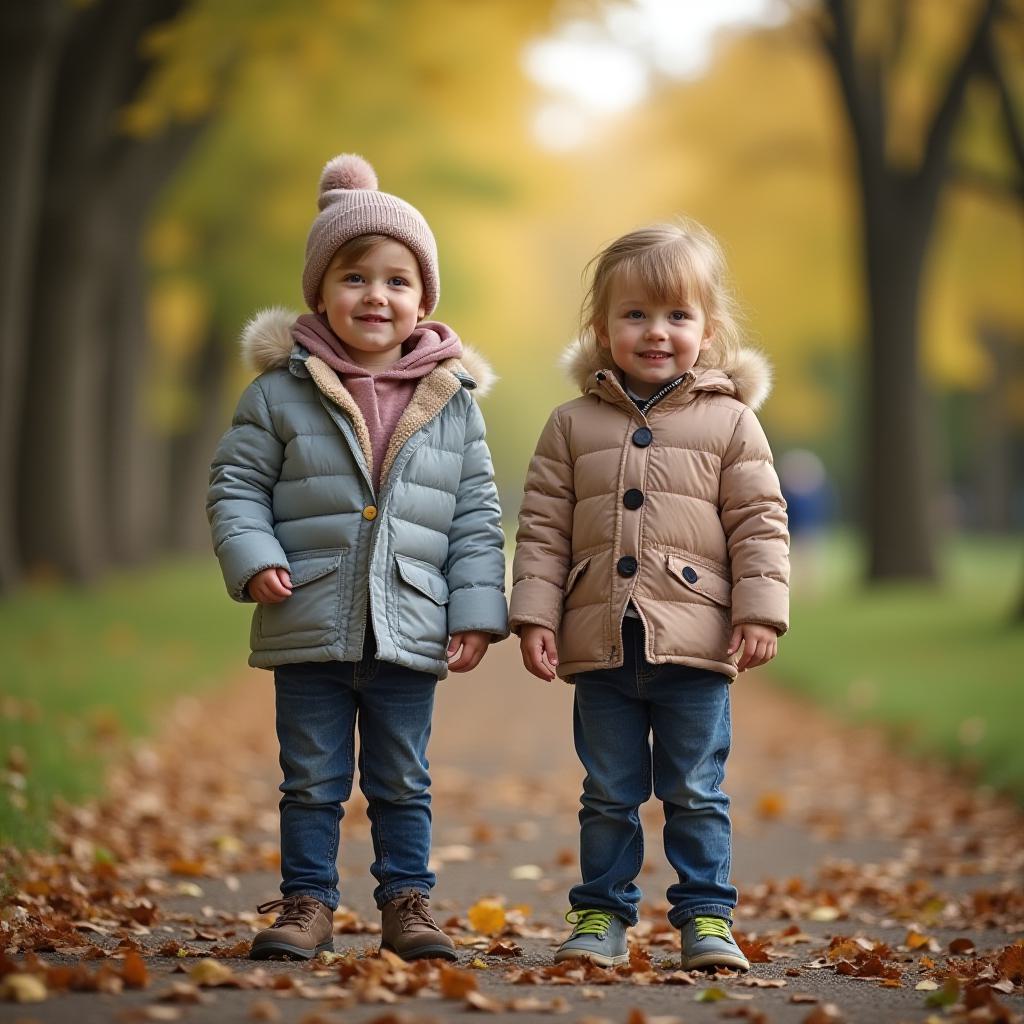}} &
        {\includegraphics[valign=c, width=\ww]{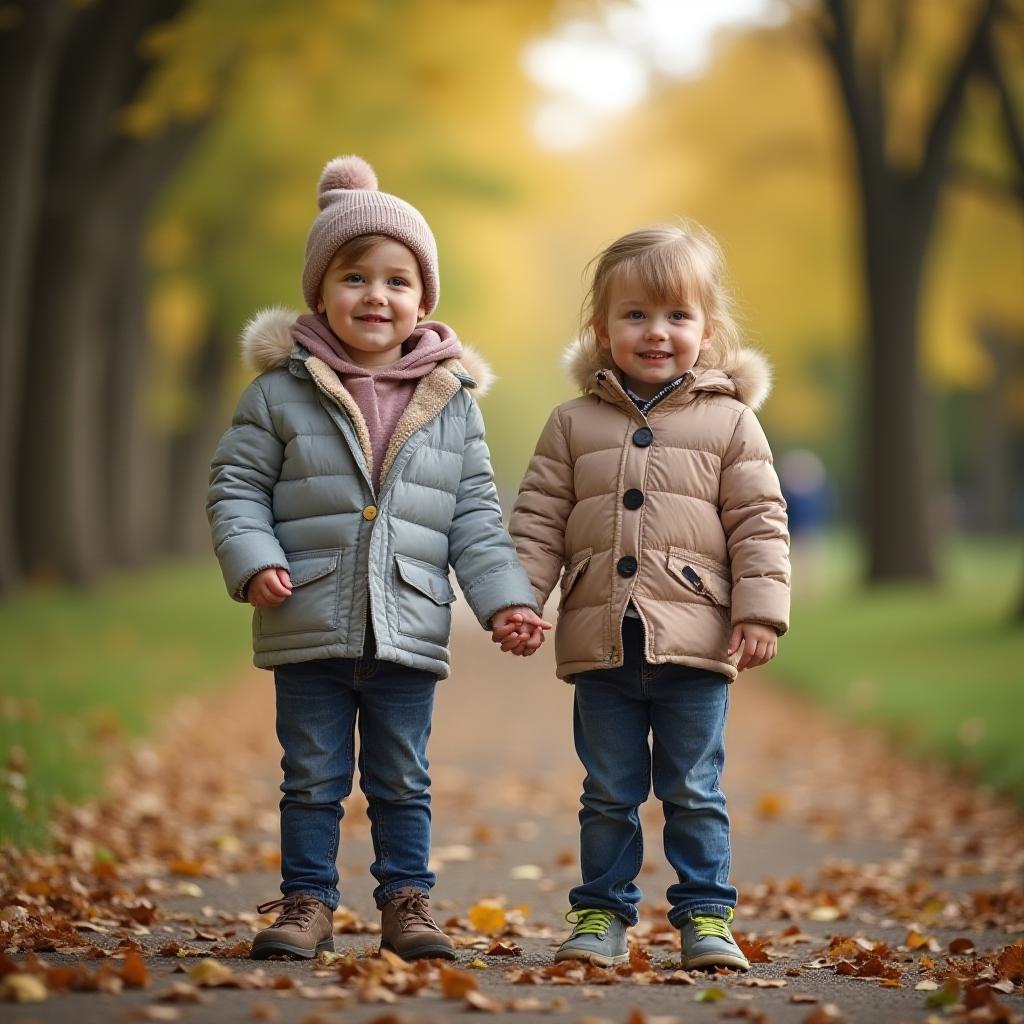}} &
        {\includegraphics[valign=c, width=\ww]{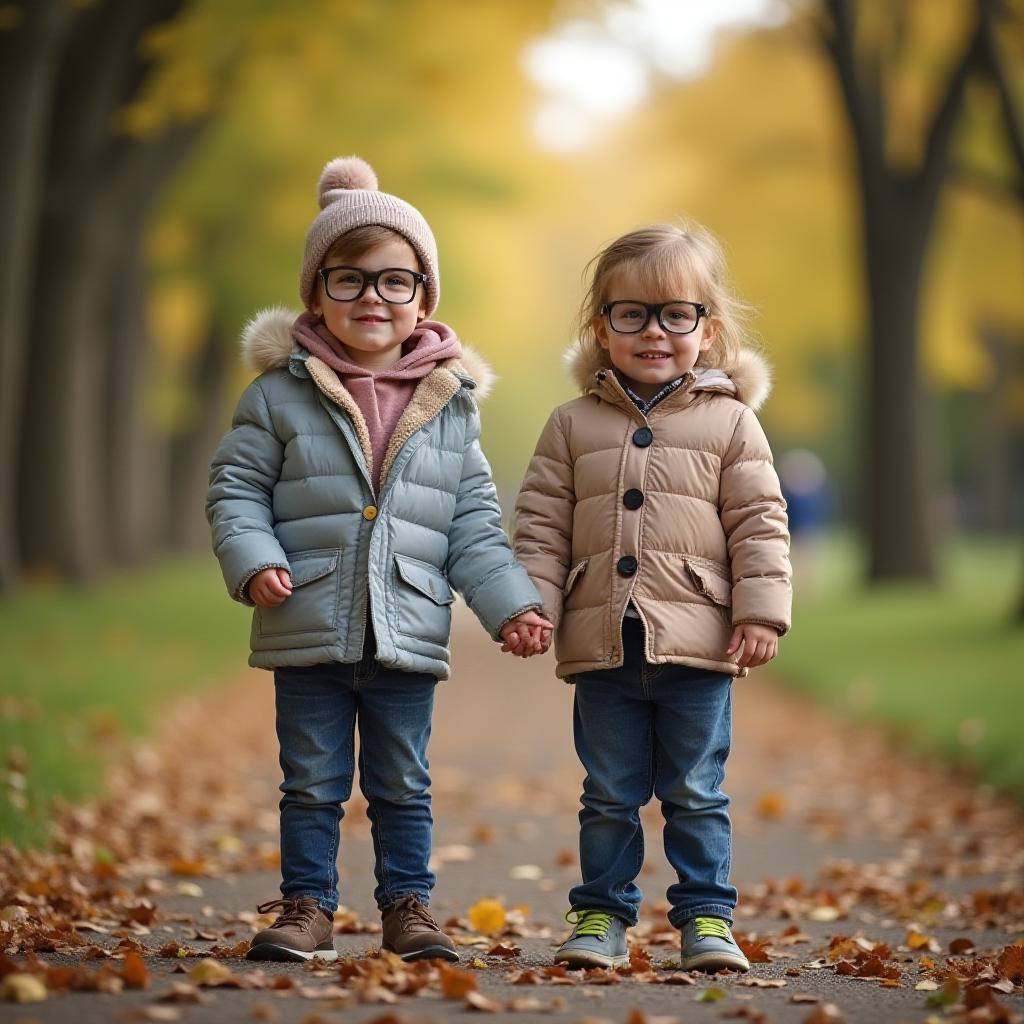}} &
        {\includegraphics[valign=c, width=\ww]{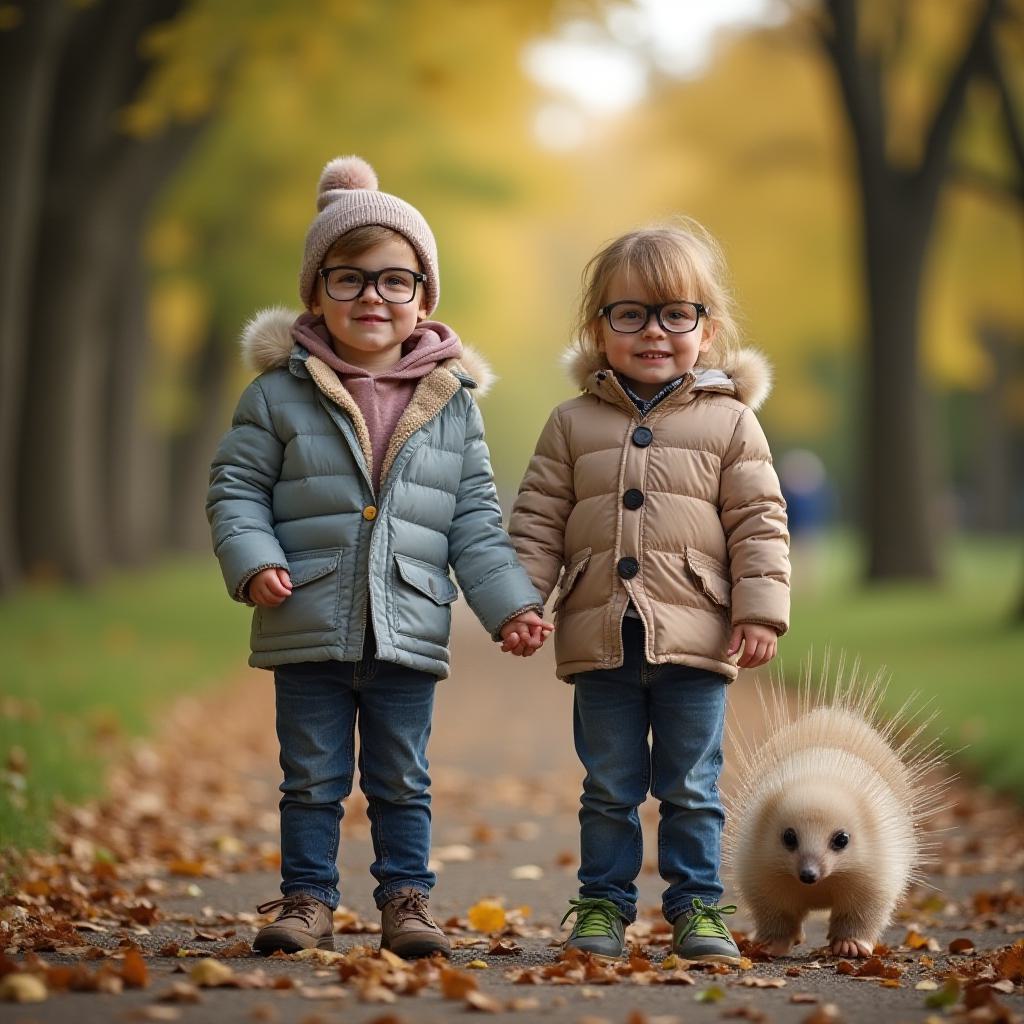}}
        \vspace{1px}
        \\

        &
        \scriptsize{Input} &
        \scriptsize{\prompt{Holding hands}} &
        \scriptsize{\prompt{Wearing glasses}} &
        \scriptsize{\promptstart{Next to an albino}}
        \\

        &
        &
        &
        &
        \scriptsize{\promptend{porcupine}}
        \vspace{3px}
        \\

        \rotatebox[origin=c]{90}{\scriptsize{(2) Const. Style}} &
        {\includegraphics[valign=c, width=\ww]{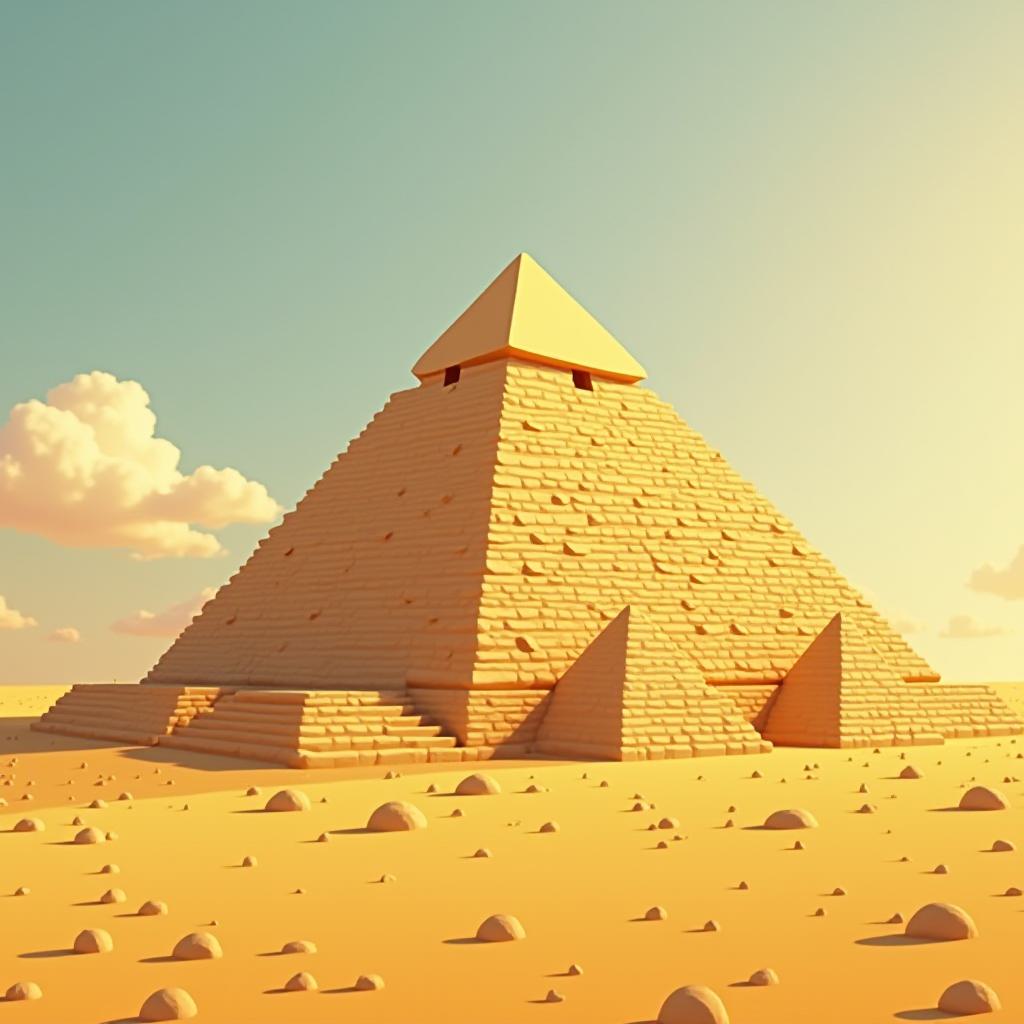}} &
        {\includegraphics[valign=c, width=\ww]{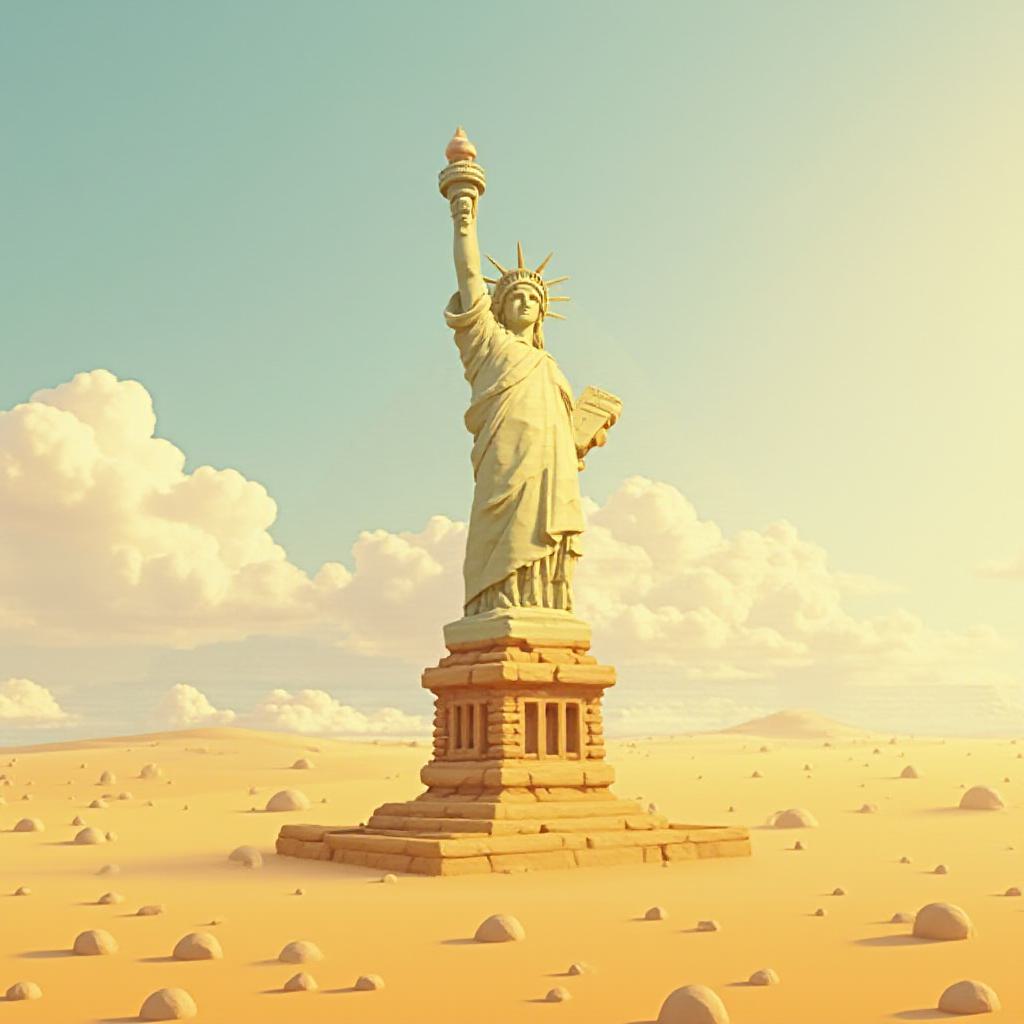}} &
        {\includegraphics[valign=c, width=\ww]{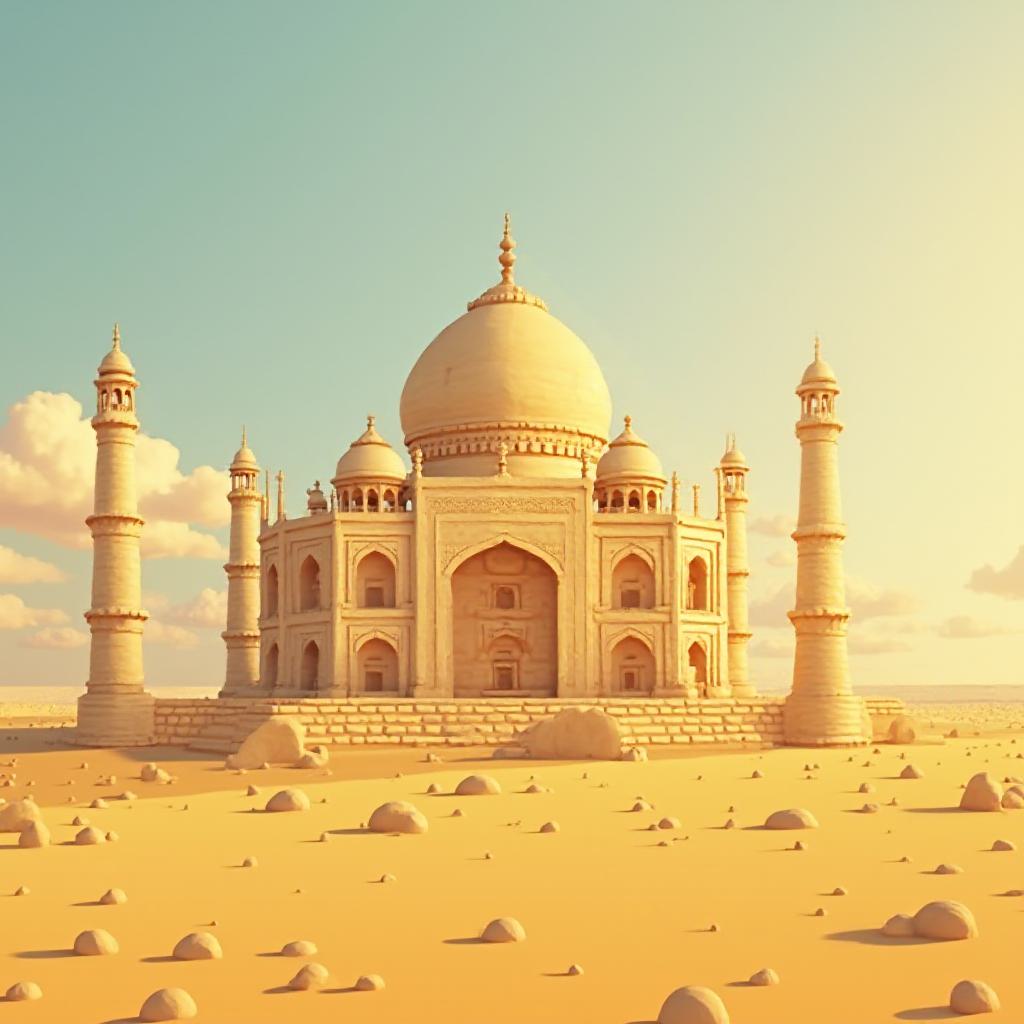}} &
        {\includegraphics[valign=c, width=\ww]{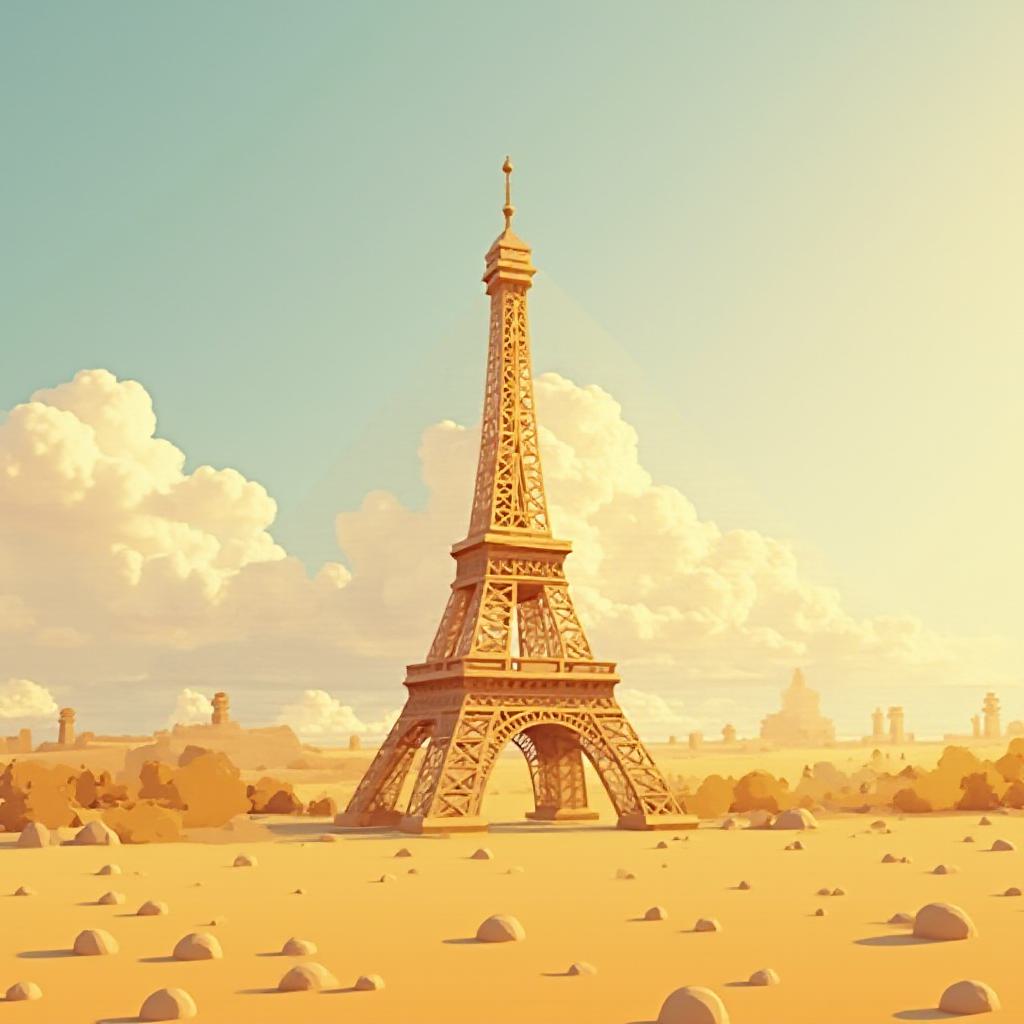}}
        \vspace{1px}
        \\

        &
        \scriptsize{Input} &
        \scriptsize{\prompt{Statue of Liberty}} &
        \scriptsize{\prompt{Taj Mahal}} &
        \scriptsize{\prompt{Eiffel Tower}}
        \vspace{3px}
        \\

        \rotatebox[origin=c]{90}{\scriptsize{(3) Text Editing}} &
        {\includegraphics[valign=c, width=\ww]{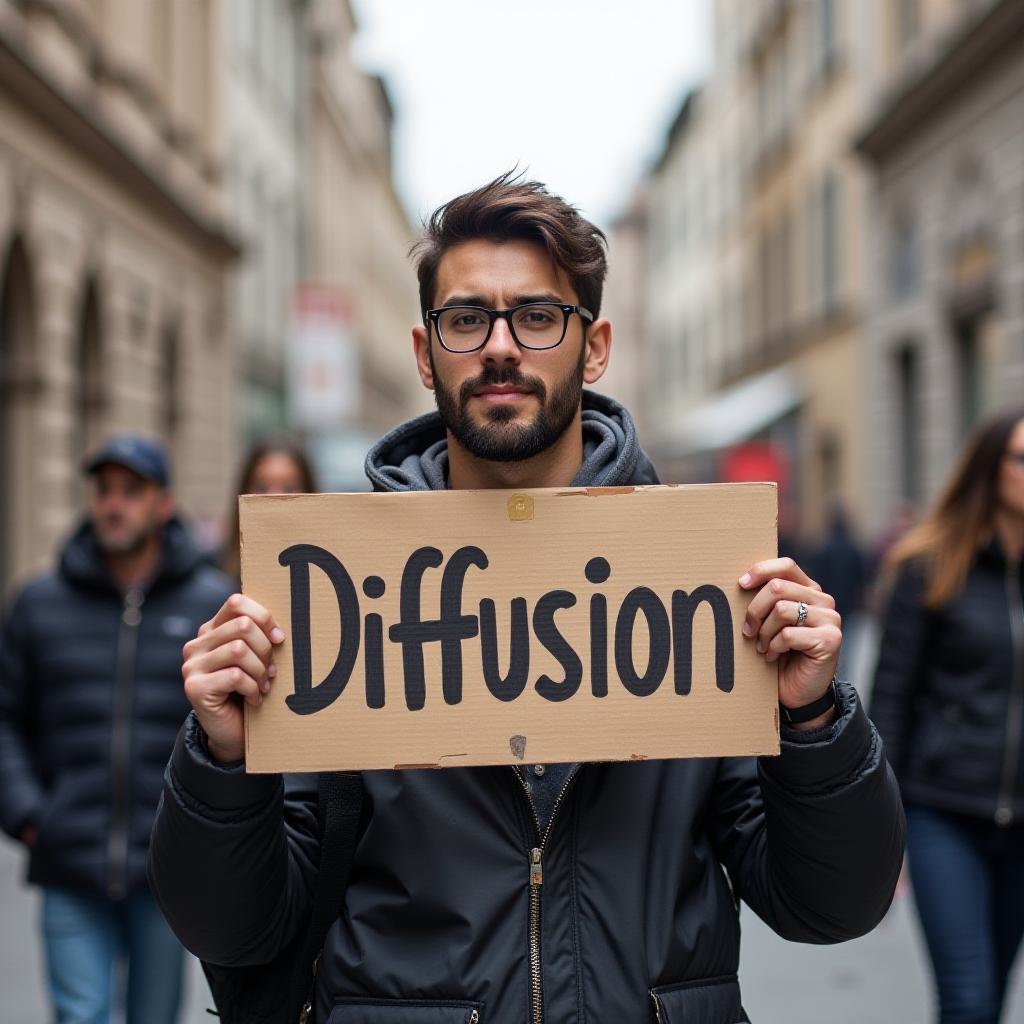}} &
        {\includegraphics[valign=c, width=\ww]{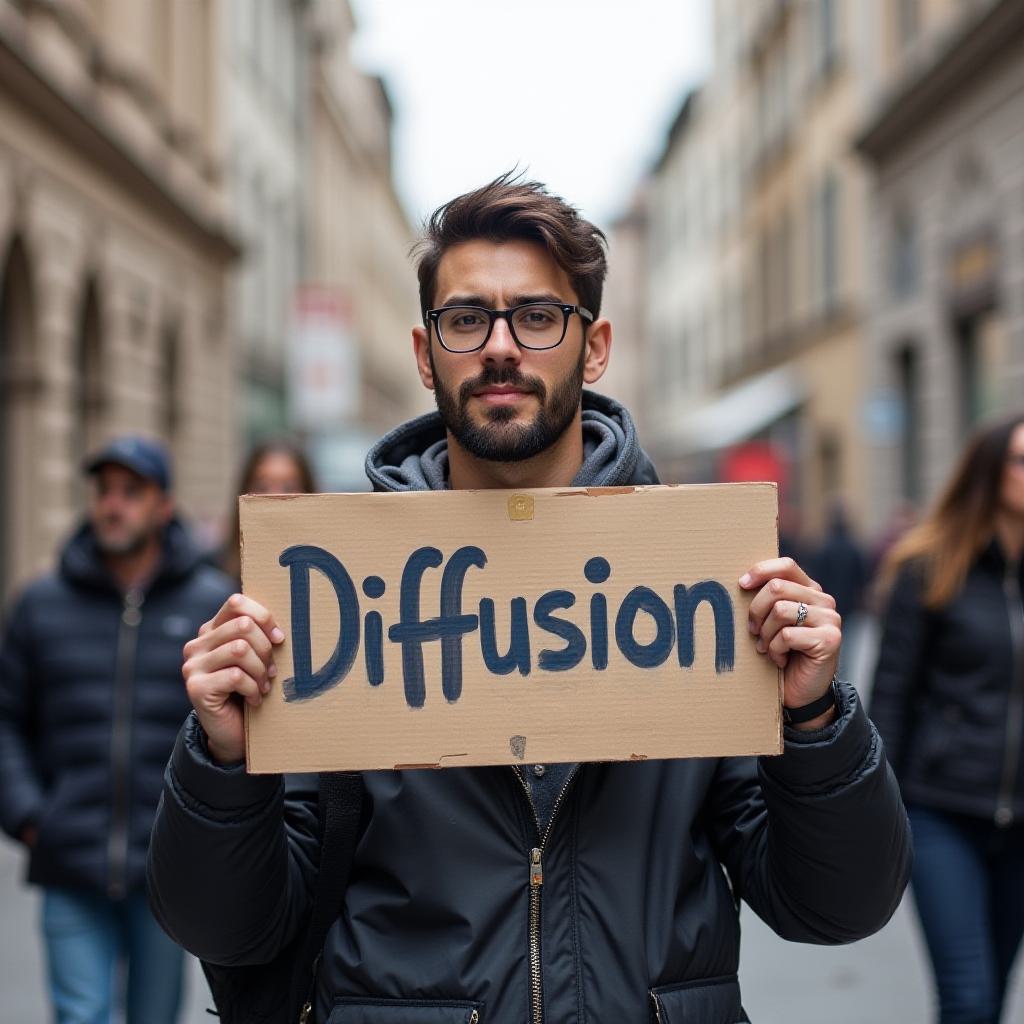}} &
        {\includegraphics[valign=c, width=\ww]{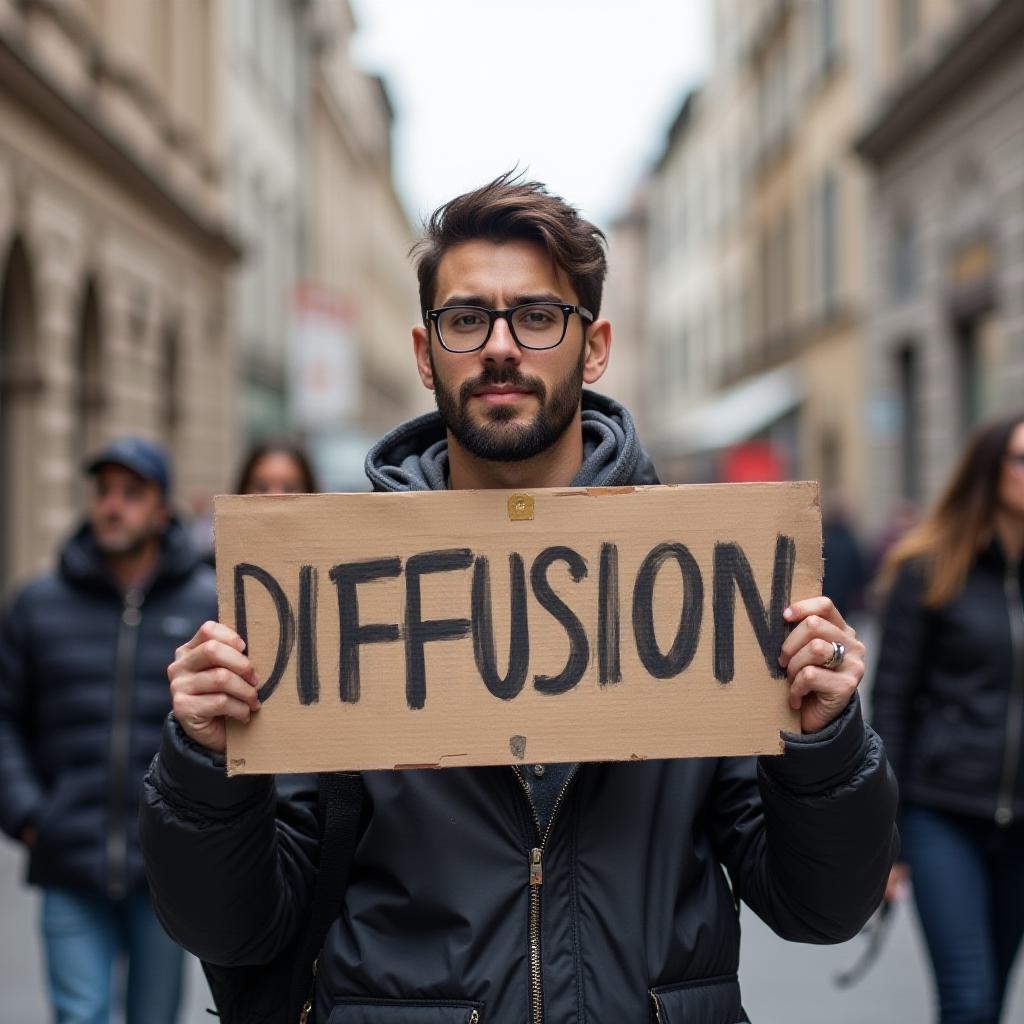}} &
        {\includegraphics[valign=c, width=\ww]{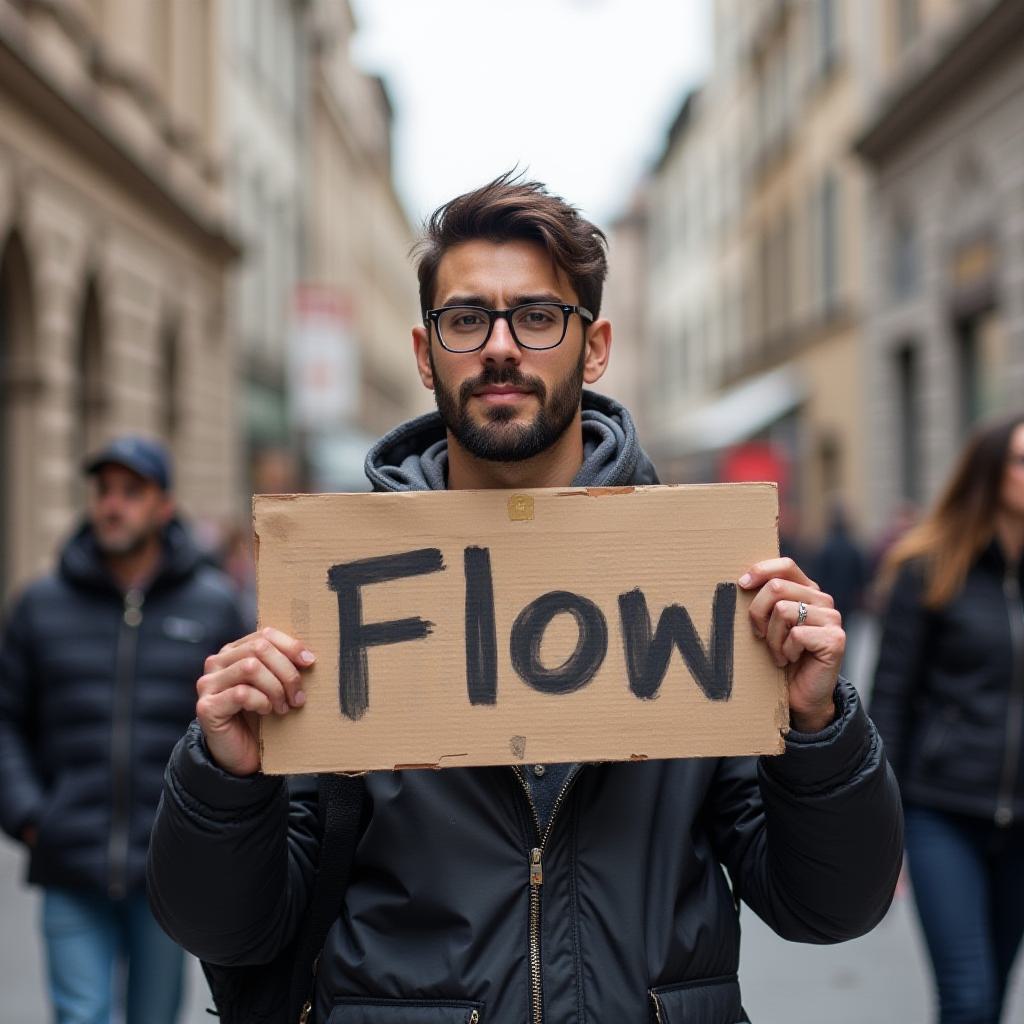}}
        \vspace{1px}
        \\

        &
        \scriptsize{Input} &
        \scriptsize{\promptstart{Man holding a}} &
        \scriptsize{\promptstart{Man holding a}} &
        \scriptsize{\promptstart{Man holding a}}
        \\

        &
        &
        \scriptsize{\textit{sign with the}} &
        \scriptsize{\textit{sign with the}} &
        \scriptsize{\textit{sign with the}}
        \\

        &
        &
        \scriptsize{\textit{text `diffusion'}} &
        \scriptsize{\textit{uppercase text}} &
        \scriptsize{\promptend{text `Flow' }}
        \\

        &
        &
        \scriptsize{\promptend{in a blue color}} &
        \scriptsize{\promptend{`DIFFUSION' }} &
        \\

    \end{tabular}
    \caption{\textbf{Applications.} Our method can be used for various applications: (1) Incremental Editing --- starting from a scene of two kids, the user can refine the image \emph{iteratively} by making the kid hold hands, then wear glasses and finally add a porcupine next to them. (2) Consistent Style --- starting from a scene with a given style, such as an animation of the Great Pyramid of Giza, the user can generate images of different places with the same style. (3) Text Editing --- given a scene that contains text, our method is able to perform text-related editing such as color change, case change and text replacement.}
    \label{fig:applications}
    \vspace{-10px}
\end{figure}

As demonstrated in \Cref{fig:applications}, our method can be used for various applications: (1) Incremental Editing --- starting from a given scene, the user can refine the image \emph{iteratively} in a step-by-step manner. (2) Consistent Style --- starting from a scene with a given style, the user can generate other images in the same style~\cite{Frenkel2024ImplicitSS, Hertz2023StyleAI, Jeong2024VisualSP}. (3) Text Editing --- given a scene that contains text, our method is able to perform text-related editing such as color change, case change and text replacement.

\section{Limitations and Conclusions}
\label{sec:limitations_and_conclusions}

\begin{figure}[tp]
    \centering
    \setlength{\tabcolsep}{0.6pt}
    \renewcommand{\arraystretch}{0.7}
    \setlength{\ww}{0.235\columnwidth}
    \begin{tabular}{cc @{\hspace{7\tabcolsep}} ccc}
        \rotatebox[origin=c]{90}{\scriptsize{(1) Style Editing}} &
        {\includegraphics[valign=c, width=\ww]{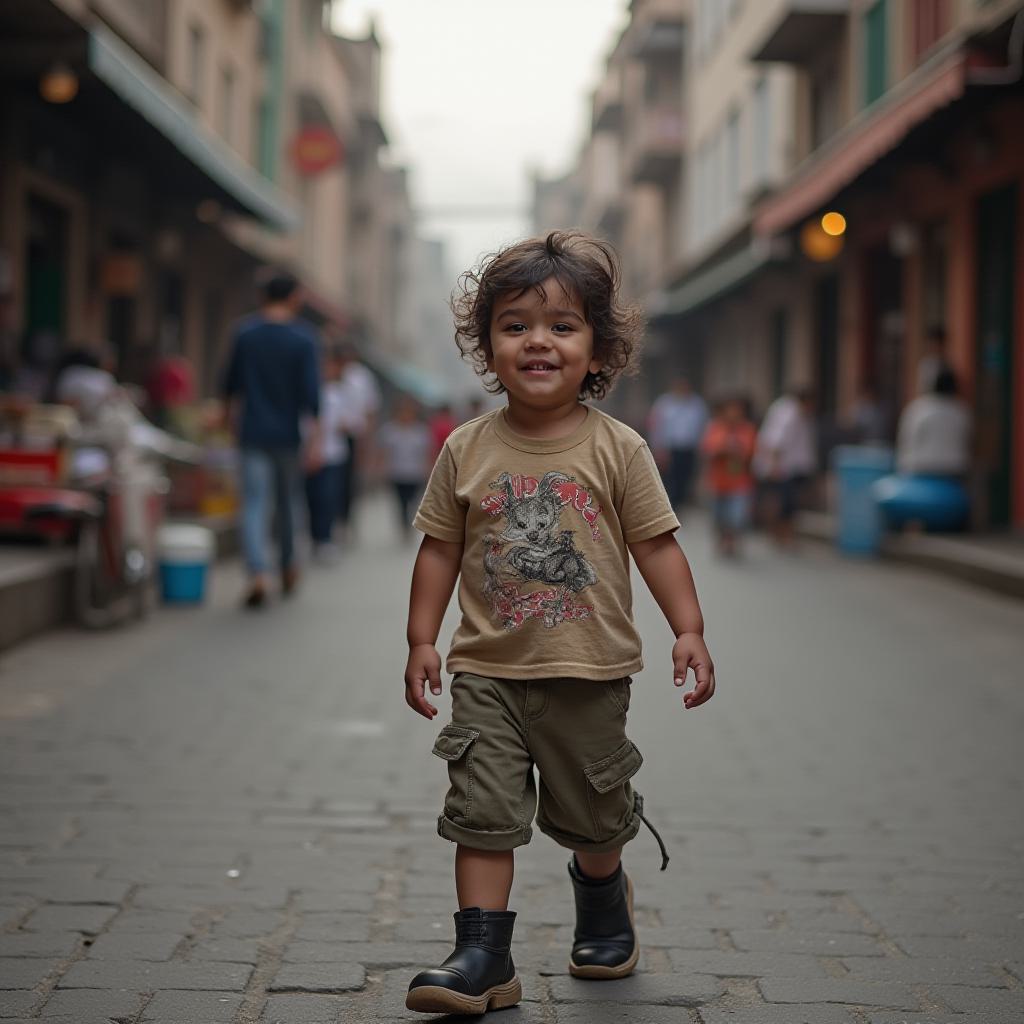}} &
        {\includegraphics[valign=c, width=\ww]{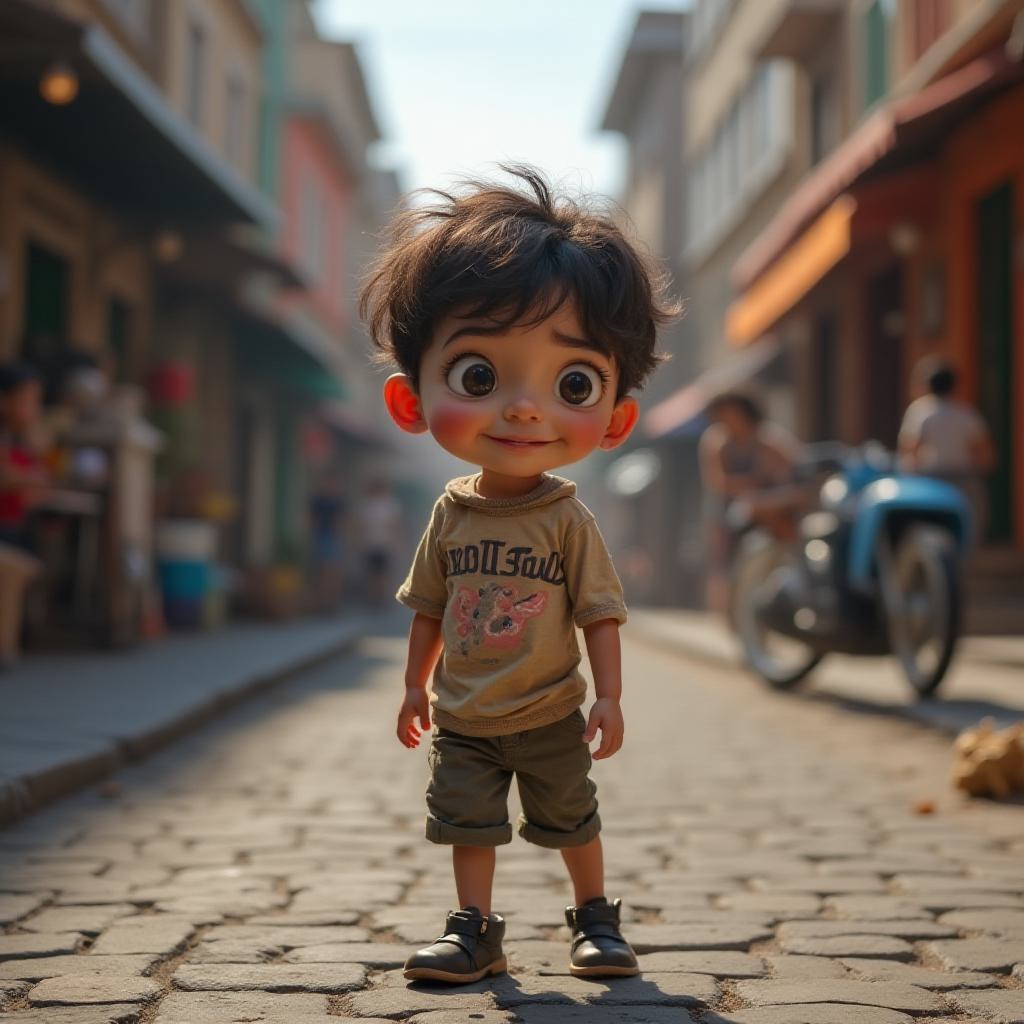}} &
        {\includegraphics[valign=c, width=\ww]{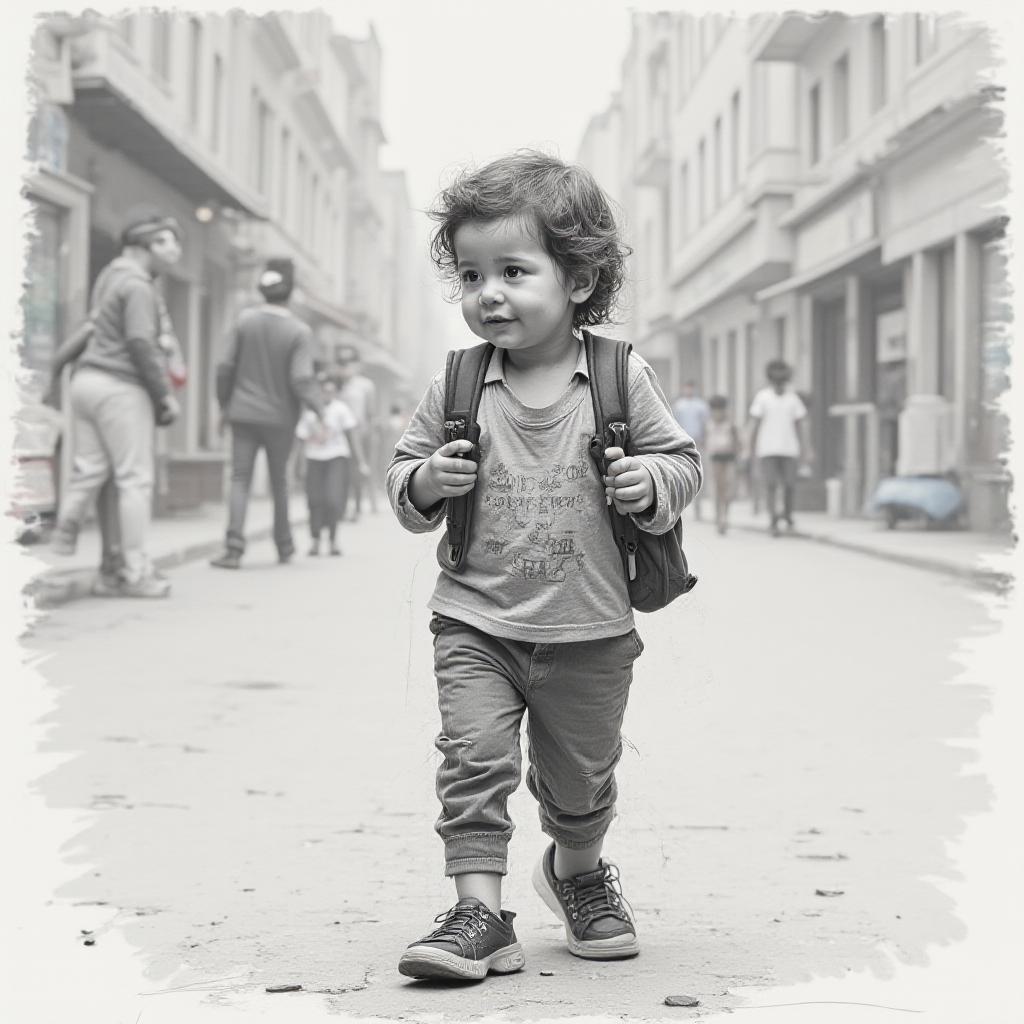}} &
        {\includegraphics[valign=c, width=\ww]{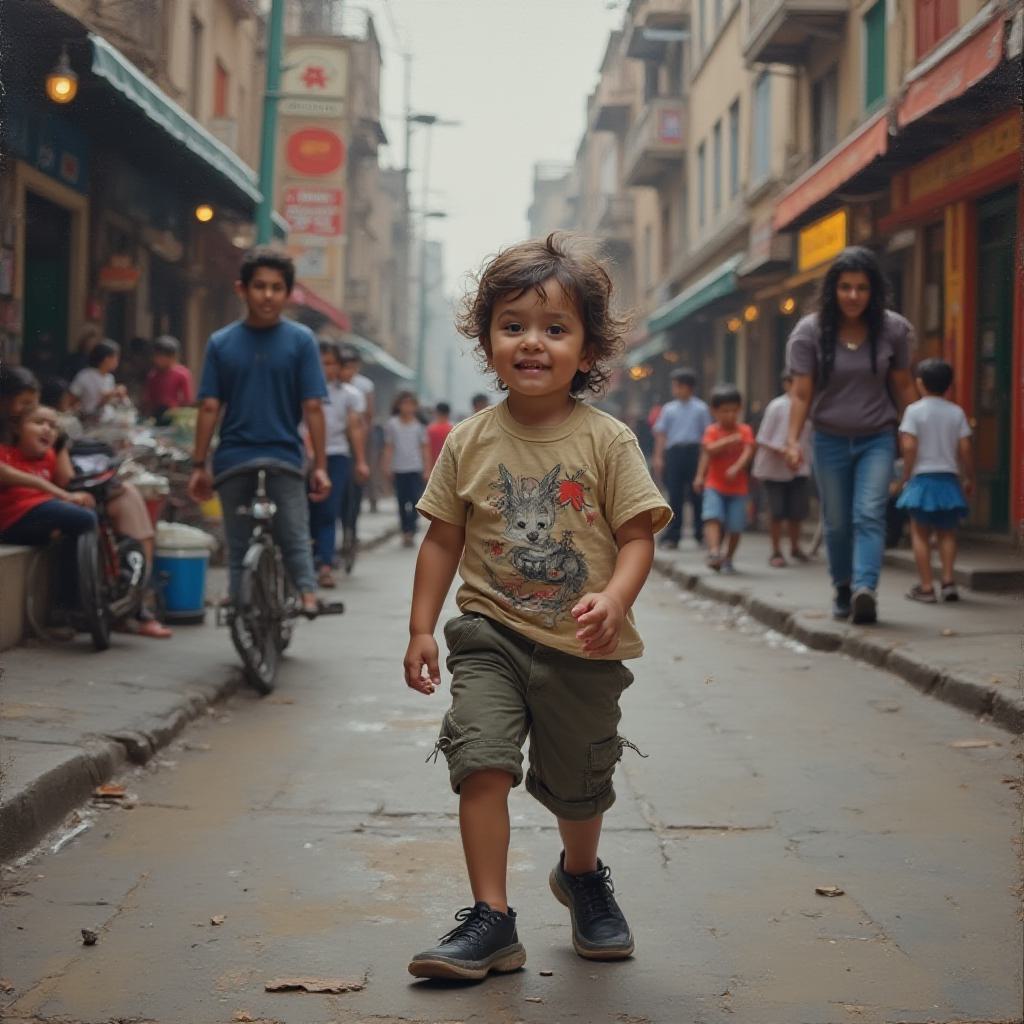}}
        \vspace{1px}
        \\

        &
        \scriptsize{Input} &
        \scriptsize{\prompt{Animation}} &
        \scriptsize{\prompt{Pencil sketch}} &
        \scriptsize{\prompt{Oil painting}}
        \vspace{3px}
        \\

        \rotatebox[origin=c]{90}{\scriptsize{(2) Object Drag.}}&
        {\includegraphics[valign=c, width=\ww]{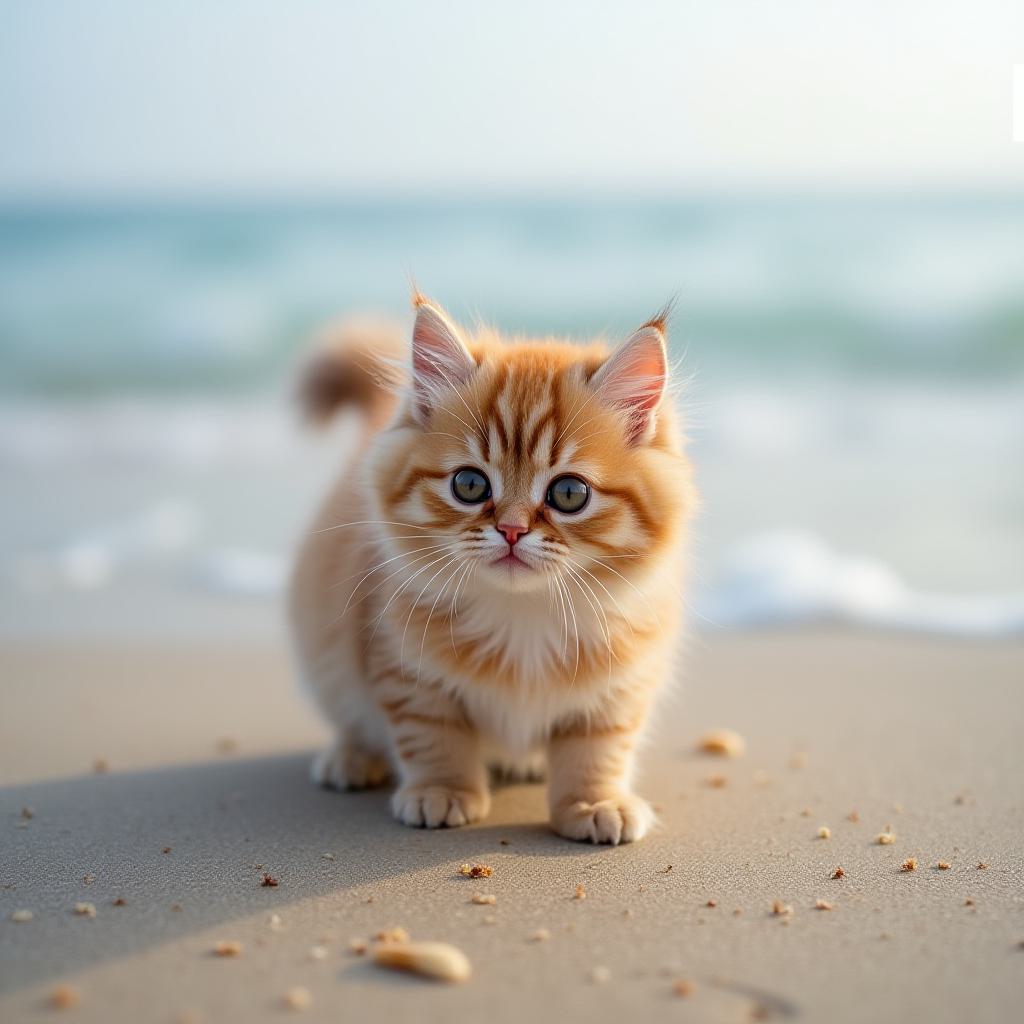}} &
        {\includegraphics[valign=c, width=\ww]{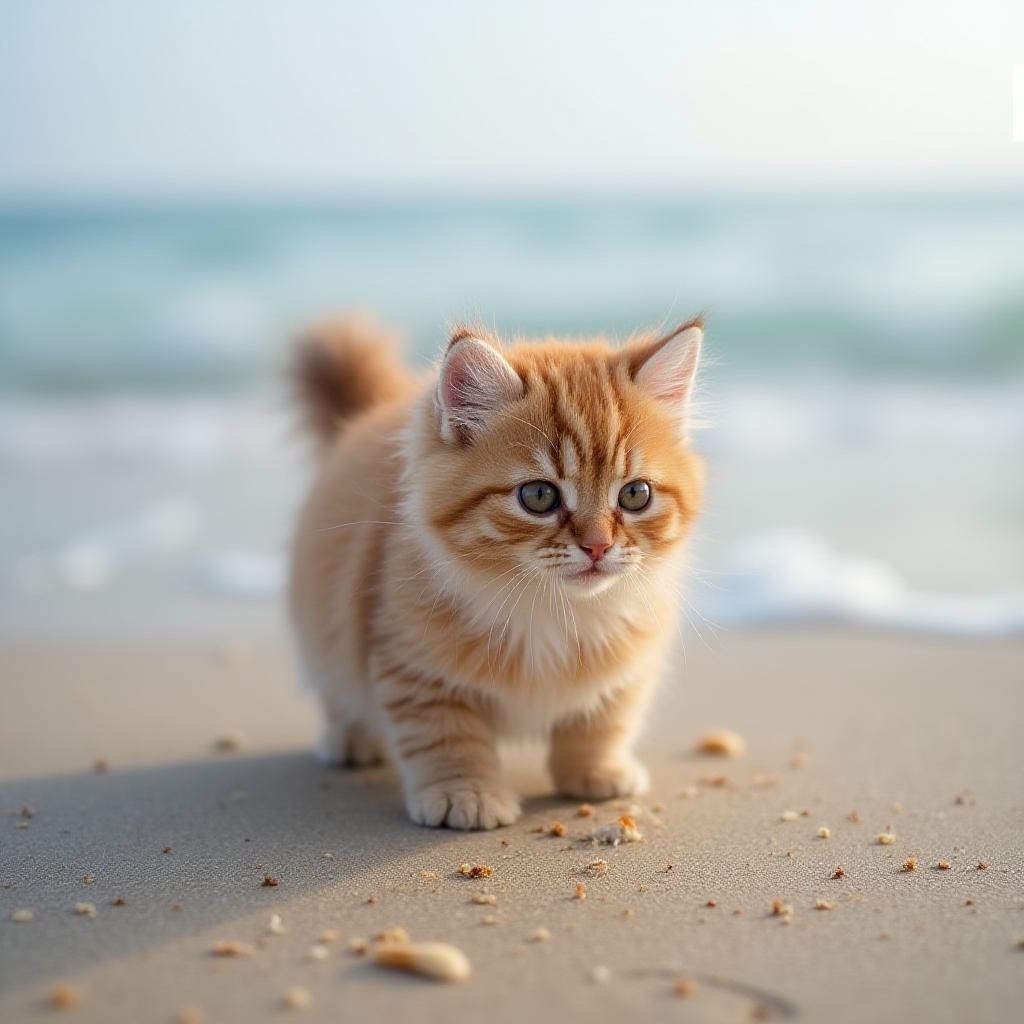}} &
        {\includegraphics[valign=c, width=\ww]{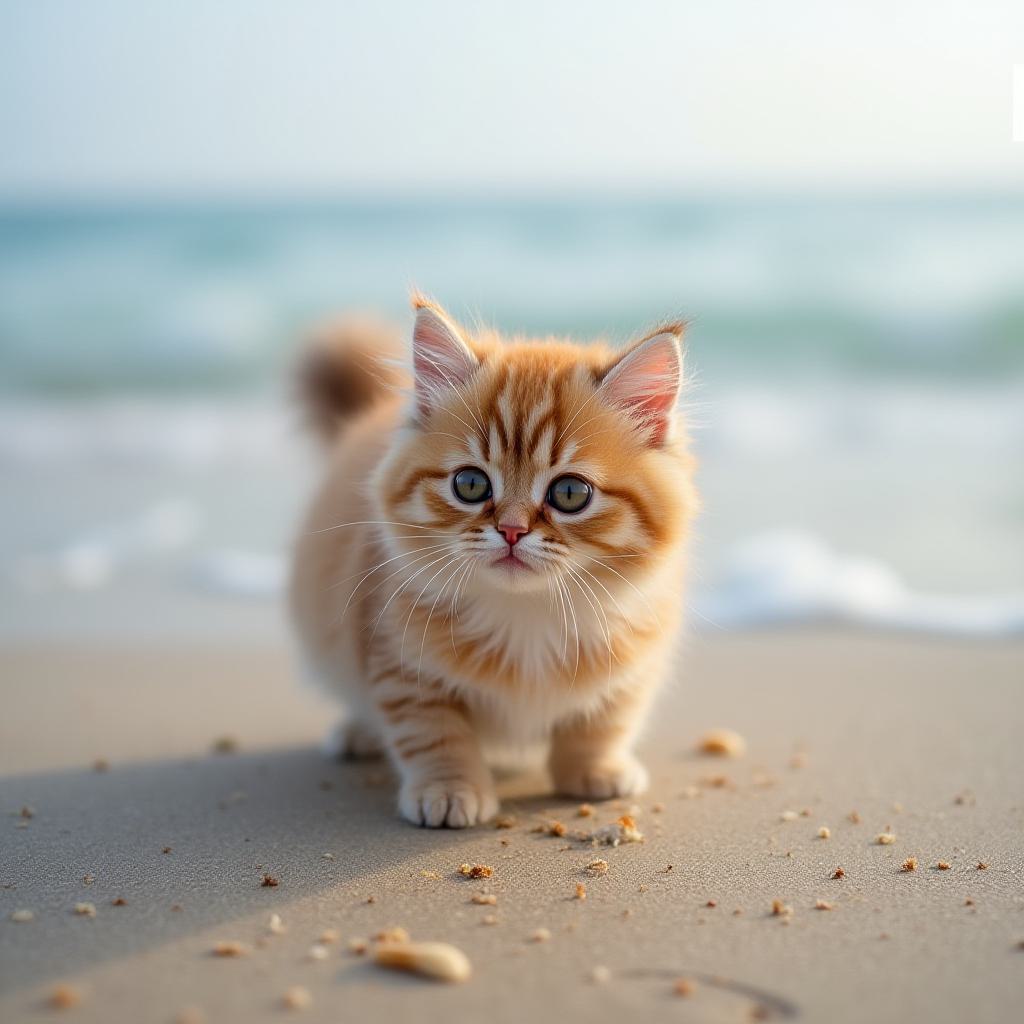}} &
        {\includegraphics[valign=c, width=\ww]{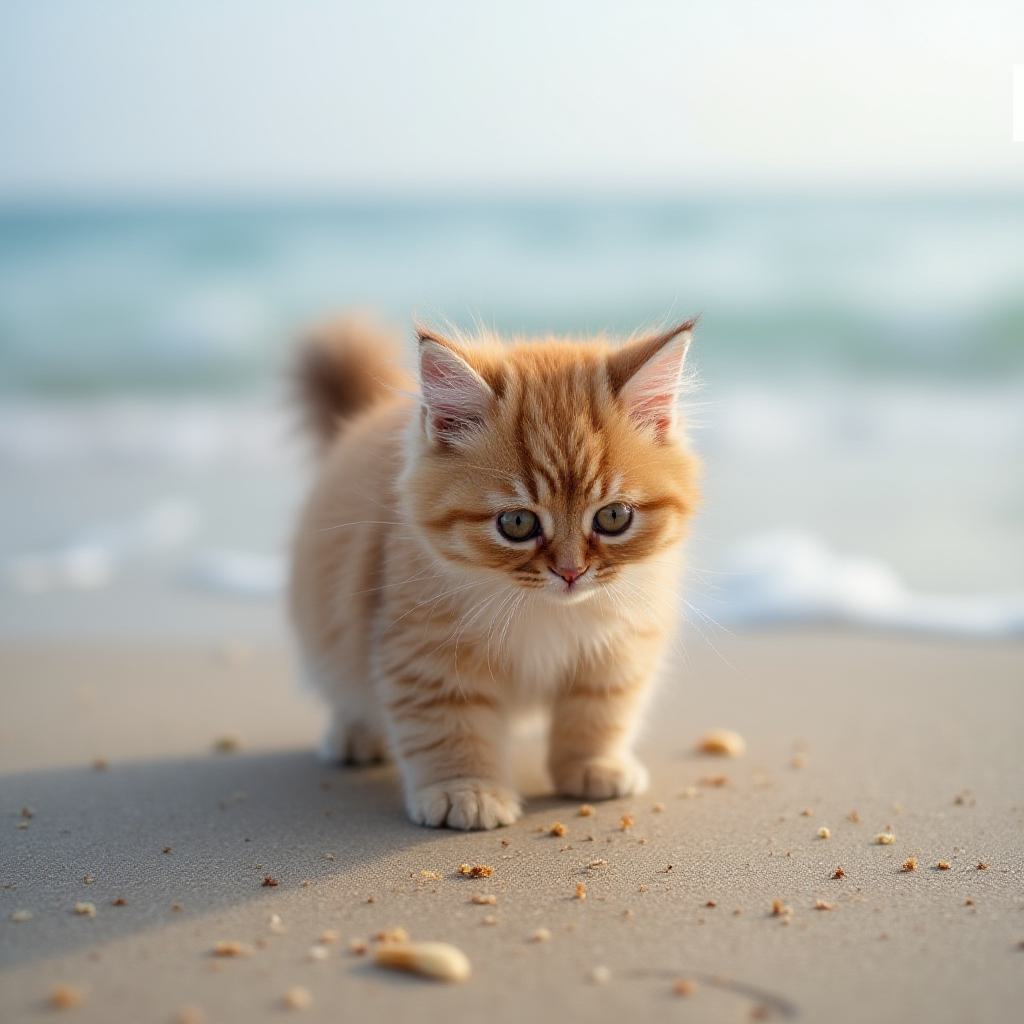}}
        \vspace{1px}
        \\

        &
        \scriptsize{Input} &
        \scriptsize{\promptstart{On the right side}} &
        \scriptsize{\promptstart{On the left side}} &
        \scriptsize{\promptstart{On the bottom}}
        \\

        &
        &
        \scriptsize{\promptend{of the frame}} &
        \scriptsize{\promptend{of the frame}} &
        \scriptsize{\promptend{of the frame}}
        \vspace{3px}
        \\

        \rotatebox[origin=c]{90}{\scriptsize{(3) Bg. Replace}} &
        {\includegraphics[valign=c, width=\ww]{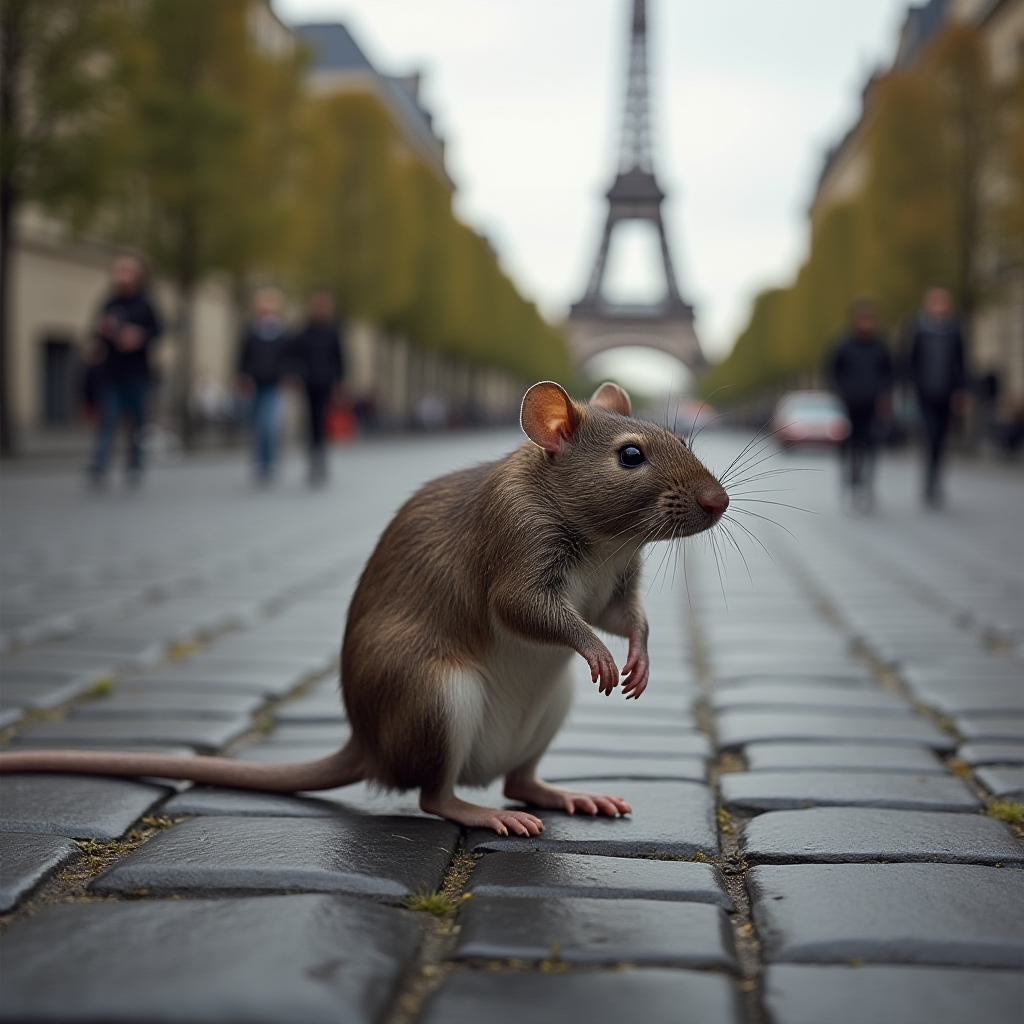}} &
        {\includegraphics[valign=c, width=\ww]{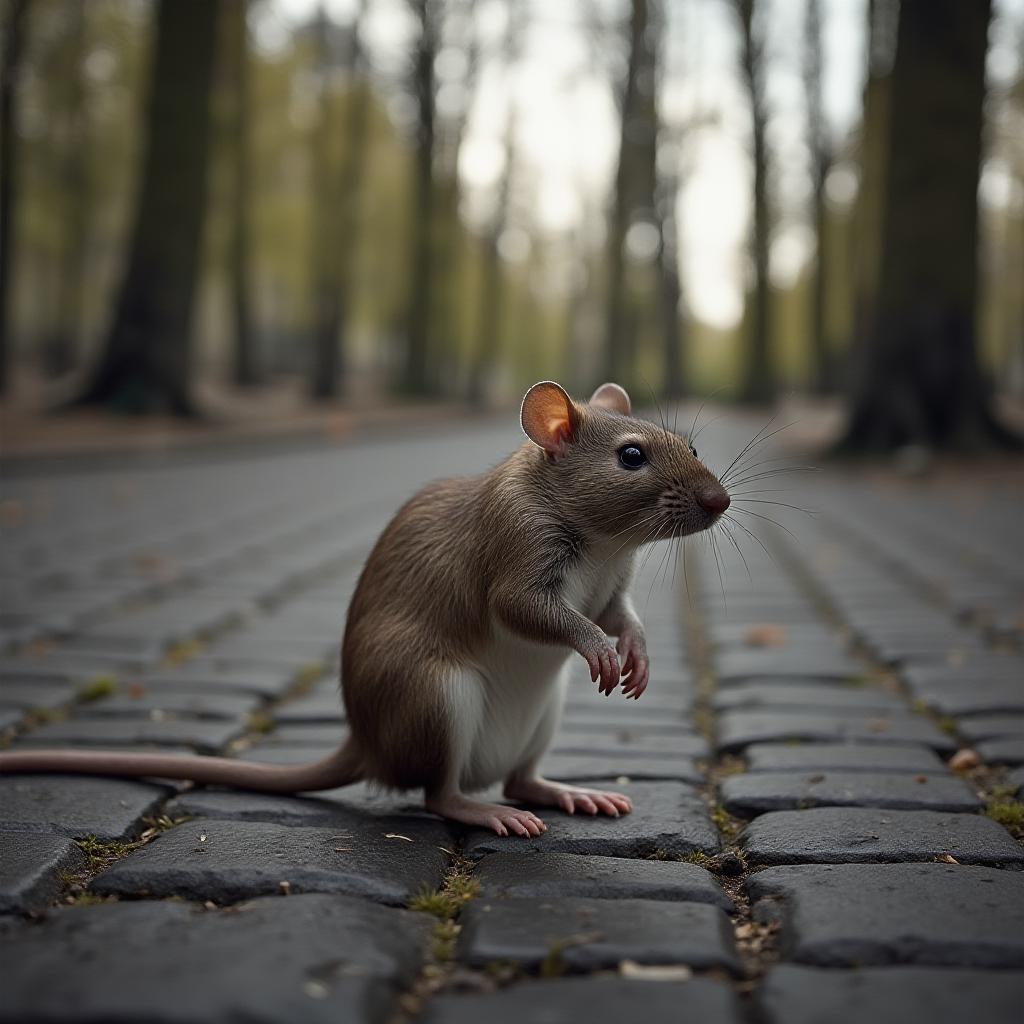}} &
        {\includegraphics[valign=c, width=\ww]{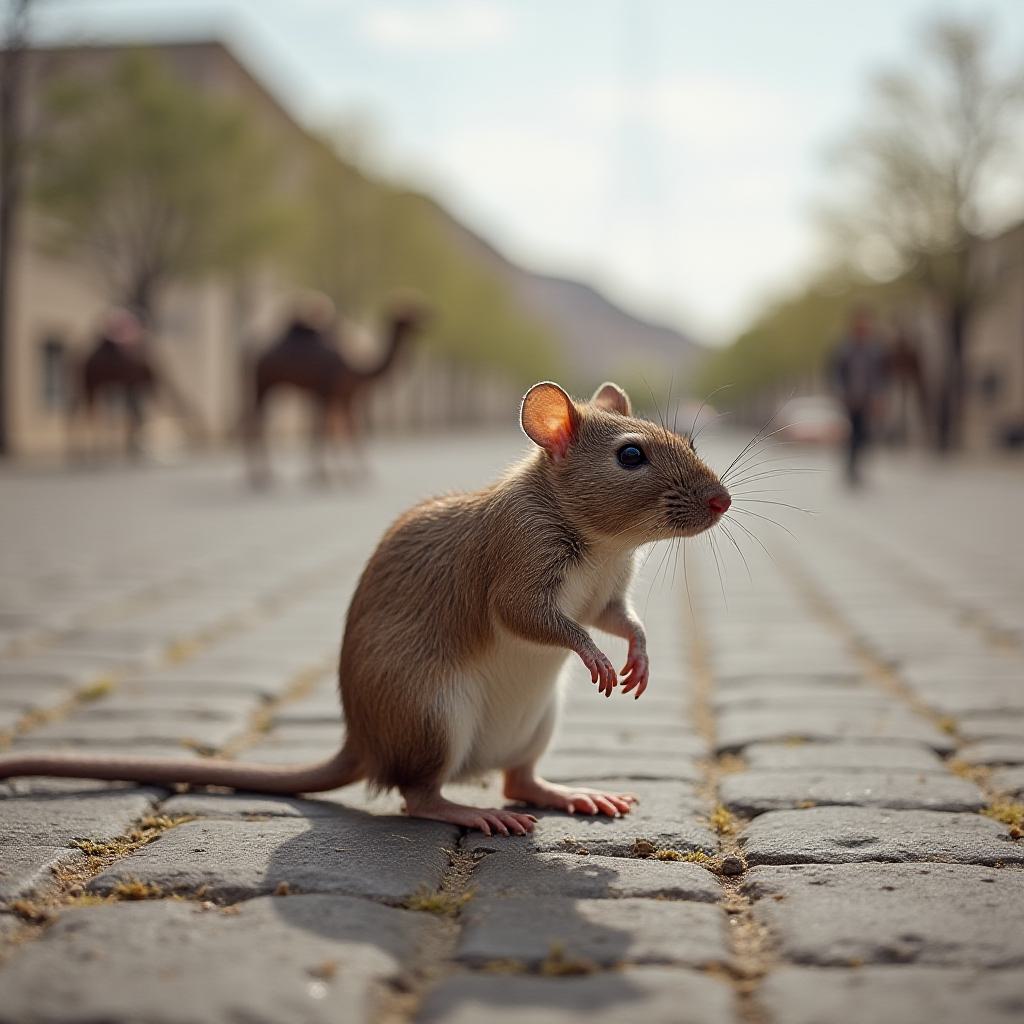}} &
        {\includegraphics[valign=c, width=\ww]{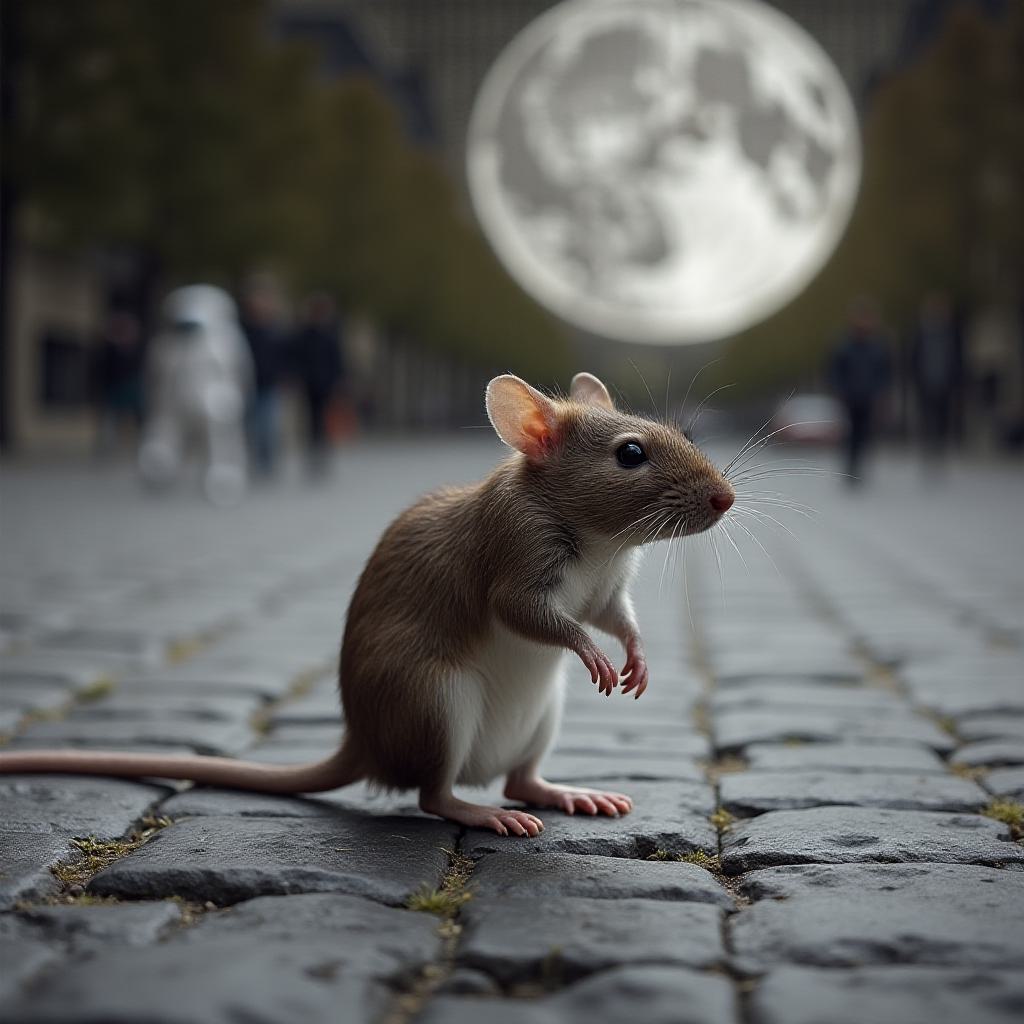}}
        \vspace{1px}
        \\

        &
        \scriptsize{Input} &
        \scriptsize{\prompt{In the forest}} &
        \scriptsize{\prompt{In the desert}} &
        \scriptsize{\prompt{On the moon}}
        \\

    \end{tabular}
    \vspace{-3px}
    \caption{\textbf{Limitations.} Our method suffers from the following limitations: (1) Style Editing --- given a photorealistic image of a boy, our method struggles with changing its style to an animation (the identity of the boy also changes), to pencil sketch (changes only to black\&white) or to an oil painting (mainly makes the image smoother). (2) Object Dragging --- given an image of a cat, our method is unable to drag it into different locations in the image, but changes the gaze of the cat instead. (3) Background Replacement --- given an image of a rat on the road, our method unable to replace its background entirely (the road leaks).}
    \label{fig:limitations}
    \vspace{-10px}
\end{figure}

As demonstrated in \Cref{fig:limitations}, our method suffers from the following limitations: (1) Style Editing --- given an input image in one style  (\eg, photorealistic), our method struggles with changing it to a different style (\eg, oil painting), as it relies on attention injection (\Cref{sec:image_editing}). (2) Object Dragging --- given an image with an object, our method is unable to drag it~\cite{avrahami2024diffuhaul} into different locations in the image, as text-to-image models often struggle~\cite{spatext_2023_CVPR} with spatial prompt adherence. (3) Background Replacement --- given an input image, our method struggles with replacing its background entirely with no leakage~\cite{Dahary2024BeYB}.

In conclusion, we present Stable Flow, a training-free method for image editing that enables \emph{various} image editing tasks using the attention injection of the \emph{same} vital layers group. We believe that our fully-automated approach of detecting vital layers may be also beneficial for other use-cases, such as generative models pruning and distillation. We hope that our layer analysis will inspire more work in the field of generative models and expanding the possibilities for creative expression.

\clearpage

\textbf{Acknowledgments.} We thank Omer Dahary for his valuable help and feedback. This work was supported in part by the Israel Science Foundation (grants 1574/21 and 2203/24).

{
    \small
    \bibliographystyle{ieeenat_fullname}
    \bibliography{egbib}

\begin{thebibliography}{102}
\providecommand{\natexlab}[1]{#1}
\providecommand{\url}[1]{\texttt{#1}}
\expandafter\ifx\csname urlstyle\endcsname\relax
  \providecommand{\doi}[1]{doi: #1}\else
  \providecommand{\doi}{doi: \begingroup \urlstyle{rm}\Url}\fi

\bibitem[Abdal et~al.(2019)Abdal, Qin, and Wonka]{abdal2019image2stylegan}
Rameen Abdal, Yipeng Qin, and Peter Wonka.
\newblock Image2stylegan: How to embed images into the stylegan latent space?
\newblock In \emph{Proceedings of the IEEE/CVF International Conference on Computer Vision}, pages 4432--4441, 2019.

\bibitem[Abdal et~al.(2020)Abdal, Qin, and Wonka]{abdal2020image2stylegan++}
Rameen Abdal, Yipeng Qin, and Peter Wonka.
\newblock Image2stylegan++: How to edit the embedded images?
\newblock In \emph{Proceedings of the IEEE/CVF conference on computer vision and pattern recognition}, pages 8296--8305, 2020.

\bibitem[Abdin et~al.(2024)Abdin, Jacobs, Awan, Aneja, Awadallah, Awadalla, Bach, Bahree, Bakhtiari, Behl, Benhaim, Bilenko, Bjorck, Bubeck, Cai, Mendes, Chen, Chaudhary, Chopra, Giorno, de~Rosa, Dixon, Eldan, Iter, Goswami, Gunasekar, Haider, Hao, Hewett, Huynh, Javaheripi, Jin, Kauffmann, Karampatziakis, Kim, Kim, Khademi, Kurilenko, Lee, Lee, Li, Liang, Liu, Lin, Lin, Madan, Mitra, Modi, Nguyen, Norick, Patra, Perez-Becker, Portet, Pryzant, Qin, Radmilac, Rosset, Roy, Saarikivi, Saied, Salim, Santacroce, Shah, Shang, Sharma, Song, Ruwase, Vaddamanu, Wang, Ward, Wang, Witte, Wyatt, Xu, Xu, Yadav, Yang, Yang, Yu, Zhang, Zhang, Zhang, Zhang, Zhang, Zhang, and Zhou]{Abdin2024Phi3TR}
Marah Abdin, Sam~Ade Jacobs, Ammar~Ahmad Awan, Jyoti Aneja, Ahmed Awadallah, Hany~Hassan Awadalla, Nguyen Bach, Amit Bahree, Arash Bakhtiari, Harkirat~Singh Behl, Alon Benhaim, Misha Bilenko, Johan Bjorck, S{\'e}bastien Bubeck, Martin Cai, Caio C'esar~Teodoro Mendes, Weizhu Chen, Vishrav Chaudhary, Parul Chopra, Allison~Del Giorno, Gustavo de Rosa, Matthew Dixon, Ronen Eldan, Dan Iter, Abhishek Goswami, Suriya Gunasekar, Emman Haider, Junheng Hao, Russell~J. Hewett, Jamie Huynh, Mojan Javaheripi, Xin Jin, Piero Kauffmann, Nikos Karampatziakis, Dongwoo Kim, Young~Jin Kim, Mahoud Khademi, Lev Kurilenko, James~R. Lee, Yin~Tat Lee, Yuanzhi Li, Chen Liang, Weishung Liu, Eric Lin, Zeqi Lin, Piyush Madan, Arindam Mitra, Hardik Modi, Anh Nguyen, Brandon Norick, Barun Patra, Daniel Perez-Becker, Thomas Portet, Reid Pryzant, Heyang Qin, Marko Radmilac, Corby Rosset, Sambudha Roy, Olli Saarikivi, Amin Saied, Adil Salim, Michael Santacroce, Shital Shah, Ning Shang, Hiteshi Sharma, Xianmin Song, Olatunji Ruwase, Praneetha Vaddamanu, Xin Wang, Rachel Ward, Guanhua Wang, Philipp Witte, Michael Wyatt, Can Xu, Jiahang Xu, Sonali Yadav, Fan Yang, Ziyi Yang, Donghan Yu, Cheng-Yuan Zhang, Cyril Zhang, Jianwen Zhang, Li~Lyna Zhang, Yi Zhang, Yunan Zhang, and Xiren Zhou.
\newblock Phi-3 technical report: A highly capable language model locally on your phone.
\newblock \emph{ArXiv}, abs/2404.14219, 2024.

\bibitem[Alaluf et~al.(2021)Alaluf, Tov, Mokady, Gal, and Bermano]{alaluf2021hyperstyle}
Yuval Alaluf, Omer Tov, Ron Mokady, Rinon Gal, and Amit~Haim Bermano.
\newblock Hyperstyle: Stylegan inversion with hypernetworks for real image editing.
\newblock \emph{2022 IEEE/CVF Conference on Computer Vision and Pattern Recognition (CVPR)}, pages 18490--18500, 2021.

\bibitem[Alaluf et~al.(2023)Alaluf, Garibi, Patashnik, Averbuch-Elor, and Cohen-Or]{Alaluf2023CrossImageAF}
Yuval Alaluf, Daniel Garibi, Or Patashnik, Hadar Averbuch-Elor, and Daniel Cohen-Or.
\newblock Cross-image attention for zero-shot appearance transfer.
\newblock In \emph{International Conference on Computer Graphics and Interactive Techniques}, 2023.

\bibitem[Albergo and Vanden-Eijnden(2022)]{Albergo2022BuildingNF}
Michael~S Albergo and Eric Vanden-Eijnden.
\newblock Building normalizing flows with stochastic interpolants.
\newblock \emph{ArXiv}, abs/2209.15571, 2022.

\bibitem[Amazon(2024)]{amt}
Amazon.
\newblock Amazon mechanical turk.
\newblock \url{https://www.mturk.com/}, 2024.

\bibitem[Arar et~al.(2024)Arar, Voynov, Hertz, Avrahami, Fruchter, Pritch, Cohen-Or, and Shamir]{arar2024Palp}
Moab Arar, Andrey Voynov, Amir Hertz, Omri Avrahami, Shlomi Fruchter, Yael Pritch, Daniel Cohen-Or, and Ariel Shamir.
\newblock Palp: Prompt aligned personalization of text-to-image models.
\newblock 2024.

\bibitem[Avrahami et~al.(2022)Avrahami, Lischinski, and Fried]{blended_2022_CVPR}
Omri Avrahami, Dani Lischinski, and Ohad Fried.
\newblock Blended diffusion for text-driven editing of natural images.
\newblock In \emph{Proceedings of the IEEE/CVF Conference on Computer Vision and Pattern Recognition (CVPR)}, pages 18208--18218, 2022.

\bibitem[Avrahami et~al.(2023{\natexlab{a}})Avrahami, Fried, and Lischinski]{avrahami2023blendedlatent}
Omri Avrahami, Ohad Fried, and Dani Lischinski.
\newblock Blended latent diffusion.
\newblock \emph{ACM Trans. Graph.}, 42\penalty0 (4), 2023{\natexlab{a}}.

\bibitem[Avrahami et~al.(2023{\natexlab{b}})Avrahami, Hayes, Gafni, Gupta, Taigman, Parikh, Lischinski, Fried, and Yin]{spatext_2023_CVPR}
Omri Avrahami, Thomas Hayes, Oran Gafni, Sonal Gupta, Yaniv Taigman, Devi Parikh, Dani Lischinski, Ohad Fried, and Xi Yin.
\newblock Spatext: Spatio-textual representation for controllable image generation.
\newblock In \emph{Proceedings of the IEEE/CVF Conference on Computer Vision and Pattern Recognition (CVPR)}, pages 18370--18380, 2023{\natexlab{b}}.

\bibitem[Avrahami et~al.(2024{\natexlab{a}})Avrahami, Gal, Chechik, Fried, Lischinski, Vahdat, and Nie]{avrahami2024diffuhaul}
Omri Avrahami, Rinon Gal, Gal Chechik, Ohad Fried, Dani Lischinski, Arash Vahdat, and Weili Nie.
\newblock Diffuhaul: A training-free method for object dragging in images.
\newblock \emph{arXiv preprint arXiv:2406.01594}, 2024{\natexlab{a}}.

\bibitem[Avrahami et~al.(2024{\natexlab{b}})Avrahami, Hertz, Vinker, Arar, Fruchter, Fried, Cohen-Or, and Lischinski]{avrahami2024chosen}
Omri Avrahami, Amir Hertz, Yael Vinker, Moab Arar, Shlomi Fruchter, Ohad Fried, Daniel Cohen-Or, and Dani Lischinski.
\newblock The chosen one: Consistent characters in text-to-image diffusion models.
\newblock In \emph{ACM SIGGRAPH 2024 Conference Papers}, New York, NY, USA, 2024{\natexlab{b}}. Association for Computing Machinery.

\bibitem[Bar-Tal et~al.(2022)Bar-Tal, Ofri-Amar, Fridman, Kasten, and Dekel]{BarTal2022Text2LIVETL}
Omer Bar-Tal, Dolev Ofri-Amar, Rafail Fridman, Yoni Kasten, and Tali Dekel.
\newblock Text2live: Text-driven layered image and video editing.
\newblock \emph{ArXiv}, abs/2204.02491, 2022.

\bibitem[Bau et~al.(2019)Bau, Strobelt, Peebles, Wulff, Zhou, Zhu, and Torralba]{Bau2019SemanticPM}
David Bau, Hendrik Strobelt, William~S. Peebles, Jonas Wulff, Bolei Zhou, Jun-Yan Zhu, and Antonio Torralba.
\newblock Semantic photo manipulation with a generative image prior.
\newblock \emph{ACM Transactions on Graphics (TOG)}, 38:\penalty0 1 -- 11, 2019.

\bibitem[Bau et~al.(2021)Bau, Andonian, Cui, Park, Jahanian, Oliva, and Torralba]{Bau2021PaintBW}
David Bau, Alex Andonian, Audrey Cui, YeonHwan Park, Ali Jahanian, Aude Oliva, and Antonio Torralba.
\newblock Paint by word.
\newblock \emph{ArXiv}, abs/2103.10951, 2021.

\bibitem[Brack et~al.(2023)Brack, Friedrich, Kornmeier, Tsaban, Schramowski, Kersting, and Passos]{Brack2023LEDITSLI}
Manuel Brack, Felix Friedrich, Katharina Kornmeier, Linoy Tsaban, Patrick Schramowski, Kristian Kersting, and Apolin'ario Passos.
\newblock Ledits++: Limitless image editing using text-to-image models.
\newblock \emph{2024 IEEE/CVF Conference on Computer Vision and Pattern Recognition (CVPR)}, pages 8861--8870, 2023.

\bibitem[Brooks et~al.(2023)Brooks, Holynski, and Efros]{brooks2022instructpix2pix}
Tim Brooks, Aleksander Holynski, and Alexei~A. Efros.
\newblock Instructpix2pix: Learning to follow image editing instructions.
\newblock In \emph{CVPR}, 2023.

\bibitem[Cao et~al.(2023)Cao, Wang, Qi, Shan, Qie, and Zheng]{cao2023masactrl}
Mingdeng Cao, Xintao Wang, Zhongang Qi, Ying Shan, Xiaohu Qie, and Yinqiang Zheng.
\newblock {MasaCtrl:} tuning-free mutual self-attention control for consistent image synthesis and editing.
\newblock In \emph{Proceedings of the IEEE/CVF International Conference on Computer Vision (ICCV)}, pages 22560--22570, 2023.

\bibitem[Caron et~al.(2021)Caron, Touvron, Misra, Jegou, Mairal, Bojanowski, and Joulin]{Caron2021EmergingPI}
Mathilde Caron, Hugo Touvron, Ishan Misra, {Herv\'{e}} Jegou, Julien Mairal, Piotr Bojanowski, and Armand Joulin.
\newblock Emerging properties in self-supervised vision transformers.
\newblock In \emph{2021 IEEE/CVF International Conference on Computer Vision (ICCV)}, pages 9630--9640, 2021.

\bibitem[Chefer et~al.(2024)Chefer, Zada, Paiss, Ephrat, Tov, Rubinstein, Wolf, Dekel, Michaeli, and Mosseri]{Chefer2024StillMovingCV}
Hila Chefer, Shiran Zada, Roni Paiss, Ariel Ephrat, Omer Tov, Michael Rubinstein, Lior Wolf, Tali Dekel, Tomer Michaeli, and Inbar Mosseri.
\newblock Still-moving: Customized video generation without customized video data.
\newblock \emph{ArXiv}, abs/2407.08674, 2024.

\bibitem[Couairon et~al.(2022)Couairon, Verbeek, Schwenk, and Cord]{couairon2022diffedit}
Guillaume Couairon, Jakob Verbeek, Holger Schwenk, and Matthieu Cord.
\newblock Diffedit: Diffusion-based semantic image editing with mask guidance.
\newblock In \emph{The Eleventh International Conference on Learning Representations}, 2022.

\bibitem[Dahary et~al.(2024)Dahary, Patashnik, Aberman, and Cohen-Or]{Dahary2024BeYB}
Omer Dahary, Or Patashnik, Kfir Aberman, and Daniel Cohen-Or.
\newblock Be yourself: Bounded attention for multi-subject text-to-image generation.
\newblock \emph{ArXiv}, abs/2403.16990, 2024.

\bibitem[Deutch et~al.(2024)Deutch, Gal, Garibi, Patashnik, and Cohen-Or]{Deutch2024TurboEditTI}
Gilad Deutch, Rinon Gal, Daniel Garibi, Or Patashnik, and Daniel Cohen-Or.
\newblock Turboedit: Text-based image editing using few-step diffusion models.
\newblock \emph{ArXiv}, abs/2408.00735, 2024.

\bibitem[Dinh et~al.(2021)Dinh, Tran, Nguyen, and Hua]{Dinh2021HyperInverterIS}
Tan~M. Dinh, A. Tran, Rang Ho~Man Nguyen, and Binh-Son Hua.
\newblock Hyperinverter: Improving stylegan inversion via hypernetwork.
\newblock \emph{2022 IEEE/CVF Conference on Computer Vision and Pattern Recognition (CVPR)}, pages 11379--11388, 2021.

\bibitem[Esser et~al.(2024)Esser, Kulal, Blattmann, Entezari, Muller, Saini, Levi, Lorenz, Sauer, Boesel, Podell, Dockhorn, English, Lacey, Goodwin, Marek, and Rombach]{Esser2024ScalingRF}
Patrick Esser, Sumith Kulal, A. Blattmann, Rahim Entezari, Jonas Muller, Harry Saini, Yam Levi, Dominik Lorenz, Axel Sauer, Frederic Boesel, Dustin Podell, Tim Dockhorn, Zion English, Kyle Lacey, Alex Goodwin, Yannik Marek, and Robin Rombach.
\newblock Scaling rectified flow transformers for high-resolution image synthesis.
\newblock \emph{ArXiv}, abs/2403.03206, 2024.

\bibitem[Fischer et~al.(2023)Fischer, Gui, Ma, Stracke, Baumann, Hu, and Ommer]{Fischer2023BoostingLD}
Johannes~S. Fischer, Ming Gui, Pingchuan Ma, Nick Stracke, Stefan~Andreas Baumann, Vincent~Tao Hu, and Bjorn Ommer.
\newblock Boosting latent diffusion with flow matching.
\newblock \emph{ArXiv}, abs/2312.07360, 2023.

\bibitem[Frenkel et~al.(2024)Frenkel, Vinker, Shamir, and Cohen-Or]{Frenkel2024ImplicitSS}
Yarden Frenkel, Yael Vinker, Ariel Shamir, and Daniel Cohen-Or.
\newblock Implicit style-content separation using b-lora.
\newblock \emph{ArXiv}, abs/2403.14572, 2024.

\bibitem[Gal et~al.(2021)Gal, Patashnik, Maron, Bermano, Chechik, and Cohen-Or]{Gal2021StyleGANNADA}
Rinon Gal, Or Patashnik, Haggai Maron, Amit~H. Bermano, Gal Chechik, and Daniel Cohen-Or.
\newblock Stylegan-nada.
\newblock \emph{ACM Transactions on Graphics (TOG)}, 41:\penalty0 1 -- 13, 2021.

\bibitem[Garibi et~al.(2024)Garibi, Patashnik, Voynov, Averbuch-Elor, and Cohen-Or]{Garibi2024ReNoiseRI}
Daniel Garibi, Or Patashnik, Andrey Voynov, Hadar Averbuch-Elor, and Daniel Cohen-Or.
\newblock Renoise: Real image inversion through iterative noising.
\newblock \emph{ArXiv}, abs/2403.14602, 2024.

\bibitem[Geyer et~al.(2023)Geyer, Bar-Tal, Bagon, and Dekel]{geyer2023tokenflow}
Michal Geyer, Omer Bar-Tal, Shai Bagon, and Tali Dekel.
\newblock Tokenflow: Consistent diffusion features for consistent video editing.
\newblock \emph{arXiv preprint arXiv:2307.10373}, 2023.

\bibitem[Goodfellow et~al.(2014)Goodfellow, Pouget-Abadie, Mirza, Xu, Warde-Farley, Ozair, Courville, and Bengio]{goodfellow2014generative}
Ian Goodfellow, Jean Pouget-Abadie, Mehdi Mirza, Bing Xu, David Warde-Farley, Sherjil Ozair, Aaron Courville, and Yoshua Bengio.
\newblock Generative adversarial nets.
\newblock \emph{Advances in neural information processing systems}, 27, 2014.

\bibitem[Han et~al.(2023)Han, Wen, Chen, Zhang, Song, Ren, Gao, Chen, Liu, Zhangli, Stathopoulos, He, Jiang, Xia, Srivastava, and Metaxas]{Han2023ProxEditIT}
Ligong Han, Song Wen, Qi Chen, Zhixing Zhang, Kunpeng Song, Mengwei Ren, Ruijiang Gao, Yuxiao Chen, Ding Liu, Qilong Zhangli, Anastasis Stathopoulos, Xiaoxiao He, Jindong Jiang, Zhaoyang Xia, Akash Srivastava, and Dimitris~N. Metaxas.
\newblock Proxedit: Improving tuning-free real image editing with proximal guidance.
\newblock \emph{2024 IEEE/CVF Winter Conference on Applications of Computer Vision (WACV)}, pages 4279--4289, 2023.

\bibitem[He et~al.(2015)He, Zhang, Ren, and Sun]{He2015DeepRL}
Kaiming He, X. Zhang, Shaoqing Ren, and Jian Sun.
\newblock Deep residual learning for image recognition.
\newblock \emph{2016 IEEE Conference on Computer Vision and Pattern Recognition (CVPR)}, pages 770--778, 2015.

\bibitem[Hertz et~al.(2022)Hertz, Mokady, Tenenbaum, Aberman, Pritch, and Cohen-or]{Hertz2022PrompttoPromptIE}
Amir Hertz, Ron Mokady, Jay Tenenbaum, Kfir Aberman, Yael Pritch, and Daniel Cohen-or.
\newblock Prompt-to-prompt image editing with cross-attention control.
\newblock In \emph{The Eleventh International Conference on Learning Representations}, 2022.

\bibitem[Hertz et~al.(2023)Hertz, Voynov, Fruchter, and Cohen-Or]{Hertz2023StyleAI}
Amir Hertz, Andrey Voynov, Shlomi Fruchter, and Daniel Cohen-Or.
\newblock Style aligned image generation via shared attention.
\newblock \emph{2024 IEEE/CVF Conference on Computer Vision and Pattern Recognition (CVPR)}, pages 4775--4785, 2023.

\bibitem[Ho et~al.(2020)Ho, Jain, and Abbeel]{ho2020denoising}
Jonathan Ho, Ajay Jain, and Pieter Abbeel.
\newblock Denoising diffusion probabilistic models.
\newblock In \emph{Proc.~NeurIPS}, 2020.

\bibitem[Horwitz et~al.(2024)Horwitz, Kahana, and Hoshen]{Horwitz2024RecoveringTP}
Eliahu Horwitz, Jonathan Kahana, and Yedid Hoshen.
\newblock Recovering the pre-fine-tuning weights of generative models.
\newblock \emph{ArXiv}, abs/2402.10208, 2024.

\bibitem[Huang et~al.(2023)Huang, Tang, Dong, Lee, and Xu]{Huang2023RegionAwareDF}
Nisha Huang, Fan Tang, Weiming Dong, Tong-Yee Lee, and Changsheng Xu.
\newblock Region-aware diffusion for zero-shot text-driven image editing.
\newblock \emph{ArXiv}, abs/2302.11797, 2023.

\bibitem[Huang et~al.(2024)Huang, Huang, Liu, Yan, Lv, Liu, Xiong, Zhang, Chen, and Cao]{Huang2024DiffusionMI}
Yi Huang, Jiancheng Huang, Yifan Liu, Mingfu Yan, Jiaxi Lv, Jianzhuang Liu, Wei Xiong, He Zhang, Shifeng Chen, and Liangliang Cao.
\newblock Diffusion model-based image editing: A survey.
\newblock \emph{ArXiv}, abs/2402.17525, 2024.

\bibitem[Huberman-Spiegelglas et~al.(2023)Huberman-Spiegelglas, Kulikov, and Michaeli]{huberman2023edit}
Inbar Huberman-Spiegelglas, Vladimir Kulikov, and Tomer Michaeli.
\newblock An edit friendly ddpm noise space: Inversion and manipulations.
\newblock \emph{arXiv e-prints}, pages arXiv--2304, 2023.

\bibitem[Jeong et~al.(2024)Jeong, Kim, Choi, Lee, and Uh]{Jeong2024VisualSP}
Jaeseok Jeong, Junho Kim, Yunjey Choi, Gayoung Lee, and Youngjung Uh.
\newblock Visual style prompting with swapping self-attention.
\newblock \emph{ArXiv}, abs/2402.12974, 2024.

\bibitem[Johnson et~al.(2016)Johnson, Alahi, and Fei-Fei]{Johnson2016PerceptualLF}
Justin Johnson, Alexandre Alahi, and Li Fei-Fei.
\newblock Perceptual losses for real-time style transfer and super-resolution.
\newblock \emph{ArXiv}, abs/1603.08155, 2016.

\bibitem[Karras et~al.(2019)Karras, Laine, and Aila]{karras2019style}
Tero Karras, Samuli Laine, and Timo Aila.
\newblock A style-based generator architecture for generative adversarial networks.
\newblock In \emph{Proceedings of the IEEE conference on computer vision and pattern recognition}, pages 4401--4410, 2019.

\bibitem[Karras et~al.(2020)Karras, Laine, Aittala, Hellsten, Lehtinen, and Aila]{karras2020analyzing}
Tero Karras, Samuli Laine, Miika Aittala, Janne Hellsten, Jaakko Lehtinen, and Timo Aila.
\newblock Analyzing and improving the image quality of stylegan.
\newblock In \emph{Proceedings of the IEEE/CVF Conference on Computer Vision and Pattern Recognition}, pages 8110--8119, 2020.

\bibitem[Kawar et~al.(2023)Kawar, Zada, Lang, Tov, Chang, Dekel, Mosseri, and Irani]{Kawar2022ImagicTR}
Bahjat Kawar, Shiran Zada, Oran Lang, Omer Tov, Huiwen Chang, Tali Dekel, Inbar Mosseri, and Michal Irani.
\newblock Imagic: Text-based real image editing with diffusion models.
\newblock In \emph{Proceedings of the IEEE/CVF Conference on Computer Vision and Pattern Recognition}, pages 6007--6017, 2023.

\bibitem[Labs(2024)]{flux}
Black~Forest Labs.
\newblock Flux.
\newblock \url{https://github.com/black-forest-labs/flux}, 2024.

\bibitem[Li et~al.(2023)Li, Zeng, Feng, Gao, Liu, Liu, Lin, Tang, Hu, Liu, and Zhang]{Li2023ZONEZI}
Shanglin Li, Bo-Wen Zeng, Yutang Feng, Sicheng Gao, Xuhui Liu, Jiaming Liu, Li Lin, Xu Tang, Yao Hu, Jianzhuang Liu, and Baochang Zhang.
\newblock Zone: Zero-shot instruction-guided local editing.
\newblock \emph{2024 IEEE/CVF Conference on Computer Vision and Pattern Recognition (CVPR)}, pages 6254--6263, 2023.

\bibitem[Lin et~al.(2014)Lin, Maire, Belongie, Hays, Perona, Ramanan, Doll{\'a}r, and Zitnick]{Lin2014MicrosoftCC}
Tsung-Yi Lin, Michael Maire, Serge~J. Belongie, James Hays, Pietro Perona, Deva Ramanan, Piotr Doll{\'a}r, and C.~Lawrence Zitnick.
\newblock Microsoft coco: Common objects in context.
\newblock In \emph{European Conference on Computer Vision}, 2014.

\bibitem[Lipman et~al.(2022)Lipman, Chen, Ben-Hamu, Nickel, and Le]{Lipman2022FlowMF}
Yaron Lipman, Ricky T.~Q. Chen, Heli Ben-Hamu, Maximilian Nickel, and Matt Le.
\newblock Flow matching for generative modeling.
\newblock \emph{ArXiv}, abs/2210.02747, 2022.

\bibitem[Liu et~al.(2022)Liu, Gong, and Liu]{Liu2022FlowSA}
Xingchao Liu, Chengyue Gong, and Qiang Liu.
\newblock Flow straight and fast: Learning to generate and transfer data with rectified flow.
\newblock \emph{ArXiv}, abs/2209.03003, 2022.

\bibitem[Meiri et~al.(2023)Meiri, Samuel, Darshan, Chechik, Avidan, and Ben-Ari]{meiri2023fixed}
Barak Meiri, Dvir Samuel, Nir Darshan, Gal Chechik, Shai Avidan, and Rami Ben-Ari.
\newblock Fixed-point inversion for text-to-image diffusion models.
\newblock \emph{arXiv preprint arXiv:2312.12540}, 2023.

\bibitem[Meng et~al.(2021)Meng, He, Song, Song, Wu, Zhu, and Ermon]{meng2021sdedit}
Chenlin Meng, Yutong He, Yang Song, Jiaming Song, Jiajun Wu, Jun-Yan Zhu, and Stefano Ermon.
\newblock Sdedit: Guided image synthesis and editing with stochastic differential equations.
\newblock In \emph{International Conference on Learning Representations}, 2021.

\bibitem[Mokady et~al.(2023)Mokady, Hertz, Aberman, Pritch, and Cohen-Or]{mokady2022null}
Ron Mokady, Amir Hertz, Kfir Aberman, Yael Pritch, and Daniel Cohen-Or.
\newblock Null-text inversion for editing real images using guided diffusion models.
\newblock In \emph{Proceedings of the IEEE/CVF Conference on Computer Vision and Pattern Recognition}, pages 6038--6047, 2023.

\bibitem[Nichol et~al.(2021)Nichol, Dhariwal, Ramesh, Shyam, Mishkin, McGrew, Sutskever, and Chen]{Nichol2021GLIDETP}
Alex Nichol, Prafulla Dhariwal, Aditya Ramesh, Pranav Shyam, Pamela Mishkin, Bob McGrew, Ilya Sutskever, and Mark Chen.
\newblock Glide: Towards photorealistic image generation and editing with text-guided diffusion models.
\newblock In \emph{International Conference on Machine Learning}, 2021.

\bibitem[Nitzan et~al.(2024)Nitzan, Wu, Zhang, Shechtman, Cohen-Or, Park, and Gharbi]{Nitzan2024LazyDT}
Yotam Nitzan, Zongze Wu, Richard Zhang, Eli Shechtman, Daniel Cohen-Or, Taesung Park, and Michael Gharbi.
\newblock Lazy diffusion transformer for interactive image editing.
\newblock \emph{ArXiv}, abs/2404.12382, 2024.

\bibitem[OpenAI(2022)]{chatgpt}
OpenAI.
\newblock {ChatGPT}.
\newblock \url{https://chat.openai.com/}, 2022.
\newblock Accessed: 2024-10-1.

\bibitem[Oquab et~al.(2023)Oquab, Darcet, Moutakanni, Vo, Szafraniec, Khalidov, Fernandez, Haziza, Massa, El-Nouby, Assran, Ballas, Galuba, Howes, Huang, Li, Misra, Rabbat, Sharma, Synnaeve, Xu, J{\'e}gou, Mairal, Labatut, Joulin, and Bojanowski]{Oquab2023DINOv2LR}
Maxime Oquab, Timoth{\'e}e Darcet, Th{\'e}o Moutakanni, Huy~Q. Vo, Marc Szafraniec, Vasil Khalidov, Pierre Fernandez, Daniel Haziza, Francisco Massa, Alaaeldin El-Nouby, Mahmoud Assran, Nicolas Ballas, Wojciech Galuba, Russ Howes, Po-Yao~(Bernie) Huang, Shang-Wen Li, Ishan Misra, Michael~G. Rabbat, Vasu Sharma, Gabriel Synnaeve, Huijiao Xu, Herv{\'e} J{\'e}gou, Julien Mairal, Patrick Labatut, Armand Joulin, and Piotr Bojanowski.
\newblock {DINOv2}: Learning robust visual features without supervision.
\newblock \emph{ArXiv}, abs/2304.07193, 2023.

\bibitem[Pan et~al.(2023)Pan, Gherardi, Xie, and Huang]{Pan2023EffectiveRI}
Zhihong Pan, Riccardo Gherardi, Xiufeng Xie, and Stephen Huang.
\newblock Effective real image editing with accelerated iterative diffusion inversion.
\newblock \emph{2023 IEEE/CVF International Conference on Computer Vision (ICCV)}, pages 15866--15875, 2023.

\bibitem[Parmar et~al.(2022)Parmar, Li, Lu, Zhang, Zhu, and Singh]{Parmar2022SpatiallyAdaptiveMS}
Gaurav Parmar, Yijun Li, Jingwan Lu, Richard Zhang, Jun-Yan Zhu, and Krishna~Kumar Singh.
\newblock Spatially-adaptive multilayer selection for gan inversion and editing.
\newblock \emph{2022 IEEE/CVF Conference on Computer Vision and Pattern Recognition (CVPR)}, pages 11389--11399, 2022.

\bibitem[Parmar et~al.(2023)Parmar, Singh, Zhang, Li, Lu, and Zhu]{Parmar2023ZeroshotIT}
Gaurav Parmar, Krishna~Kumar Singh, Richard Zhang, Yijun Li, Jingwan Lu, and Jun-Yan Zhu.
\newblock Zero-shot image-to-image translation.
\newblock \emph{ACM SIGGRAPH 2023 Conference Proceedings}, 2023.

\bibitem[Patashnik et~al.(2021)Patashnik, Wu, Shechtman, Cohen-Or, and Lischinski]{Patashnik2021StyleCLIPTM}
Or Patashnik, Zongze Wu, Eli Shechtman, Daniel Cohen-Or, and Dani Lischinski.
\newblock Styleclip: Text-driven manipulation of stylegan imagery.
\newblock \emph{2021 IEEE/CVF International Conference on Computer Vision (ICCV)}, pages 2065--2074, 2021.

\bibitem[Patashnik et~al.(2023)Patashnik, Garibi, Azuri, Averbuch-Elor, and Cohen-Or]{Patashnik2023LocalizingOS}
Or Patashnik, Daniel Garibi, Idan Azuri, Hadar Averbuch-Elor, and Daniel Cohen-Or.
\newblock Localizing object-level shape variations with text-to-image diffusion models.
\newblock \emph{2023 IEEE/CVF International Conference on Computer Vision (ICCV)}, pages 22994--23004, 2023.

\bibitem[Peebles and Xie(2022)]{Peebles2022ScalableDM}
William~S. Peebles and Saining Xie.
\newblock Scalable diffusion models with transformers.
\newblock \emph{2023 IEEE/CVF International Conference on Computer Vision (ICCV)}, pages 4172--4182, 2022.

\bibitem[Pidhorskyi et~al.(2020)Pidhorskyi, Adjeroh, and Doretto]{Pidhorskyi2020AdversarialLA}
Stanislav Pidhorskyi, Donald~A. Adjeroh, and Gianfranco Doretto.
\newblock Adversarial latent autoencoders.
\newblock \emph{2020 IEEE/CVF Conference on Computer Vision and Pattern Recognition (CVPR)}, pages 14092--14101, 2020.

\bibitem[Po et~al.(2023)Po, Yifan, Golyanik, Aberman, Barron, Bermano, Chan, Dekel, Holynski, Kanazawa, Liu, Liu, Mildenhall, Nie{\ss}ner, Ommer, Theobalt, Wonka, and Wetzstein]{Po2023StateOT}
Ryan Po, Wang Yifan, Vladislav Golyanik, Kfir Aberman, Jonathan~T. Barron, Amit~H. Bermano, Eric~Ryan Chan, Tali Dekel, Aleksander Holynski, Angjoo Kanazawa, C.~Karen Liu, Lingjie Liu, Ben Mildenhall, Matthias Nie{\ss}ner, Bjorn Ommer, Christian Theobalt, Peter Wonka, and Gordon Wetzstein.
\newblock State of the art on diffusion models for visual computing.
\newblock \emph{ArXiv}, abs/2310.07204, 2023.

\bibitem[Podell et~al.(2023)Podell, English, Lacey, Blattmann, Dockhorn, Muller, Penna, and Rombach]{Podell2023SDXLIL}
Dustin Podell, Zion English, Kyle Lacey, A. Blattmann, Tim Dockhorn, Jonas Muller, Joe Penna, and Robin Rombach.
\newblock {SDXL}: Improving latent diffusion models for high-resolution image synthesis.
\newblock \emph{ArXiv}, abs/2307.01952, 2023.

\bibitem[Radford et~al.(2021)Radford, Kim, Hallacy, Ramesh, Goh, Agarwal, Sastry, Askell, Mishkin, Clark, Krueger, and Sutskever]{Radford2021LearningTV}
Alec Radford, Jong~Wook Kim, Chris Hallacy, Aditya Ramesh, Gabriel Goh, Sandhini Agarwal, Girish Sastry, Amanda Askell, Pamela Mishkin, Jack Clark, Gretchen Krueger, and Ilya Sutskever.
\newblock Learning transferable visual models from natural language supervision.
\newblock In \emph{International Conference on Machine Learning}, 2021.

\bibitem[Ramesh et~al.(2022)Ramesh, Dhariwal, Nichol, Chu, and Chen]{ramesh2022hierarchical}
Aditya Ramesh, Prafulla Dhariwal, Alex Nichol, Casey Chu, and Mark Chen.
\newblock Hierarchical text-conditional image generation with {CLIP} latents.
\newblock \emph{arXiv preprint arXiv:2204.06125}, 2022.

\bibitem[Regev et~al.(2024)Regev, Avrahami, and Lischinski]{Regev2024Click2MaskLE}
Omer Regev, Omri Avrahami, and Dani Lischinski.
\newblock Click2mask: Local editing with dynamic mask generation.
\newblock \emph{ArXiv}, abs/2409.08272, 2024.

\bibitem[Richardson et~al.(2020)Richardson, Alaluf, Patashnik, Nitzan, Azar, Shapiro, and Cohen-Or]{Richardson2020EncodingIS}
Elad Richardson, Yuval Alaluf, Or Patashnik, Yotam Nitzan, Yaniv Azar, Stav Shapiro, and Daniel Cohen-Or.
\newblock Encoding in style: a stylegan encoder for image-to-image translation.
\newblock \emph{2021 IEEE/CVF Conference on Computer Vision and Pattern Recognition (CVPR)}, pages 2287--2296, 2020.

\bibitem[Roich et~al.(2021)Roich, Mokady, Bermano, and Cohen-Or]{Roich2021PivotalTF}
Daniel Roich, Ron Mokady, Amit~H. Bermano, and Daniel Cohen-Or.
\newblock Pivotal tuning for latent-based editing of real images.
\newblock \emph{ACM Transactions on Graphics (TOG)}, 42:\penalty0 1 -- 13, 2021.

\bibitem[Rombach et~al.(2021)Rombach, Blattmann, Lorenz, Esser, and Ommer]{Rombach2021HighResolutionIS}
Robin Rombach, A. Blattmann, Dominik Lorenz, Patrick Esser, and Bj{\"o}rn Ommer.
\newblock High-resolution image synthesis with latent diffusion models.
\newblock \emph{2022 IEEE/CVF Conference on Computer Vision and Pattern Recognition (CVPR)}, pages 10674--10685, 2021.

\bibitem[Ronneberger et~al.(2015)Ronneberger, Fischer, and Brox]{Ronneberger2015UNetCN}
Olaf Ronneberger, Philipp Fischer, and Thomas Brox.
\newblock U-net: Convolutional networks for biomedical image segmentation.
\newblock \emph{ArXiv}, abs/1505.04597, 2015.

\bibitem[Saharia et~al.(2022)Saharia, Chan, Saxena, Li, Whang, Denton, Ghasemipour, Gontijo~Lopes, Karagol~Ayan, Salimans, et~al.]{Saharia2022PhotorealisticTD}
Chitwan Saharia, William Chan, Saurabh Saxena, Lala Li, Jay Whang, Emily~L Denton, Kamyar Ghasemipour, Raphael Gontijo~Lopes, Burcu Karagol~Ayan, Tim Salimans, et~al.
\newblock Photorealistic text-to-image diffusion models with deep language understanding.
\newblock \emph{Advances in Neural Information Processing Systems}, 35:\penalty0 36479--36494, 2022.

\bibitem[Salama et~al.(2024)Salama, Kahana, Horwitz, and Hoshen]{Salama2024DatasetSR}
Mohammad Salama, Jonathan Kahana, Eliahu Horwitz, and Yedid Hoshen.
\newblock Dataset size recovery from lora weights.
\newblock \emph{ArXiv}, abs/2406.19395, 2024.

\bibitem[Sheynin et~al.(2023)Sheynin, Polyak, Singer, Kirstain, Zohar, Ashual, Parikh, and Taigman]{Sheynin2023EmuEP}
Shelly Sheynin, Adam Polyak, Uriel Singer, Yuval Kirstain, Amit Zohar, Oron Ashual, Devi Parikh, and Yaniv Taigman.
\newblock Emu edit: Precise image editing via recognition and generation tasks.
\newblock \emph{ArXiv}, abs/2311.10089, 2023.

\bibitem[Simonyan et~al.(2013)Simonyan, Vedaldi, and Zisserman]{Simonyan2013DeepIC}
Karen Simonyan, Andrea Vedaldi, and Andrew Zisserman.
\newblock Deep inside convolutional networks: Visualising image classification models and saliency maps.
\newblock \emph{CoRR}, abs/1312.6034, 2013.

\bibitem[Sohl-Dickstein et~al.(2015)Sohl-Dickstein, Weiss, Maheswaranathan, and Ganguli]{sohl2015deep}
Jascha Sohl-Dickstein, Eric Weiss, Niru Maheswaranathan, and Surya Ganguli.
\newblock Deep unsupervised learning using nonequilibrium thermodynamics.
\newblock In \emph{International Conference on Machine Learning}, pages 2256--2265. PMLR, 2015.

\bibitem[Song et~al.(2020)Song, Meng, and Ermon]{song2020denoising}
Jiaming Song, Chenlin Meng, and Stefano Ermon.
\newblock Denoising diffusion implicit models.
\newblock In \emph{International Conference on Learning Representations}, 2020.

\bibitem[Song and Ermon(2019)]{song2019generative}
Yang Song and Stefano Ermon.
\newblock Generative modeling by estimating gradients of the data distribution.
\newblock \emph{Advances in Neural Information Processing Systems}, 32, 2019.

\bibitem[Tewel et~al.(2024)Tewel, Kaduri, Gal, Kasten, Wolf, Chechik, and Atzmon]{Tewel2024TrainingFreeCT}
Yoad Tewel, Omri Kaduri, Rinon Gal, Yoni Kasten, Lior Wolf, Gal Chechik, and Yuval Atzmon.
\newblock Training-free consistent text-to-image generation.
\newblock \emph{ArXiv}, abs/2402.03286, 2024.

\bibitem[Tov et~al.(2021)Tov, Alaluf, Nitzan, Patashnik, and Cohen-Or]{Tov2021DesigningAE}
Omer Tov, Yuval Alaluf, Yotam Nitzan, Or Patashnik, and Daniel Cohen-Or.
\newblock Designing an encoder for stylegan image manipulation.
\newblock \emph{ACM Transactions on Graphics (TOG)}, 40:\penalty0 1 -- 14, 2021.

\bibitem[Tumanyan et~al.(2023)Tumanyan, Geyer, Bagon, and Dekel]{pnpDiffusion2022}
Narek Tumanyan, Michal Geyer, Shai Bagon, and Tali Dekel.
\newblock Plug-and-play diffusion features for text-driven image-to-image translation.
\newblock In \emph{Proceedings of the IEEE/CVF Conference on Computer Vision and Pattern Recognition}, pages 1921--1930, 2023.

\bibitem[Valevski et~al.(2022)Valevski, Kalman, Matias, and Leviathan]{valevski2022unitune}
Dani Valevski, Matan Kalman, Yossi Matias, and Yaniv Leviathan.
\newblock Unitune: Text-driven image editing by fine tuning an image generation model on a single image.
\newblock \emph{arXiv preprint arXiv:2210.09477}, 2022.

\bibitem[von Platen et~al.(2022)von Platen, Patil, Lozhkov, Cuenca, Lambert, Rasul, Davaadorj, and Wolf]{von-platen-etal-2022-diffusers}
Patrick von Platen, Suraj Patil, Anton Lozhkov, Pedro Cuenca, Nathan Lambert, Kashif Rasul, Mishig Davaadorj, and Thomas Wolf.
\newblock Diffusers: State-of-the-art diffusion models.
\newblock \url{https://github.com/huggingface/diffusers}, 2022.

\bibitem[Wallace et~al.(2022)Wallace, Gokul, and Naik]{Wallace2022EDICTED}
Bram Wallace, Akash Gokul, and Nikhil~Vijay Naik.
\newblock Edict: Exact diffusion inversion via coupled transformations.
\newblock \emph{2023 IEEE/CVF Conference on Computer Vision and Pattern Recognition (CVPR)}, pages 22532--22541, 2022.

\bibitem[Wang et~al.(2023{\natexlab{a}})Wang, Zhang, Birsak, and Wonka]{Wang2023InstructEditIA}
Qian Wang, Biao Zhang, Michael Birsak, and Peter Wonka.
\newblock Instructedit: Improving automatic masks for diffusion-based image editing with user instructions.
\newblock \emph{ArXiv}, abs/2305.18047, 2023{\natexlab{a}}.

\bibitem[Wang et~al.(2023{\natexlab{b}})Wang, Saharia, Montgomery, Pont-Tuset, Noy, Pellegrini, Onoe, Laszlo, Fleet, Soricut, et~al.]{wang2022imagen}
Su Wang, Chitwan Saharia, Ceslee Montgomery, Jordi Pont-Tuset, Shai Noy, Stefano Pellegrini, Yasumasa Onoe, Sarah Laszlo, David~J Fleet, Radu Soricut, et~al.
\newblock Imagen editor and editbench: Advancing and evaluating text-guided image inpainting.
\newblock In \emph{Proceedings of the IEEE/CVF Conference on Computer Vision and Pattern Recognition}, pages 18359--18369, 2023{\natexlab{b}}.

\bibitem[Wolf et~al.(2020)Wolf, Debut, Sanh, Chaumond, Delangue, Moi, Cistac, Rault, Louf, Funtowicz, Davison, Shleifer, von Platen, Ma, Jernite, Plu, Xu, Scao, Gugger, Drame, Lhoest, and Rush]{wolf-etal-2020-transformers}
Thomas Wolf, Lysandre Debut, Victor Sanh, Julien Chaumond, Clement Delangue, Anthony Moi, Pierric Cistac, Tim Rault, Rémi Louf, Morgan Funtowicz, Joe Davison, Sam Shleifer, Patrick von Platen, Clara Ma, Yacine Jernite, Julien Plu, Canwen Xu, Teven~Le Scao, Sylvain Gugger, Mariama Drame, Quentin Lhoest, and Alexander~M. Rush.
\newblock Transformers: State-of-the-art natural language processing.
\newblock In \emph{Proceedings of the 2020 Conference on Empirical Methods in Natural Language Processing: System Demonstrations}, pages 38--45, Online, 2020. Association for Computational Linguistics.

\bibitem[Wu et~al.(2022)Wu, Ge, Wang, Lei, Gu, Hsu, Shan, Qie, and Shou]{Wu2022TuneAVideoOT}
Jay~Zhangjie Wu, Yixiao Ge, Xintao Wang, Weixian Lei, Yuchao Gu, Wynne Hsu, Ying Shan, Xiaohu Qie, and Mike~Zheng Shou.
\newblock Tune-a-video: One-shot tuning of image diffusion models for text-to-video generation.
\newblock \emph{2023 IEEE/CVF International Conference on Computer Vision (ICCV)}, pages 7589--7599, 2022.

\bibitem[Xia et~al.(2021)Xia, Zhang, Yang, Xue, Zhou, and Yang]{Xia2021GANIA}
Weihao Xia, Yulun Zhang, Yujiu Yang, Jing-Hao Xue, Bolei Zhou, and Ming-Hsuan Yang.
\newblock Gan inversion: A survey.
\newblock \emph{IEEE Transactions on Pattern Analysis and Machine Intelligence}, 45:\penalty0 3121--3138, 2021.

\bibitem[Xie et~al.(2022)Xie, Zhang, Lin, Hinz, and Zhang]{Xie2022SmartBrushTA}
Shaoan Xie, Zhifei Zhang, Zhe Lin, Tobias Hinz, and Kun Zhang.
\newblock Smartbrush: Text and shape guided object inpainting with diffusion model.
\newblock \emph{2023 IEEE/CVF Conference on Computer Vision and Pattern Recognition (CVPR)}, pages 22428--22437, 2022.

\bibitem[Yang et~al.(2022)Yang, Gu, Zhang, Zhang, Chen, Sun, Chen, and Wen]{Yang2022PaintBE}
Binxin Yang, Shuyang Gu, Bo Zhang, Ting Zhang, Xuejin Chen, Xiaoyan Sun, Dong Chen, and Fang Wen.
\newblock Paint by example: Exemplar-based image editing with diffusion models.
\newblock \emph{2023 IEEE/CVF Conference on Computer Vision and Pattern Recognition (CVPR)}, pages 18381--18391, 2022.

\bibitem[Yang et~al.(2023)Yang, Gui, Wang, Chen, Zhuang, and Shen]{Yang2023ObjectawareIA}
Zhen Yang, Dinggang Gui, Wen Wang, Hao Chen, Bohan Zhuang, and Chunhua Shen.
\newblock Object-aware inversion and reassembly for image editing.
\newblock \emph{ArXiv}, abs/2310.12149, 2023.

\bibitem[Zeiler and Fergus(2013)]{Zeiler2013VisualizingAU}
Matthew~D. Zeiler and Rob Fergus.
\newblock Visualizing and understanding convolutional networks.
\newblock \emph{ArXiv}, abs/1311.2901, 2013.

\bibitem[Zhang et~al.(2023{\natexlab{a}})Zhang, Mo, Chen, Sun, and Su]{Zhang2023MagicBrush}
Kai Zhang, Lingbo Mo, Wenhu Chen, Huan Sun, and Yu Su.
\newblock Magicbrush: A manually annotated dataset for instruction-guided image editing.
\newblock In \emph{Advances in Neural Information Processing Systems}, 2023{\natexlab{a}}.

\bibitem[Zhang et~al.(2018)Zhang, Isola, Efros, Shechtman, and Wang]{Zhang2018TheUE}
Richard Zhang, Phillip Isola, Alexei~A. Efros, Eli Shechtman, and Oliver Wang.
\newblock The unreasonable effectiveness of deep features as a perceptual metric.
\newblock \emph{2018 IEEE/CVF Conference on Computer Vision and Pattern Recognition}, pages 586--595, 2018.

\bibitem[Zhang(2024)]{zhang2024fast}
Shiwen Zhang.
\newblock Fast imagic: Solving overfitting in text-guided image editing via disentangled {UN}et with forgetting mechanism and unified vision-language optimization.
\newblock In \emph{UniReps: 2nd Edition of the Workshop on Unifying Representations in Neural Models}, 2024.

\bibitem[Zhang et~al.(2023{\natexlab{b}})Zhang, Xiao, and Huang]{Zhang2023ForgeditTG}
Shiwen Zhang, Shuai Xiao, and Weilin Huang.
\newblock Forgedit: Text guided image editing via learning and forgetting.
\newblock \emph{ArXiv}, abs/2309.10556, 2023{\natexlab{b}}.

\bibitem[Zhu et~al.(2020{\natexlab{a}})Zhu, Shen, Zhao, and Zhou]{zhu2020domain}
Jiapeng Zhu, Yujun Shen, Deli Zhao, and Bolei Zhou.
\newblock In-domain gan inversion for real image editing.
\newblock In \emph{European conference on computer vision}, pages 592--608. Springer, 2020{\natexlab{a}}.

\bibitem[Zhu et~al.(2020{\natexlab{b}})Zhu, Abdal, Qin, and Wonka]{Zhu2020ImprovedSE}
Peihao Zhu, Rameen Abdal, Yipeng Qin, and Peter Wonka.
\newblock Improved stylegan embedding: Where are the good latents?
\newblock \emph{ArXiv}, abs/2012.09036, 2020{\natexlab{b}}.

\end{thebibliography}
}

\clearpage
\appendix
\twocolumn[
    \centering
    \Large
    \textbf{\thetitle} \\
    Supplementary Material \\
    \vspace{1.0em}
]

\section{Implementation Details}
\label{sec:implementation_details}

In \Cref{sec:method_details}, we start by providing implementation details for our method. Next, in \Cref{sec:baselines_details}, we provide the implementation details for the baselines we compared our method against. Later, in \Cref{sec:automatic_metrics_details}, we provide the implementation details for the automatic evaluations dataset and metrics. Finally, in \Cref{sec:user_study_details} we provide the full details of the user study we conducted.

\subsection{Method Implementation Details}
\label{sec:method_details}

\begin{figure*}[t]
    \centering
    \includegraphics[width=1\linewidth]{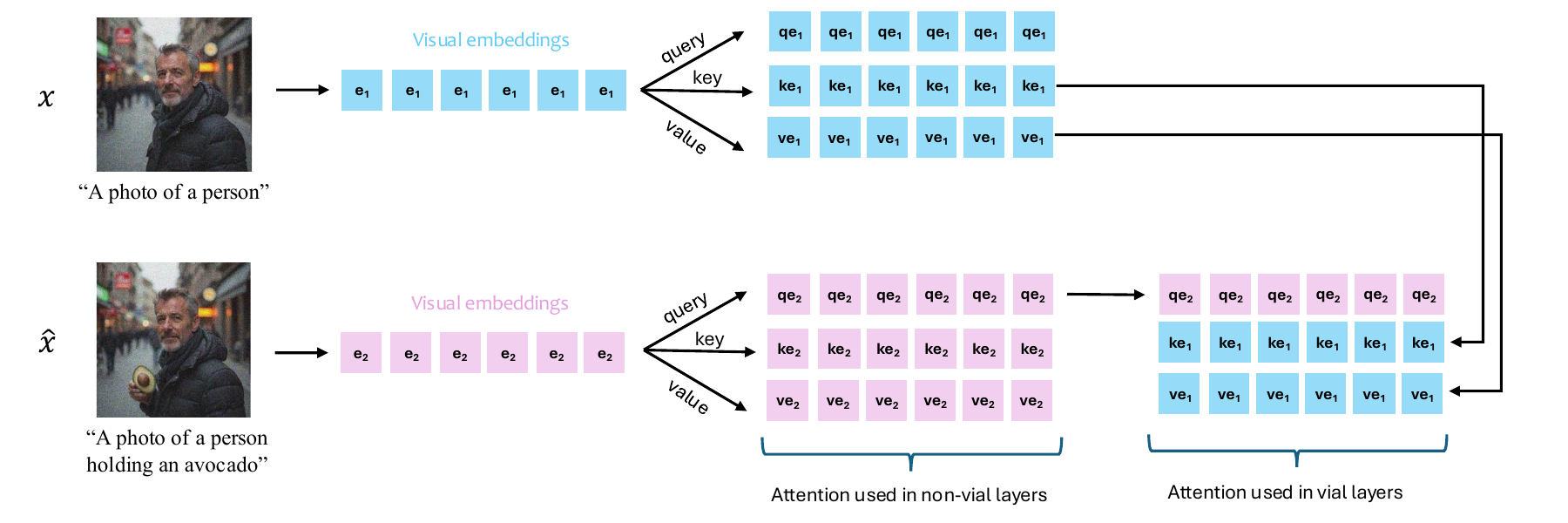}
    \caption{\textbf{Attention Injection.} We adapt the self-attention injection mechanism, previously shown effective for image editing in UNet-based diffusion models, to the DiT-based FLUX architecture. Since each DiT layer processes a sequence of image and text embeddings, we propose generating both the reference image $x$ and generated image $\hat{x}$ in parallel while \emph{selectively replacing} the attention \emph{keys and values} that correspond to the image embeddings of $\hat{x}$ with those of $x$. This replacement is performed only within the vital layers set.}
    \label{fig:attention_injection}
\end{figure*}

As described in 
\Cref{sec:layers_importance},
we started by collecting a dataset of $k=64$ text prompts using ChatGPT~\cite{chatgpt}. We instructed it to generate text prompts describing a diverse set of objects in different environments, with the focus on one main object. Then, we sampled $k$ seeds denoted by $S$ and used them to generate $k$ corresponding images $G_{\textit{ref}}$. Next, for each layer $l$, we bypass it by taking only the residual connection values. For each bypass, we generate $k$ images using the same seed set $S$ demoted by $G_{l}$. All the images were generated using Euler sampler in 15 steps and a guidance scale of $3.5$.

Next, to evaluate the effect of each layer $l$ on the final result, we compared the generated images $G_{l}$ with their corresponding images $G_{\textit{ref}}$ using the DINOv2~\cite{Oquab2023DINOv2LR} perceptual similarity metric. We term the layers that effect the generated image the most (\ie, the layers with the lowest perceptual similarity) as vital layers, while the rest of the layers as non-vital layers. We found that the vital layers in the FLUX.1-dev model~\cite{flux} are $[0, 1, 2, 17, 18, 25, 28, 53, 54, 56]$. For visualization results, please refer to \Cref{sec:layer_bypassing_visualization}. We empirically found that layer 2 can be removed from this set. In addition, the vital layers for the Stable Diffusion 3 (SD3)~\cite{Esser2024ScalingRF} model vital layers are: $[0, 7, 8, 9]$. For more details, please refer to \Cref{sec:sd3_results}.

In addition, as mentioned in
\Cref{sec:image_editing},
We adapt the self-attention injection mechanism, previously  to be effective for image and video editing~\cite{Wu2022TuneAVideoOT, cao2023masactrl} in UNet-based diffusion models, to the DiT-based FLUX architecture. Since each DiT layer processes a sequence of image and text embeddings, we propose generating both the reference image $x$ and generated image $\hat{x}$ in parallel while \emph{selectively replacing} the image embeddings of $\hat{x}$ with those of $x$, but only within the vital layers set. A full visualization can be found in \Cref{fig:attention_injection}.

Lastly, the variance list of the perceptual similarity of the different layers, as explained in
\Cref{sec:layers_importance},
is as follows: [0.222, 0.041, 0.076, 0.08, 0.123, 0.101, 0.135, 0.124, 0.112, 0.105, 0.097, 0.12, 0.118, 0.086, 0.116, 0.067, 0.065, 0.116, 0.146, 0.065, 0.098, 0.061, 0.076, 0.077, 0.072, 0.086, 0.069, 0.067, 0.081, 0.091, 0.074, 0.062, 0.061, 0.044, 0.04, 0.054, 0.036, 0.038, 0.037, 0.04, 0.066, 0.04, 0.034, 0.044, 0.044, 0.031, 0.033, 0.036, 0.03, 0.032, 0.026, 0.026, 0.026, 0.079, 0.039, 0.037, 0.026].

\subsection{Baselines Implementation Details}
\label{sec:baselines_details}

As explained in 
\Cref{sec:comparisons},
we compare our method against the following baselines: SDEdit~\cite{meng2021sdedit}, P2P+NTI~\cite{Hertz2022PrompttoPromptIE, mokady2022null}, Instruct-P2P~\cite{brooks2022instructpix2pix}, MagicBrush~\cite{Zhang2023MagicBrush}, and MasaCTRL~\cite{cao2023masactrl}. We reimplement SDEdit using the FLUX.1-dev model~\cite{flux}, and use the official implementation for the rest of the baselines.

We adapt the text prompts based on the baseline type: for SDEdit~\cite{meng2021sdedit}, P2P+NTI~\cite{Hertz2022PrompttoPromptIE, mokady2022null}, and MasaCTRL~\cite{cao2023masactrl}, we used the standard text prompt describing the desired edited scene (\eg, \prompt{A photo of a man with a red hat}). For the instruction-based baselines Instruct-P2P~\cite{brooks2022instructpix2pix} and MagicBrush~\cite{Zhang2023MagicBrush} we adapted the style to fit an instructional format (\eg, \prompt{Make the person wear a red hat}).

We used the following third-party implementations in this project:
\begin{itemize}
    \item \textbf{FLUX.1-dev} model~\cite{flux} HuggingFace Diffusers~\cite{von-platen-etal-2022-diffusers} implementation at \url{https://github.com/huggingface/diffusers}
    \item \textbf{P2P+NTI}~\cite{Hertz2022PrompttoPromptIE, mokady2022null} official implementation at \url{https://github.com/google/prompt-to-prompt}
    \item \textbf{Instruct-P2P}~\cite{brooks2022instructpix2pix} official implementation at \url{https://github.com/timothybrooks/instruct-pix2pix}
    \item \textbf{MagicBrush}~\cite{Zhang2023MagicBrush} official implementation at \url{https://github.com/OSU-NLP-Group/MagicBrush}
    \item \textbf{MasaCTRL}~\cite{cao2023masactrl} official implementation at \url{https://github.com/TencentARC/MasaCtrl}
    \item \textbf{DINOv2}~\cite{Oquab2023DINOv2LR} ViT-g/14 implementation by HuggingFace Transformers~\cite{wolf-etal-2020-transformers} at \url{https://github.com/huggingface/transformers}.
    \item \textbf{DINOv1}~\cite{Caron2021EmergingPI} ViT-B/16 implementation by HuggingFace Transformers~\cite{wolf-etal-2020-transformers} at \url{https://github.com/huggingface/transformers}.
    \item \textbf{CLIP}~\cite{Radford2021LearningTV} ViT-L/14 implementation by HuggingFace Transformers~\cite{wolf-etal-2020-transformers} implementation at \url{https://github.com/huggingface/transformers}
    \item \textbf{LPIPS}~\cite{Zhang2018TheUE} official implementation at \url{https://github.com/richzhang/PerceptualSimilarity}.
\end{itemize}

\subsection{Automatic Metrics Implementation Details}
\label{sec:automatic_metrics_details}

As explained in 
\Cref{sec:comparisons},
we prepare an evaluation dataset based on the COCO~\cite{Lin2014MicrosoftCC} validation dataset. We begin by filtering the dataset automatically to include at least one prominent non-rigid body. More specifically, we filter only images containing humans or animals that at least one of them is prominent enough, but not too small, \ie, the prominent non-rigid body occupies at least 5\% of the image but no more than 33\%. Next, for each image, we apply various image editing tasks (non-rigid editing, object addition, object replacement, and scene editing) that take into account the prominent object from a list of different combinations, resulting in a total dataset of 3,200 samples. Examples of images from this dataset can be seen in \Cref{fig:baselines_qualitative_comparison_automatic}.

We evaluate the editing results using three metrics: (1) \clipimg which measures the similarity between the input image and the edited image by calculating the normalized cosine similarity of their CLIP image embeddings. (2) \cliptxt which measures the similarity between the edited image and the target editing prompt by calculating the normalized cosine similarity between the CLIP image embedding and the target text CLIP embedding. (3) \clipdir~\cite{Patashnik2021StyleCLIPTM, Gal2021StyleGANNADA} which measures the similarity between the direction of the prompt change and the direction of the image change.

\subsection{User Study Details}
\label{sec:user_study_details}

\begin{figure}[th]
    \includegraphics[width=1\linewidth]{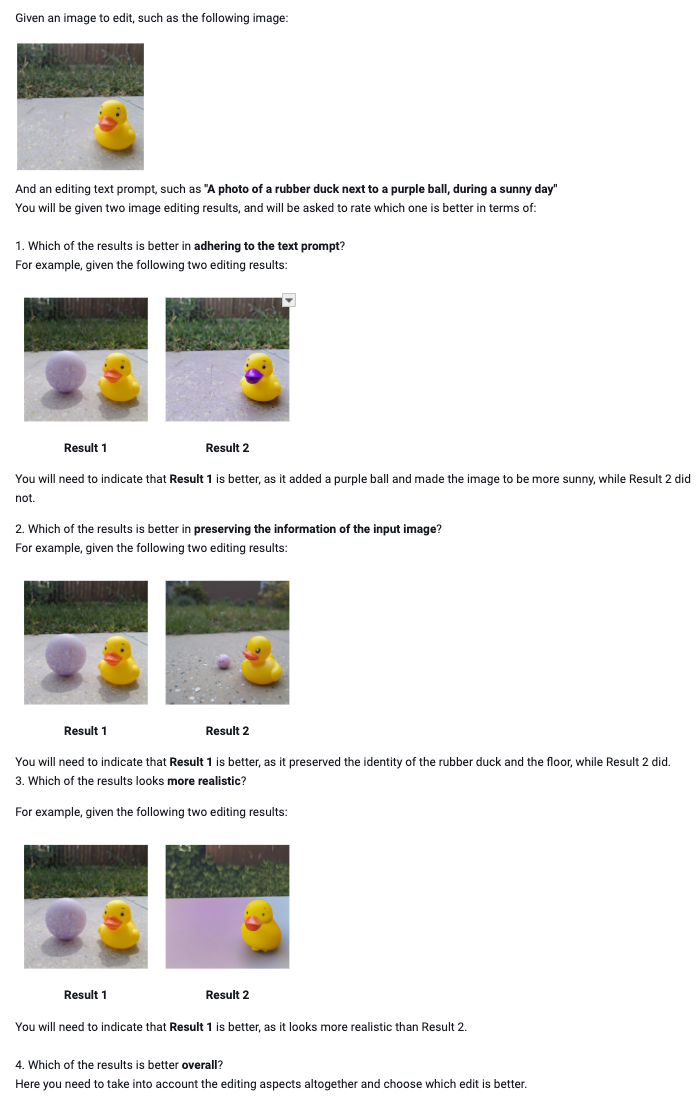}
    
    \caption{\textbf{User Study Instructions.} We provide the complete instructions for the user study we conducted using Amazon Mechanical Turk (AMT)~\cite{amt} to compare our method with each baseline.}
    \label{fig:user_study_instructions}
\end{figure}

\begin{figure}[th]
    \includegraphics[width=1\linewidth]{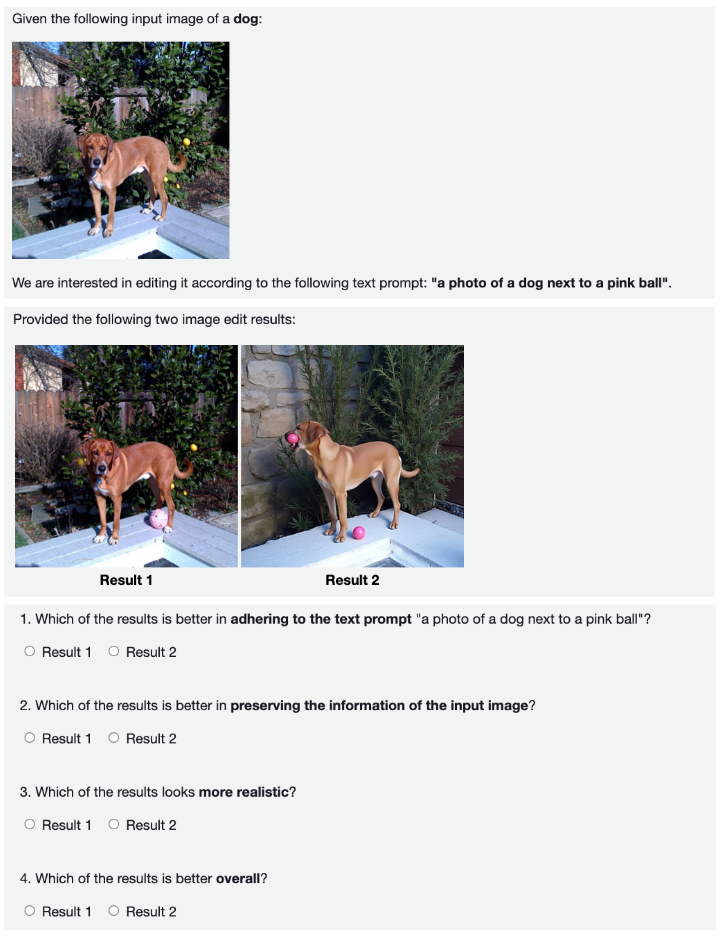}
    
    \caption{\textbf{User Study Trial.} We provide an example of a trial task in the user study conducted using Amazon Mechanical Turk (AMT)~\cite{amt}. Users were asked four questions of a two-alternative forced-choice format. Complete instructions are shown in \Cref{fig:user_study_instructions}.}
    \label{fig:user_study_trial}
\end{figure}

\begin{table}
    \centering
    \caption{\textbf{User Study Statistical Significance.} A binomial statistical test of the user study results suggests that our results are statistically significant (p-value $< 5\%$).}
    \begin{adjustbox}{width=1\columnwidth}
        \begin{tabular}{>{\columncolor[gray]{0.95}}lcccc}
            \toprule

            \textbf{Ours vs} & 
            Prompt Adher.&
            Image Pres.&
            Realism&
            Overall
            \\

            &
            p-value &
            p-value &
            p-value &
            p-value
            \\
            
            \midrule

            SDEdit~\cite{meng2021sdedit} &
            $< 1\mathrm{e}{-8}$ &
            $< 1\mathrm{e}{-8}$ &
            $< 1\mathrm{e}{-6}$ &
            $< 1\mathrm{e}{-8}$
            \\

            P2P+NTI~\cite{Hertz2022PrompttoPromptIE, mokady2022null} &
            $< 1\mathrm{e}{-8}$ &
            $< 1\mathrm{e}{-8}$ &
            $< 1\mathrm{e}{-8}$ &
            $< 6\mathrm{e}{-8}$
            \\

            Instruct-P2P~\cite{brooks2022instructpix2pix} &
            $< 1\mathrm{e}{-8}$ &
            $< 1\mathrm{e}{-8}$ &
            $< 1\mathrm{e}{-8}$ &
            $< 2\mathrm{e}{-4}$
            \\

            MagicBrush~\cite{Zhang2023MagicBrush} &
            $< 5\mathrm{e}{-5}$ &
            $< 1\mathrm{e}{-8}$ &
            $< 1\mathrm{e}{-8}$ &
            $< 1\mathrm{e}{-8}$
            \\

            MasaCTRL~\cite{cao2023masactrl} &
            $< 1\mathrm{e}{-8}$ &
            $< 1\mathrm{e}{-8}$ &
            $< 1\mathrm{e}{-8}$ &
            $< 1\mathrm{e}{-8}$
            \\
            
            \bottomrule
        \end{tabular}
    \end{adjustbox}
    \label{tab:user_study_statistical_significance}
\end{table}

As described in 
\Cref{sec:user_study}
we conducted an extensive user study using the Amazon Mechanical Turk (AMT)~\cite{amt} platform, using automatically generated test examples, as explained in \Cref{sec:automatic_metrics_details}. We compared all the baselines with our method using a standard two-alternative forced-choice format. The users were given full instructions, as can be seen in \Cref{fig:user_study_instructions}. Then, for each study trial, as shown in \Cref{fig:user_study_trial}, users were presented with an image and an instruction \prompt{Given the following input image of a \{CATEGORY\}} where \{CATEGORY\} is the COCO category of the prominent object. The users were given two editing results --- one from our method and one from the baseline, and were asked the following questions:
\begin{enumerate}
    \item \prompt{Which of the results is better in adhering to the text prompt \{PROMPT\}?}, where \{PROMPT\} is the editing target prompt.
    \item \prompt{Which of the results is better in preserving the information of the input image?}
    \item \prompt{Which of the results looks more realistic?}
    \item \prompt{Which of the results is better in overall?}
\end{enumerate}

We collected five ratings per sample, resulting in 320 ratings per baseline, for a total of 1,920 responses. The time allotted per task was one hour, to allow raters to properly evaluate the results without time pressure. A binomial statistical test of the user study results, as presented in \Cref{tab:user_study_statistical_significance}, suggests that our results are statistically significant (p-value $< 5\%$).

\section{Additional Experiments}
\label{sec:additional_experiments}

In \Cref{sec:additional_comparisons}, we start by providing additional comparisons and results of our method. Then, in \Cref{sec:different_perceptual_metrics}, we present experiments on using different perceptual metrics. Following that, in \Cref{sec:number_of_vital_layers}, we test the effect of different sizes for vital layer set. Next, in \Cref{sec:latent_nudging_experiment}, we provide latent nudging experiments. Furthermore, in \Cref{sec:layer_bypassing_visualization} we present a full visualization of our layer bypassing method. Finally, in \Cref{sec:sd3_results}, we test our method on the Stable Diffusion 3 backbone.

\subsection{Additional Comparisons and Results}
\label{sec:additional_comparisons}

\begin{figure*}[tp]
    \centering
    \setlength{\tabcolsep}{0.6pt}
    \renewcommand{\arraystretch}{0.8}
    \setlength{\ww}{0.138\linewidth}
    \begin{tabular}{c @{\hspace{10\tabcolsep}} cccccc}

        \footnotesize{Input} &
        \footnotesize{SDEdit}~\cite{meng2021sdedit} &
        \footnotesize{P2P+NTI}~\cite{Hertz2022PrompttoPromptIE, mokady2022null} &
        \footnotesize{Instruct-P2P}~\cite{brooks2022instructpix2pix} &
        \footnotesize{MagicBrush}~\cite{Zhang2023MagicBrush} &
        \footnotesize{MasaCTRL}~\cite{cao2023masactrl} &
        \footnotesize{Stable Flow (ours)}
        \vspace{2px}
        \\

        {\includegraphics[valign=c, width=\ww]{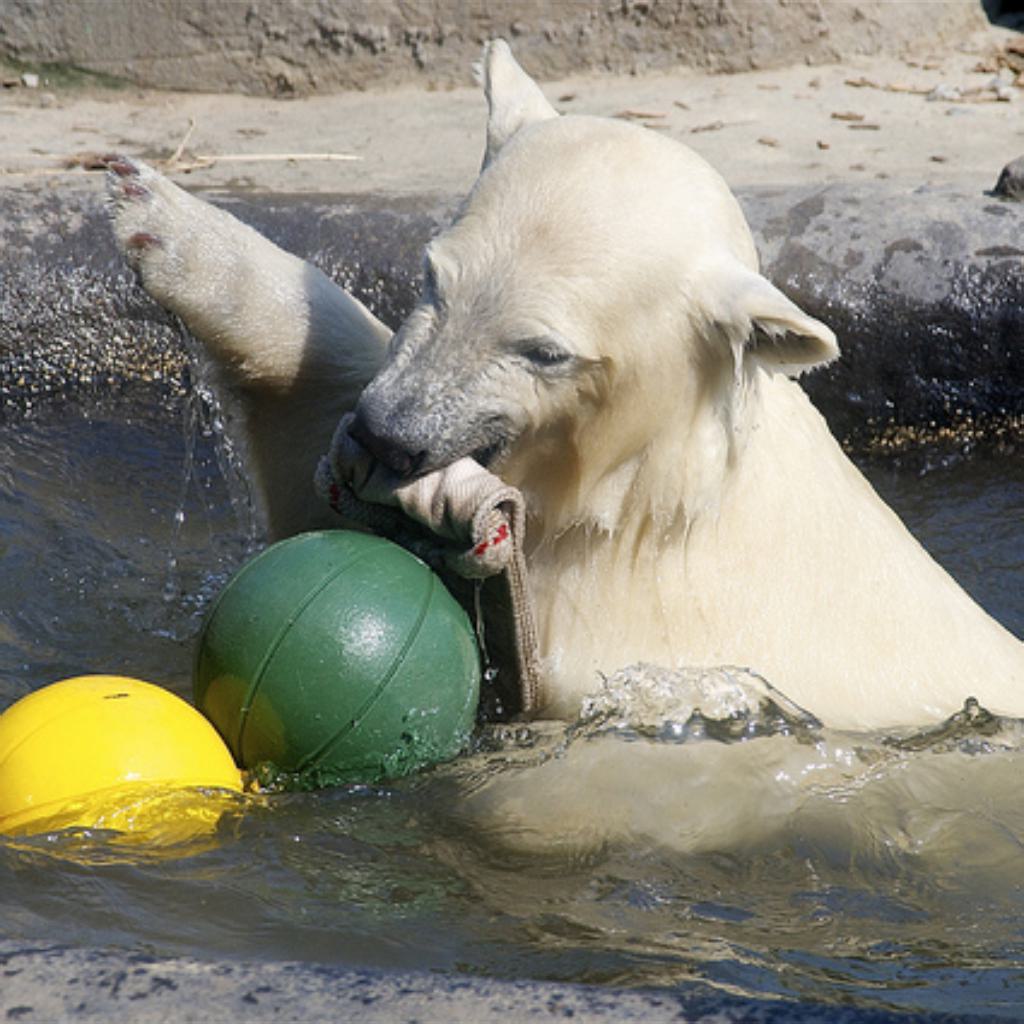}} &
        {\includegraphics[valign=c, width=\ww]{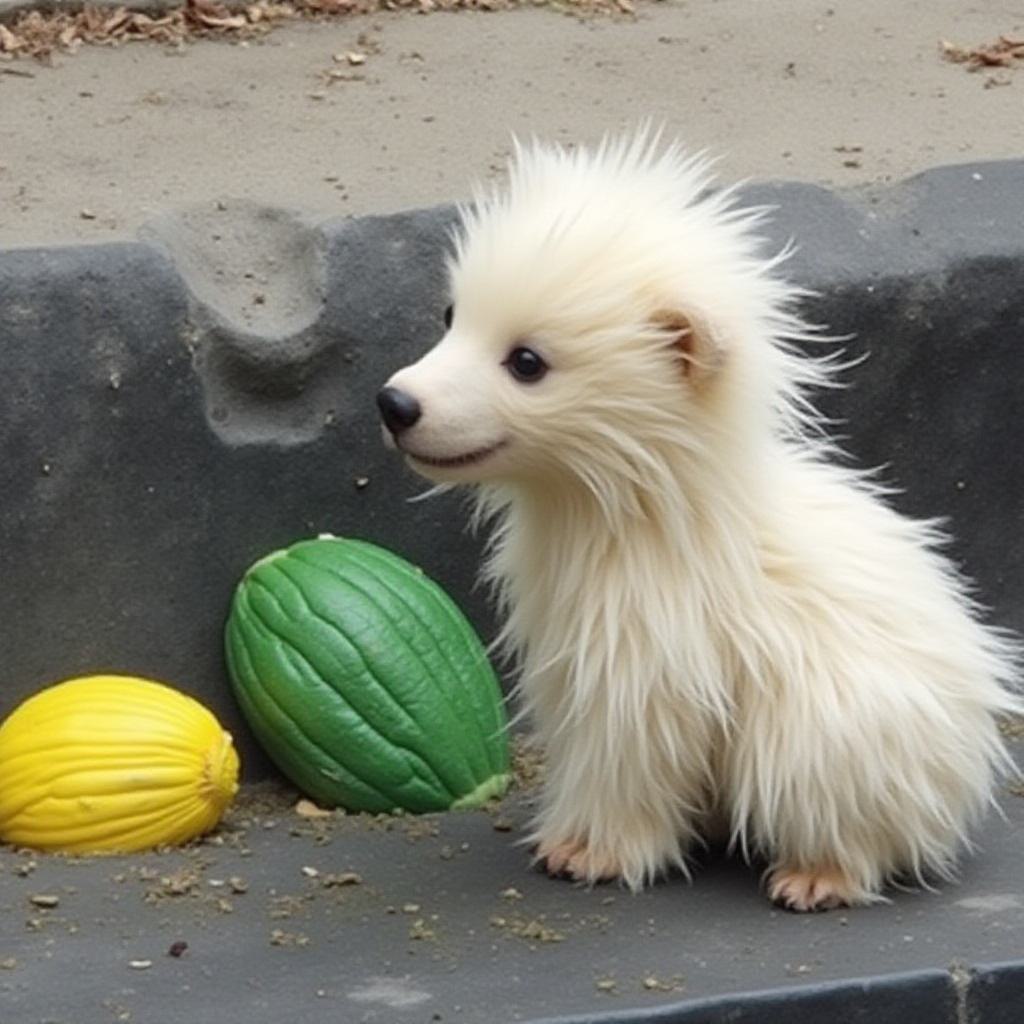}} &
        {\includegraphics[valign=c, width=\ww]{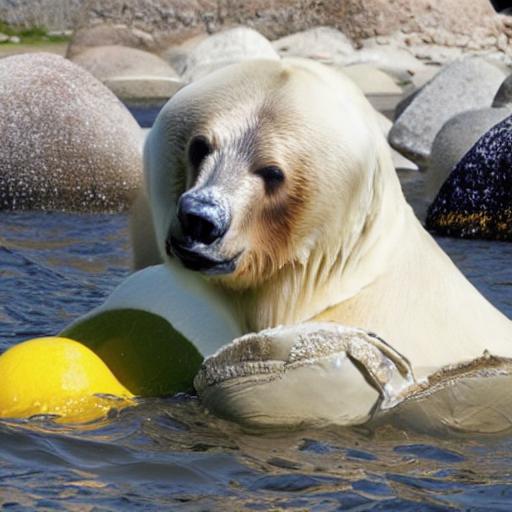}} &
        {\includegraphics[valign=c, width=\ww]{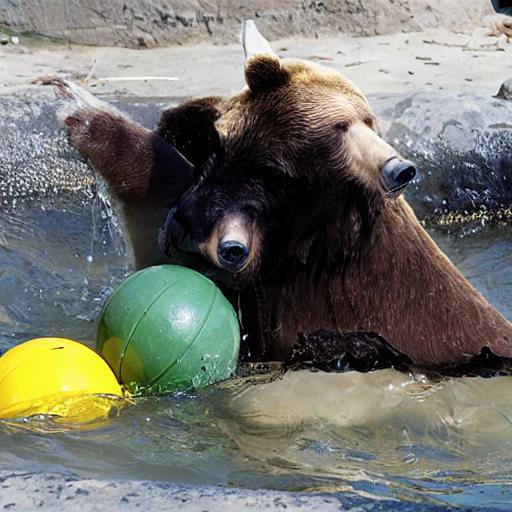}} &
        {\includegraphics[valign=c, width=\ww]{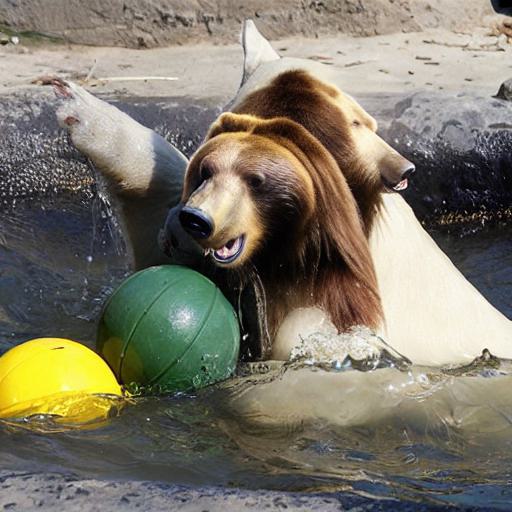}} &
        {\includegraphics[valign=c, width=\ww]{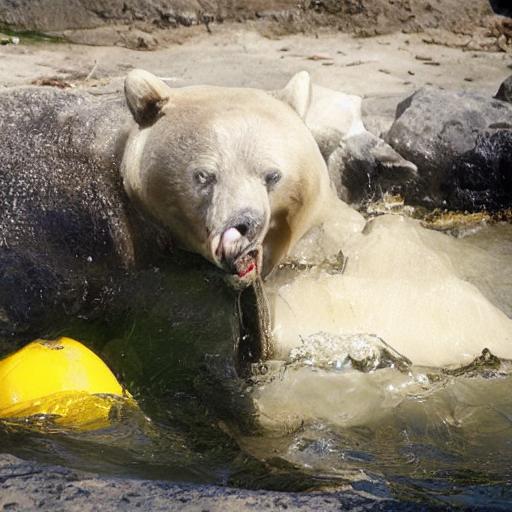}} &
        {\includegraphics[valign=c, width=\ww]{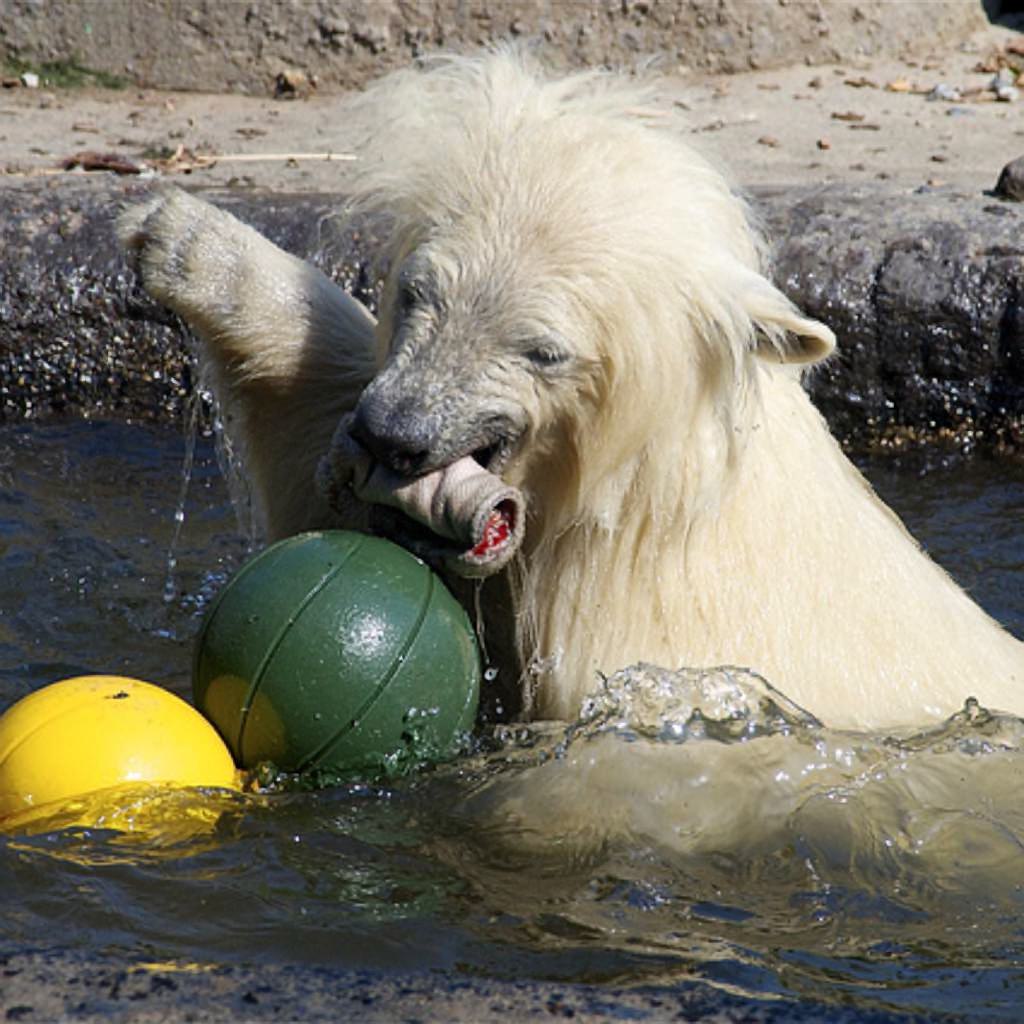}}
        \vspace{1px}
        \\

        &
        \multicolumn{6}{c}{\small{\prompt{A photo of a bear with a long hair}}}
        \vspace{5px}
        \\

        {\includegraphics[valign=c, width=\ww]{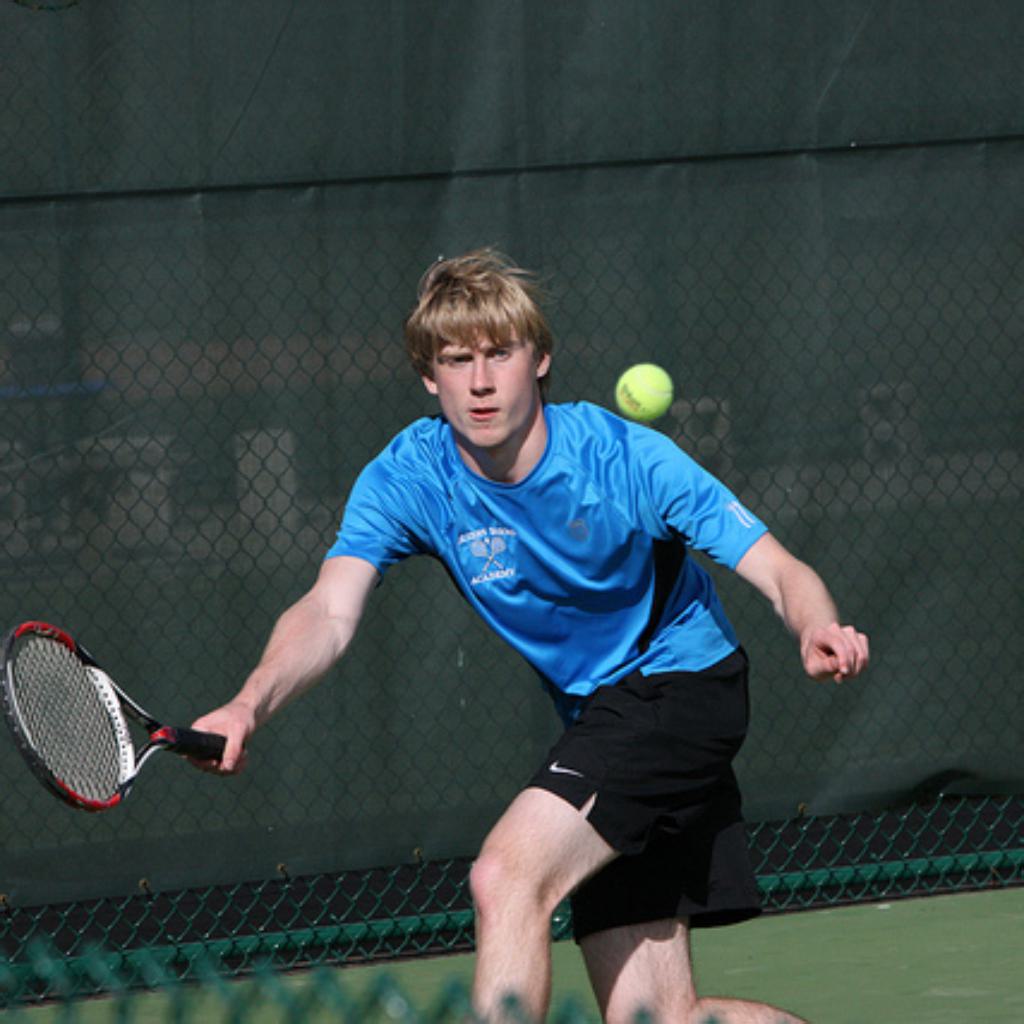}} &
        {\includegraphics[valign=c, width=\ww]{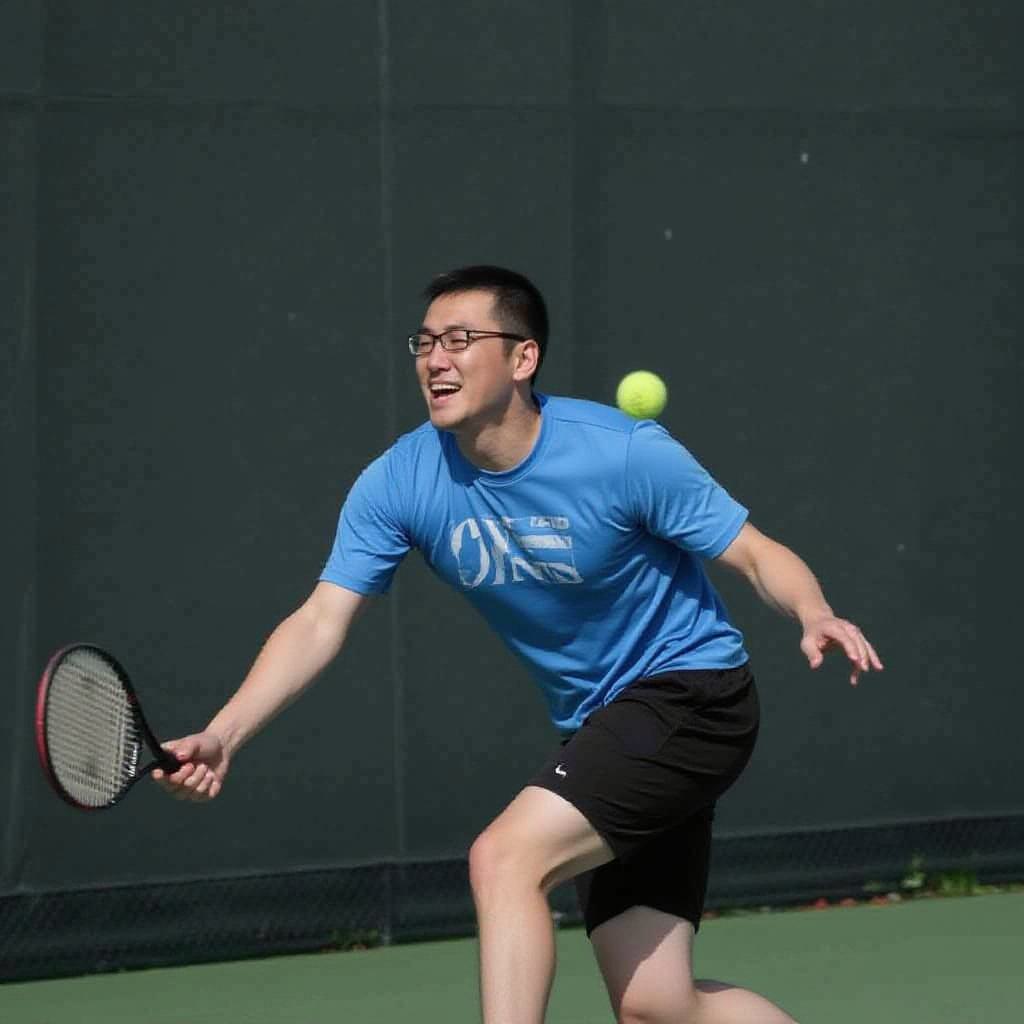}} &
        {\includegraphics[valign=c, width=\ww]{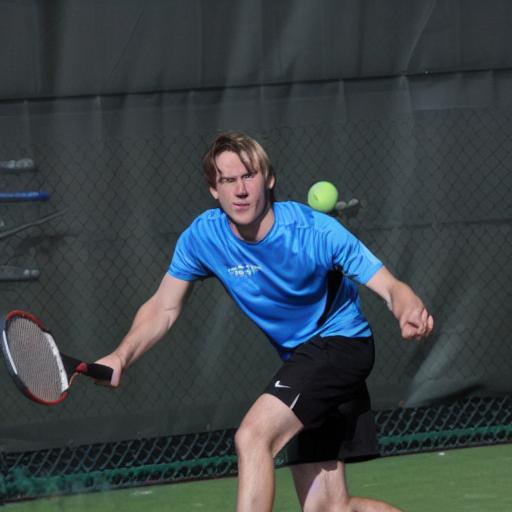}} &
        {\includegraphics[valign=c, width=\ww]{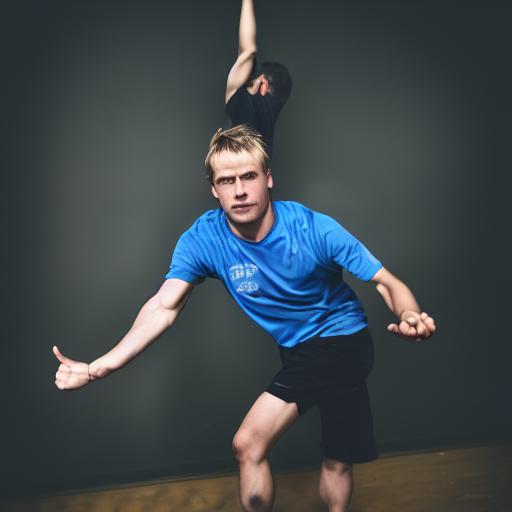}} &
        {\includegraphics[valign=c, width=\ww]{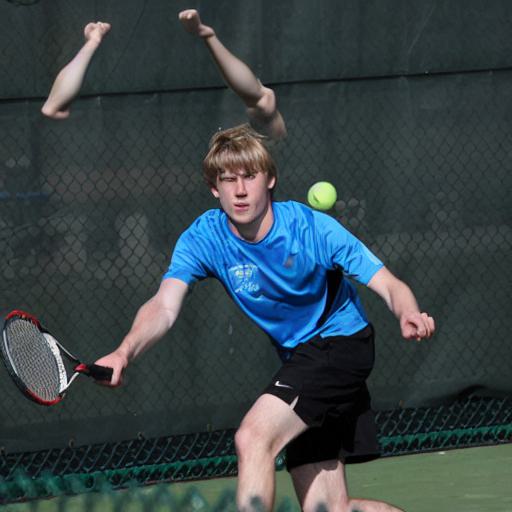}} &
        {\includegraphics[valign=c, width=\ww]{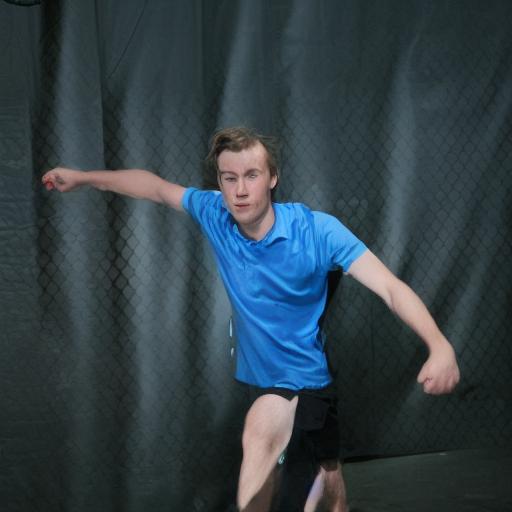}} &
        {\includegraphics[valign=c, width=\ww]{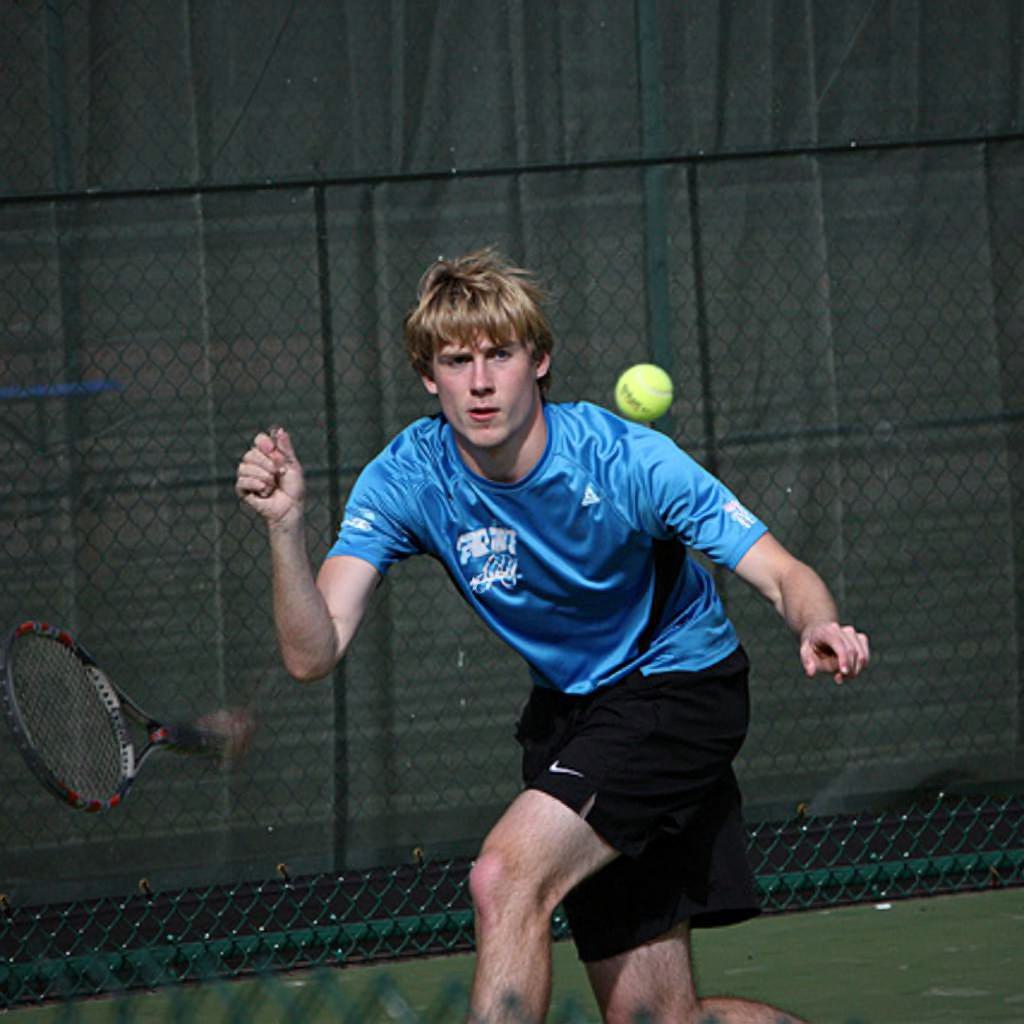}}
        \vspace{1px}
        \\

        &
        \multicolumn{6}{c}{\small{\prompt{A photo of a man raising his hand}}}
        \vspace{5px}
        \\

        {\includegraphics[valign=c, width=\ww]{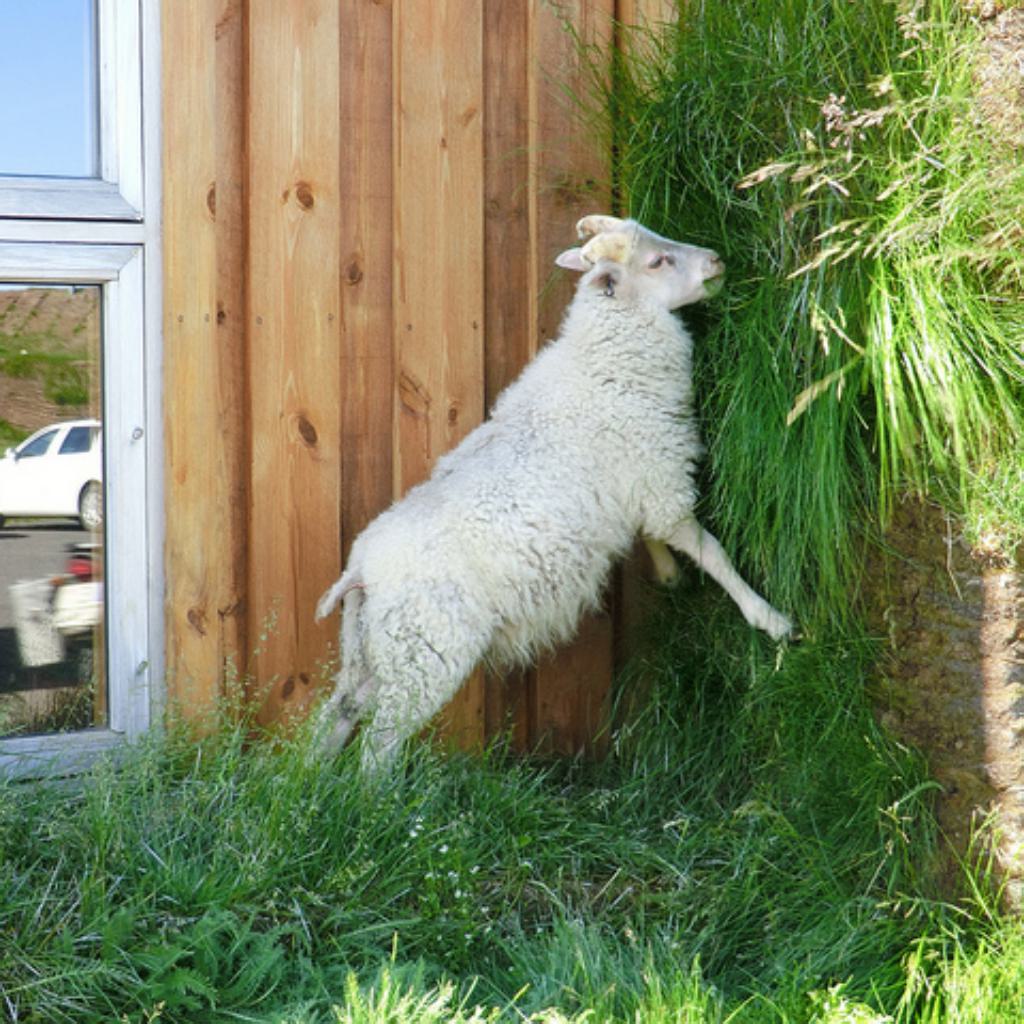}} &
        {\includegraphics[valign=c, width=\ww]{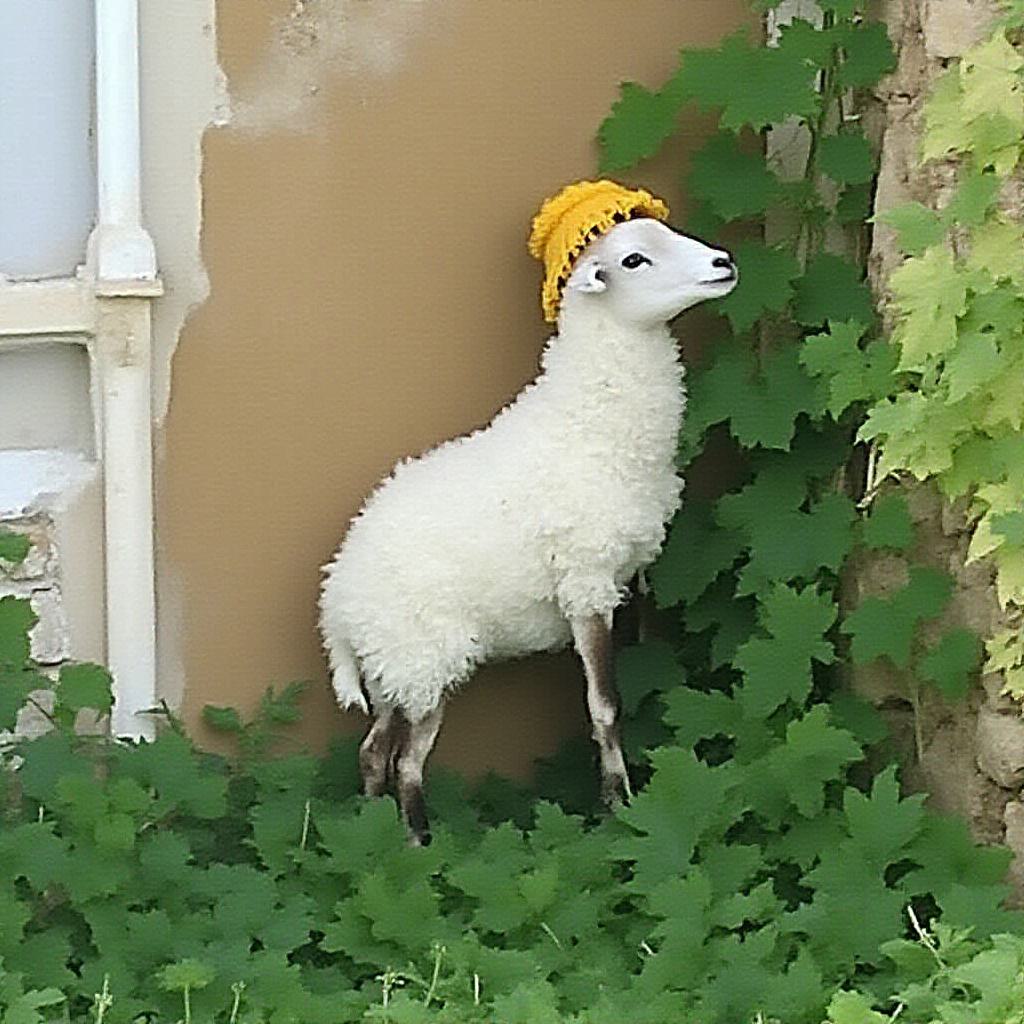}} &
        {\includegraphics[valign=c, width=\ww]{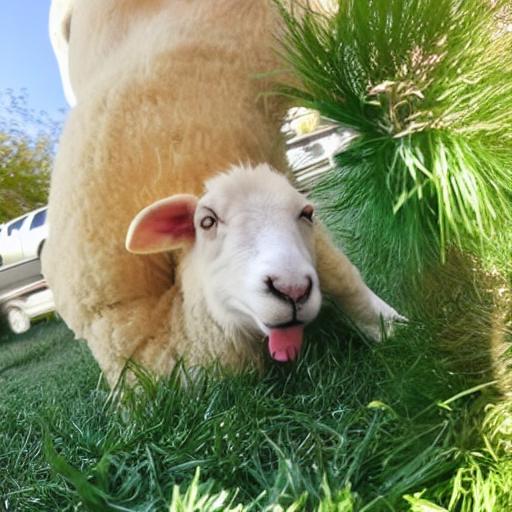}} &
        {\includegraphics[valign=c, width=\ww]{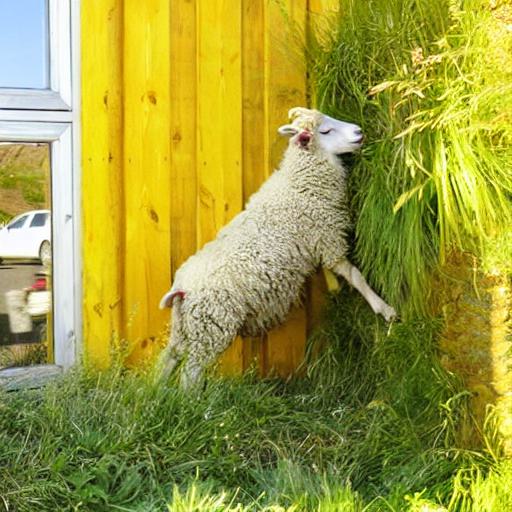}} &
        {\includegraphics[valign=c, width=\ww]{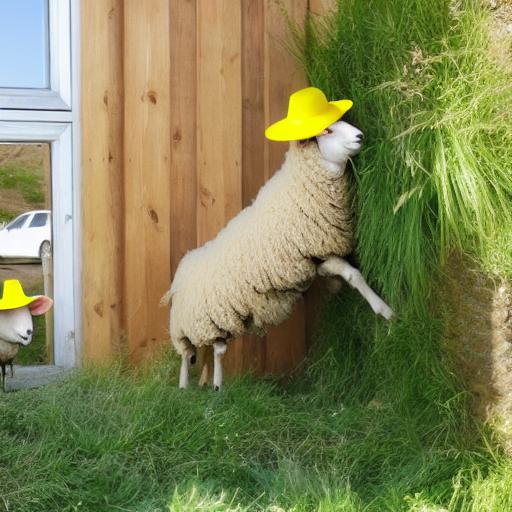}} &
        {\includegraphics[valign=c, width=\ww]{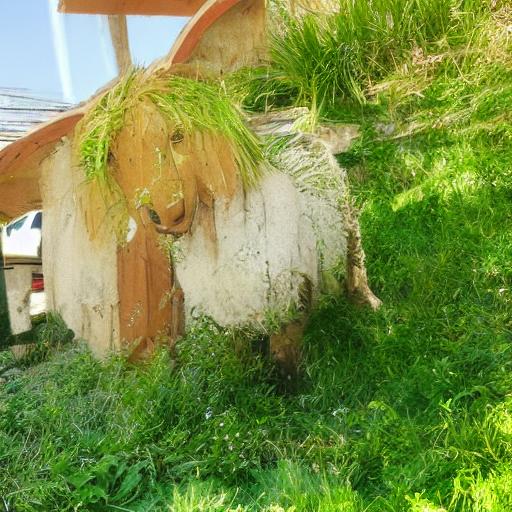}} &
        {\includegraphics[valign=c, width=\ww]{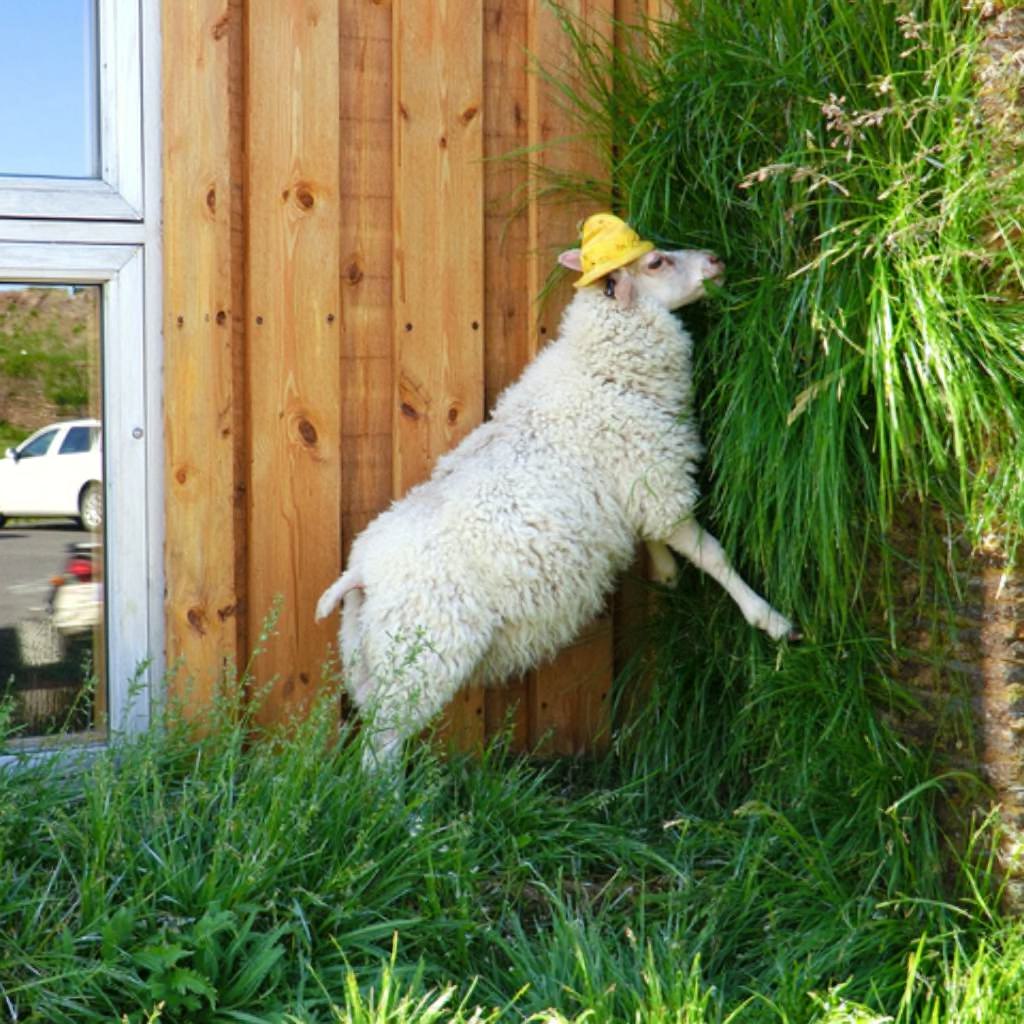}}
        \vspace{1px}
        \\

        &
        \multicolumn{6}{c}{\small{\prompt{A photo of a sheep with a yellow hat}}}
        \vspace{5px}
        \\

        {\includegraphics[valign=c, width=\ww]{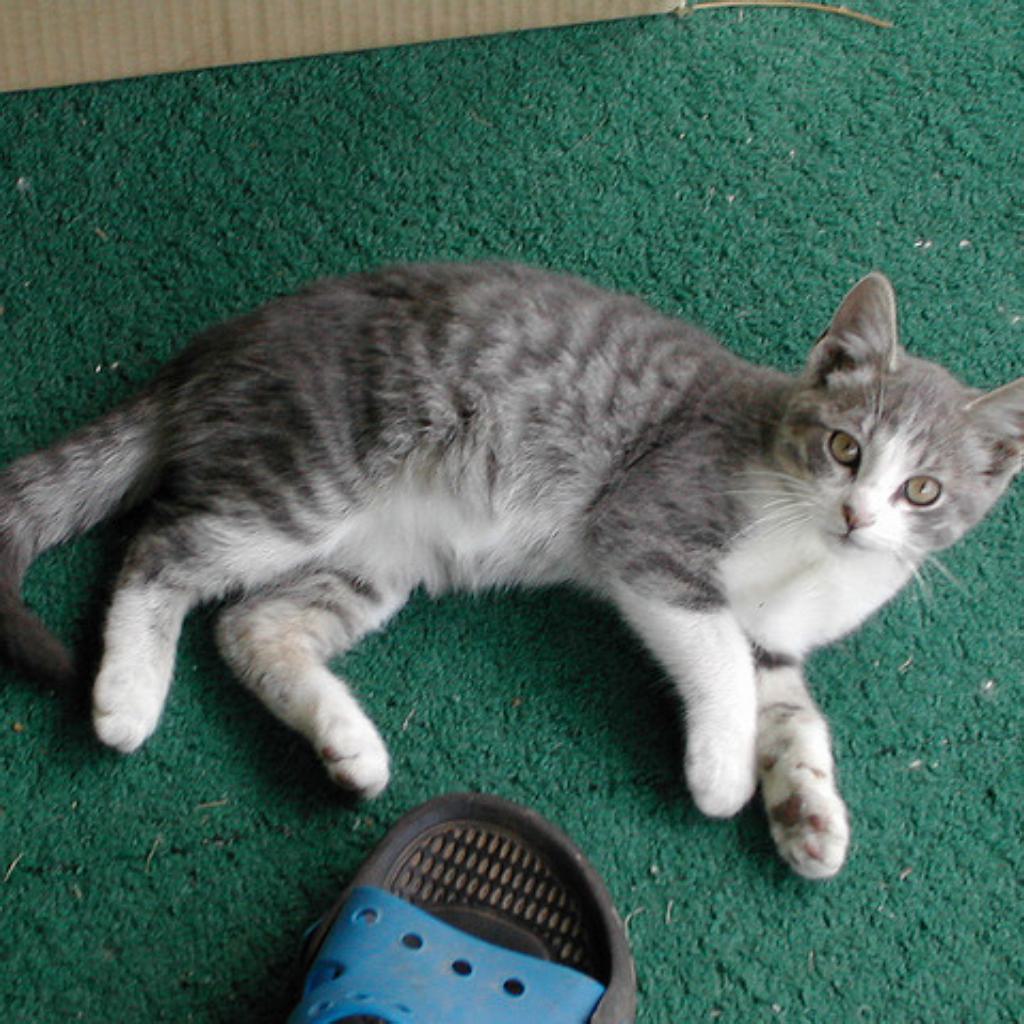}} &
        {\includegraphics[valign=c, width=\ww]{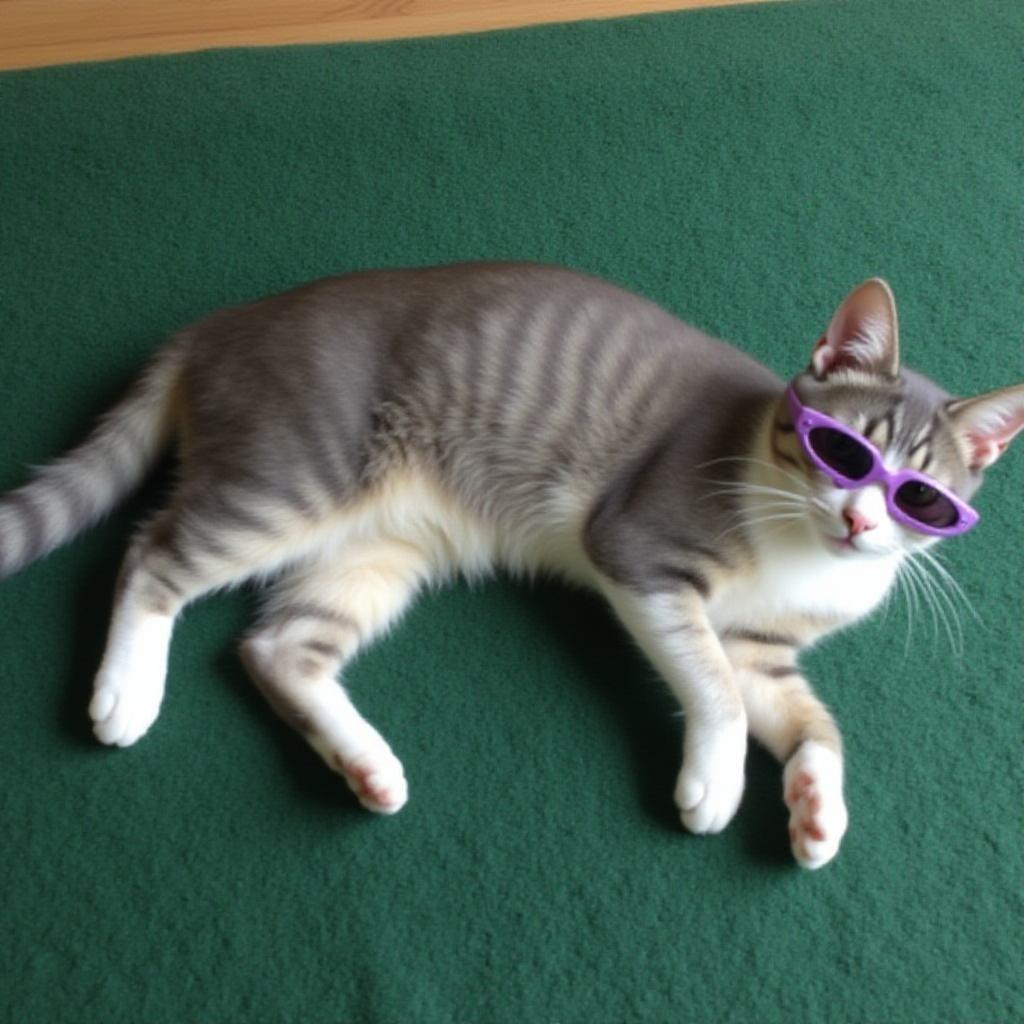}} &
        {\includegraphics[valign=c, width=\ww]{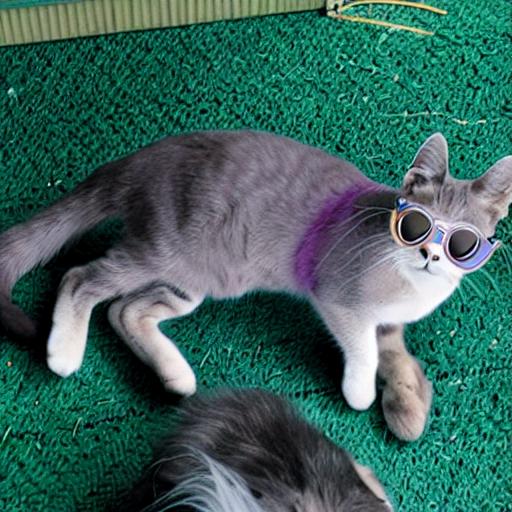}} &
        {\includegraphics[valign=c, width=\ww]{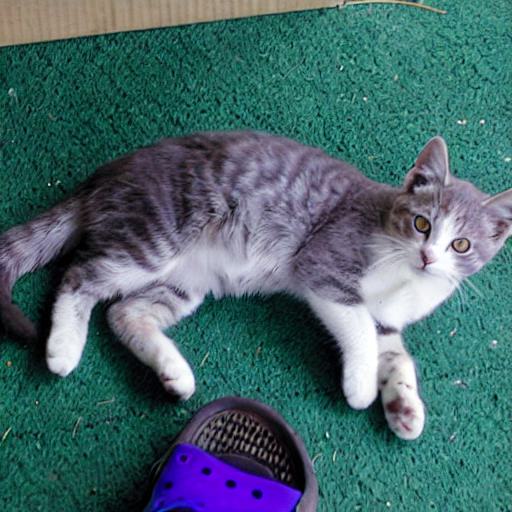}} &
        {\includegraphics[valign=c, width=\ww]{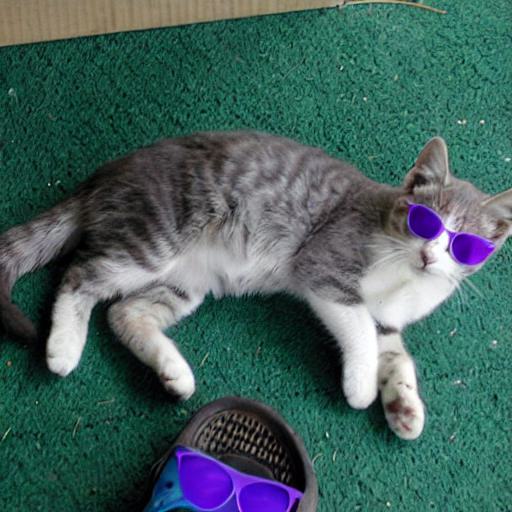}} &
        {\includegraphics[valign=c, width=\ww]{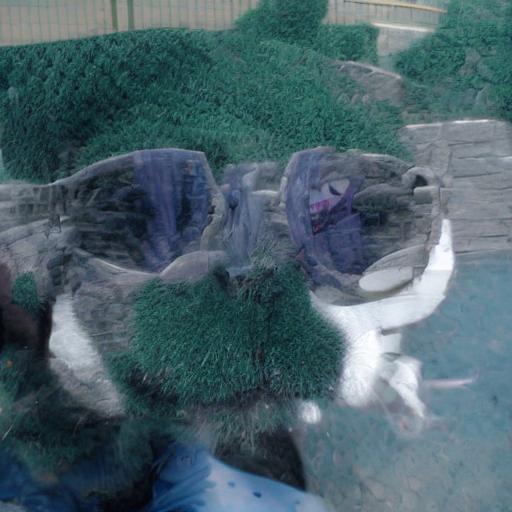}} &
        {\includegraphics[valign=c, width=\ww]{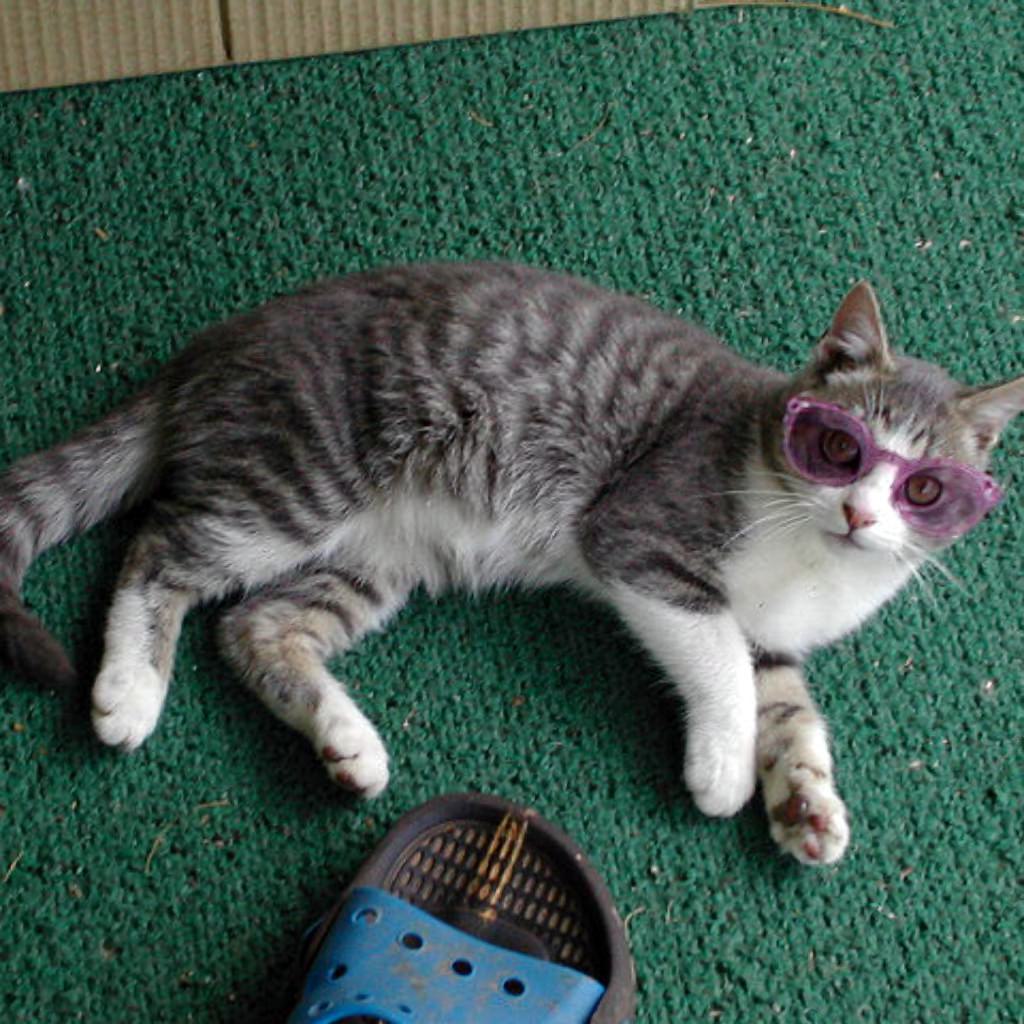}}
        \vspace{1px}
        \\

        &
        \multicolumn{6}{c}{\small{\prompt{A photo of a cat wearing purple sunglasses}}}
        \vspace{5px}
        \\

        {\includegraphics[valign=c, width=\ww]{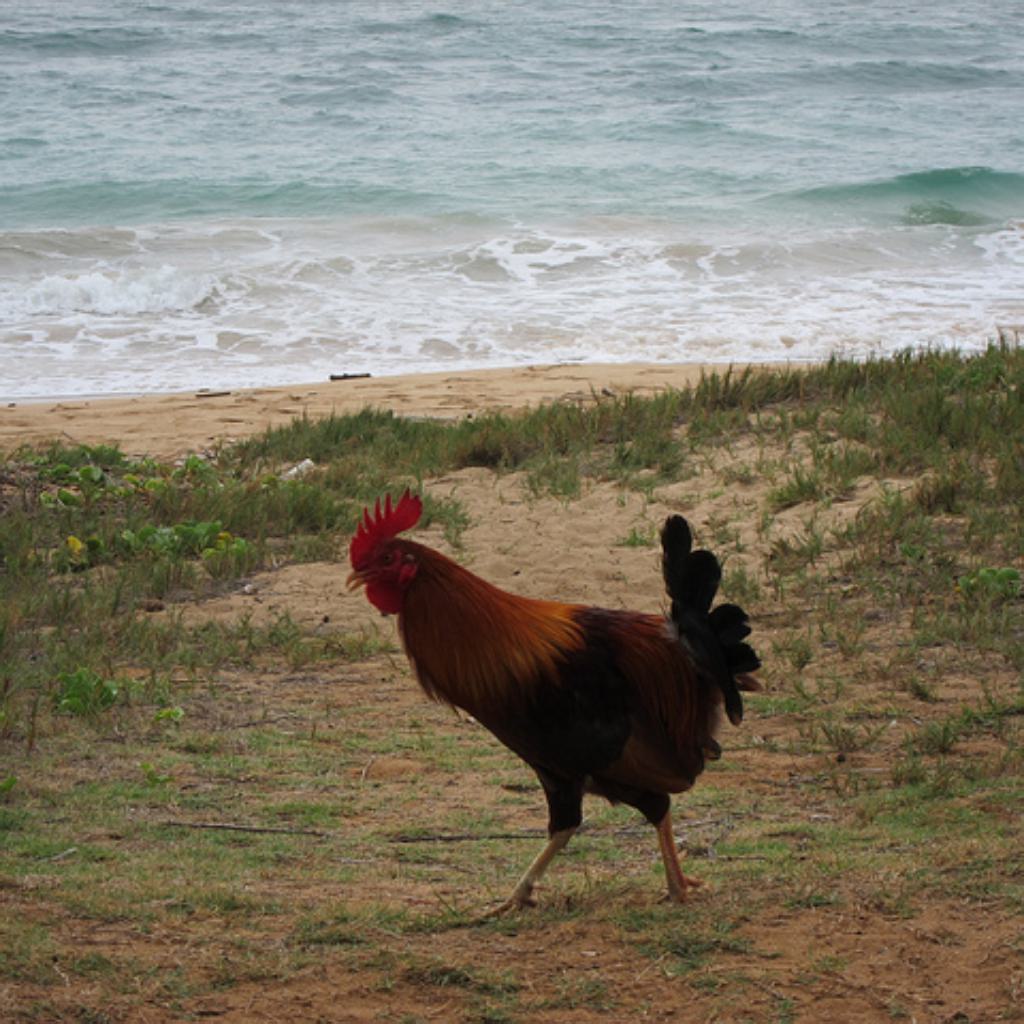}} &
        {\includegraphics[valign=c, width=\ww]{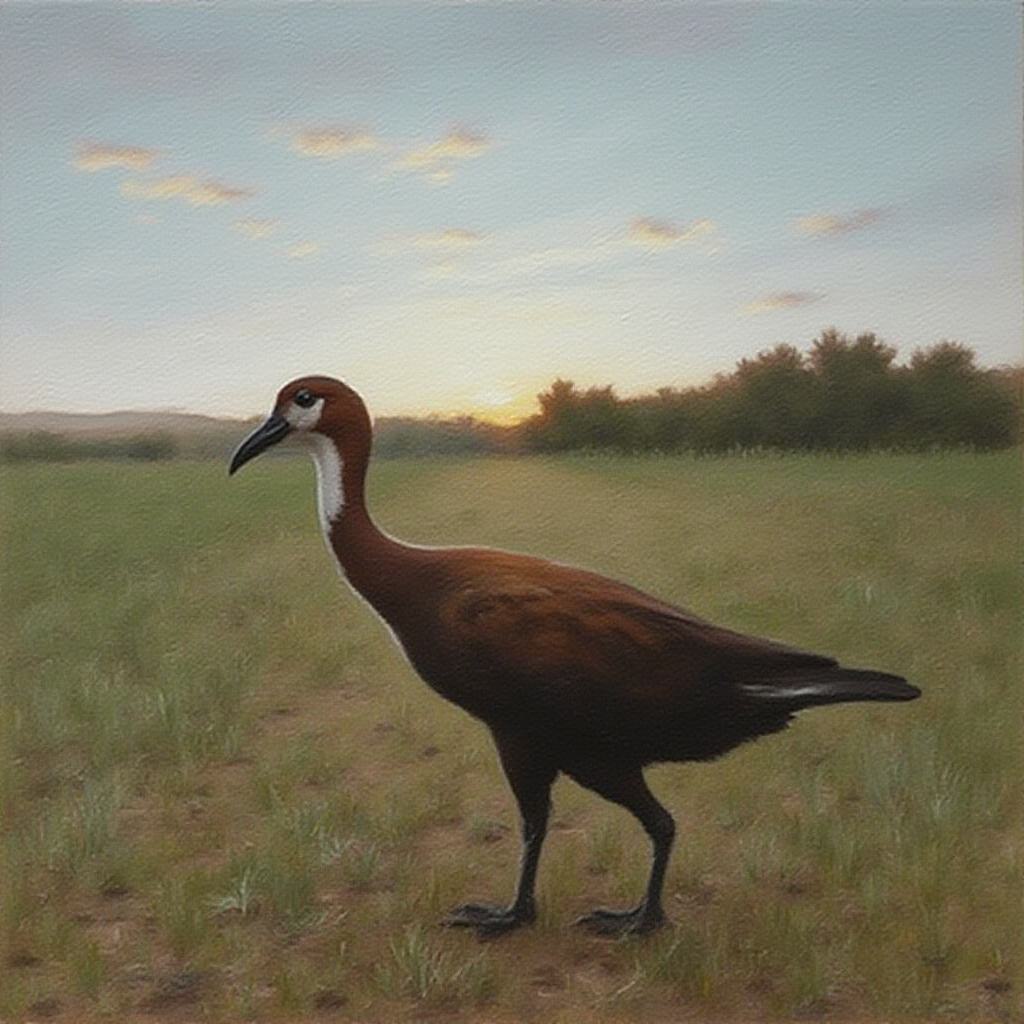}} &
        {\includegraphics[valign=c, width=\ww]{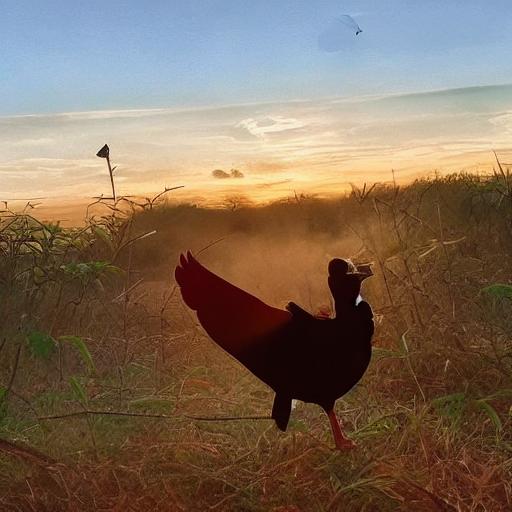}} &
        {\includegraphics[valign=c, width=\ww]{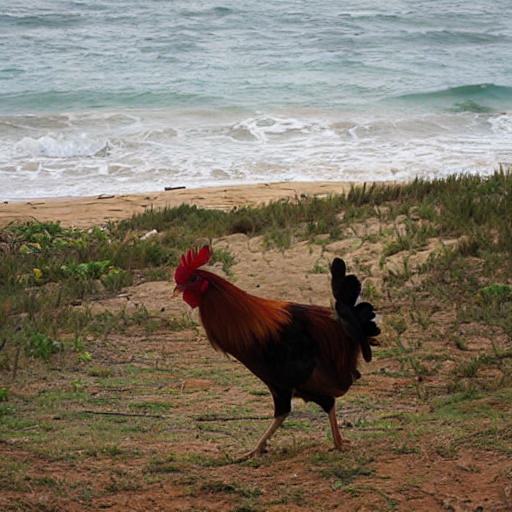}} &
        {\includegraphics[valign=c, width=\ww]{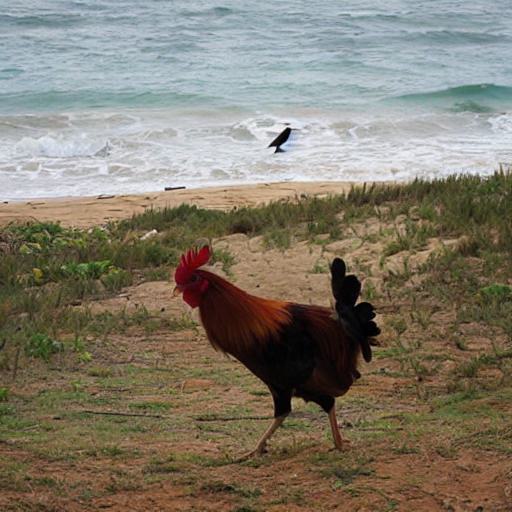}} &
        {\includegraphics[valign=c, width=\ww]{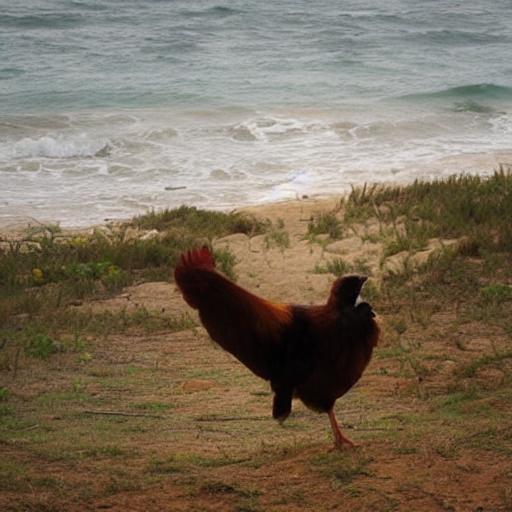}} &
        {\includegraphics[valign=c, width=\ww]{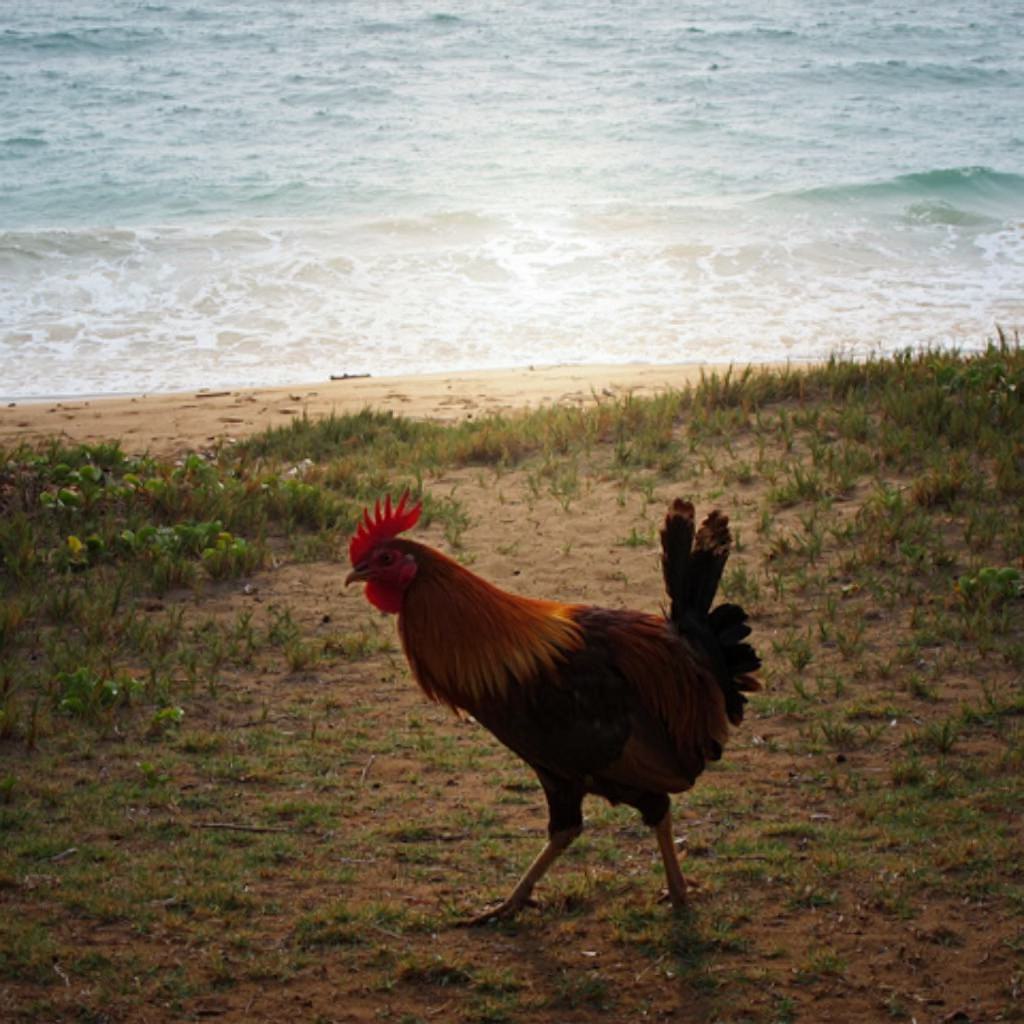}}
        \vspace{1px}
        \\

        &
        \multicolumn{6}{c}{\small{\prompt{A photo of a chicken during sunset}}}
        \vspace{5px}
        \\

        {\includegraphics[valign=c, width=\ww]{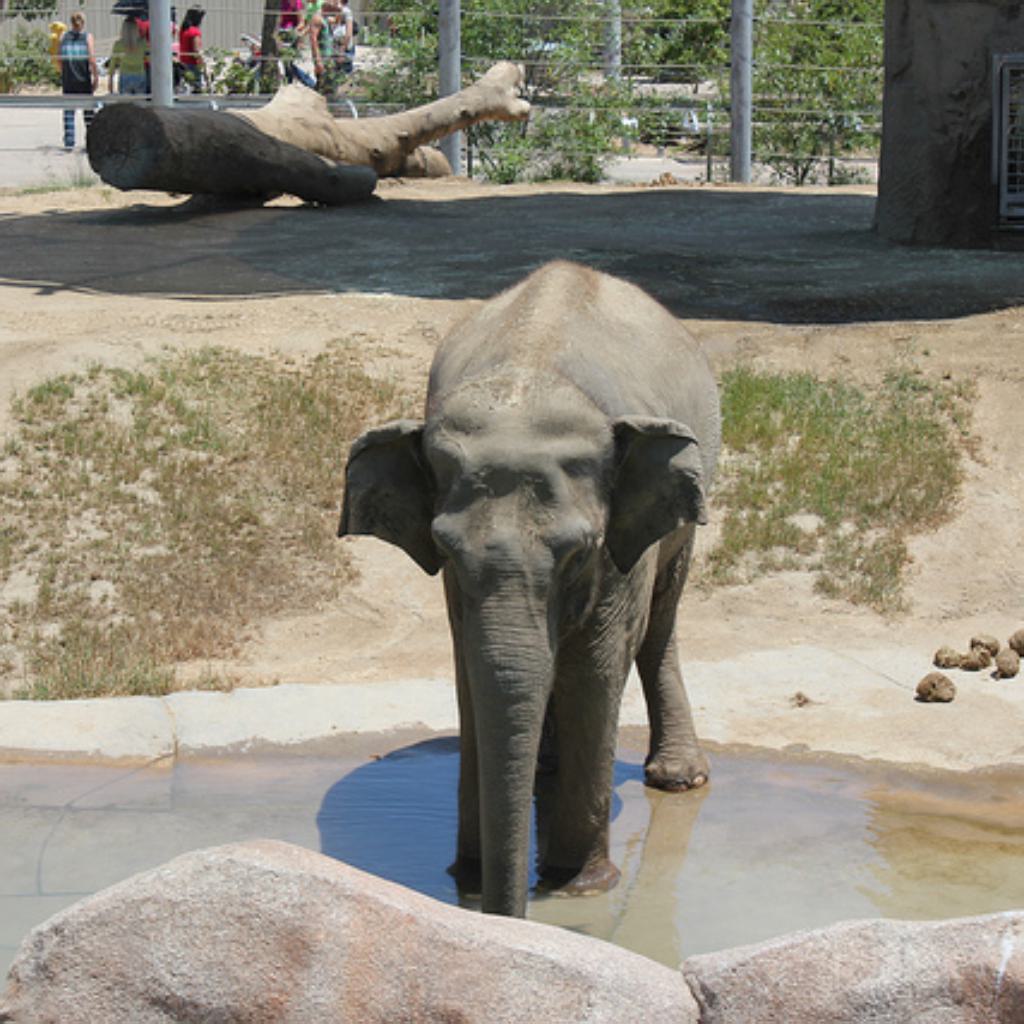}} &
        {\includegraphics[valign=c, width=\ww]{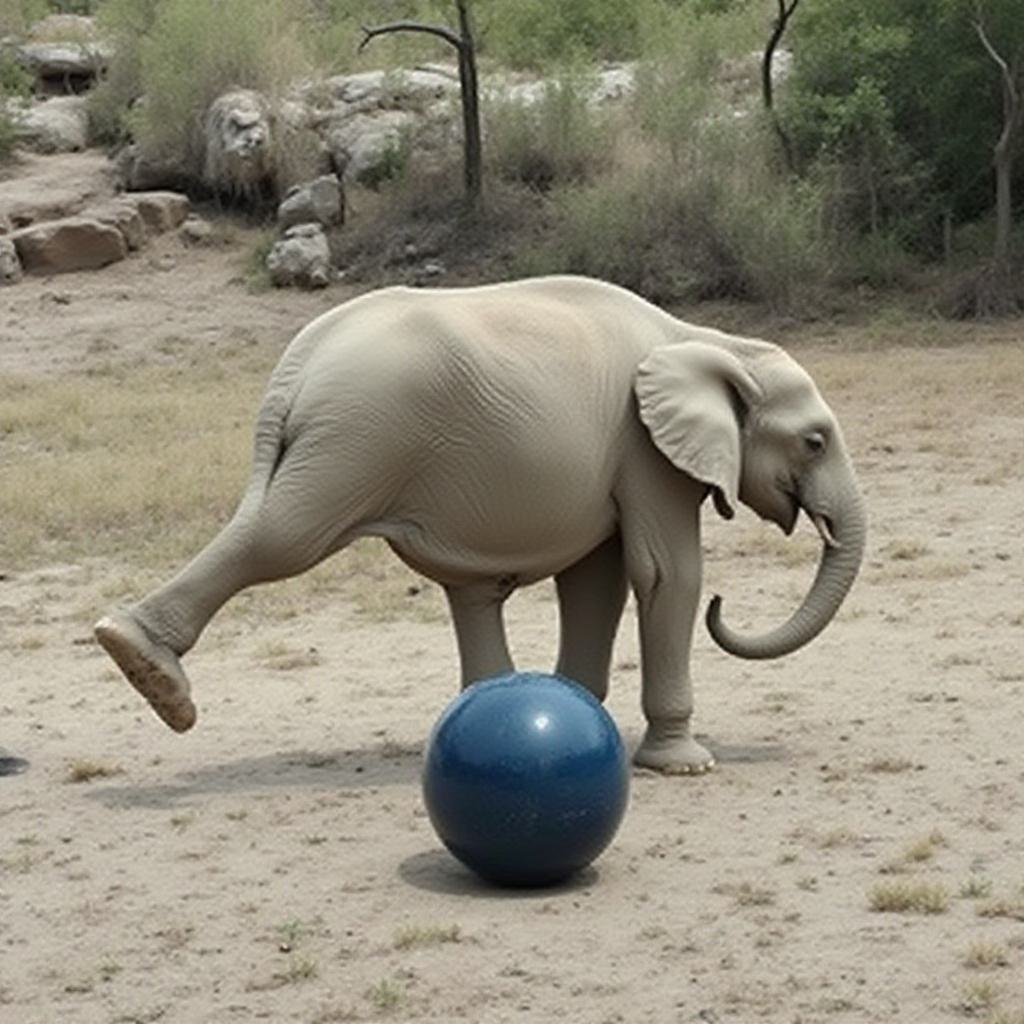}} &
        {\includegraphics[valign=c, width=\ww]{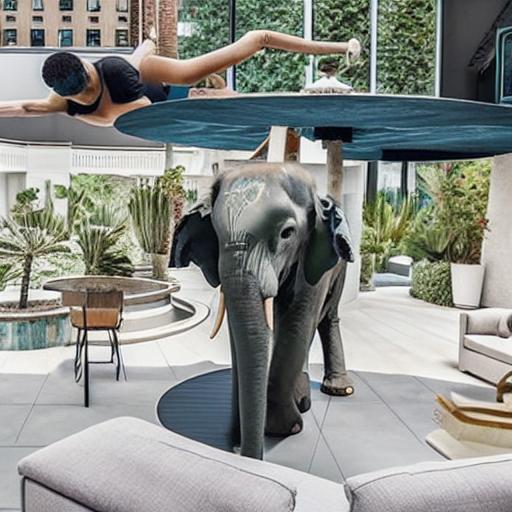}} &
        {\includegraphics[valign=c, width=\ww]{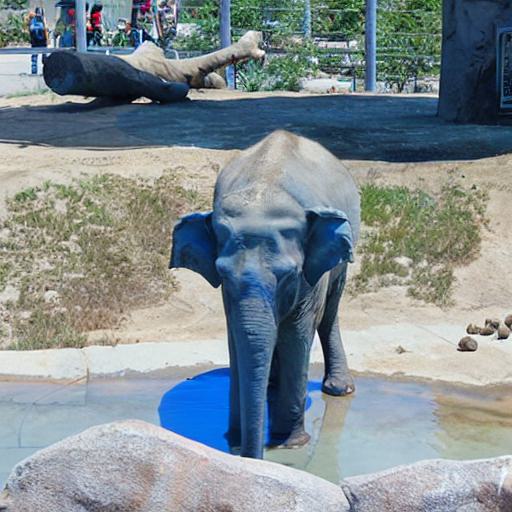}} &
        {\includegraphics[valign=c, width=\ww]{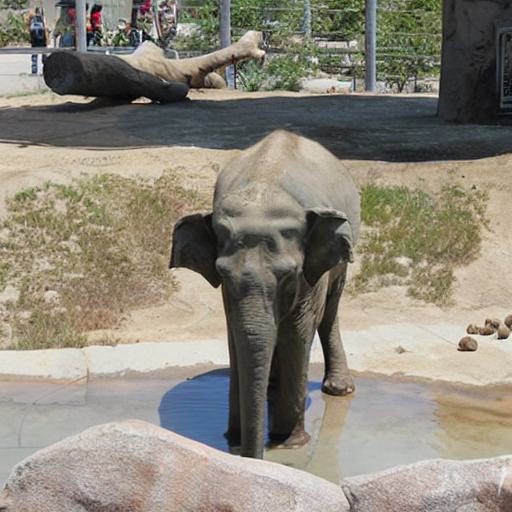}} &
        {\includegraphics[valign=c, width=\ww]{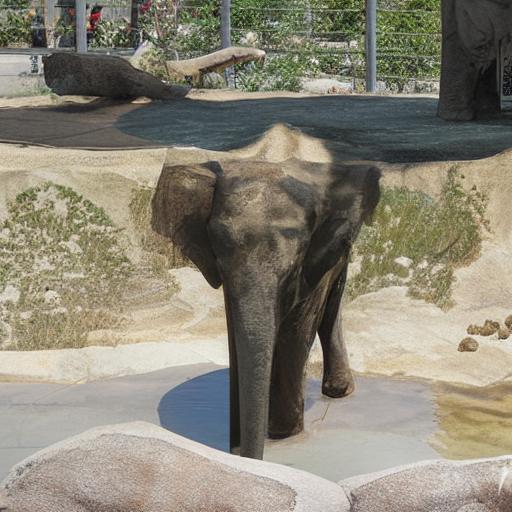}} &
        {\includegraphics[valign=c, width=\ww]{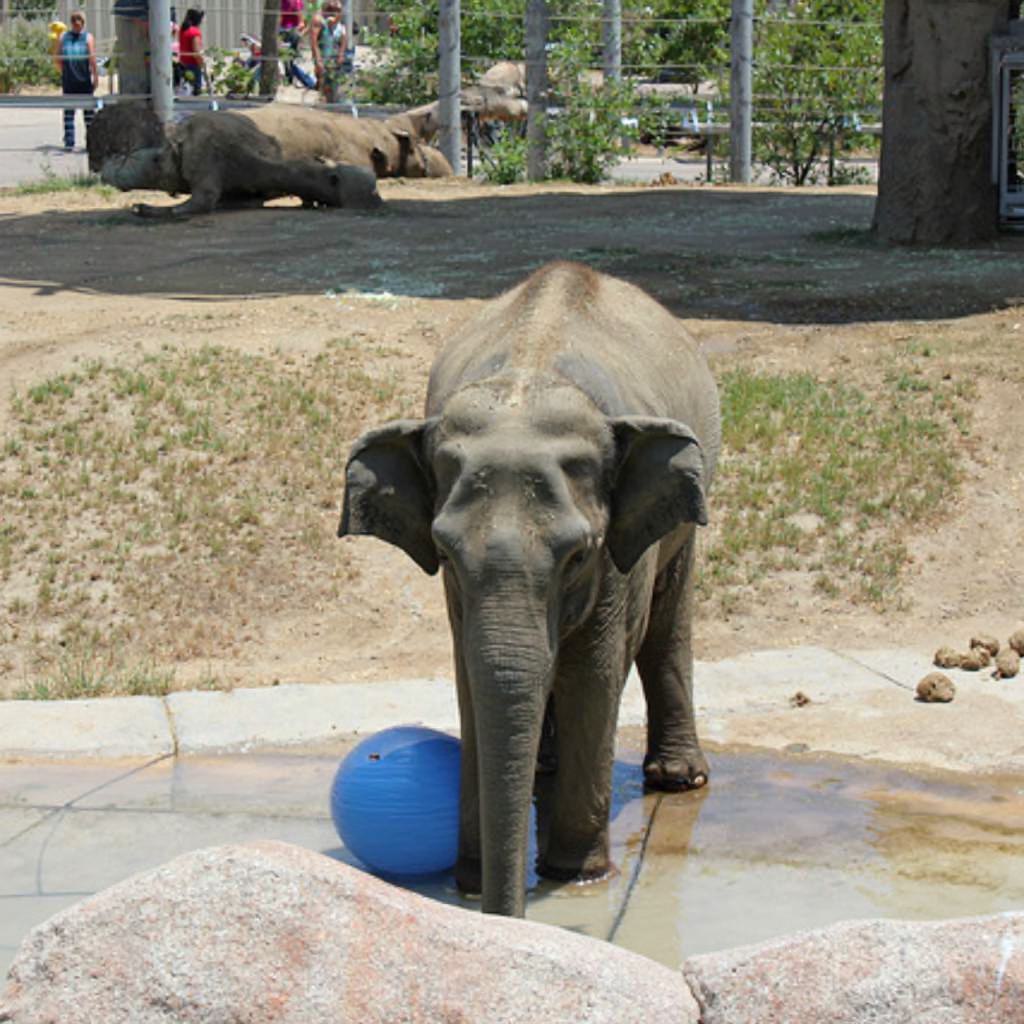}}
        \vspace{1px}
        \\

        &
        \multicolumn{6}{c}{\small{\prompt{A photo of an elephant next to a blue ball}}}
        \vspace{8px}
        \\

    \end{tabular}
    \vspace{-3px}
    \caption{\textbf{Baselines Qualitative Comparison on Automatic Dataset.} As explained in 
    \Cref{sec:comparisons},
    we compare our method against the baselines on real images extracted from the COCO~\cite{Lin2014MicrosoftCC} dataset. We find that SDEdit~\cite{meng2021sdedit} struggles with preserving the object identities and backgrounds (\eg, bear and chicken examples). P2P+NTI ~\cite{Hertz2022PrompttoPromptIE, mokady2022null} struggles with preserving object identities (\eg, bear and person examples) and with adding new objects (\eg, missing hat in the sheep example and missing ball in the elephant example). Instruct-P2P~\cite{brooks2022instructpix2pix} and MagicBrush~\cite{Zhang2023MagicBrush} struggle with non-rigid editing (\eg, person raising hand). MasaCTRL~\cite{cao2023masactrl} struggles with preserving object identities (\eg, bear and person examples) and adding new objects (\eg, sheep and cat examples). Our method, on the other hand, is able to adhere to the editing prompt while preserving the identities.}
    \label{fig:baselines_qualitative_comparison_automatic}
\end{figure*}

\begin{figure*}[tp]
    \centering
    \setlength{\tabcolsep}{0.6pt}
    \renewcommand{\arraystretch}{0.8}
    \setlength{\ww}{0.16\linewidth}
    \begin{tabular}{c @{\hspace{10\tabcolsep}} ccccc}

        \footnotesize{Input} &
        \footnotesize{(1) Inj. all layers} &
        \footnotesize{(2) Inj. non-vital layers} &
        \footnotesize{(3) Extension all layers} &
        \footnotesize{(4) w/o latent nudging} &
        \footnotesize{Stable Flow (ours)}
        \vspace{2px}
        \\

        {\includegraphics[valign=c, width=\ww]{figures/qualitative_comparison_automatic/assets/bear/inp.jpg}} &
        {\includegraphics[valign=c, width=\ww]{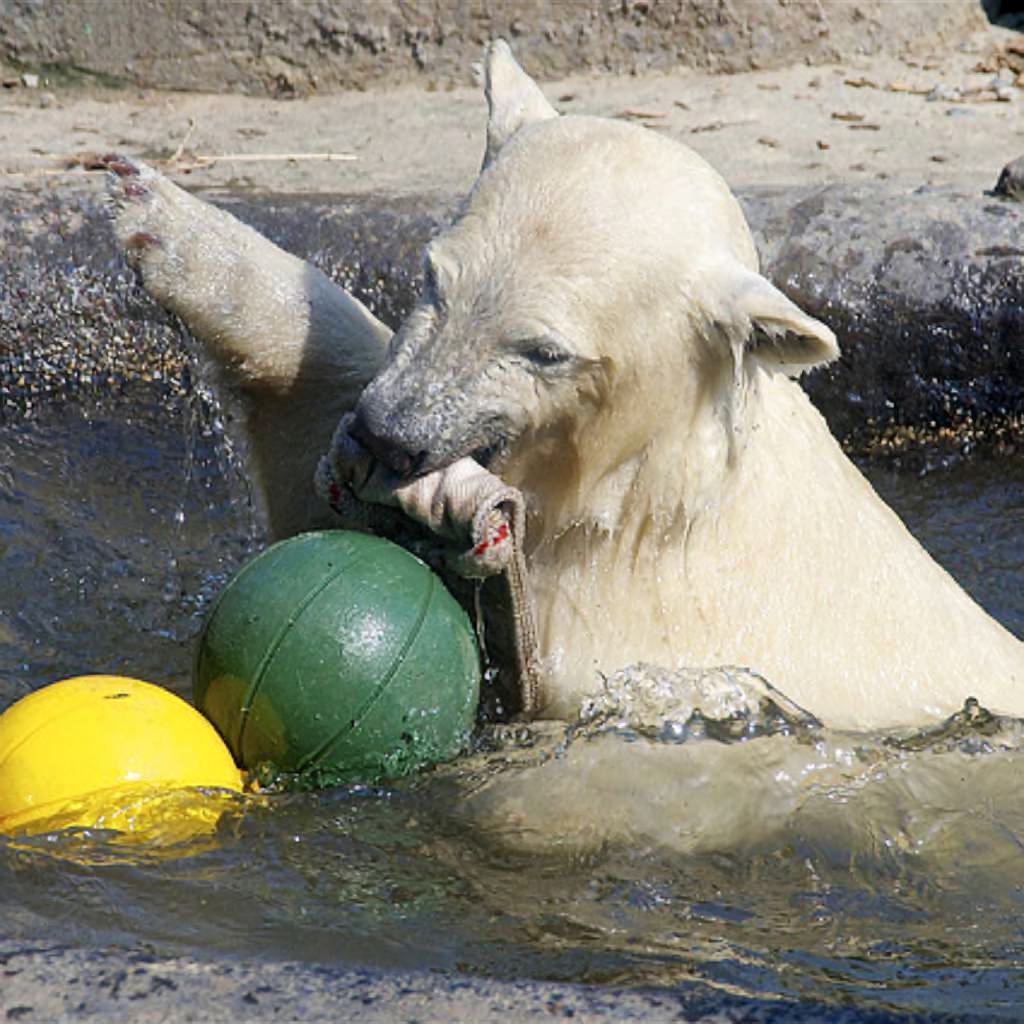}} &
        {\includegraphics[valign=c, width=\ww]{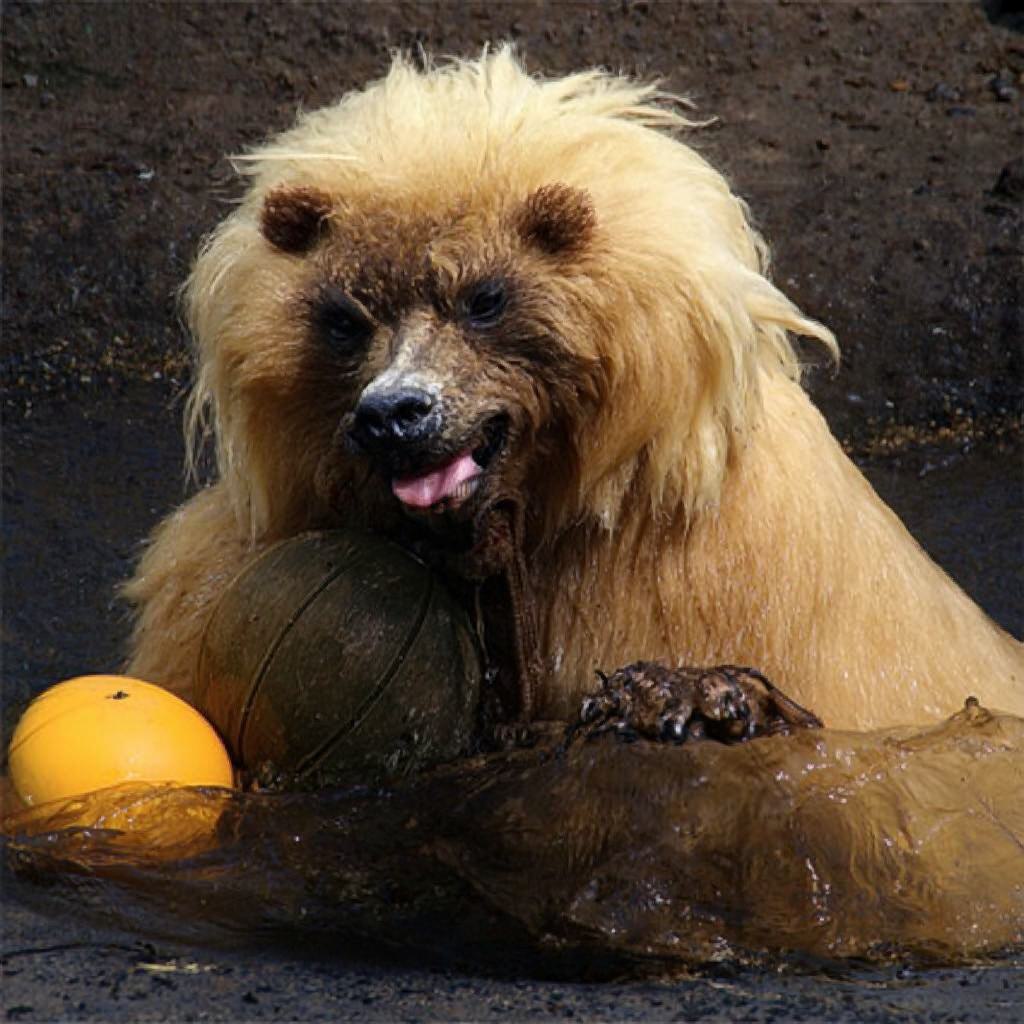}} &
        {\includegraphics[valign=c, width=\ww]{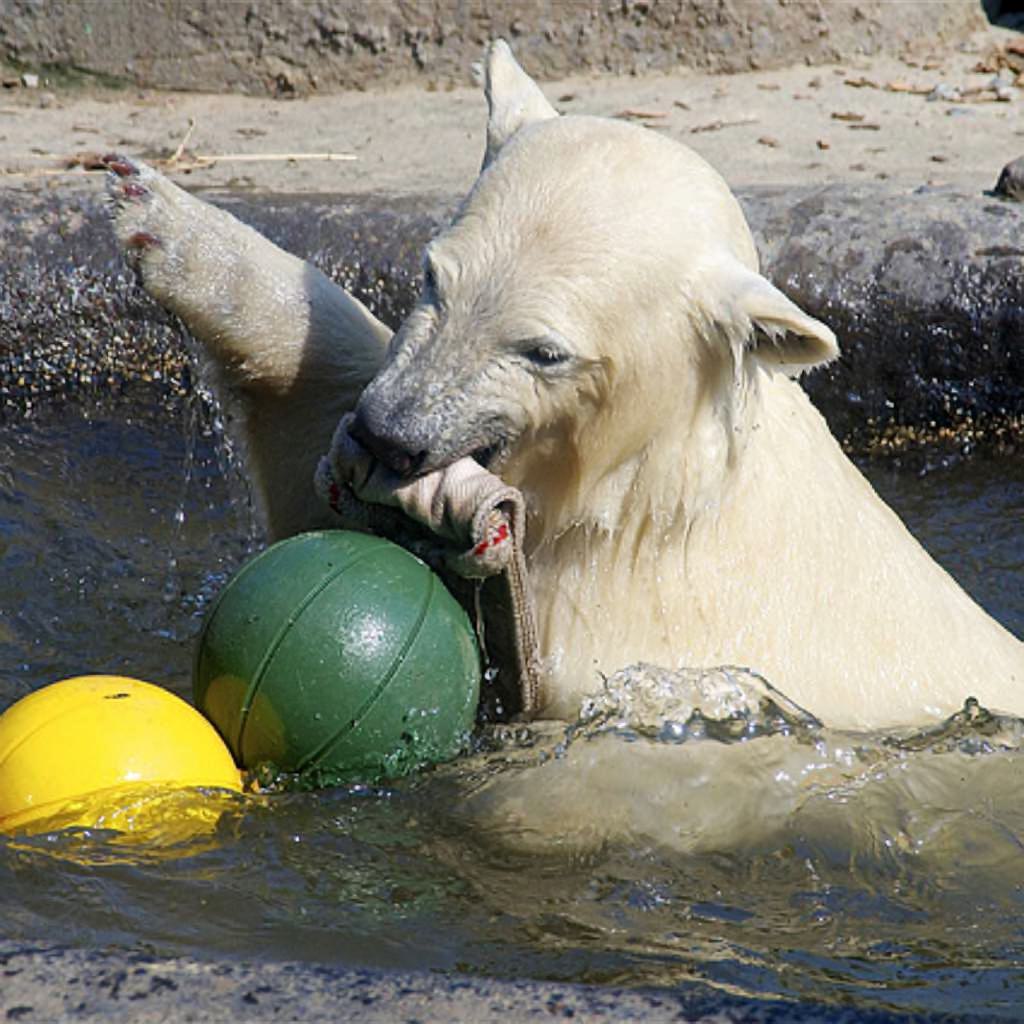}} &
        {\includegraphics[valign=c, width=\ww]{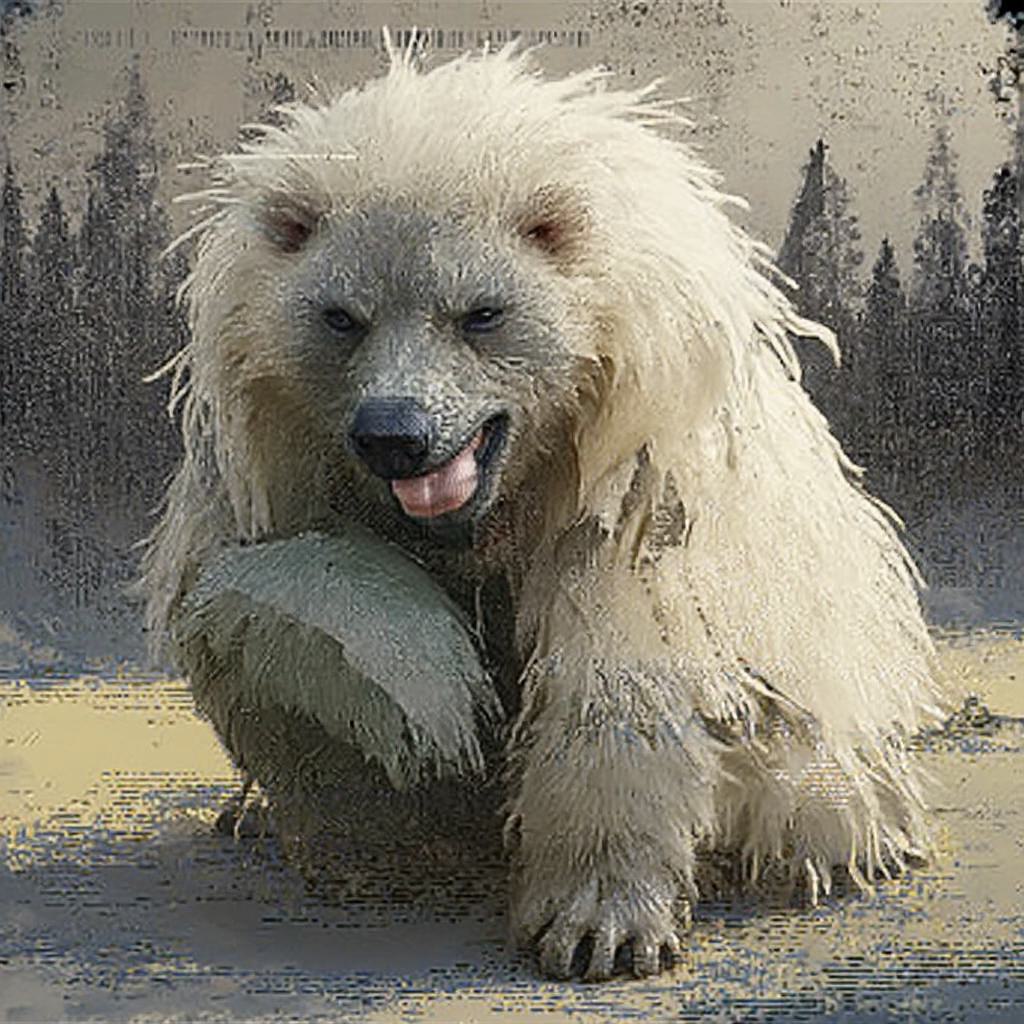}} &
        {\includegraphics[valign=c, width=\ww]{figures/qualitative_comparison_automatic/assets/bear/ours.jpg}}
        \vspace{1px}
        \\

        &
        \multicolumn{5}{c}{\small{\prompt{A photo of a bear with a long hair}}}
        \vspace{5px}
        \\

        {\includegraphics[valign=c, width=\ww]{figures/qualitative_comparison_automatic/assets/player/inp.jpg}} &
        {\includegraphics[valign=c, width=\ww]{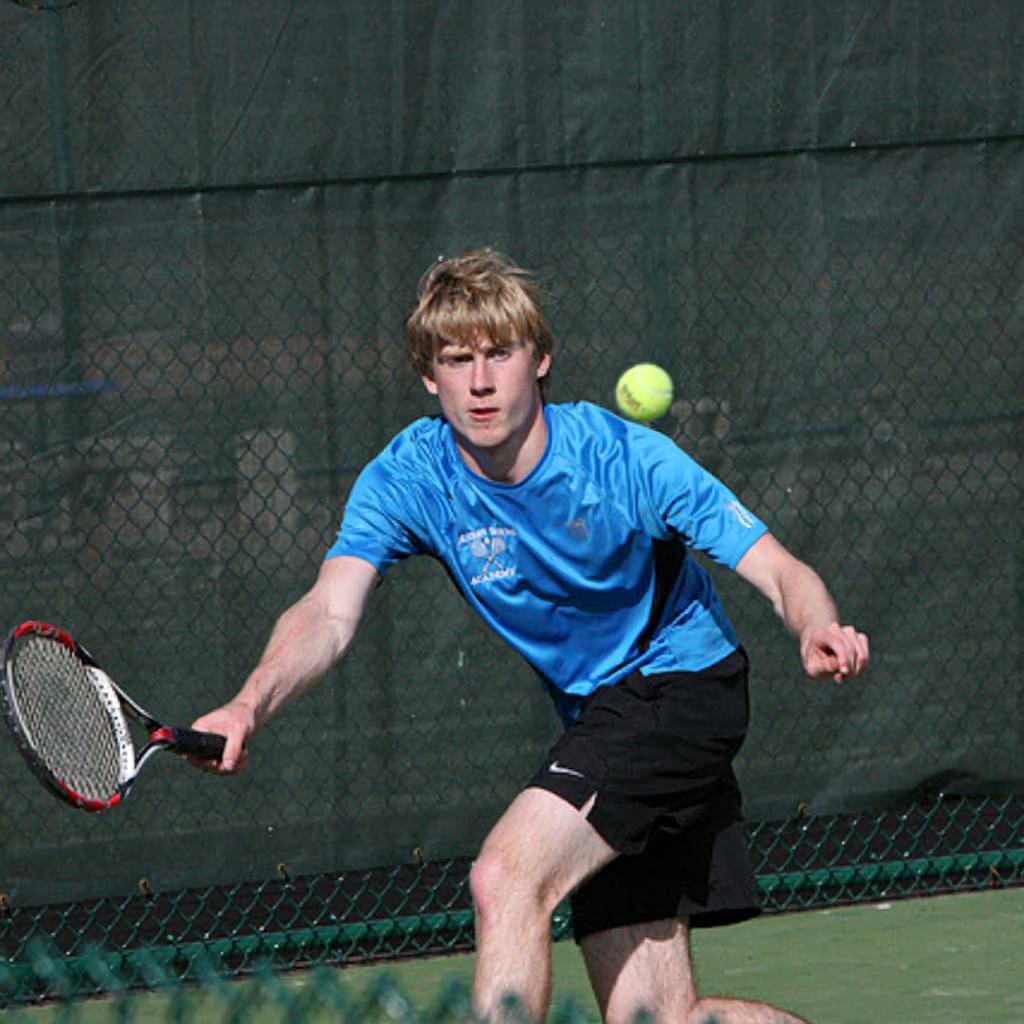}} &
        {\includegraphics[valign=c, width=\ww]{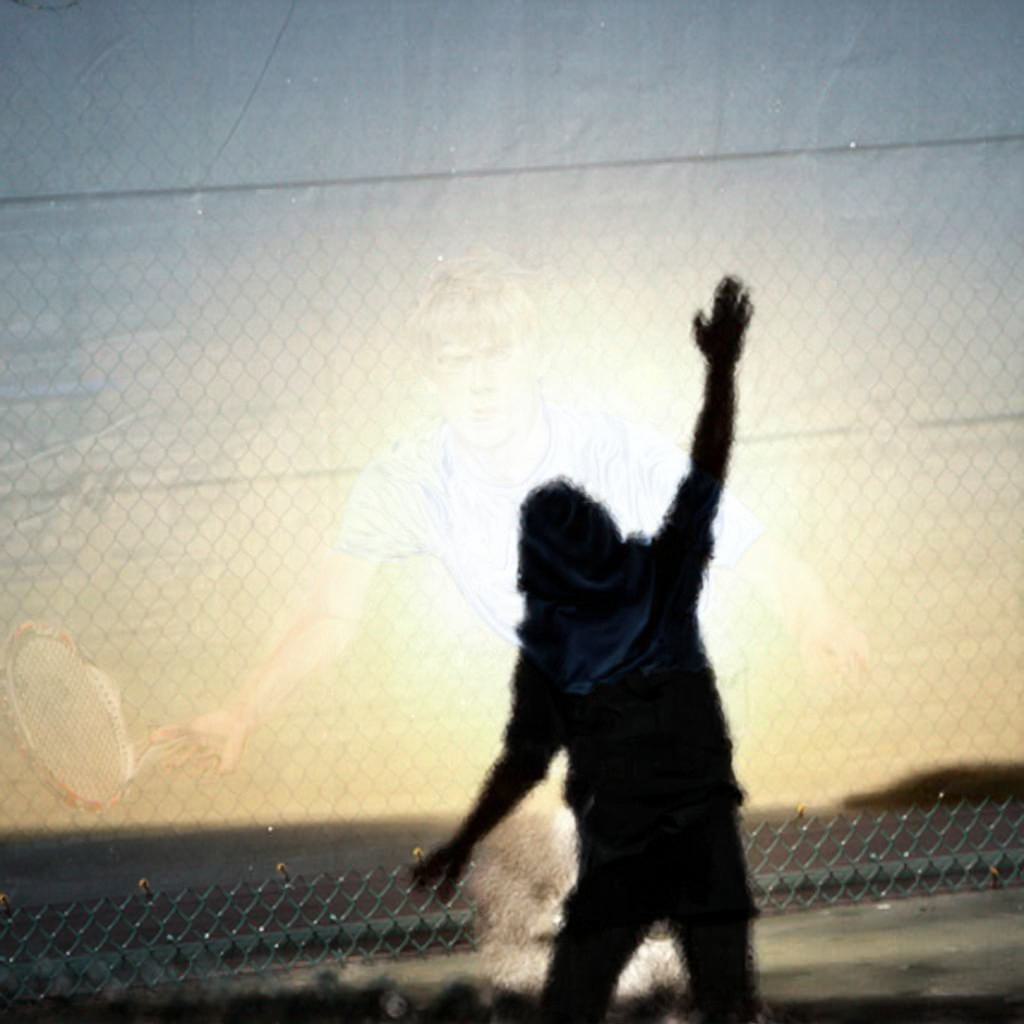}} &
        {\includegraphics[valign=c, width=\ww]{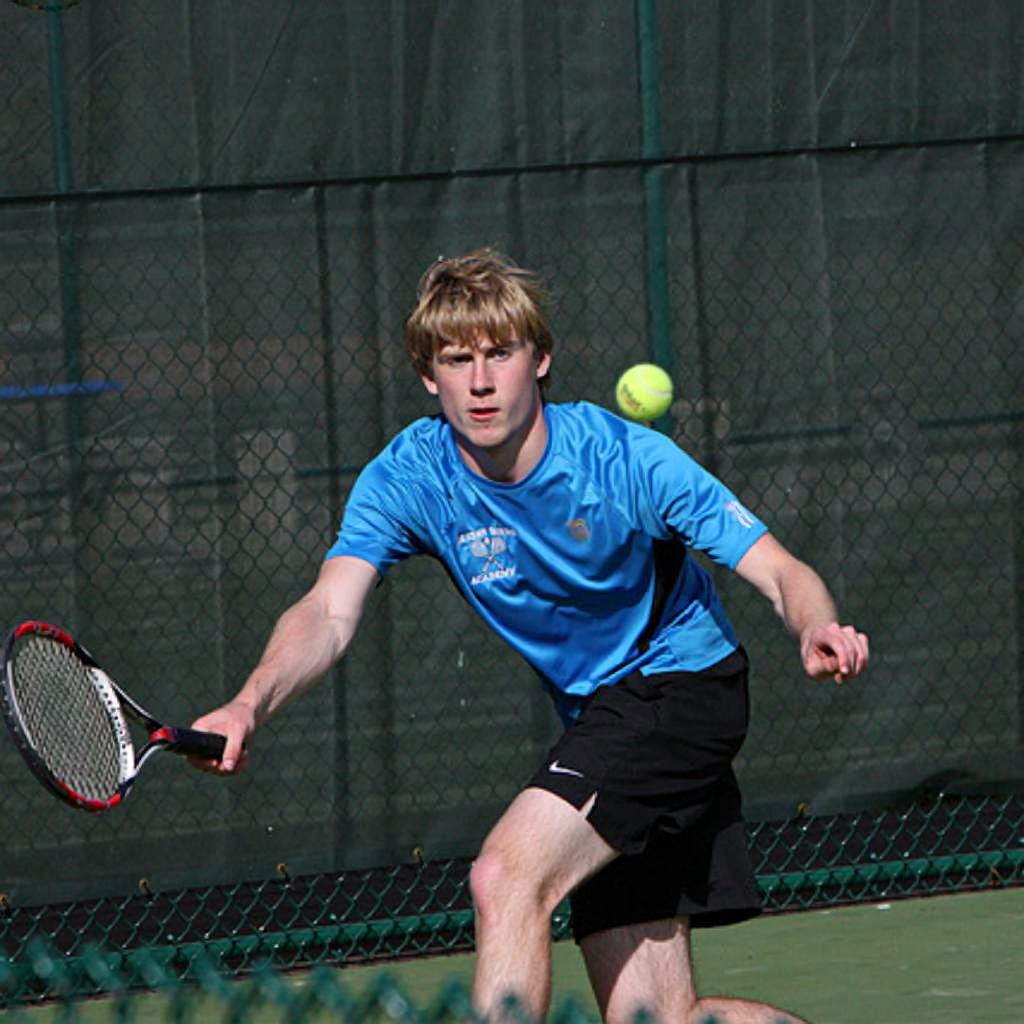}} &
        {\includegraphics[valign=c, width=\ww]{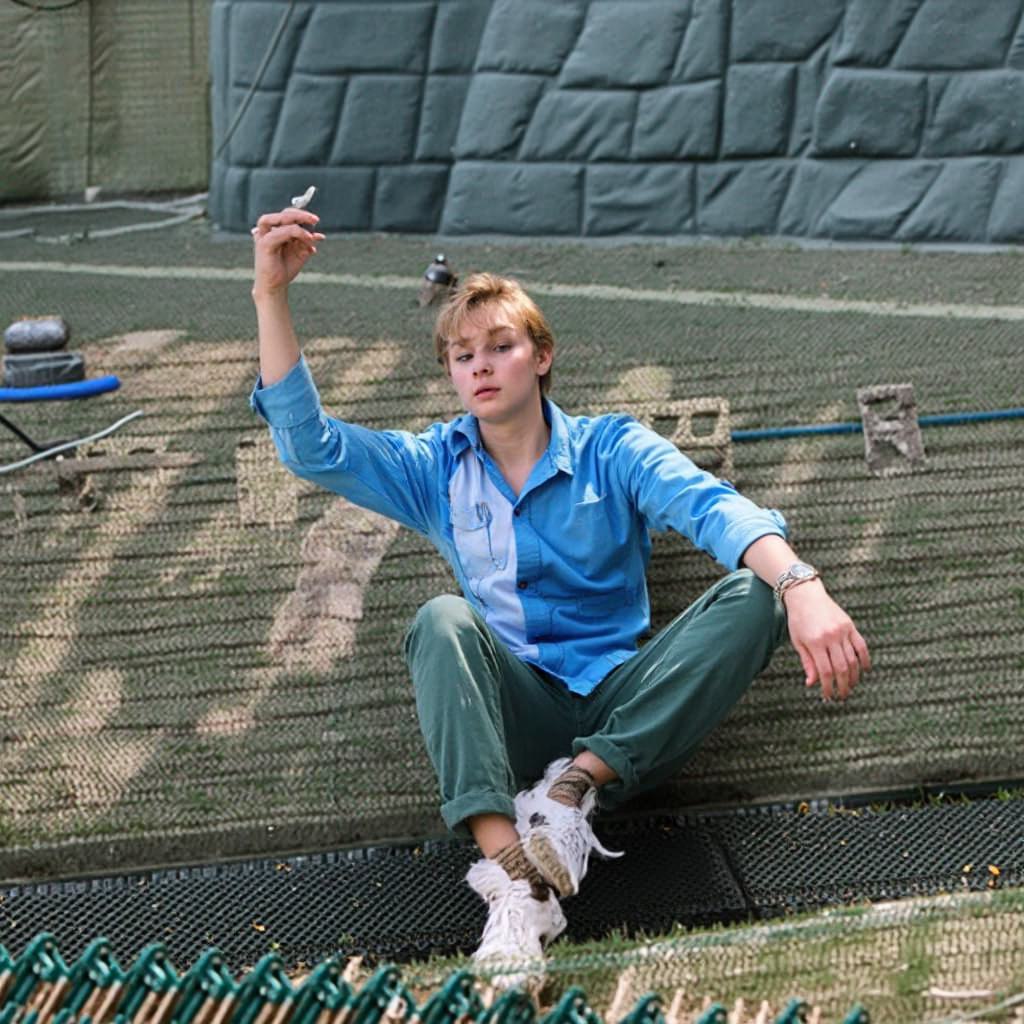}} &
        {\includegraphics[valign=c, width=\ww]{figures/qualitative_comparison_automatic/assets/player/ours.jpg}}
        \vspace{1px}
        \\

        &
        \multicolumn{5}{c}{\small{\prompt{A photo of a man raising his hand}}}
        \vspace{5px}
        \\

        {\includegraphics[valign=c, width=\ww]{figures/qualitative_comparison_automatic/assets/sheep/inp.jpg}} &
        {\includegraphics[valign=c, width=\ww]{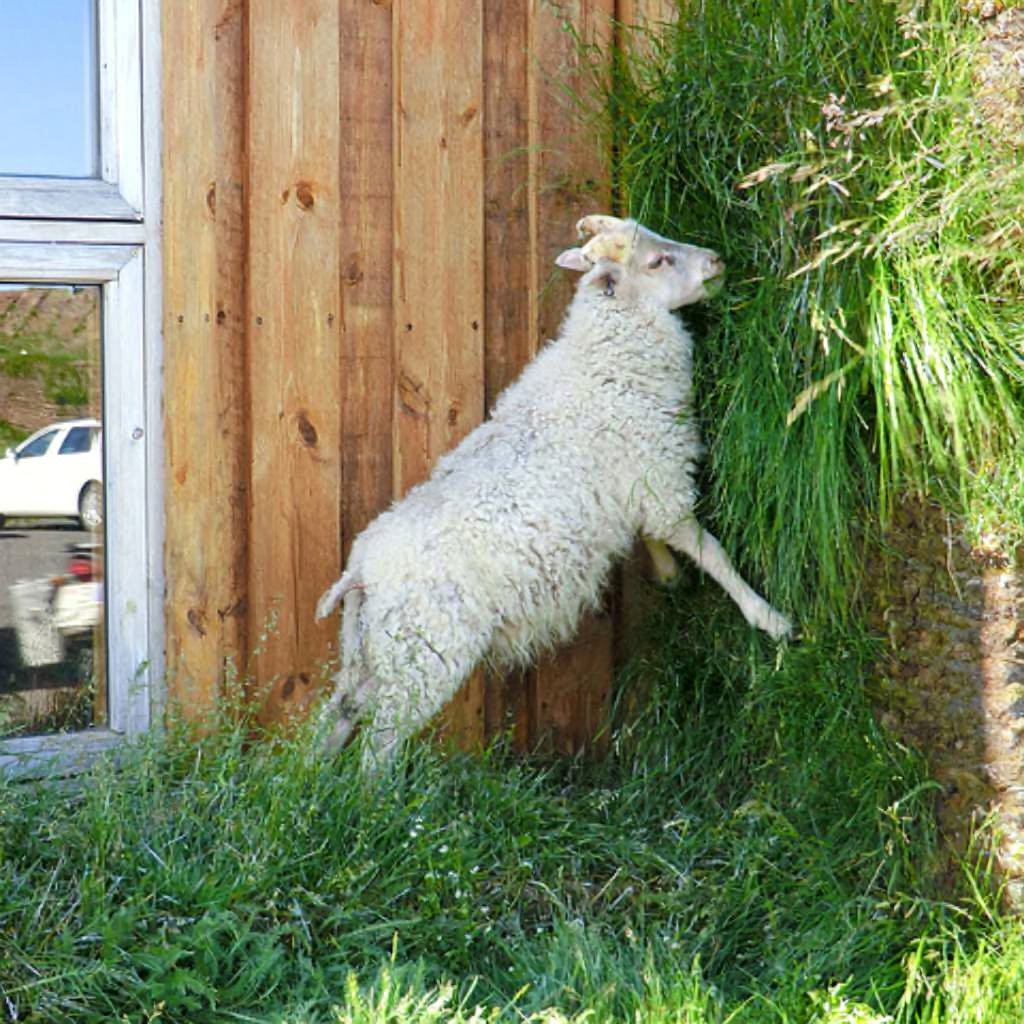}} &
        {\includegraphics[valign=c, width=\ww]{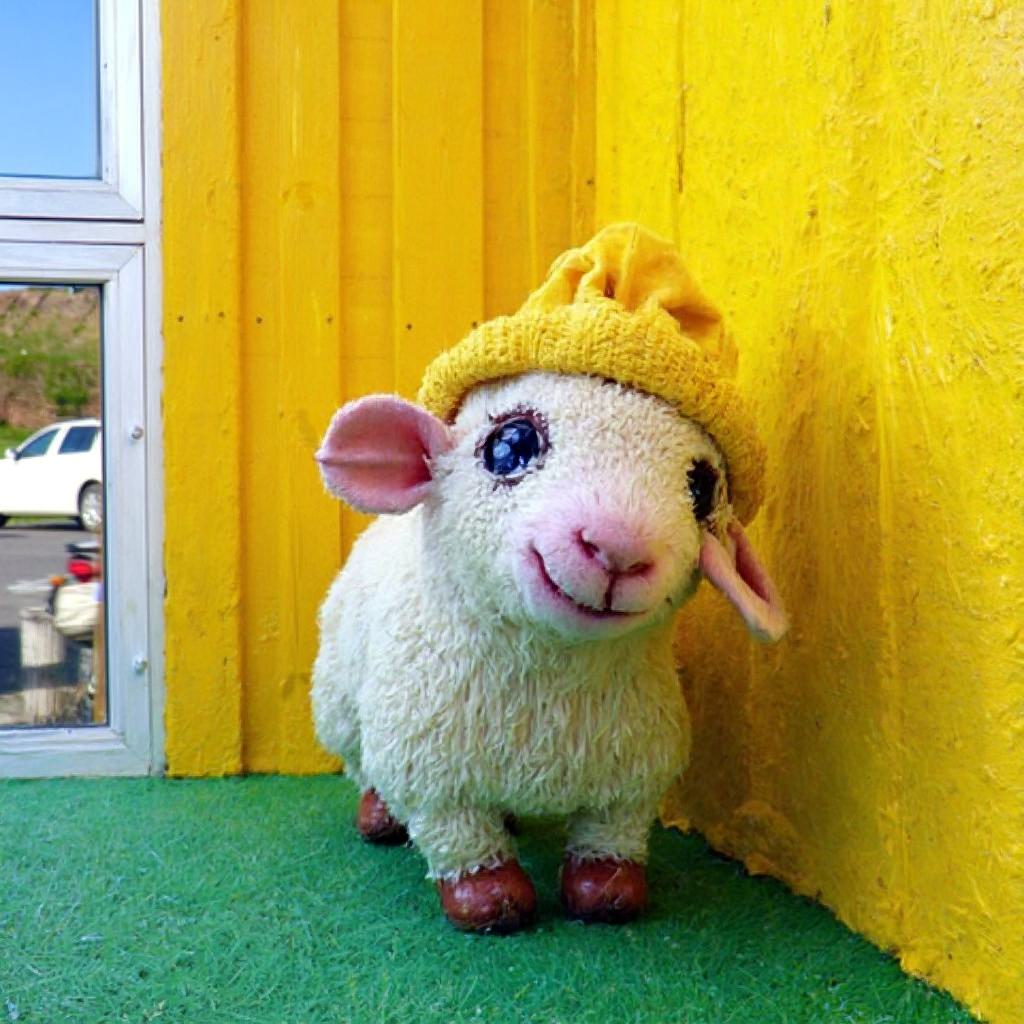}} &
        {\includegraphics[valign=c, width=\ww]{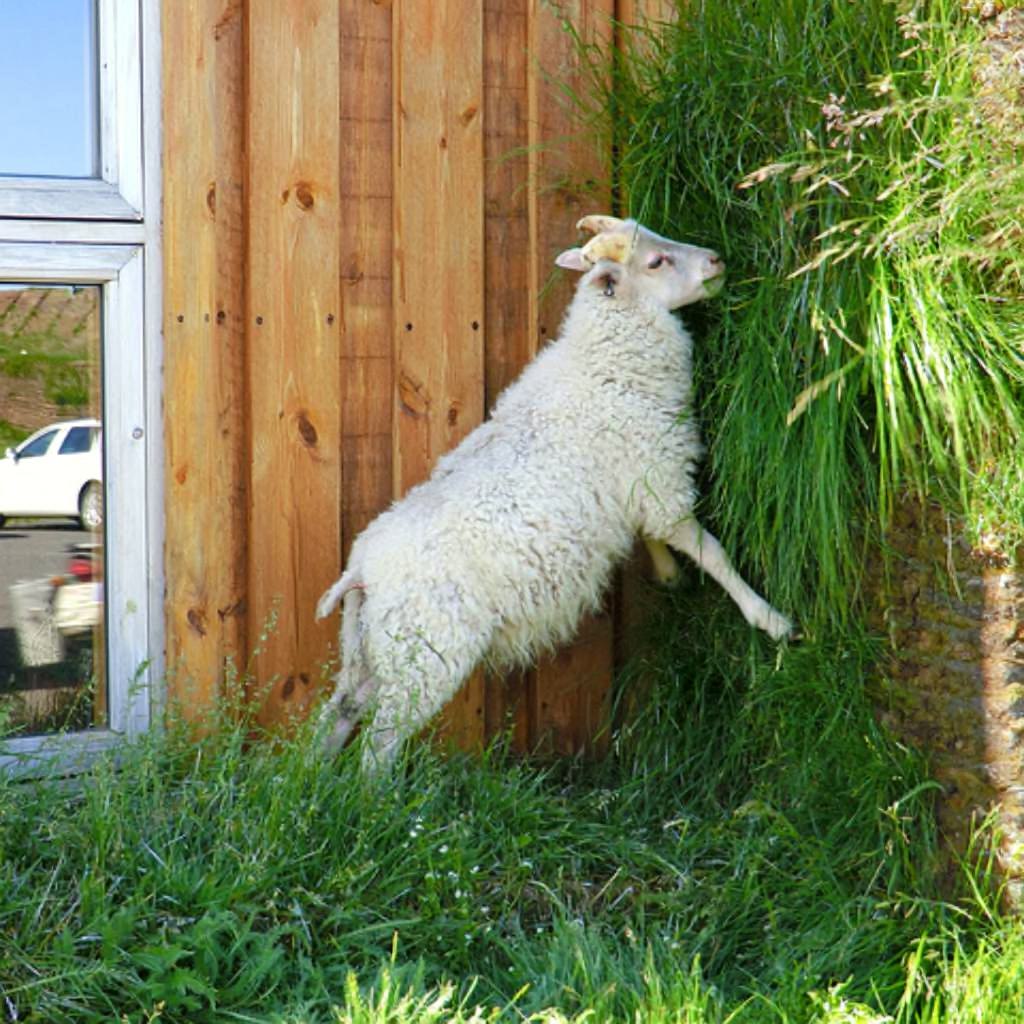}} &
        {\includegraphics[valign=c, width=\ww]{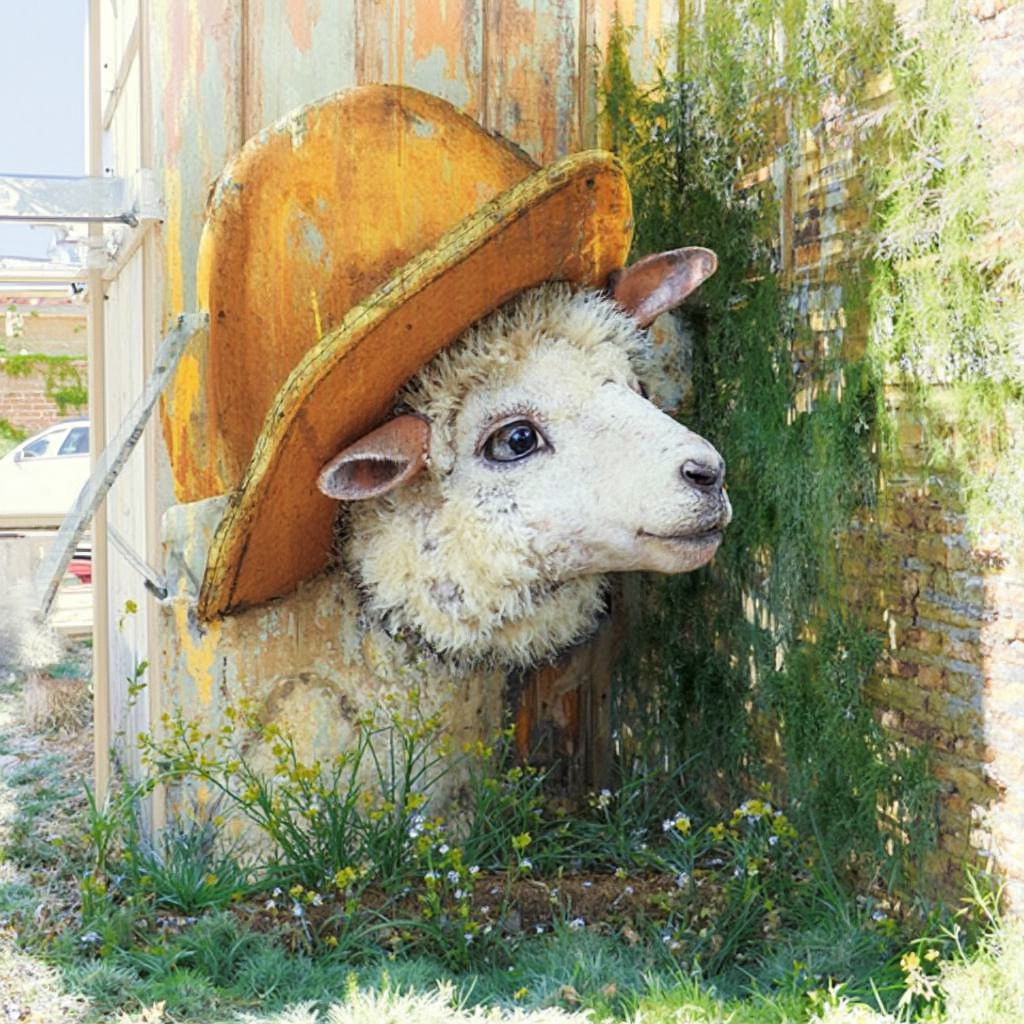}} &
        {\includegraphics[valign=c, width=\ww]{figures/qualitative_comparison_automatic/assets/sheep/ours.jpg}}
        \vspace{1px}
        \\

        &
        \multicolumn{5}{c}{\small{\prompt{A photo of a sheep with a yellow hat}}}
        \vspace{5px}
        \\

        {\includegraphics[valign=c, width=\ww]{figures/qualitative_comparison_automatic/assets/cat/inp.jpg}} &
        {\includegraphics[valign=c, width=\ww]{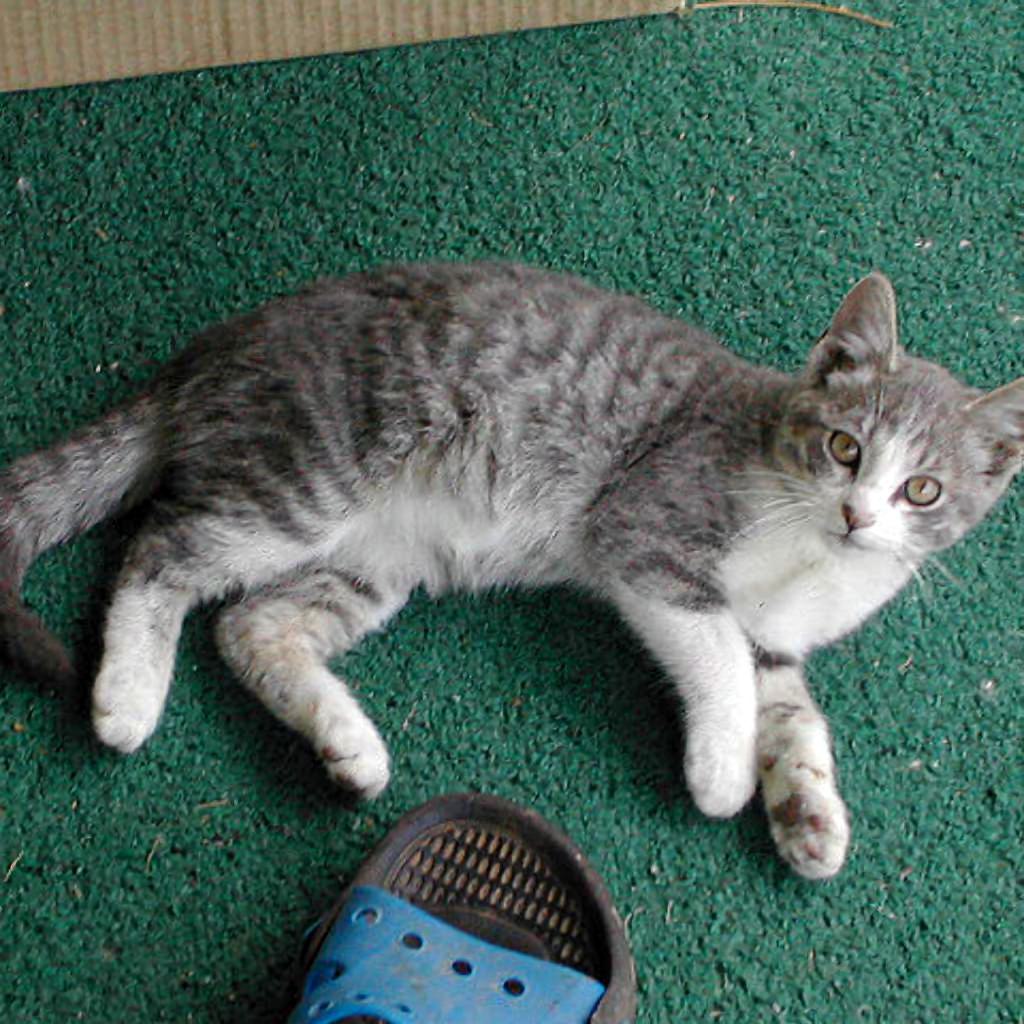}} &
        {\includegraphics[valign=c, width=\ww]{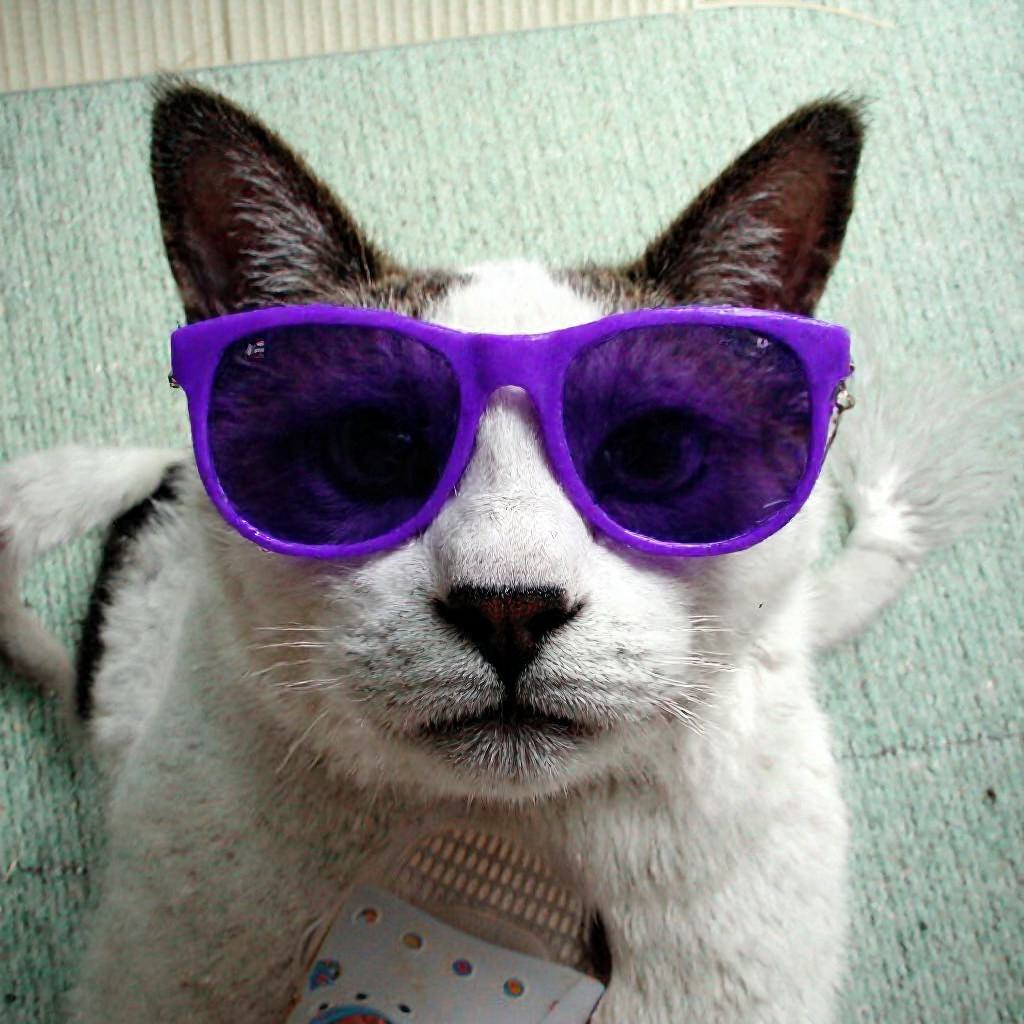}} &
        {\includegraphics[valign=c, width=\ww]{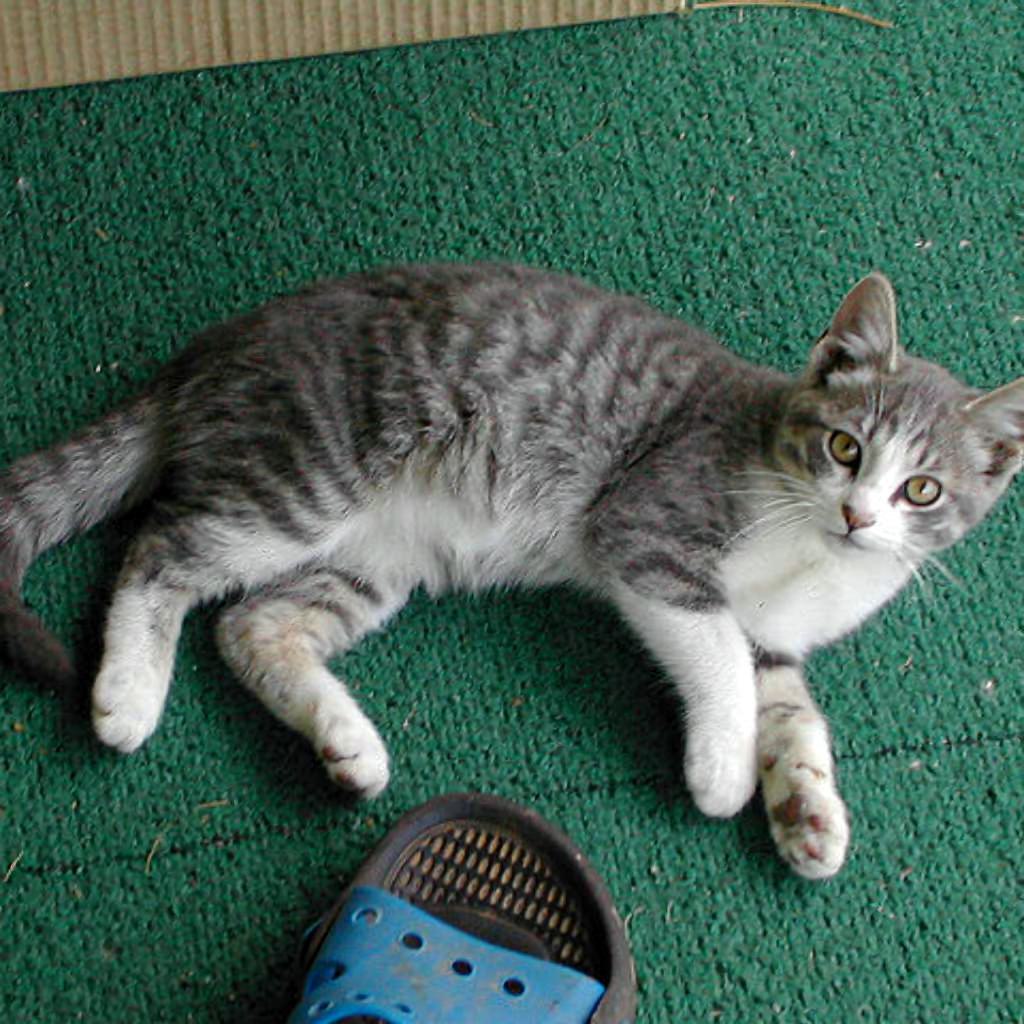}} &
        {\includegraphics[valign=c, width=\ww]{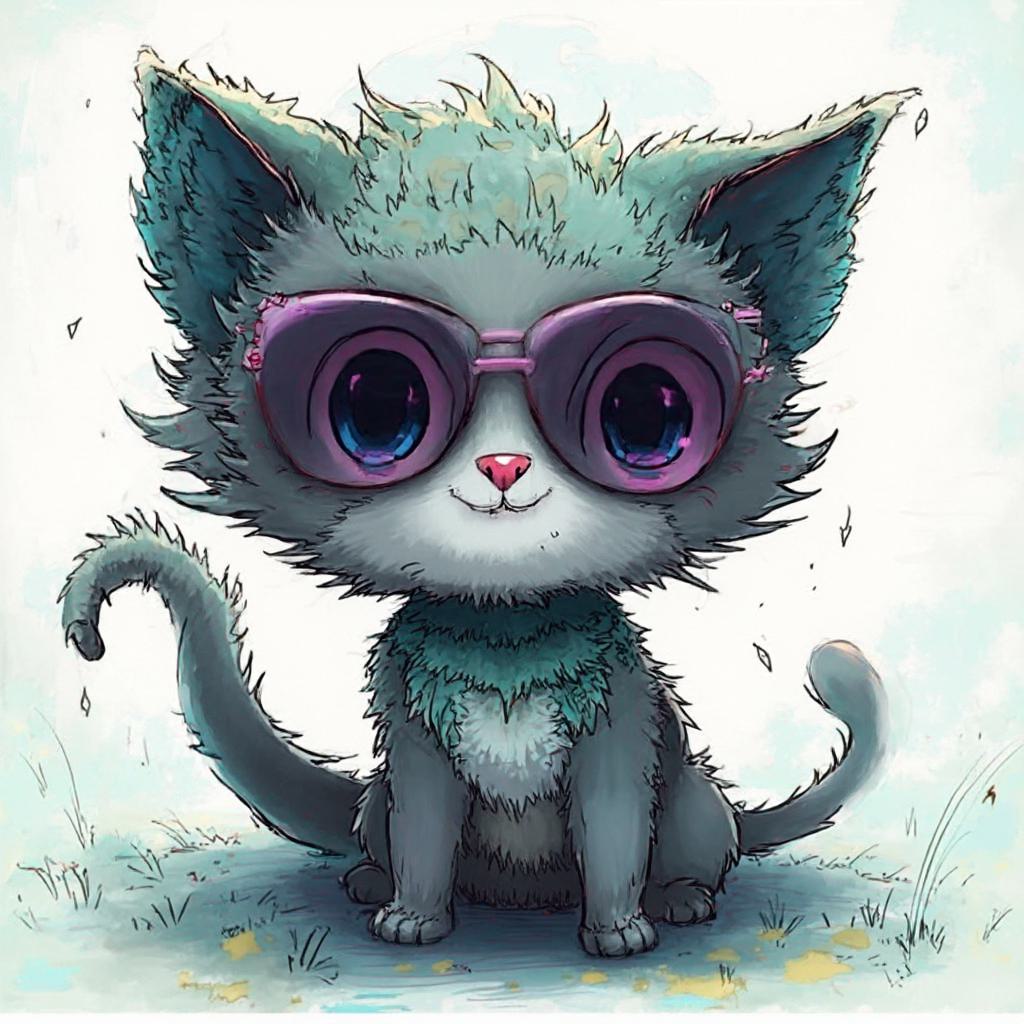}} &
        {\includegraphics[valign=c, width=\ww]{figures/qualitative_comparison_automatic/assets/cat/ours.jpg}}
        \vspace{1px}
        \\

        &
        \multicolumn{5}{c}{\small{\prompt{A photo of a cat wearing purple sunglasses}}}
        \vspace{5px}
        \\

        {\includegraphics[valign=c, width=\ww]{figures/qualitative_comparison_automatic/assets/chicken/inp.jpg}} &
        {\includegraphics[valign=c, width=\ww]{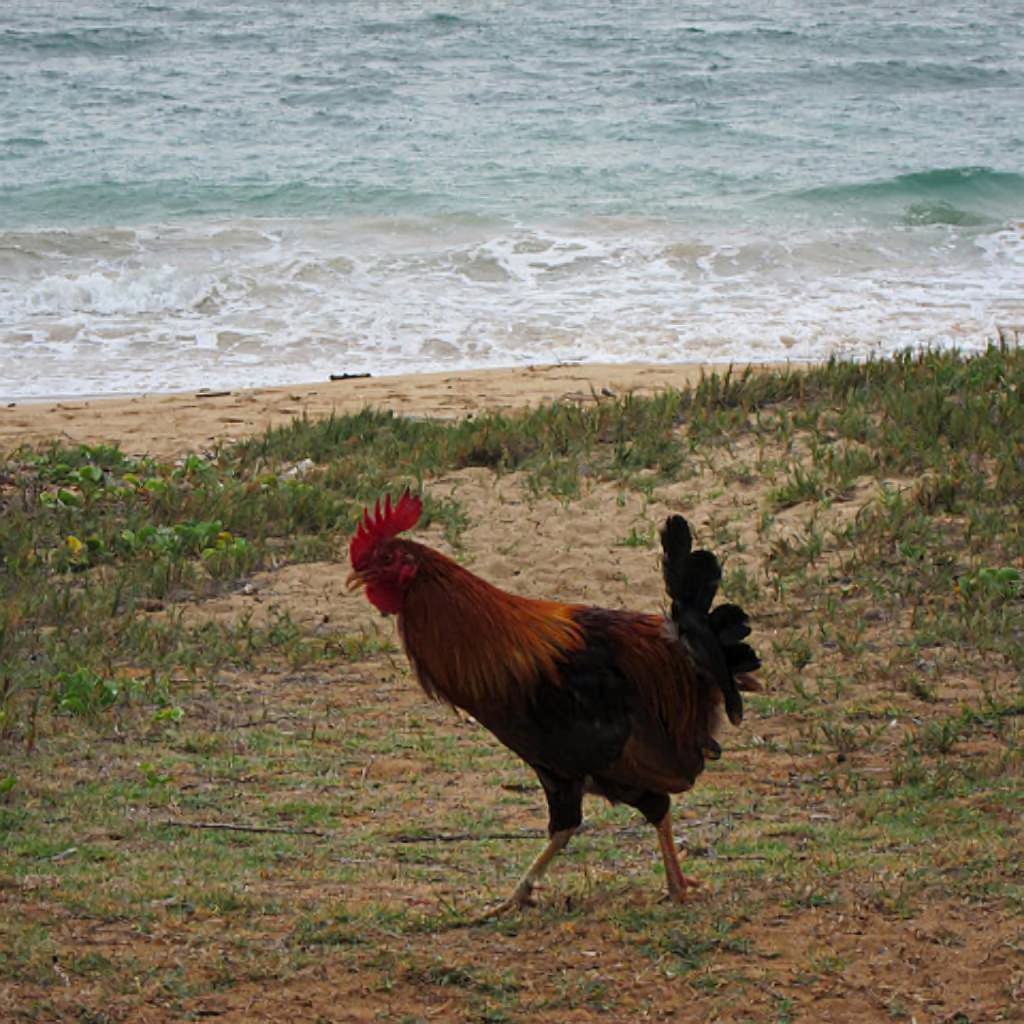}} &
        {\includegraphics[valign=c, width=\ww]{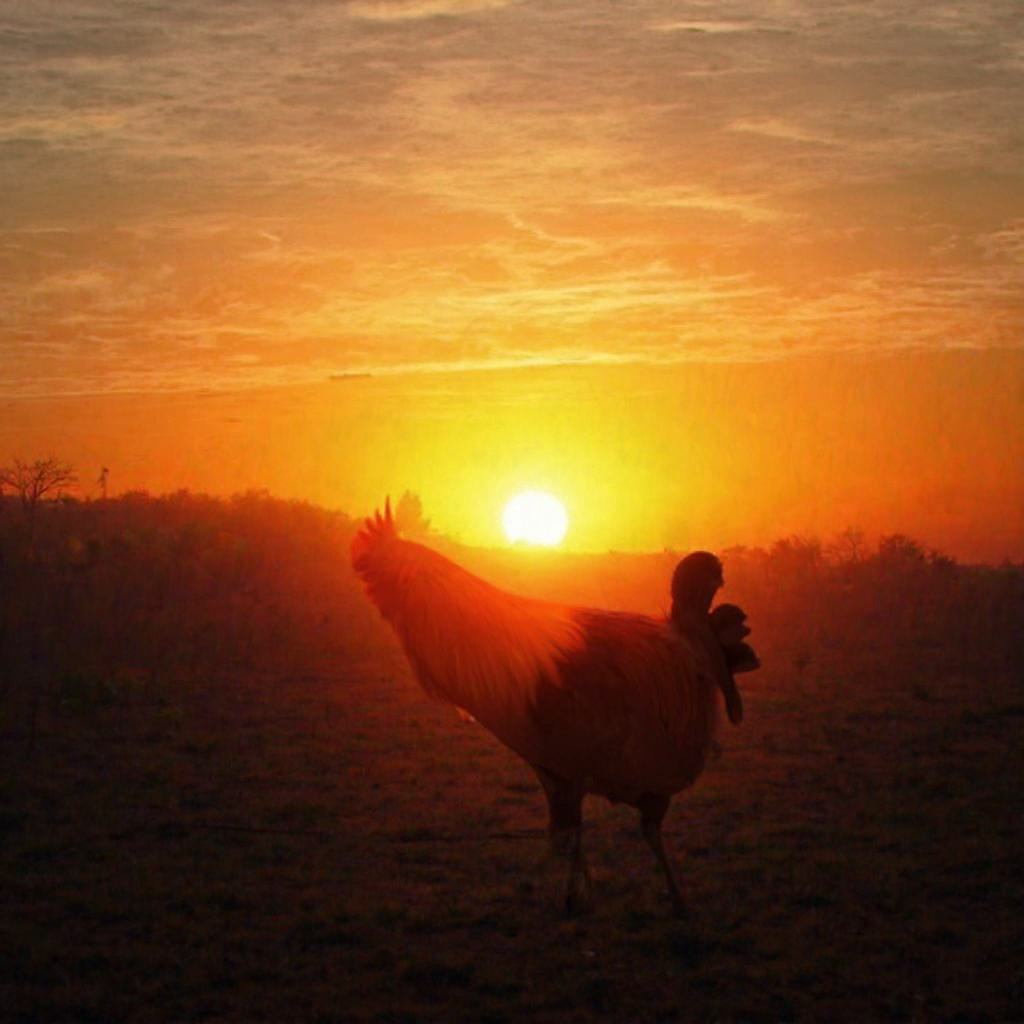}} &
        {\includegraphics[valign=c, width=\ww]{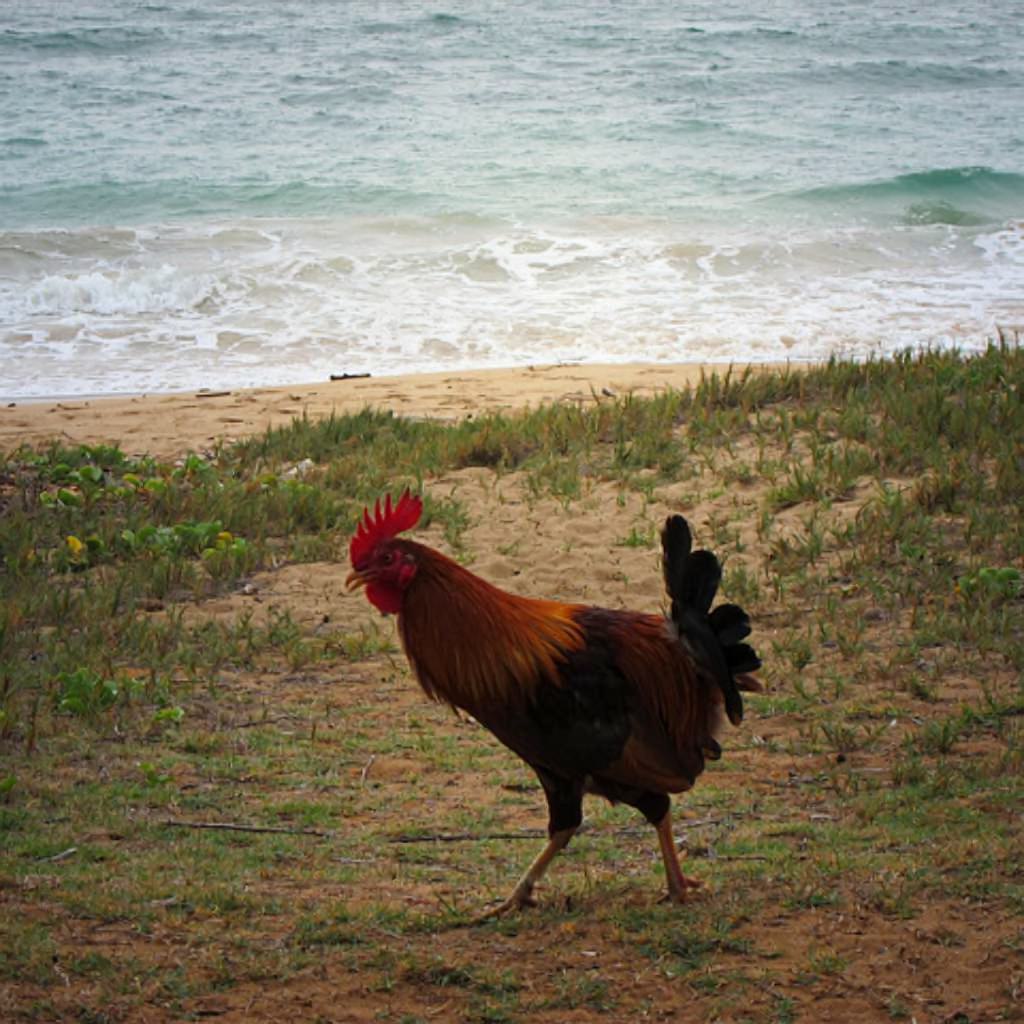}} &
        {\includegraphics[valign=c, width=\ww]{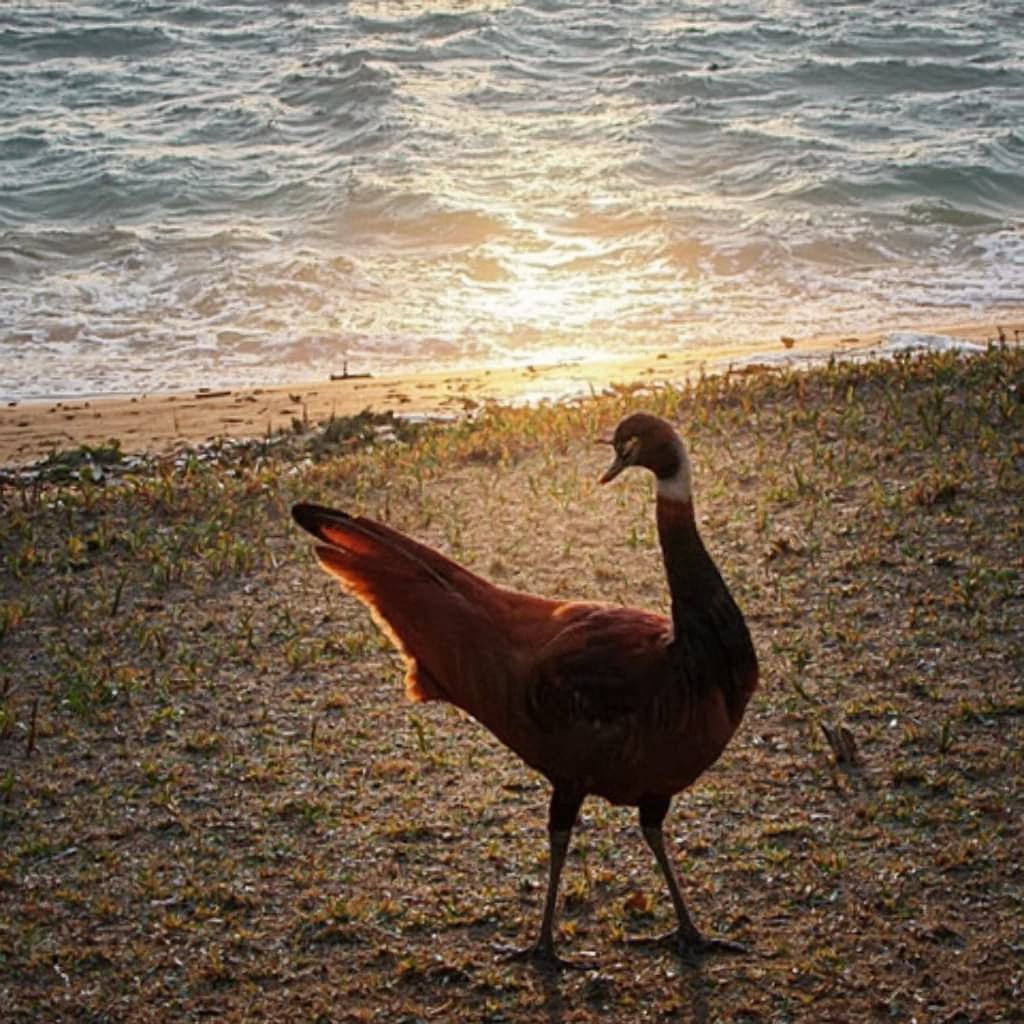}} &
        {\includegraphics[valign=c, width=\ww]{figures/qualitative_comparison_automatic/assets/chicken/ours.jpg}}
        \vspace{1px}
        \\

        &
        \multicolumn{5}{c}{\small{\prompt{A photo of a chicken during sunset}}}
        \vspace{5px}
        \\

        {\includegraphics[valign=c, width=\ww]{figures/qualitative_comparison_automatic/assets/elephant/inp.jpg}} &
        {\includegraphics[valign=c, width=\ww]{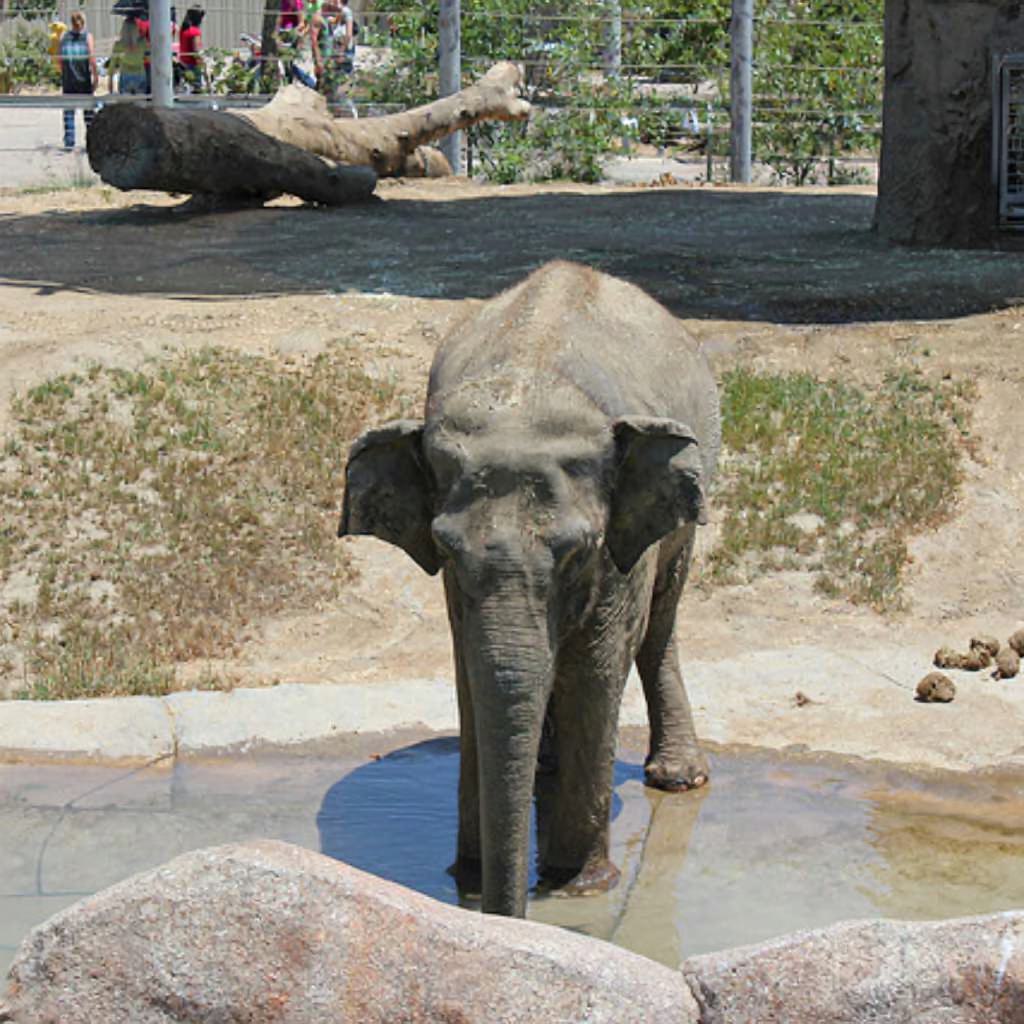}} &
        {\includegraphics[valign=c, width=\ww]{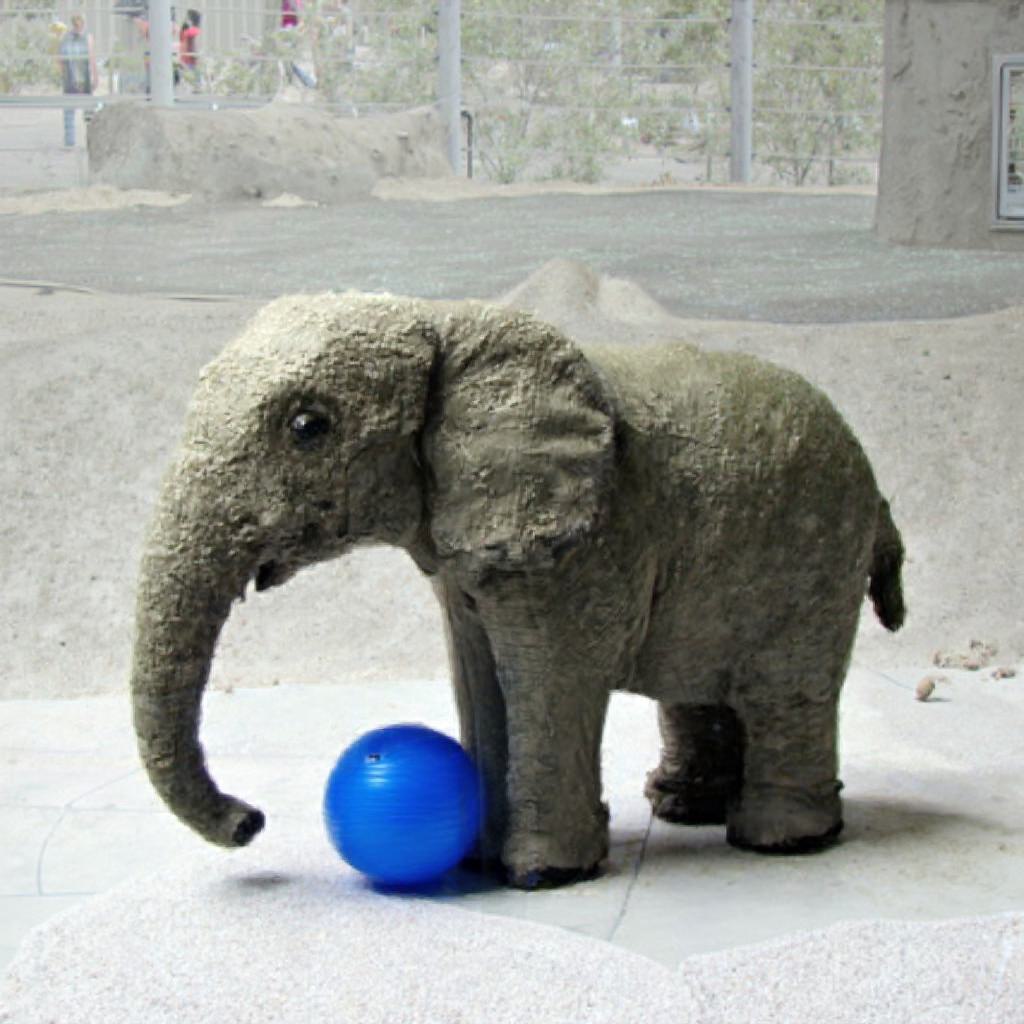}} &
        {\includegraphics[valign=c, width=\ww]{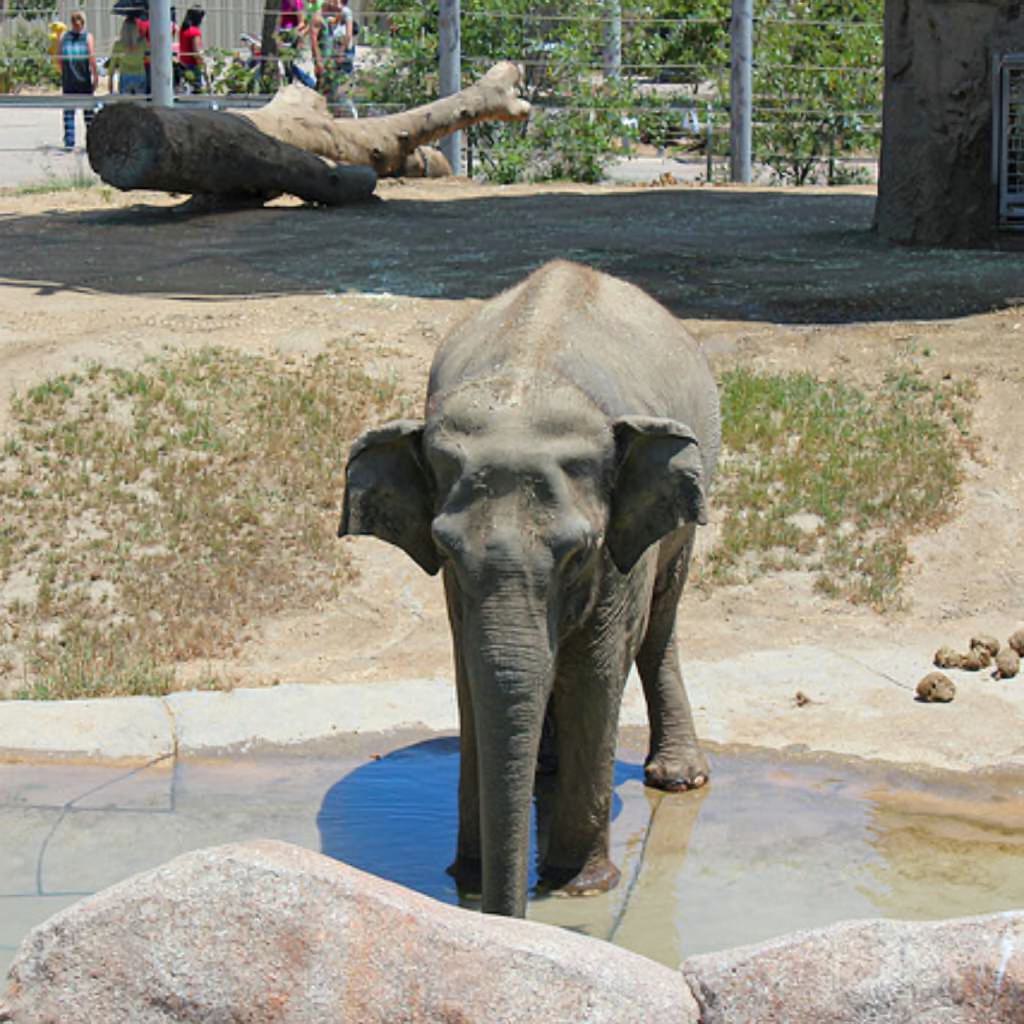}} &
        {\includegraphics[valign=c, width=\ww]{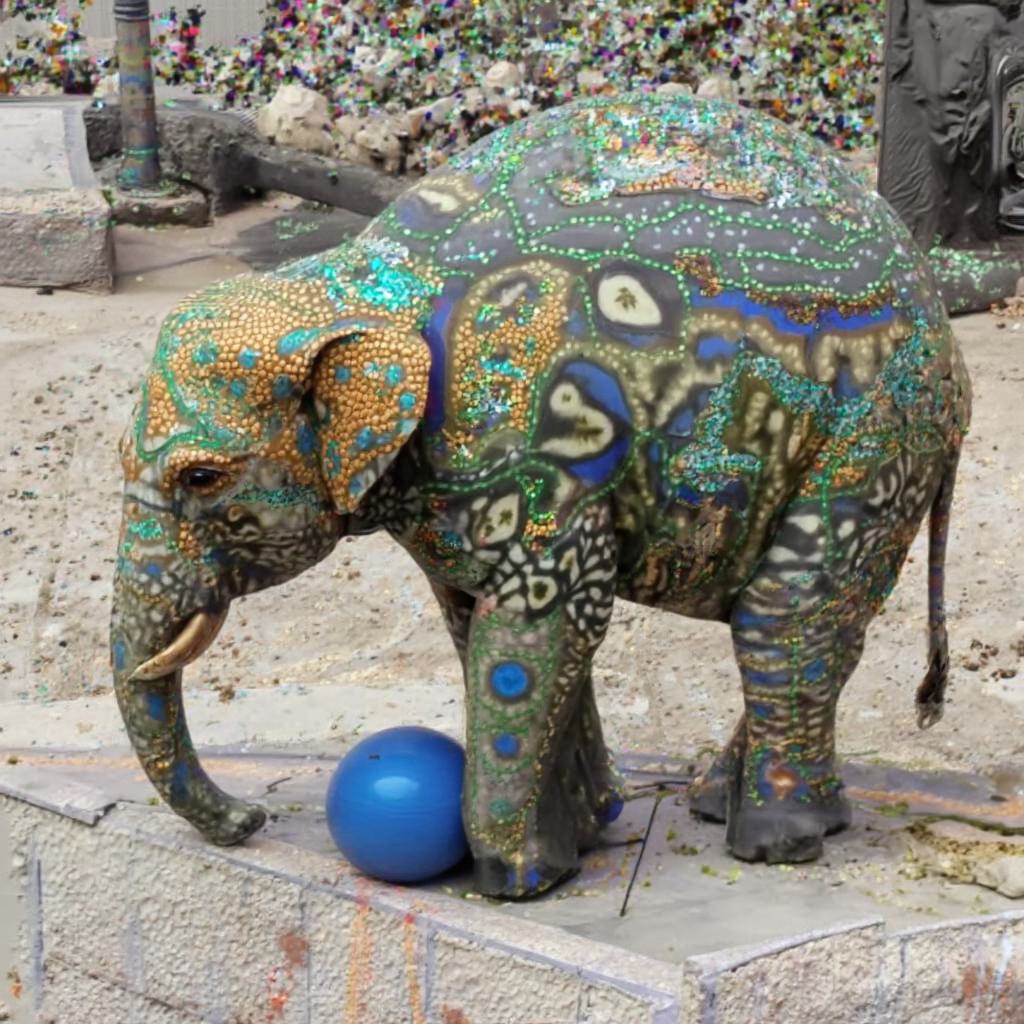}} &
        {\includegraphics[valign=c, width=\ww]{figures/qualitative_comparison_automatic/assets/elephant/ours.jpg}}
        \vspace{1px}
        \\

        &
        \multicolumn{5}{c}{\small{\prompt{A photo of an elephant next to a blue ball}}}
        \\

    \end{tabular}
    \vspace{-3px}
    \caption{\textbf{Ablations Qualitative Comparison on Automatic Dataset.} As explained in 
    \Cref{sec:ablation_study},
    we compare our method against several ablation cases on real images extracted from the COCO~\cite{Lin2014MicrosoftCC} dataset. As can be seen, we found that (1) performing attention injection in all the layers or performing (3) an attention extension in all the layers encourages the model to directly copy the input image while neglecting the target prompt. In addition, (2) performing an attention extension in the non-vital layers or (4) removing the latent nudging reduces the input image similarity significantly.}
    \label{fig:ablation_qualitative_comparison_automatic}
\end{figure*}

\begin{figure*}[tp]
    \centering
    \setlength{\tabcolsep}{0.6pt}
    \renewcommand{\arraystretch}{0.8}
    \setlength{\ww}{0.24\linewidth}
    \begin{tabular}{c @{\hspace{10\tabcolsep}} ccc}
        \includegraphics[valign=c, width=\ww]{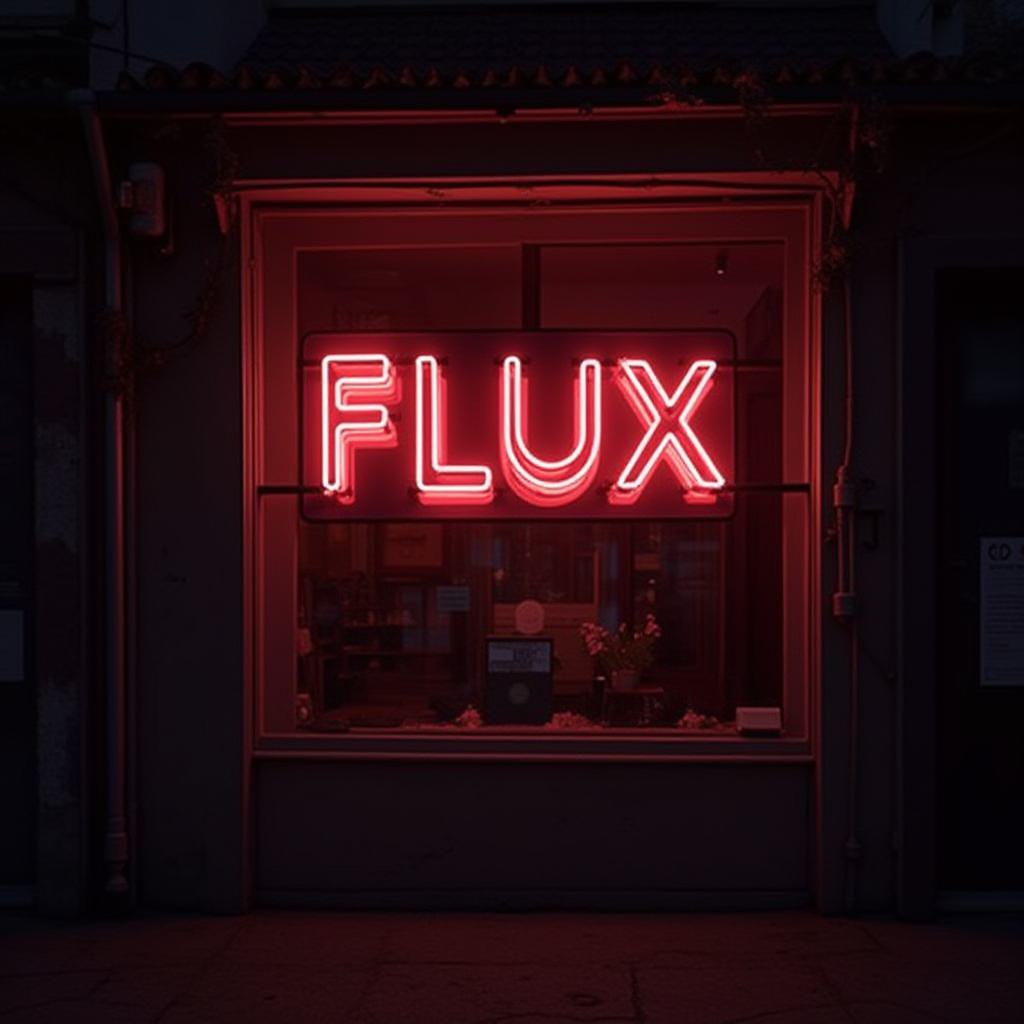} &
        \includegraphics[valign=c, width=\ww]{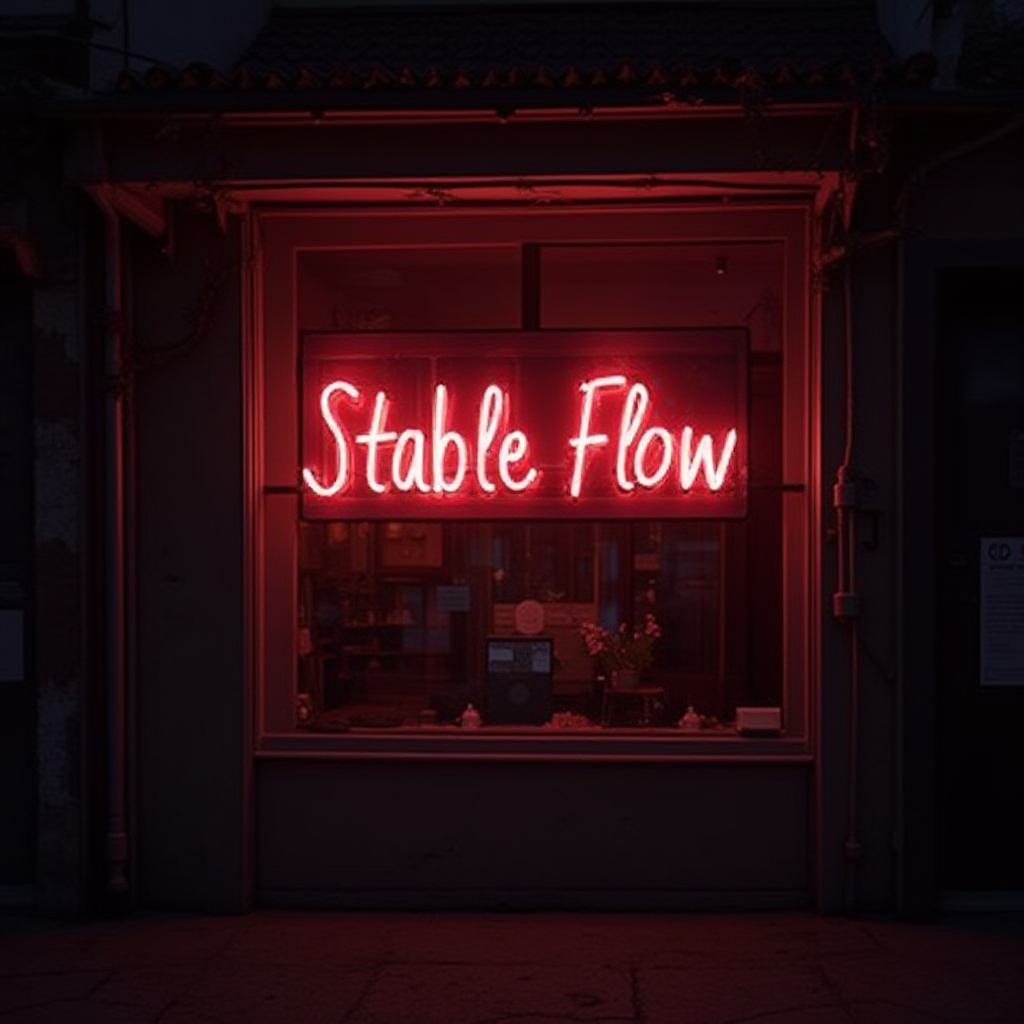} &
        \includegraphics[valign=c, width=\ww]{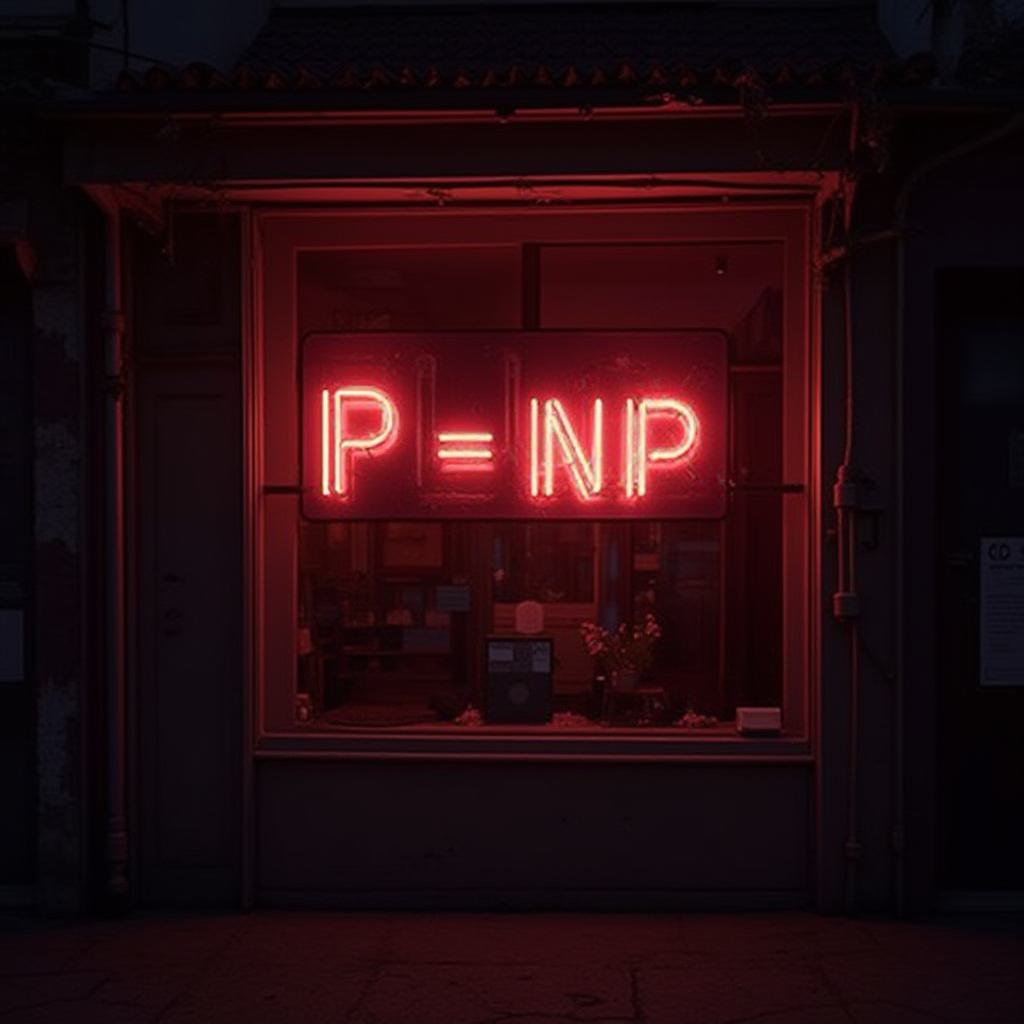} &
        \includegraphics[valign=c, width=\ww]{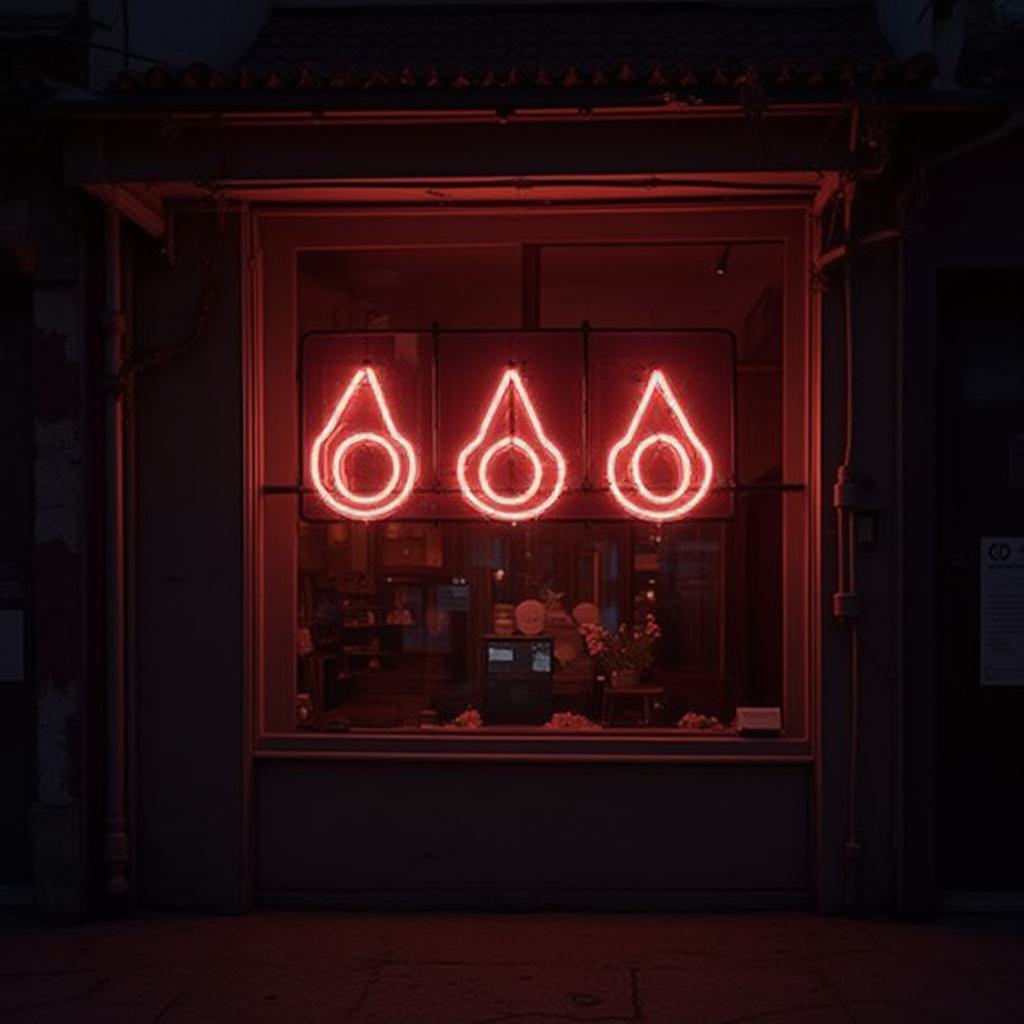}
        \vspace{3px}
        \\

        \small{Input} &
        \small{\prompt{A `Stable Flow' neon sign}} &
        \small{\prompt{A `P = NP' neon sign}} &
        \small{\prompt{A neon sign of avocados}}
        \vspace{10px}
        \\

        \includegraphics[valign=c, width=\ww]{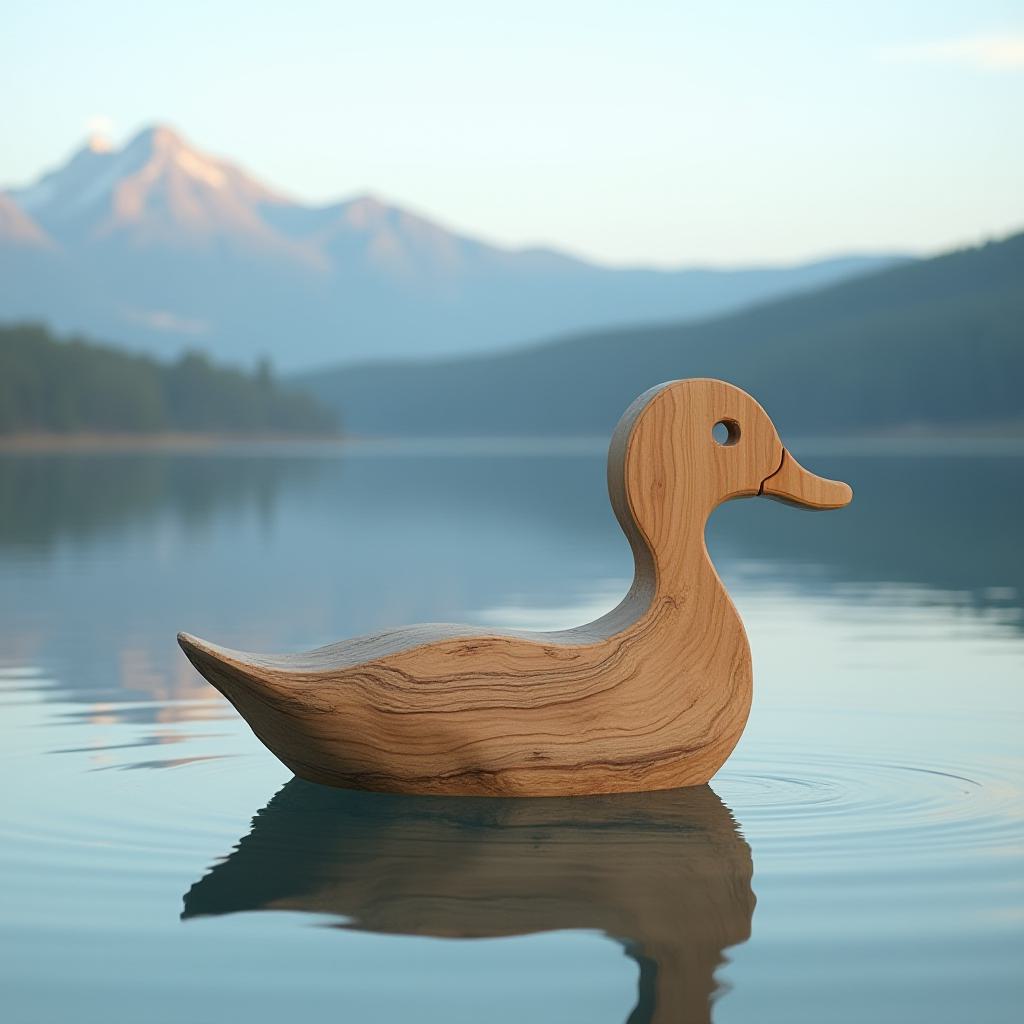} &
        \includegraphics[valign=c, width=\ww]{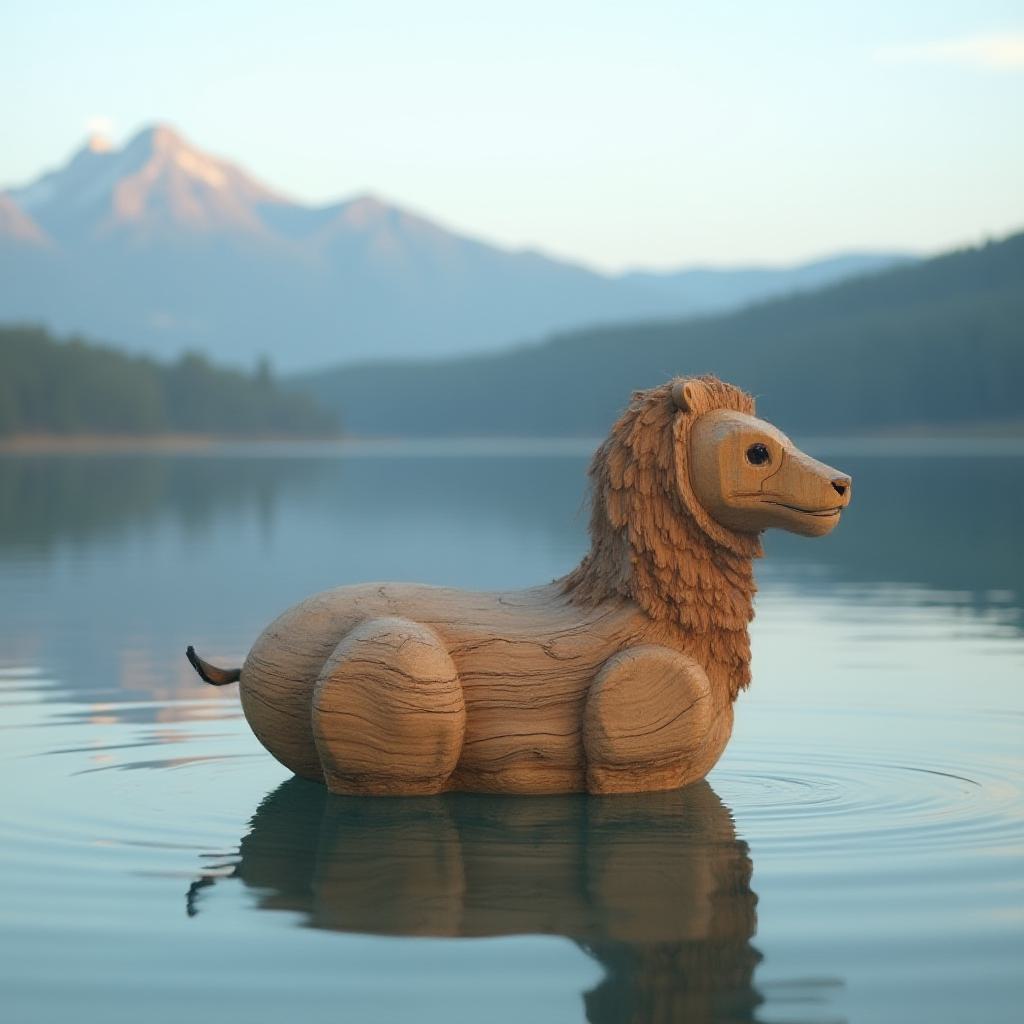} &
        \includegraphics[valign=c, width=\ww]{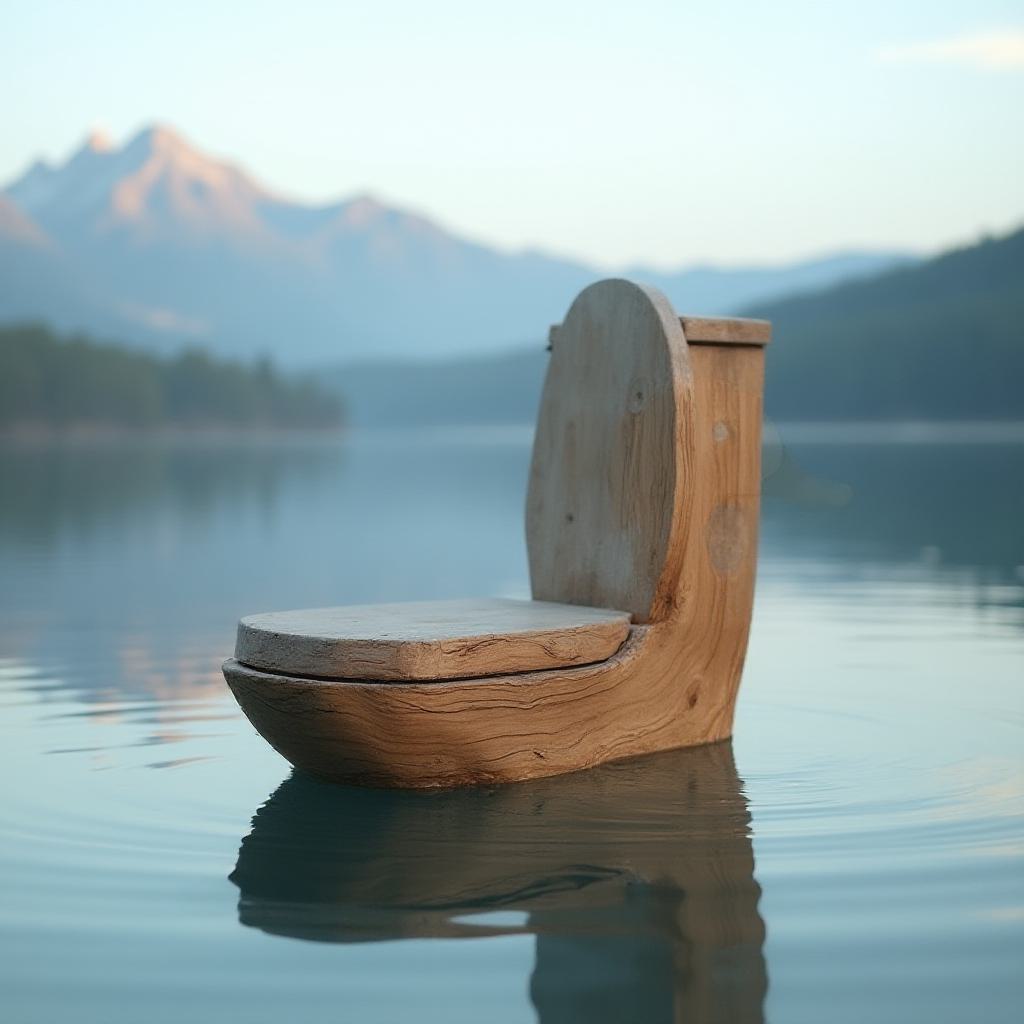} &
        \includegraphics[valign=c, width=\ww]{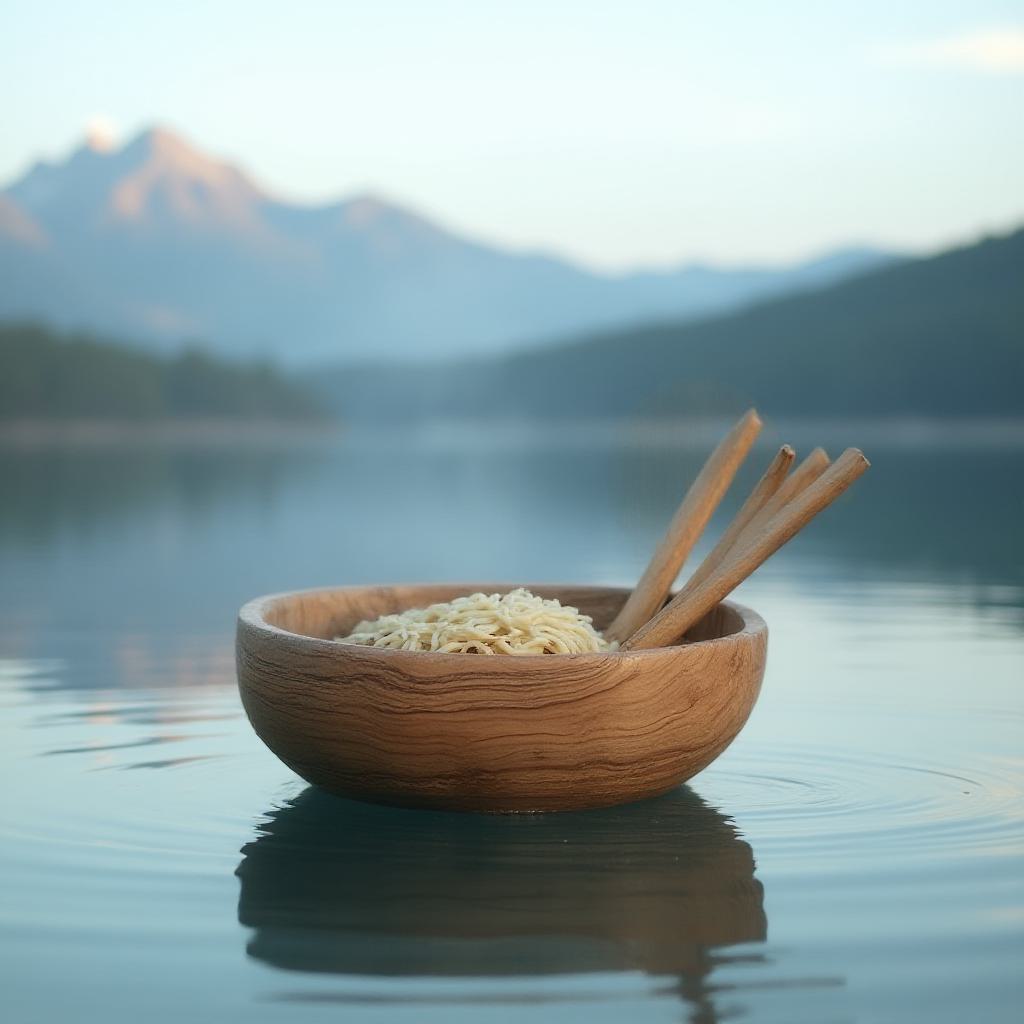}
        \vspace{3px}
        \\

        \small{Input} &
        \small{\prompt{A wooden lion}} &
        \small{\prompt{A wooden toilet}} &
        \small{\prompt{A wooden noodles bowl}}
        \vspace{10px}
        \\

        \includegraphics[valign=c, width=\ww]{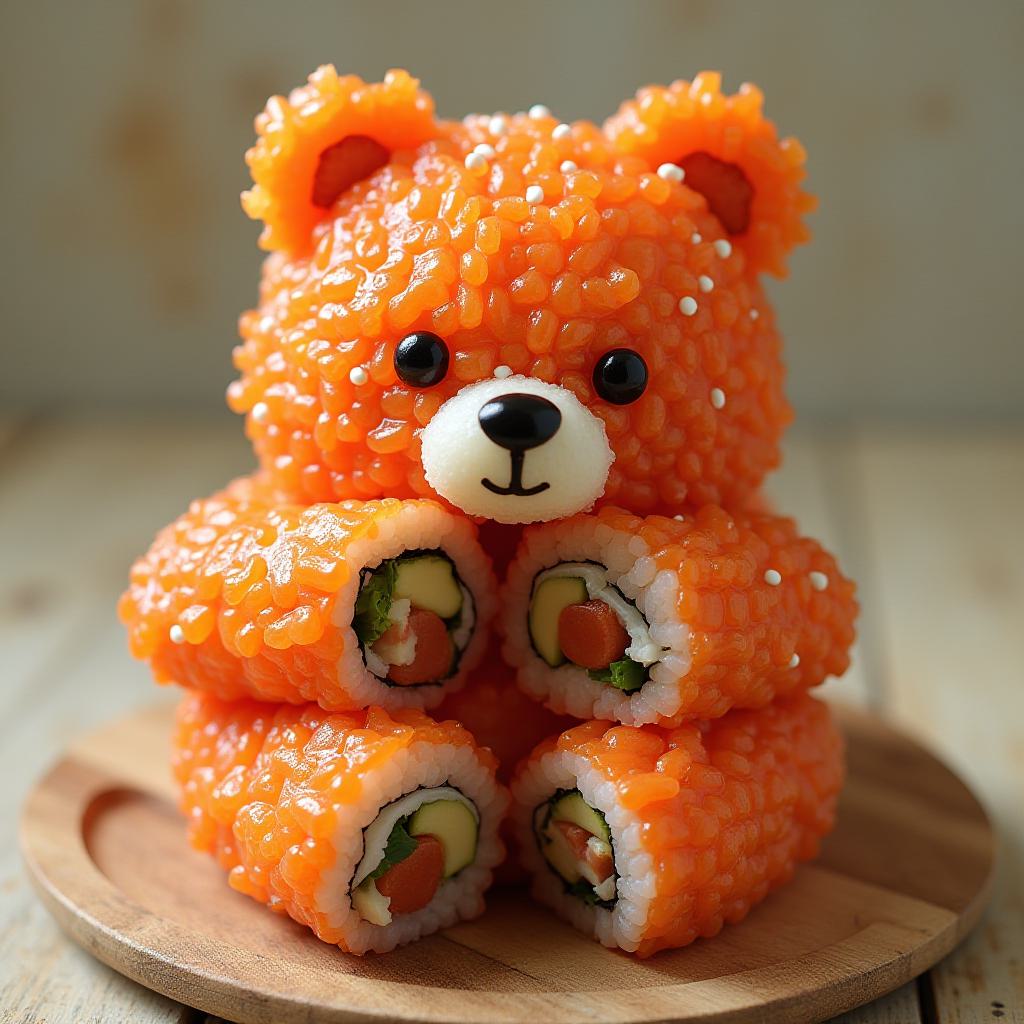} &
        \includegraphics[valign=c, width=\ww]{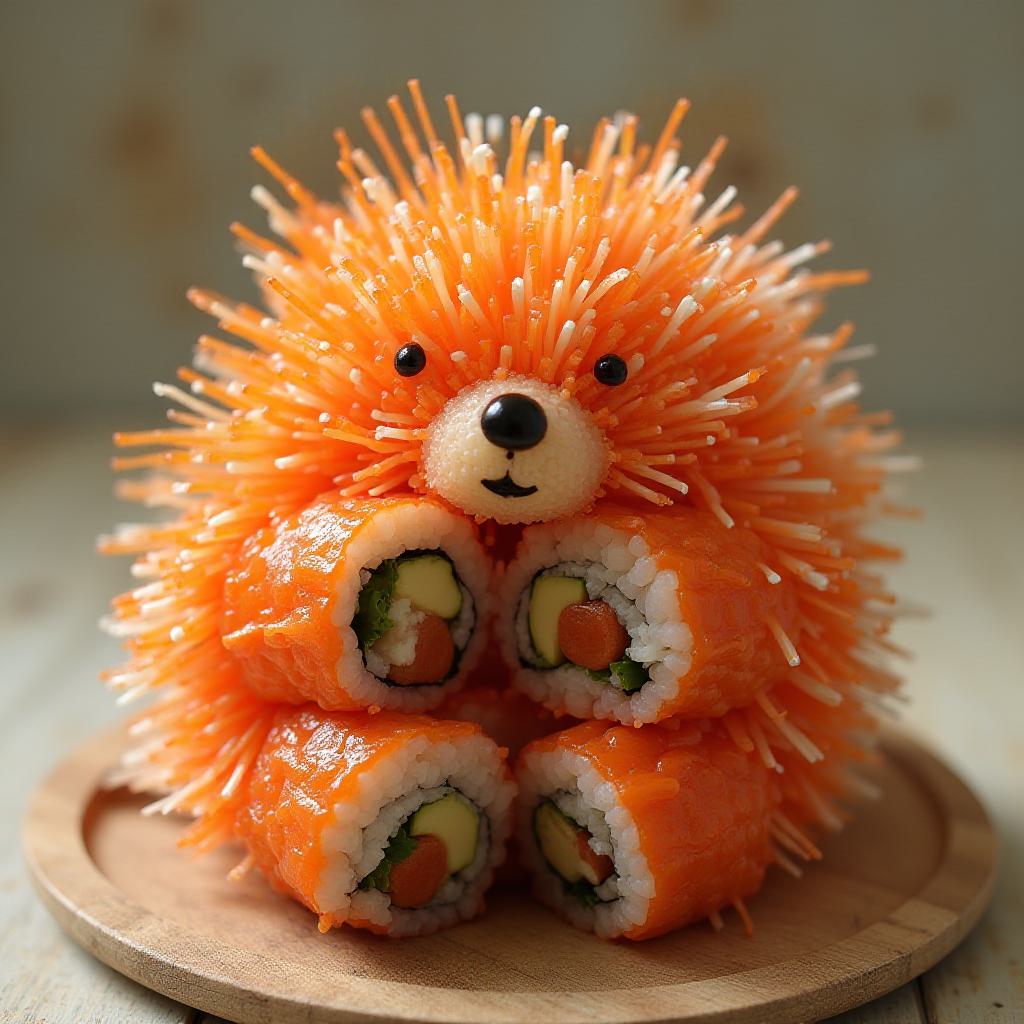} &
        \includegraphics[valign=c, width=\ww]{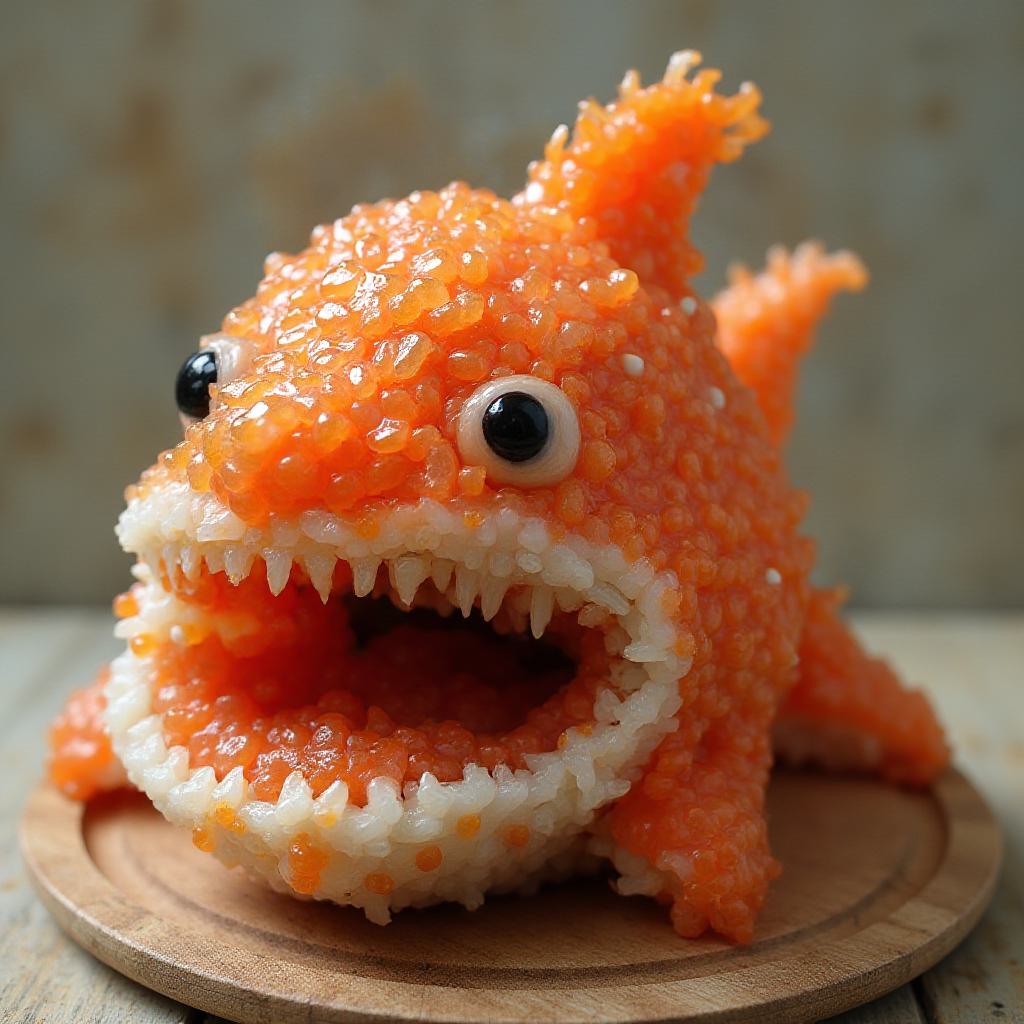} &
        \includegraphics[valign=c, width=\ww]{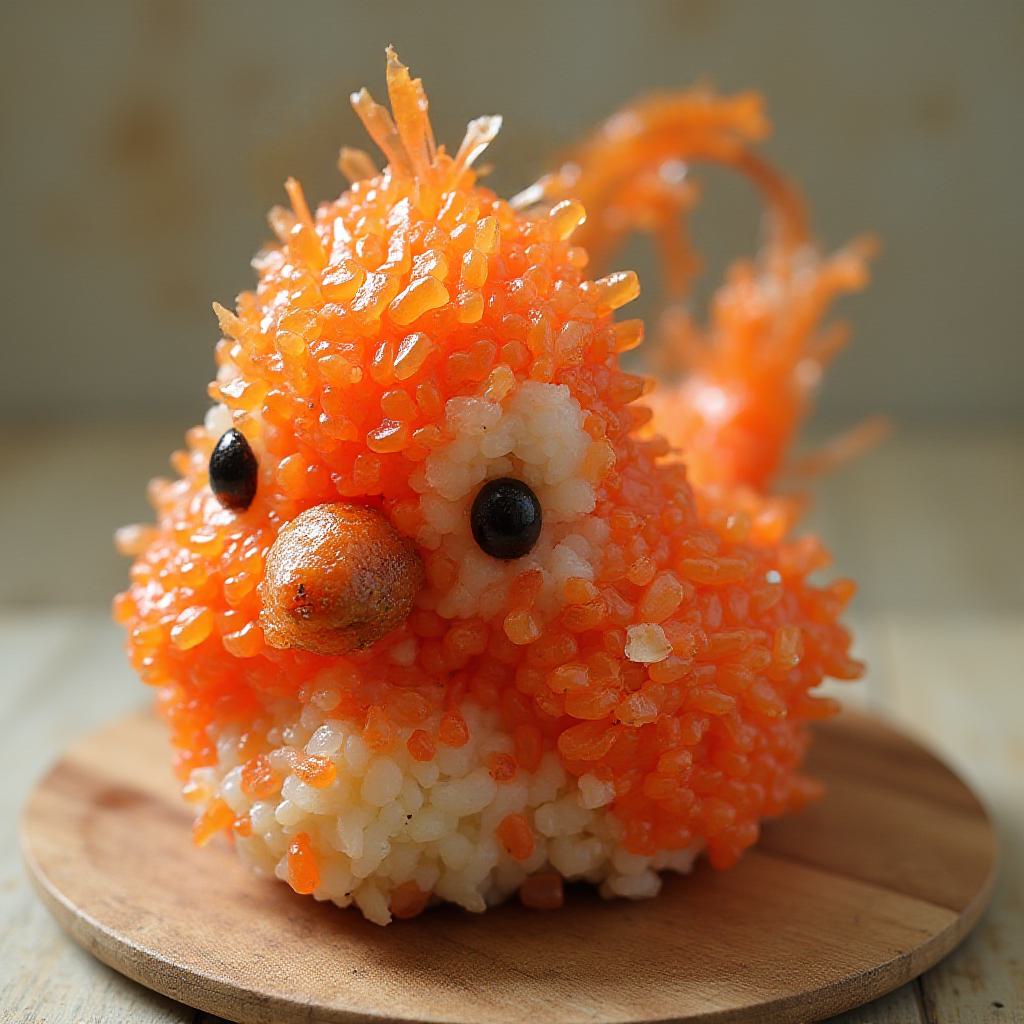}
        \vspace{3px}
        \\

        \small{Input} &
        \small{\prompt{A hedgehog}} &
        \small{\prompt{A shark}} &
        \small{\prompt{A bird}}
        \vspace{10px}
        \\

        \includegraphics[valign=c, width=\ww]{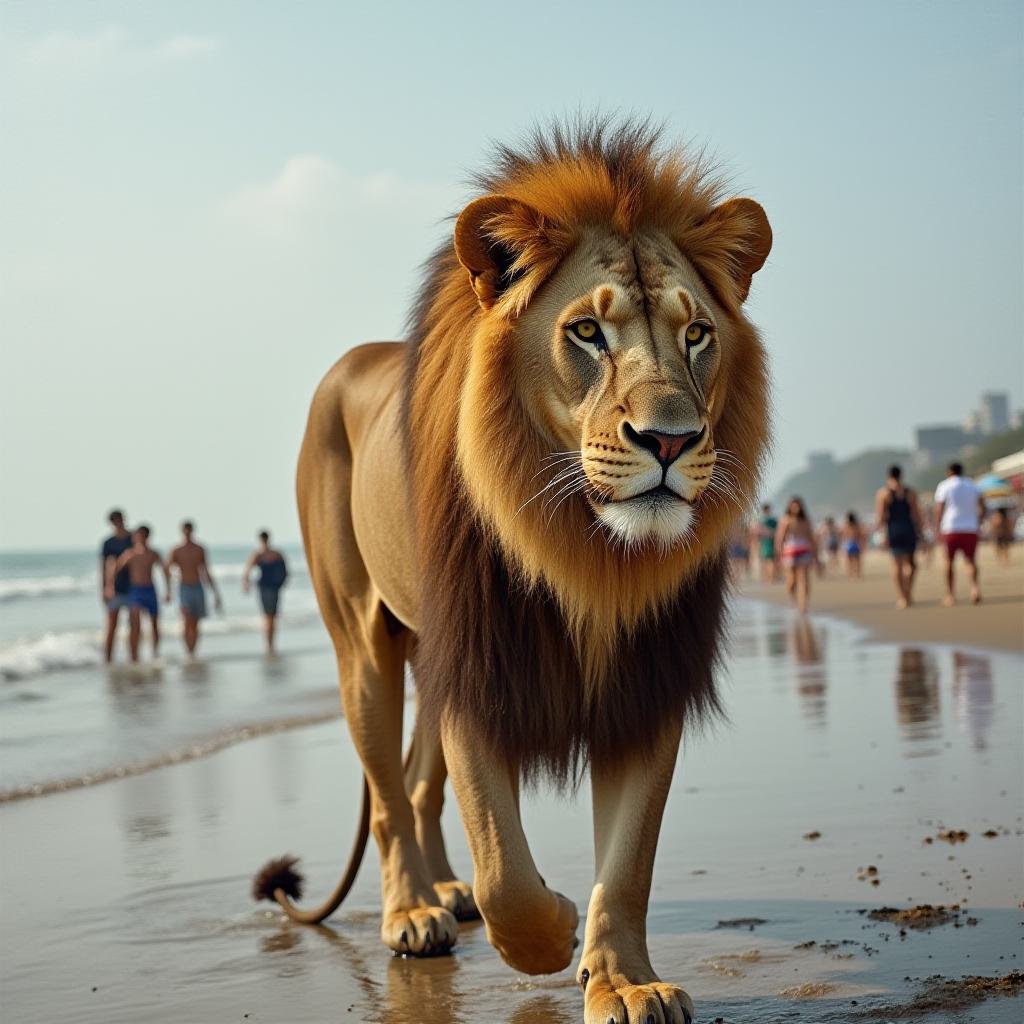} &
        \includegraphics[valign=c, width=\ww]{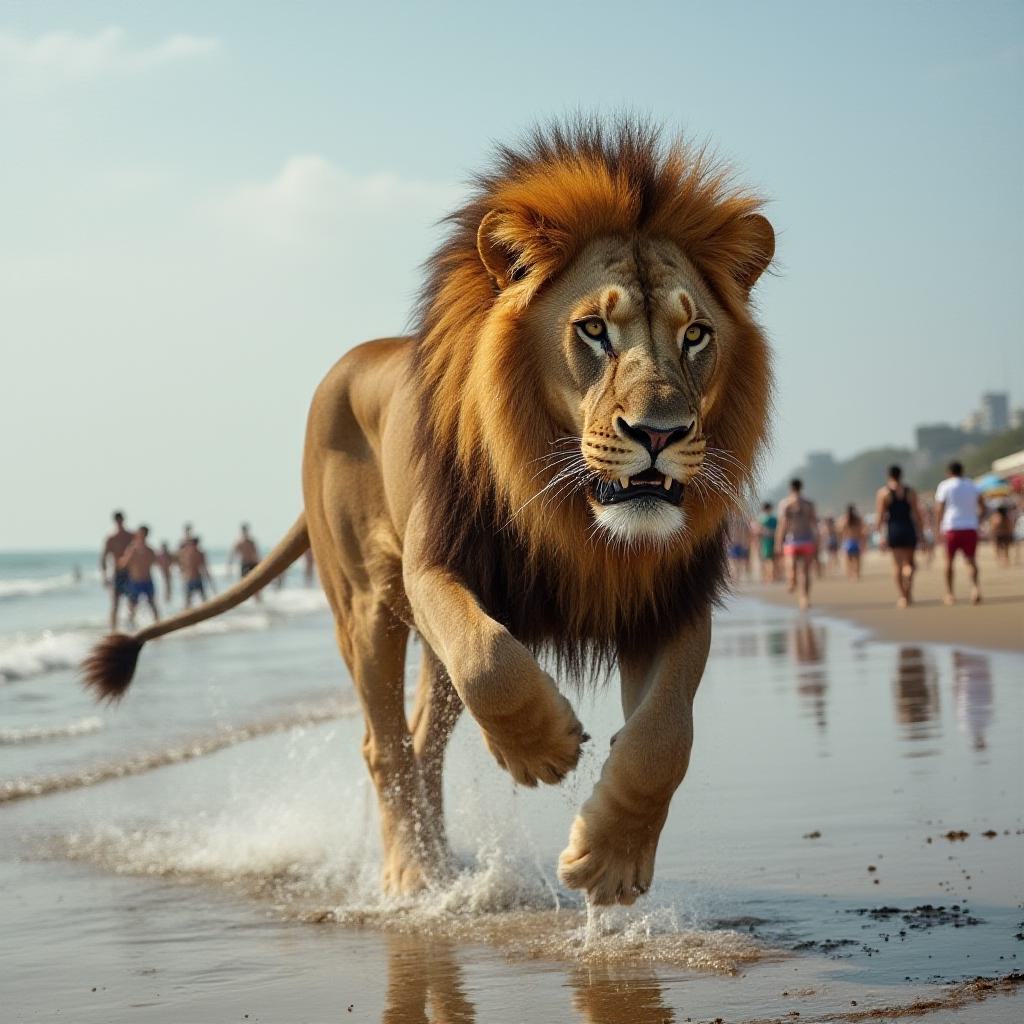} &
        \includegraphics[valign=c, width=\ww]{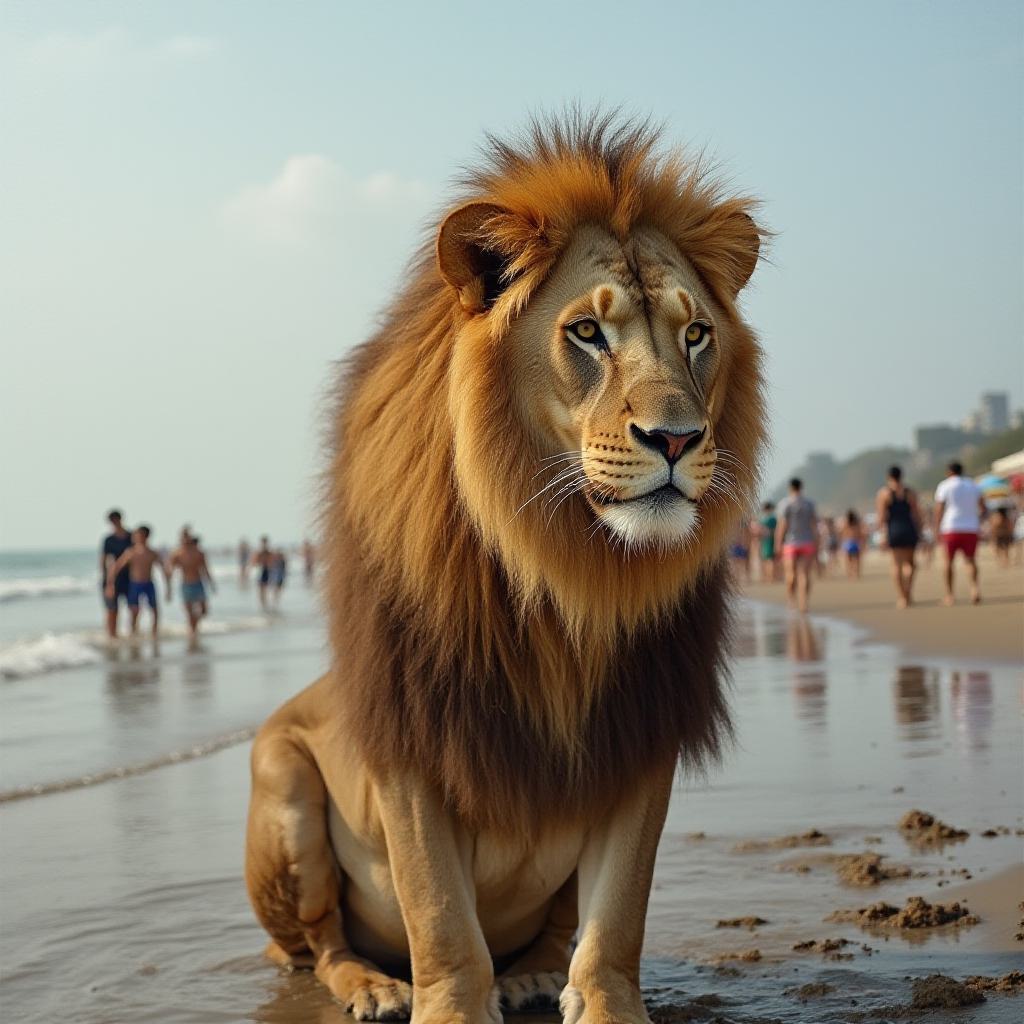} &
        \includegraphics[valign=c, width=\ww]{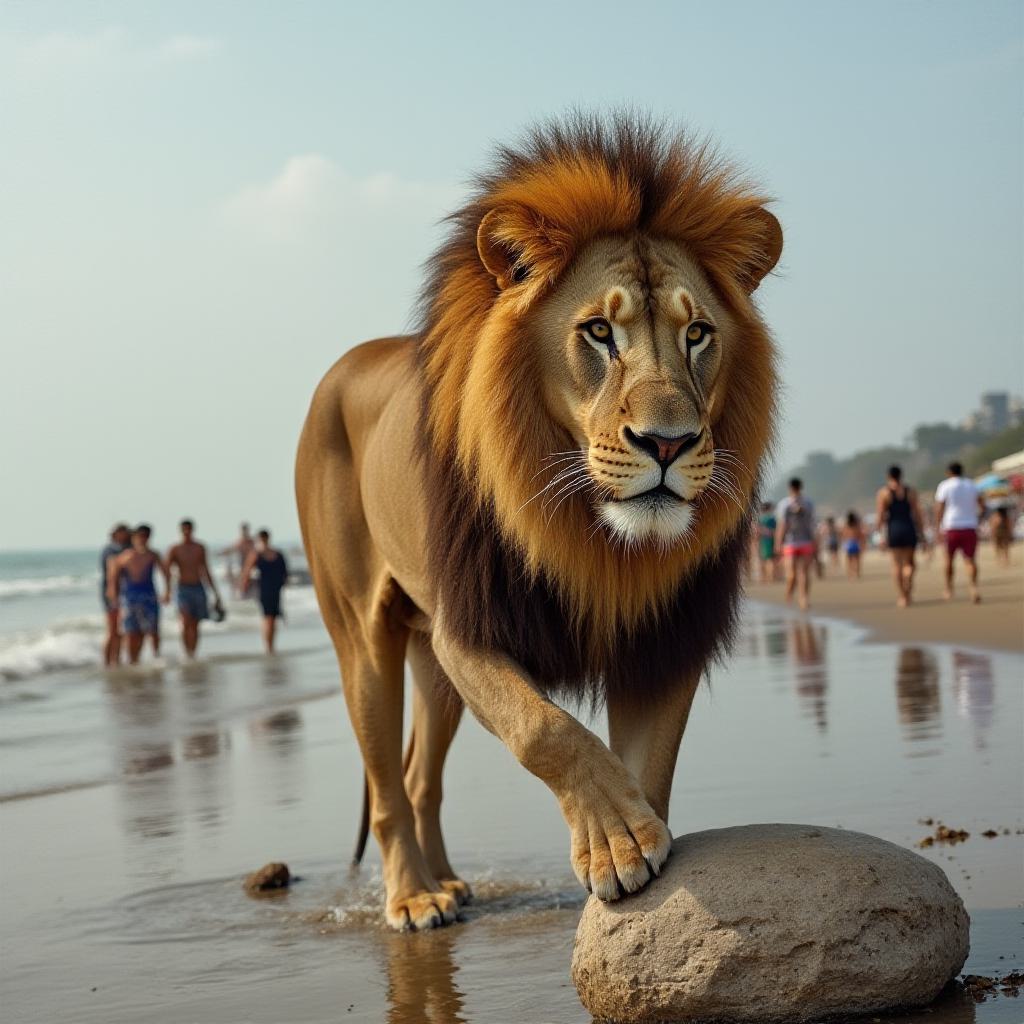}
        \vspace{3px}
        \\

        \small{Input} &
        \small{\prompt{Jumping}} &
        \small{\prompt{Sitting}} &
        \small{\prompt{Putting its paw on a stone}}
        \vspace{10px}
        \\

    \end{tabular}
    \caption{\textbf{Additional Results.} We provide various editing results of our method. These different edits are done using the \emph{same} vital layer set.}
    \label{fig:additional_results1}
    \vspace{10px}
\end{figure*}

\begin{figure*}[tp]
    \centering
    \setlength{\tabcolsep}{0.6pt}
    \renewcommand{\arraystretch}{0.8}
    \setlength{\ww}{0.24\linewidth}
    \begin{tabular}{c @{\hspace{10\tabcolsep}} ccc}
        \includegraphics[valign=c, width=\ww]{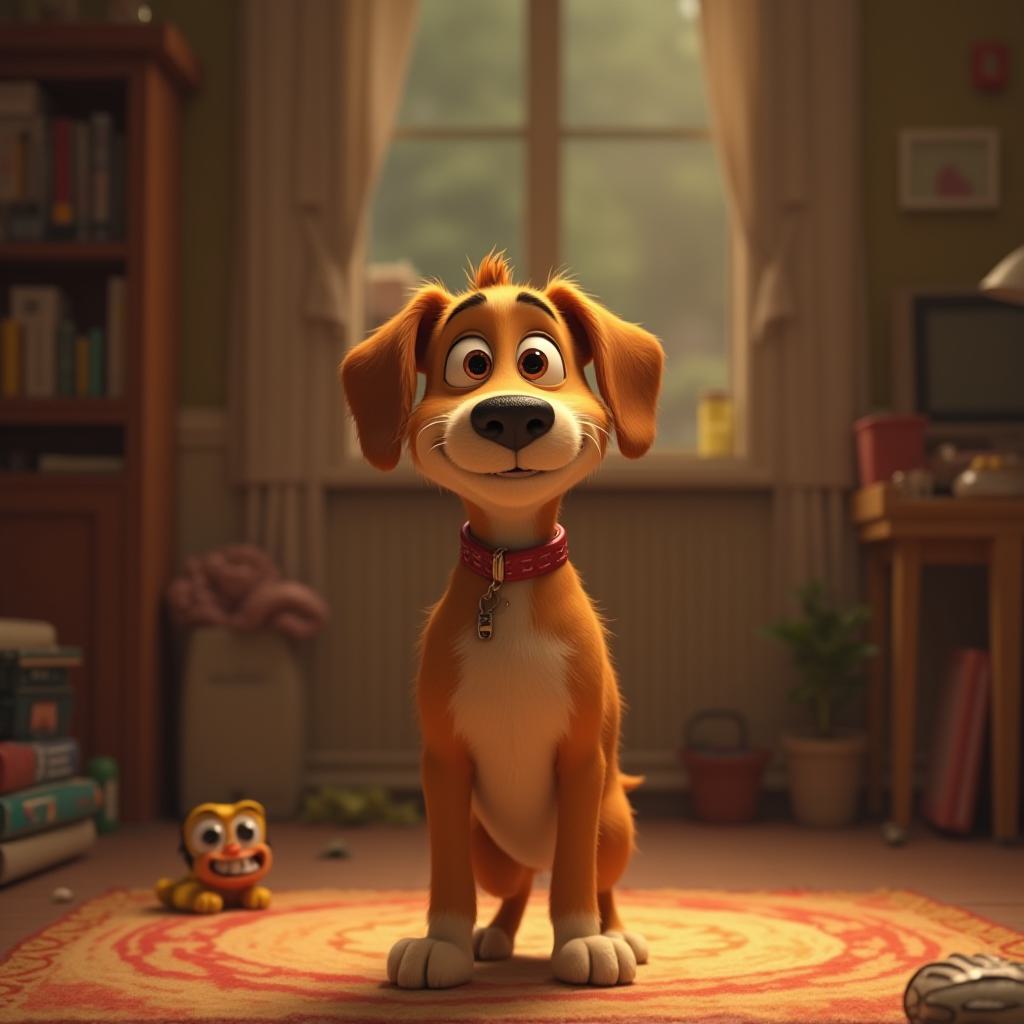} &
        \includegraphics[valign=c, width=\ww]{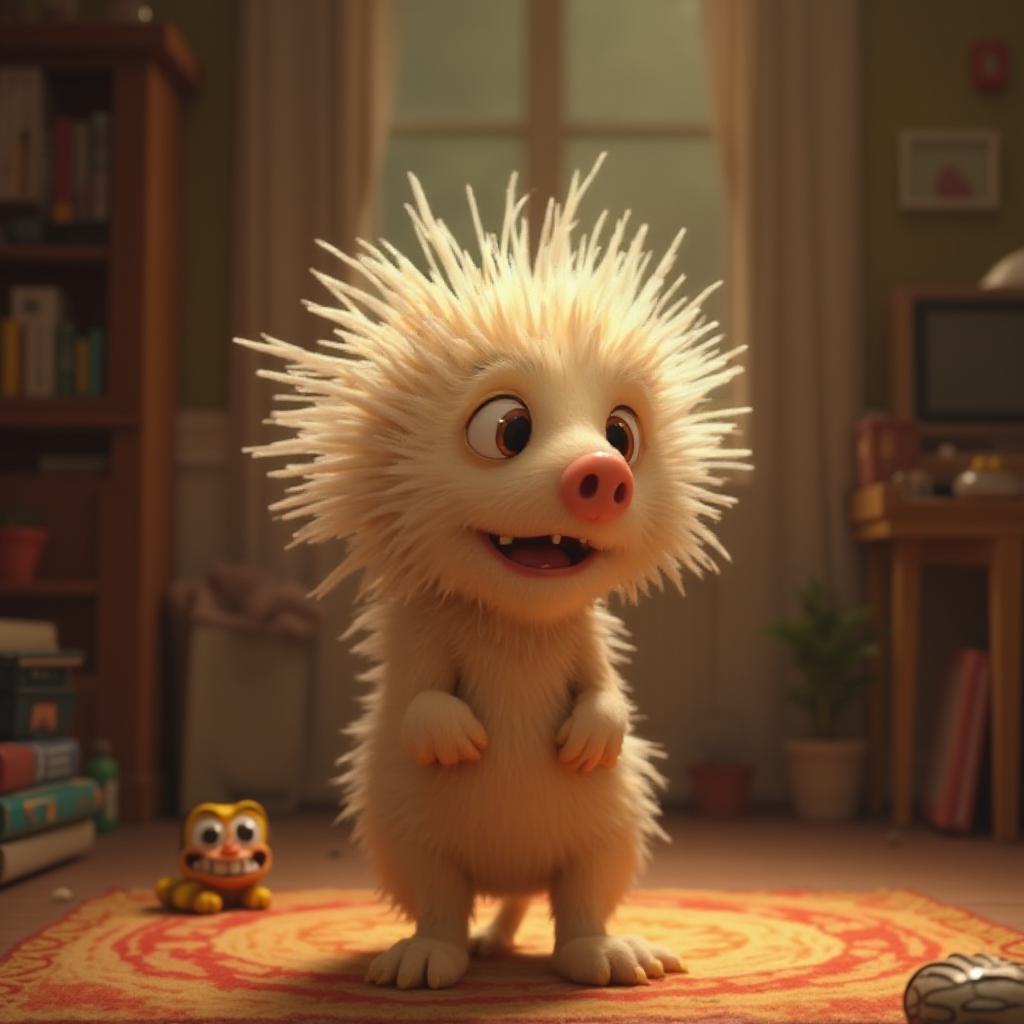} &
        \includegraphics[valign=c, width=\ww]{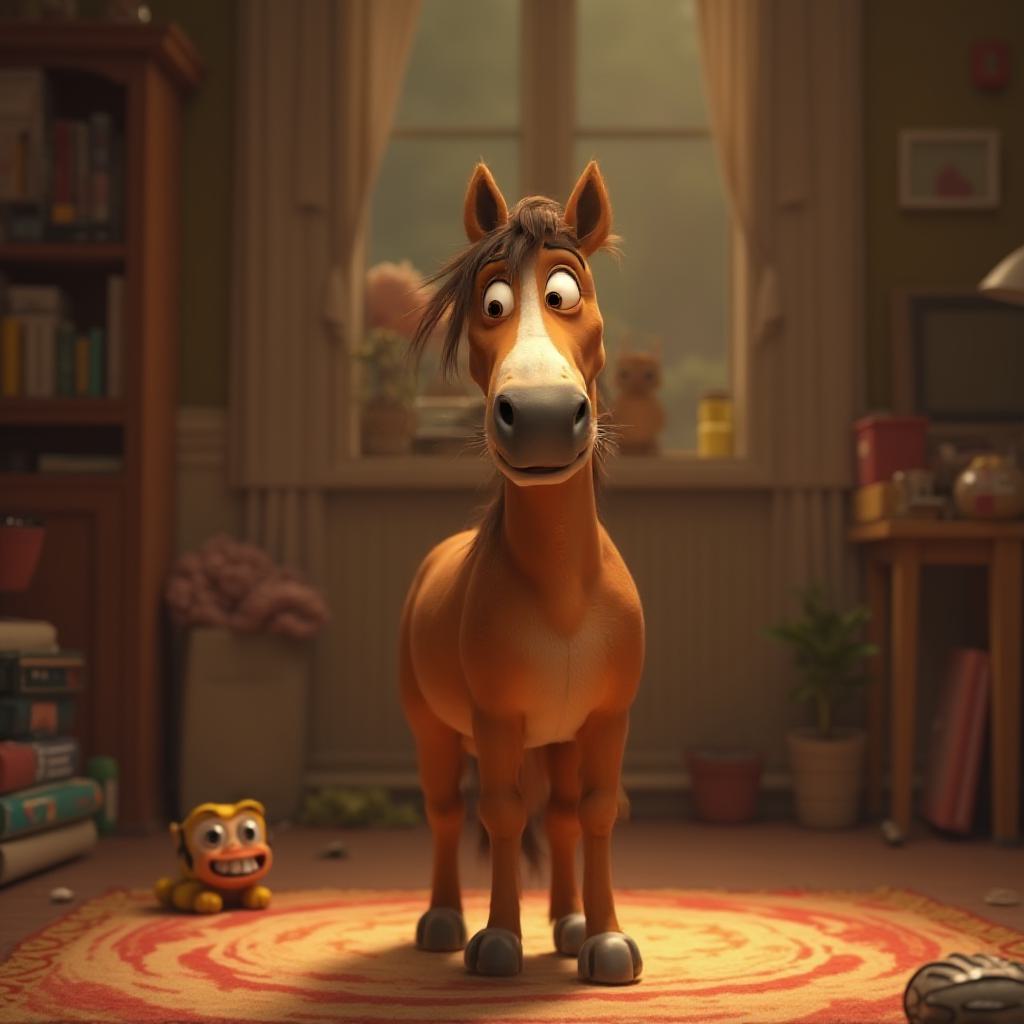} &
        \includegraphics[valign=c, width=\ww]{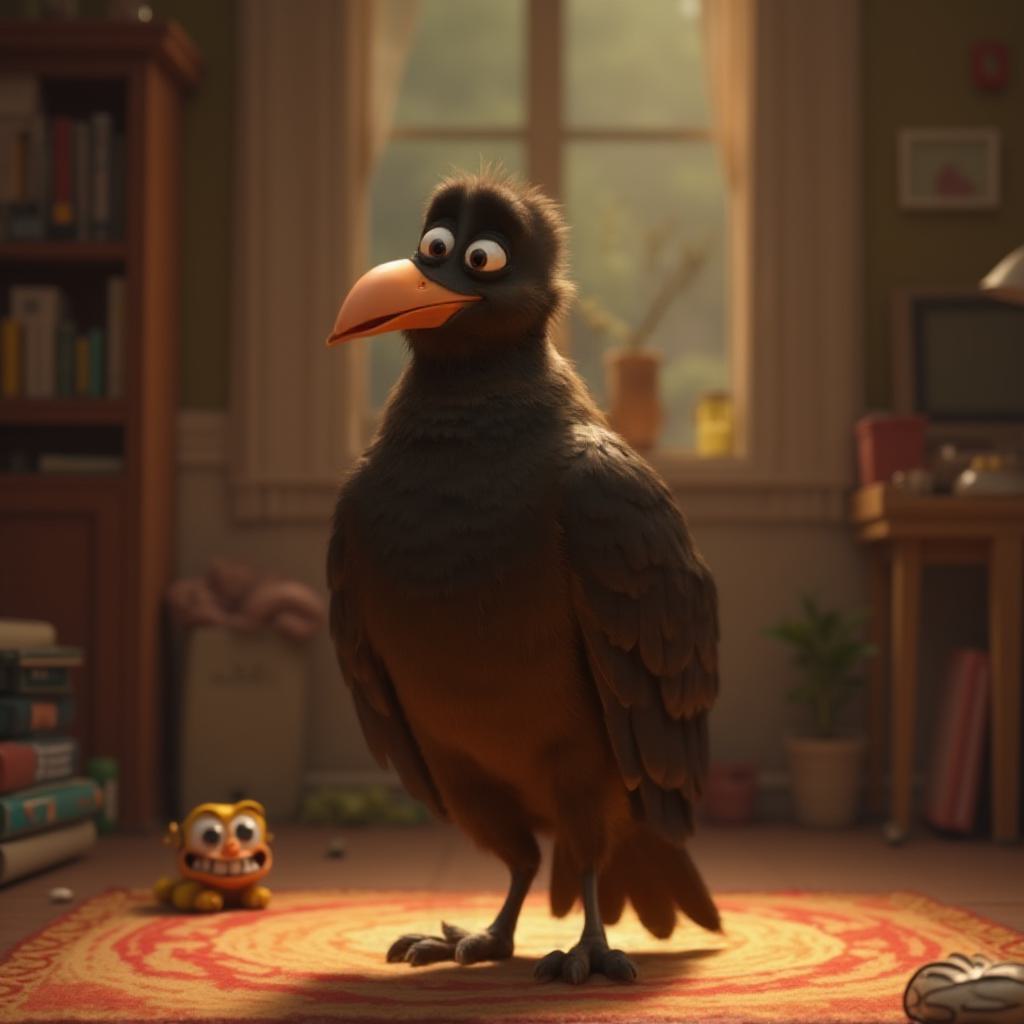}
        \vspace{3px}
        \\

        \small{Input} &
        \small{\prompt{An albino porcupine}} &
        \small{\prompt{A horse}} &
        \small{\prompt{A crow}}
        \vspace{10px}
        \\

        \includegraphics[valign=c, width=\ww]{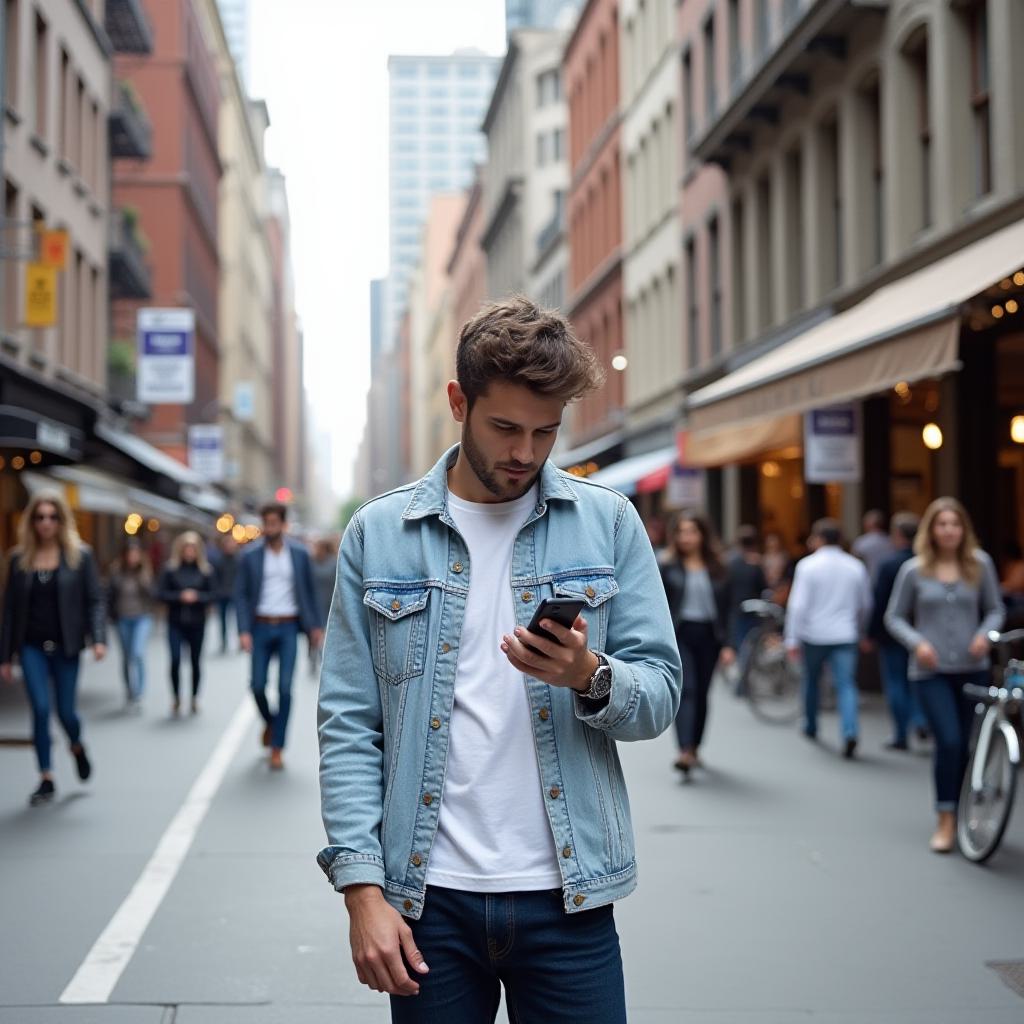} &
        \includegraphics[valign=c, width=\ww]{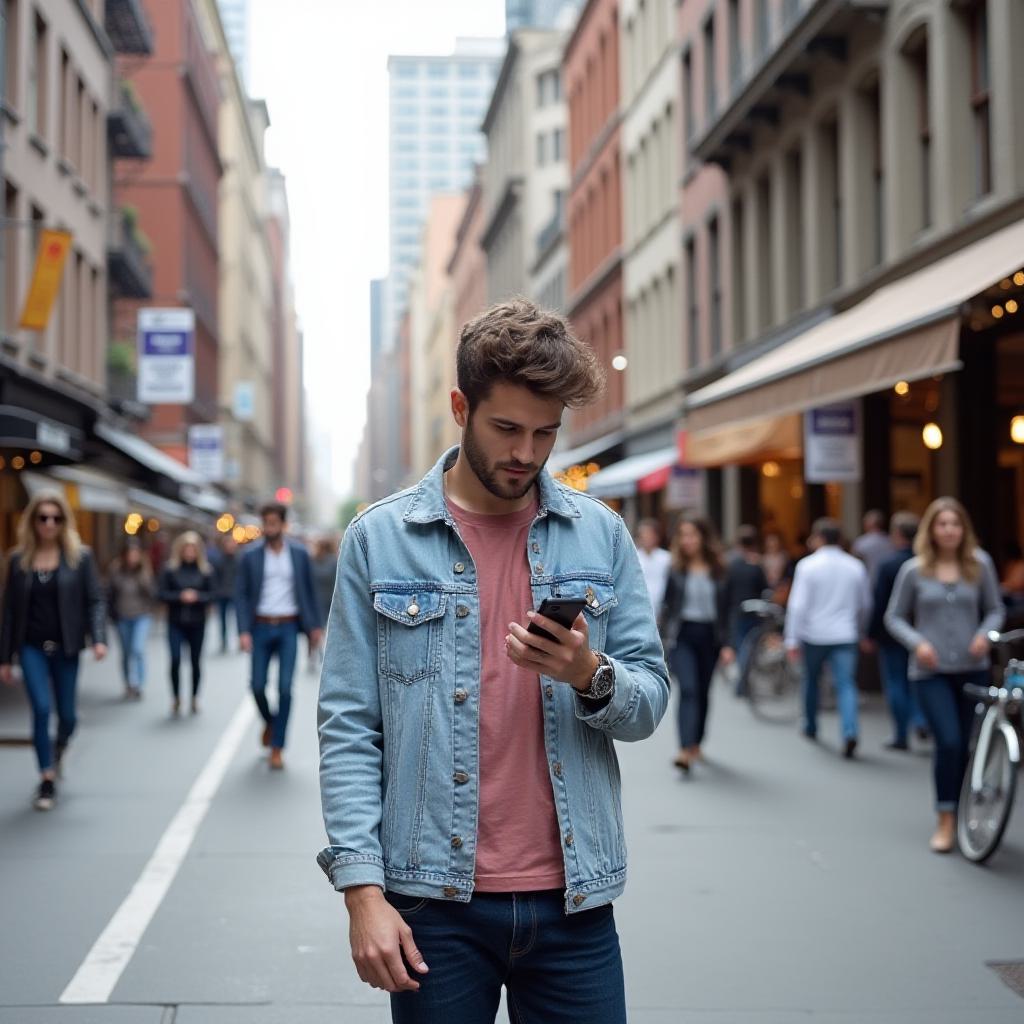} &
        \includegraphics[valign=c, width=\ww]{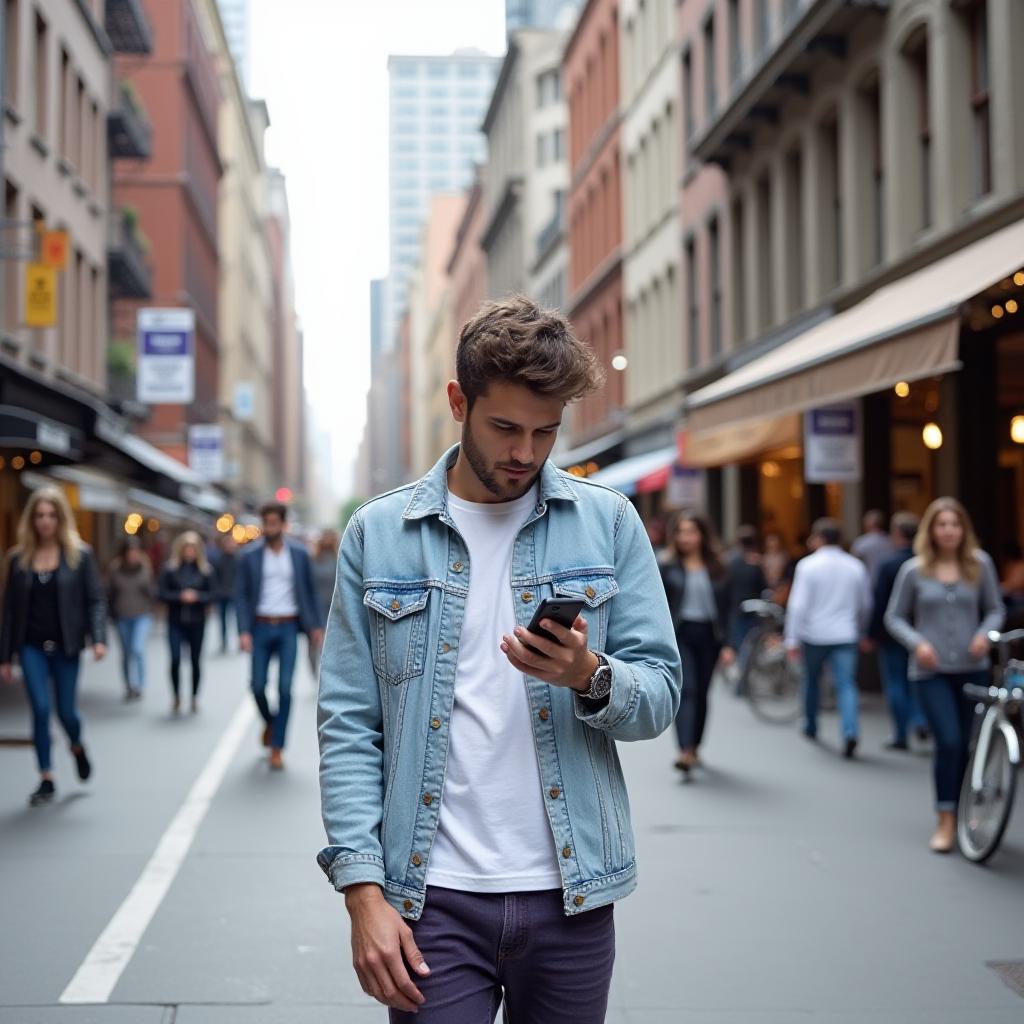} &
        \includegraphics[valign=c, width=\ww]{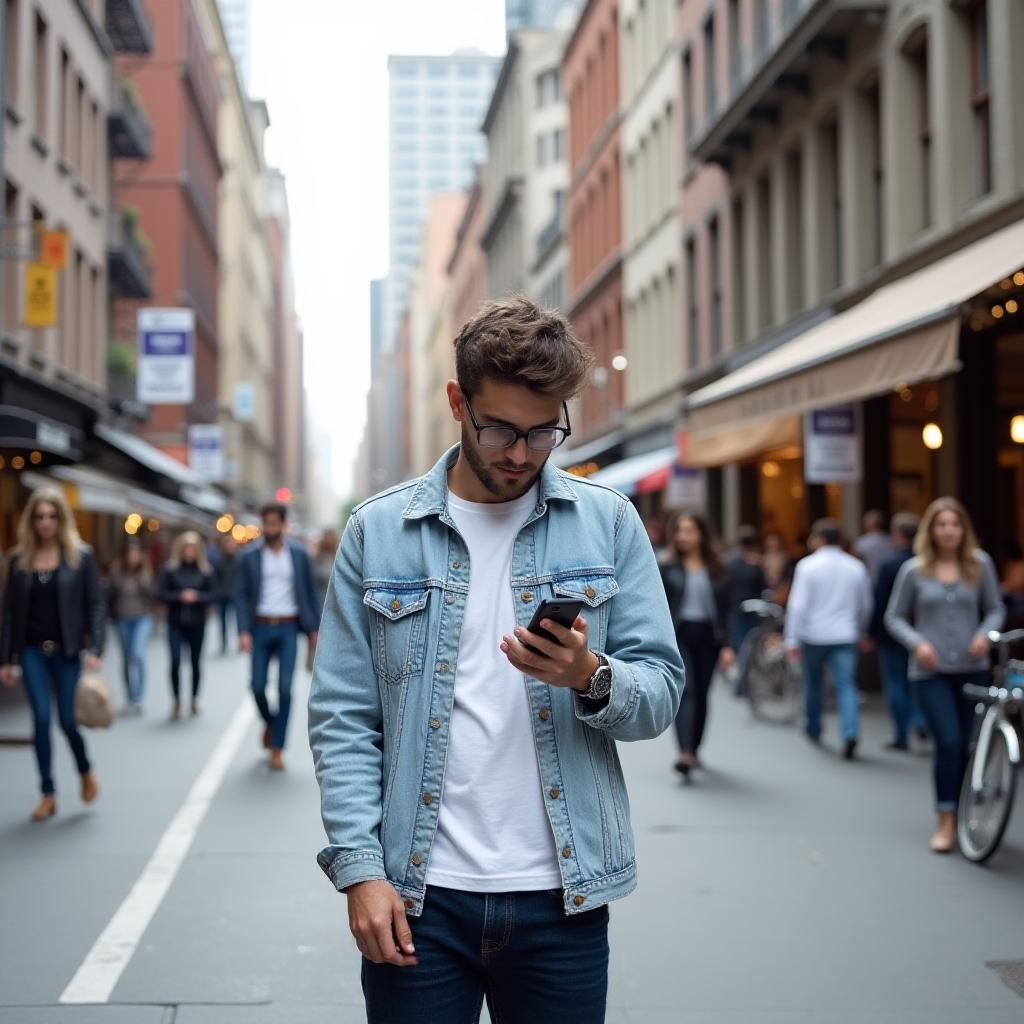}
        \vspace{3px}
        \\

        \small{Input} &
        \small{\prompt{Wearing a red shirt}} &
        \small{\prompt{Wearing purple jeans}} &
        \small{\prompt{Wearing glasses}}
        \vspace{10px}
        \\

        \includegraphics[valign=c, width=\ww]{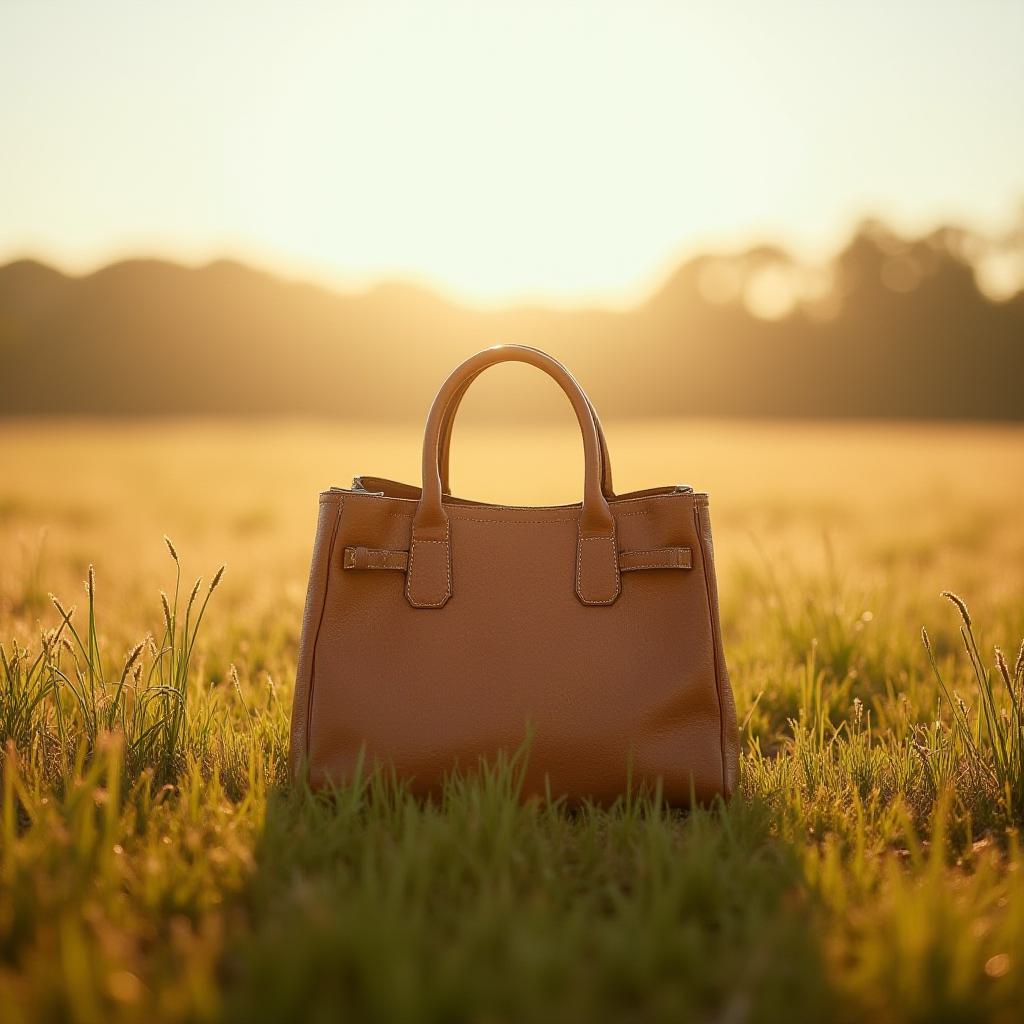} &
        \includegraphics[valign=c, width=\ww]{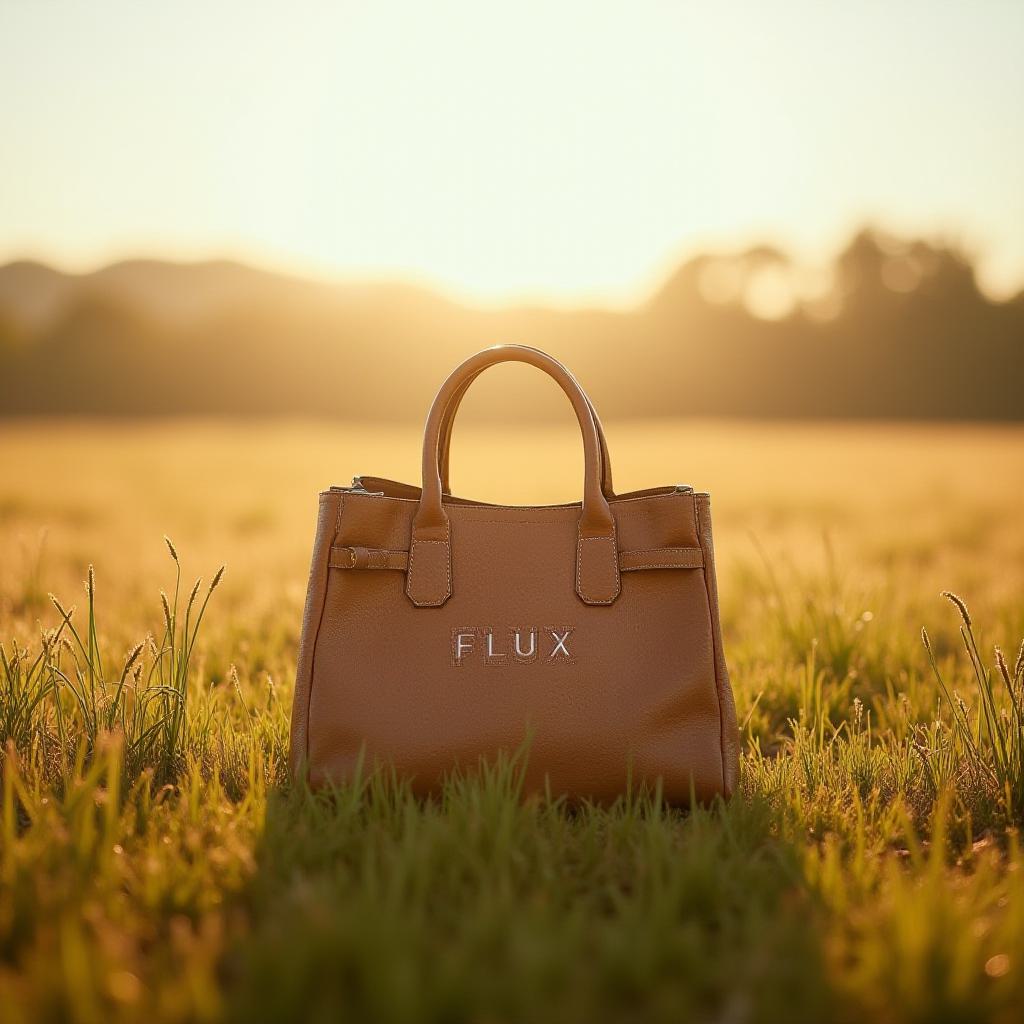} &
        \includegraphics[valign=c, width=\ww]{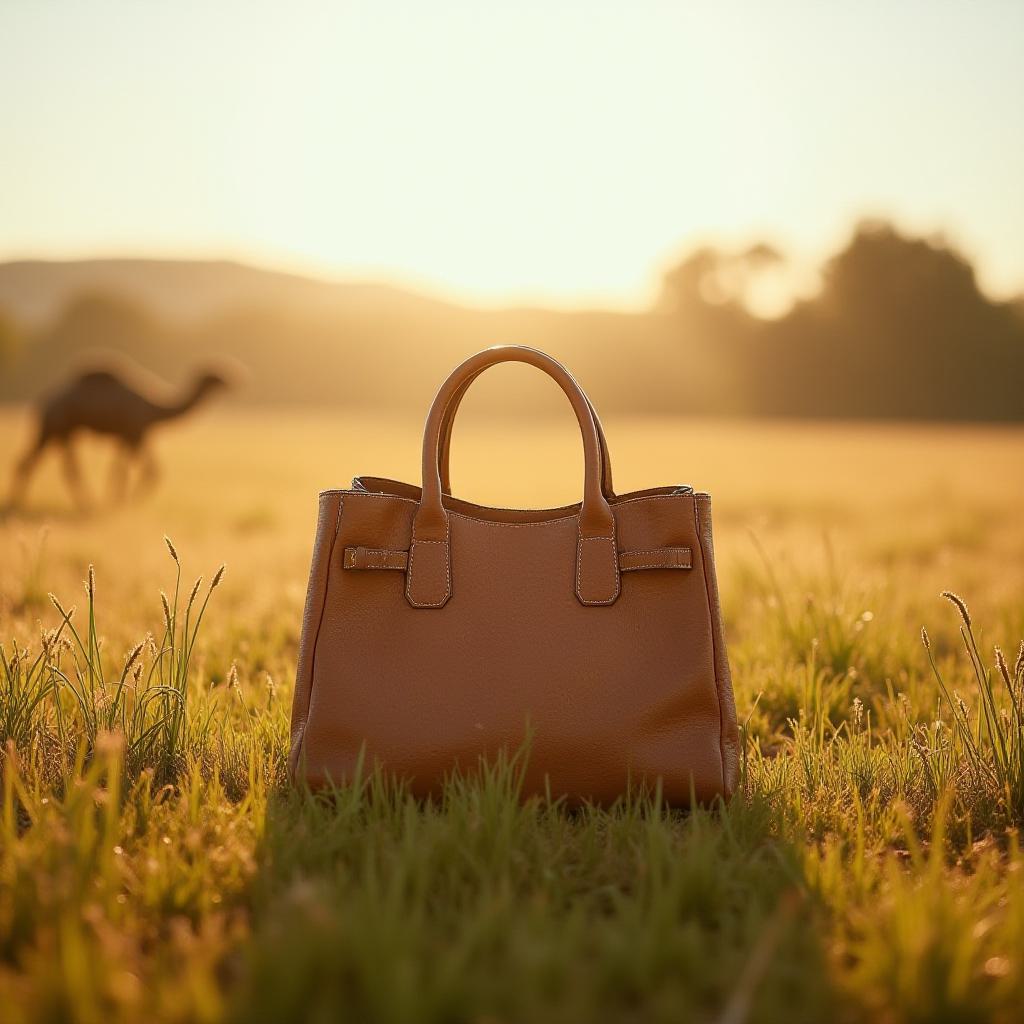} &
        \includegraphics[valign=c, width=\ww]{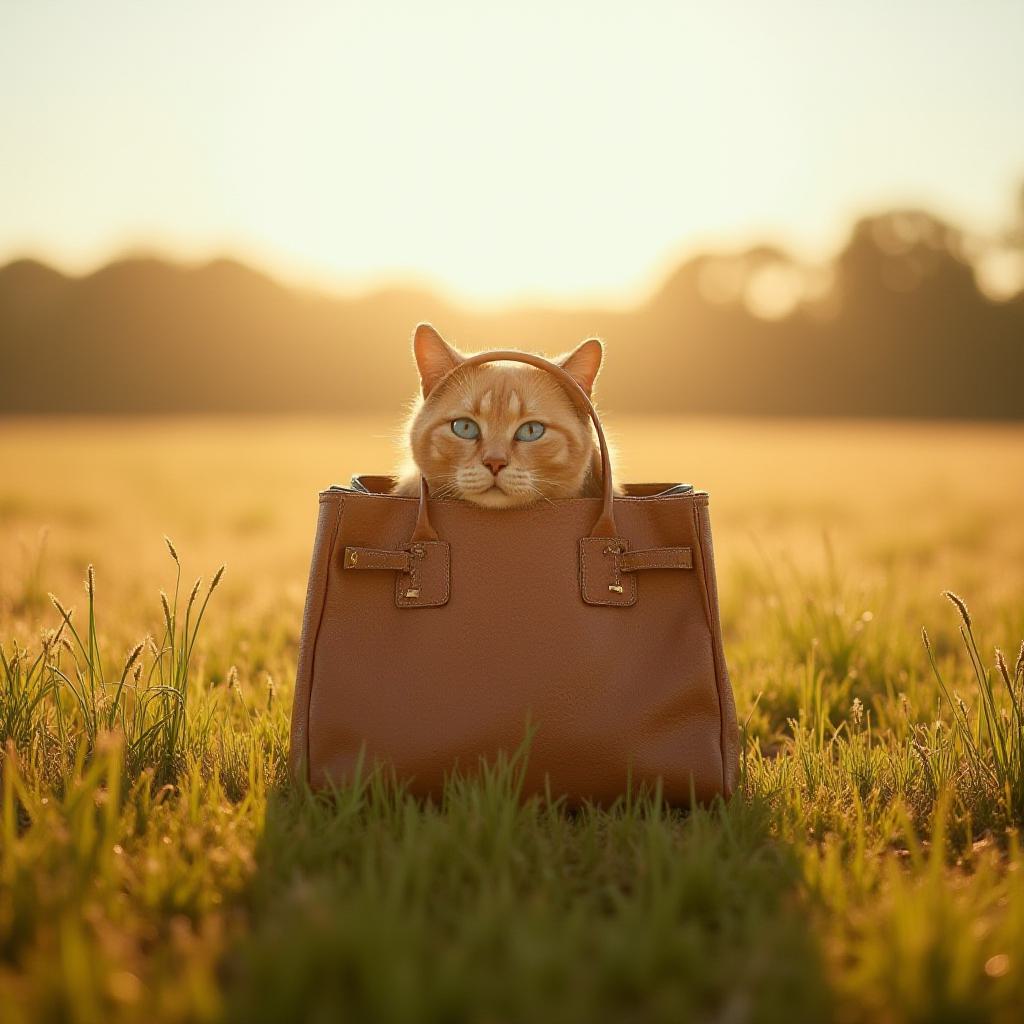}
        \vspace{3px}
        \\

        \small{Input} &
        \small{\promptstart{The text `FLUX' is written}} &
        \small{\prompt{A camel in the background}} &
        \small{\prompt{A cat inside the bag}}
        \\

        &
        \small{\promptend{on the bag}} &
        &
        \vspace{10px}
        \\

        \includegraphics[valign=c, width=\ww]{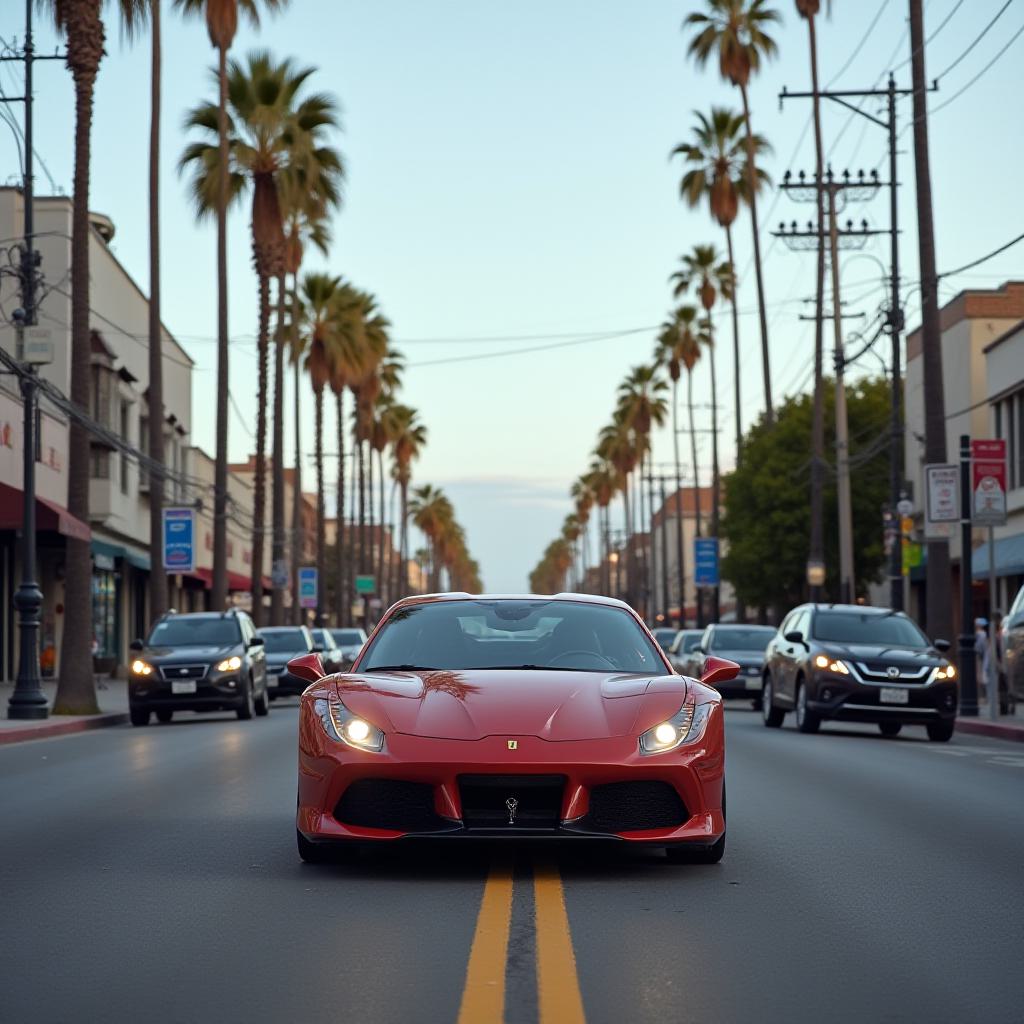} &
        \includegraphics[valign=c, width=\ww]{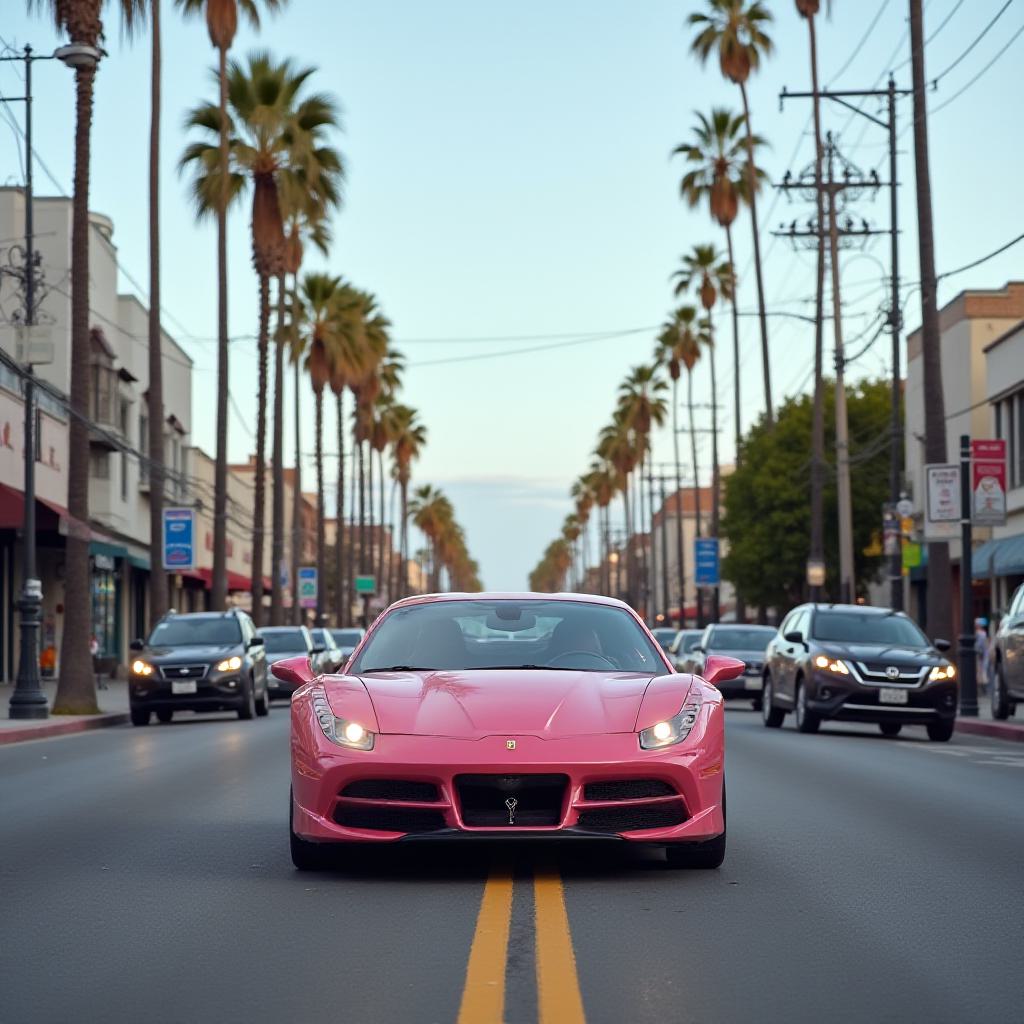} &
        \includegraphics[valign=c, width=\ww]{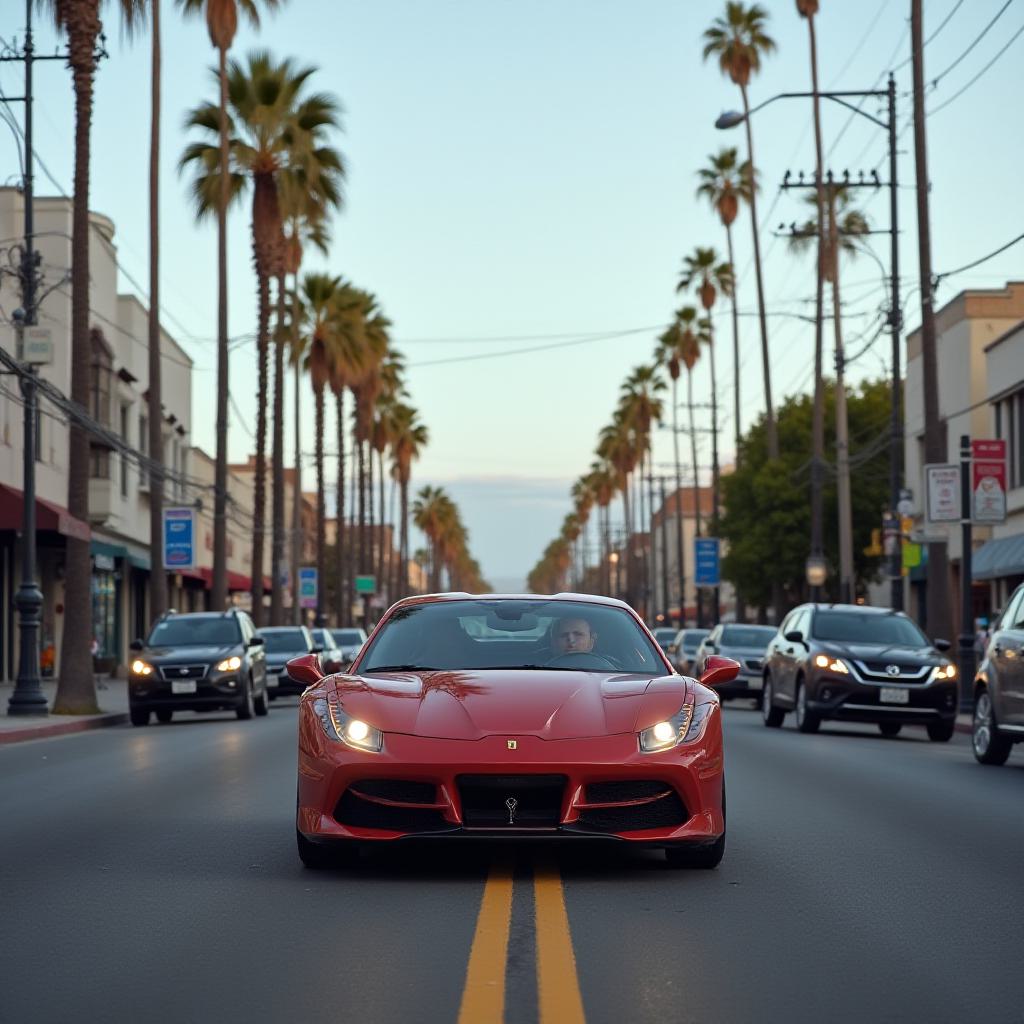} &
        \includegraphics[valign=c, width=\ww]{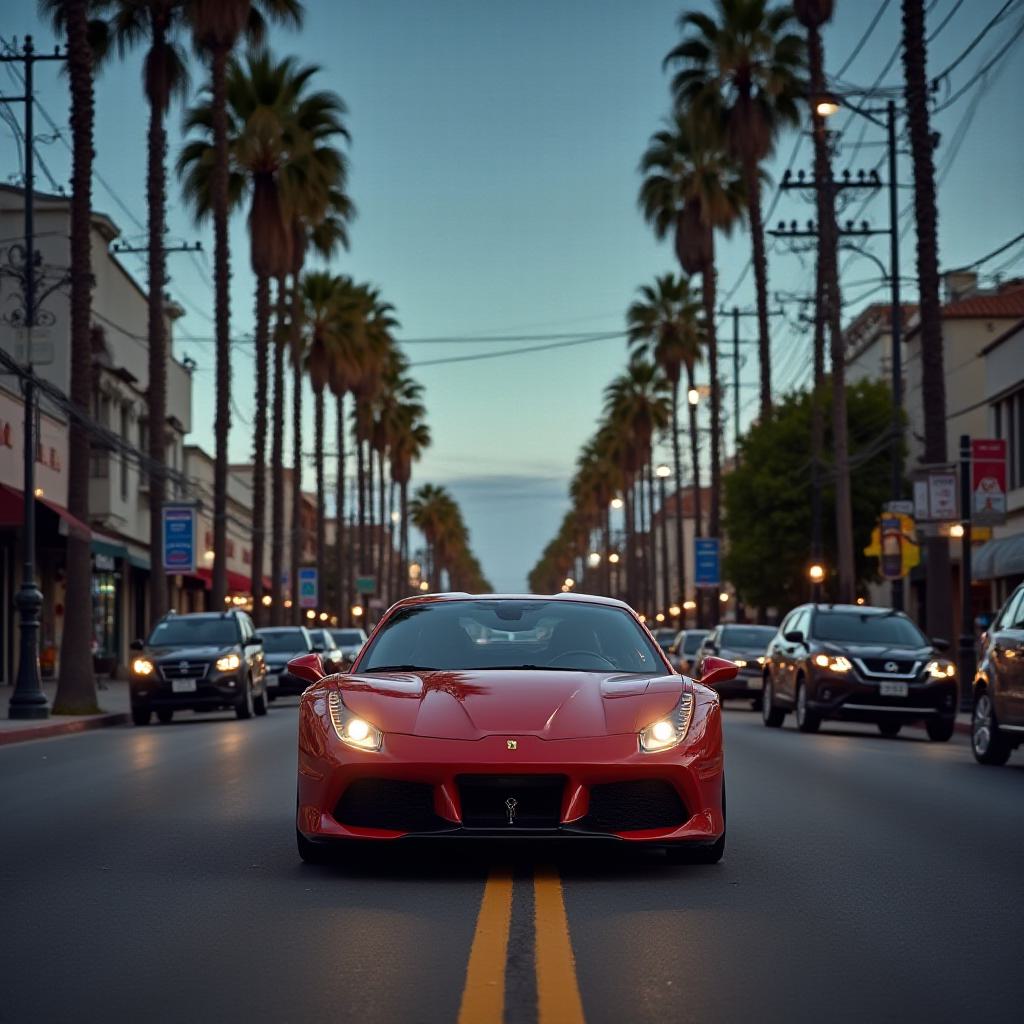}
        \vspace{3px}
        \\

        \small{Input} &
        \small{\prompt{A pink car}} &
        \small{\prompt{A man driving the car}} &
        \small{\prompt{In the evening}}
        \vspace{10px}
        \\

    \end{tabular}
    \caption{\textbf{Additional Results.} We provide various editing results of our method. These different edits are done using the \emph{same} vital layer set.}
    \label{fig:additional_results2}
\end{figure*}

\begin{figure*}[tp]
    \centering
    \setlength{\tabcolsep}{0.6pt}
    \renewcommand{\arraystretch}{0.8}
    \setlength{\ww}{0.238\linewidth}
    \begin{tabular}{c @{\hspace{10\tabcolsep}} ccc}
        \includegraphics[valign=c, width=\ww]{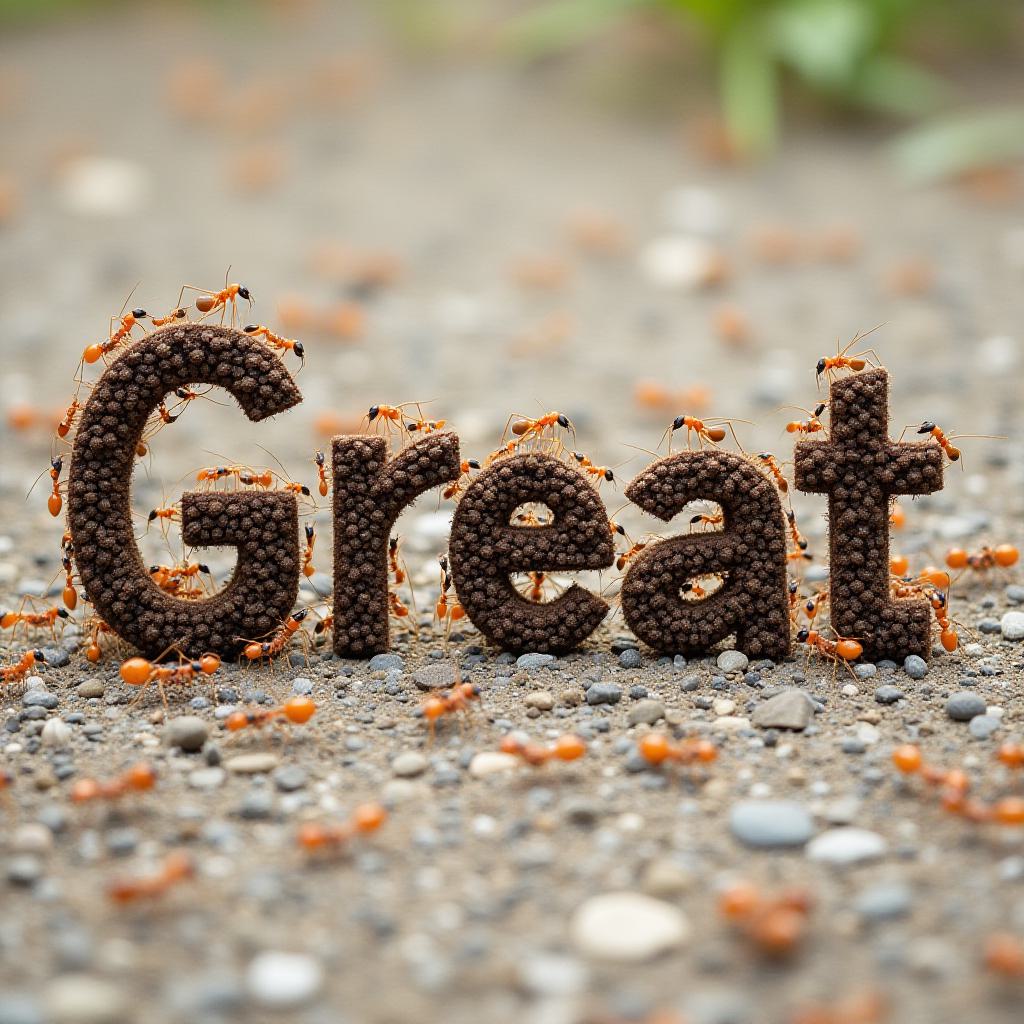} &
        \includegraphics[valign=c, width=\ww]{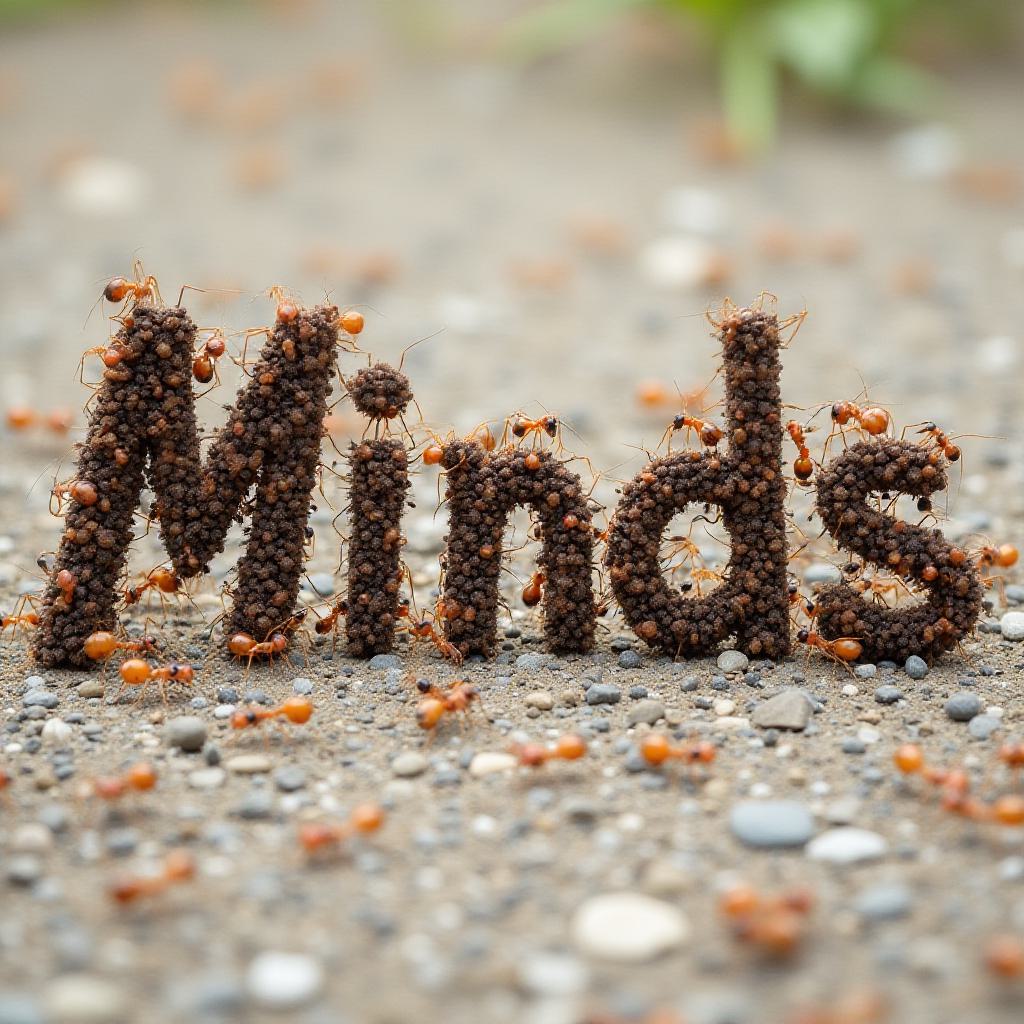} &
        \includegraphics[valign=c, width=\ww]{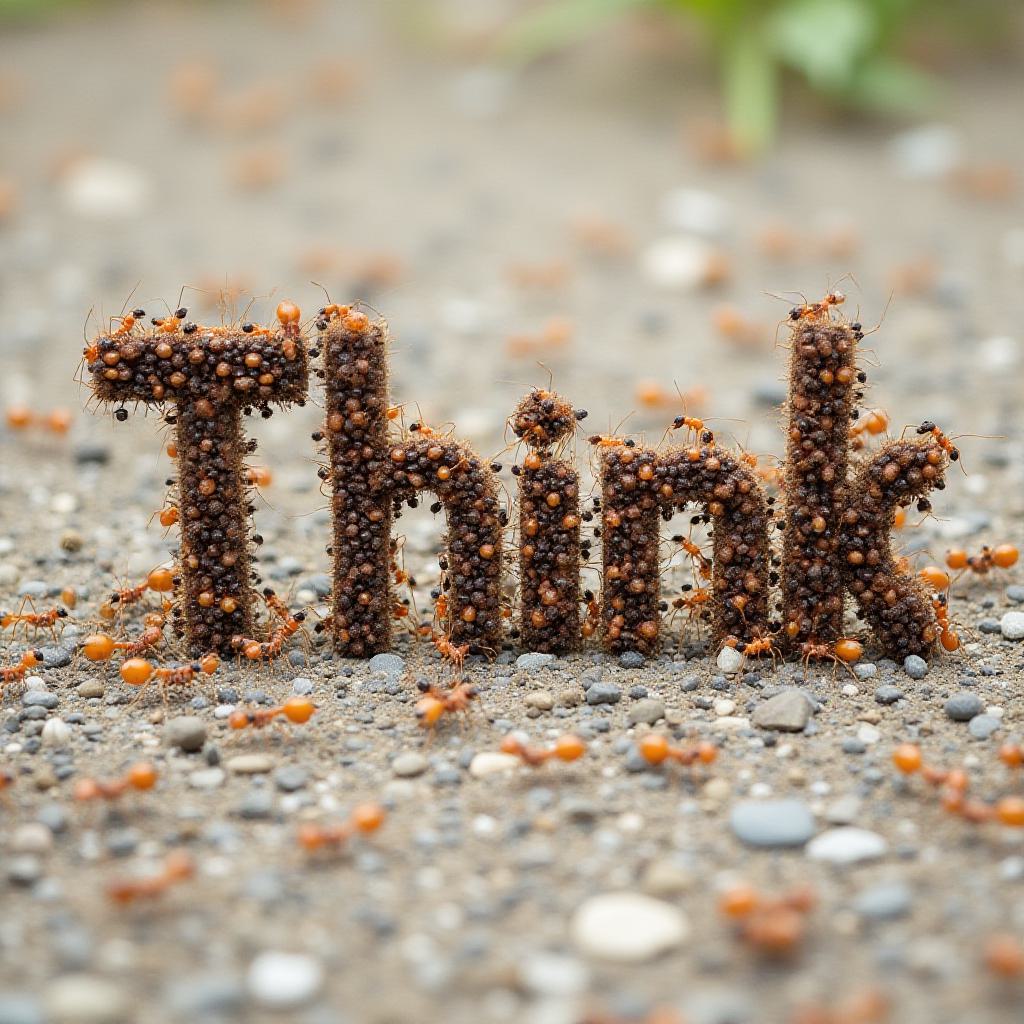} &
        \includegraphics[valign=c, width=\ww]{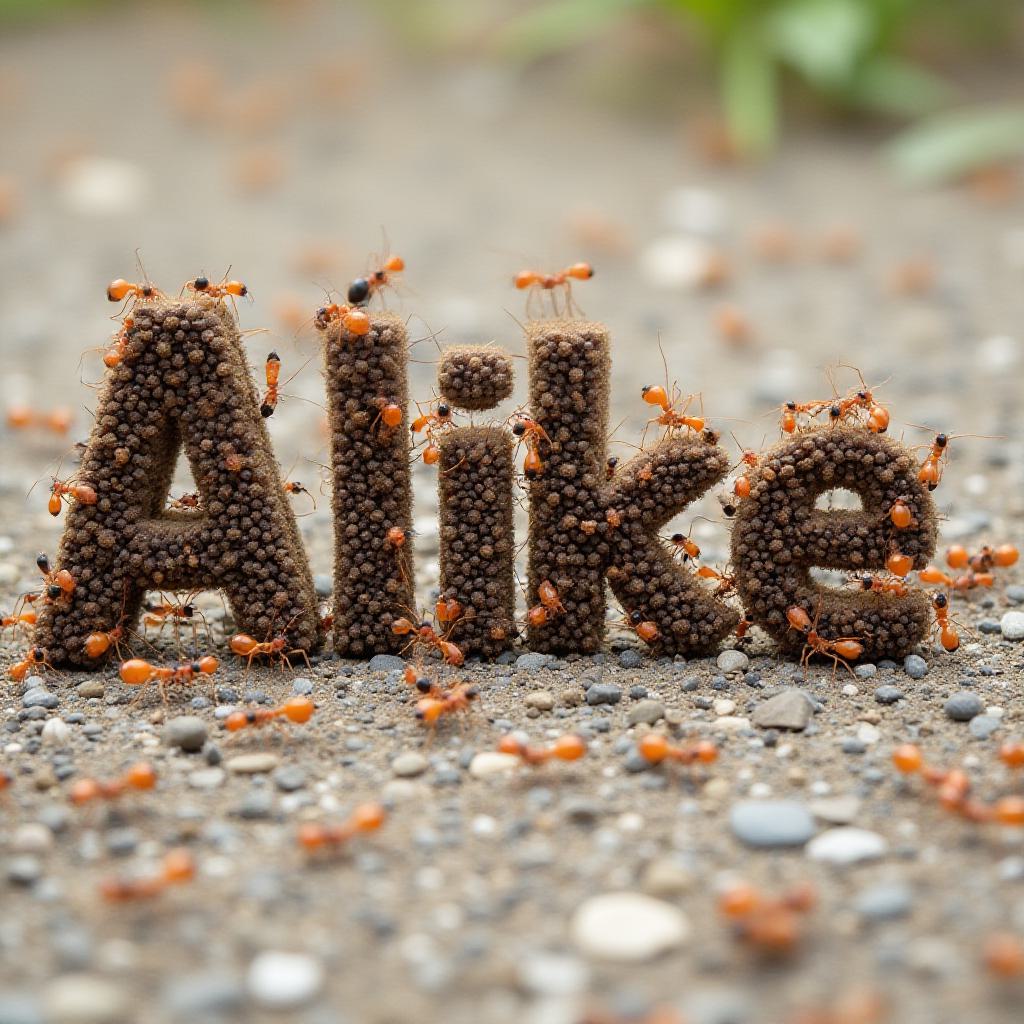}
        \vspace{5px}
        \\

        \includegraphics[valign=c, width=\ww]{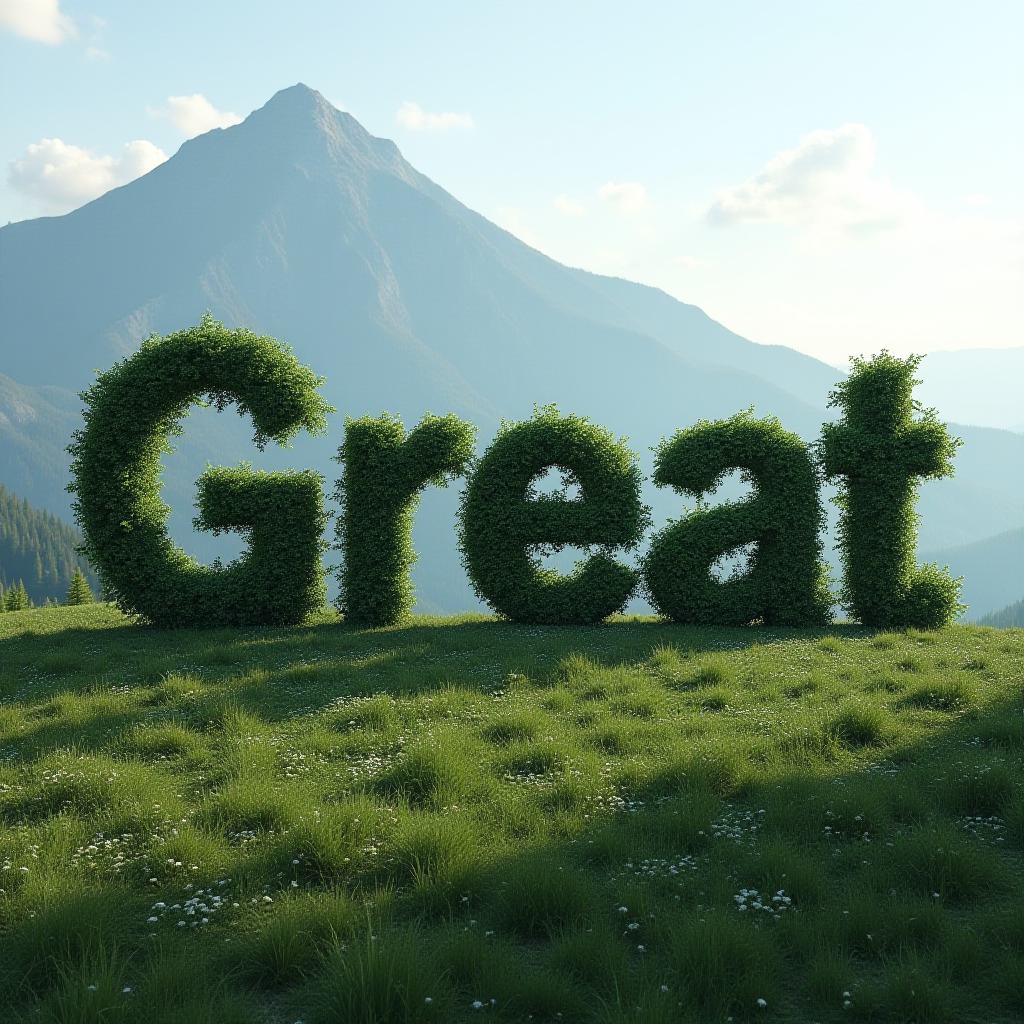} &
        \includegraphics[valign=c, width=\ww]{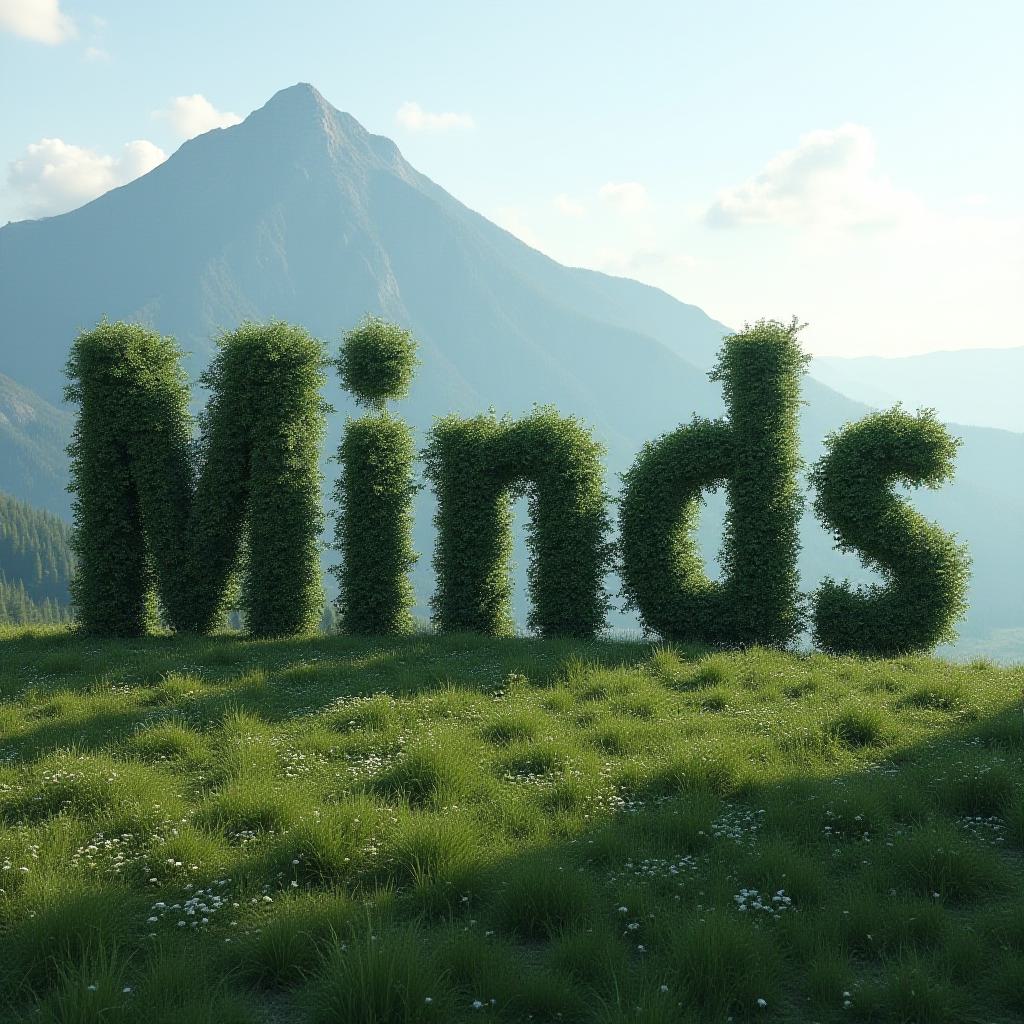} &
        \includegraphics[valign=c, width=\ww]{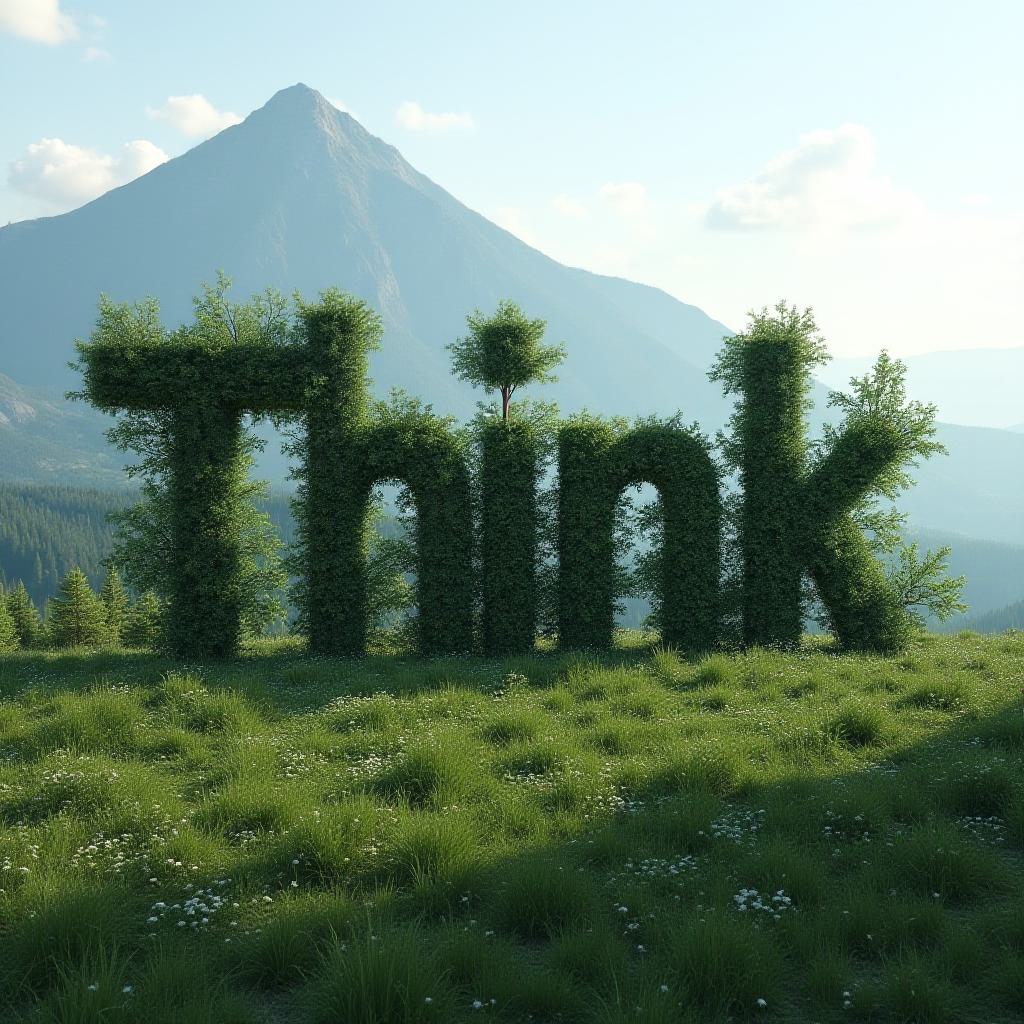} &
        \includegraphics[valign=c, width=\ww]{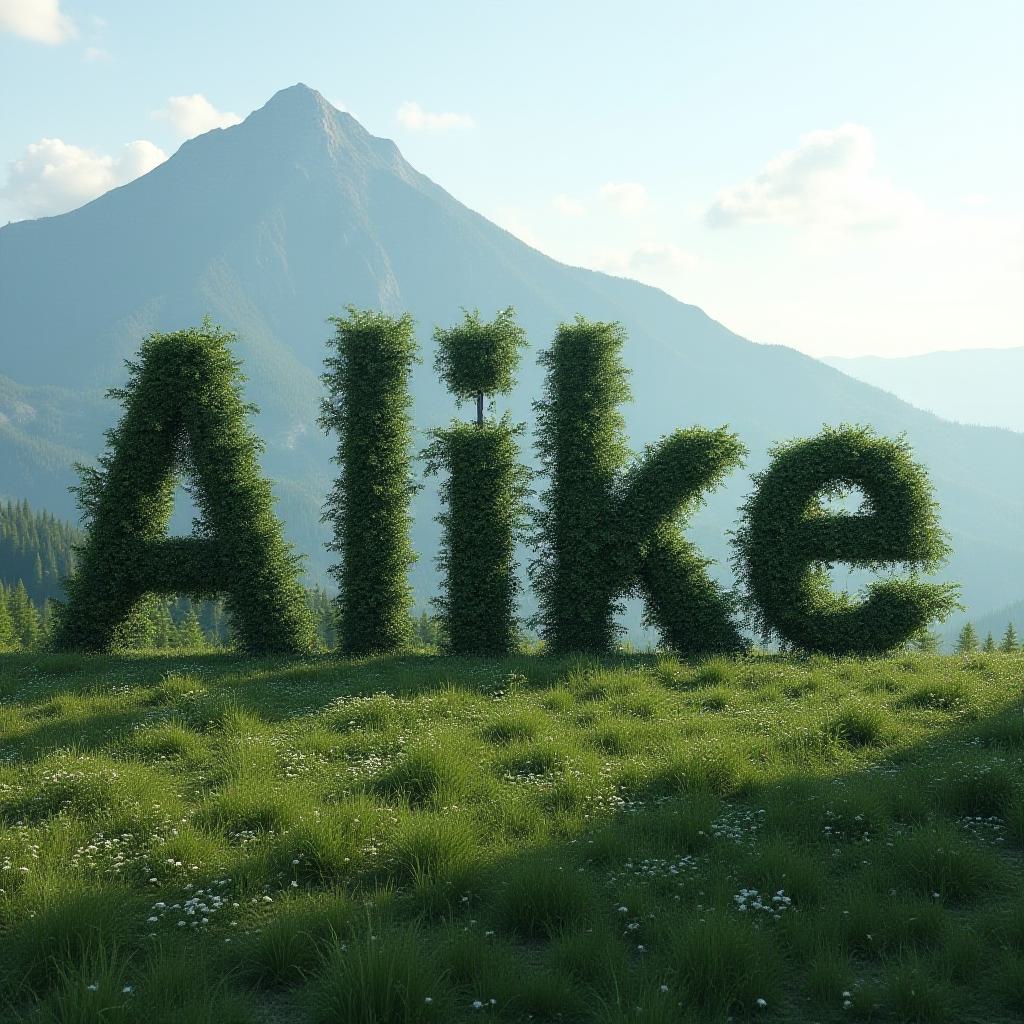}
        \vspace{5px}
        \\

        \includegraphics[valign=c, width=\ww]{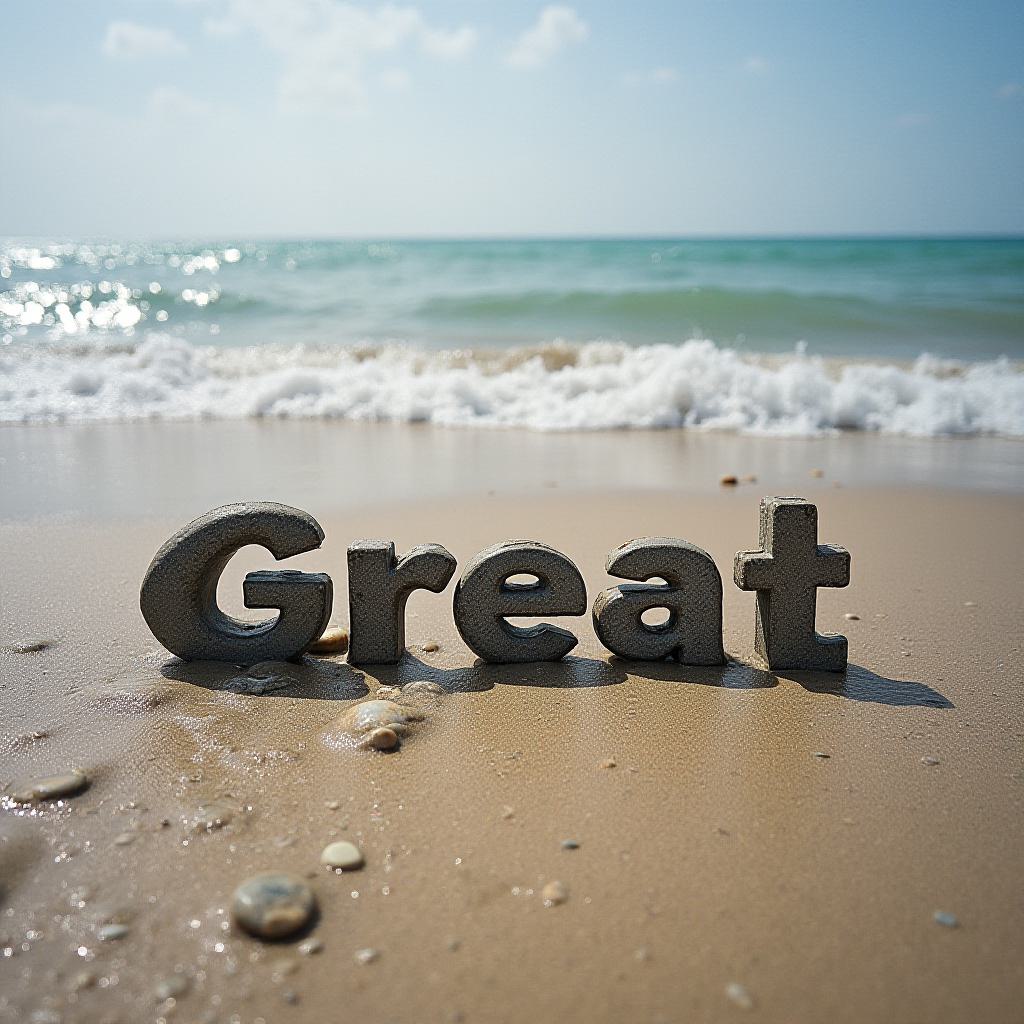} &
        \includegraphics[valign=c, width=\ww]{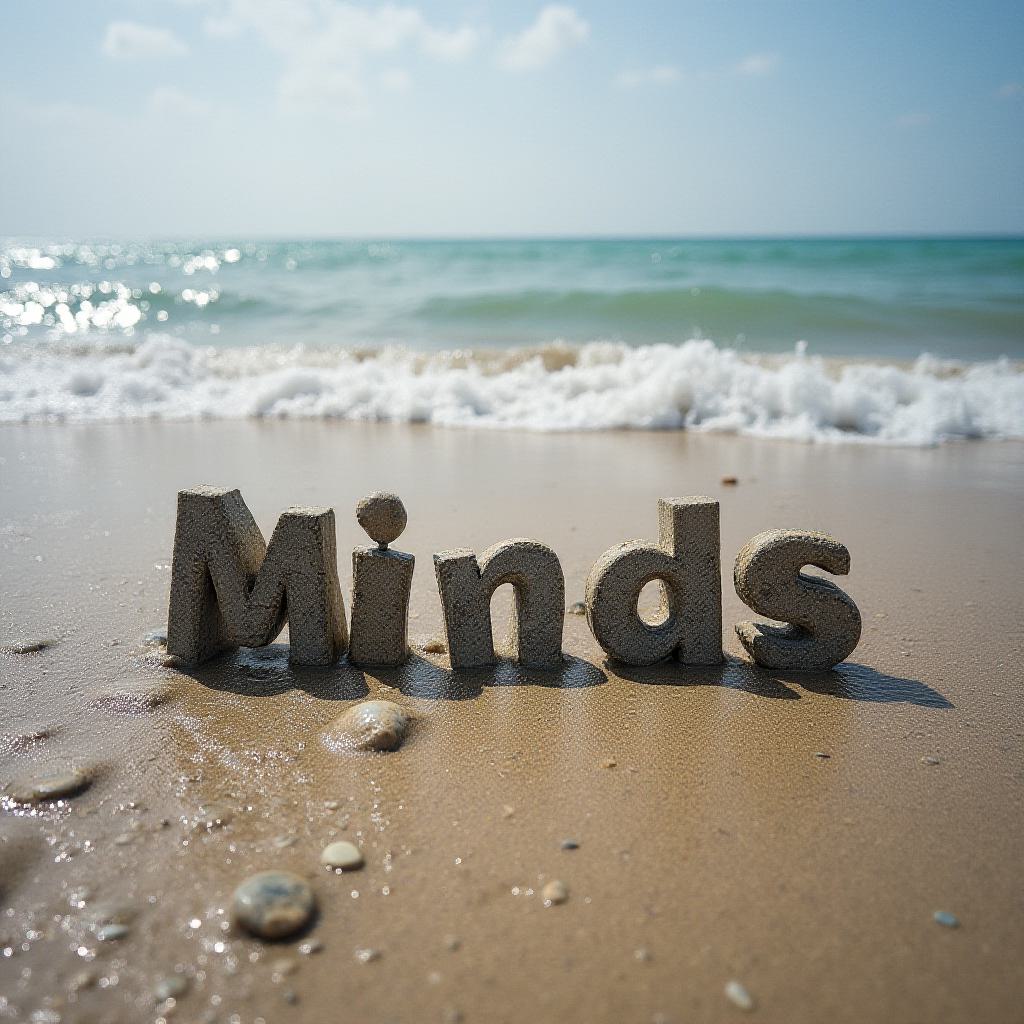} &
        \includegraphics[valign=c, width=\ww]{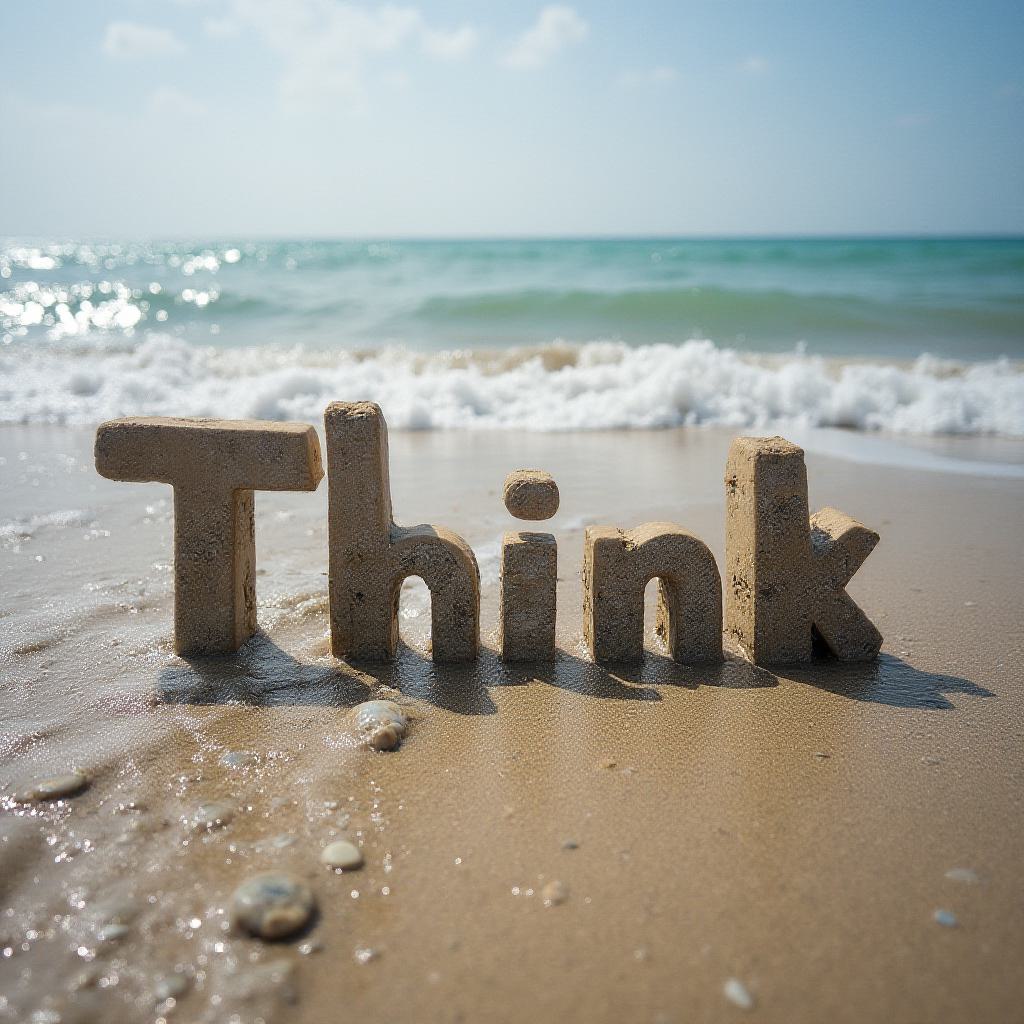} &
        \includegraphics[valign=c, width=\ww]{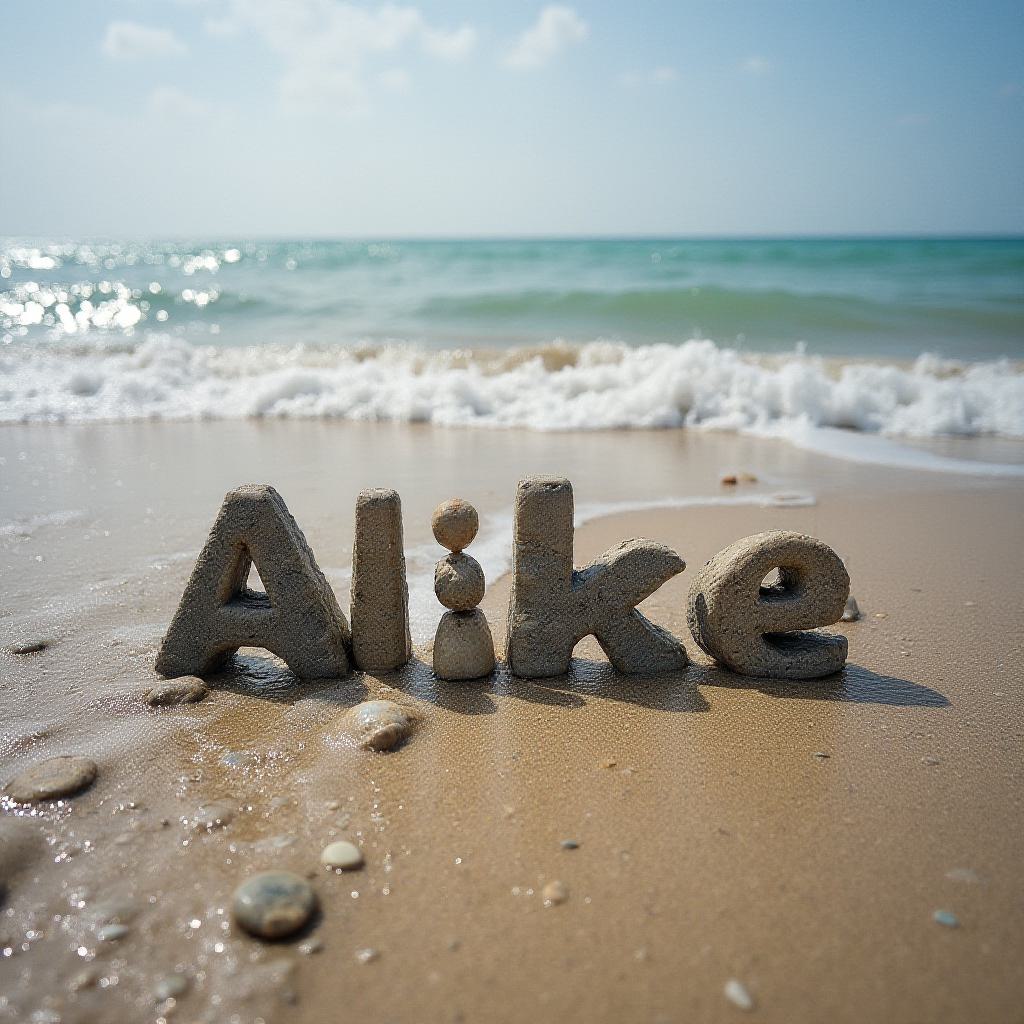}
        \vspace{5px}
        \\

        \includegraphics[valign=c, width=\ww]{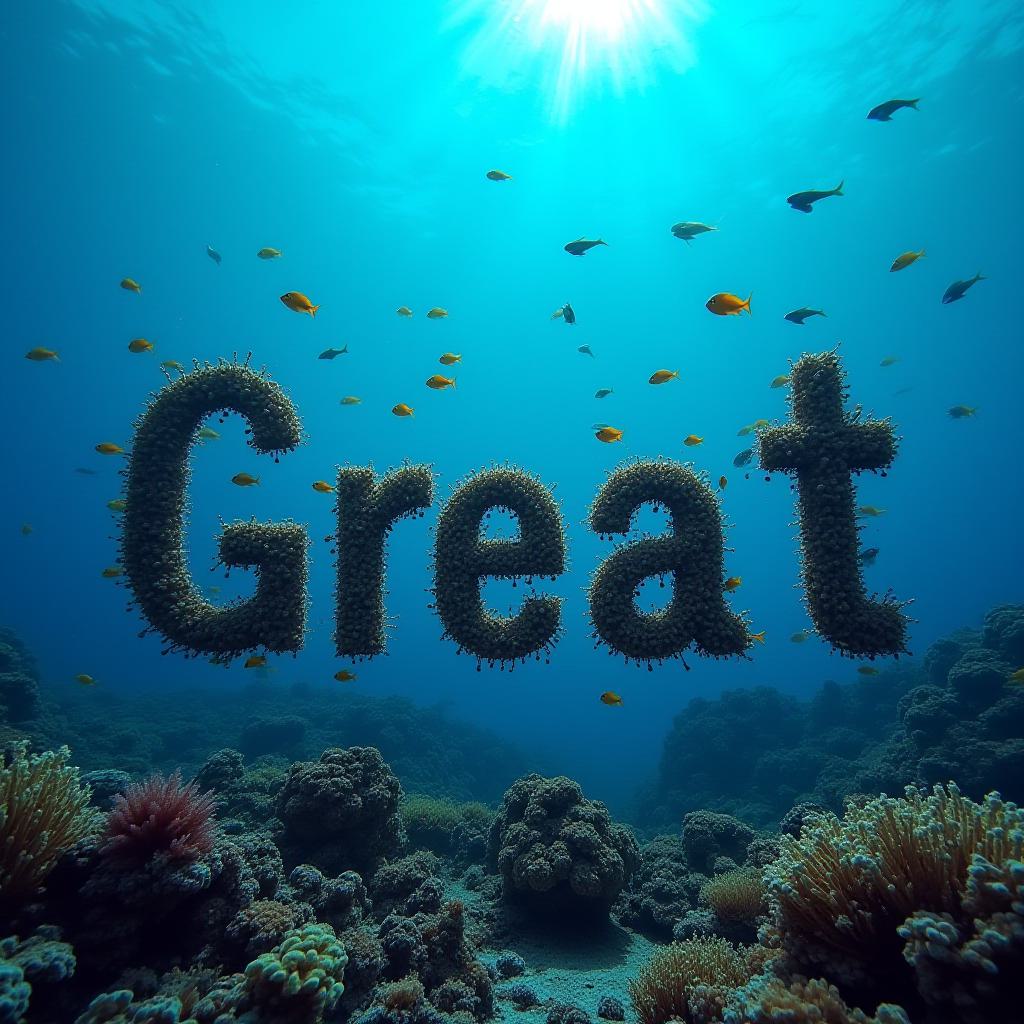} &
        \includegraphics[valign=c, width=\ww]{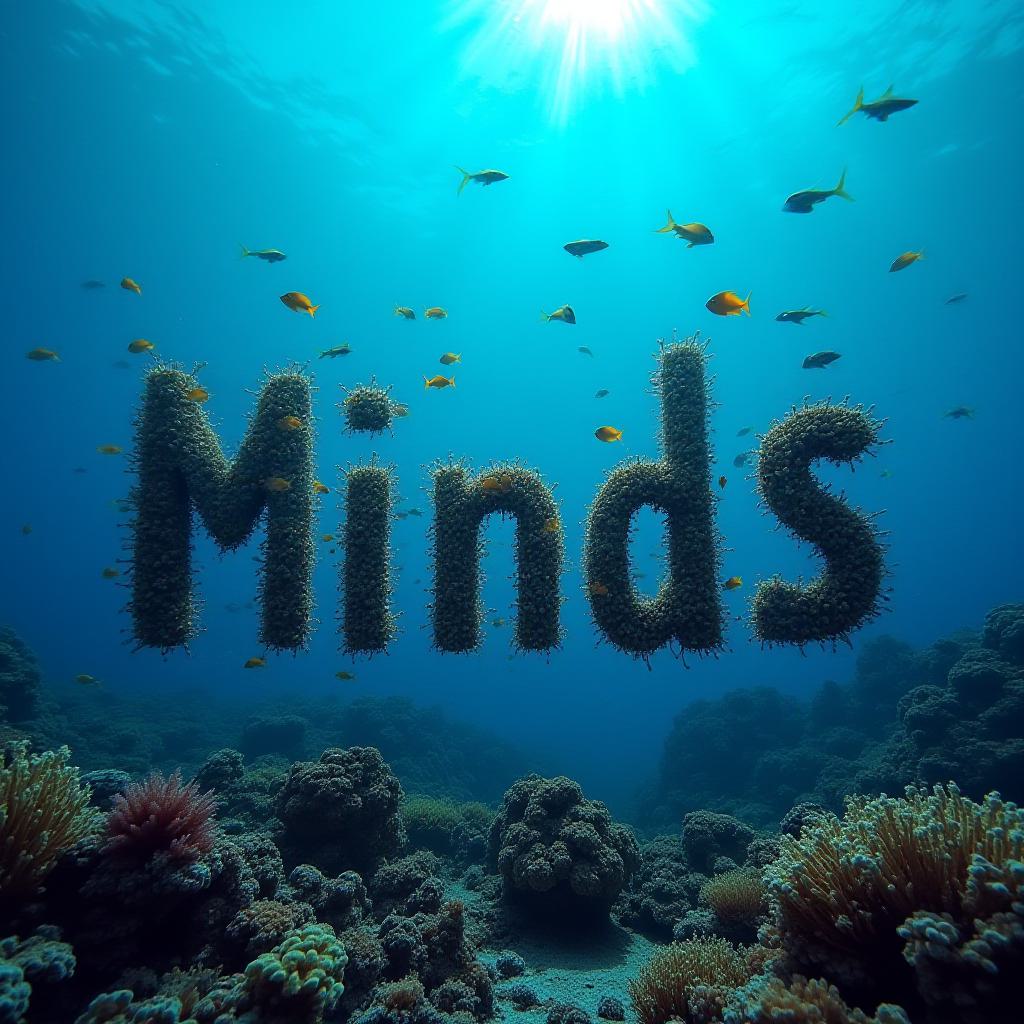} &
        \includegraphics[valign=c, width=\ww]{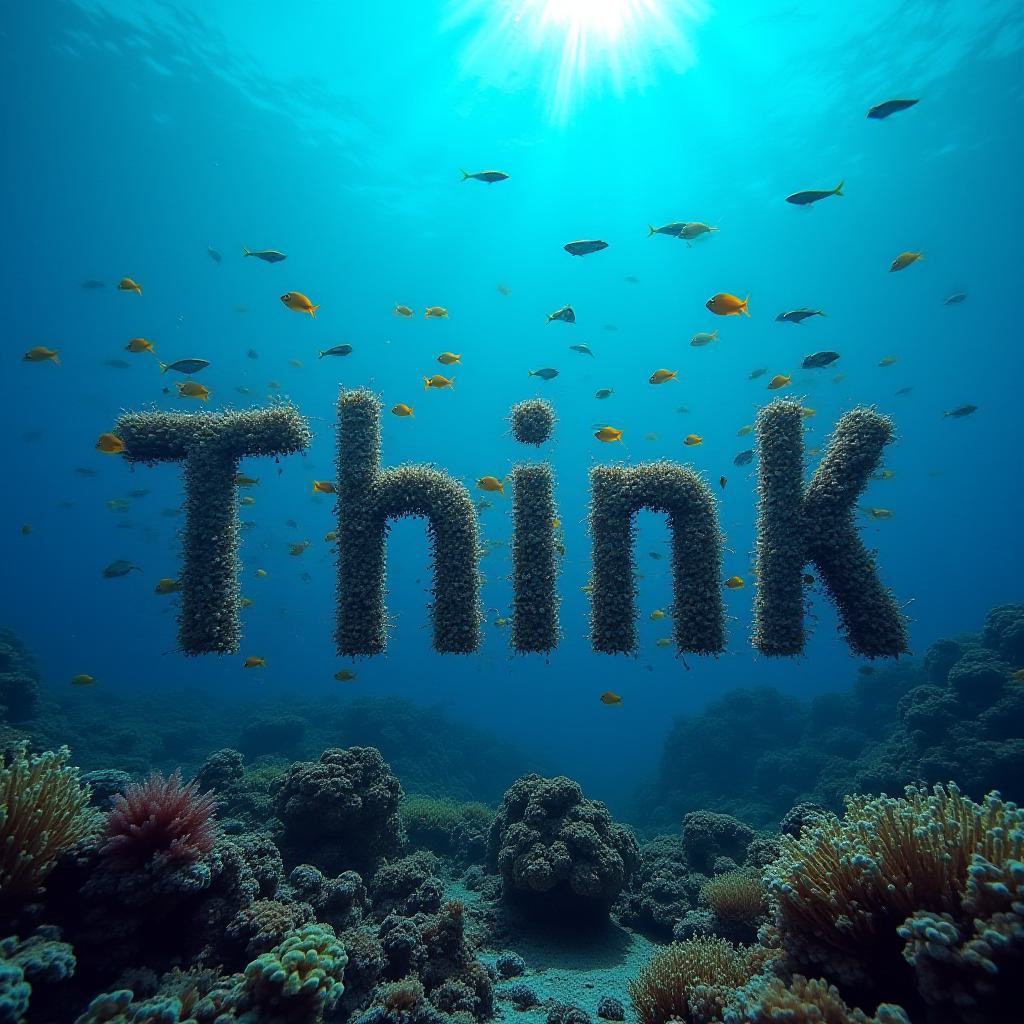} &
        \includegraphics[valign=c, width=\ww]{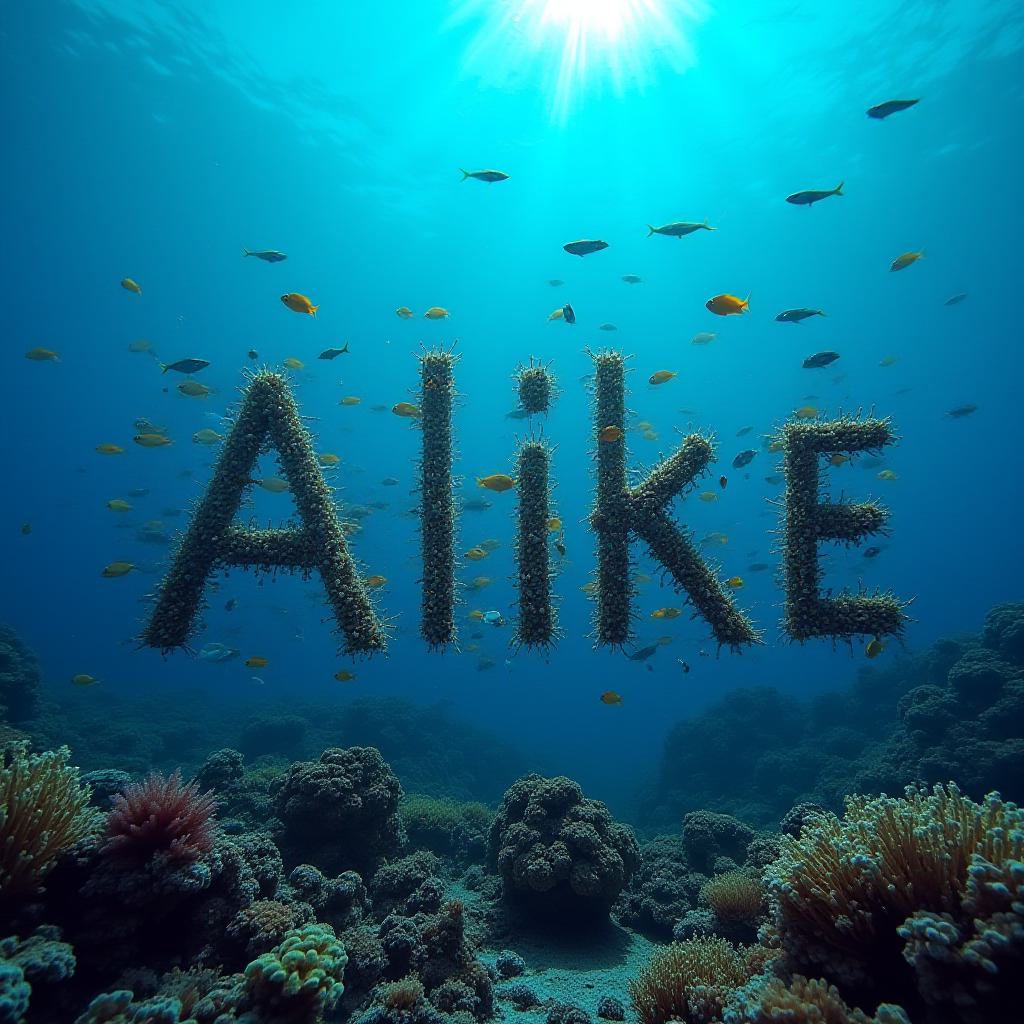}
        \vspace{5px}
        \\

        \includegraphics[valign=c, width=\ww]{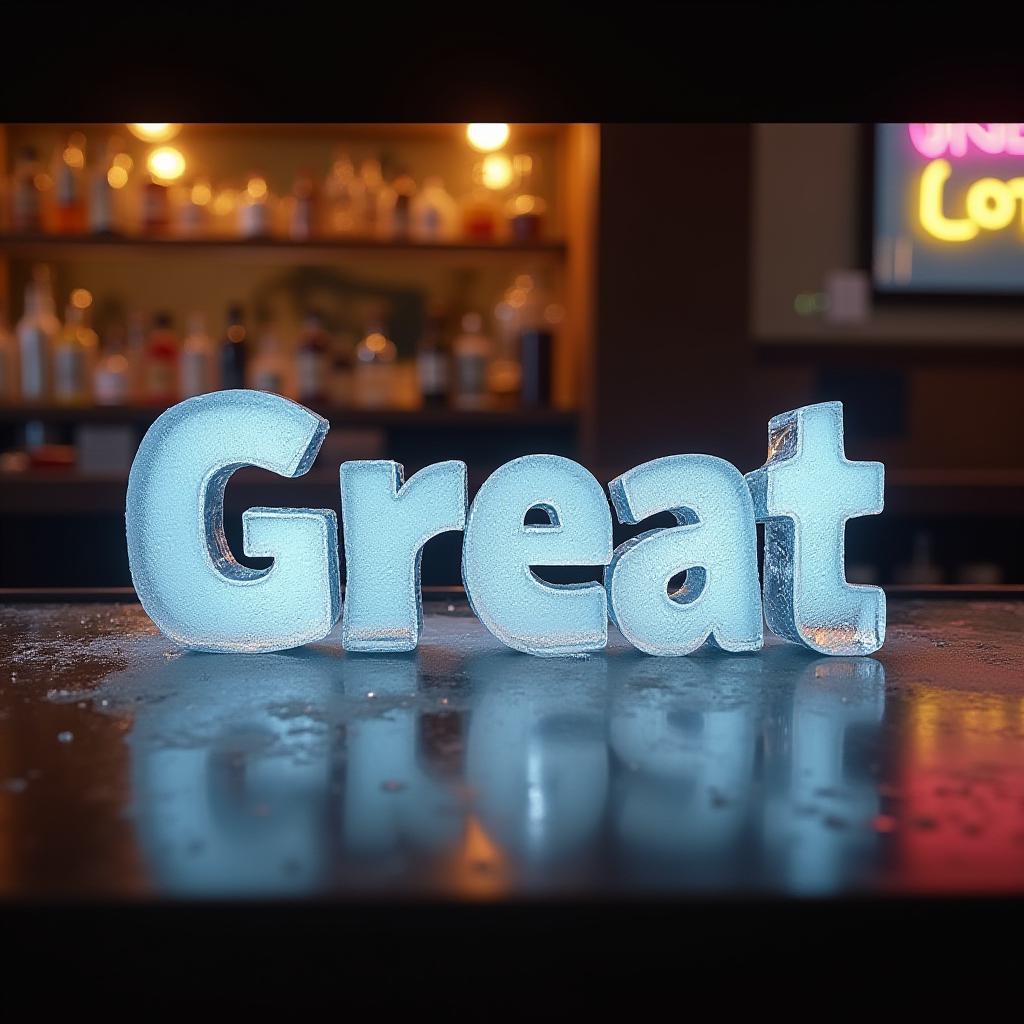} &
        \includegraphics[valign=c, width=\ww]{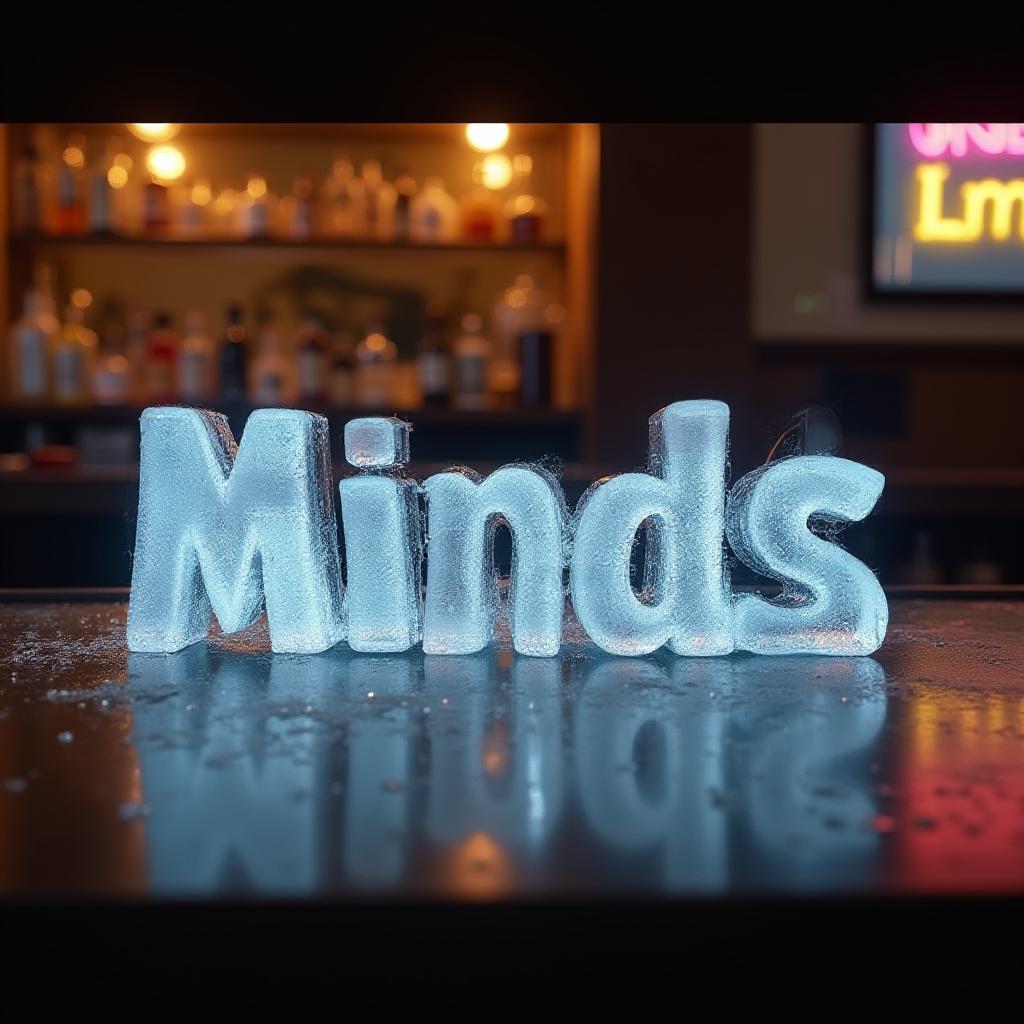} &
        \includegraphics[valign=c, width=\ww]{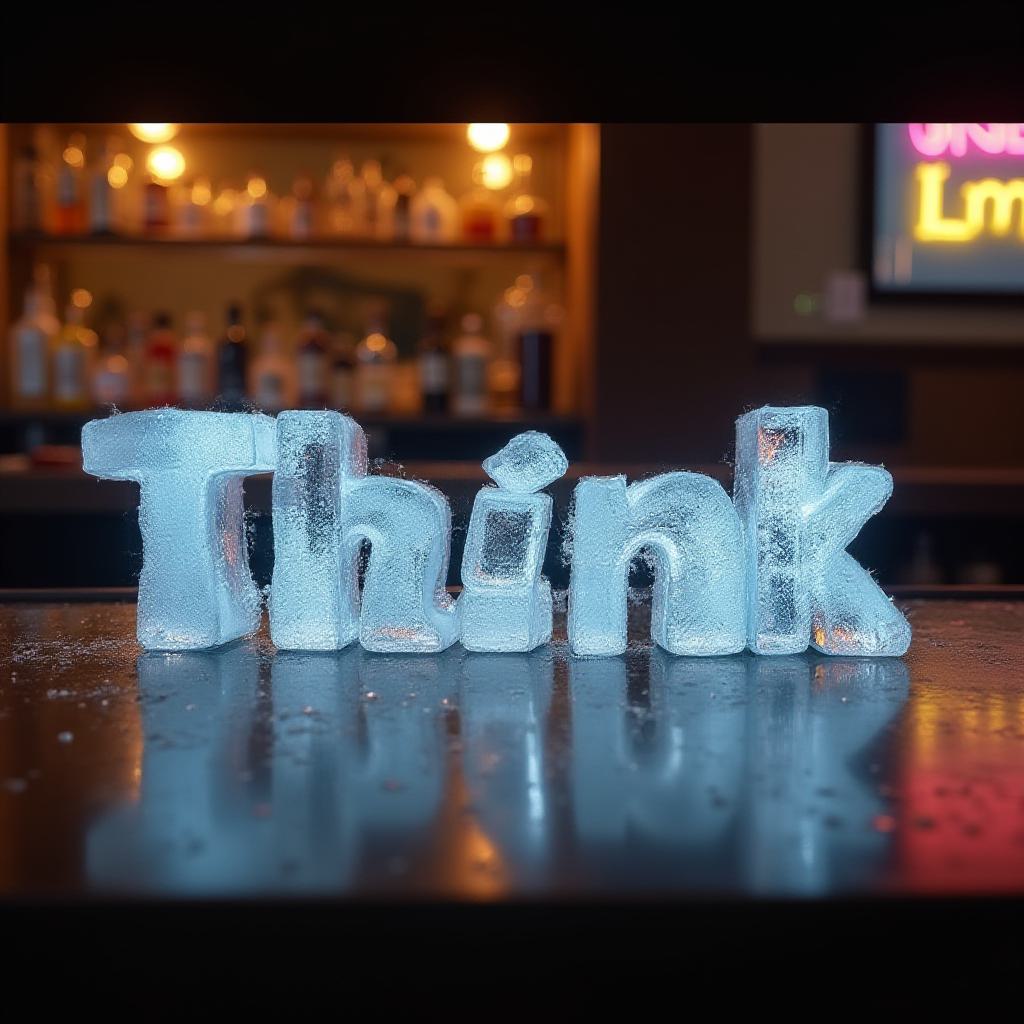} &
        \includegraphics[valign=c, width=\ww]{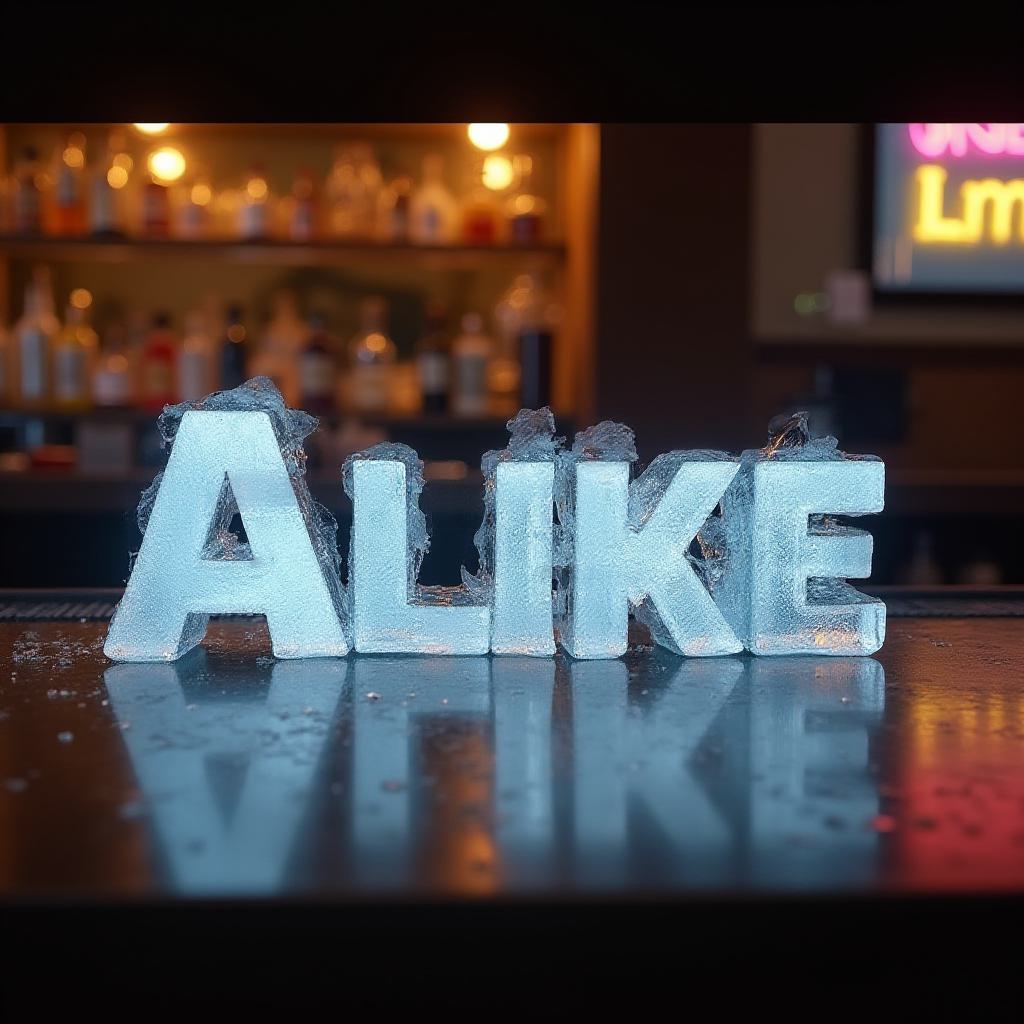}
        \vspace{3px}
        \\

        \small{Input} &
        \small{\prompt{Minds}} &
        \small{\prompt{Think}} &
        \small{\prompt{Alike}}
        \\

    \end{tabular}
    \caption{\textbf{Additional Results.} Given an input image that contain a text, our method cat edit the text while keeping the background and style.}
    \label{fig:additional_results3}
\end{figure*}

In \Cref{fig:baselines_qualitative_comparison_automatic} we provide an additional qualitative comparison of our method against the baselines on real images extracted from the COCO~\cite{Lin2014MicrosoftCC} dataset, as explained in 
\Cref{sec:comparisons}.
As can be seen, SDEdit~\cite{meng2021sdedit} struggles with preserving the object identities and backgrounds (\eg, the bear and chicken examples). P2P+NTI ~\cite{Hertz2022PrompttoPromptIE, mokady2022null} struggles with preserving object identities (\eg, the bear and person examples) and with adding new objects (\eg, the missing hat in the sheep example and missing ball in the elephant example). Instruct-P2P~\cite{brooks2022instructpix2pix} and MagicBrush~\cite{Zhang2023MagicBrush} struggle with non-rigid editing (\eg, the person raising hand example). MasaCTRL~\cite{cao2023masactrl} struggles with preserving object identities (\eg, the bear and person examples) and adding new objects (\eg, the sheep and cat examples). Our method, on the other hand, is able to adhere to the editing prompt while preserving the identities.

Next, in \Cref{fig:ablation_qualitative_comparison_automatic}, we provide a qualitative comparison of the ablated cases that are explained in, 
\Cref{sec:ablation_study}.
As can be seen, we found that (1) performing attention injection in all the layers or performing (3) an attention extension in all the layers, encourages the model to directly copy the input image while neglecting the target prompt. In addition, (2) performing an attention extension in the non-vital layers or (4) removing the latent nudging reduces the input image similarity significantly.

Finally, in Figures \ref{fig:additional_results1} and \ref{fig:additional_results2}, we present additional image editing results using our method.

\subsection{Different Perceptual Metrics}
\label{sec:different_perceptual_metrics}

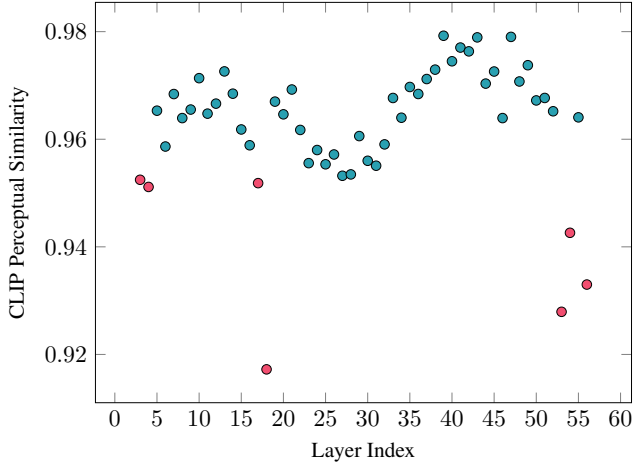
\begin{figure}[t]
    \begin{tikzpicture} [thick,scale=0.9, every node/.style={scale=1}]
        \begin{axis}[
            xlabel={\small{Layer Index}},
            ylabel={\small{CLIP Perceptual Similarity}},
            compat=newest,
            xtick distance=5,
            width=9.5cm,
            height=7.5cm,
            scatter/classes={
                a={mark=*, fill=nonvitalcolor},
                b={mark=*, fill=vitalcolor}
            }
            ]
            \addplot[scatter, only marks, scatter src=explicit symbolic]
            table[meta=label] {
                x y label
                3 0.9524642825126648 b
                4 0.9511328935623169 b
                5 0.9653081297874451 a
                6 0.9586458802223206 a
                7 0.9684000015258789 a
                8 0.963932454586029 a
                9 0.9655178785324097 a
                10 0.9713525772094727 a
                11 0.9647583961486816 a
                12 0.966618537902832 a
                13 0.9726079702377319 a
                14 0.9684693813323975 a
                15 0.961816668510437 a
                16 0.9588773250579834 a
                17 0.9518409967422485 b
                18 0.9172269105911255 b
                19 0.9669922590255737 a
                20 0.9646296501159668 a
                21 0.9692423939704895 a
                22 0.961728572845459 a
                23 0.9555586576461792 a
                24 0.9579951763153076 a
                25 0.955342173576355 a
                26 0.9571689367294312 a
                27 0.9532103538513184 a
                28 0.9534640312194824 a
                29 0.9605858325958252 a
                30 0.9559949636459351 a
                31 0.9550789594650269 a
                32 0.9590439200401306 a
                33 0.967668890953064 a
                34 0.9639999866485596 a
                35 0.9697138071060181 a
                36 0.9684097170829773 a
                37 0.9711917638778687 a
                38 0.9729384779930115 a
                39 0.9792381525039673 a
                40 0.9744912385940552 a
                41 0.9770424962043762 a
                42 0.9763350486755371 a
                43 0.9789327383041382 a
                44 0.9703418016433716 a
                45 0.9725882411003113 a
                46 0.9639327526092529 a
                47 0.9790254831314087 a
                48 0.9707351326942444 a
                49 0.9737594127655029 a
                50 0.9672143459320068 a
                51 0.9676733016967773 a
                52 0.9651976823806763 a
                53 0.9279221296310425 b
                54 0.942609429359436 b
                55 0.964067816734314 a
                56 0.9329898357391357 b
            };
            \end{axis}
    \end{tikzpicture}

    \caption{\textbf{Layer Removal Quantitative Comparison Using CLIP.} As explained in \Cref{sec:different_perceptual_metrics}, we measured the effect of removing each layer of the model by calculating the \emph{CLIP}~\cite{Radford2021LearningTV} perceptual similarity between the generated images with and without this layer. Lower perceptual similarity indicates significant changes in the generated images. As can be seen, removing certain layers significantly affects the generated images, while others have minimal impact. Importantly, influential layers are distributed across the transformer rather than concentrated in specific regions. Note that the first vital layers were omitted for clarity (as their perceptual similarity approached zero).}
    \label{fig:layer_removal_quantiative_clip}
\end{figure}

\begin{figure}[t]
    \begin{tikzpicture} [thick,scale=0.9, every node/.style={scale=1}]
        \begin{axis}[
            xlabel={\small{Layer Index}},
            ylabel={\small{DINOv1 Perceptual Similarity}},
            compat=newest,
            xtick distance=5,
            width=9.5cm,
            height=7.5cm,
            scatter/classes={
                a={mark=*, fill=nonvitalcolor},
                b={mark=*, fill=vitalcolor}
            }
            ]
            \addplot[scatter, only marks, scatter src=explicit symbolic]
            table[meta=label] {
                x y label
                3 0.8859560489654541 a
                4 0.8807618021965027 a
                5 0.9133185148239136 a
                6 0.890884518623352 a
                7 0.9067306518554688 a
                8 0.8891706466674805 a
                9 0.9108058214187622 a
                10 0.92772376537323 a
                11 0.9060047268867493 a
                12 0.9168164730072021 a
                13 0.9420595169067383 a
                14 0.9167765378952026 a
                15 0.8989335298538208 a
                16 0.8900765180587769 a
                17 0.8631271719932556 b
                18 0.778456449508667 b
                19 0.8964021801948547 a
                20 0.8851877450942993 a
                21 0.9078318476676941 a
                22 0.8843109607696533 a
                23 0.8816295266151428 a
                24 0.8728955388069153 a
                25 0.8591060638427734 b
                26 0.8785309195518494 a
                27 0.8719003200531006 a
                28 0.8616998195648193 b
                29 0.8913426995277405 a
                30 0.8836356401443481 a
                31 0.8792330622673035 a
                32 0.8854716420173645 a
                33 0.9109123945236206 a
                34 0.9098462462425232 a
                35 0.9213109016418457 a
                36 0.9207352995872498 a
                37 0.9261149168014526 a
                38 0.9307715892791748 a
                39 0.9394903182983398 a
                40 0.932727575302124 a
                41 0.9457308053970337 a
                42 0.9364351630210876 a
                43 0.9341707229614258 a
                44 0.9192824363708496 a
                45 0.924848198890686 a
                46 0.9195623397827148 a
                47 0.944649875164032 a
                48 0.9243229627609253 a
                49 0.9229422211647034 a
                50 0.9147465229034424 a
                51 0.9051727056503296 a
                52 0.911866307258606 a
                53 0.8070423603057861 b
                54 0.8458688259124756 b
                55 0.9061789512634277 a
                56 0.8417607545852661 b
            };
            \end{axis}
    \end{tikzpicture}

    \caption{\textbf{Layer Removal Quantitative Comparison Using DINOv1.} As explained in \Cref{sec:different_perceptual_metrics}, we measured the effect of removing each layer of the model by calculating the \emph{DINOv1}~\cite{Caron2021EmergingPI} perceptual similarity between the generated images with and without this layer. Lower perceptual similarity indicates significant changes in the generated images. As can be seen, removing certain layers significantly affects the generated images, while others have minimal impact. Importantly, influential layers are distributed across the transformer rather than concentrated in specific regions. Note that the first vital layers were omitted for clarity (as their perceptual similarity approached zero).}
    \label{fig:layer_removal_quantiative_dino_v1}
\end{figure}
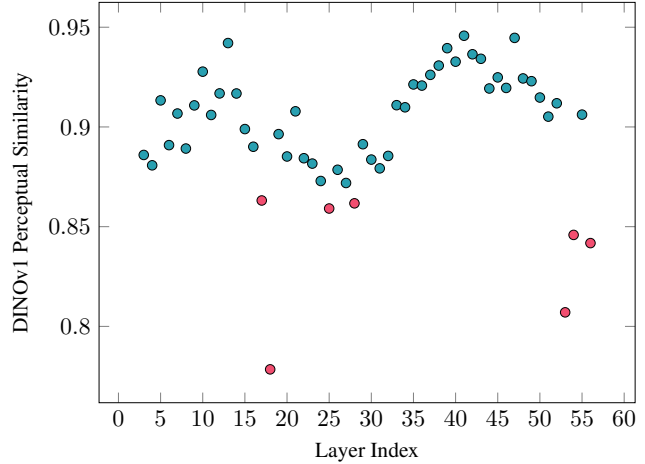

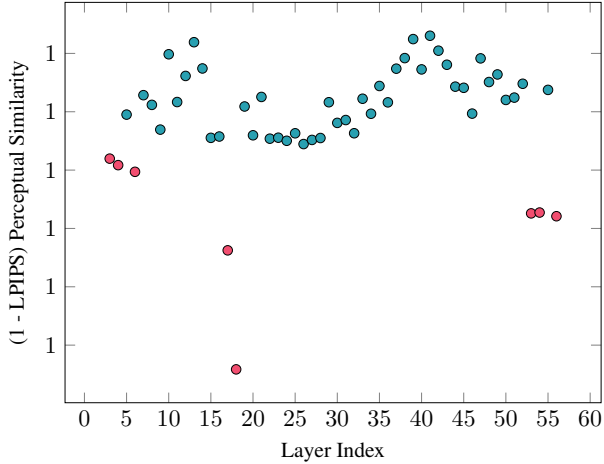
\begin{figure}[t]
    \begin{tikzpicture} [thick,scale=0.9, every node/.style={scale=1}]
        \begin{axis}[
            xlabel={\small{Layer Index}},
            ylabel={\small{(1 - LPIPS) Perceptual Similarity}},
            compat=newest,
            xtick distance=5,
            width=9.5cm,
            height=7.5cm,
            scatter/classes={
                a={mark=*, fill=nonvitalcolor},
                b={mark=*, fill=vitalcolor}
            }
            ]
            \addplot[scatter, only marks, scatter src=explicit symbolic]
            table[meta=label] {
                x y label
                3 0.9994391202926636 b
                4 0.9994168281555176 b
                5 0.9995900392532349 a
                6 0.9993939399719238 b
                7 0.9996566772460938 a
                8 0.9996234178543091 a
                9 0.9995386004447937 a
                10 0.9997968673706055 a
                11 0.9996331334114075 a
                12 0.9997225999832153 a
                13 0.9998382329940796 a
                14 0.9997482299804688 a
                15 0.9995104074478149 a
                16 0.9995149374008179 a
                17 0.9991247653961182 b
                18 0.9987167716026306 b
                19 0.999617874622345 a
                20 0.9995191693305969 a
                21 0.9996505975723267 a
                22 0.9995075464248657 a
                23 0.9995108842849731 a
                24 0.9995003342628479 a
                25 0.9995260238647461 a
                26 0.9994890689849854 a
                27 0.9995032548904419 a
                28 0.9995098114013672 a
                29 0.9996324777603149 a
                30 0.9995616674423218 a
                31 0.9995714426040649 a
                32 0.9995261430740356 a
                33 0.9996444582939148 a
                34 0.9995930194854736 a
                35 0.9996881484985352 a
                36 0.9996320605278015 a
                37 0.9997475147247314 a
                38 0.9997837543487549 a
                39 0.9998489022254944 a
                40 0.9997450113296509 a
                41 0.999860405921936 a
                42 0.999809205532074 a
                43 0.9997609853744507 a
                44 0.9996858835220337 a
                45 0.9996819496154785 a
                46 0.9995934963226318 a
                47 0.9997828006744385 a
                48 0.999701738357544 a
                49 0.999727725982666 a
                50 0.9996403455734253 a
                51 0.9996483325958252 a
                52 0.9996957778930664 a
                53 0.9992514848709106 b
                54 0.9992543458938599 b
                55 0.9996747970581055 a
                56 0.9992417097091675 b
            };
            \end{axis}
    \end{tikzpicture}

    \caption{\textbf{Layer Removal Quantitative Comparison Using LPIPS.} As explained in \Cref{sec:different_perceptual_metrics}, we measured the effect of removing each layer of the model by calculating the \emph{(1 - LPIPS)}~\cite{Zhang2018TheUE} perceptual similarity between the generated images with and without this layer. Lower perceptual similarity indicates significant changes in the generated images. As can be seen, removing certain layers significantly affects the generated images, while others have minimal impact. Importantly, influential layers are distributed across the transformer rather than concentrated in specific regions. Note that the first vital layers were omitted for clarity (as their perceptual similarity approached zero).}
    \label{fig:layer_removal_quantiative_lpips}
\end{figure}

\begin{figure*}[tp]
    \centering
    \setlength{\tabcolsep}{0.6pt}
    \renewcommand{\arraystretch}{0.8}
    \setlength{\ww}{0.24\linewidth}
    \begin{tabular}{c @{\hspace{10\tabcolsep}} ccc}

        Input &
        DINO~\cite{Caron2021EmergingPI, Oquab2023DINOv2LR} &
        CLIP~\cite{Radford2021LearningTV} &
        LPIPS~\cite{Zhang2018TheUE}
        \vspace{2px}
        \\

        \includegraphics[valign=c, width=\ww]{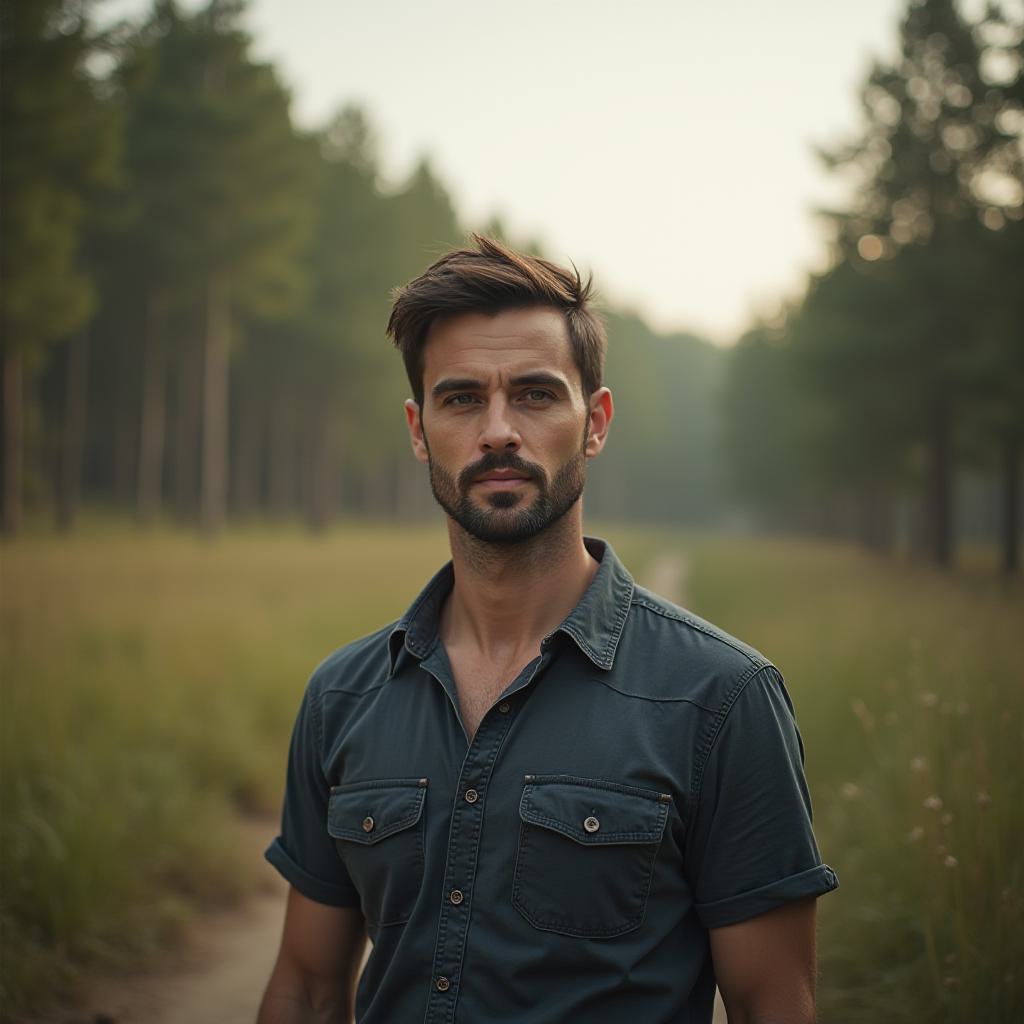} &
        \includegraphics[valign=c, width=\ww]{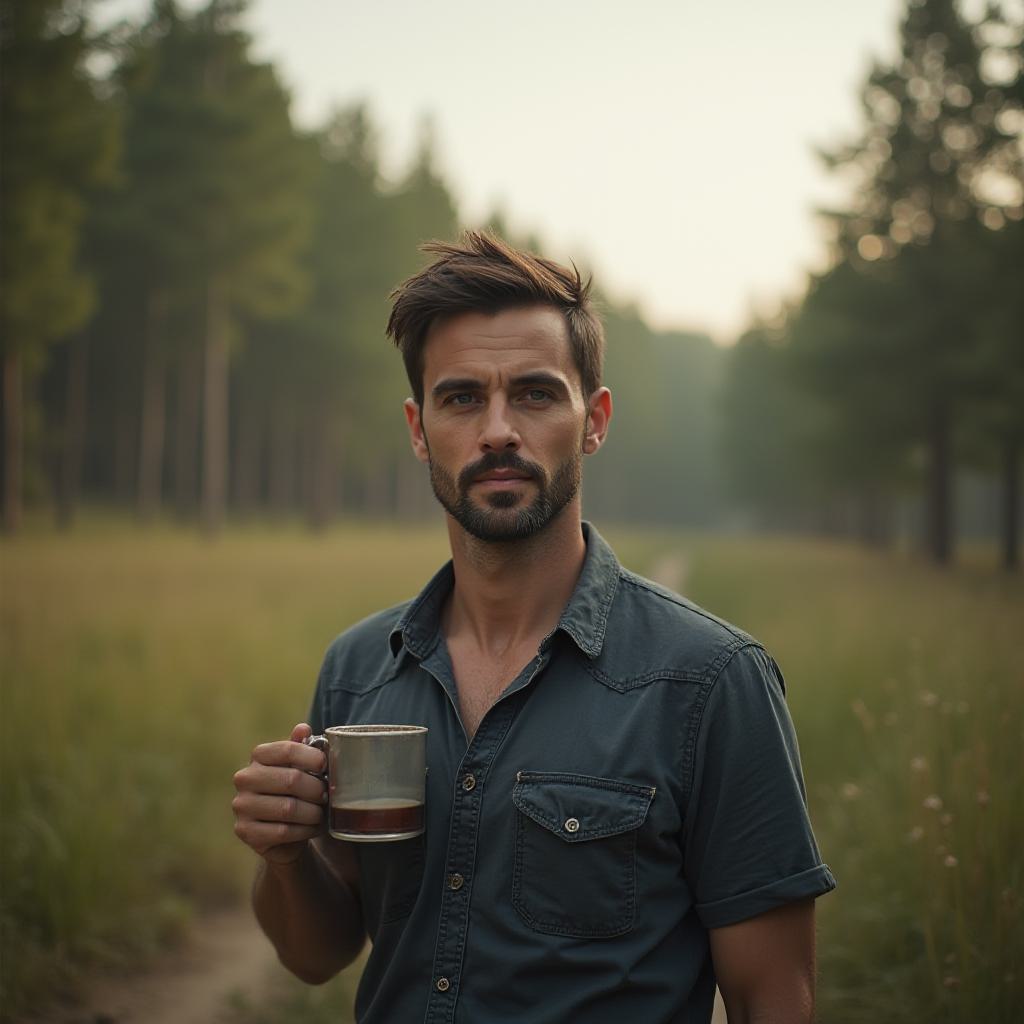} &
        \includegraphics[valign=c, width=\ww]{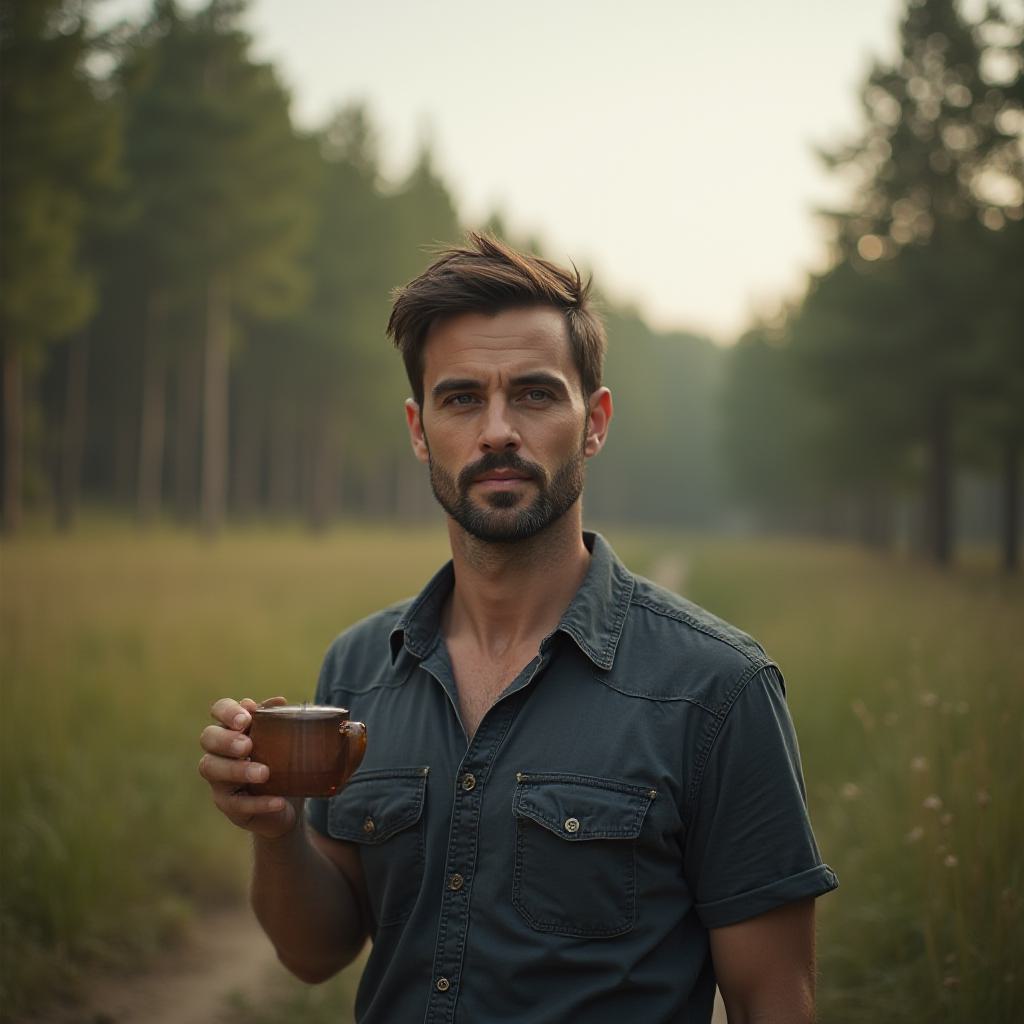} &
        \includegraphics[valign=c, width=\ww]{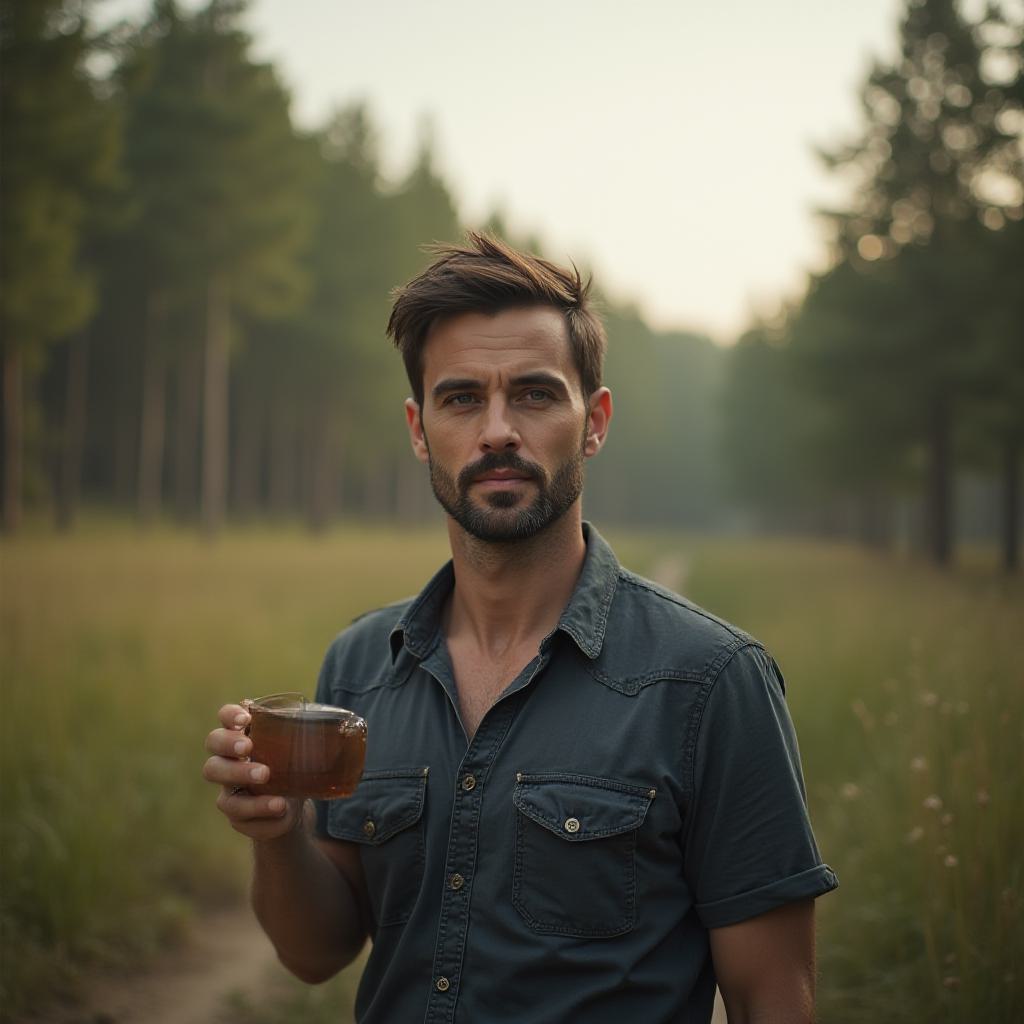}
        \vspace{3px}
        \\

        &
        \multicolumn{3}{c}{\prompt{A man holding a cup of tea}}
        \vspace{10px}
        \\

        \includegraphics[valign=c, width=\ww]{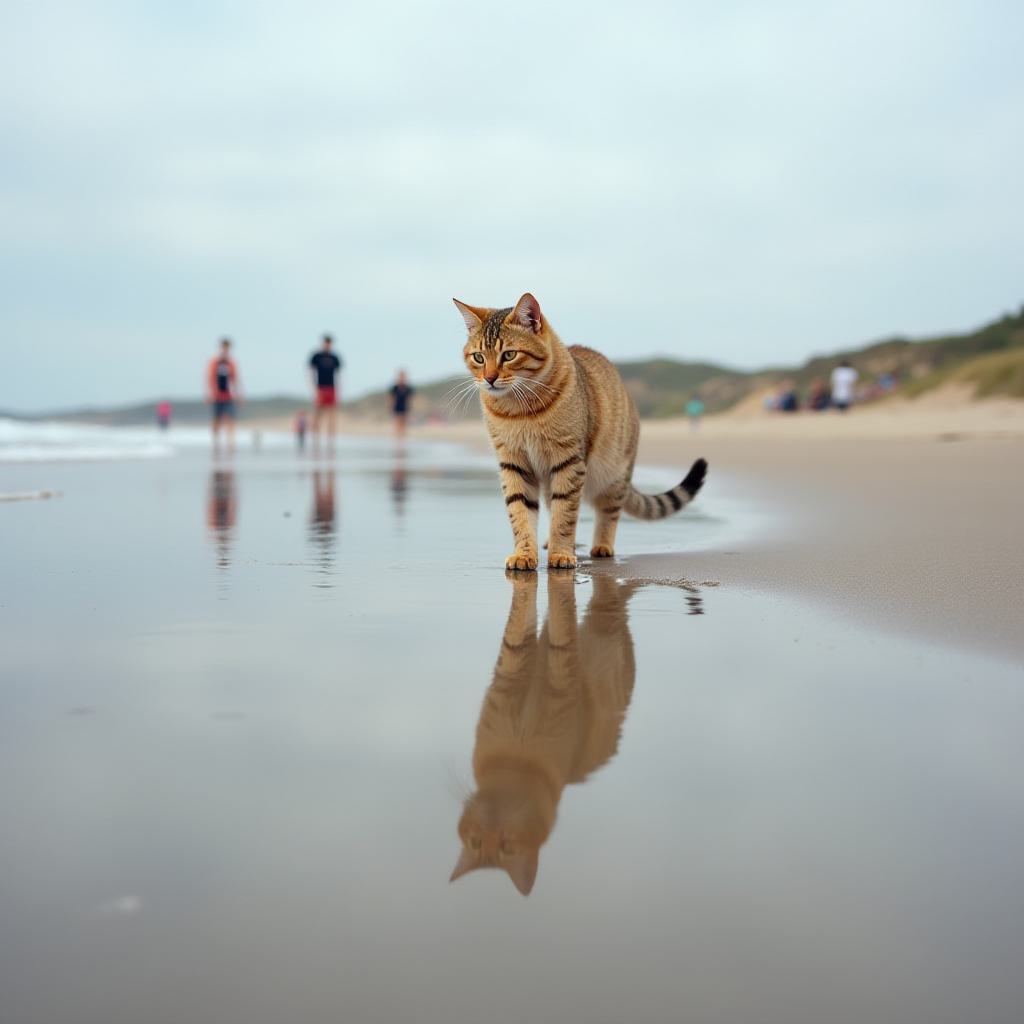} &
        \includegraphics[valign=c, width=\ww]{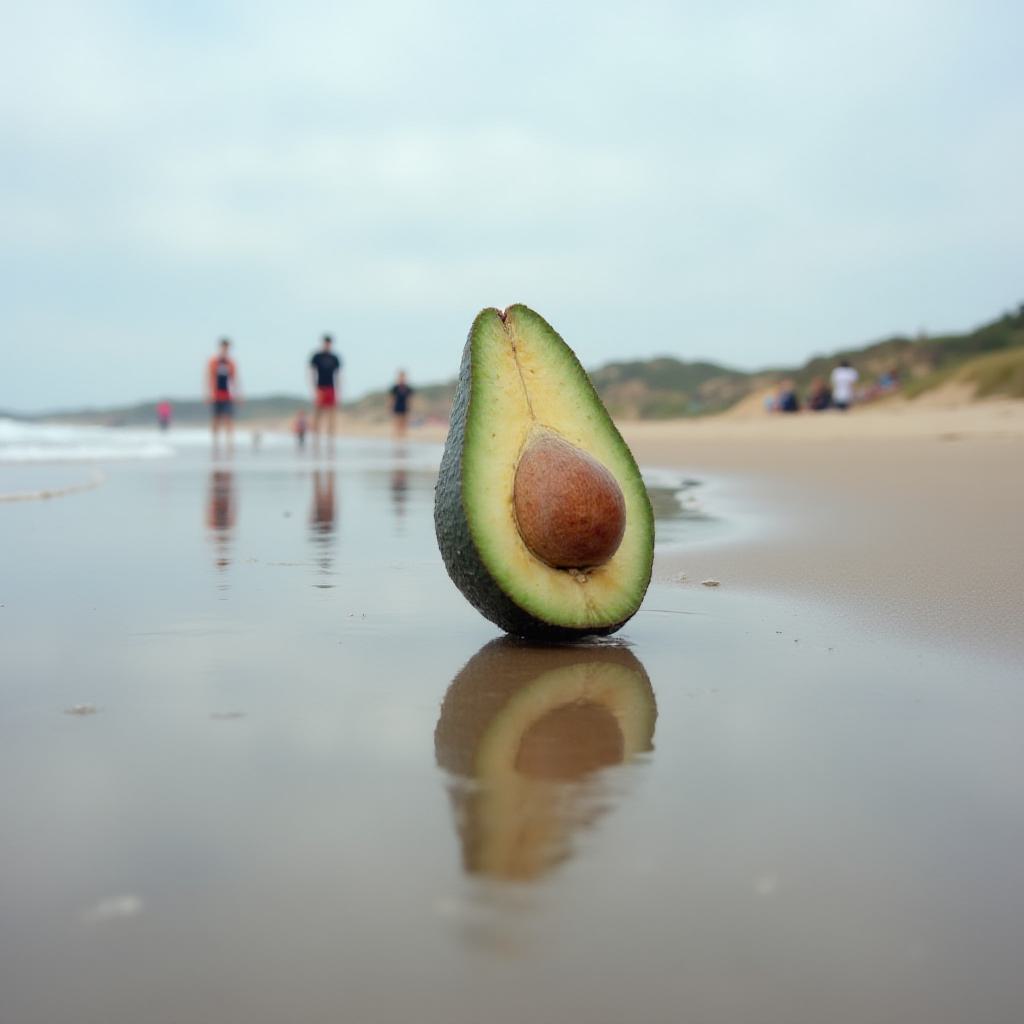} &
        \includegraphics[valign=c, width=\ww]{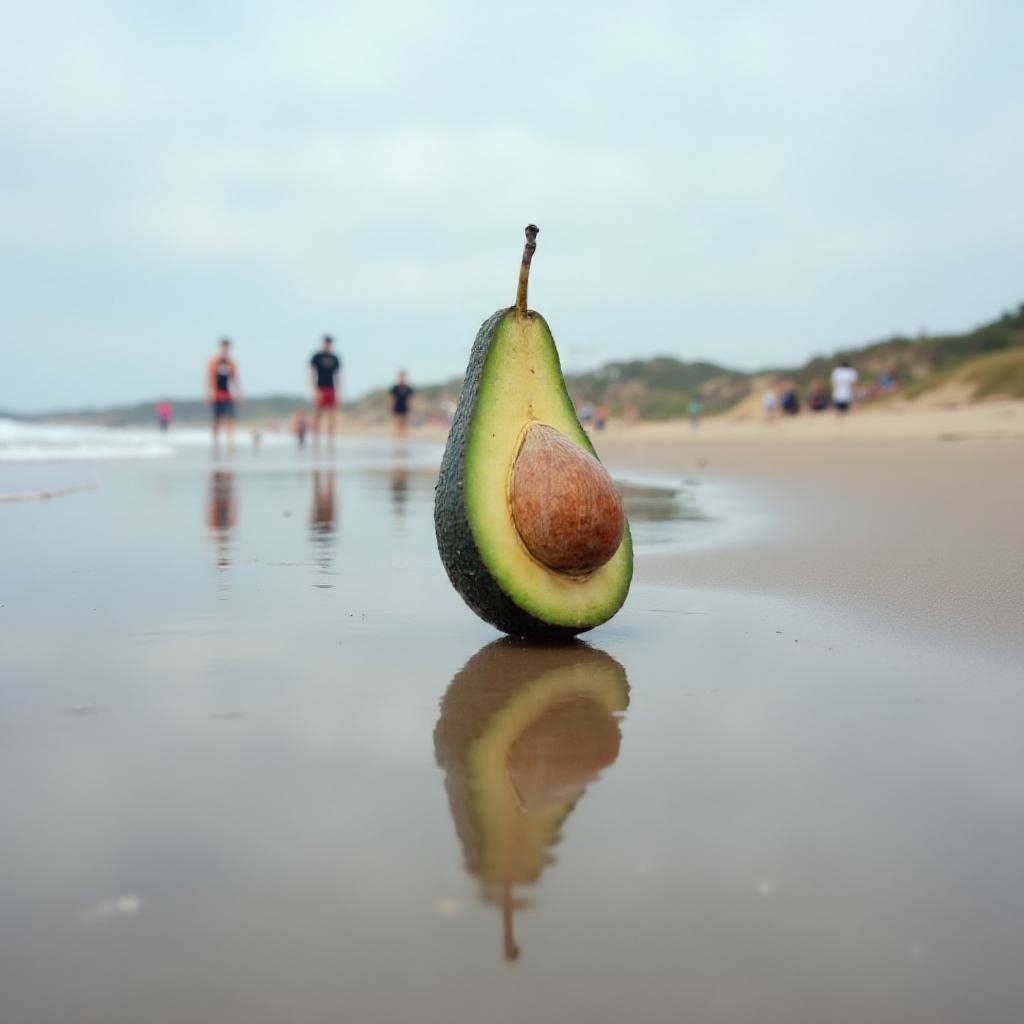} &
        \includegraphics[valign=c, width=\ww]{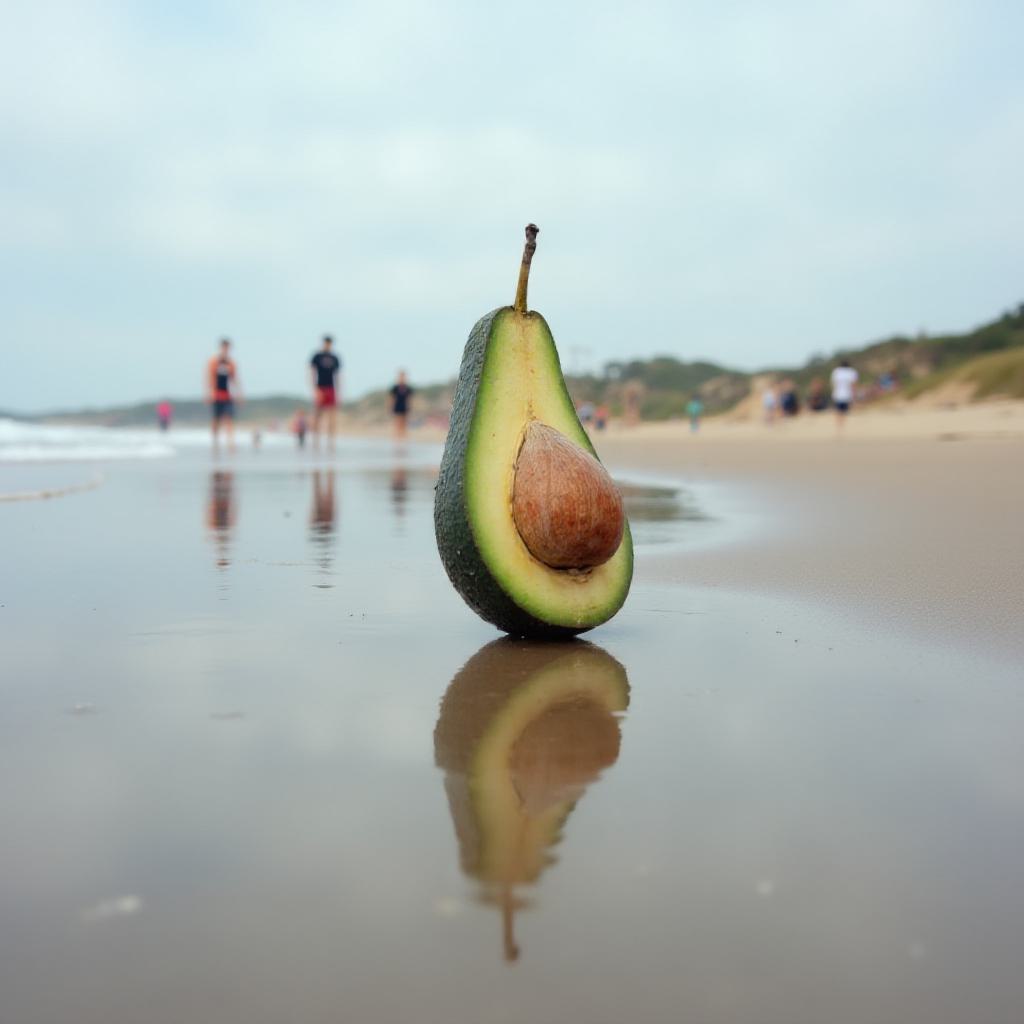}
        \vspace{3px}
        \\

        &
        \multicolumn{3}{c}{\prompt{An avocado at the beach}}
        \vspace{10px}
        \\

        \includegraphics[valign=c, width=\ww]{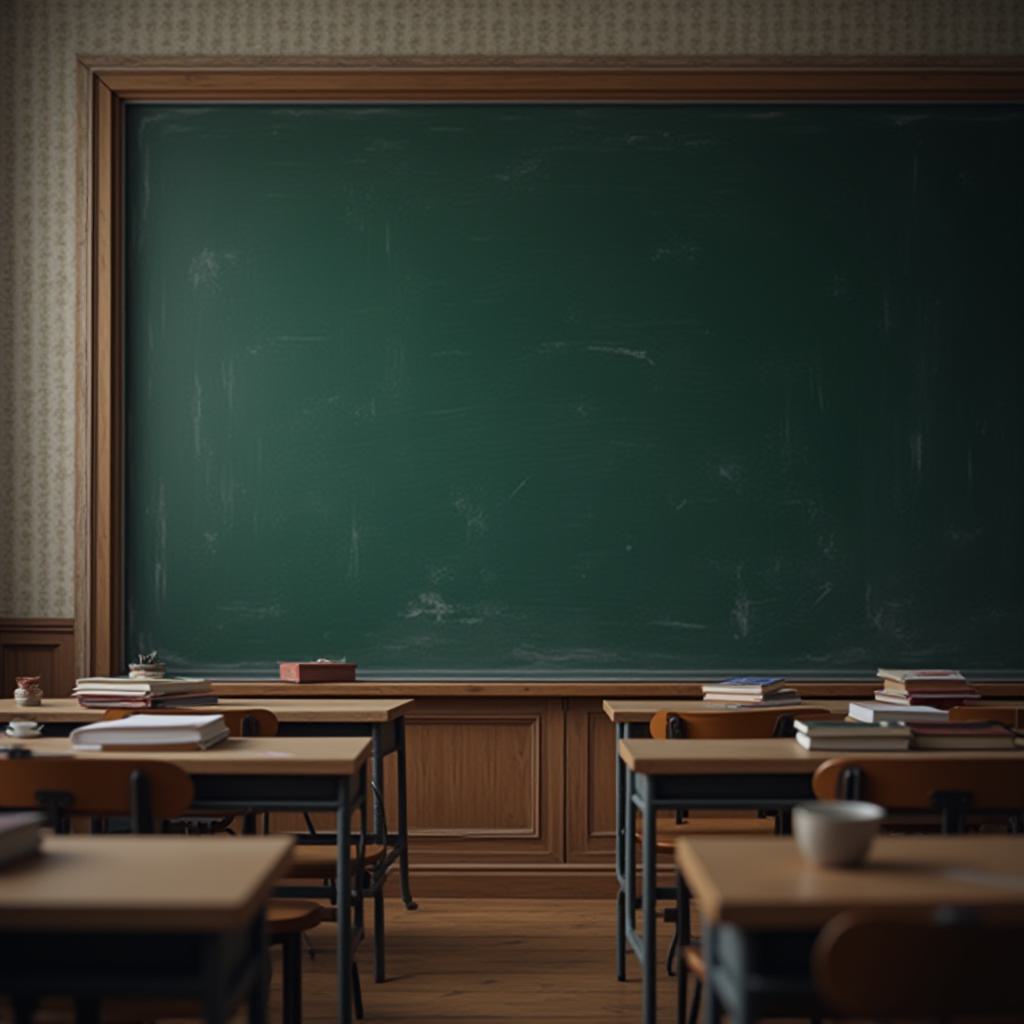} &
        \includegraphics[valign=c, width=\ww]{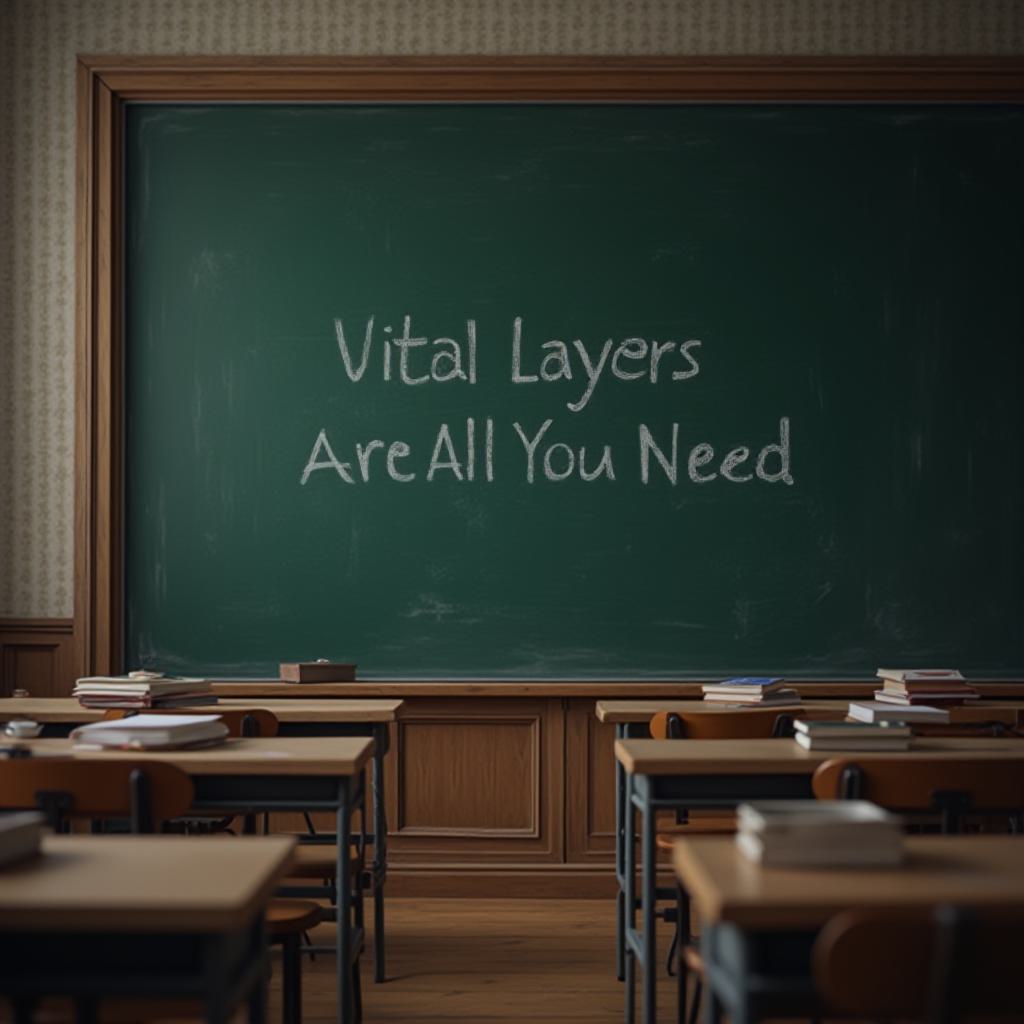} &
        \includegraphics[valign=c, width=\ww]{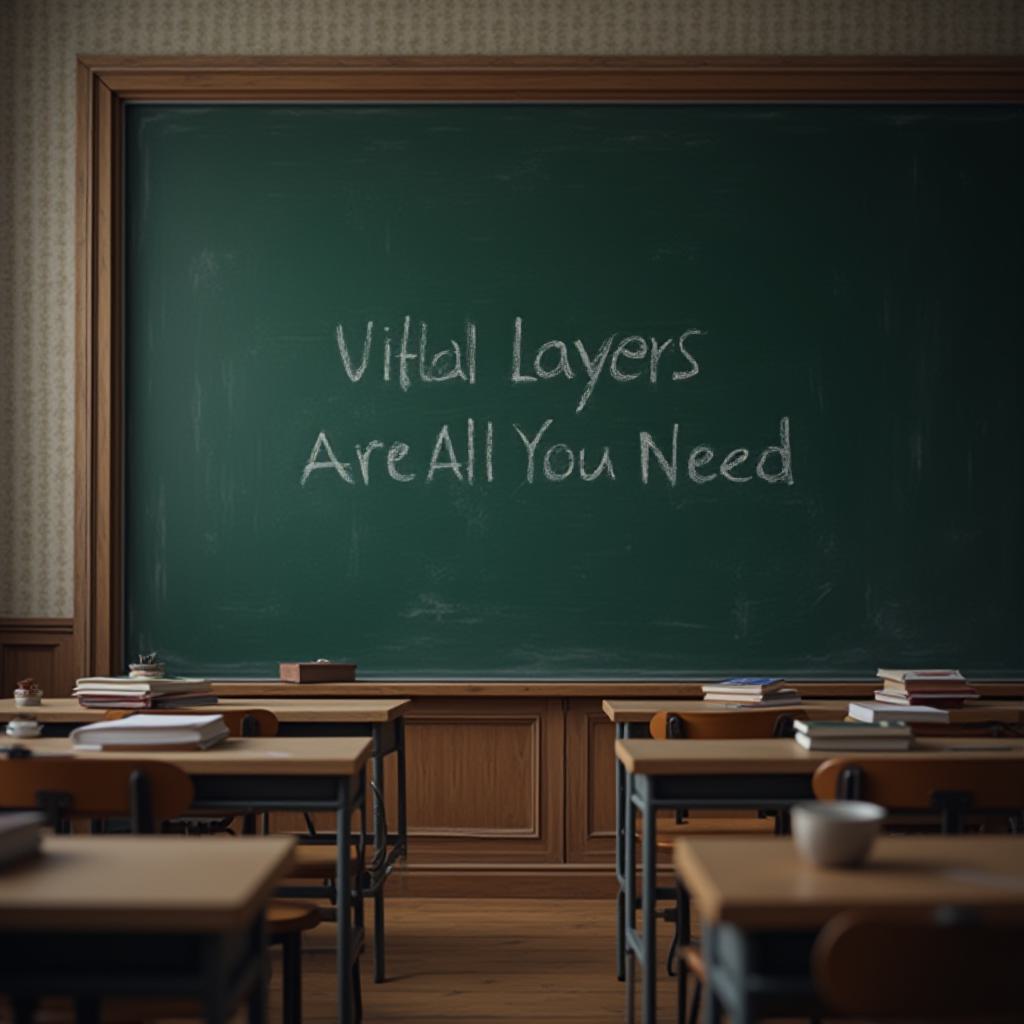} &
        \includegraphics[valign=c, width=\ww]{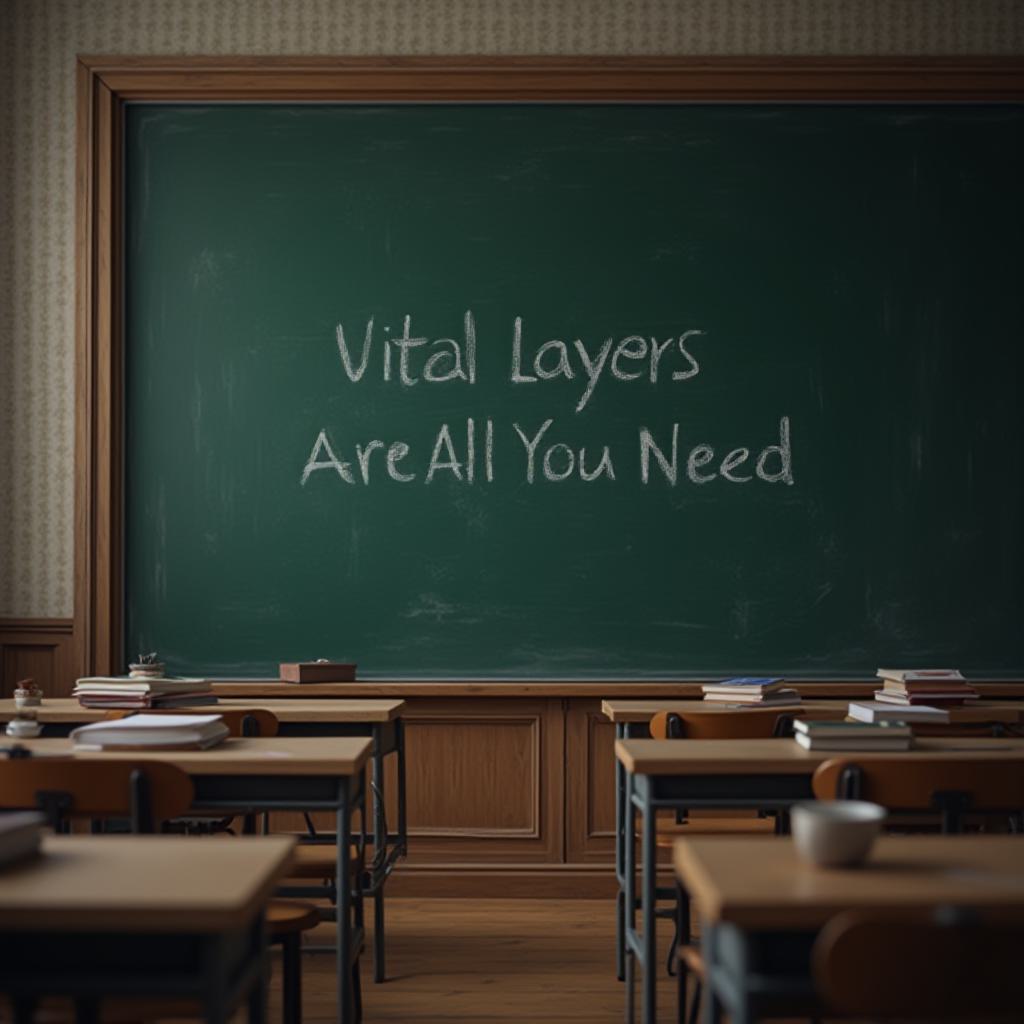}
        \vspace{3px}
        \\

        &
        \multicolumn{3}{c}{\prompt{A blackboard with the text ``Vital Layers Are All You Need'' }}
        \vspace{10px}
        \\

        \includegraphics[valign=c, width=\ww]{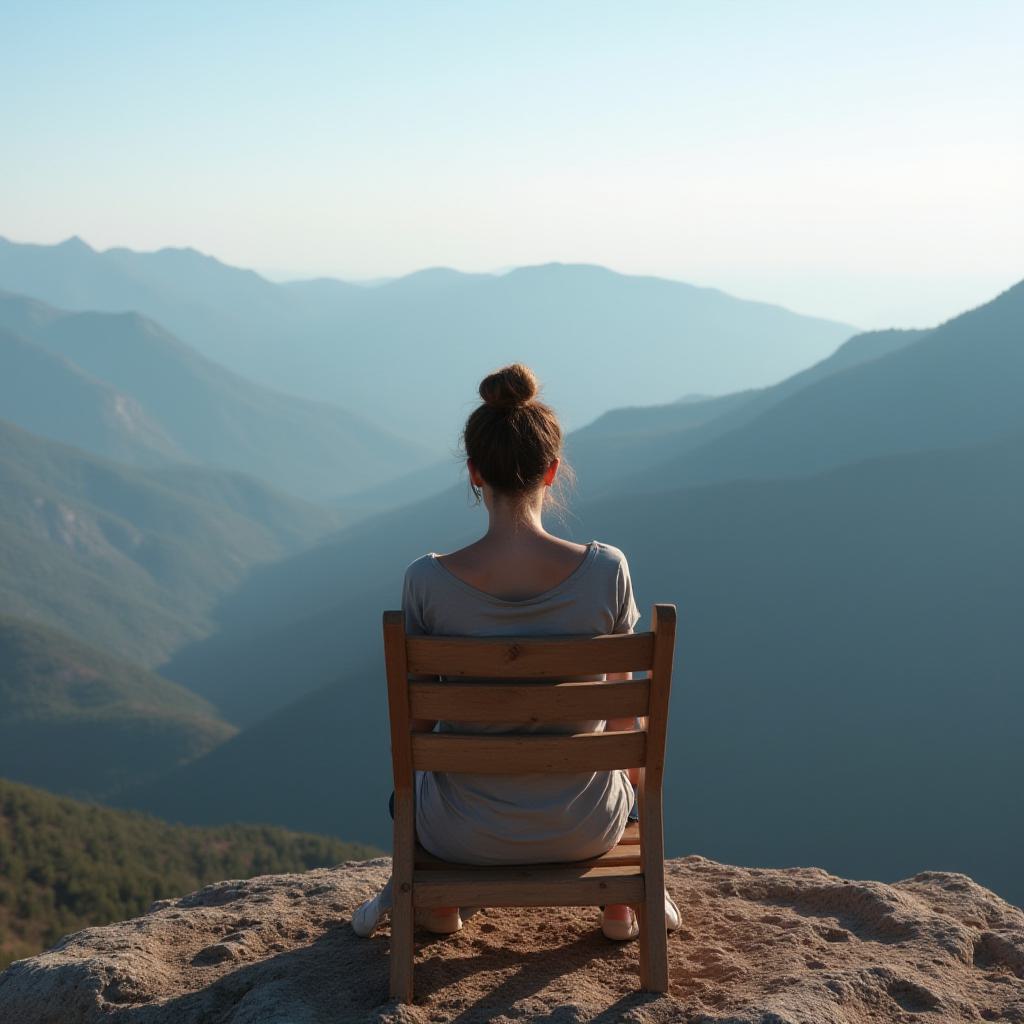} &
        \includegraphics[valign=c, width=\ww]{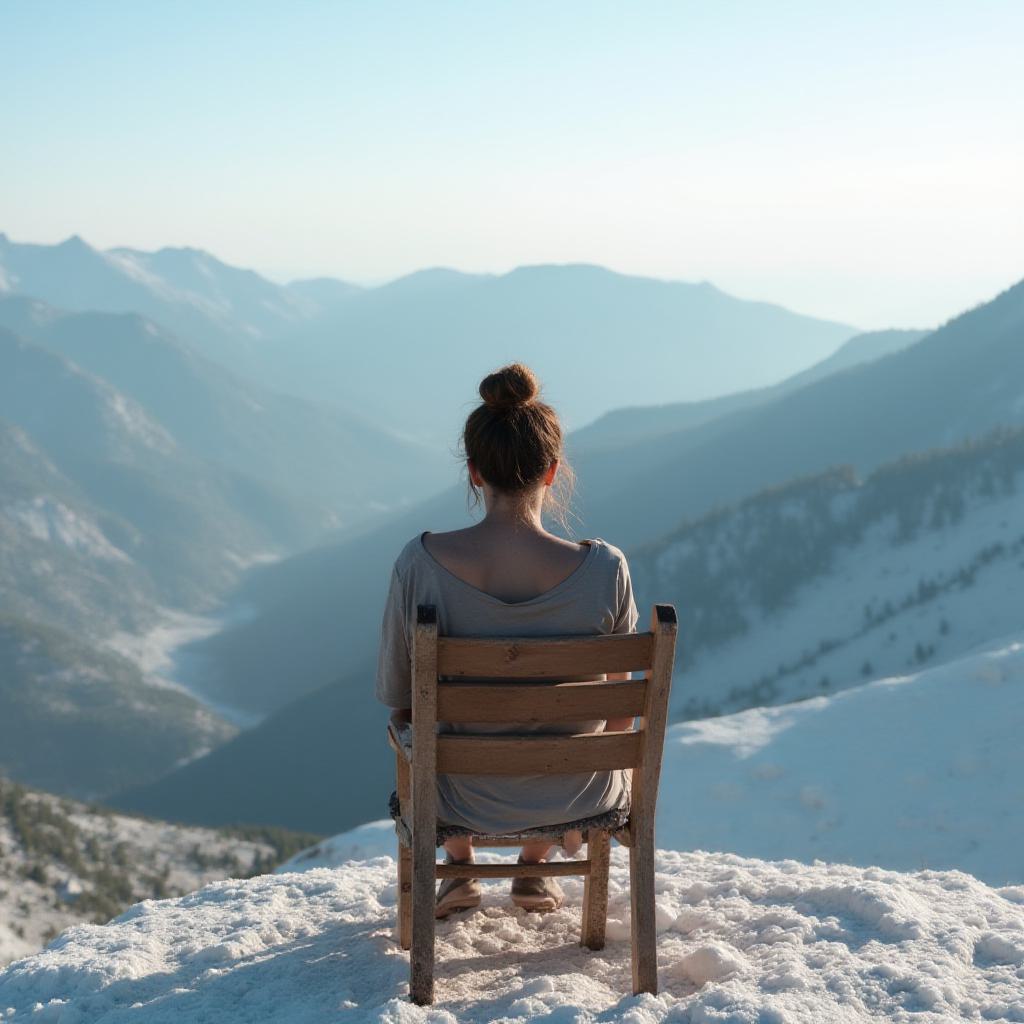} &
        \includegraphics[valign=c, width=\ww]{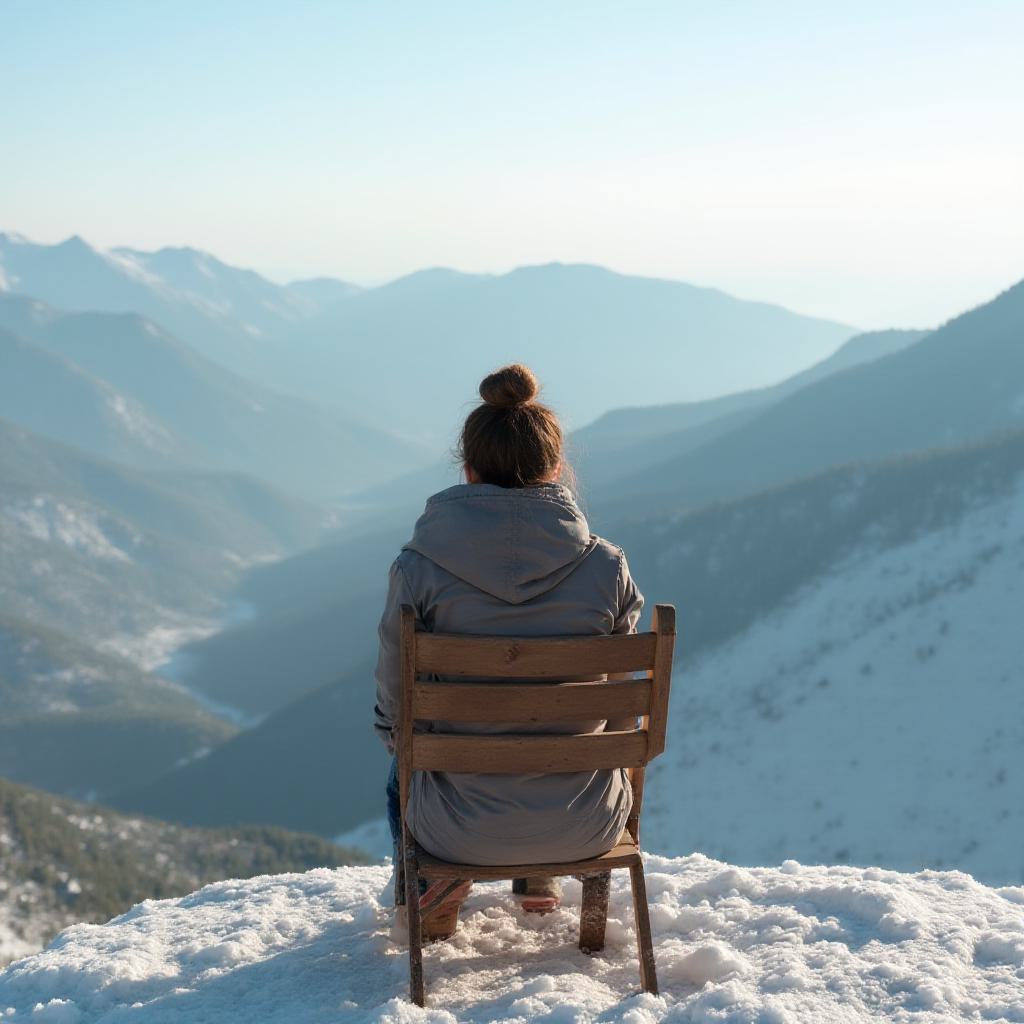} &
        \includegraphics[valign=c, width=\ww]{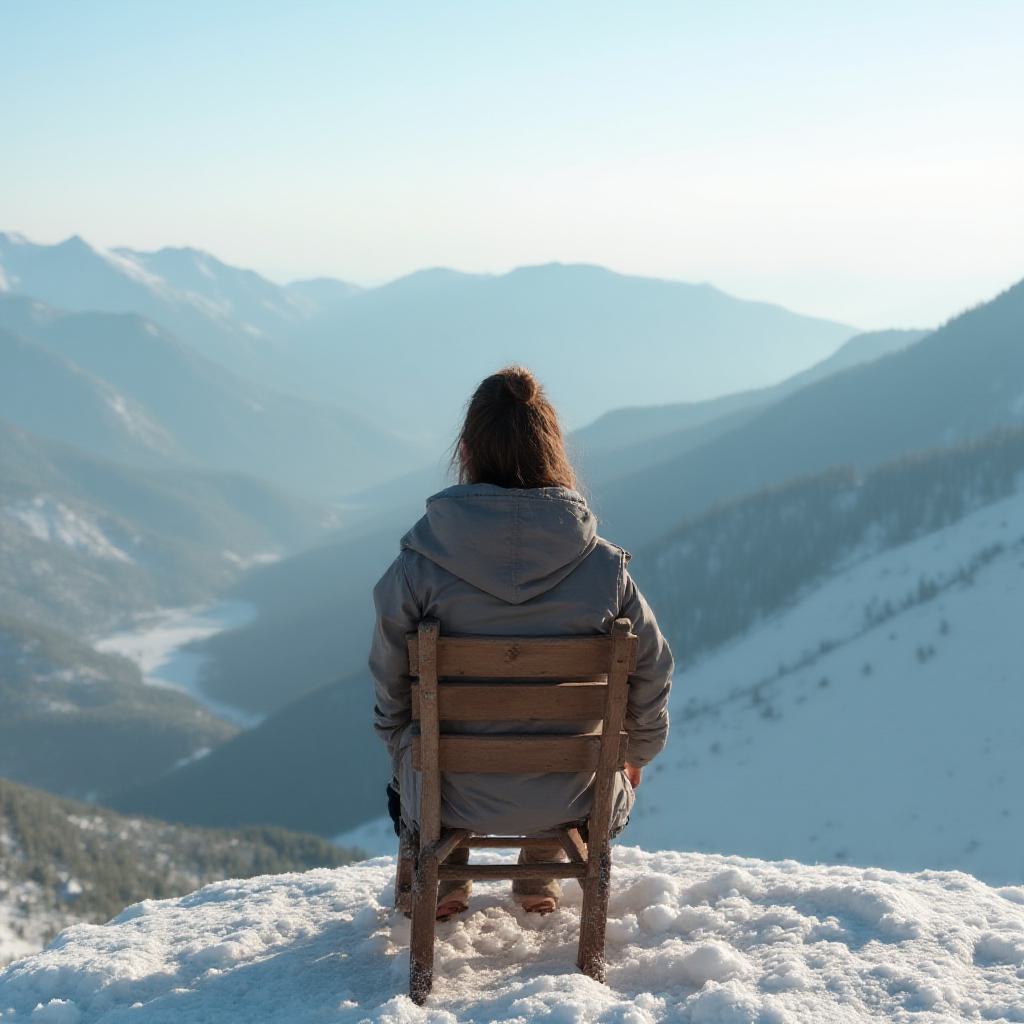}
        \vspace{3px}
        \\

        &
        \multicolumn{3}{c}{\prompt{The floor is covered with snow}}
        \\

    \end{tabular}
    \caption{\textbf{Metrics Qualitative Comparison.} As described in \Cref{sec:different_perceptual_metrics}, we also experimented with other perceptual metrics. We found DINOv2~\cite{Oquab2023DINOv2LR} and DINOv1~\cite{Caron2021EmergingPI} to produce the same set of vital layer. While CLIP~\cite{Radford2021LearningTV} and LPIPS~\cite{Zhang2018TheUE} replaced two layers in the vital layers set (though they include most of the vital layer set as in DINO). As can be seen, the differences between these sets are negligible when editing images.}
    \label{fig:metrics_qualitative_comparison}
\end{figure*}

As explained in
\Cref{sec:layers_importance},
we assess the impact of each layer by measuring the perceptual similarity between $G_{\textit{ref}}$ and $G_{\ell}$ using DINOv2~\cite{Oquab2023DINOv2LR}. It raises the question of the importance of the specific perceptual~\cite{Johnson2016PerceptualLF} similarity metric when determining the vital layers. 

To this end, we also experiment with different perceptual metrics: DINOv1~\cite{Caron2021EmergingPI}, CLIP~\cite{Radford2021LearningTV}, and LPIPS~\cite{Zhang2018TheUE}. In Figures \ref{fig:layer_removal_quantiative_clip}, \ref{fig:layer_removal_quantiative_dino_v1} and \ref{fig:layer_removal_quantiative_lpips} we plot the perceptual similarity per layer for each of these metrics. The vital layers, ordered by vitality, as defined in
\Cref{eqn:vitality},
for each metric are:
\begin{itemize}
    \item DINOv2 --- $[1, 0, 2, 18, 53, 28, 54, 17, 56, 25]$.
    \item DINOv1 --- $[1,  0,  2, 18, 53, 56, 54, 25, 28, 17]$.
    \item CLIP --- $[2, 0, 1, 18, 53, 56, 54, 4, 17, 3]$.
    \item LPIPS --- $[0, 1, 2, 18, 17, 56, 53, 54, 6, 4]$.
\end{itemize}
As can be seen, the vital set $V$ is equivalent for DINOv2 and DINOv1 (even though there is a disagreement on the order). In addition, all the metrics include the set of $\{1, 0, 2, 18, 53, 54, 17, 56\}$ to be included in the vital set, while DINOv1 and DINOv2 suggest also including $\{28, 25\}$, CLIP suggests including $\{3, 4\}$ instead and LPIPS suggests including $\{6, 4\}$ instead. In \Cref{fig:metrics_qualitative_comparison} we edited images with these slightly different vital layer sets, and found the differences to be negligible in practice.

\subsection{VLM-Based Quantiative Metric}
\label{sec:vlm_based_metric}

\begin{table}
    \centering
    \caption{\textbf{VLM-Based quantitative comparison.} For each method, we used Phi-3.5-vision~\cite{Abdin2024Phi3TR} VLM to compute the percentage of editing results that follow the text prompt and of the results that change only the essential parts of the image. P2P+NTI~\cite{Hertz2022PrompttoPromptIE, mokady2022null}, Instruct-P2P~\cite{brooks2022instructpix2pix}, and MasaCTRL~\cite{cao2023masactrl} suffer from low similarity to the text prompt. SDEdit~\cite{Yang2022PaintBE} and MagicBrush~\cite{Zhang2023MagicBrush} adhere more to the text prompt, but they struggle with avoiding unintended changes.}
    \vspace{-5px}
    \begin{adjustbox}{width=1\columnwidth}
        \begin{tabular}{>{\columncolor[gray]{0.95}}lcc}
            \toprule

            \textbf{Method} & 
            Text Following $(\uparrow)$ &
            Modify only essential $(\uparrow)$
            \\
            
            \midrule

            SDEdit [88] &
            86.66\% &
            21.66\%
            \\

            P2P+NTI [33, 51] &
            68.33\% &
            26.66\%
            \\

            Instruct-P2P [17] &
            33.33\% &
            26.66\%
            \\

            MagicBrush [91] &
            \textbf{88.33\%} &
            46.66\%
            \\

            MasaCTRL [18] &
            33.33\% &
            06.66\%
            \\

            \midrule

            Stable Flow (ours) &
            83.33\% &
            \textbf{61.66\%}
            \\
            
            \bottomrule
        \end{tabular}
    \end{adjustbox}
    \label{tab:vlm_quantitative_comparison}
\end{table}

As explained in
\Cref{sec:comparisons},
We evaluated the editing results using three widely-used CLIP-based metrics: \clipimg, \cliptxt, and \clipdir. In addition, we experimented with a VLM-based metric using the Phi-3.5-vision~\cite{Abdin2024Phi3TR} VLM that was trained specifically for the task of multiple image comparison. For each input image $x$, editing prompt $p$, and editing result $\hat{x}$, we computed the following two metrics: (1) \emph{Text Following} --- we presented the VLM the edited image $\hat{x}$ and the editing prompt $p$ and asked it ``Does this image correspond to the text $p$? Answer yes or no.''. (2) \emph{Modify only essential parts} --- we extracted the prompt instruction and presented it, along with $x$ and $\hat{x}$, to the VLM and asked it ``Is the only difference between these two images the text PROMPT? Answer yes or no''. For each metric, we calculated the number of times that the VLM answered ``yes''. As demonstrated in \Cref{tab:vlm_quantitative_comparison}, the VLM-based metric follows the same trend as the CLIP-based metrics:P2P+NTI~\cite{Hertz2022PrompttoPromptIE, mokady2022null}, Instruct-P2P~\cite{brooks2022instructpix2pix}, and MasaCTRL~\cite{cao2023masactrl} suffer from low similarity to the text prompt. SDEdit~\cite{Yang2022PaintBE} and MagicBrush~\cite{Zhang2023MagicBrush} adhere more to the text prompt, but they struggle with avoiding unintended changes.

\subsection{Number of Vital Layers}
\label{sec:number_of_vital_layers}

\begin{figure*}[tp]
    \centering
    \setlength{\tabcolsep}{0.6pt}
    \renewcommand{\arraystretch}{0.8}
    \setlength{\ww}{0.16\linewidth}
    \begin{tabular}{c @{\hspace{10\tabcolsep}} ccccc}

        Input &
        $100\%$ &
        $80\%$ &
        $60\%$ &
        $40\%$ &
        $20\%$
        \vspace{2px}
        \\

        \includegraphics[valign=c, width=\ww]{figures/metrics_qualitative_comparison/assets/man/inp.jpg} &
        \includegraphics[valign=c, width=\ww]{figures/metrics_qualitative_comparison/assets/man/dino.jpg} &
        \includegraphics[valign=c, width=\ww]{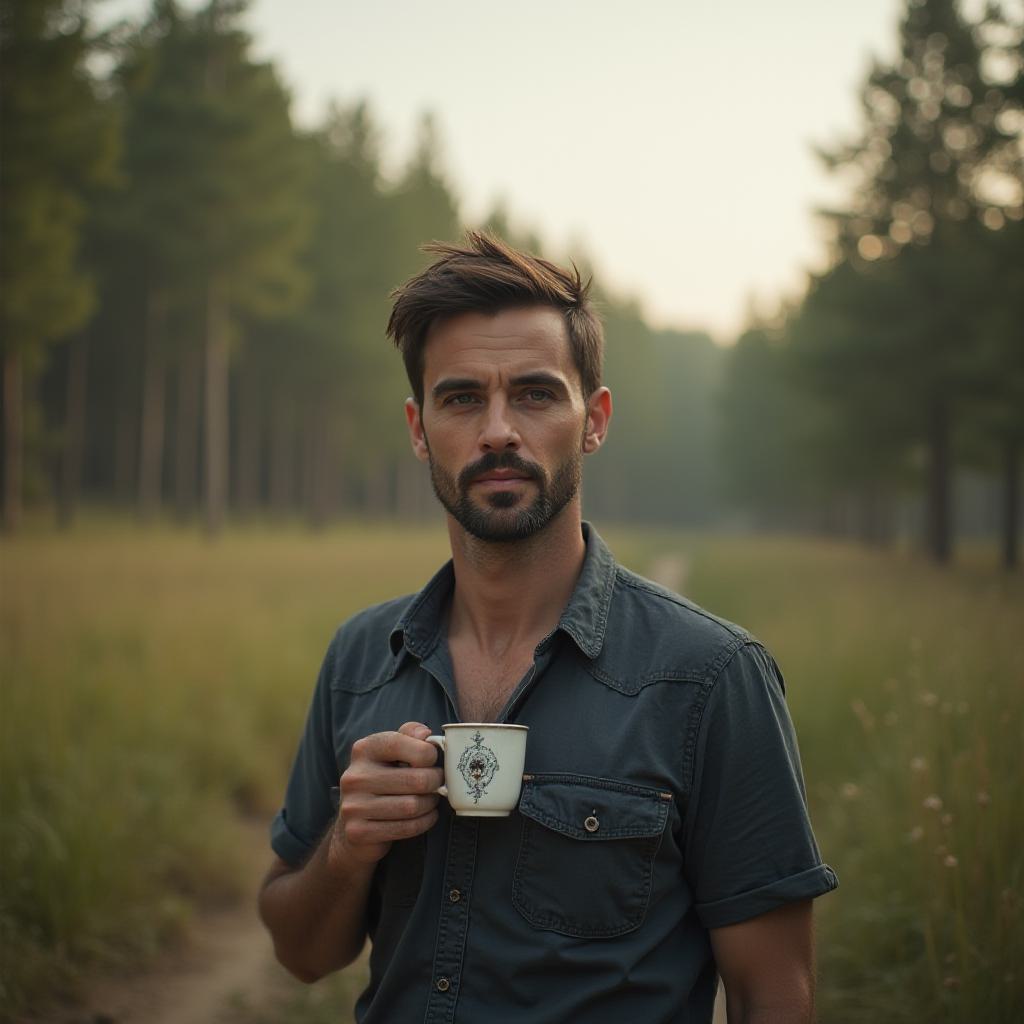} &
        \includegraphics[valign=c, width=\ww]{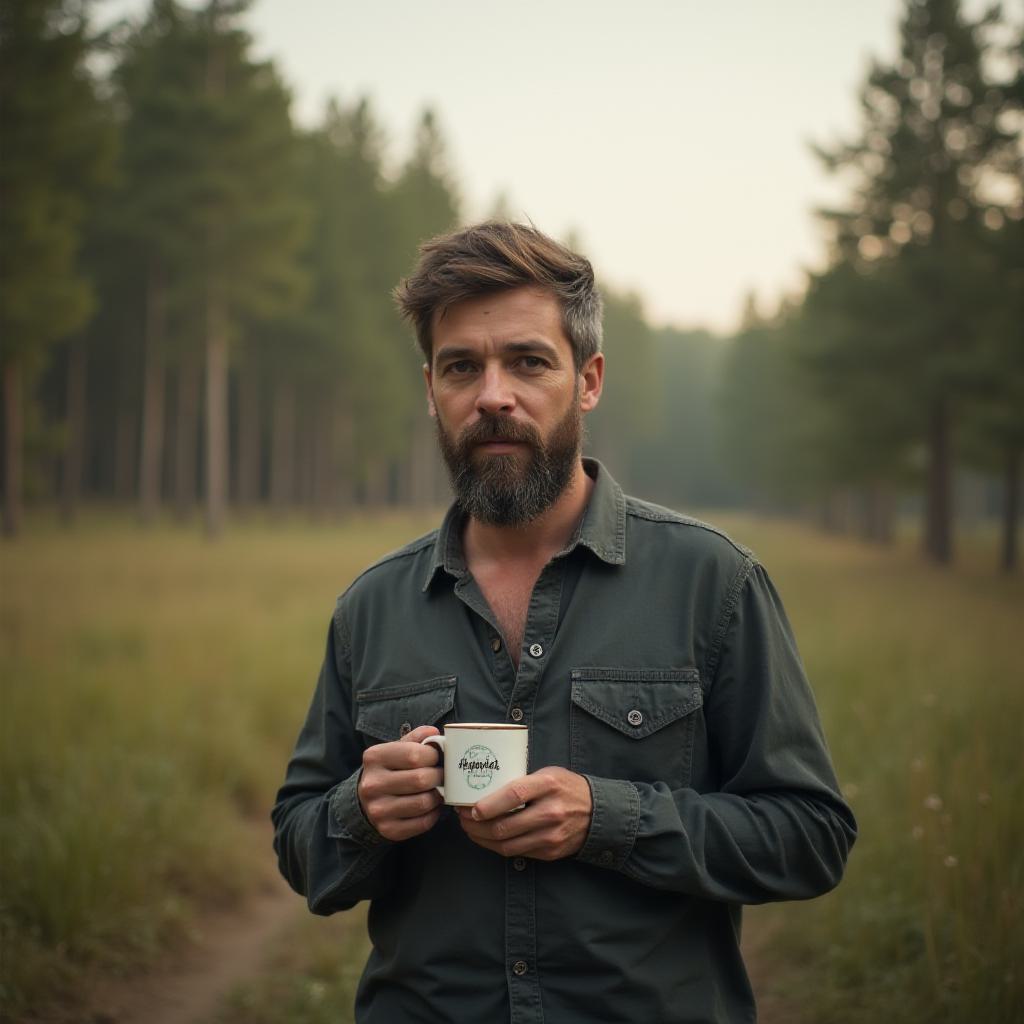} &
        \includegraphics[valign=c, width=\ww]{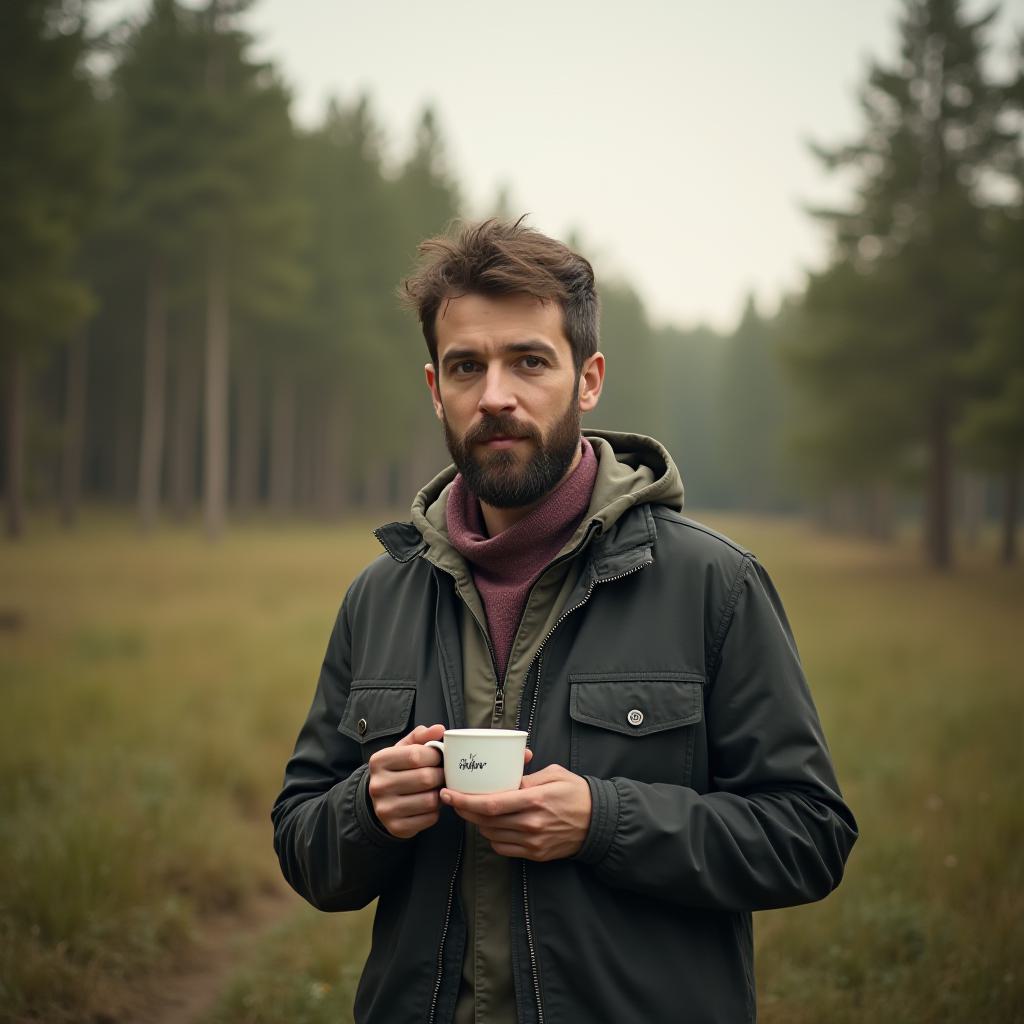} &
        \includegraphics[valign=c, width=\ww]{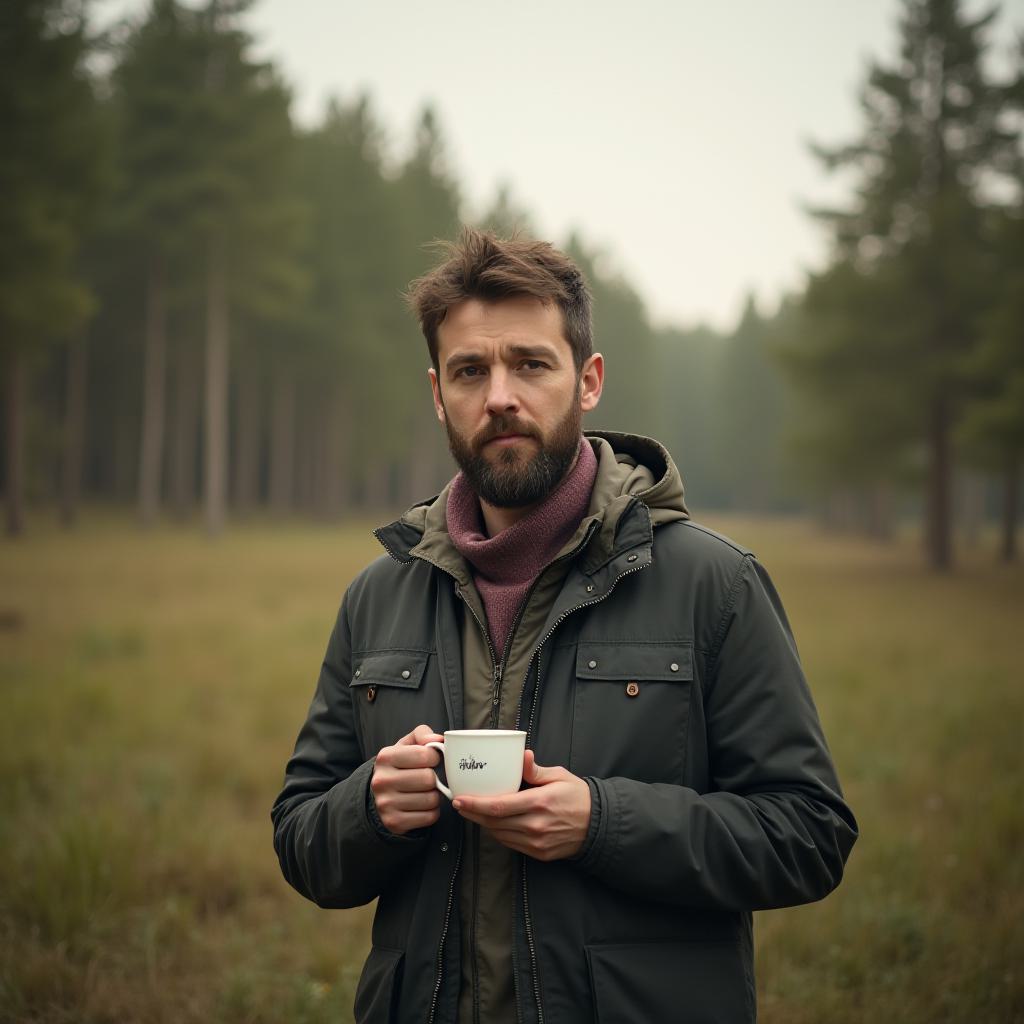}
        \vspace{3px}
        \\

        &
        \multicolumn{5}{c}{\prompt{A man holding a cup of tea}}
        \vspace{10px}
        \\

        \includegraphics[valign=c, width=\ww]{figures/metrics_qualitative_comparison/assets/cat/inp.jpg} &
        \includegraphics[valign=c, width=\ww]{figures/metrics_qualitative_comparison/assets/cat/dino.jpg} &
        \includegraphics[valign=c, width=\ww]{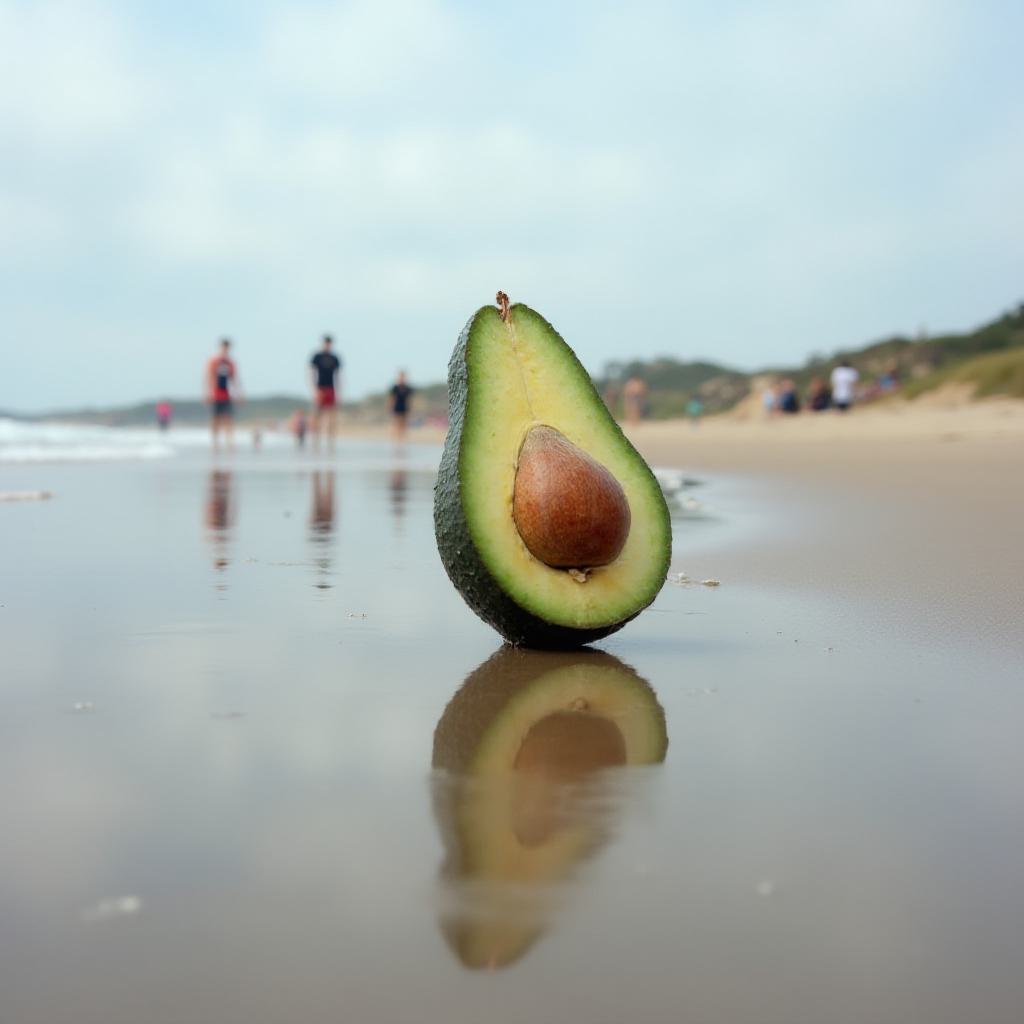} &
        \includegraphics[valign=c, width=\ww]{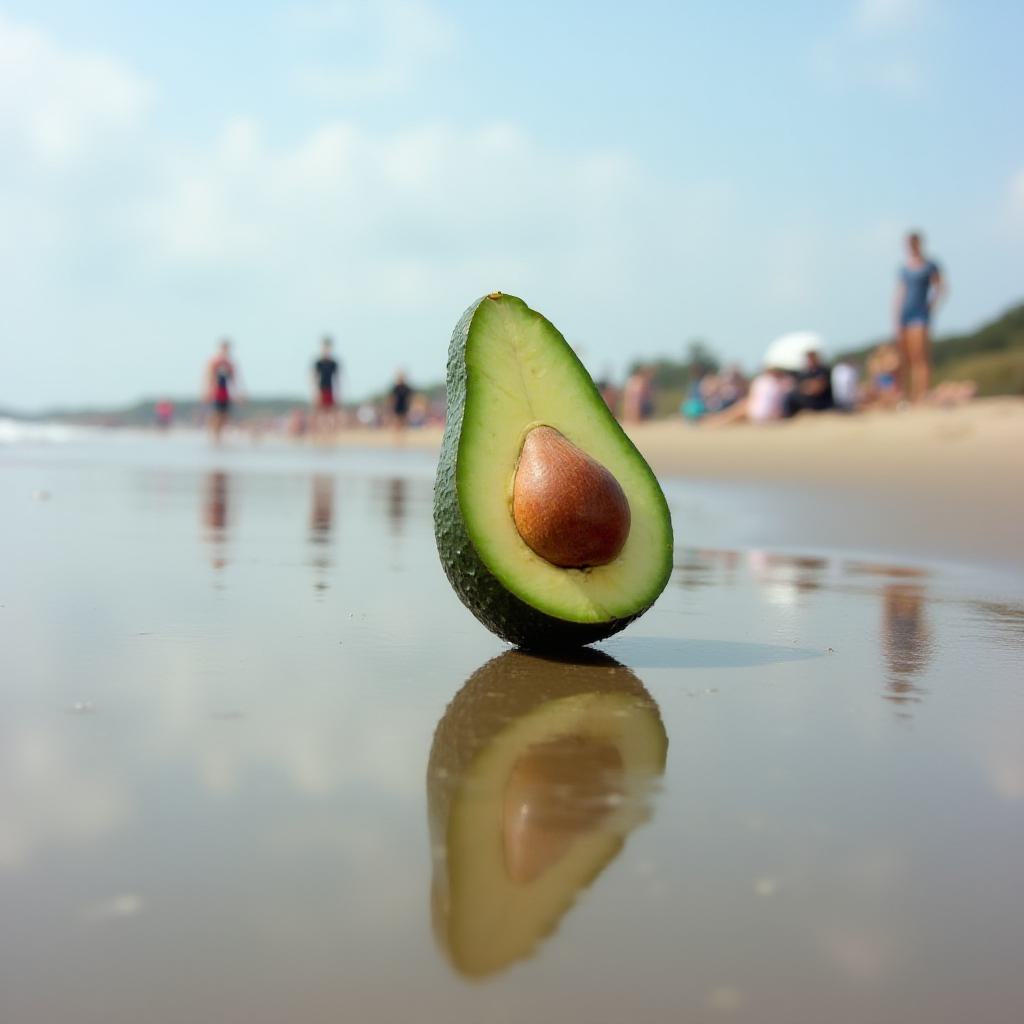} &
        \includegraphics[valign=c, width=\ww]{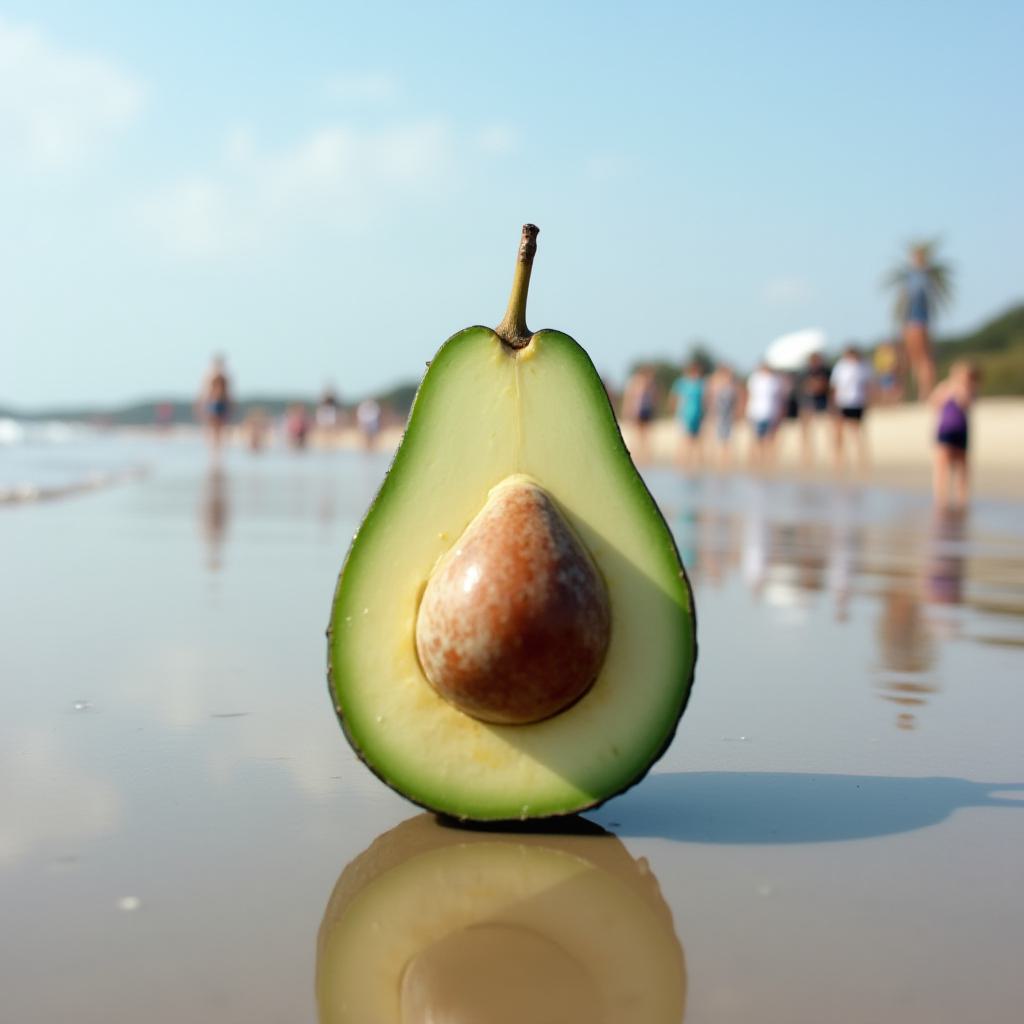} &
        \includegraphics[valign=c, width=\ww]{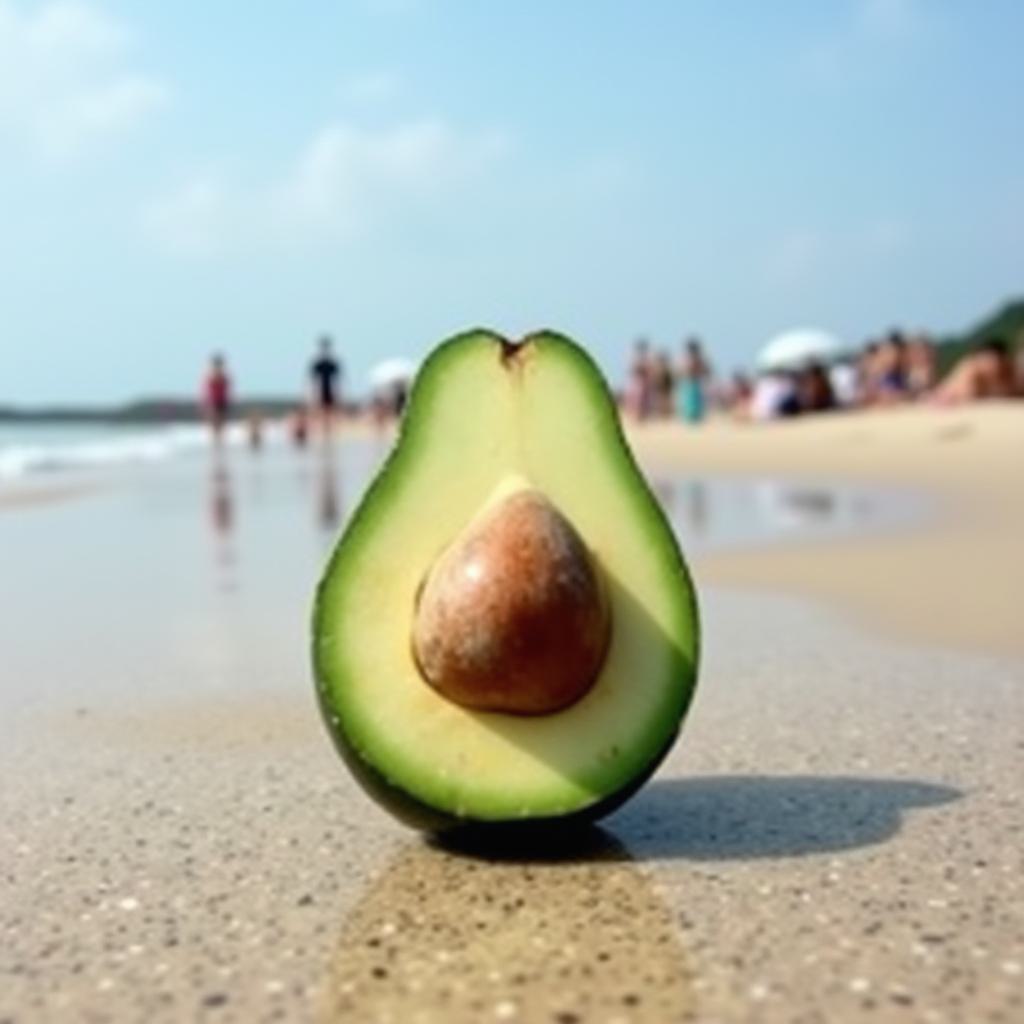}
        \vspace{3px}
        \\

        &
        \multicolumn{5}{c}{\prompt{An avocado at the beach}}
        \vspace{10px}
        \\

        \includegraphics[valign=c, width=\ww]{figures/metrics_qualitative_comparison/assets/blackboard/inp.jpg} &
        \includegraphics[valign=c, width=\ww]{figures/metrics_qualitative_comparison/assets/blackboard/dino.jpg} &
        \includegraphics[valign=c, width=\ww]{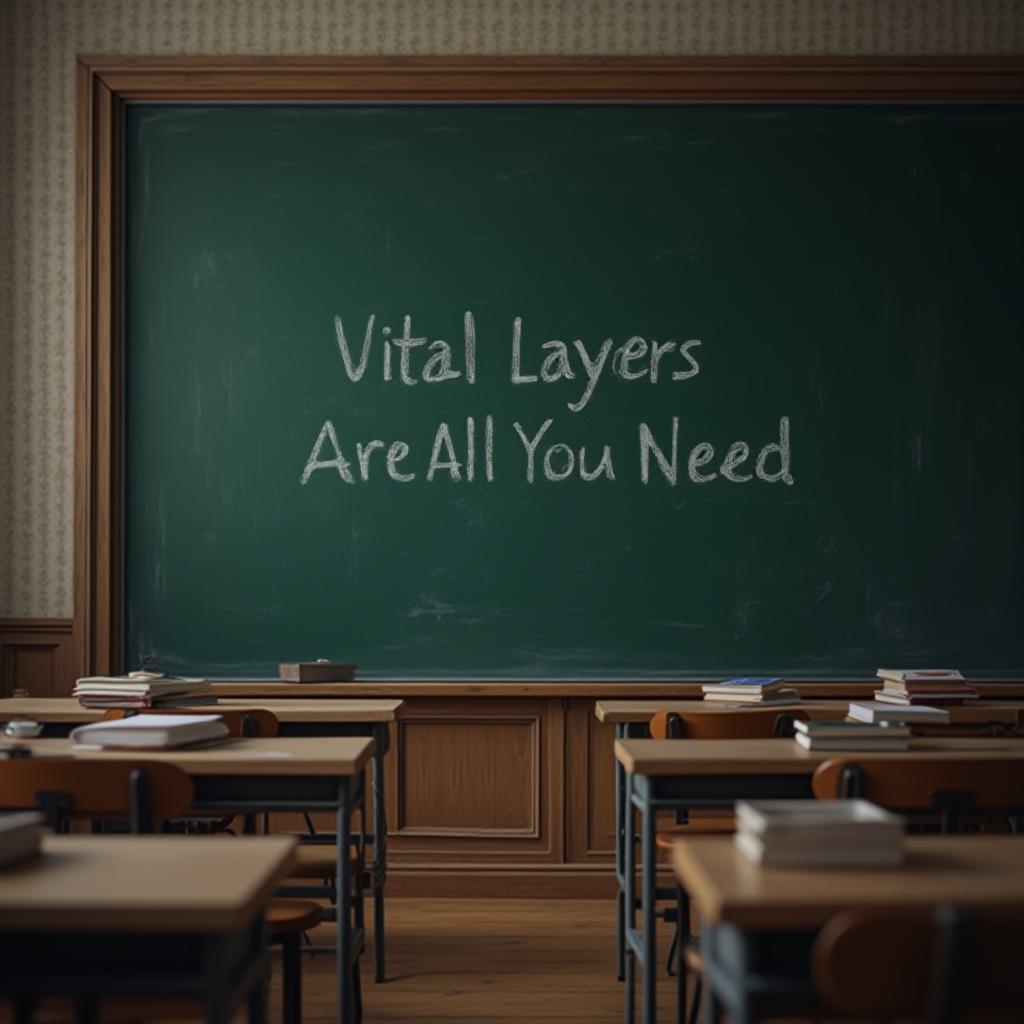} &
        \includegraphics[valign=c, width=\ww]{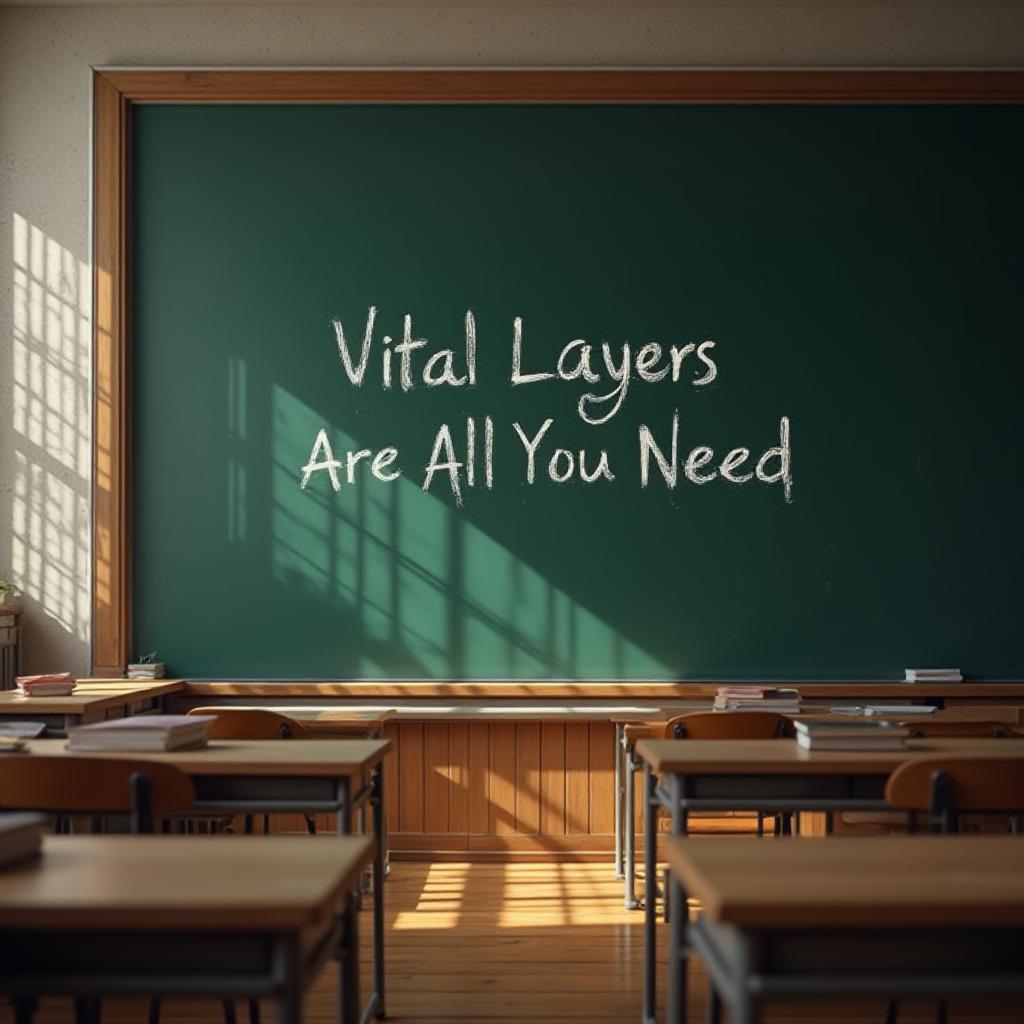} &
        \includegraphics[valign=c, width=\ww]{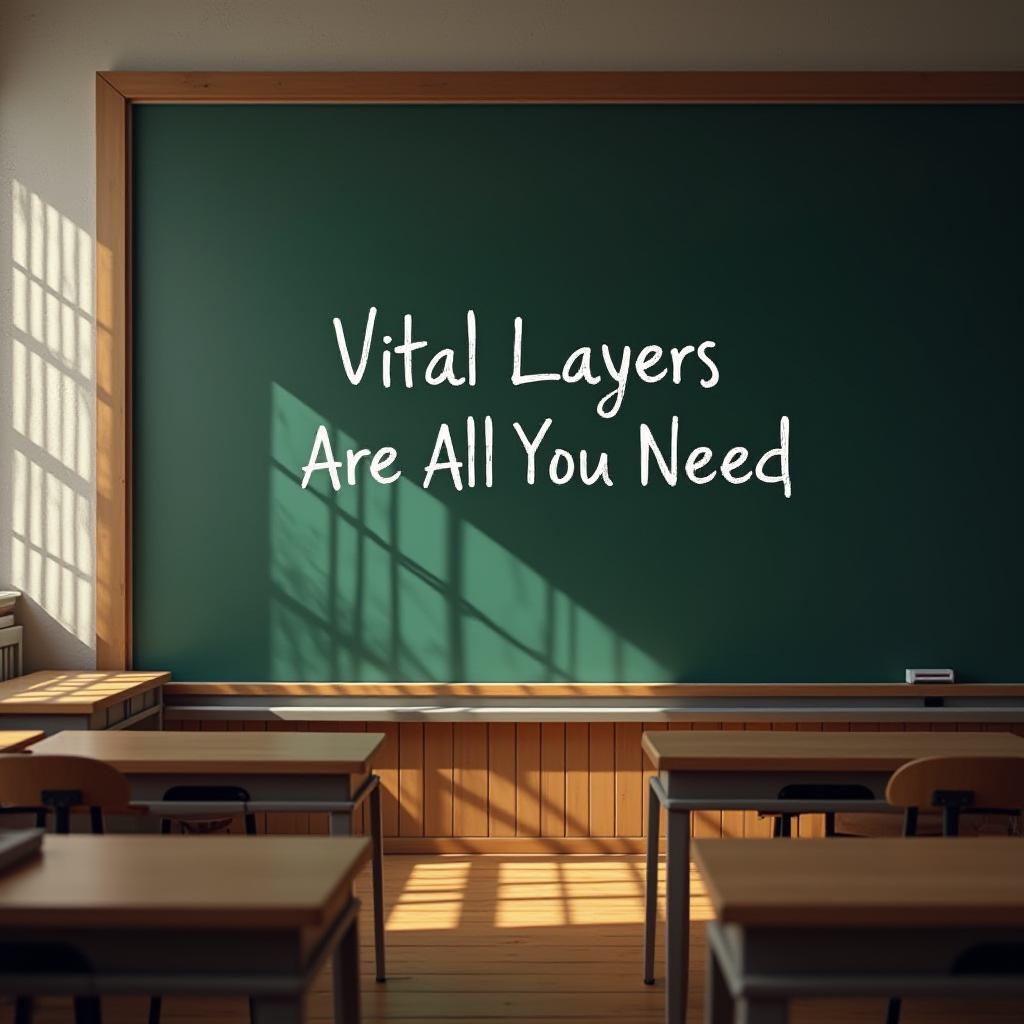} &
        \includegraphics[valign=c, width=\ww]{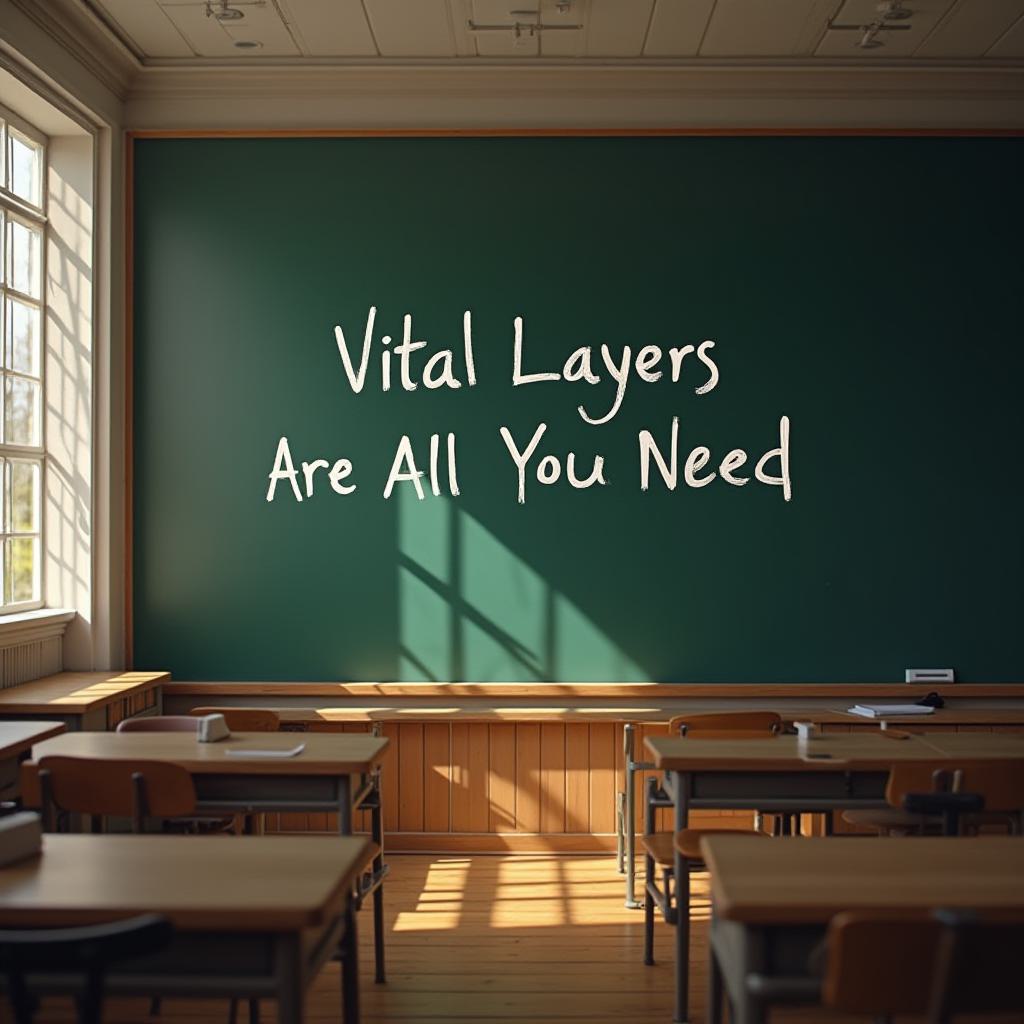}
        \vspace{3px}
        \\

        &
        \multicolumn{5}{c}{\prompt{A blackboard with the text ``Vital Layers Are All You Need'' }}
        \vspace{10px}
        \\

        \includegraphics[valign=c, width=\ww]{figures/metrics_qualitative_comparison/assets/woman/inp.jpg} &
        \includegraphics[valign=c, width=\ww]{figures/metrics_qualitative_comparison/assets/woman/dino.jpg} &
        \includegraphics[valign=c, width=\ww]{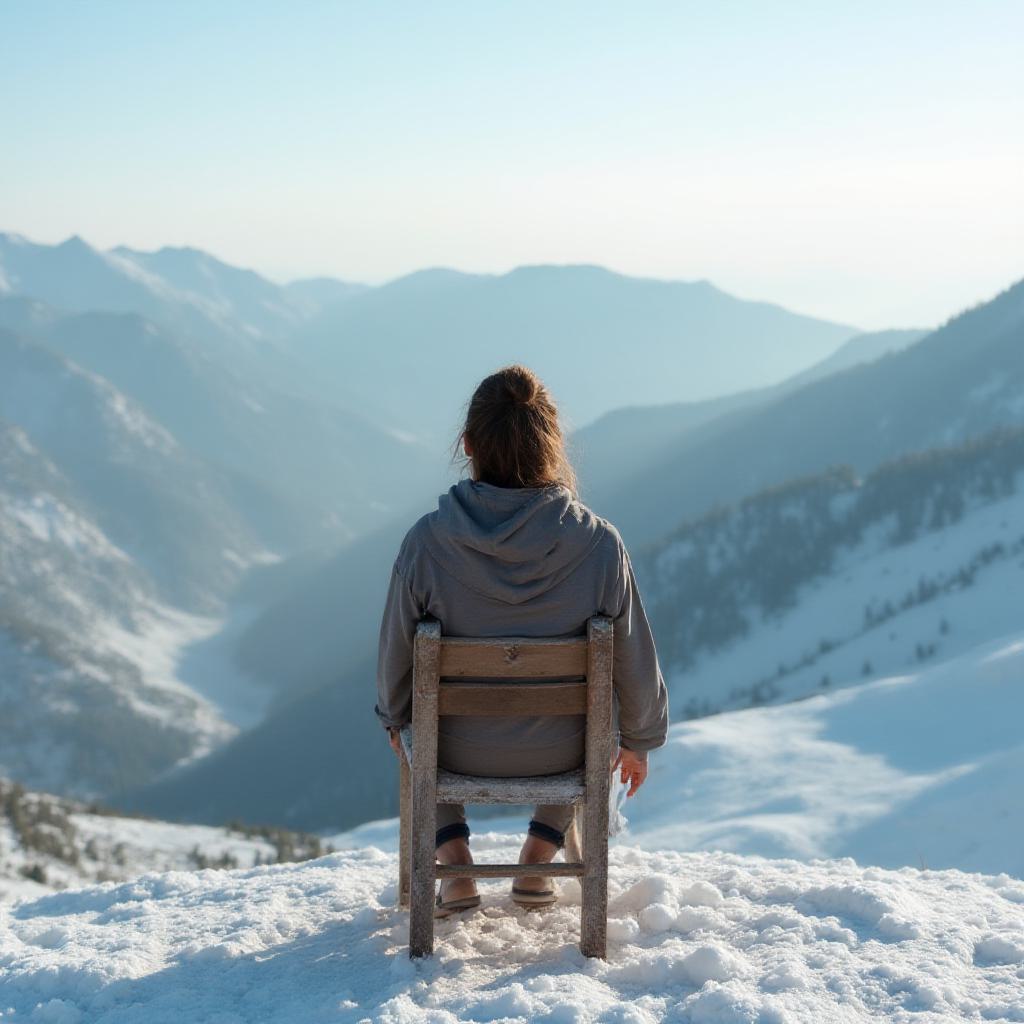} &
        \includegraphics[valign=c, width=\ww]{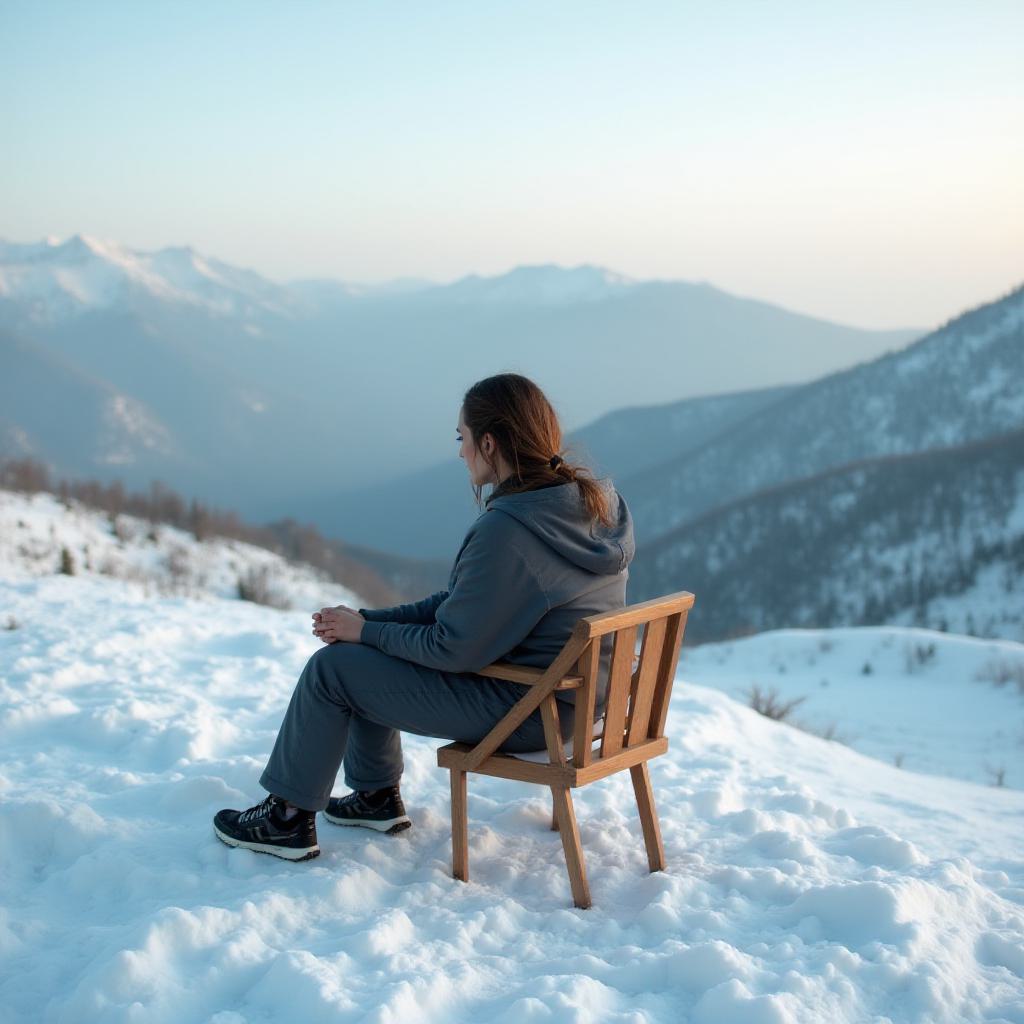} &
        \includegraphics[valign=c, width=\ww]{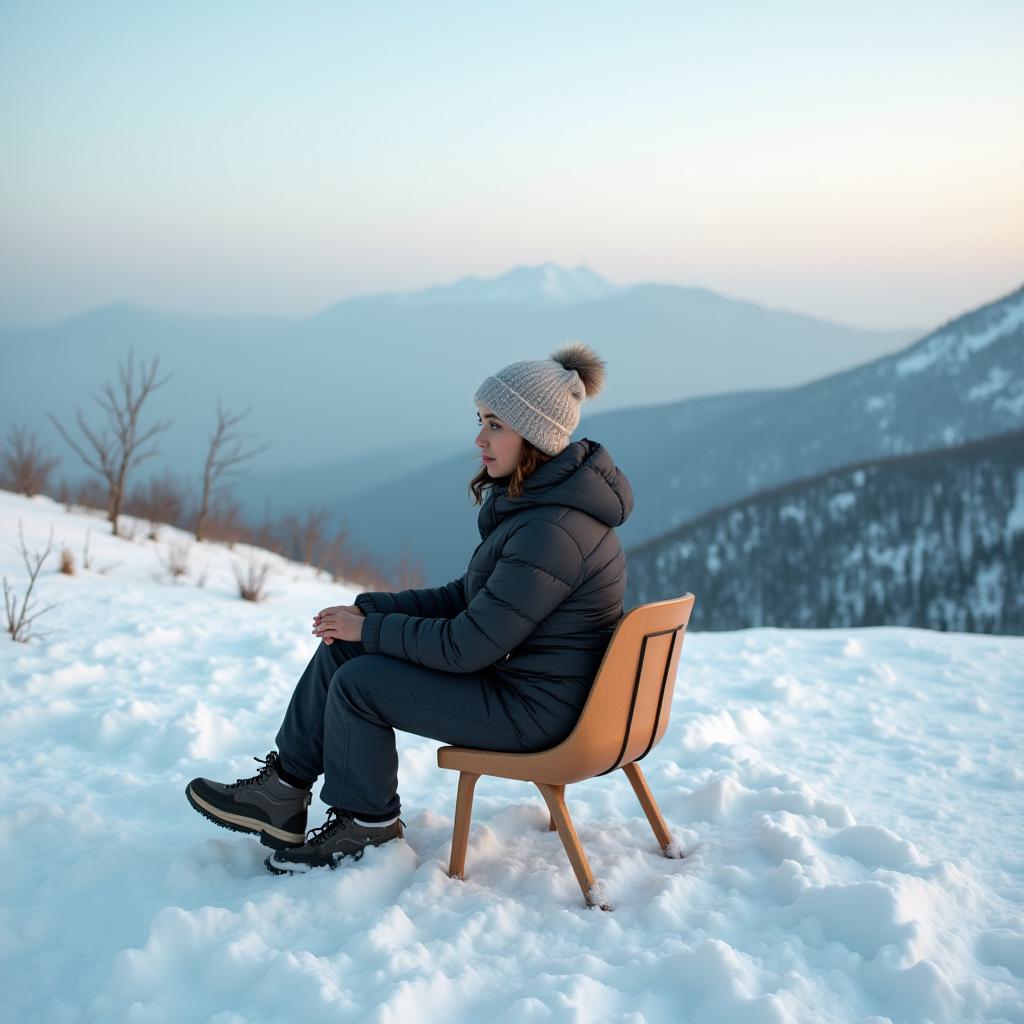} &
        \includegraphics[valign=c, width=\ww]{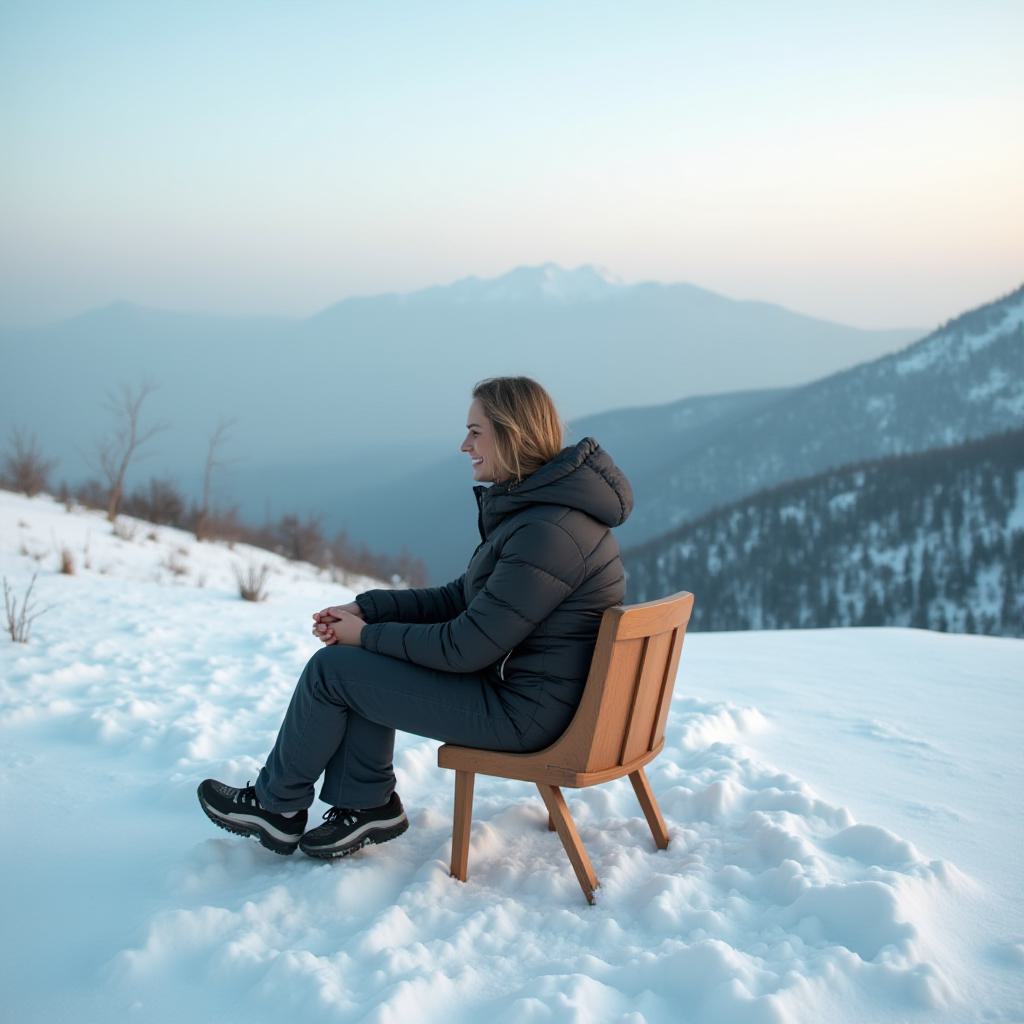}
        \vspace{3px}
        \\

        &
        \multicolumn{5}{c}{\prompt{The floor is covered with snow}}
        \\

    \end{tabular}
    \caption{\textbf{Number of Vital Layers Comparison.} As explained in \Cref{sec:number_of_vital_layers}, we experimented with choosing a different portion of the calculated vital layer set $V$. As can be seen, when removing $20\%$ of the vital layer set, the changes are negligible. However, when removing more than that, the editing results include unintended changes, such as identity changes (\eg, the man and woman examples) and background changes (\eg, the cat and blackboard examples).}
    \label{fig:number_of_vital_layers}
\end{figure*}

The somewhat agnostic nature of our method to the specific perceptual metric, as described in \Cref{sec:different_perceptual_metrics}, raises the question of the importance of the entire vital layer set $V$ to the editing task. To this end, in \Cref{fig:number_of_vital_layers} we experimented with omitting a growing number of vital layers and testing the editing results. As can be seen, when removing $20\%$ of the vital layer set, the changes are negligible. However, when removing more than that, the editing results include unintended changes, such as identity changes (\eg, man and woman examples) and background changes (\eg, cat and blackboard examples). This is consistent with the results from \Cref{sec:different_perceptual_metrics} that show that the least vital layers for each perceptual metric are less important for the image editing task.

\subsection{Latent Nudging Experiments}
\label{sec:latent_nudging_experiment}

\begin{figure*}[tp]
    \centering
    \setlength{\tabcolsep}{0.6pt}
    \renewcommand{\arraystretch}{0.8}
    \setlength{\ww}{0.19\linewidth}
    \begin{tabular}{c @{\hspace{10\tabcolsep}} cccc}
        \raisebox{-1.3\height}[0pt][0pt]{\includegraphics[valign=c, width=\ww]{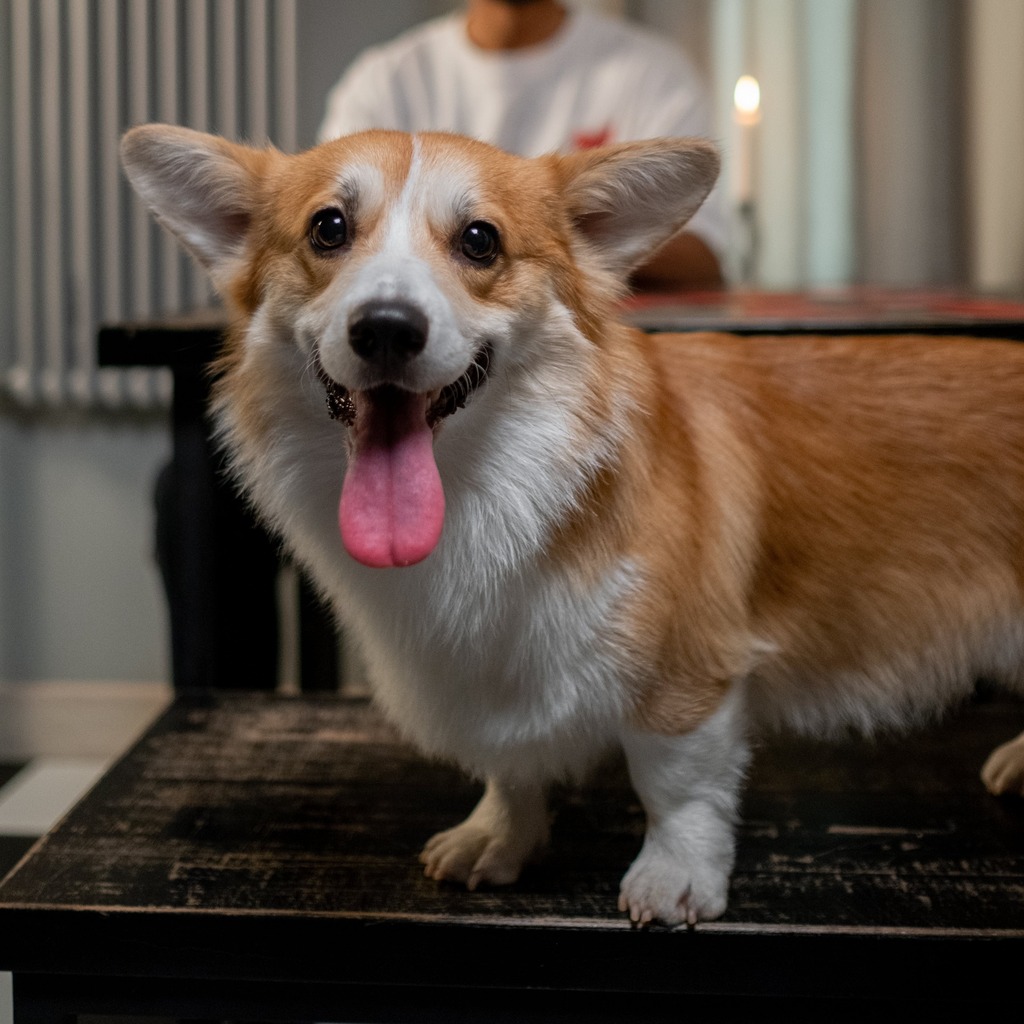}} &
        \includegraphics[valign=c, width=\ww]{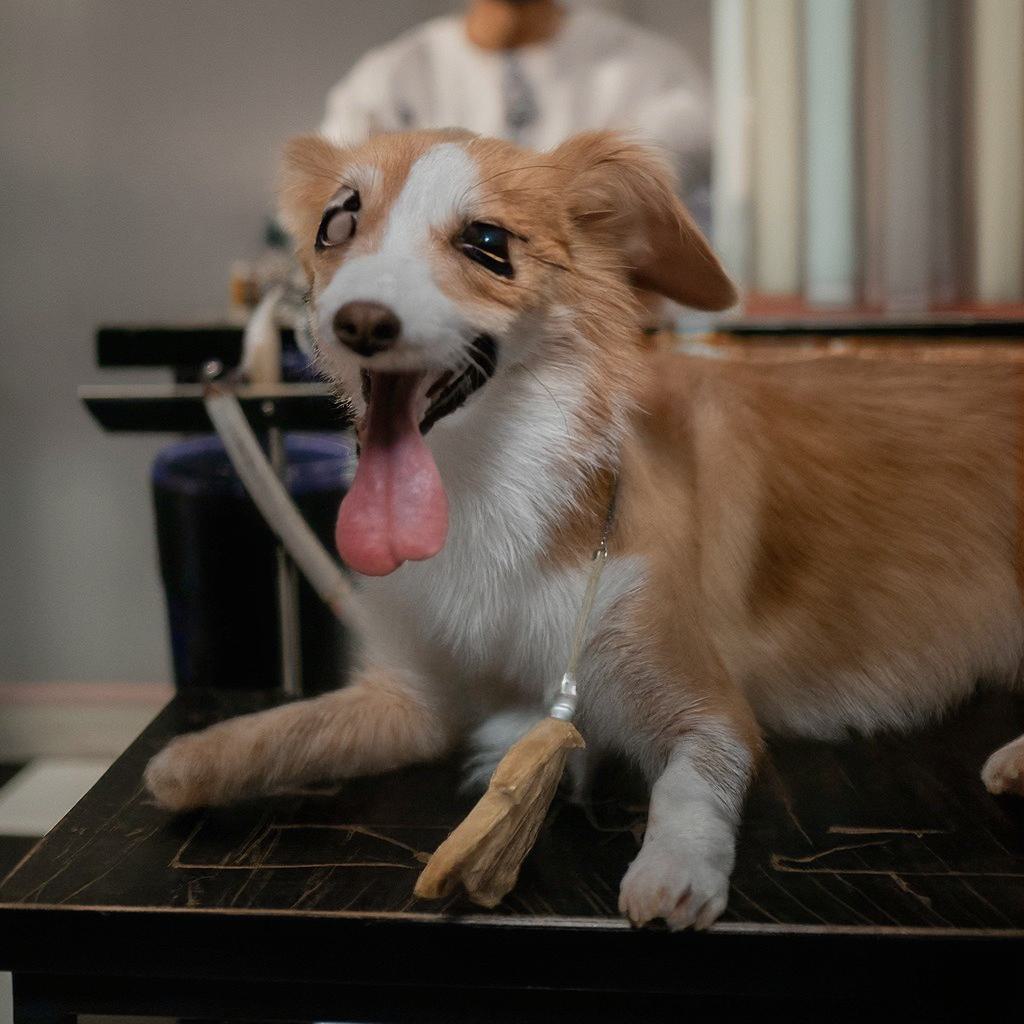} &
        \includegraphics[valign=c, width=\ww]{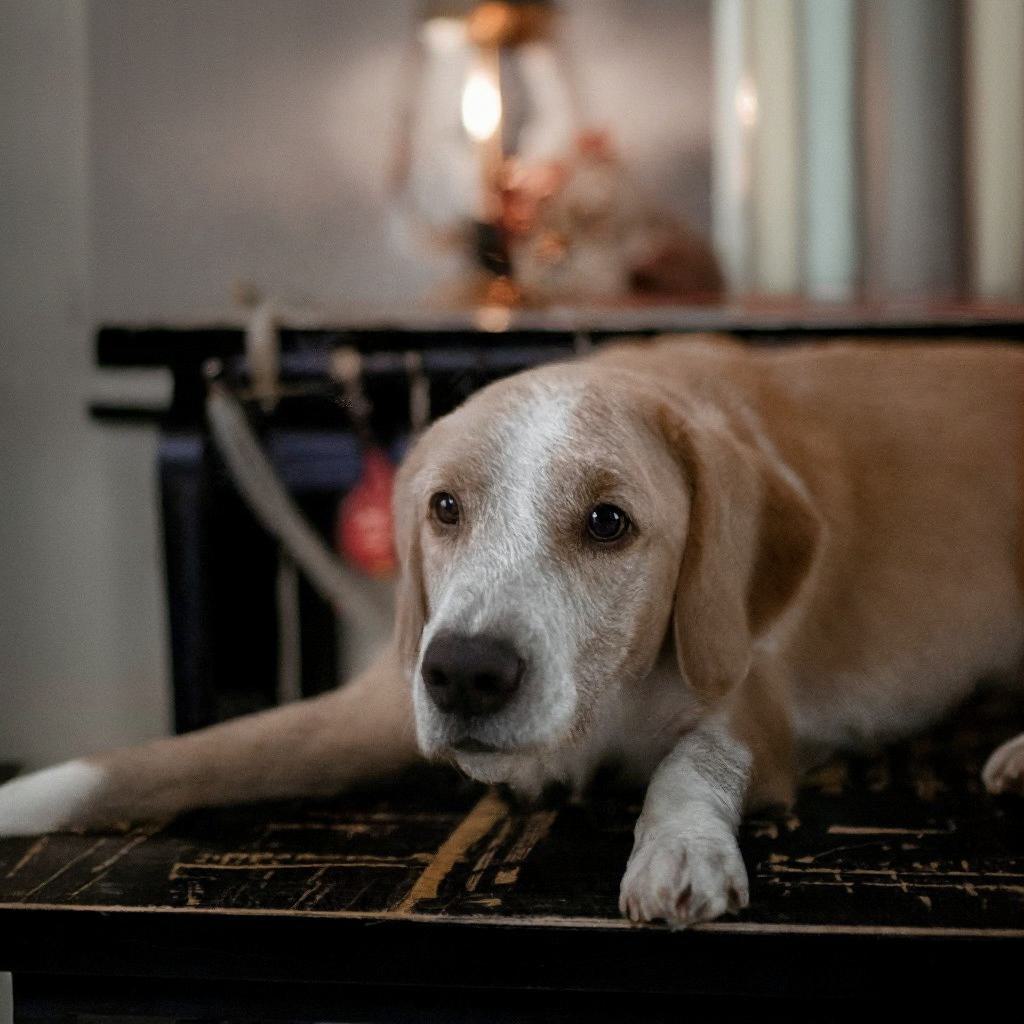} &
        \includegraphics[valign=c, width=\ww]{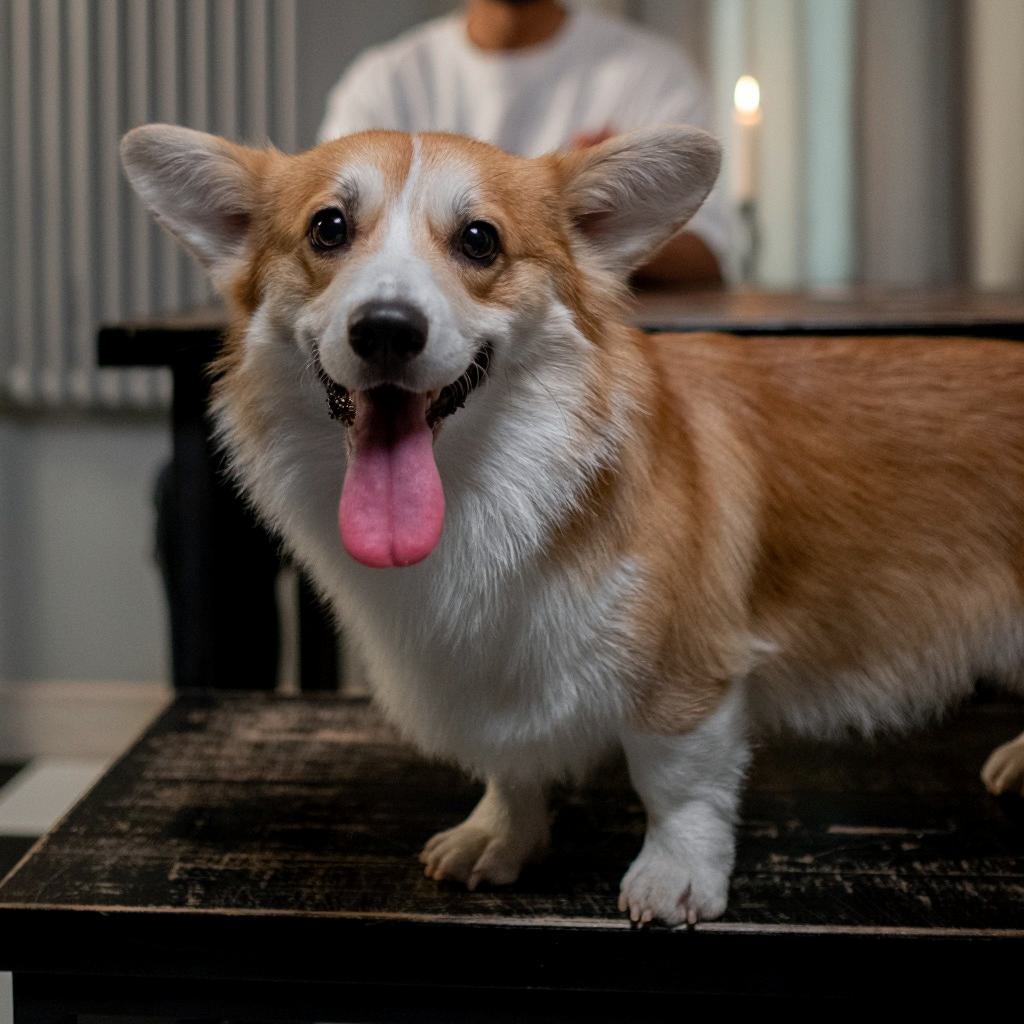} &
        \includegraphics[valign=c, width=\ww]{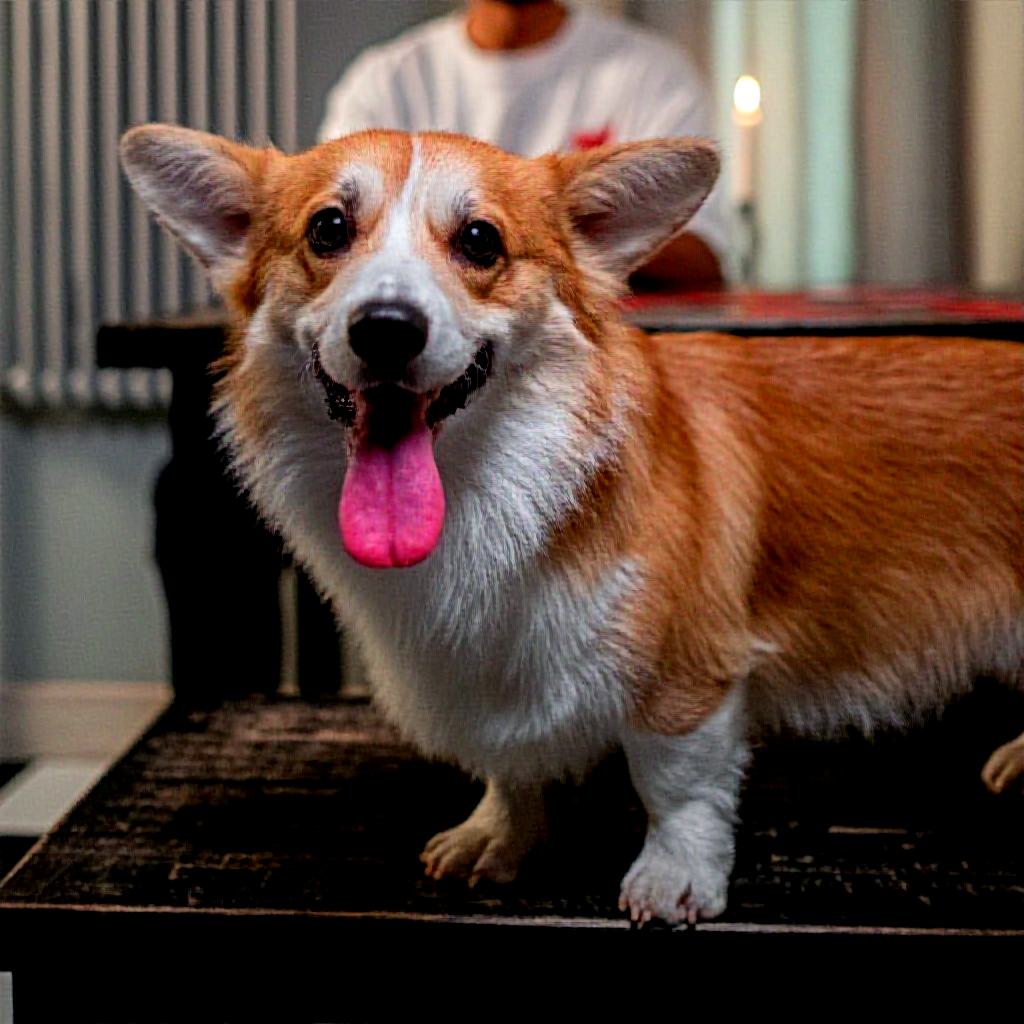}
        \vspace{3px}
        \\

        \raisebox{-9.3\height}[0pt][0pt]{Input}&
        $\lambda=1.0$ &
        $\lambda=1.1$ &
        \highlight{$\lambda=1.15$} &
        $\lambda=3.0$
        \vspace{10px}
        \\

        &
        \includegraphics[valign=c, width=\ww]{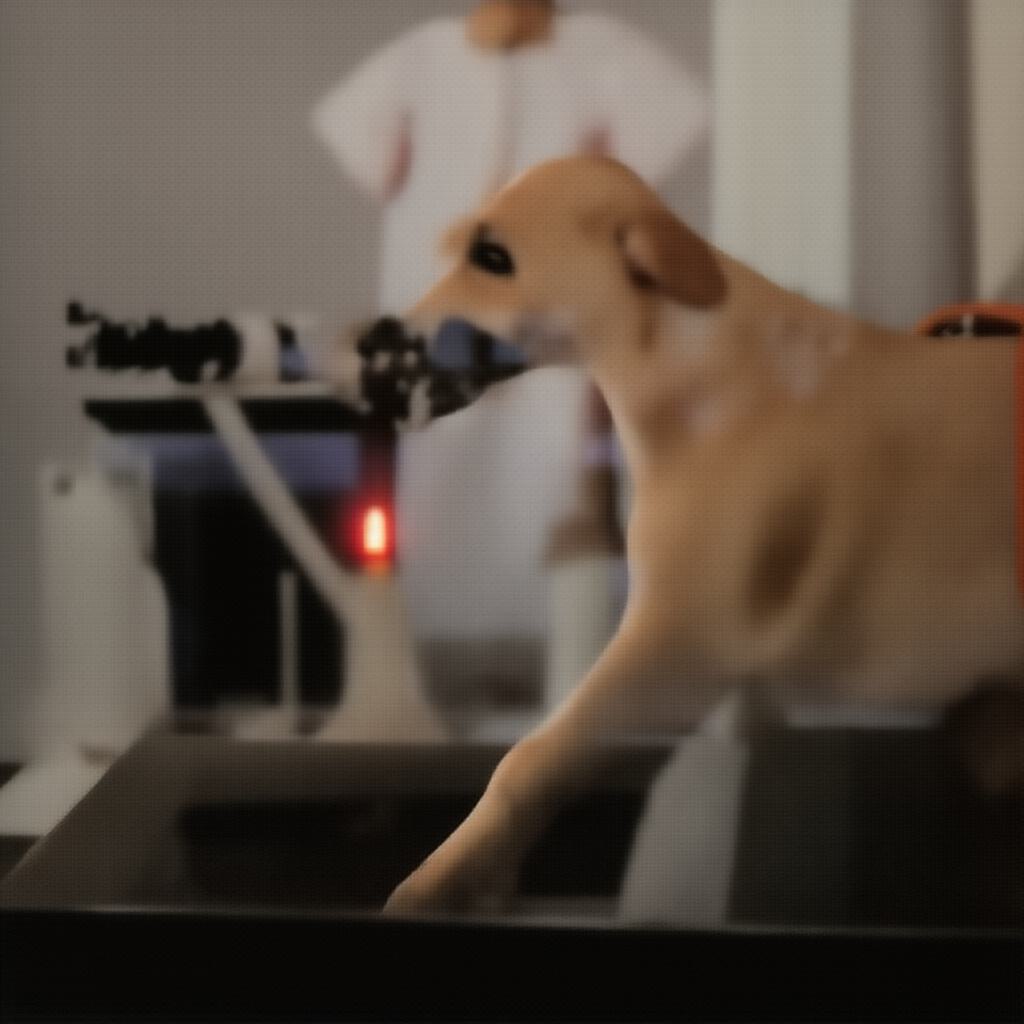} &
        \includegraphics[valign=c, width=\ww]{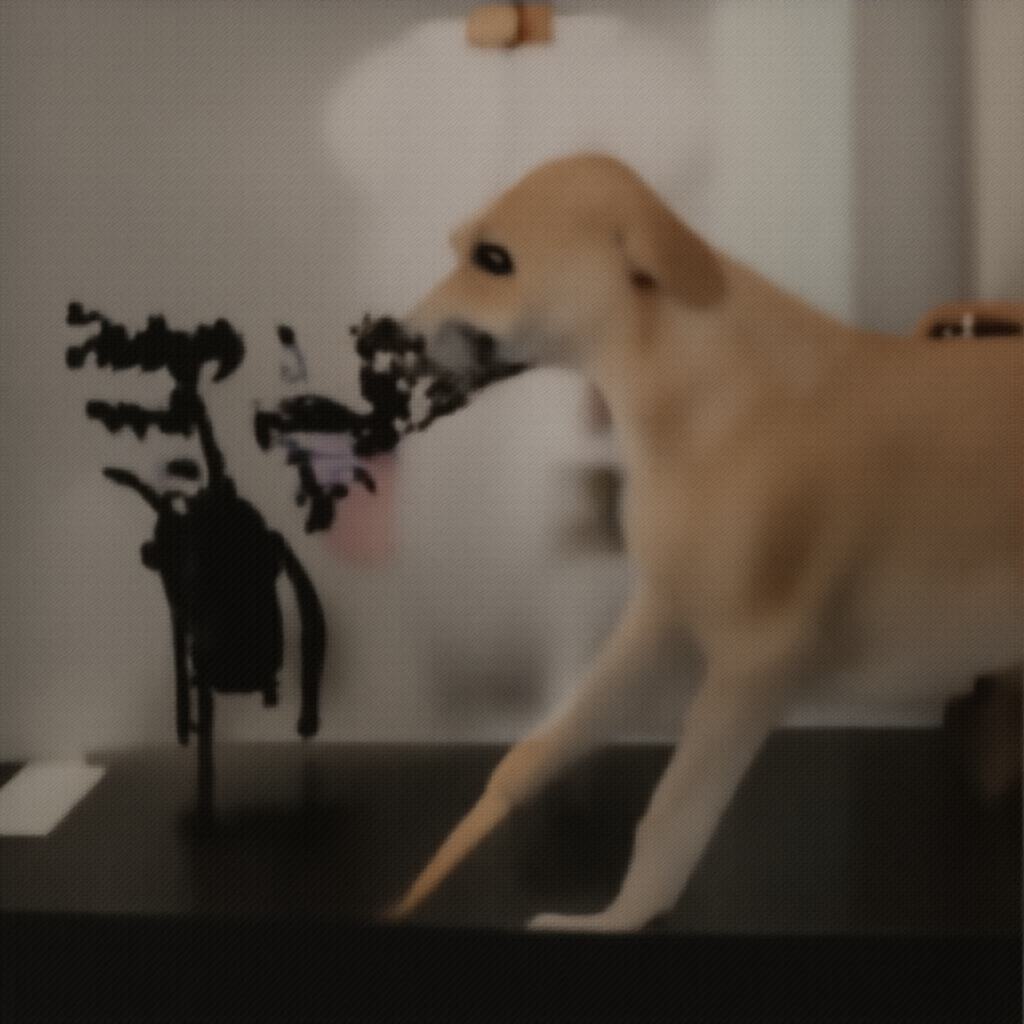} &
        \includegraphics[valign=c, width=\ww]{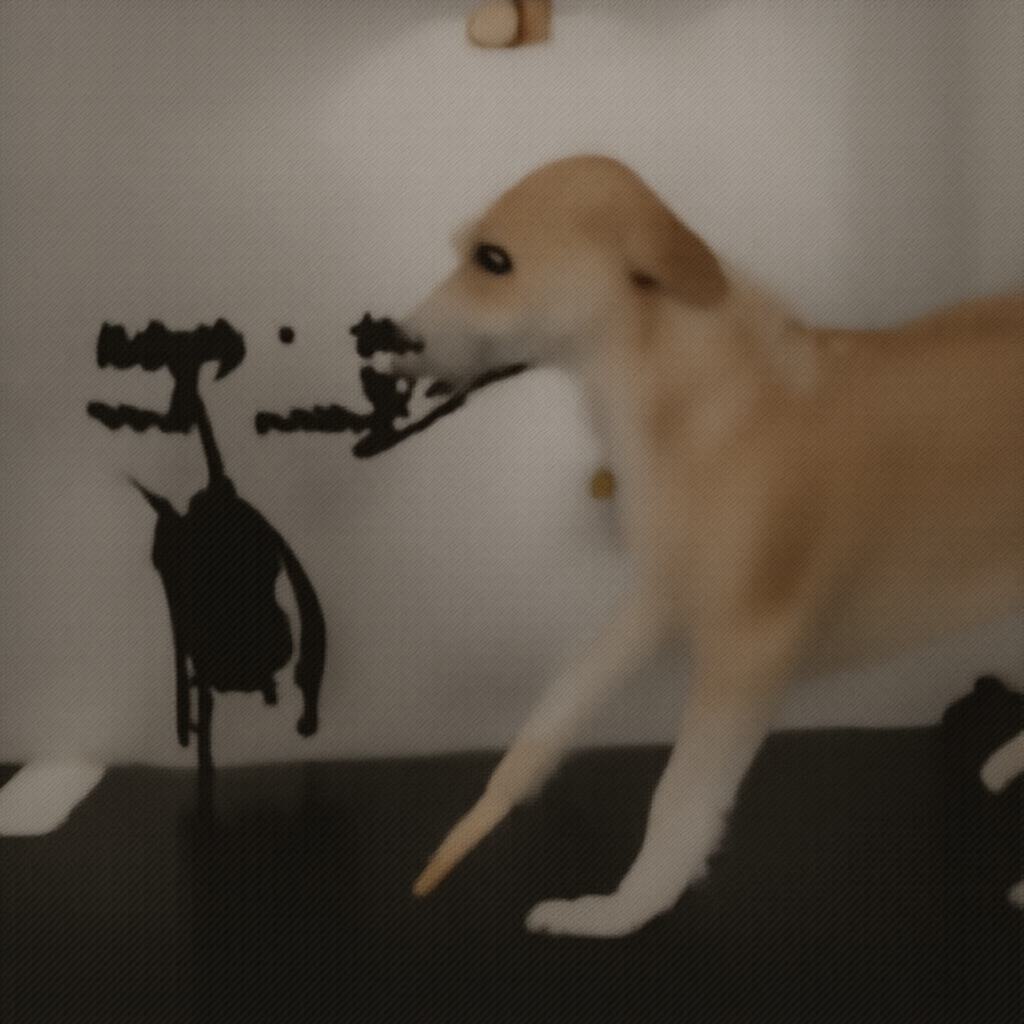} &
        \includegraphics[valign=c, width=\ww]{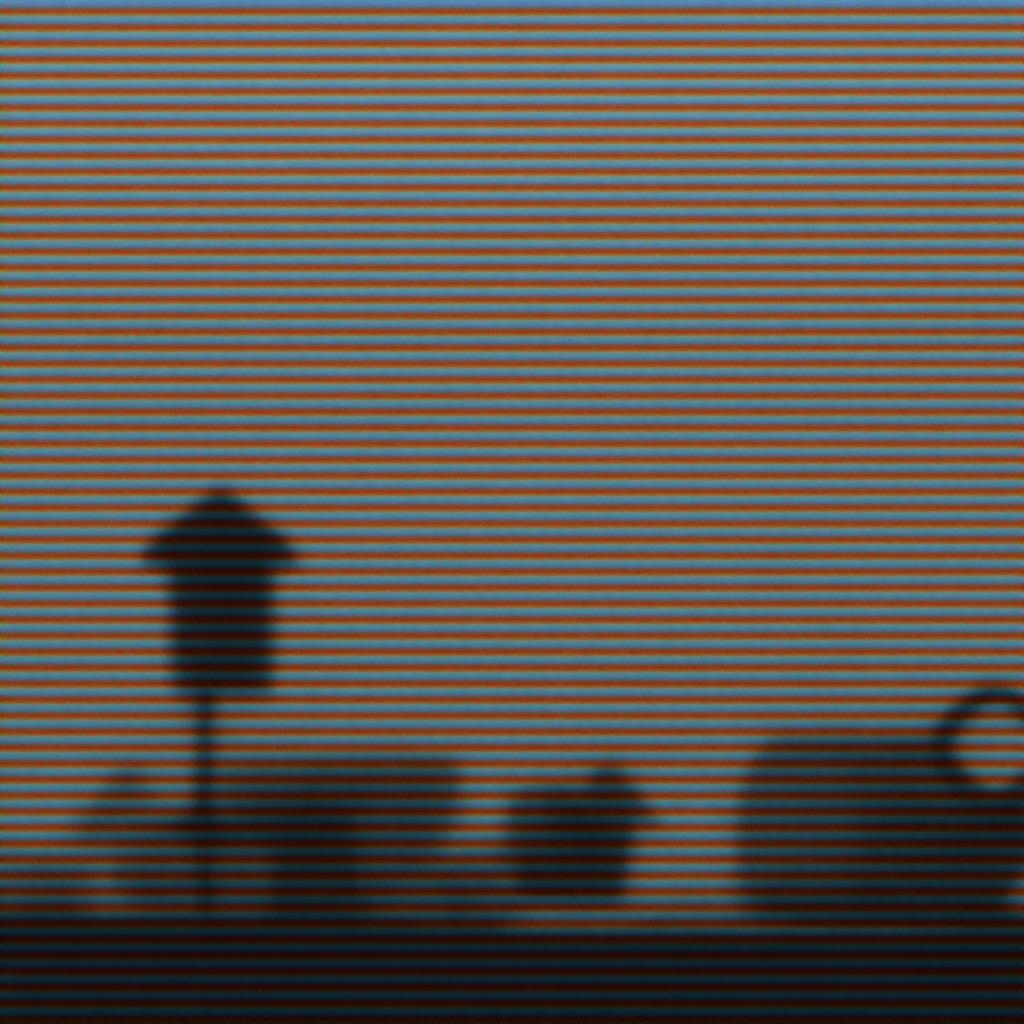}
        \vspace{3px}
        \\

        &
        $\lambda=0.9$ &
        $\lambda=0.8$ &
        $\lambda=0.7$ &
        $\lambda=0.6$
        \\

    \end{tabular}
    \caption{\textbf{Latent Nudging Values.} As described in \Cref{sec:latent_nudging_experiment}, we empirically tested different values for the latent nudging hyperparameter $\lambda$. In our experiments, we performed inversion using the inverse Euler ODE solver with a high number of 1,000 inversion (and denoising) steps, to reduce the inversion error. However, even when using such a high number of inversion/denoising steps, we notice that when not using latent nudging (\ie, $\lambda=1.0$), the reconstruction quality is poor (notice the eyes and the legs of the dog). Next, we found that $\lambda=1.15$ is the smallest value that enables full reconstruction using the inverse Euler solver. Furthermore, nudging values that are too high (\eg, $\lambda=3.0$) result in saturated images. Lastly, we notice that reducing nudging values ($\lambda<1.0$) severely damages the reconstruction quality.}
    \label{fig:latent_nudging_values}
\end{figure*}

\begin{figure*}[tp]
    \centering
    \setlength{\tabcolsep}{1.0pt}
    \renewcommand{\arraystretch}{0.8}
    \setlength{\ww}{0.23\linewidth}
    \begin{tabular}{cccc}

        Input &
        Reconstruction &
        Caching only &
        Latent Nudging
        \vspace{2px}
        \\

        \includegraphics[valign=c, width=\ww]{figures/qualitative_comparison/assets/rabbit/inp.jpg} &
        \includegraphics[valign=c, width=\ww]{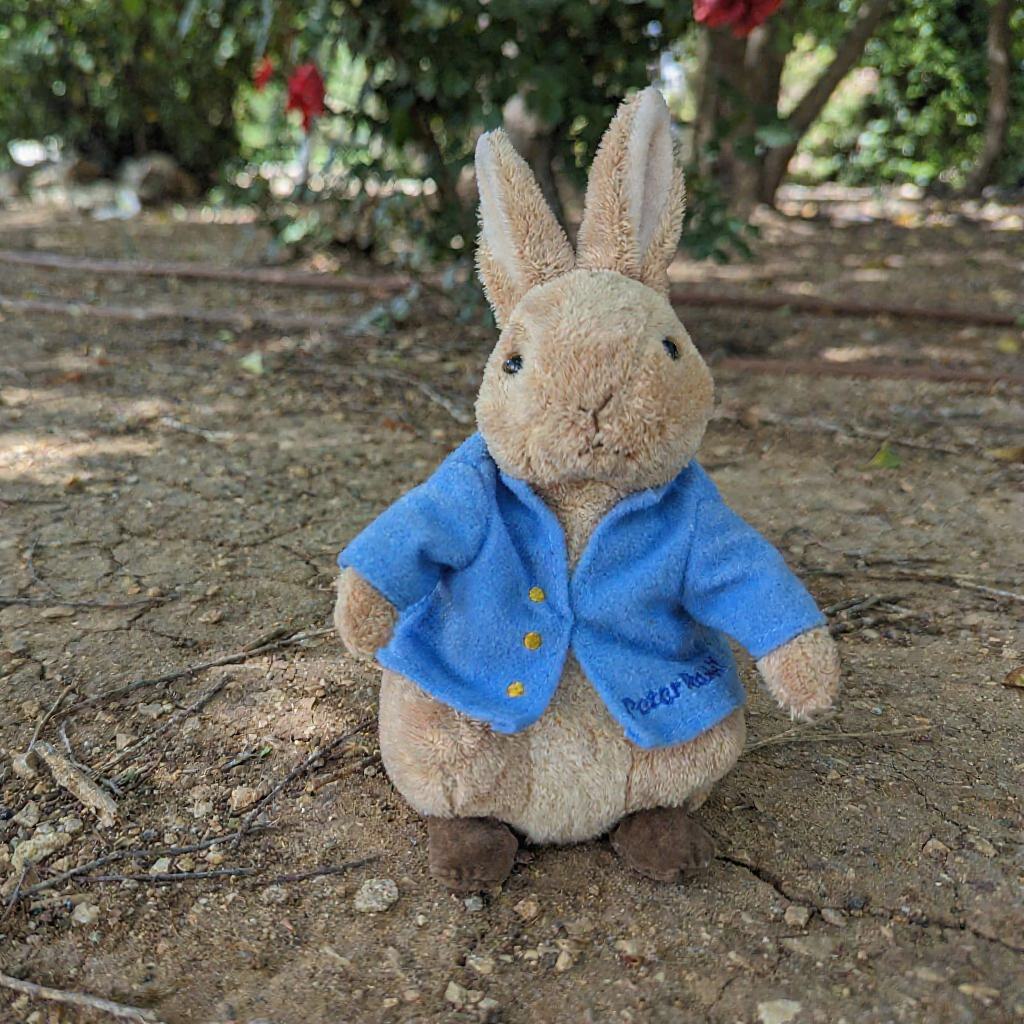} &
        \includegraphics[valign=c, width=\ww]{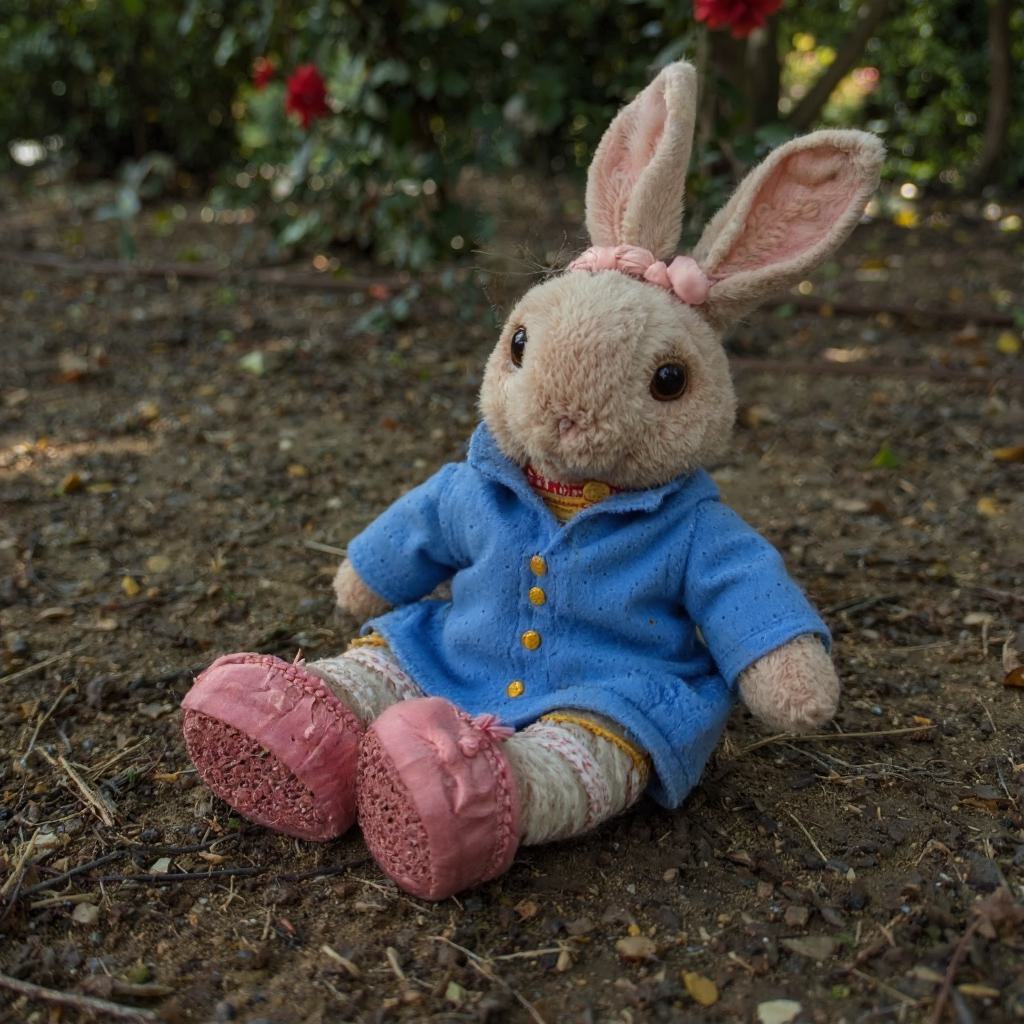} &
        \includegraphics[valign=c, width=\ww]{figures/qualitative_comparison/assets/rabbit/ours.jpg}
        \vspace{1px}
        \\

        &
        &
        \multicolumn{2}{c}{\small{\promptstart{A rabbit toy sitting and wearing}}}
        \\

        &
        &
        \multicolumn{2}{c}{\small{\promptend{pink socks during the late afternoon}}}
        \vspace{10px}
        \\

        \includegraphics[valign=c, width=\ww]{figures/qualitative_comparison/assets/cat/inp.jpg} &
        \includegraphics[valign=c, width=\ww]{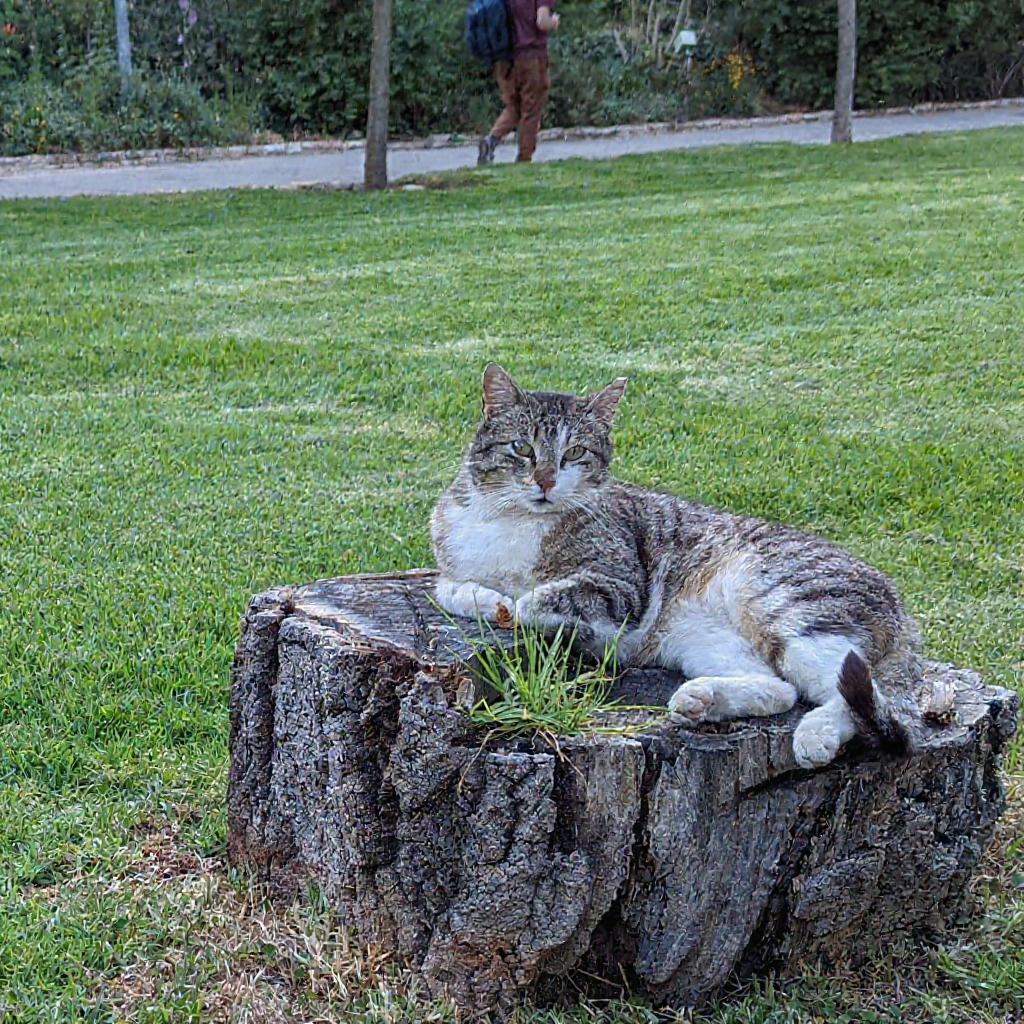} &
        \includegraphics[valign=c, width=\ww]{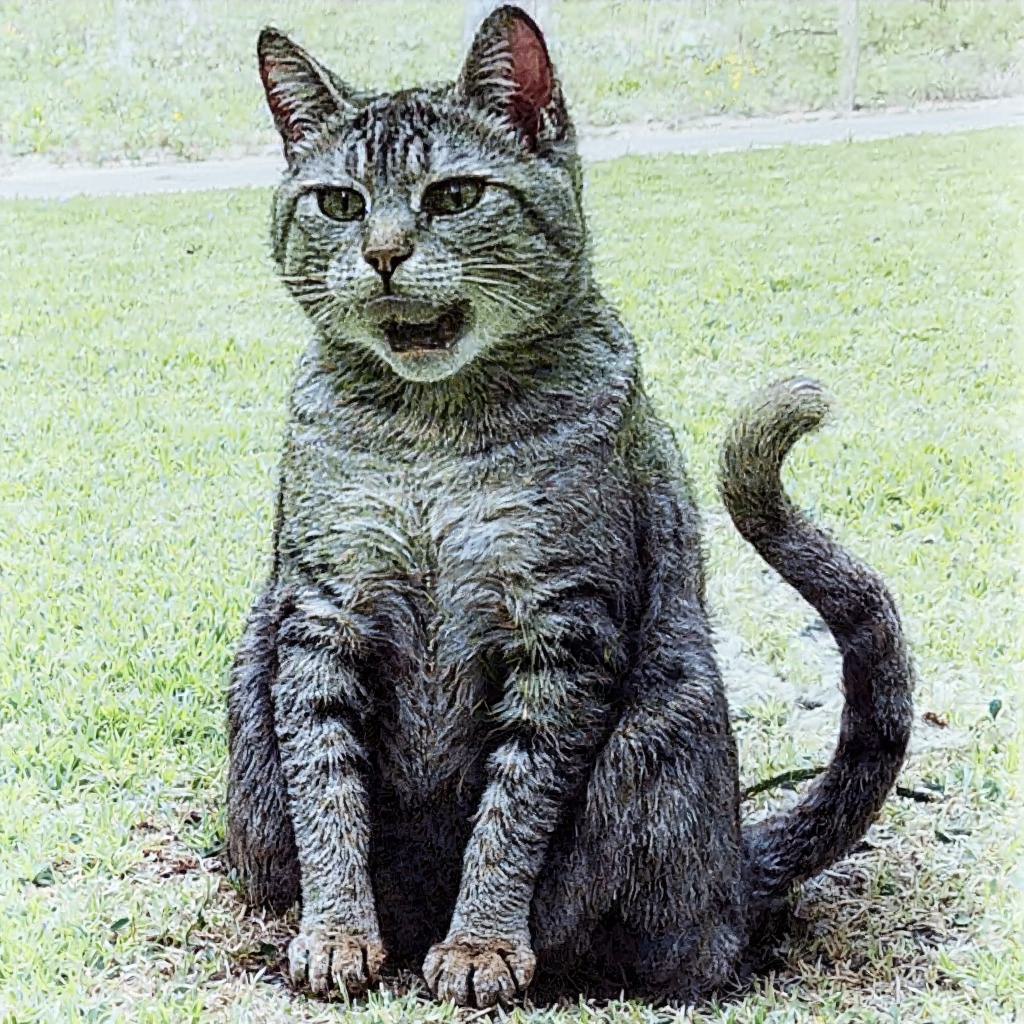} &
        \includegraphics[valign=c, width=\ww]{figures/qualitative_comparison/assets/cat/ours.jpg}
        \vspace{1px}
        \\

        &
        &
        \multicolumn{2}{c}{\small{\prompt{The cat is yelling and raising its paw}}}
        \vspace{10px}
        \\

        \includegraphics[valign=c, width=\ww]{figures/qualitative_comparison/assets/duck/inp.jpg} &
        \includegraphics[valign=c, width=\ww]{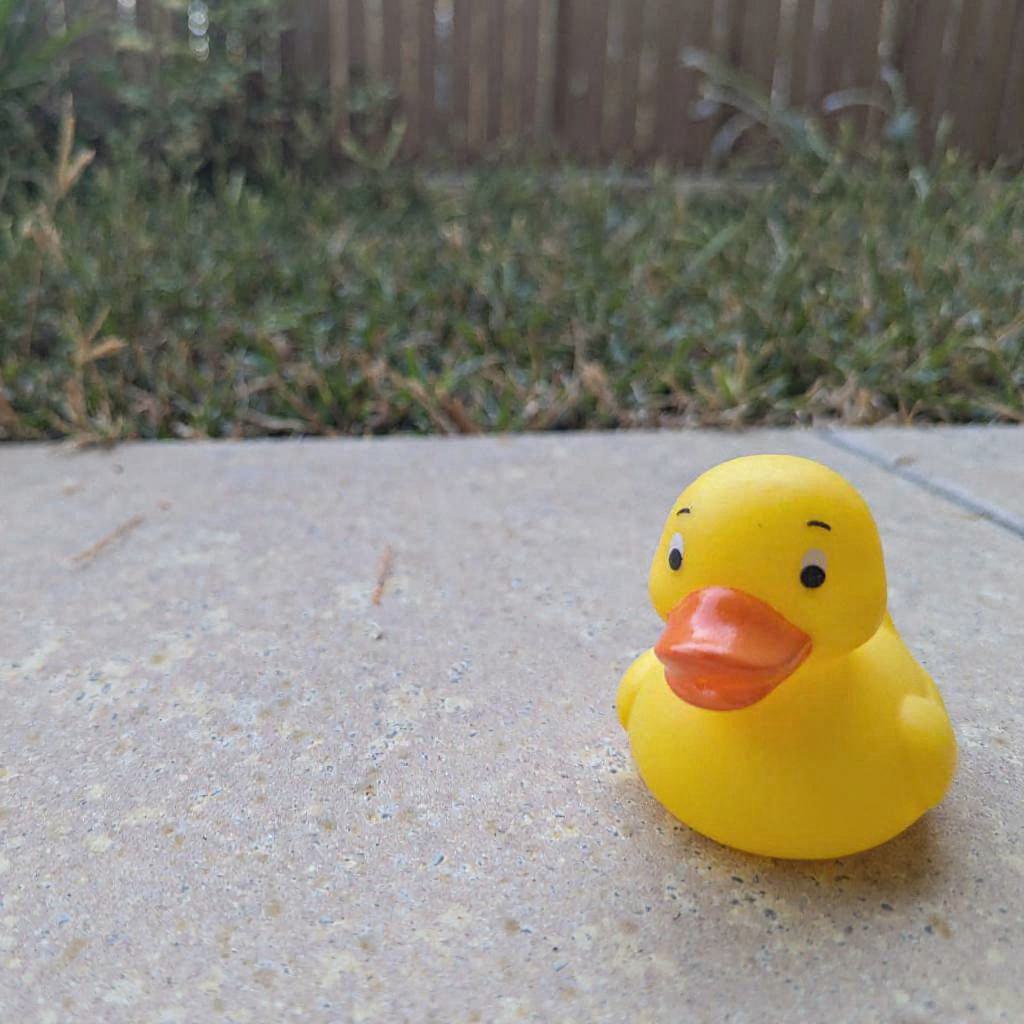} &
        \includegraphics[valign=c, width=\ww]{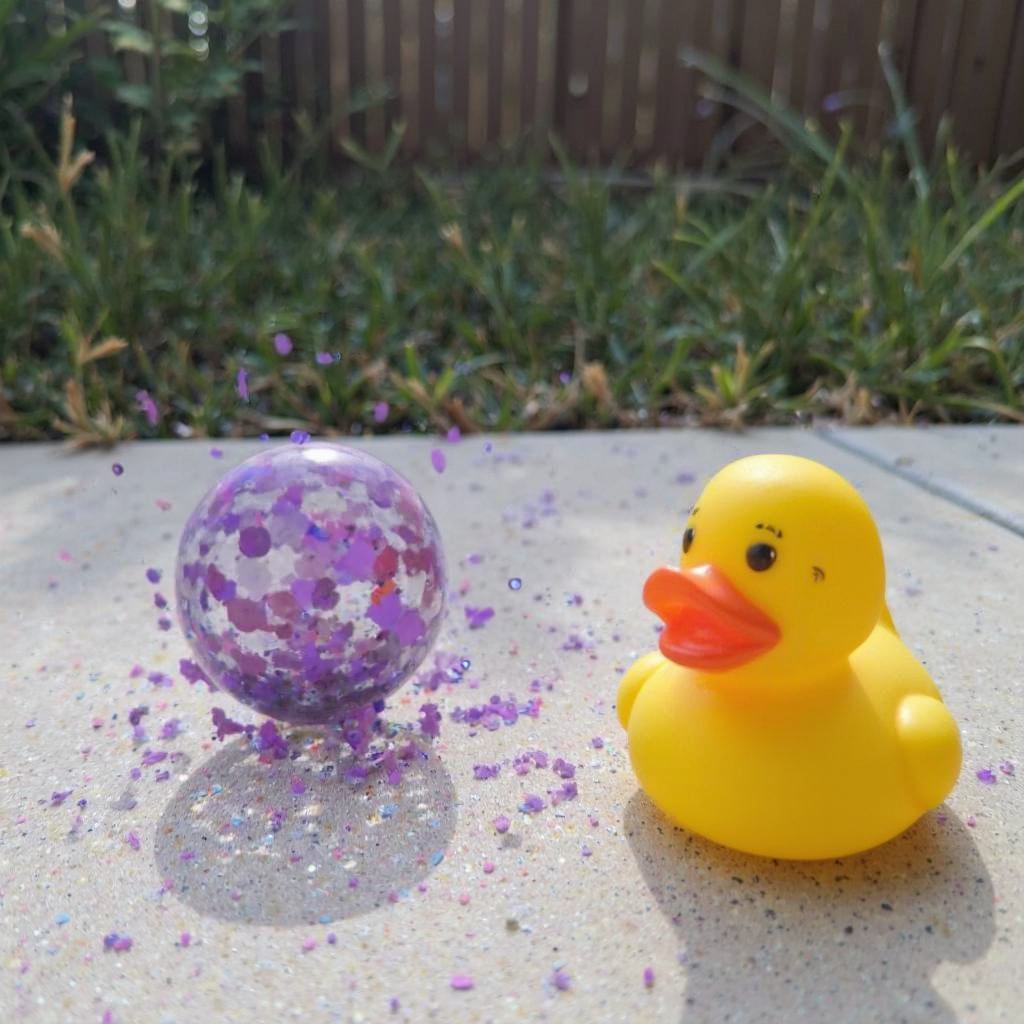} &
        \includegraphics[valign=c, width=\ww]{figures/qualitative_comparison/assets/duck/ours.jpg}
        \vspace{1px}
        \\

        &
        &
        \multicolumn{2}{c}{\small{\prompt{A rubber duck next to a purple ball during a sunny day}}}
        \vspace{10px}
        \\

        \includegraphics[valign=c, width=\ww]{figures/qualitative_comparison/assets/man/inp.jpg} &
        \includegraphics[valign=c, width=\ww]{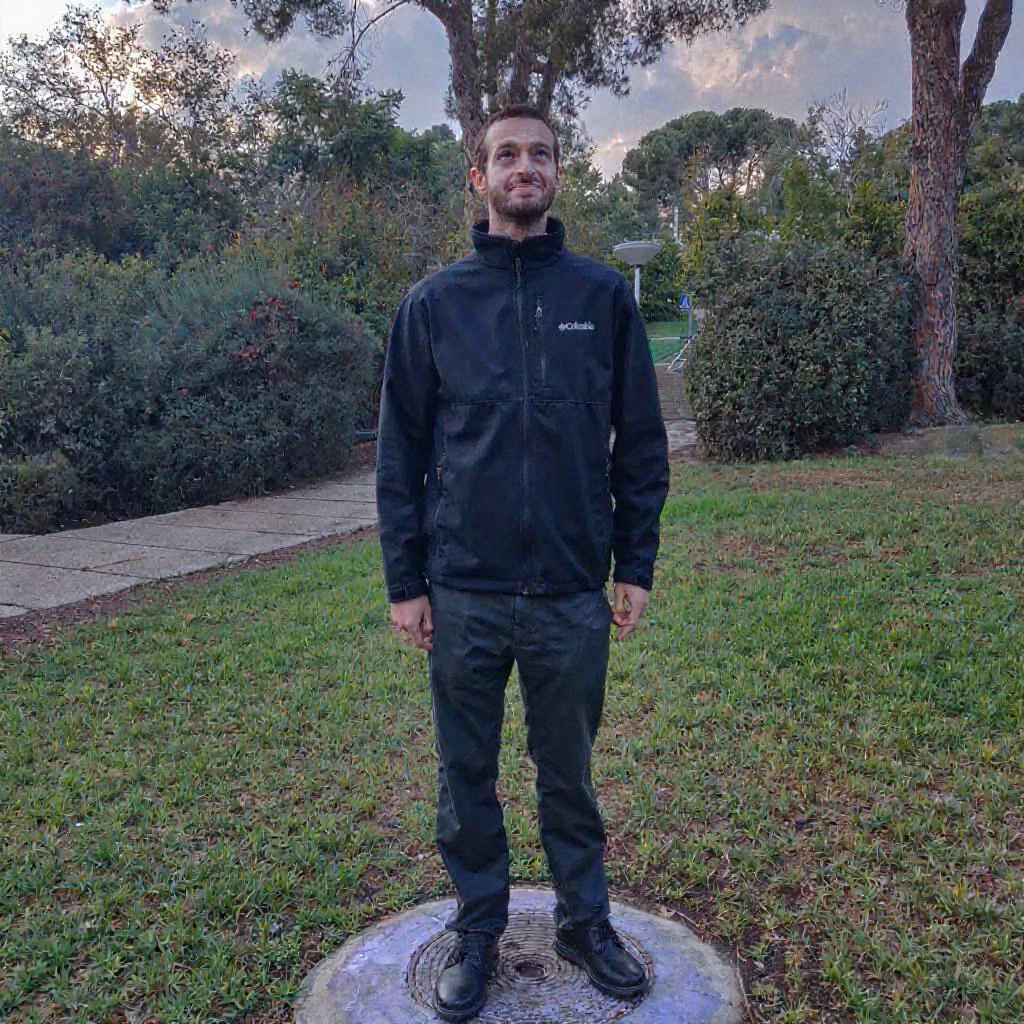} &
        \includegraphics[valign=c, width=\ww]{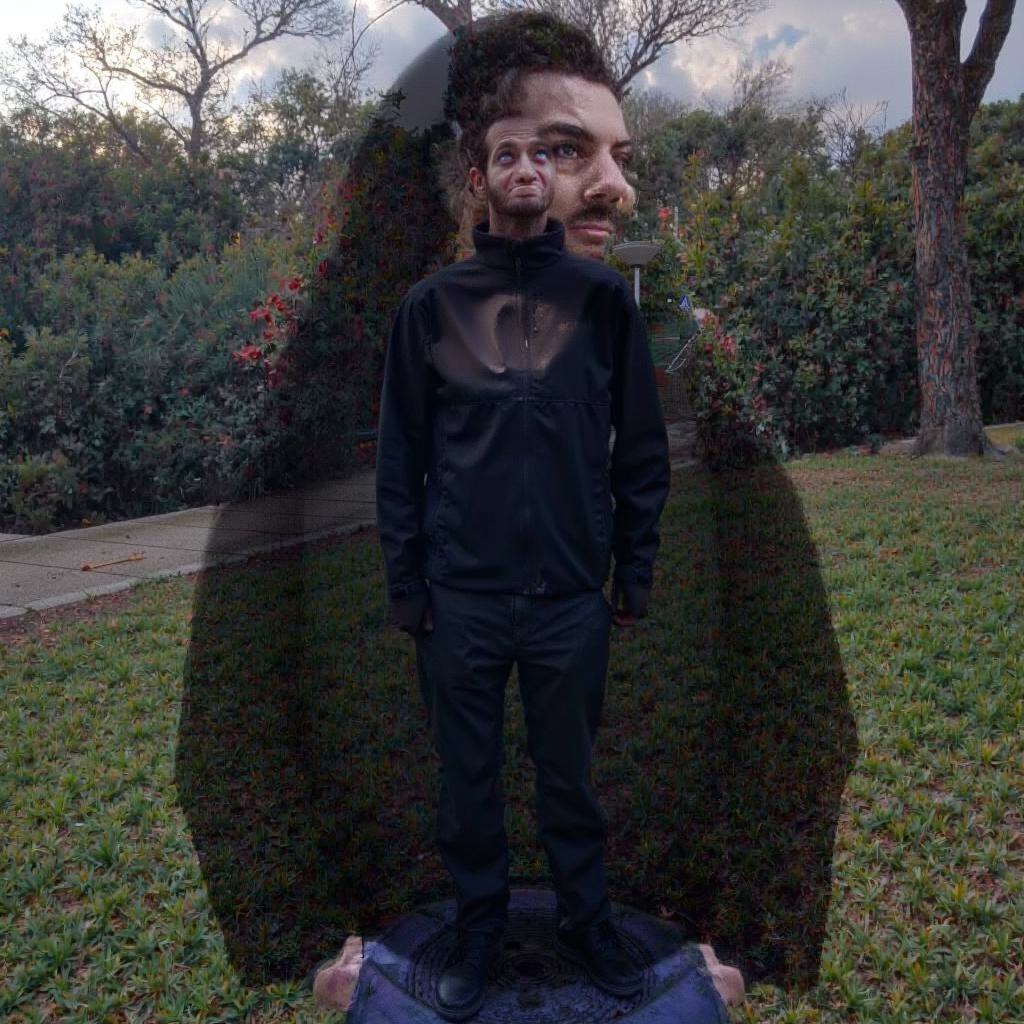} &
        \includegraphics[valign=c, width=\ww]{figures/qualitative_comparison/assets/man/ours.jpg}
        \vspace{1px}
        \\

        &
        &
        \multicolumn{2}{c}{\small{\prompt{A man with a long hair}}}
        \vspace{10px}
        \\

    \end{tabular}
    \caption{\textbf{Latent Caching.} As explained in \Cref{sec:latent_nudging_experiment}, we also tested a latent caching approach~\cite{avrahami2024diffuhaul}, in which we saved the series of latents during the inversion process without applying latent nudging. As can be seen, this approach indeed achieves perfect inversion (second column), but (third column) still struggles with preserving the identities while editing the image (\eg, the rabbit and duck examples) or significantly alters the image (\eg, the cat and man examples). On the other hand, our method with the latent nudging (fourth column) is able to preserve the identities during editing.}
    \label{fig:latent_caching}
\end{figure*}

As described in Section 3.3 of the main paper,
we proposed using a latent nudging technique to avoid the bad reconstruction quality of vanilla inverse Euler ODE solver. We suggest multiplying the initial latent $z_0$ by a small scalar $\lambda=1.15$ to slightly offset it from the training distribution. As shown in \Cref{fig:latent_nudging_values}, we empirically tested different values for the latent nudging hyperparameter $\lambda$. We performed inversion using the inverse Euler ODE solver with a high number of 1,000 inversion (and denoising) steps, to reduce the inversion error. However, even when using such a high number of inversion/denoising steps, we notice that when not using latent nudging (\ie, $\lambda=1.0$), the reconstruction quality is poor (notice the eyes and the legs of the dog). Next, we found that $\lambda=1.15$ is the smallest value that enables full reconstruction using the inverse Euler solver. Furthermore, nudging values that are too high (\eg, $\lambda=3.0$) result in saturated images. Lastly, we notice that decreasing nudging values (\ie, $\lambda<1.0$) severely damages the reconstruction quality.

In addition, we experiment with a simpler inversion variant based on latent caching (termed \emph{DDPM bucketing} in DiffUHaul~\cite{avrahami2024diffuhaul}), in which we saved the series of latents during the inversion process without applying latent nudging. As shown in \Cref{fig:latent_caching}, this approach indeed achieves perfect inversion (second column), but (third column) still struggles with preserving the identities while editing the image (\eg, the rabbit and duck examples) or significantly alters the image (\eg, the cat and man examples). On the other hand, our method (fourth column) with the latent nudging is able to preserve the identities during editing. In practice, we found that using latent caching in addition to latent nudging enables inversion with a lower number of steps (50 steps), hence, this is the approach we used.

\subsection{Layer Bypassing Visualization}
\label{sec:layer_bypassing_visualization}

\begin{figure*}[tp]
    \centering
    \setlength{\tabcolsep}{2.5pt}
    \renewcommand{\arraystretch}{1.0}
    \setlength{\ww}{0.32\linewidth}
    \begin{tabular}{ccc}

        \includegraphics[valign=c, width=\ww, frame]{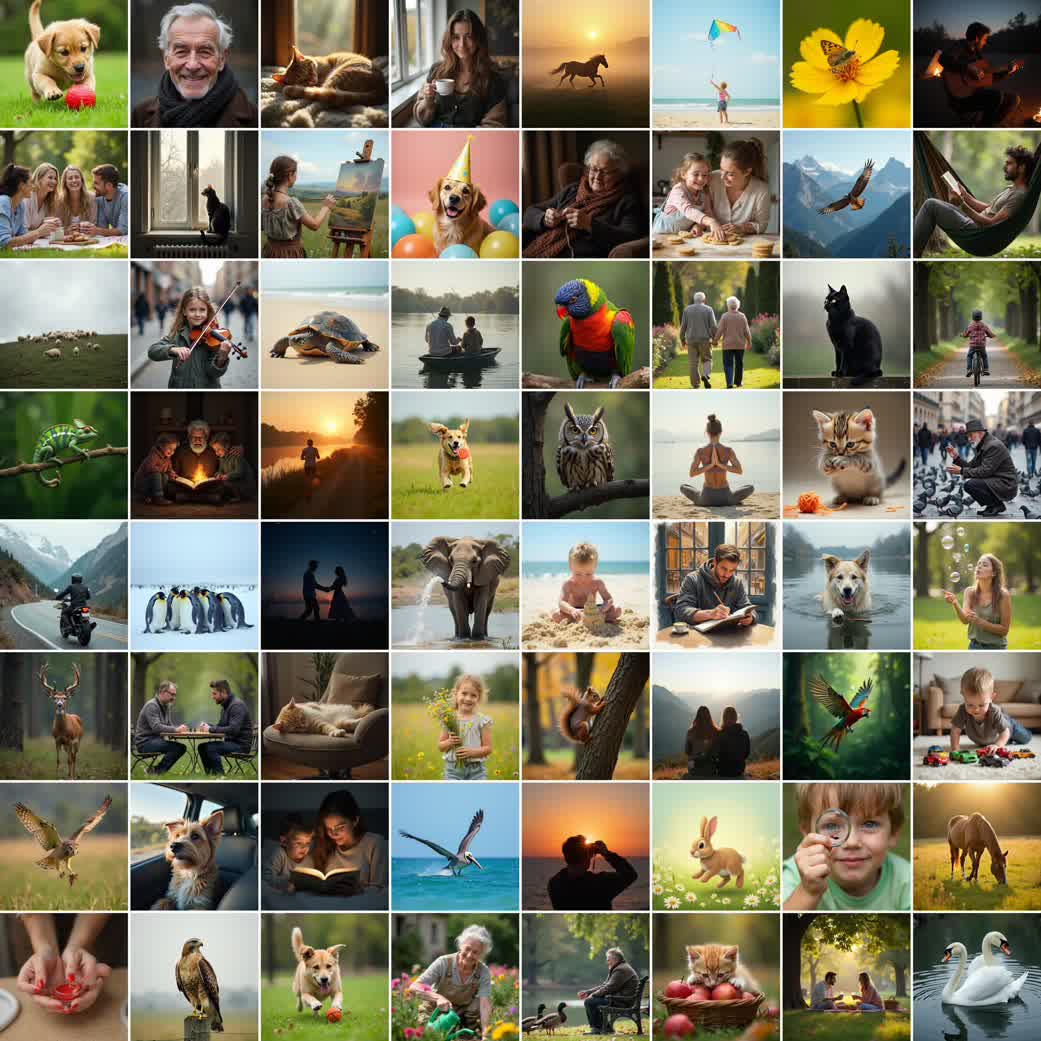} &
        \includegraphics[valign=c, width=\ww, frame]{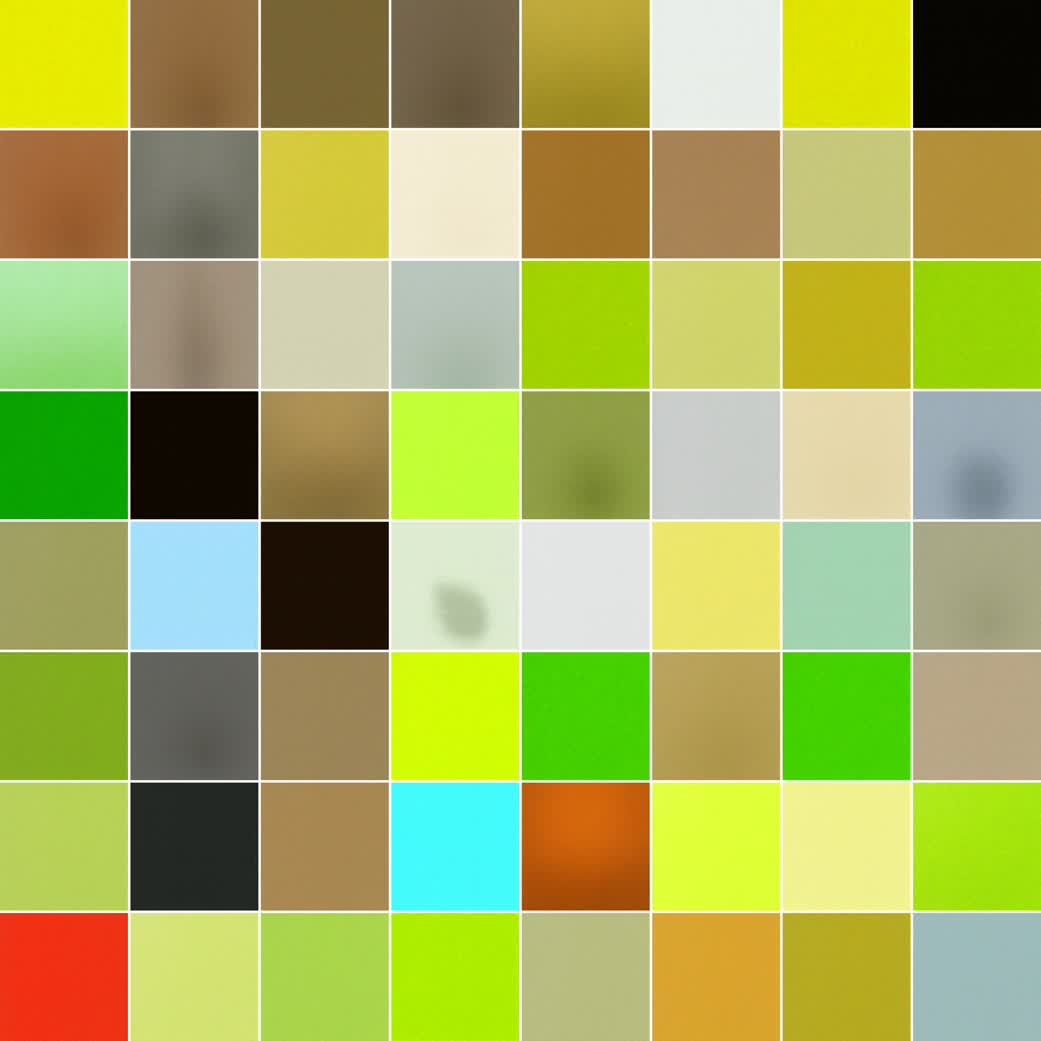} &
        \includegraphics[valign=c, width=\ww, frame]{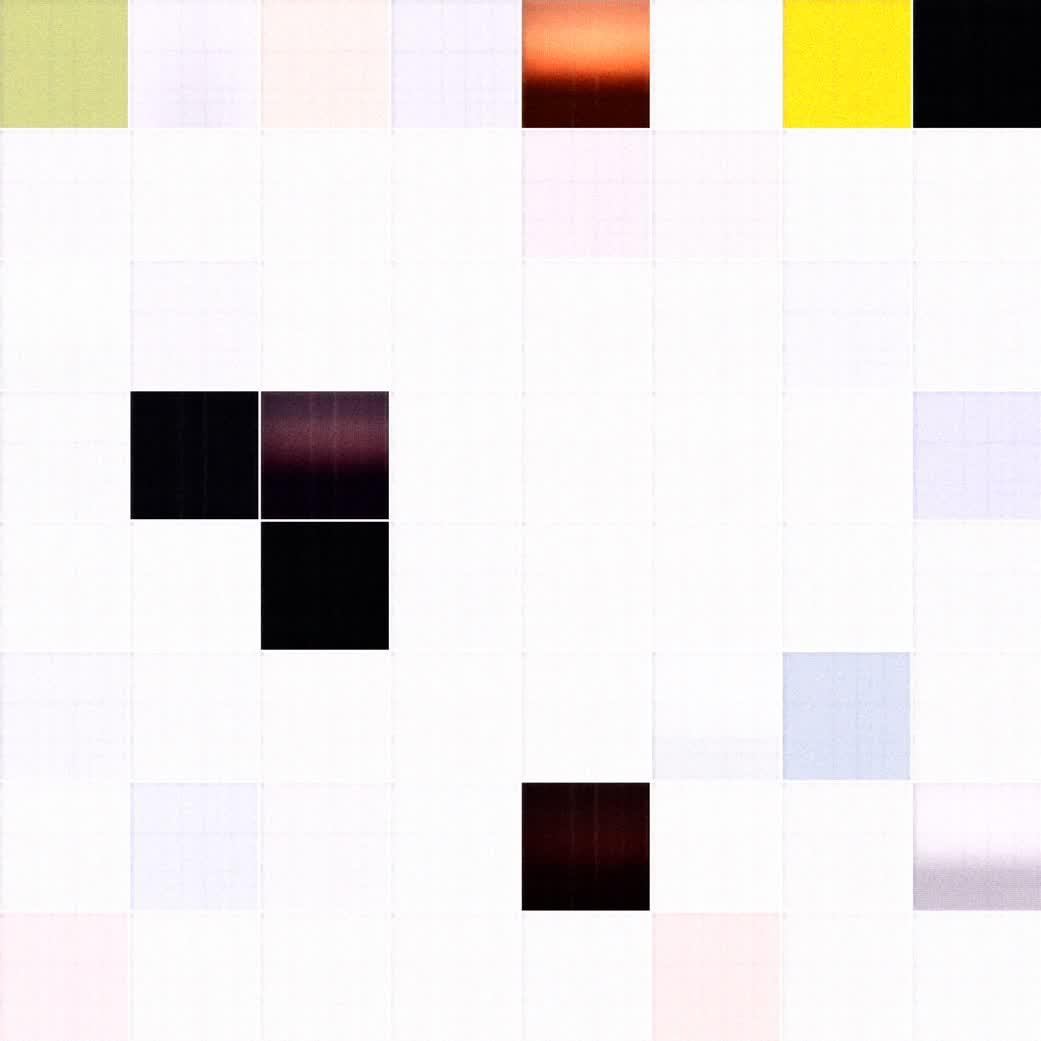}
        \vspace{3px}
        \\

        $G_{\textit{ref}}$ &
        \vital{$G_{0}$} &
        \vital{$G_{1}$}
        \vspace{15px}
        \\

        \includegraphics[valign=c, width=\ww, frame]{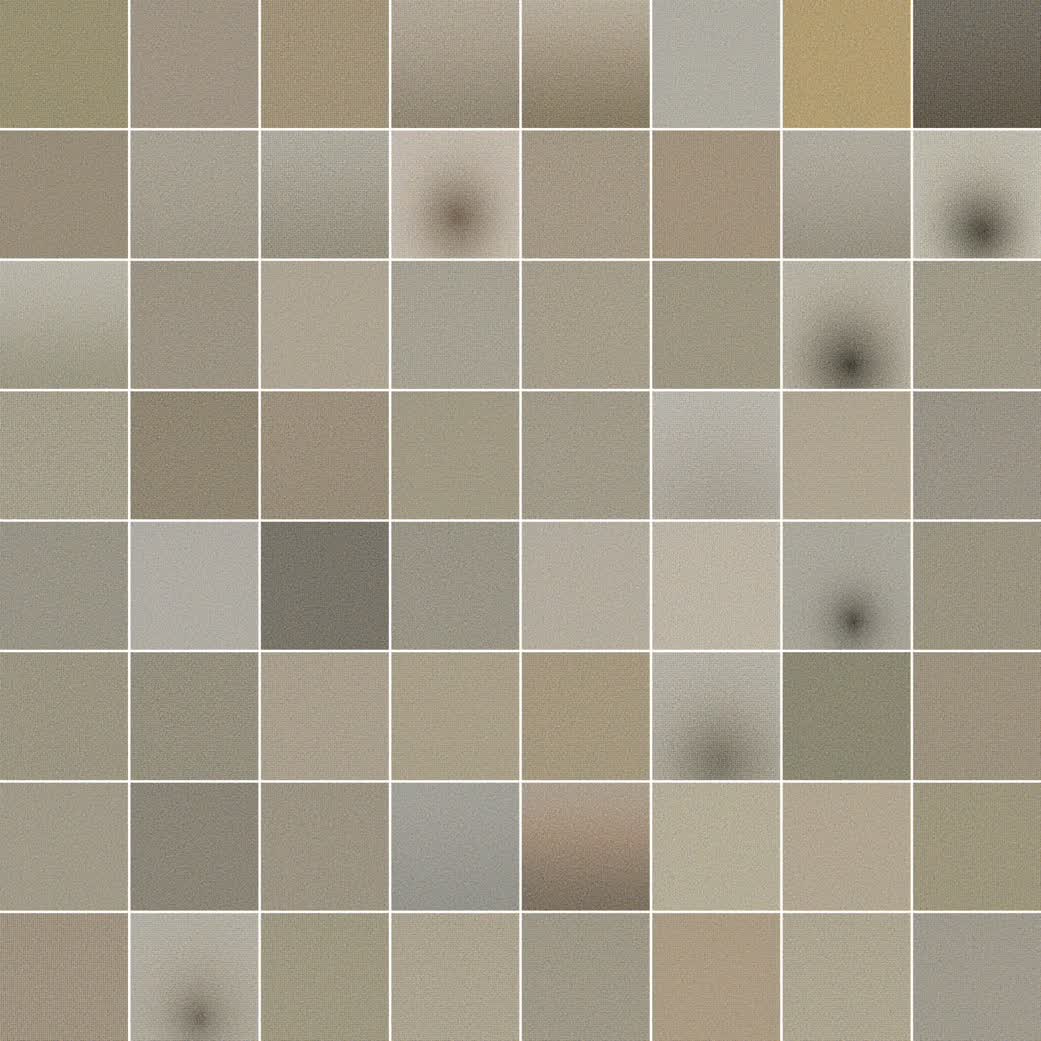} &
        \includegraphics[valign=c, width=\ww, frame]{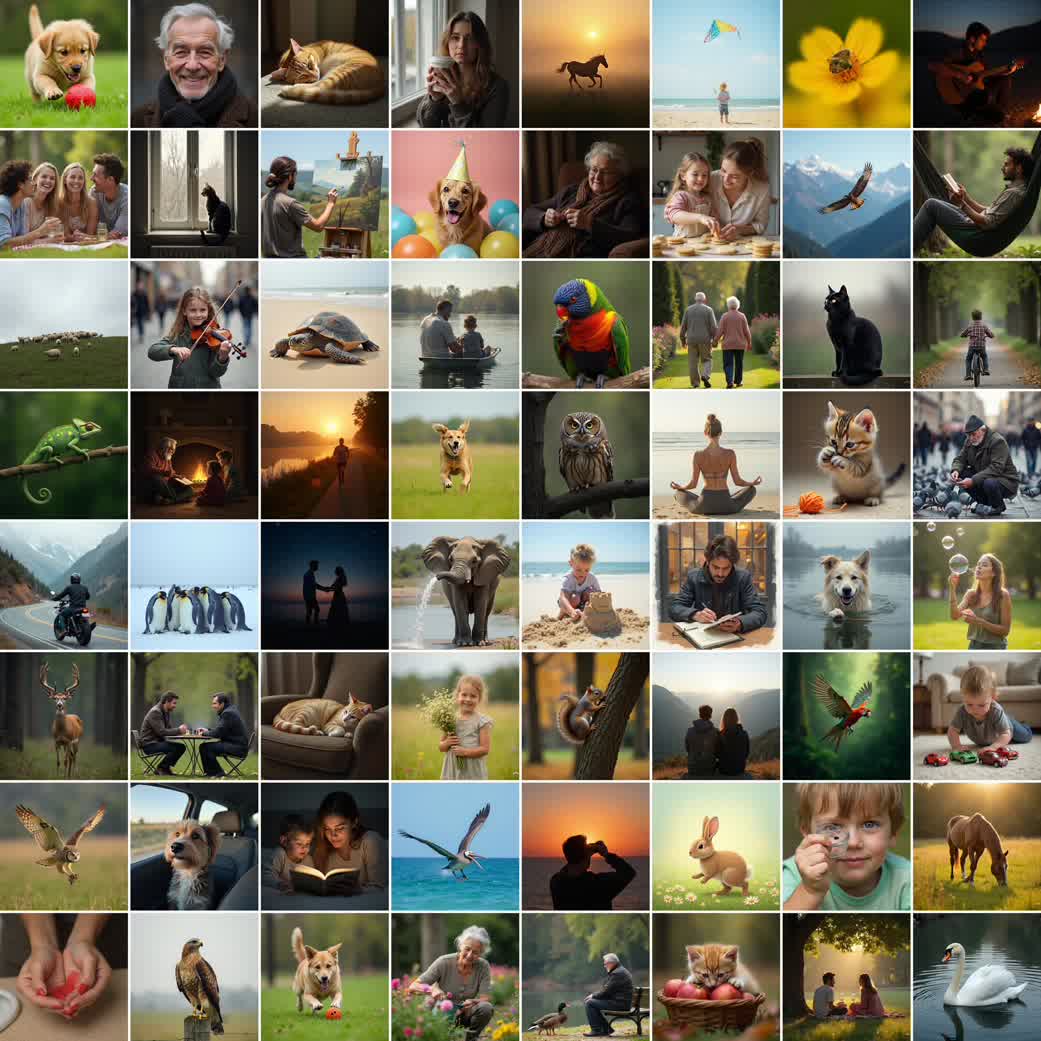} &
        \includegraphics[valign=c, width=\ww, frame]{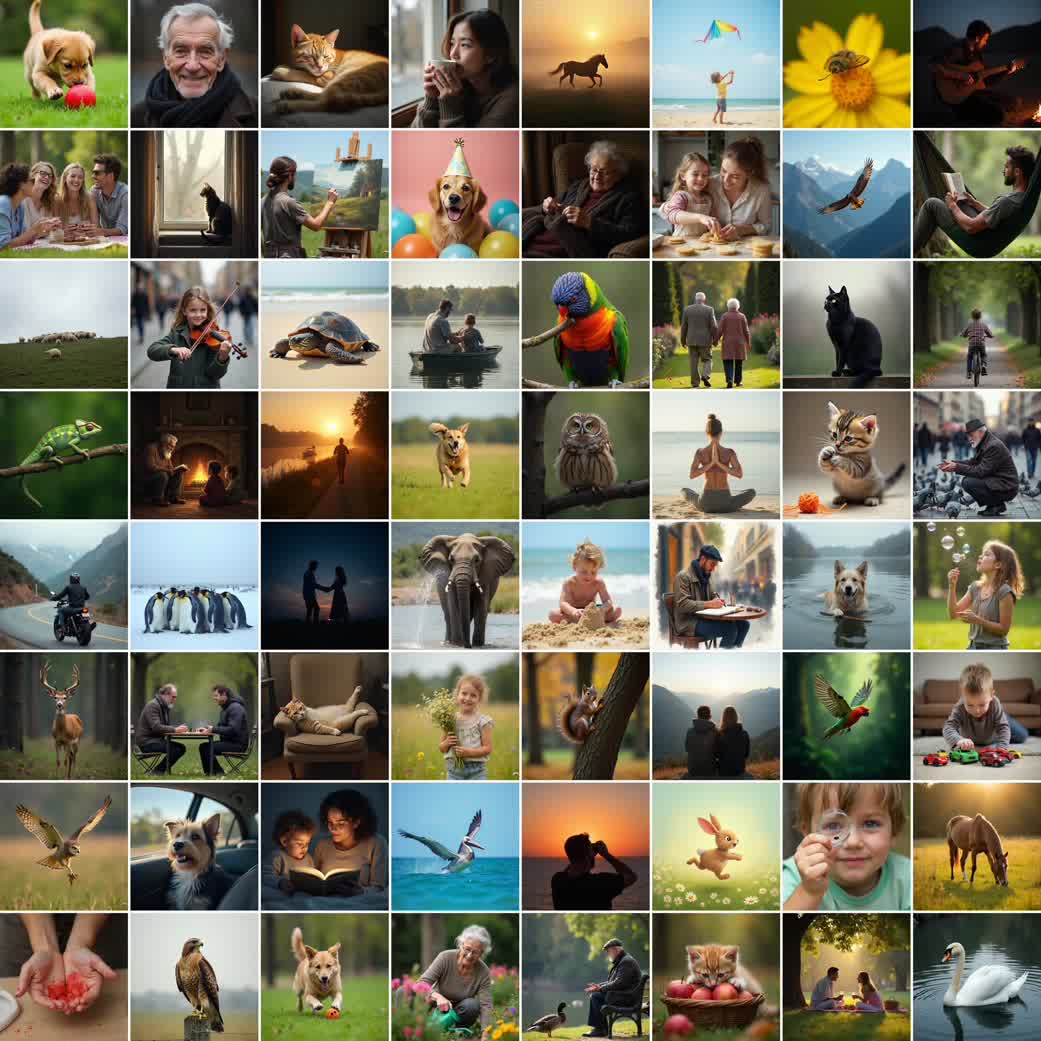}
        \vspace{3px}
        \\

        \vital{$G_{2}$} &
        \nonvital{$G_{3}$} &
        \nonvital{$G_{4}$}
        \vspace{15px}
        \\

        \includegraphics[valign=c, width=\ww, frame]{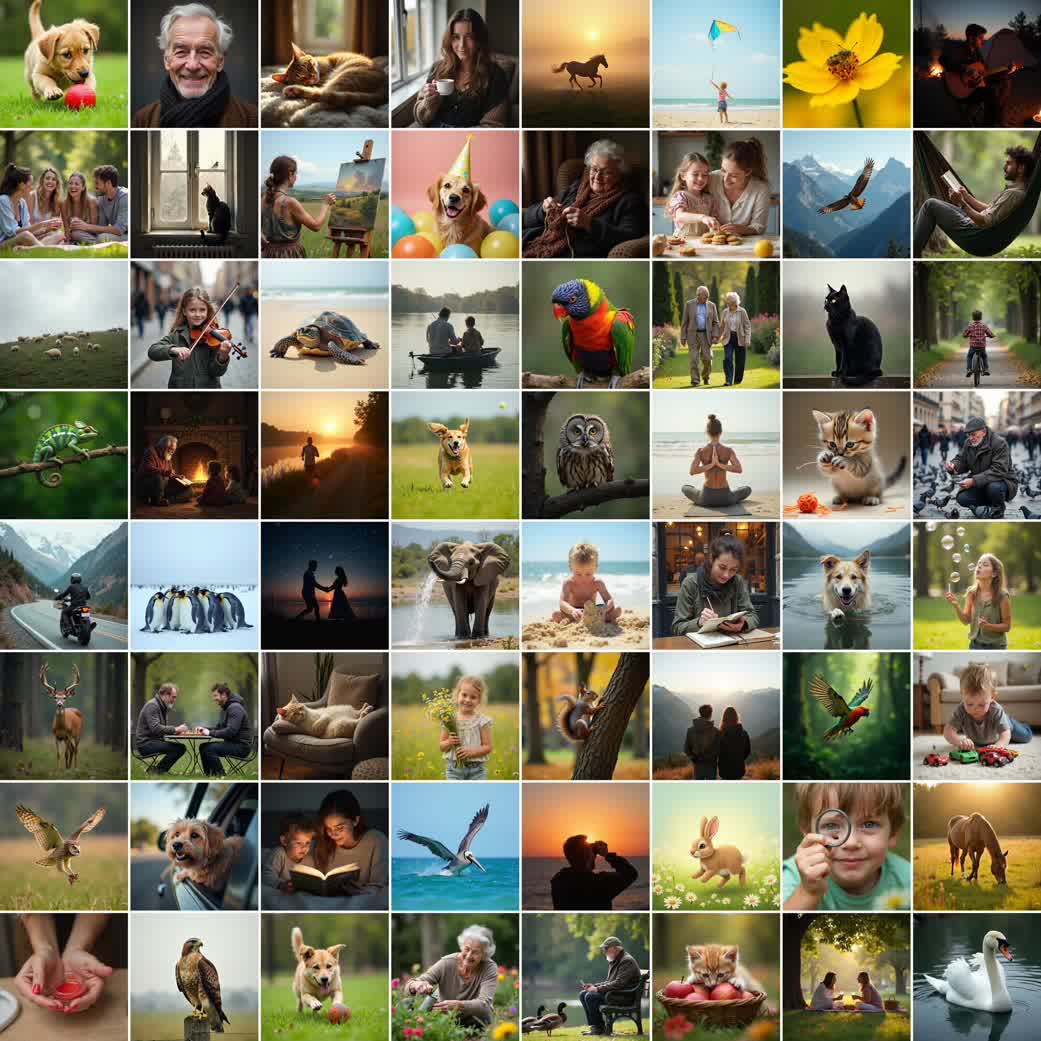} &
        \includegraphics[valign=c, width=\ww, frame]{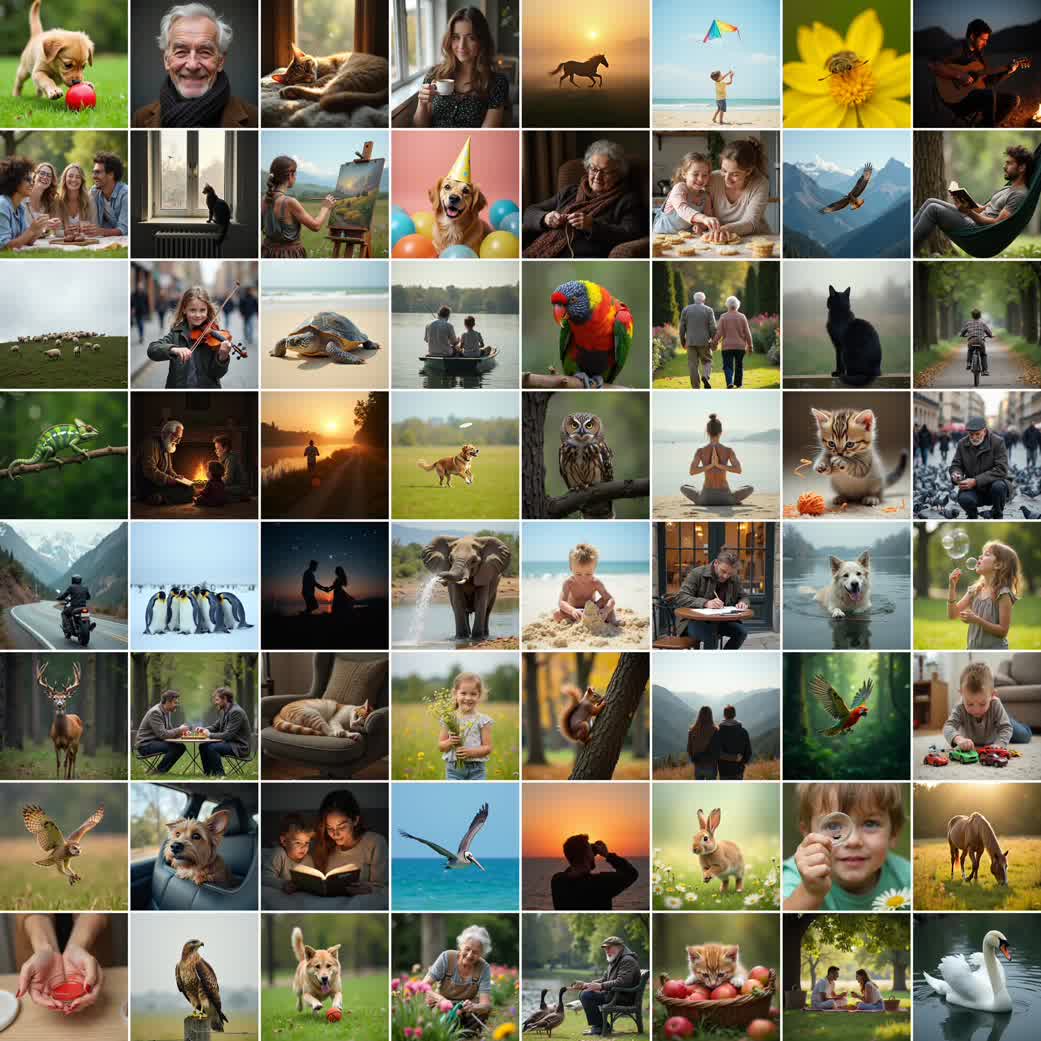} &
        \includegraphics[valign=c, width=\ww, frame]{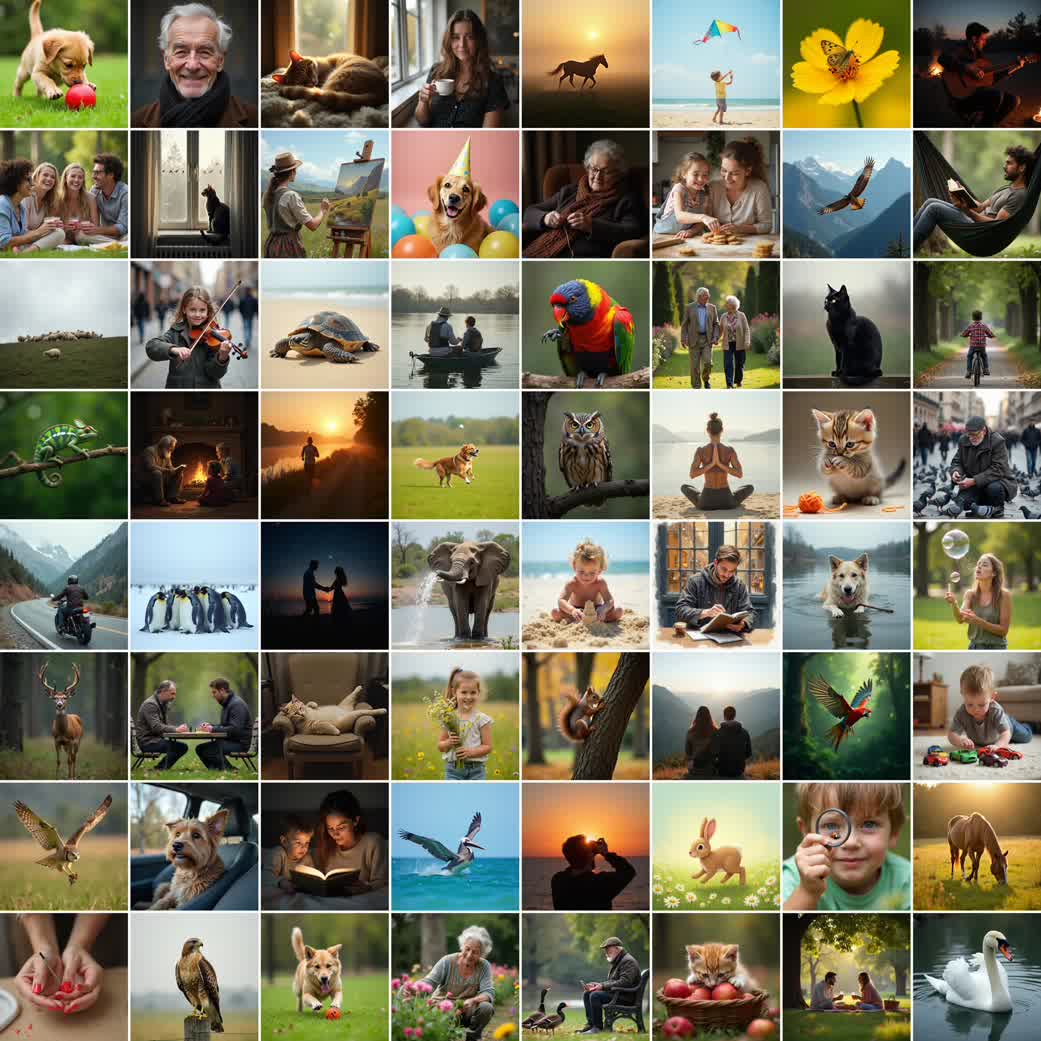}
        \vspace{3px}
        \\

        \nonvital{$G_{5}$} &
        \nonvital{$G_{6}$} &
        \nonvital{$G_{7}$}
        \vspace{3px}
        \\

    \end{tabular}
    \caption{\textbf{Full Layer Bypassing Visualization for Flux.} We visualize the individual layer bypassing study we conducted, as described in \Cref{sec:layer_bypassing_visualization}. We start by generating a set of images $G_{\textit{ref}}$ using a fixed set of seeds and prompts. Then, we bypass each layer $\ell$ by using its residual connection and generate the set of images $G_{\ell}$ using the same fixed set of prompts and seeds. In this visualization, \vital{$G_{0}$} -- \vital{$G_{2}$} are \vital{vital layers}, while \nonvital{$G_{3}$} -- \nonvital{$G_{7}$} are \nonvital{non-vital layers}.}
    \label{fig:full_flux_bypassing_1}
\end{figure*}

\begin{figure*}[tp]
    \centering
    \setlength{\tabcolsep}{2.5pt}
    \renewcommand{\arraystretch}{1.0}
    \setlength{\ww}{0.32\linewidth}
    \begin{tabular}{ccc}

        \includegraphics[valign=c, width=\ww, frame]{figures/full_bypassing_visualization_flux/assets/src.jpg} &
        \includegraphics[valign=c, width=\ww, frame]{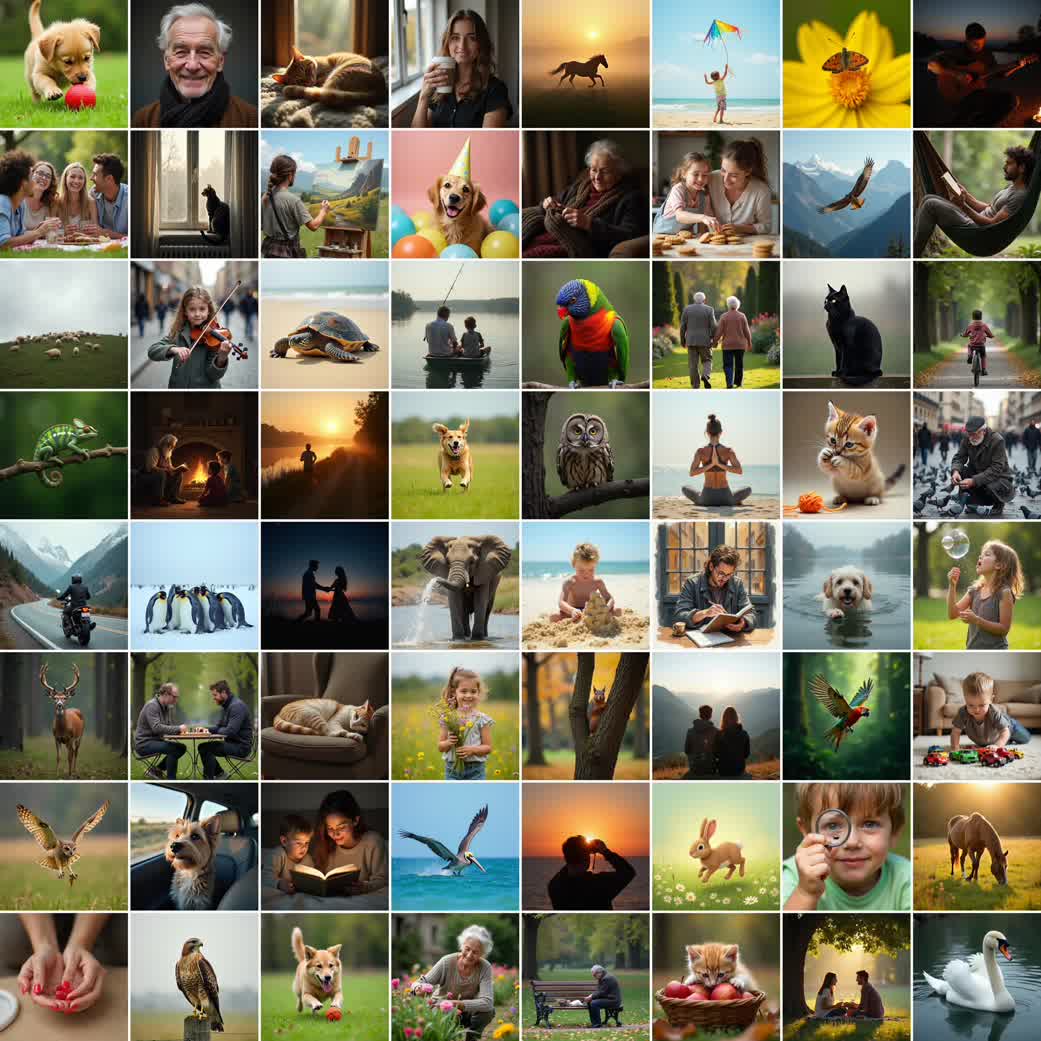} &
        \includegraphics[valign=c, width=\ww, frame]{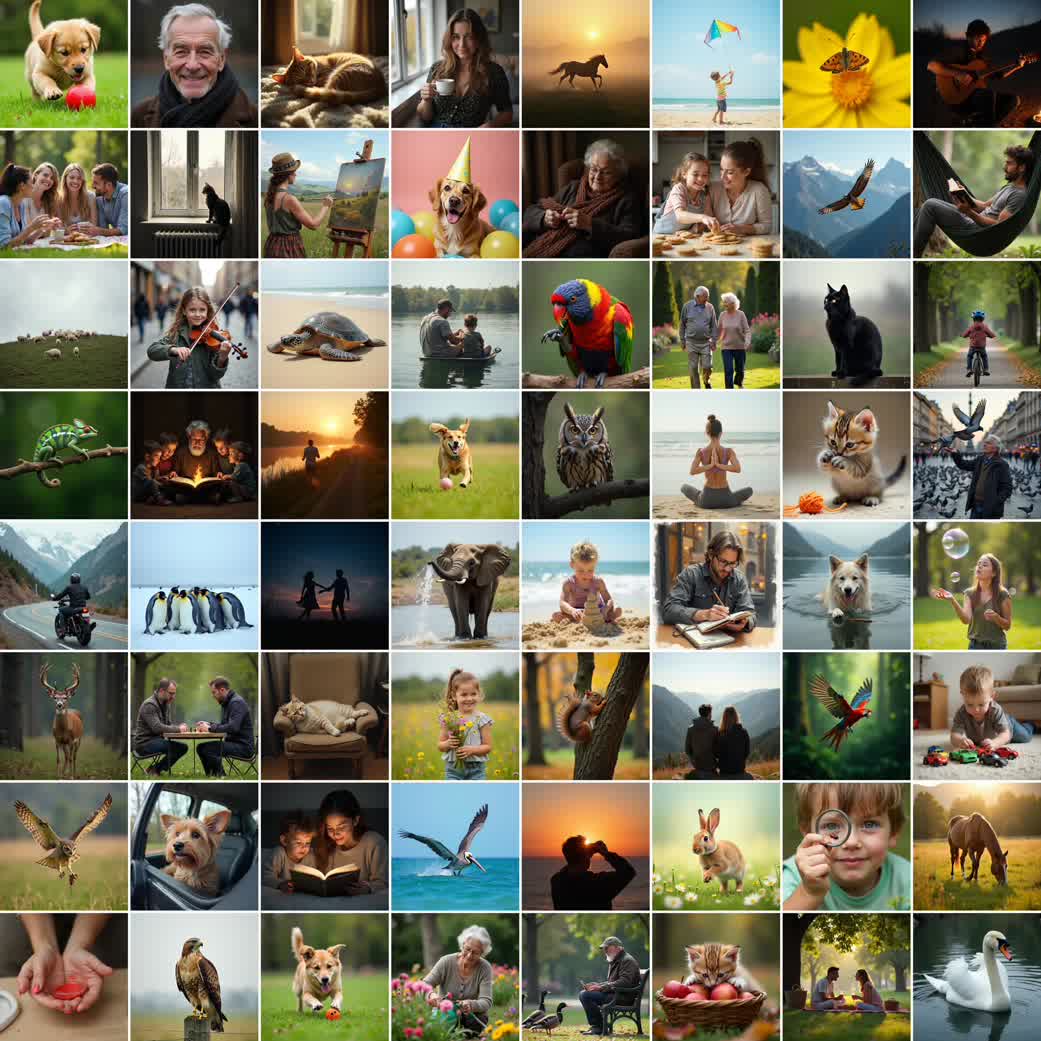}
        \vspace{3px}
        \\

        $G_{\textit{ref}}$ &
        \nonvital{$G_{8}$} &
        \nonvital{$G_{9}$}
        \vspace{15px}
        \\

        \includegraphics[valign=c, width=\ww, frame]{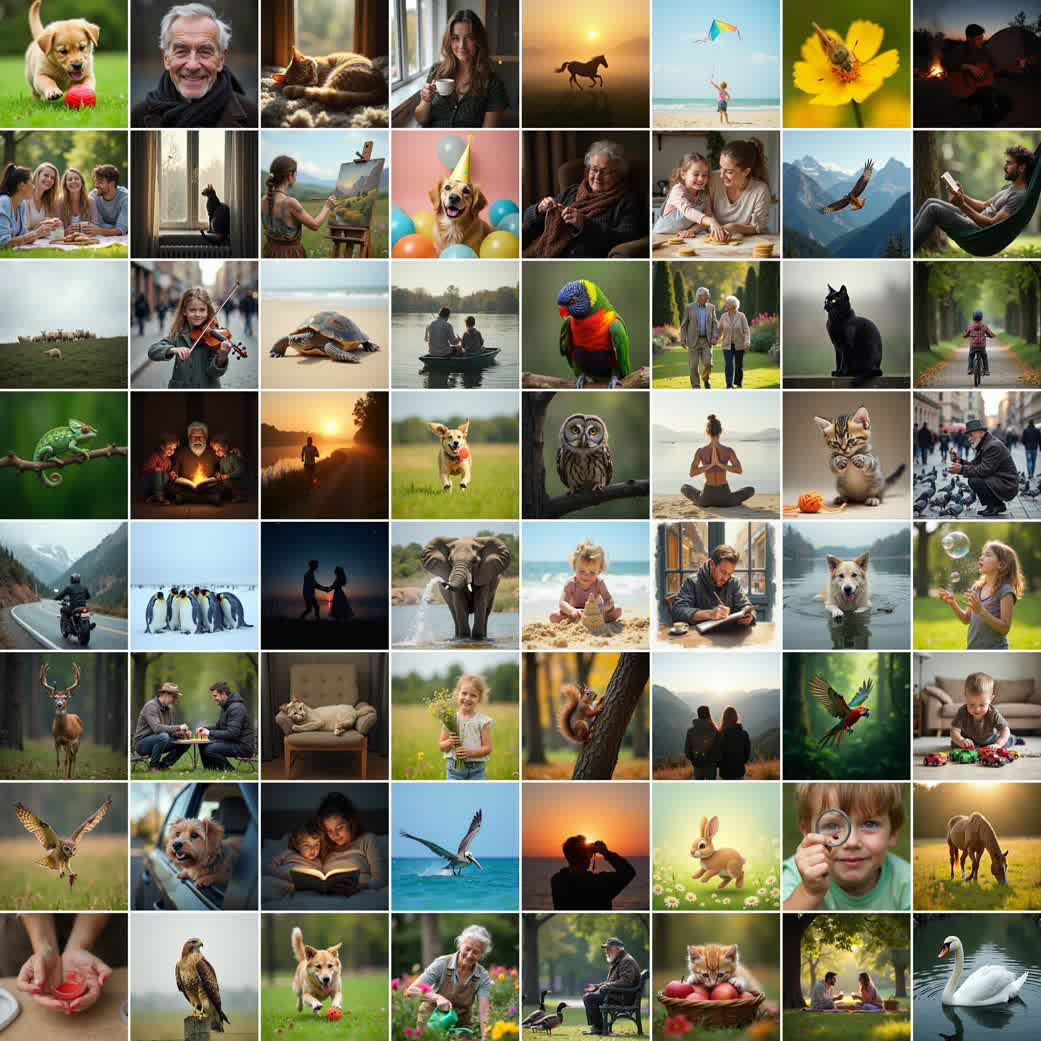} &
        \includegraphics[valign=c, width=\ww, frame]{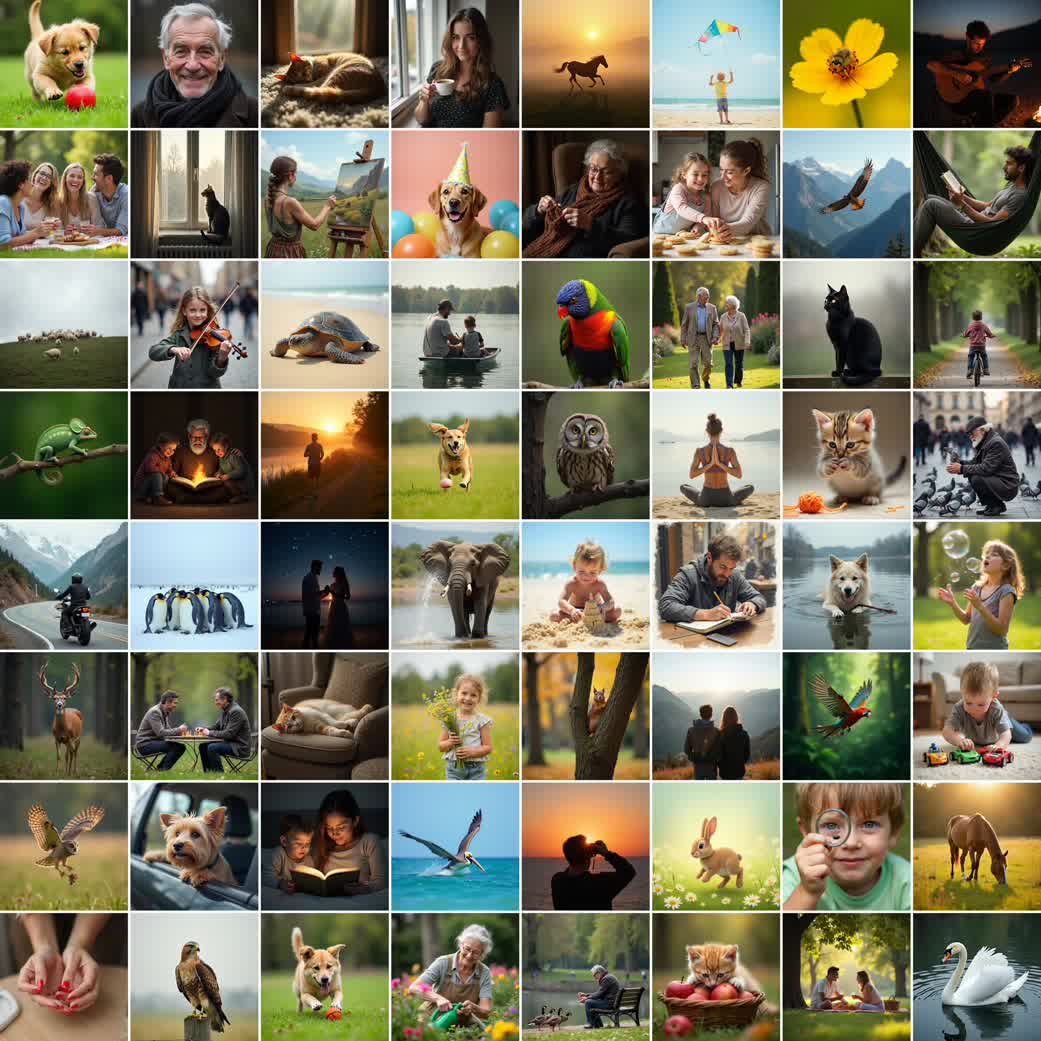} &
        \includegraphics[valign=c, width=\ww, frame]{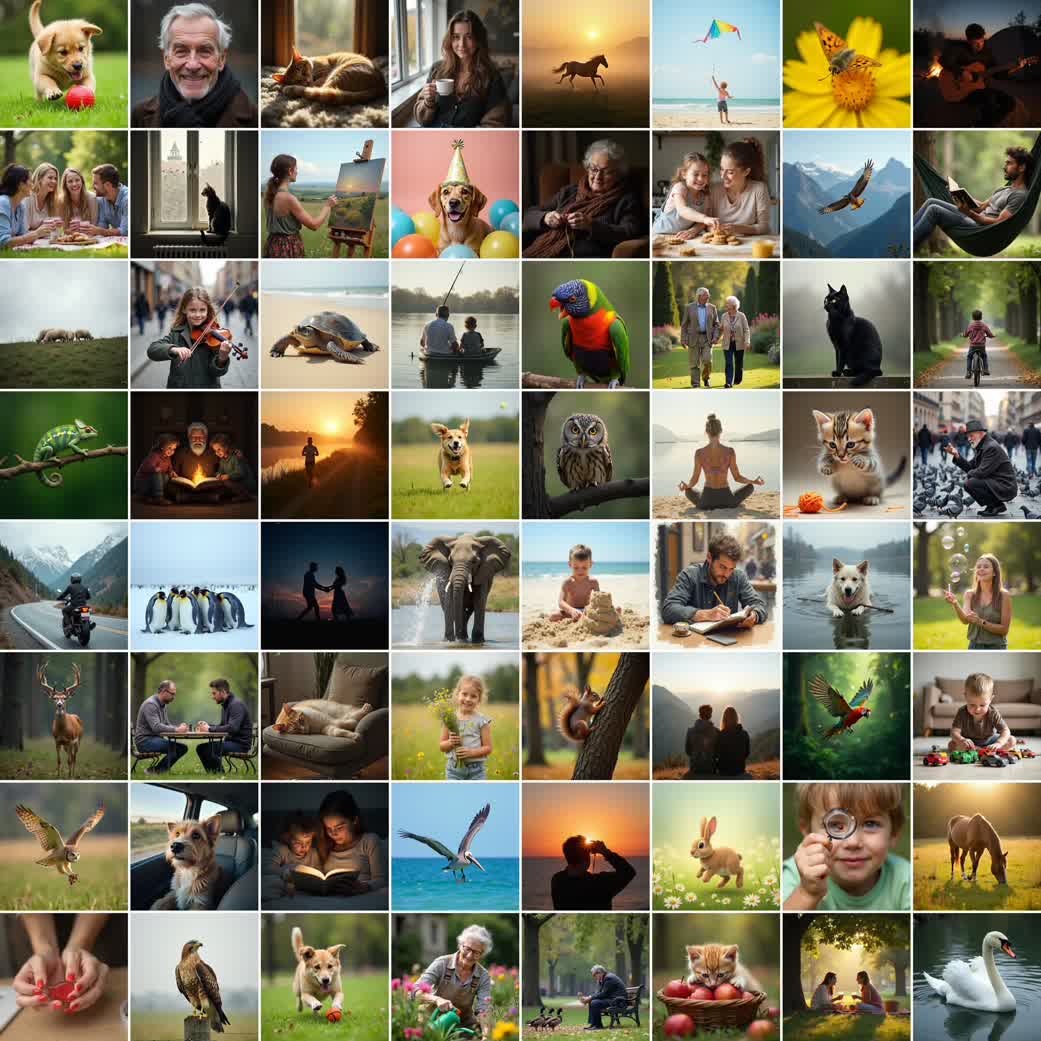}
        \vspace{3px}
        \\

        \nonvital{$G_{10}$} &
        \nonvital{$G_{11}$} &
        \nonvital{$G_{12}$}
        \vspace{15px}
        \\

        \includegraphics[valign=c, width=\ww, frame]{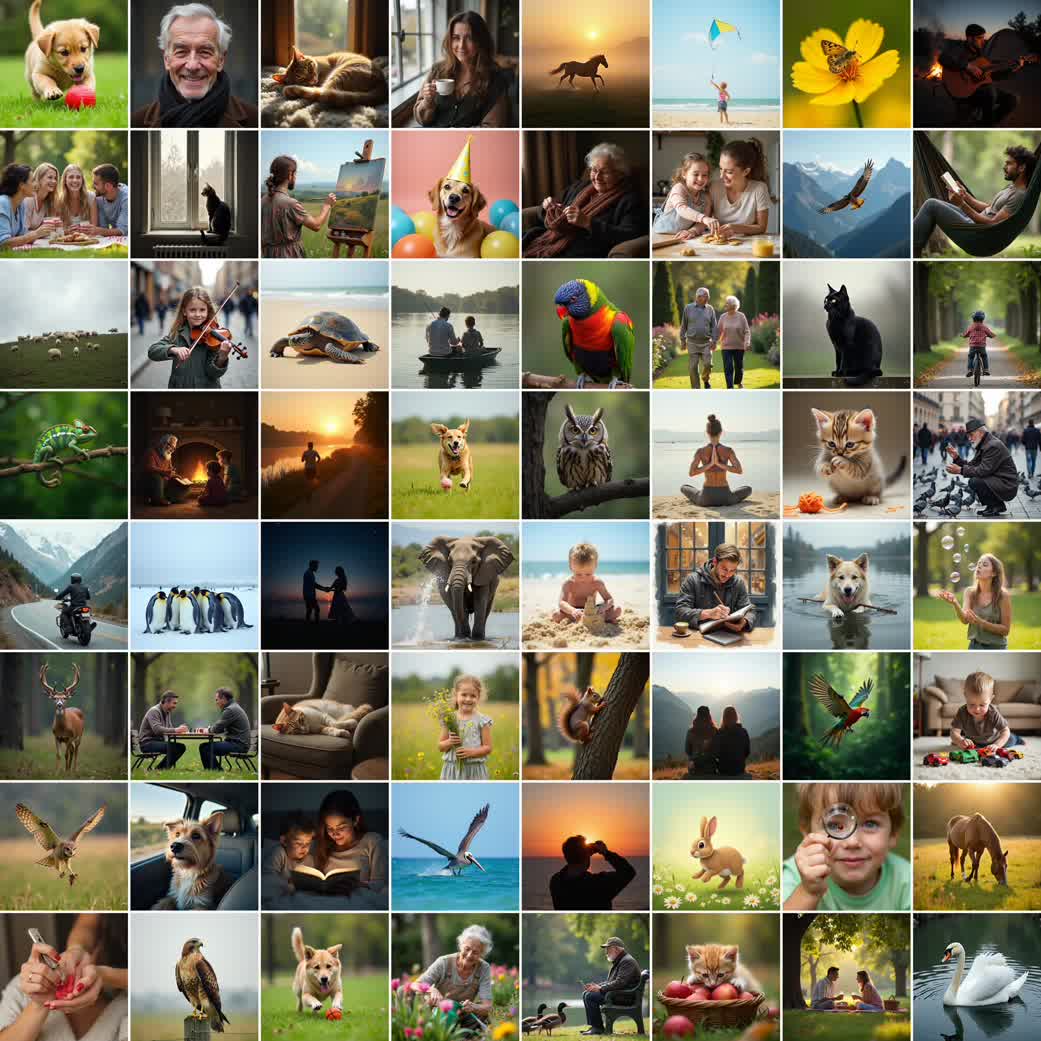} &
        \includegraphics[valign=c, width=\ww, frame]{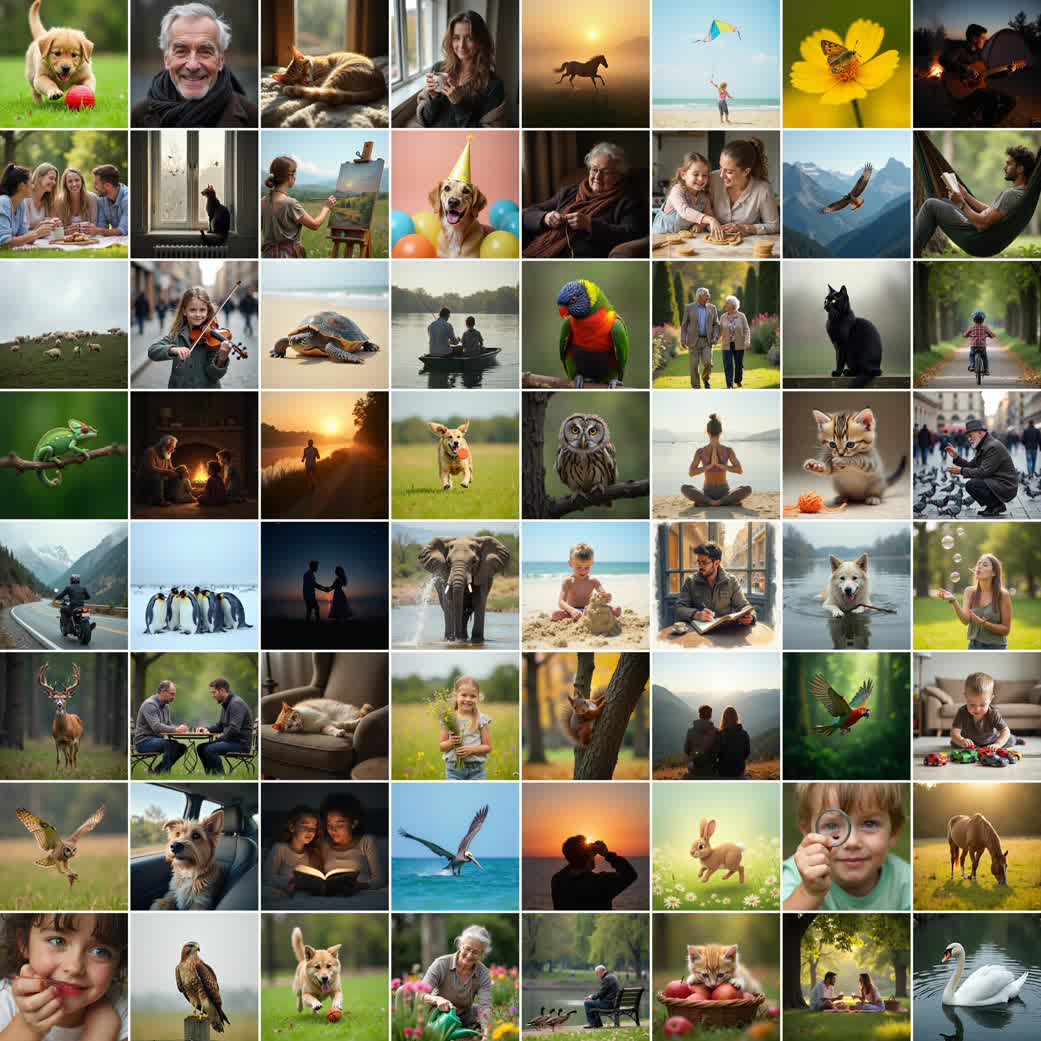} &
        \includegraphics[valign=c, width=\ww, frame]{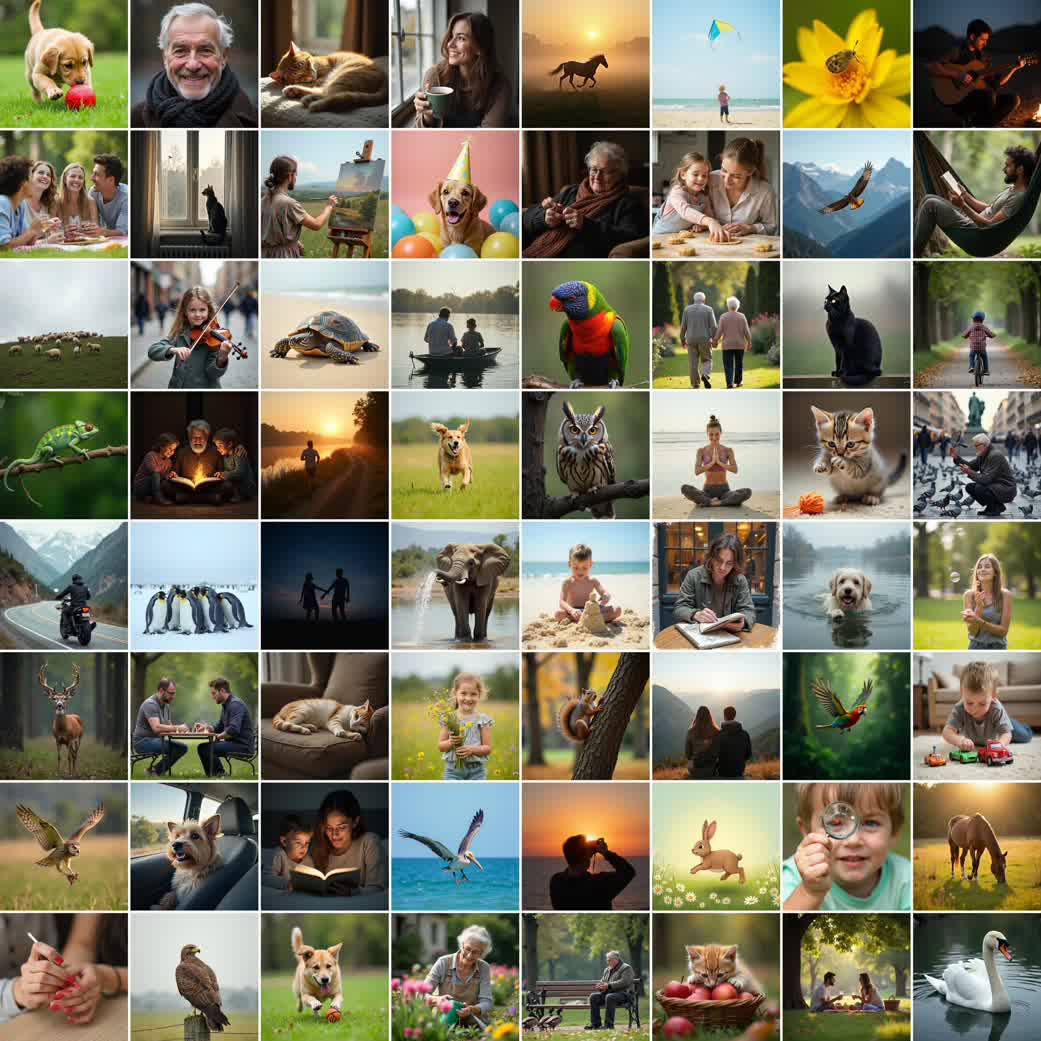}
        \vspace{3px}
        \\

        \nonvital{$G_{13}$} &
        \nonvital{$G_{14}$} &
        \nonvital{$G_{15}$}
        \vspace{3px}
        \\

    \end{tabular}
    \caption{\textbf{Full Layer Bypassing Visualization for Flux.} We visualize the individual layer bypassing study we conducted, as described in \Cref{sec:layer_bypassing_visualization}. We start by generating a set of images $G_{\textit{ref}}$ using a fixed set of seeds and prompts. Then, we bypass each layer $\ell$ by using its residual connection and generate the set of images $G_{\ell}$ using the same fixed set of prompts and seeds. In this visualization, \nonvital{$G_8$} -- \nonvital{$G_{15}$} are all \nonvital{non-vital layers}.}
    \label{fig:full_flux_bypassing_2}
\end{figure*}

\begin{figure*}[tp]
    \centering
    \setlength{\tabcolsep}{2.5pt}
    \renewcommand{\arraystretch}{1.0}
    \setlength{\ww}{0.32\linewidth}
    \begin{tabular}{ccc}

        \includegraphics[valign=c, width=\ww, frame]{figures/full_bypassing_visualization_flux/assets/src.jpg} &
        \includegraphics[valign=c, width=\ww, frame]{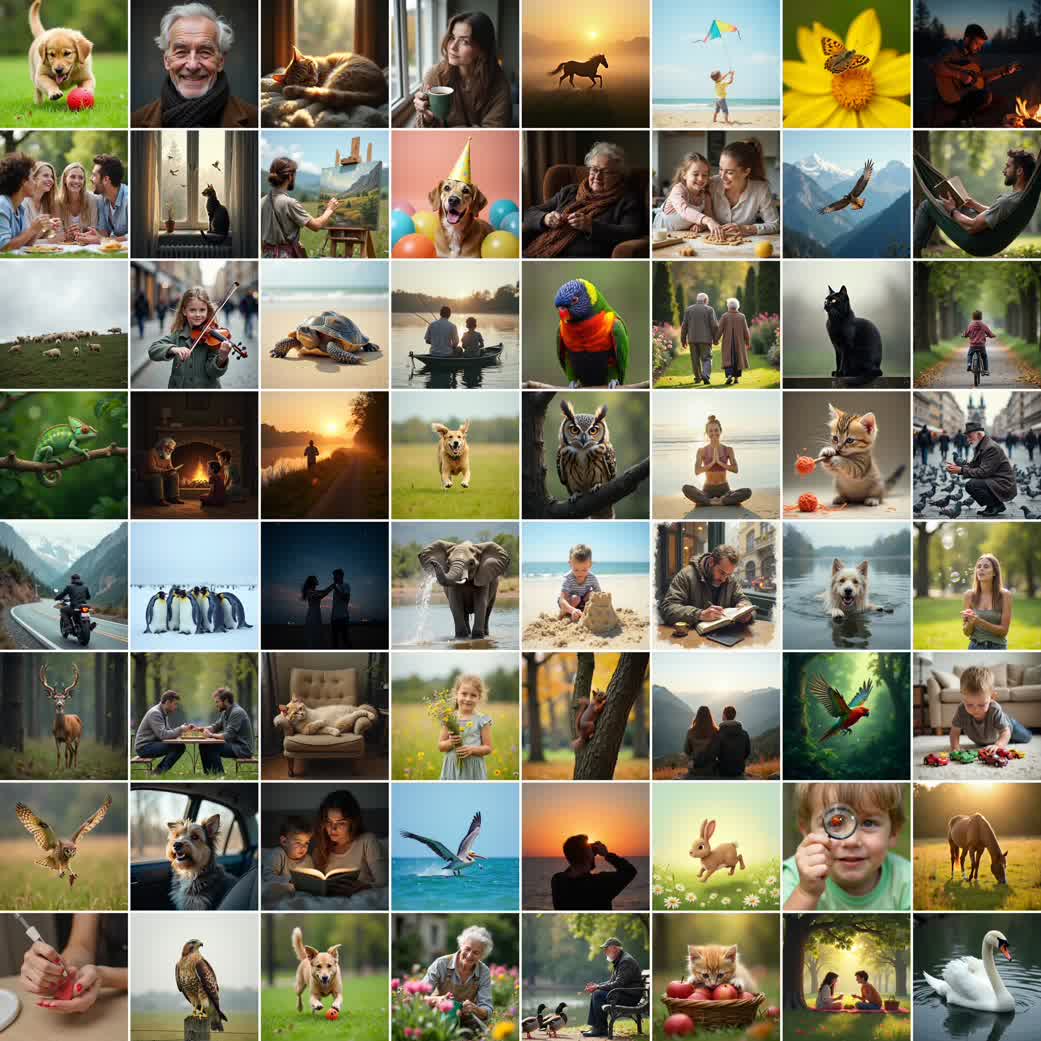} &
        \includegraphics[valign=c, width=\ww, frame]{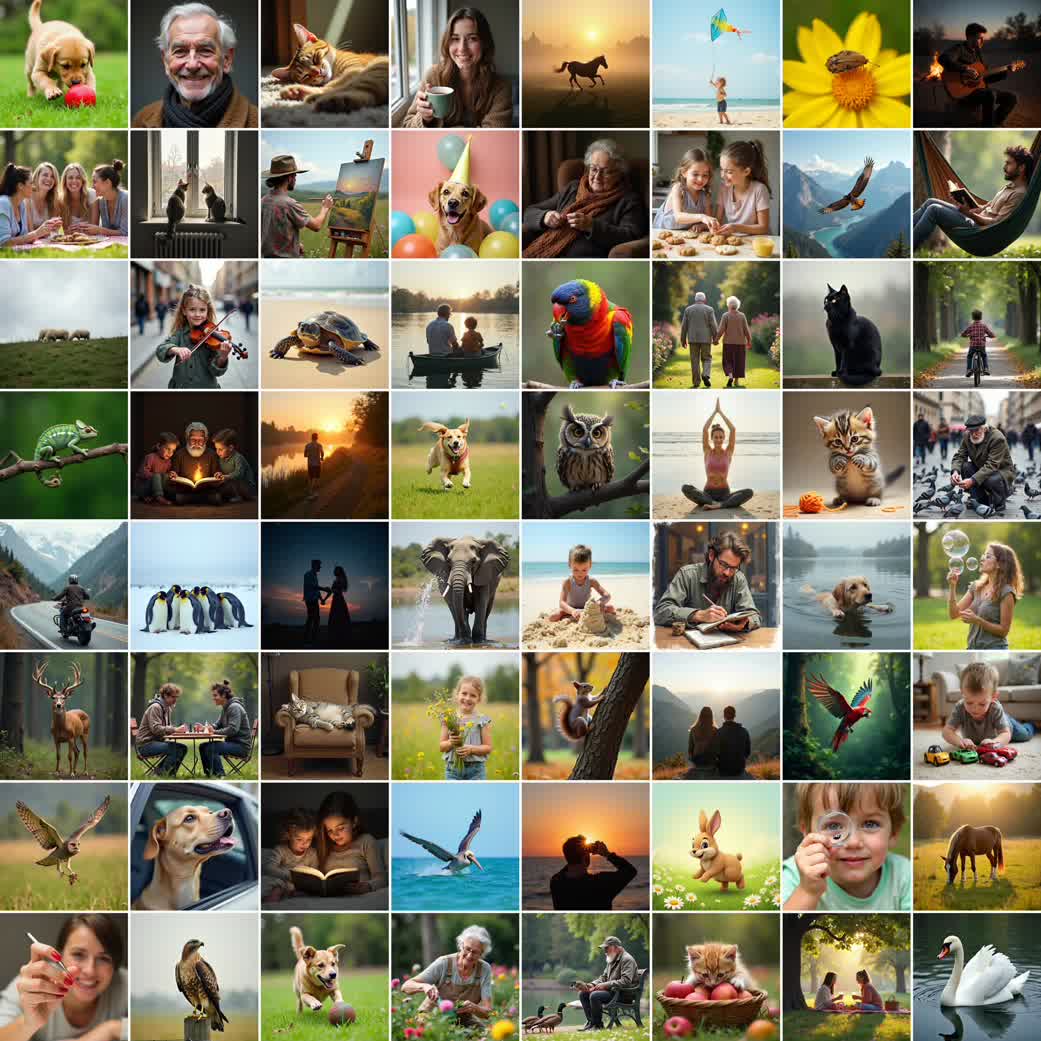}
        \vspace{3px}
        \\

        $G_{\textit{ref}}$ &
        \nonvital{$G_{16}$} &
        \vital{$G_{17}$}
        \vspace{15px}
        \\

        \includegraphics[valign=c, width=\ww, frame]{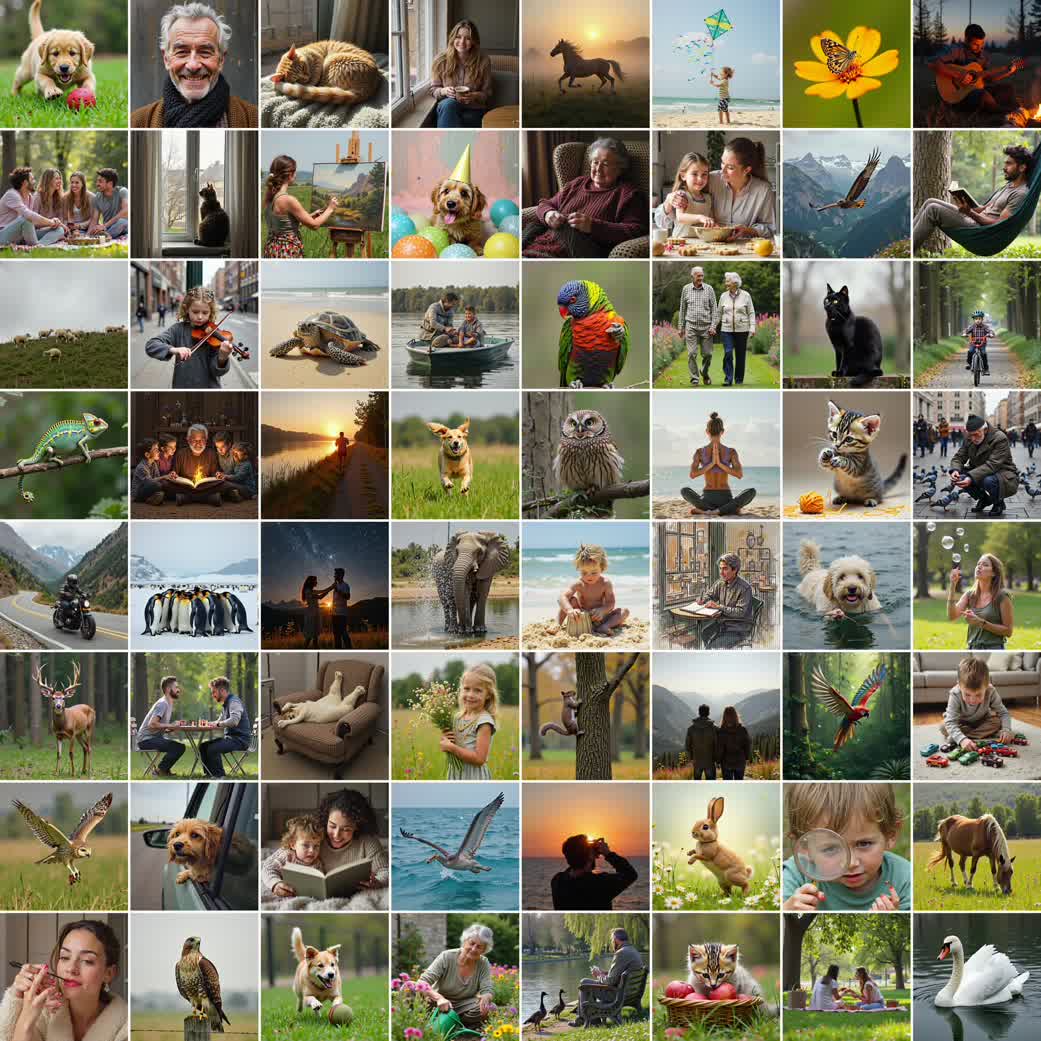} &
        \includegraphics[valign=c, width=\ww, frame]{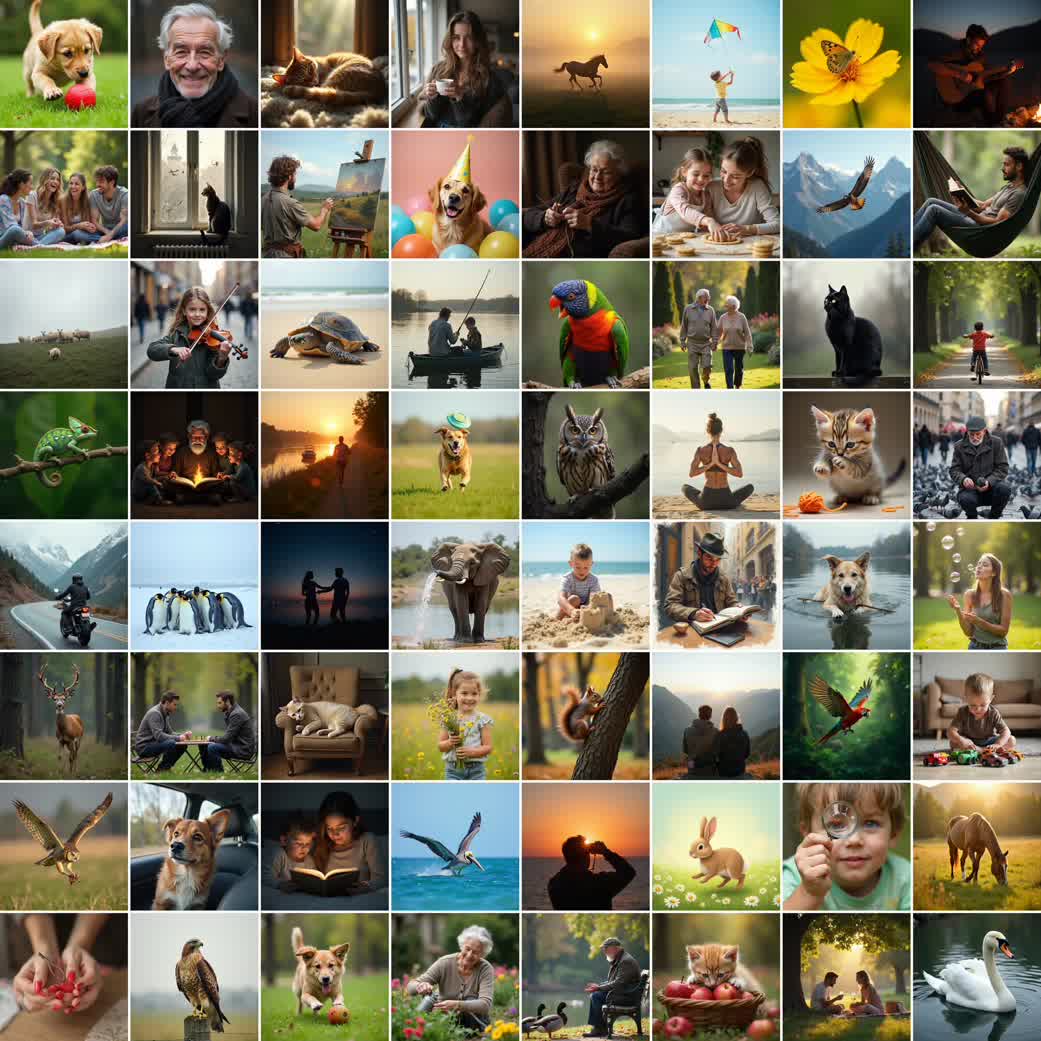} &
        \includegraphics[valign=c, width=\ww, frame]{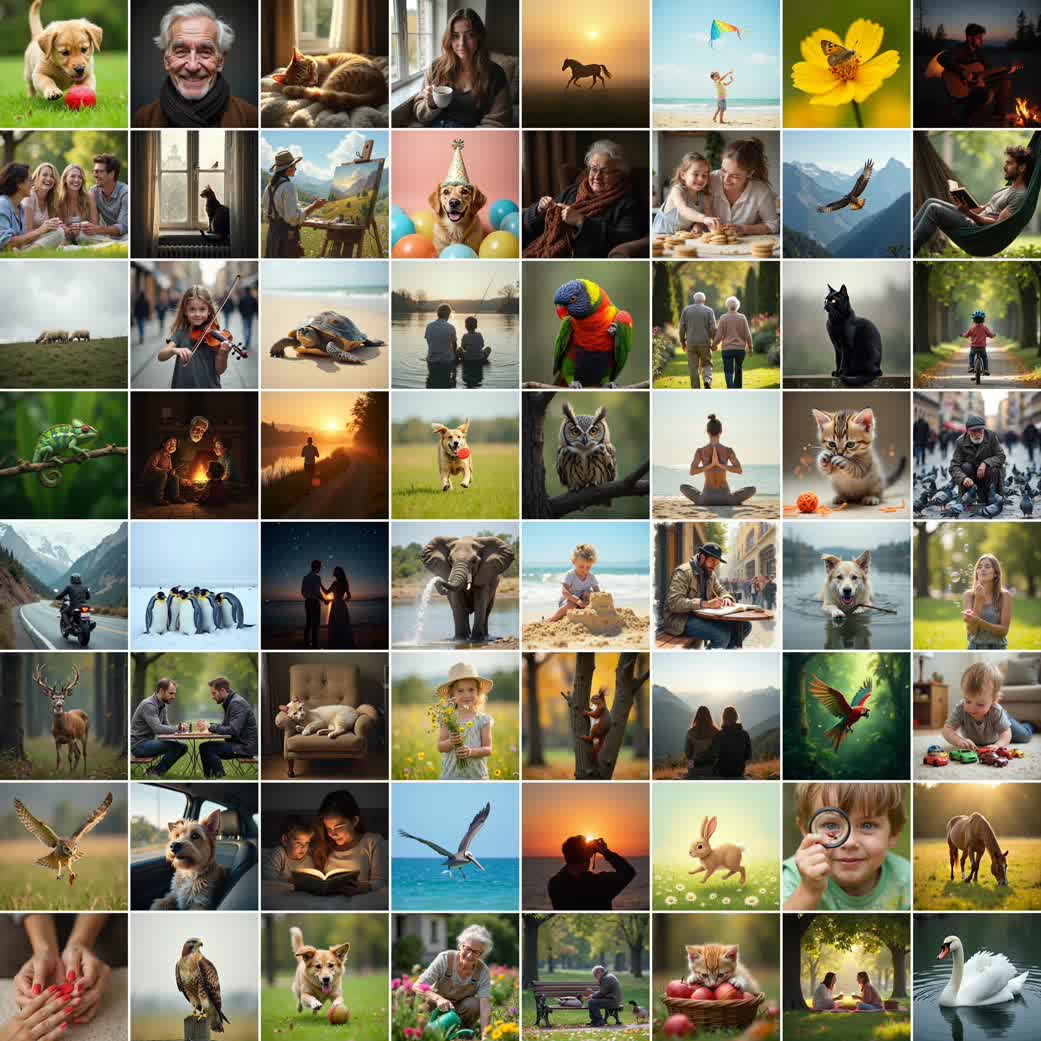}
        \vspace{3px}
        \\

        \vital{$G_{18}$} &
        \nonvital{$G_{19}$} &
        \nonvital{$G_{20}$}
        \vspace{15px}
        \\

        \includegraphics[valign=c, width=\ww, frame]{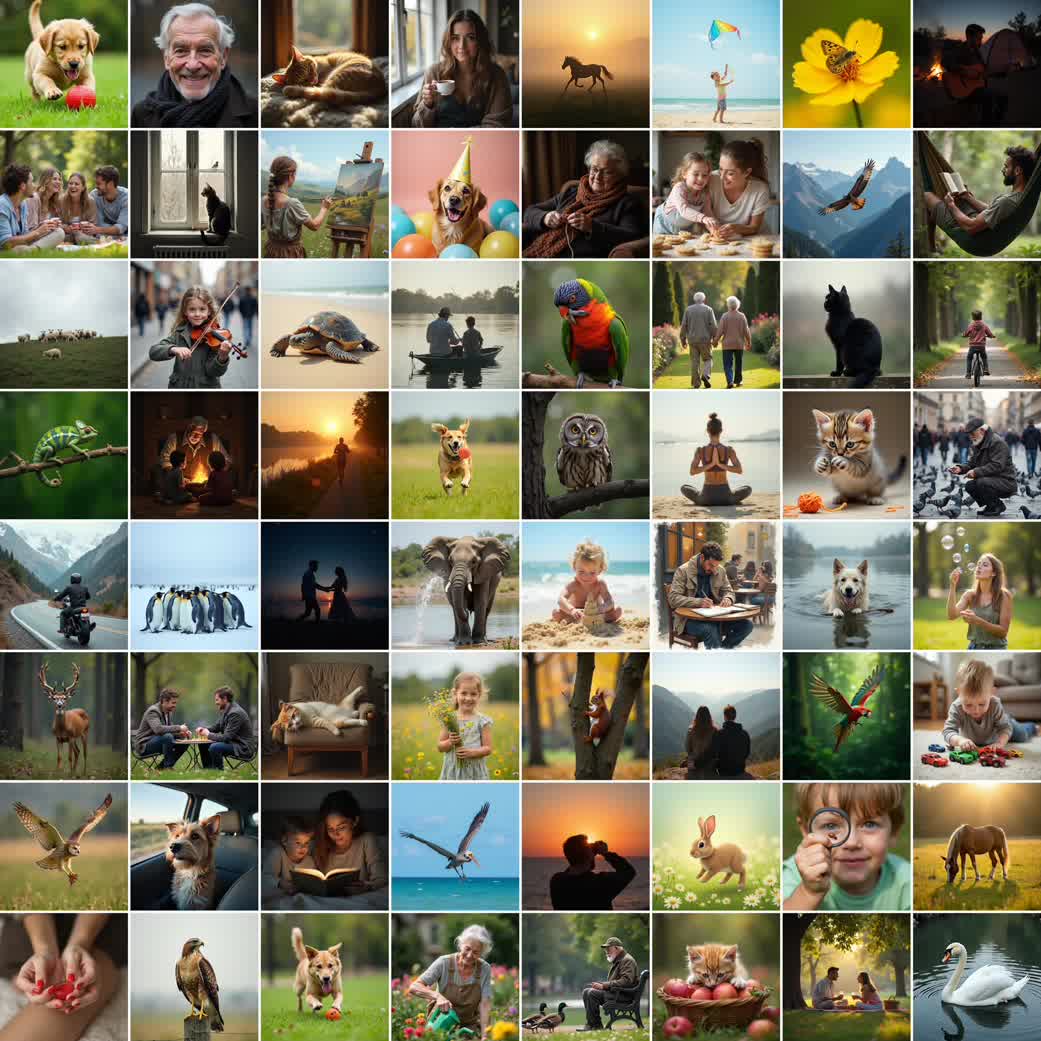} &
        \includegraphics[valign=c, width=\ww, frame]{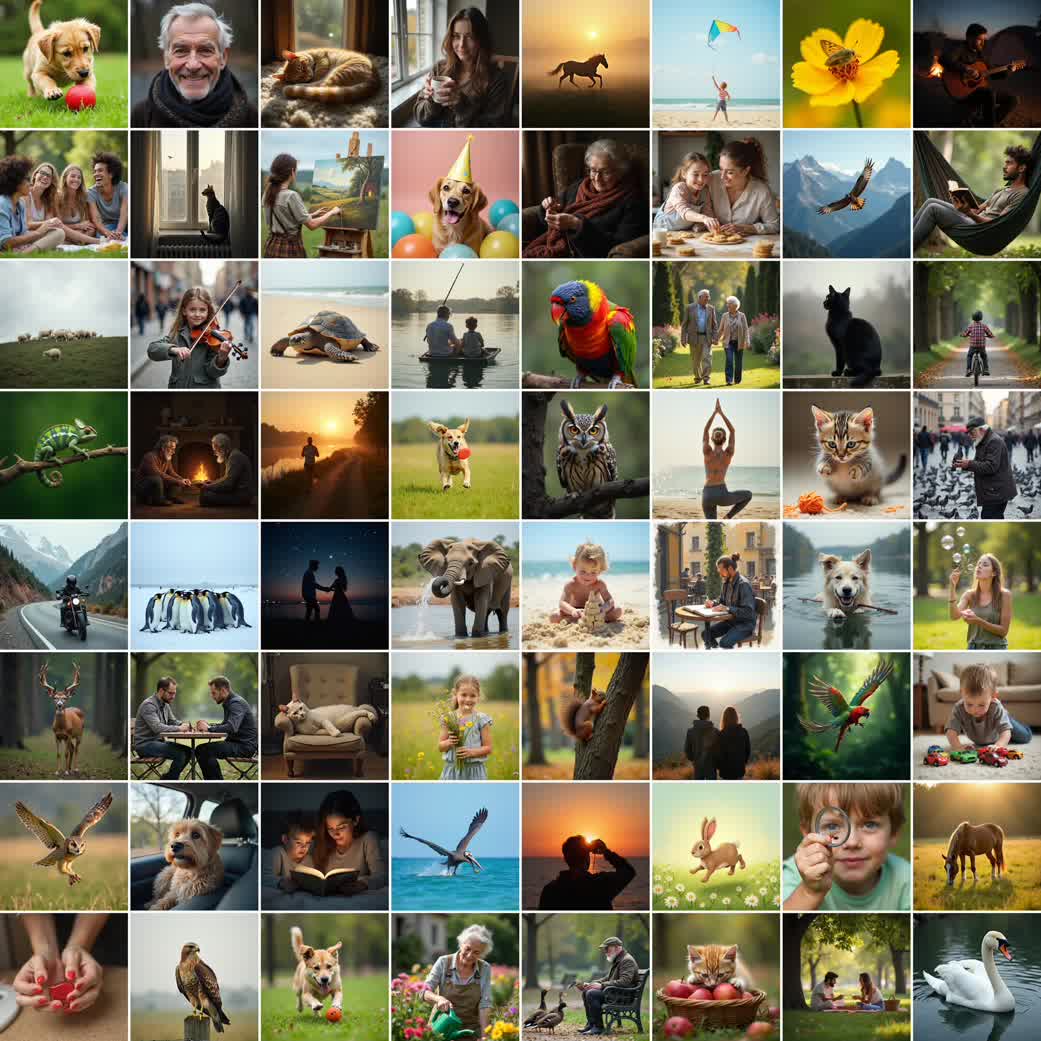} &
        \includegraphics[valign=c, width=\ww, frame]{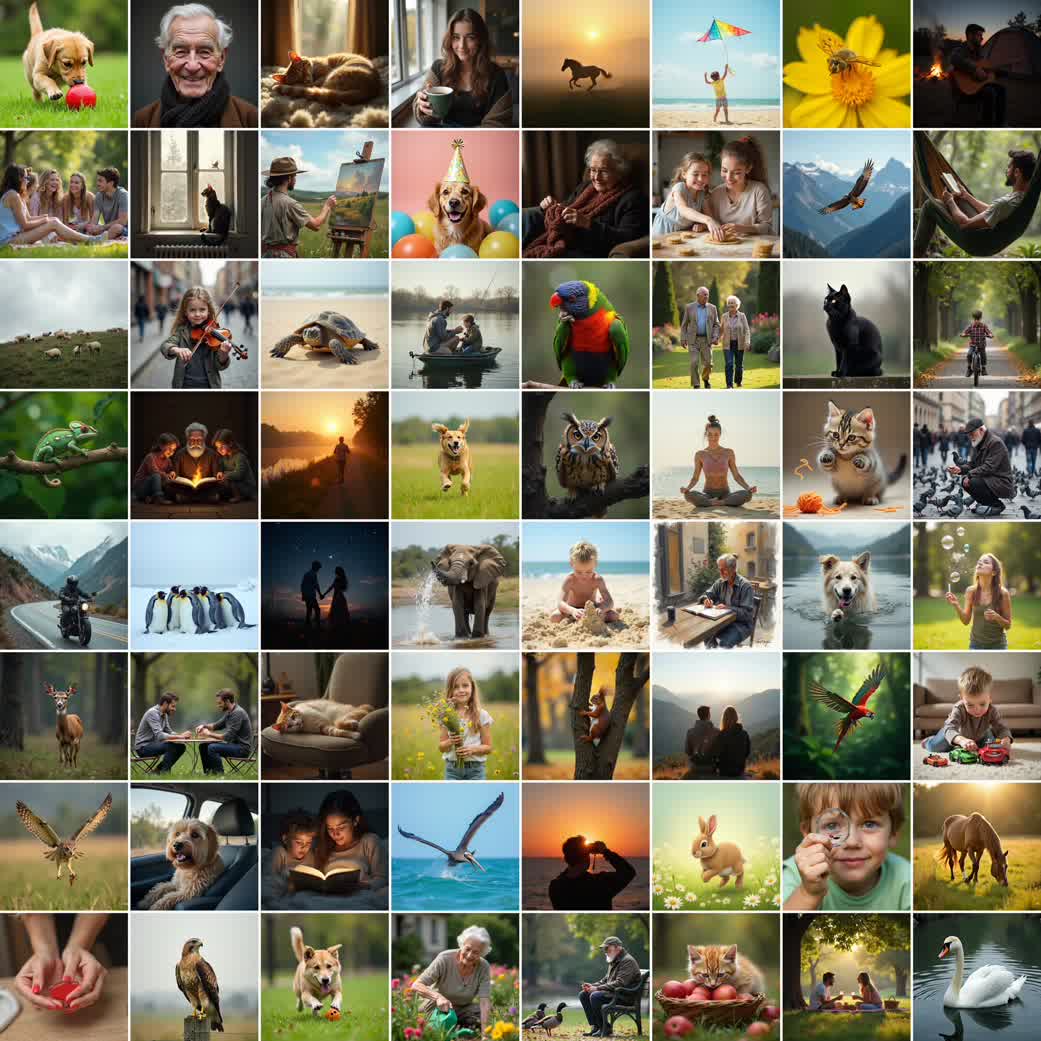}
        \vspace{3px}
        \\

        \nonvital{$G_{21}$} &
        \nonvital{$G_{22}$} &
        \nonvital{$G_{23}$}
        \vspace{3px}
        \\

    \end{tabular}
    \caption{\textbf{Full Layer Bypassing Visualization for Flux.} We visualize the individual layer bypassing study we conducted, as described in \Cref{sec:layer_bypassing_visualization}. We start by generating a set of images $G_{\textit{ref}}$ using a fixed set of seeds and prompts. Then, we bypass each layer $\ell$ by using its residual connection and generate the set of images $G_{\ell}$ using the same fixed set of prompts and seeds. In this visualization, \vital{$G_{17}$} and \vital{$G_{18}$} are \vital{vital layers}, while \nonvital{$G_{16}$} and \nonvital{$G_{19}$} -- \nonvital{$G_{23}$} are \nonvital{non-vital layers}.}
    \label{fig:full_flux_bypassing_3}
\end{figure*}

\begin{figure*}[tp]
    \centering
    \setlength{\tabcolsep}{2.5pt}
    \renewcommand{\arraystretch}{1.0}
    \setlength{\ww}{0.32\linewidth}
    \begin{tabular}{ccc}

        \includegraphics[valign=c, width=\ww, frame]{figures/full_bypassing_visualization_flux/assets/src.jpg} &
        \includegraphics[valign=c, width=\ww, frame]{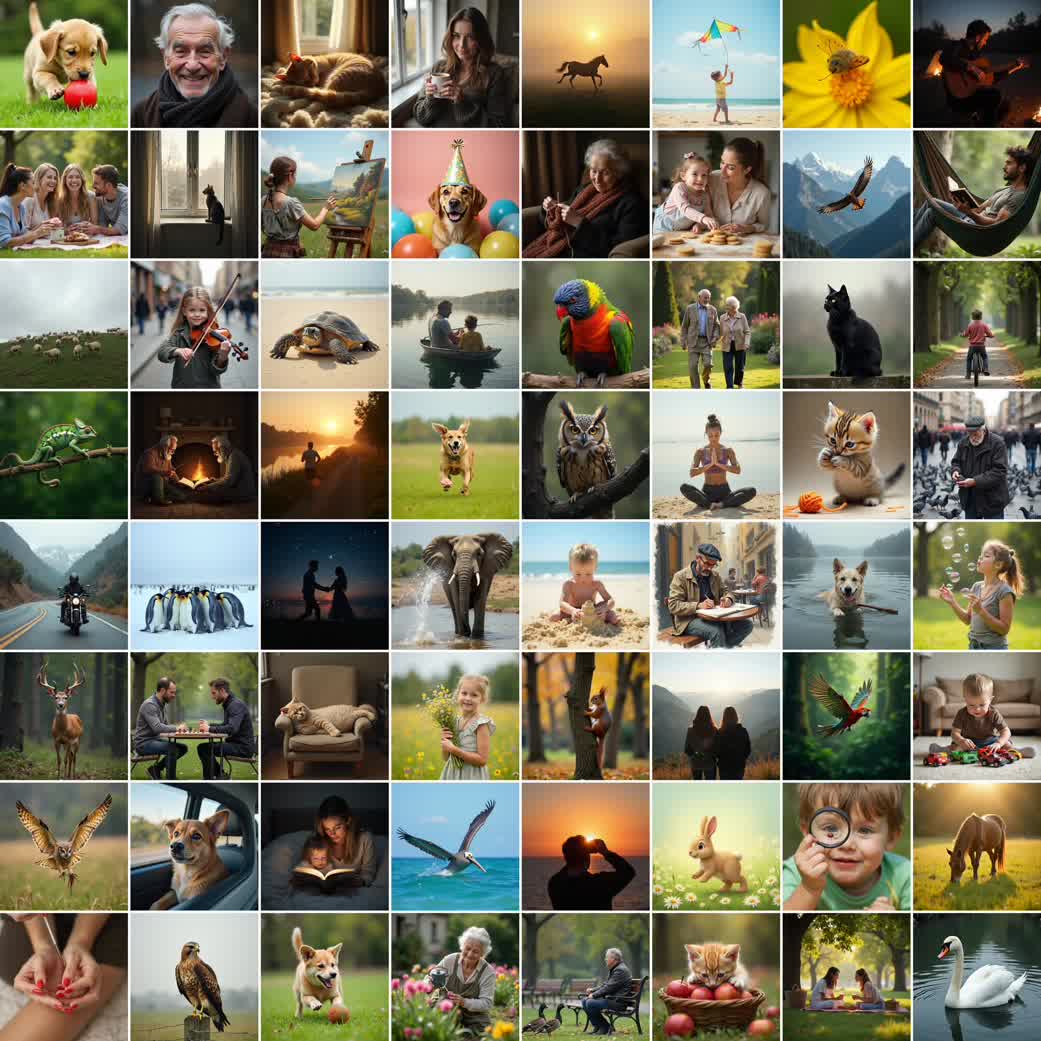} &
        \includegraphics[valign=c, width=\ww, frame]{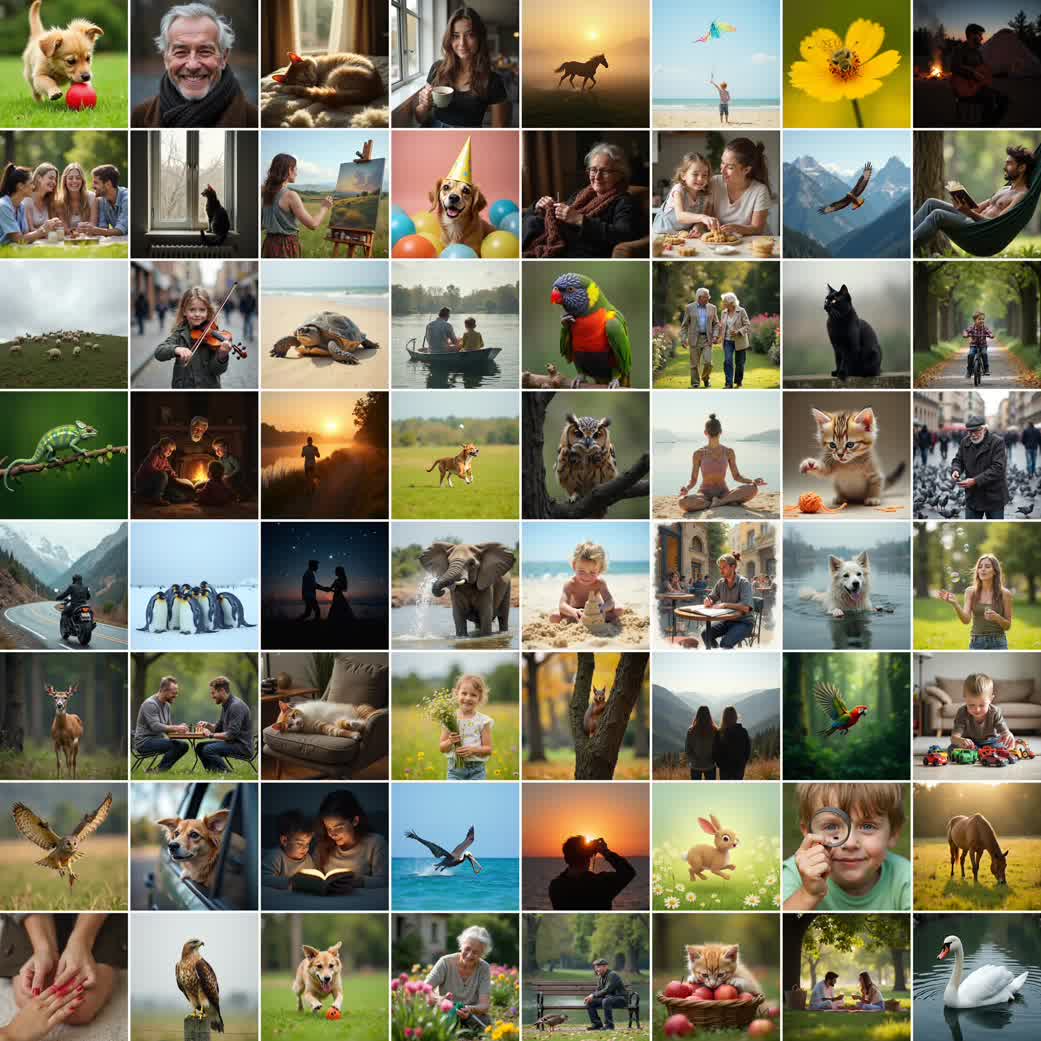}
        \vspace{3px}
        \\

        $G_{\textit{ref}}$ &
        \nonvital{$G_{24}$} &
        \vital{$G_{25}$}
        \vspace{15px}
        \\

        \includegraphics[valign=c, width=\ww, frame]{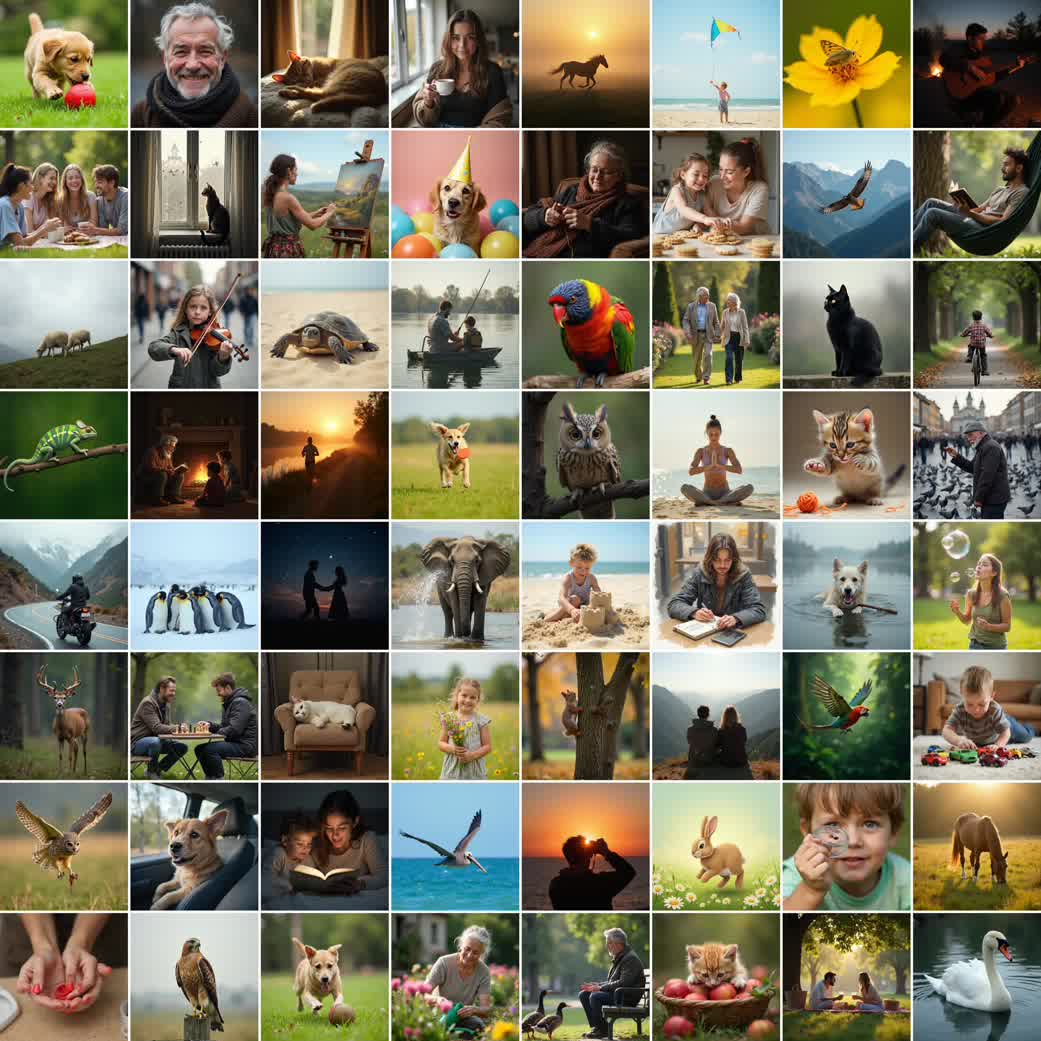} &
        \includegraphics[valign=c, width=\ww, frame]{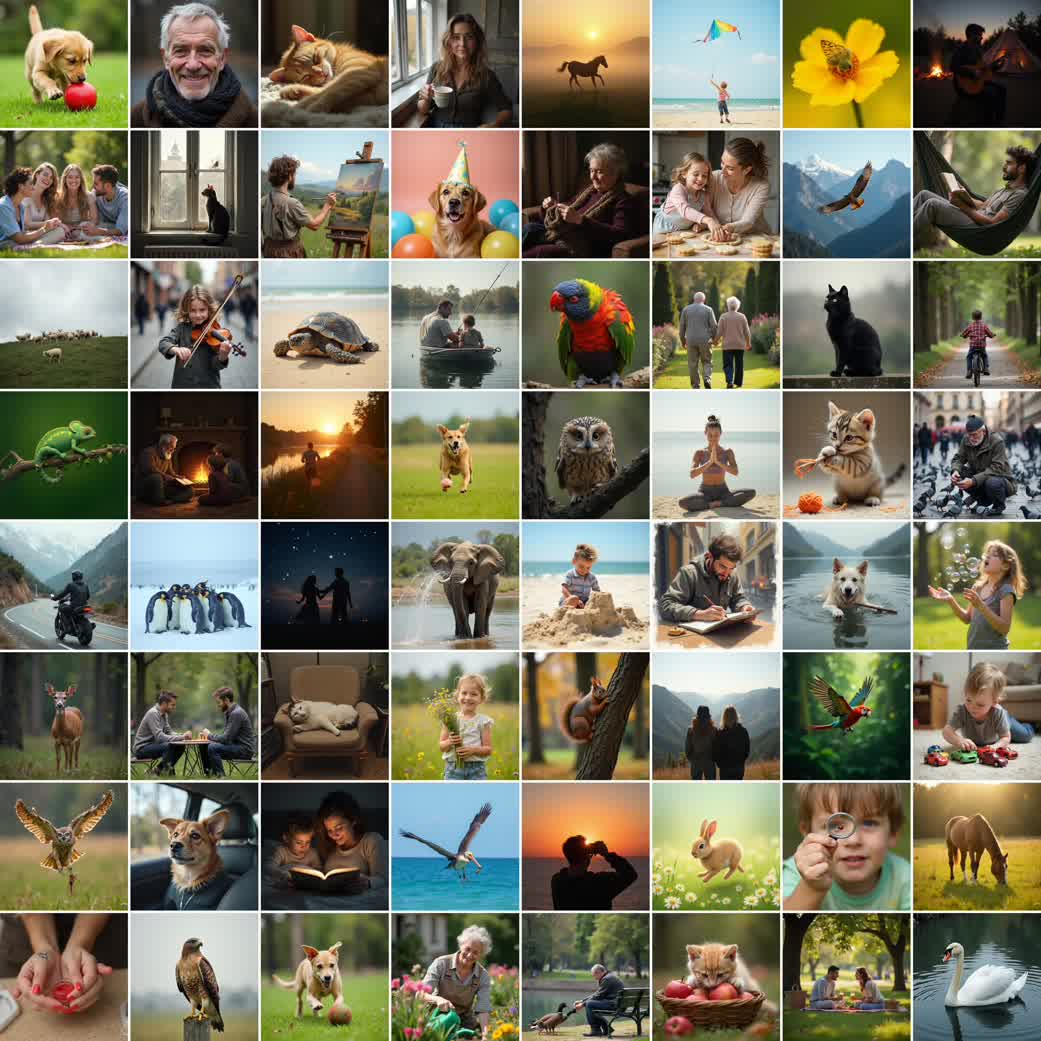} &
        \includegraphics[valign=c, width=\ww, frame]{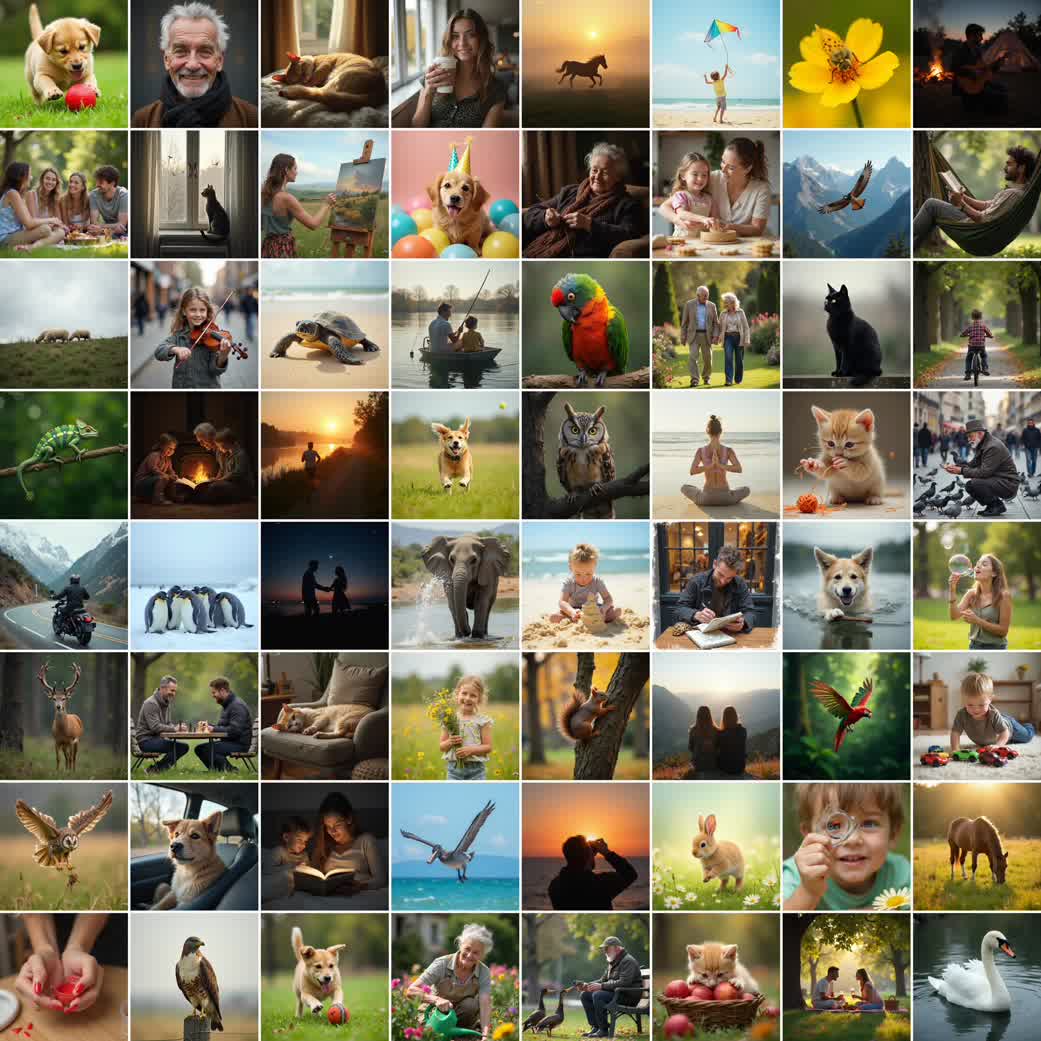}
        \vspace{3px}
        \\

        \nonvital{$G_{26}$} &
        \nonvital{$G_{27}$} &
        \vital{$G_{28}$}
        \vspace{15px}
        \\

        \includegraphics[valign=c, width=\ww, frame]{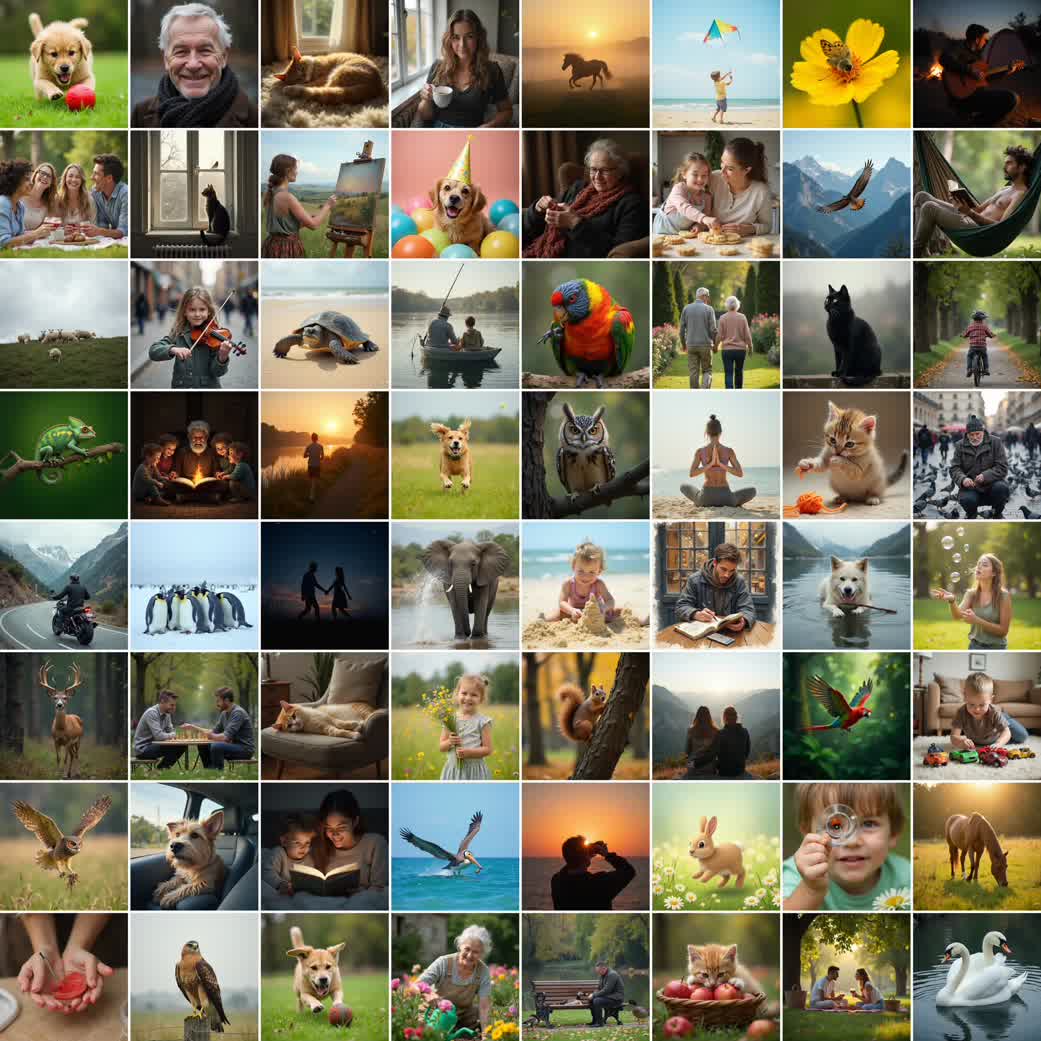} &
        \includegraphics[valign=c, width=\ww, frame]{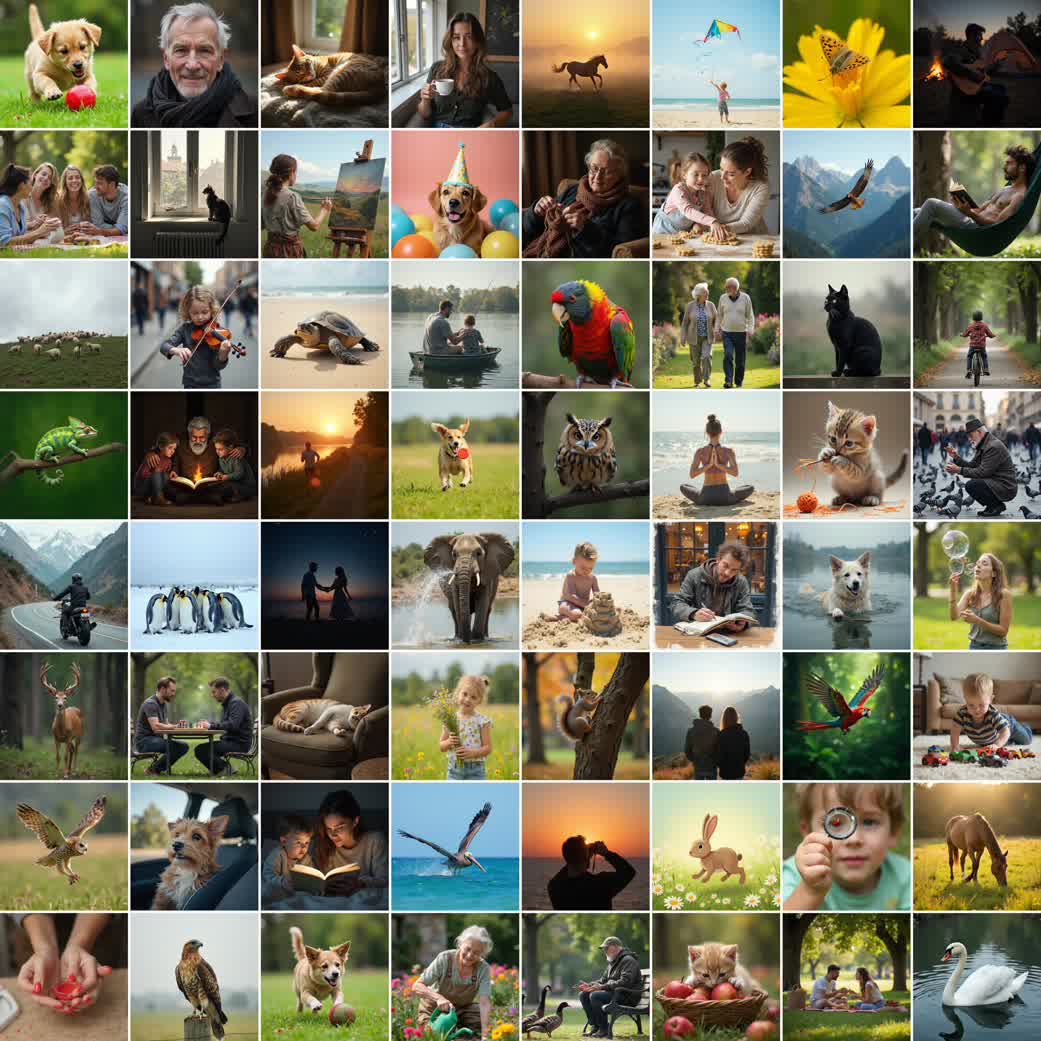} &
        \includegraphics[valign=c, width=\ww, frame]{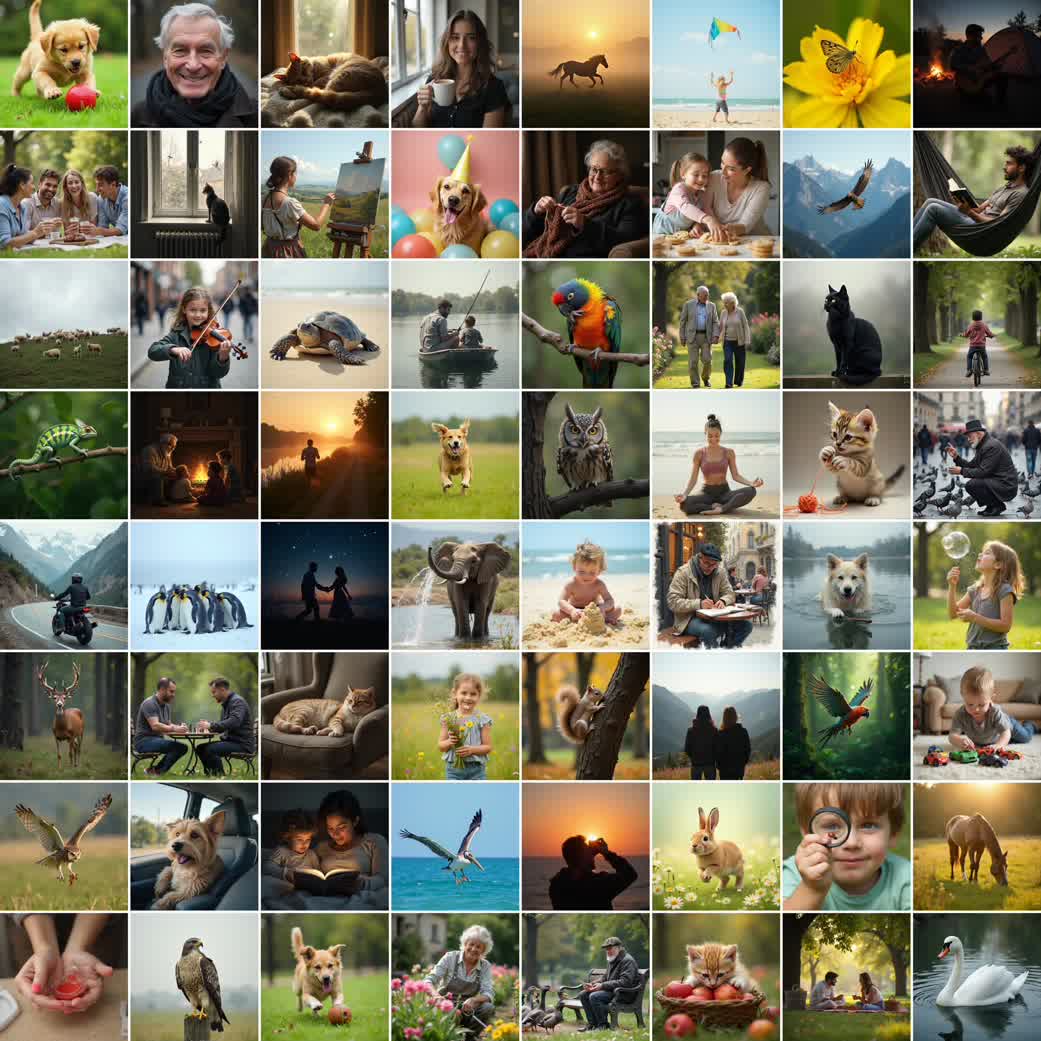}
        \vspace{3px}
        \\

        \nonvital{$G_{29}$} &
        \nonvital{$G_{30}$} &
        \nonvital{$G_{31}$}
        \vspace{3px}
        \\

    \end{tabular}
    \caption{\textbf{Full Layer Bypassing Visualization for Flux.} We visualize the individual layer bypassing study we conducted, as described in \Cref{sec:layer_bypassing_visualization}. We start by generating a set of images $G_{\textit{ref}}$ using a fixed set of seeds and prompts. Then, we bypass each layer $\ell$ by using its residual connection and generate the set of images $G_{\ell}$ using the same fixed set of prompts and seeds. In this visualization, \vital{$G_{25}$} and \vital{$G_{28}$} are \vital{vital layers}, while \nonvital{$G_{24}$}, \nonvital{$G_{26}$} -- \nonvital{$G_{27}$} and \nonvital{$G_{29}$} -- \nonvital{$G_{31}$} are \nonvital{non-vital layers}.}
    \label{fig:full_flux_bypassing_4}
\end{figure*}

\begin{figure*}[tp]
    \centering
    \setlength{\tabcolsep}{2.5pt}
    \renewcommand{\arraystretch}{1.0}
    \setlength{\ww}{0.32\linewidth}
    \begin{tabular}{ccc}

        \includegraphics[valign=c, width=\ww, frame]{figures/full_bypassing_visualization_flux/assets/src.jpg} &
        \includegraphics[valign=c, width=\ww, frame]{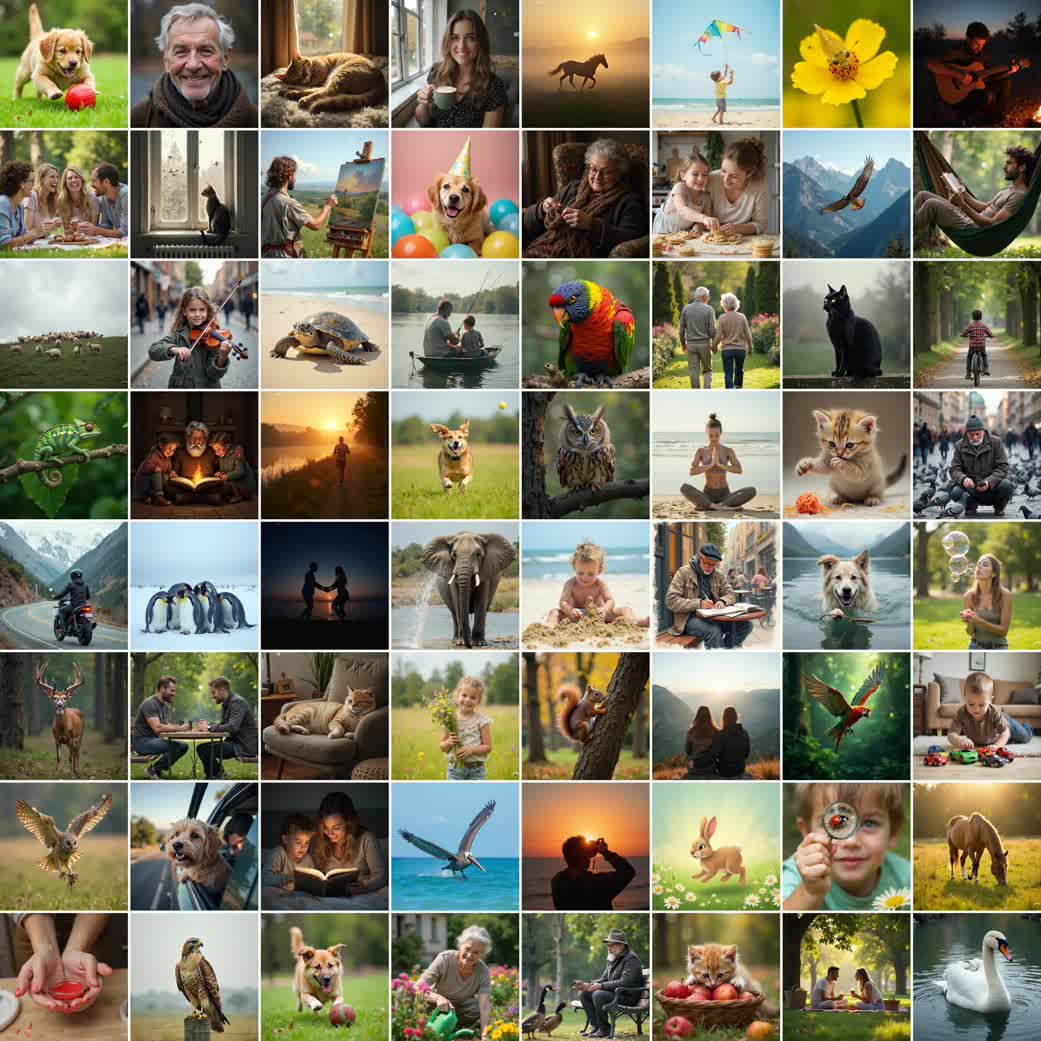} &
        \includegraphics[valign=c, width=\ww, frame]{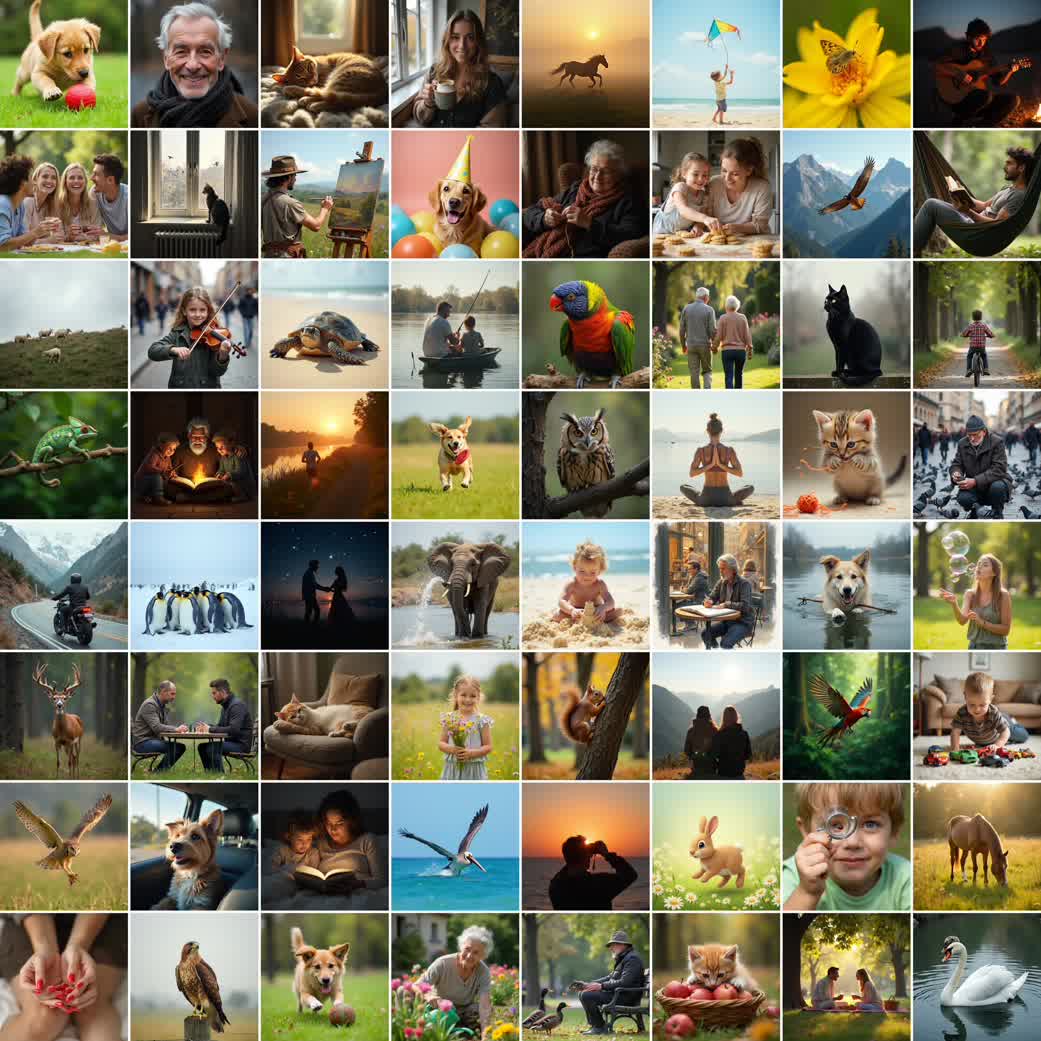}
        \vspace{3px}
        \\

        $G_{\textit{ref}}$ &
        \nonvital{$G_{32}$} &
        \nonvital{$G_{33}$}
        \vspace{15px}
        \\

        \includegraphics[valign=c, width=\ww, frame]{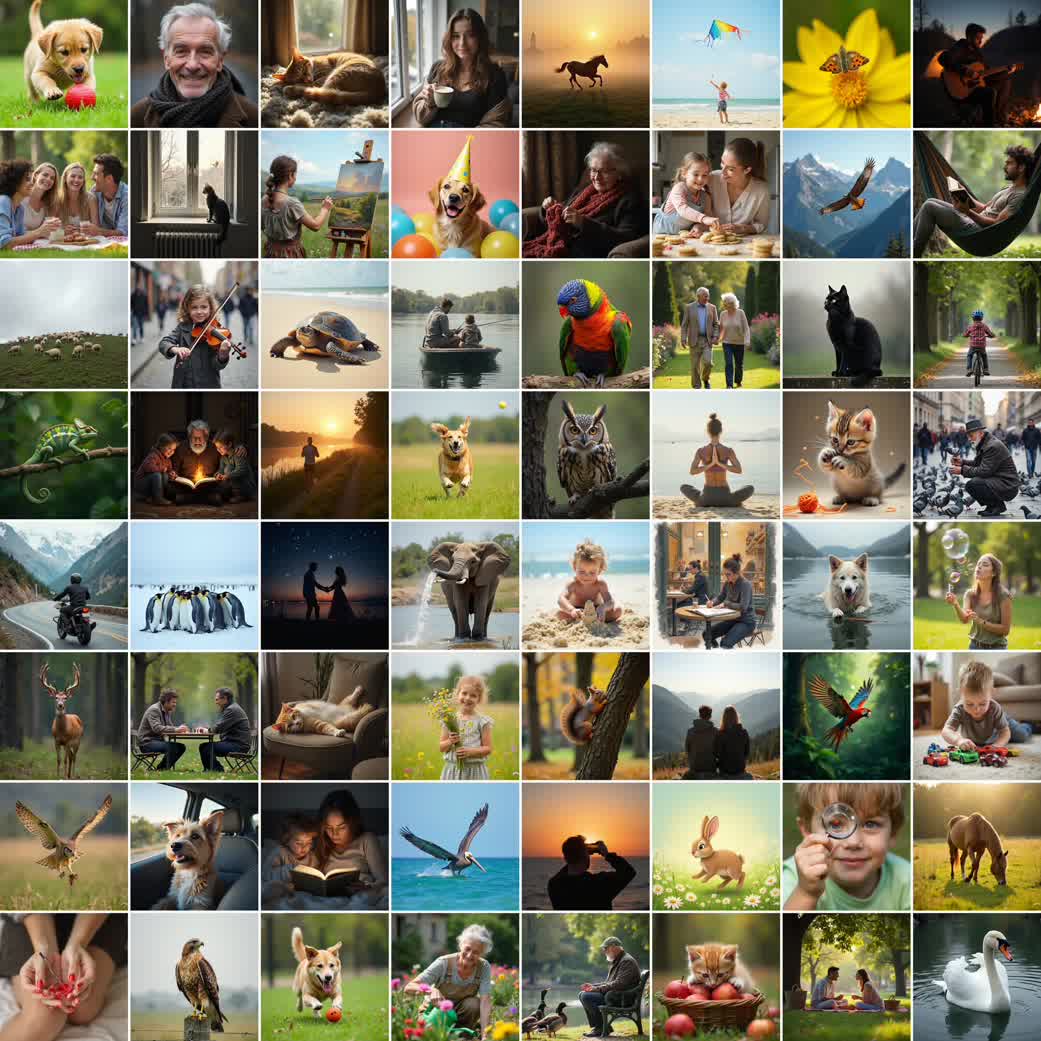} &
        \includegraphics[valign=c, width=\ww, frame]{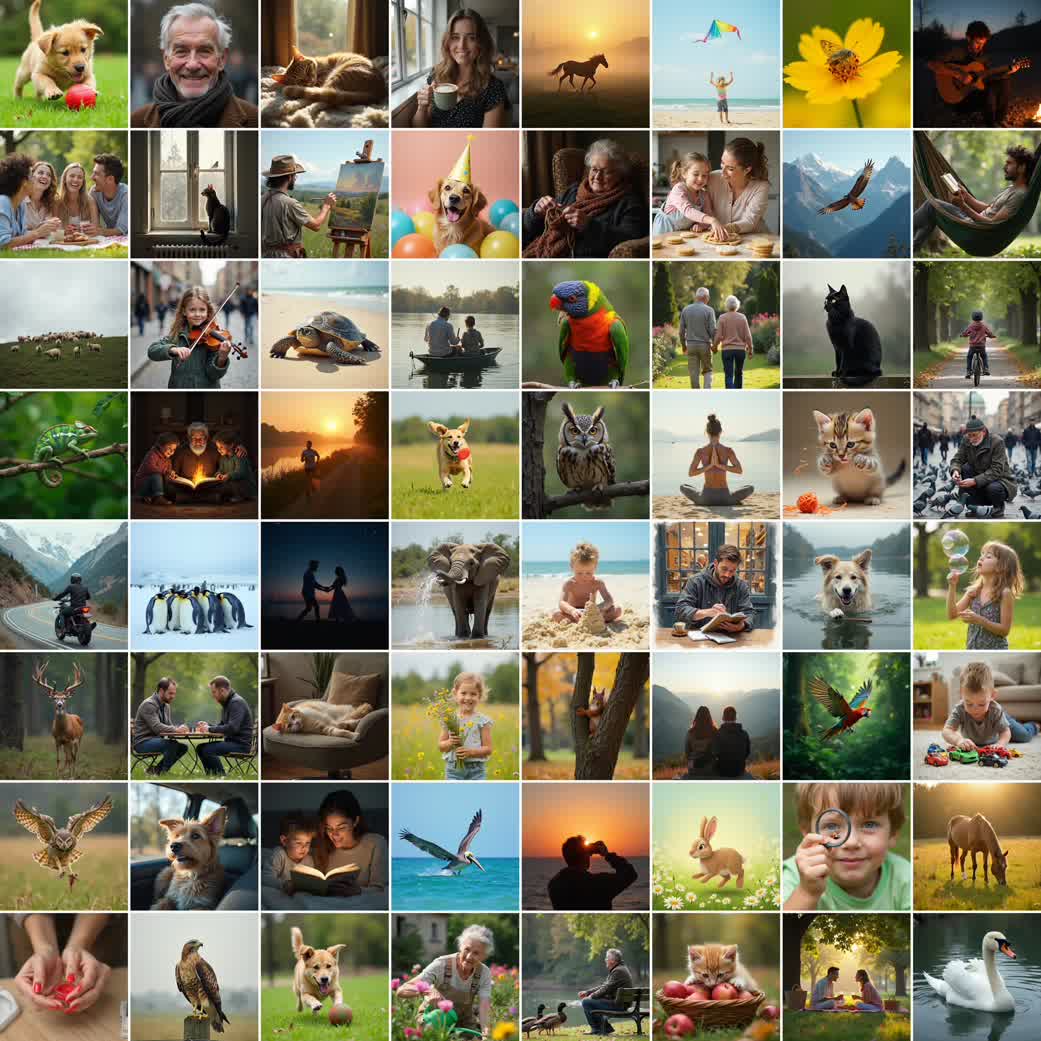} &
        \includegraphics[valign=c, width=\ww, frame]{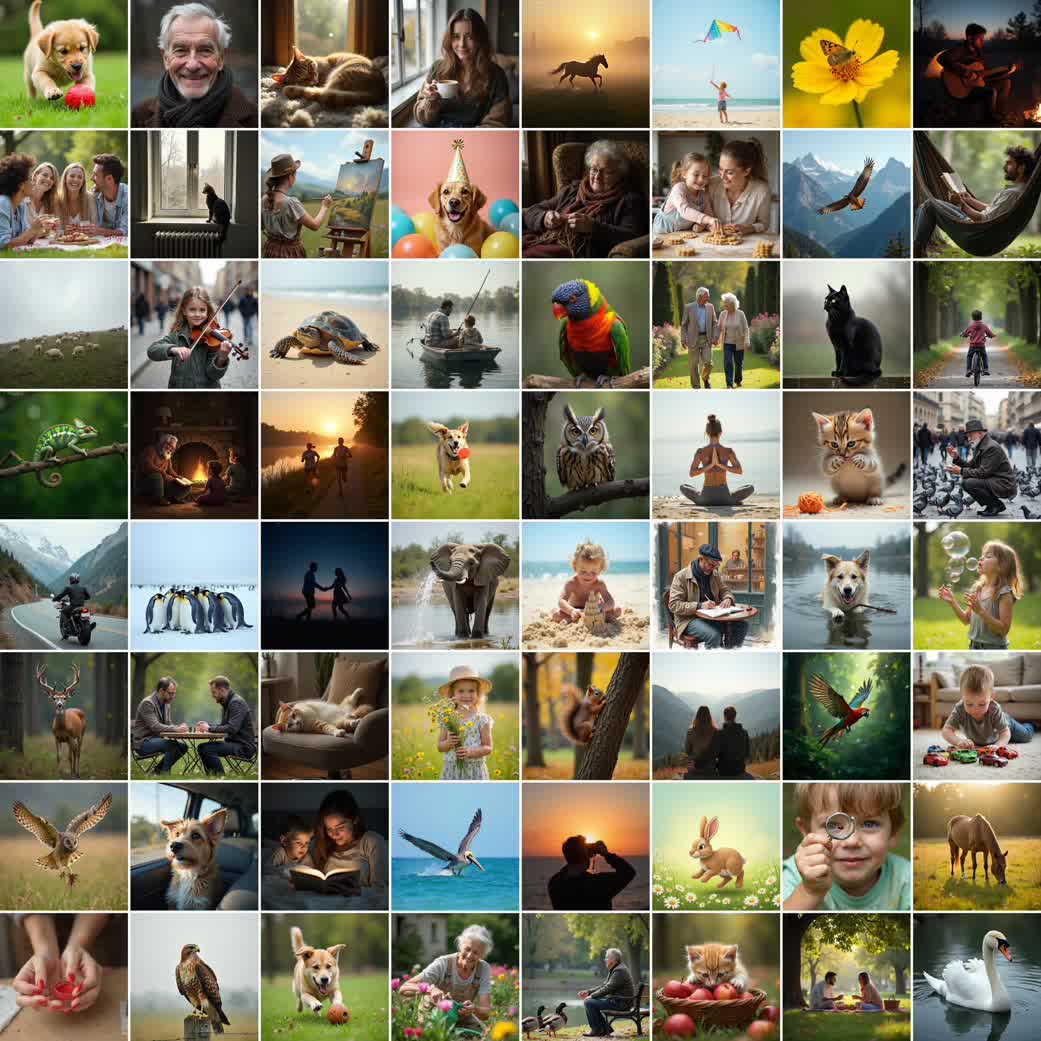}
        \vspace{3px}
        \\

        \nonvital{$G_{34}$} &
        \nonvital{$G_{35}$} &
        \nonvital{$G_{36}$}
        \vspace{15px}
        \\

        \includegraphics[valign=c, width=\ww, frame]{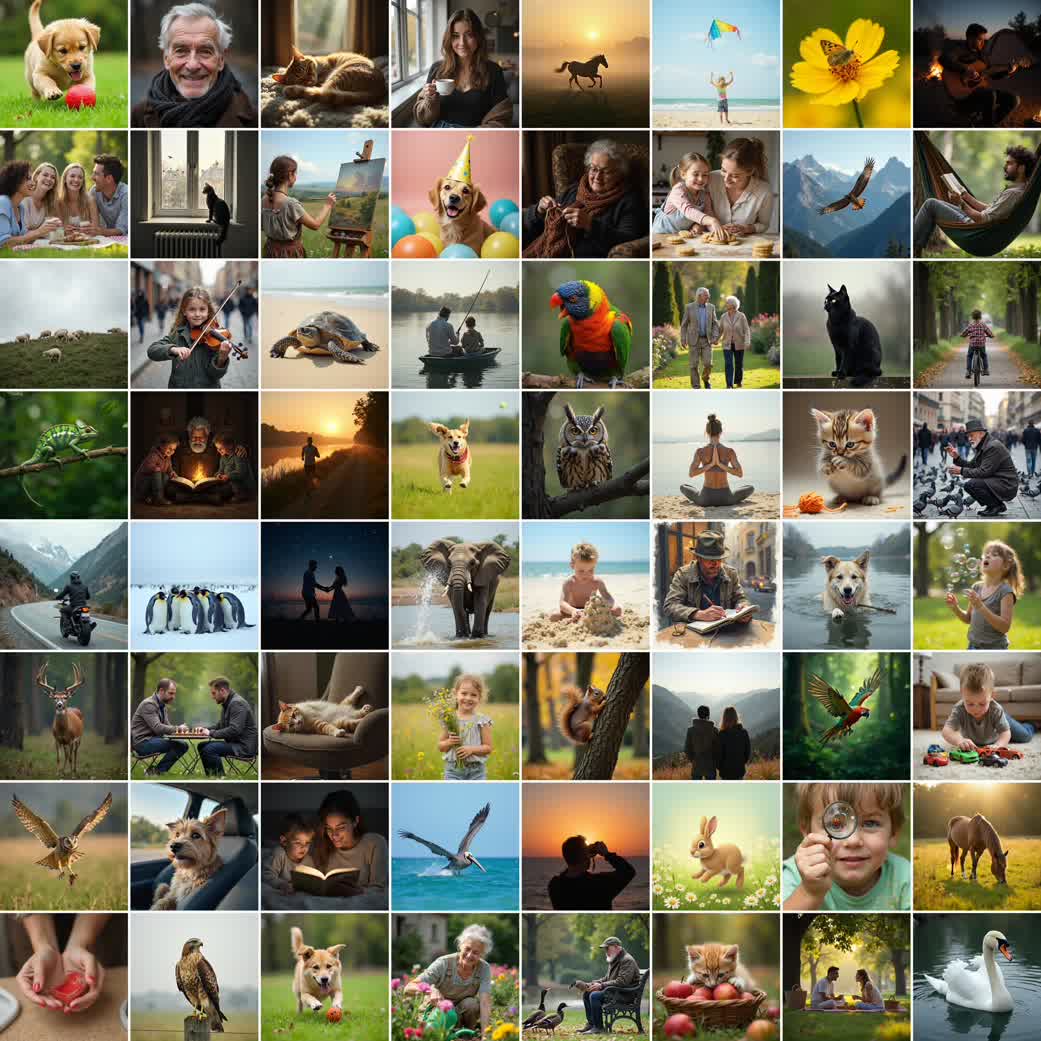} &
        \includegraphics[valign=c, width=\ww, frame]{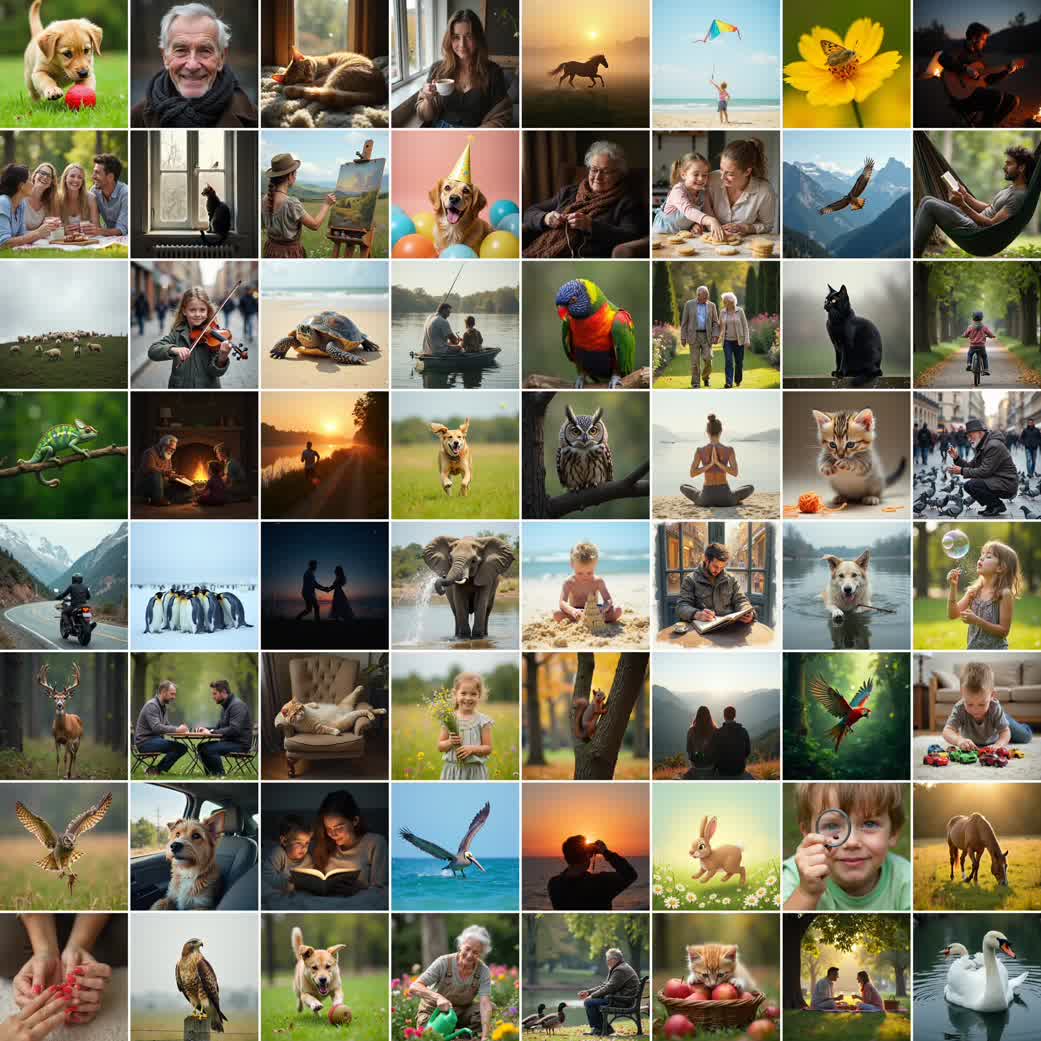} &
        \includegraphics[valign=c, width=\ww, frame]{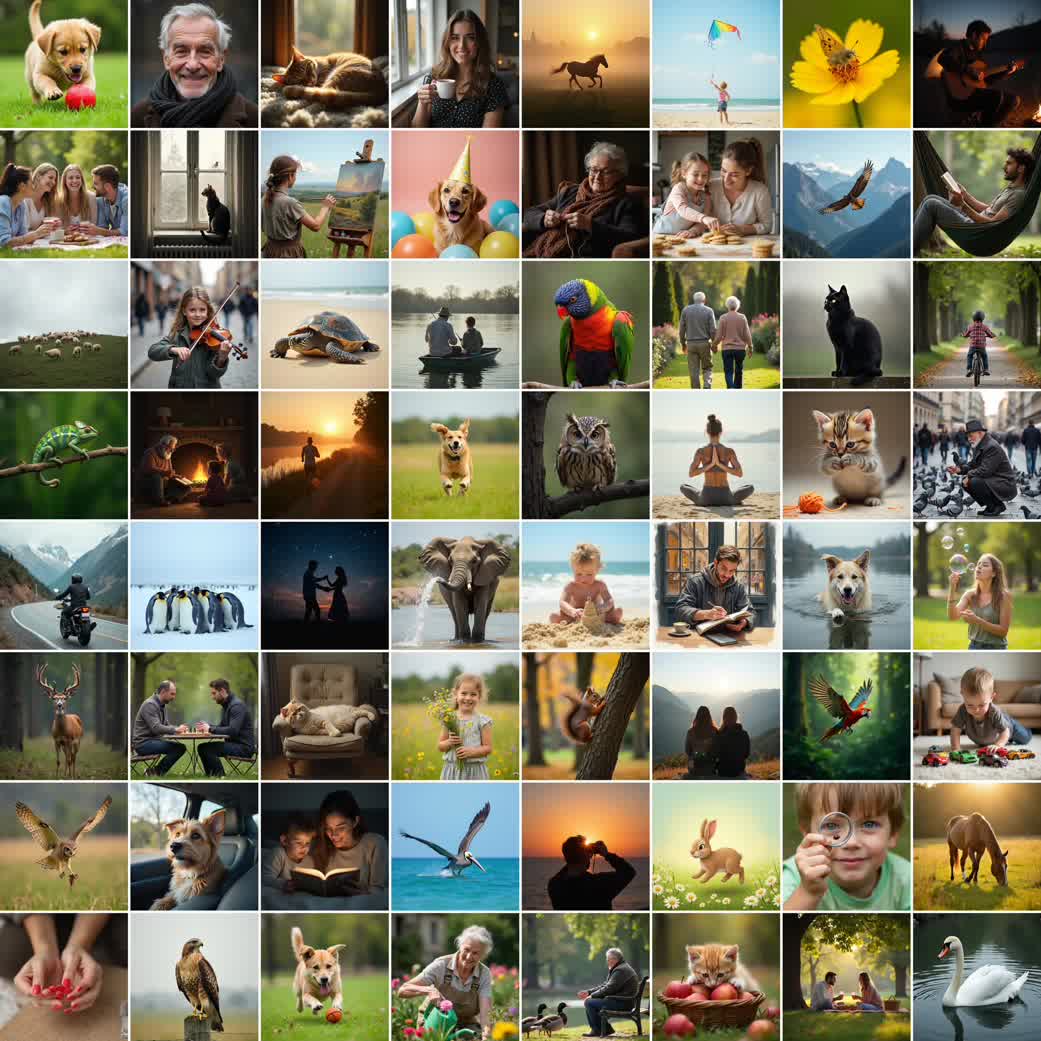}
        \vspace{3px}
        \\

        \nonvital{$G_{37}$} &
        \nonvital{$G_{38}$} &
        \nonvital{$G_{39}$}
        \vspace{3px}
        \\

    \end{tabular}
    \caption{\textbf{Full Layer Bypassing Visualization for Flux.} We visualize the individual layer bypassing study we conducted, as described in \Cref{sec:layer_bypassing_visualization}. We start by generating a set of images $G_{\textit{ref}}$ using a fixed set of seeds and prompts. Then, we bypass each layer $\ell$ by using its residual connection and generate the set of images $G_{\ell}$ using the same fixed set of prompts and seeds. In this visualization, \nonvital{$G_{31}$} -- \nonvital{$G_{39}$} are \nonvital{non-vital layers}.}
    \label{fig:full_flux_bypassing_5}
\end{figure*}

\begin{figure*}[tp]
    \centering
    \setlength{\tabcolsep}{2.5pt}
    \renewcommand{\arraystretch}{1.0}
    \setlength{\ww}{0.32\linewidth}
    \begin{tabular}{ccc}

        \includegraphics[valign=c, width=\ww, frame]{figures/full_bypassing_visualization_flux/assets/src.jpg} &
        \includegraphics[valign=c, width=\ww, frame]{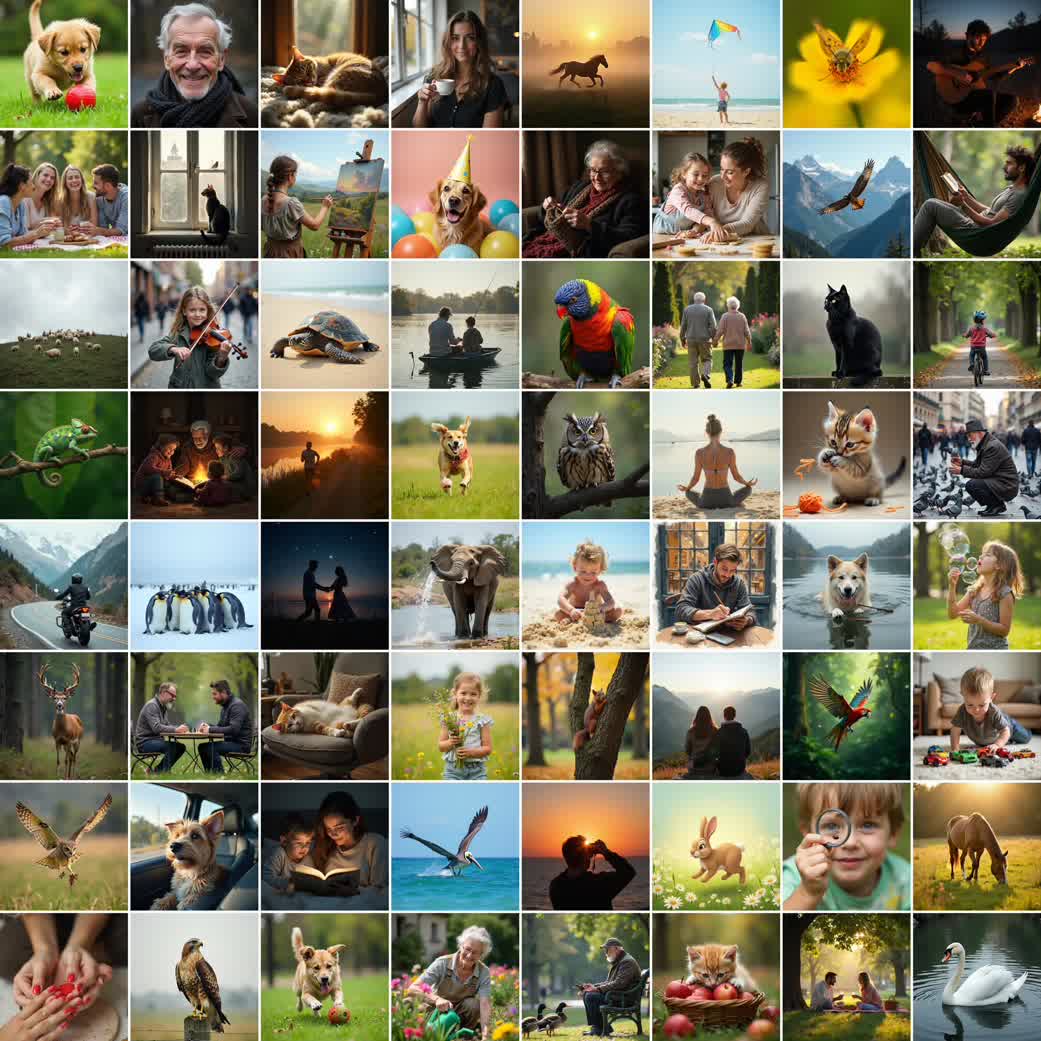} &
        \includegraphics[valign=c, width=\ww, frame]{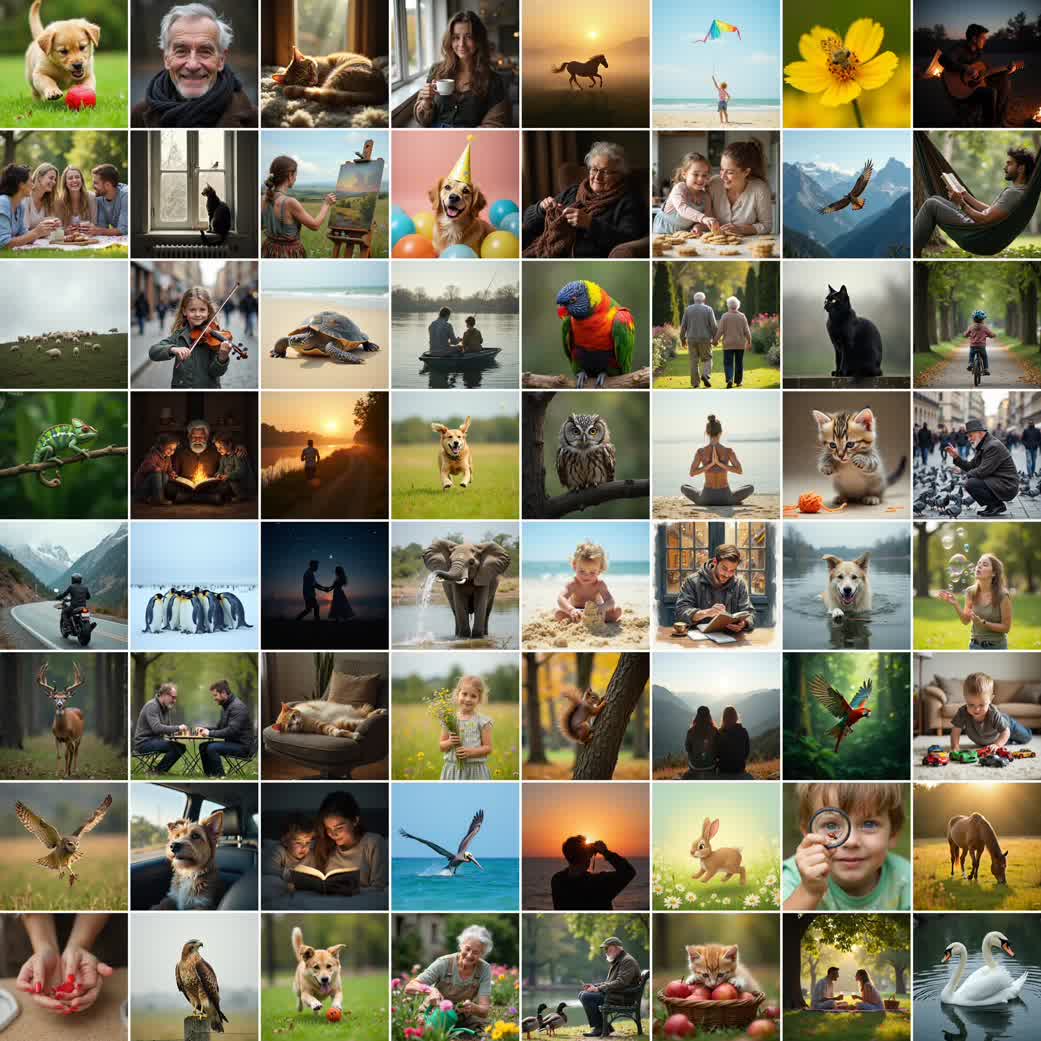}
        \vspace{3px}
        \\

        $G_{\textit{ref}}$ &
        \nonvital{$G_{40}$} &
        \nonvital{$G_{41}$}
        \vspace{15px}
        \\

        \includegraphics[valign=c, width=\ww, frame]{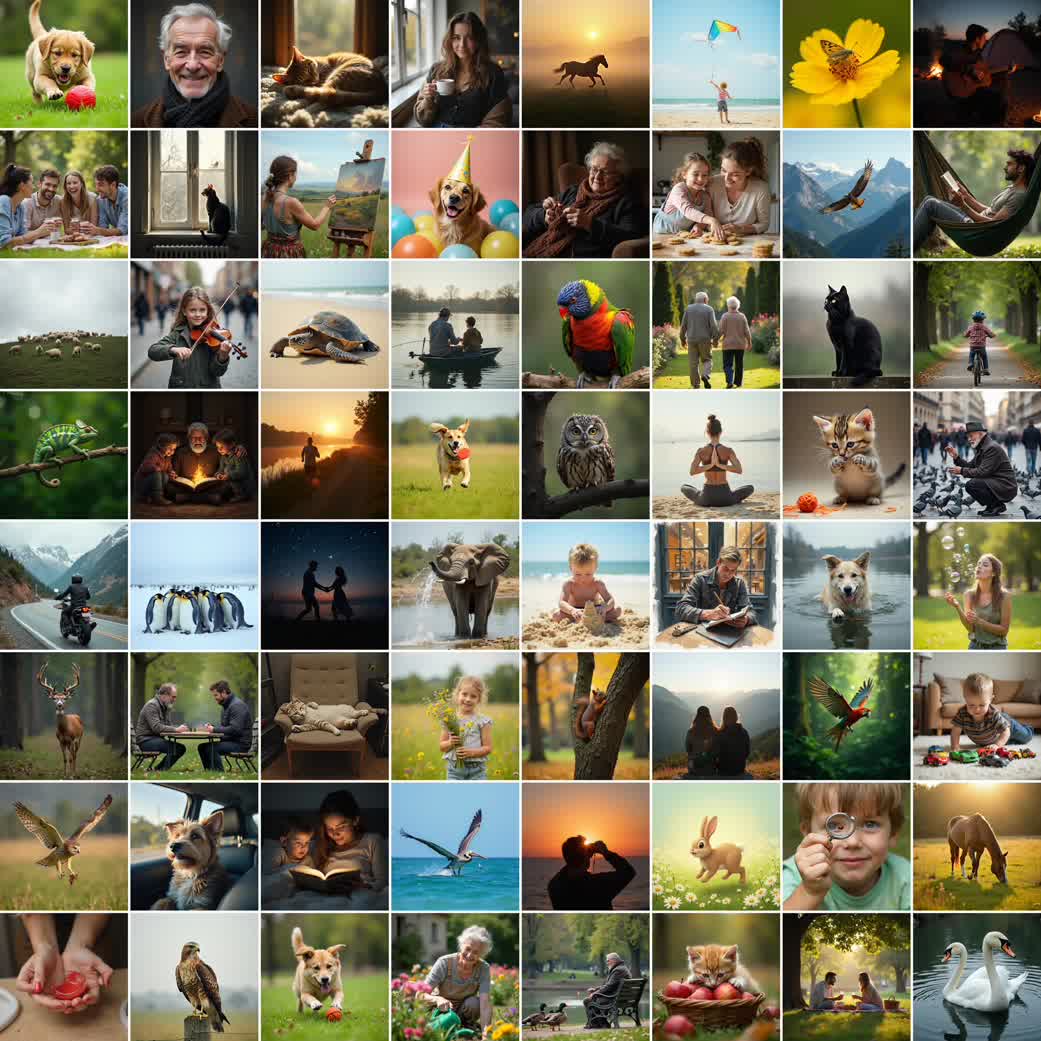} &
        \includegraphics[valign=c, width=\ww, frame]{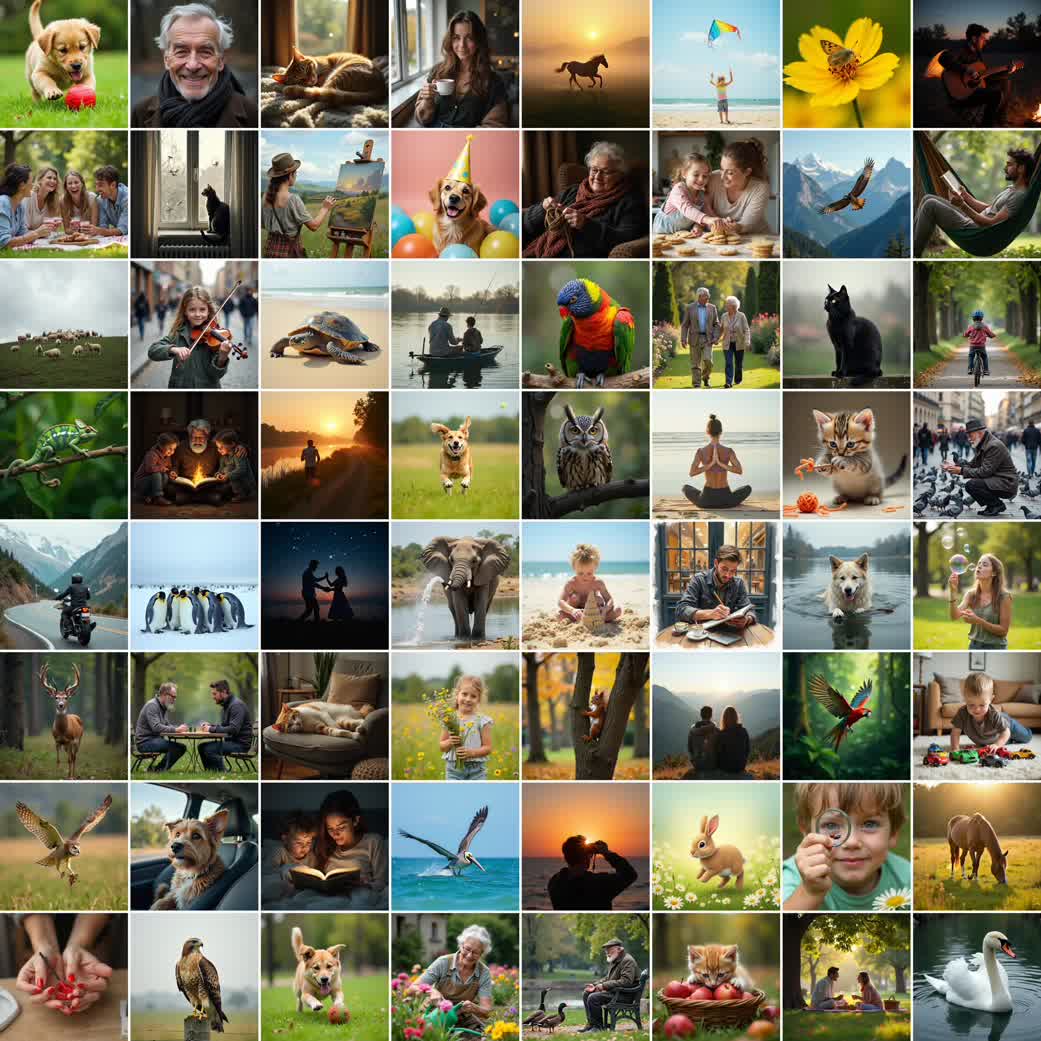} &
        \includegraphics[valign=c, width=\ww, frame]{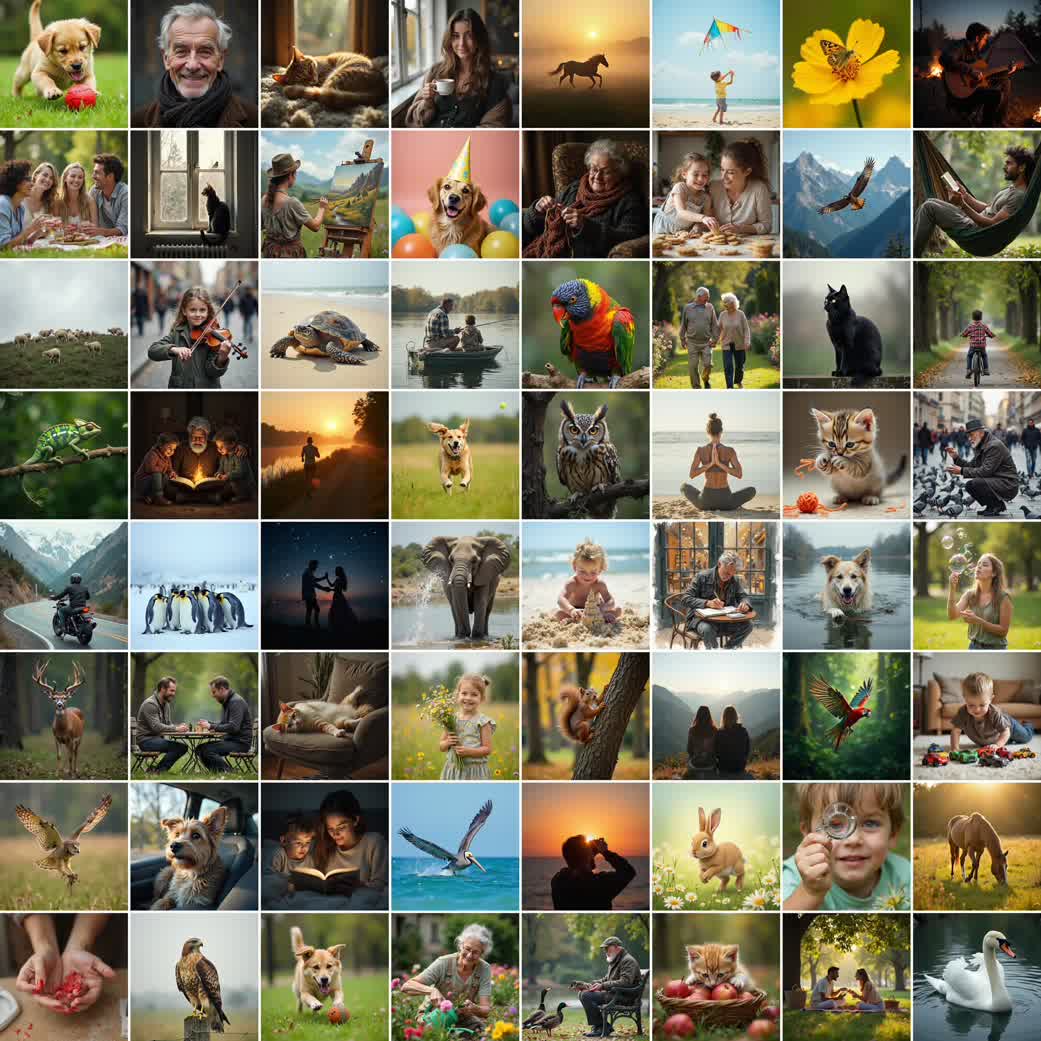}
        \vspace{3px}
        \\

        \nonvital{$G_{42}$} &
        \nonvital{$G_{43}$} &
        \nonvital{$G_{44}$}
        \vspace{15px}
        \\

        \includegraphics[valign=c, width=\ww, frame]{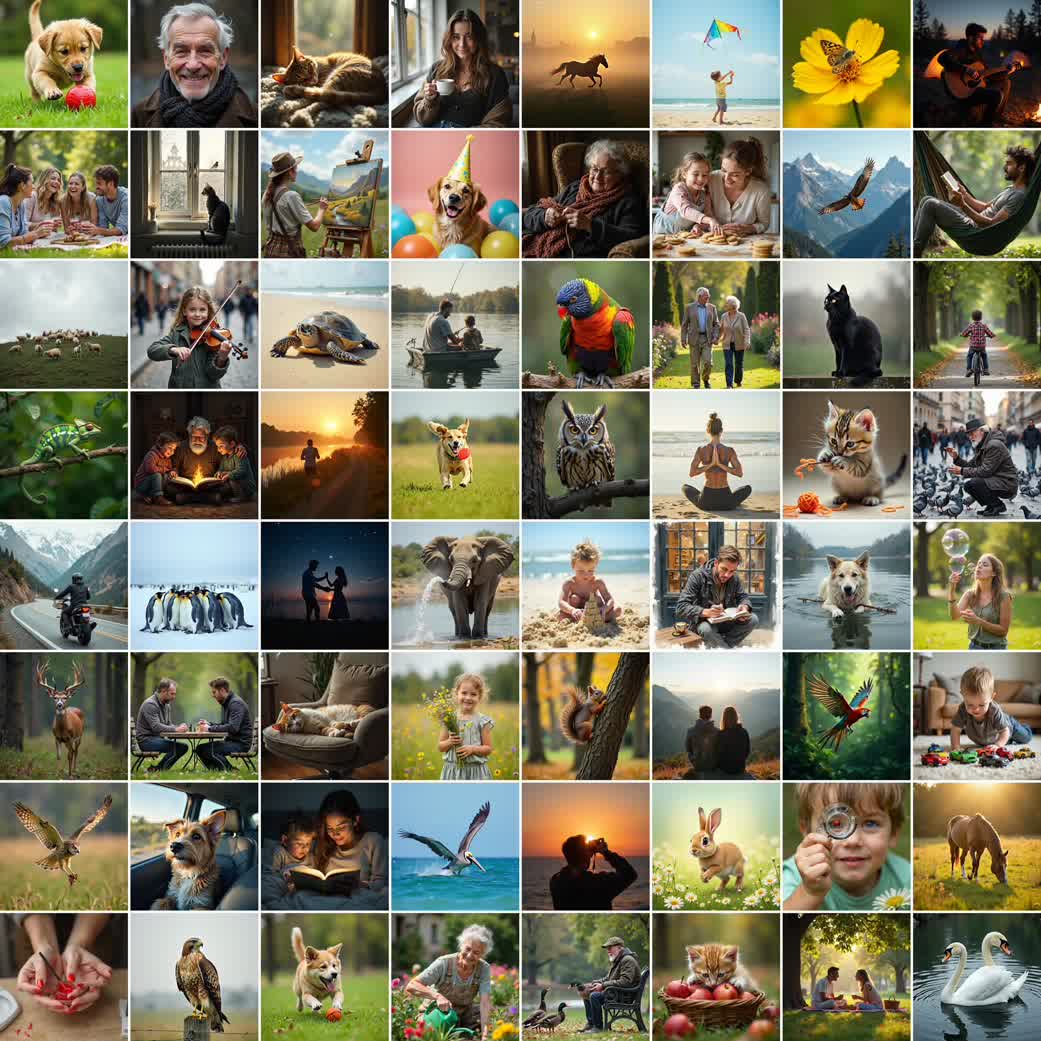} &
        \includegraphics[valign=c, width=\ww, frame]{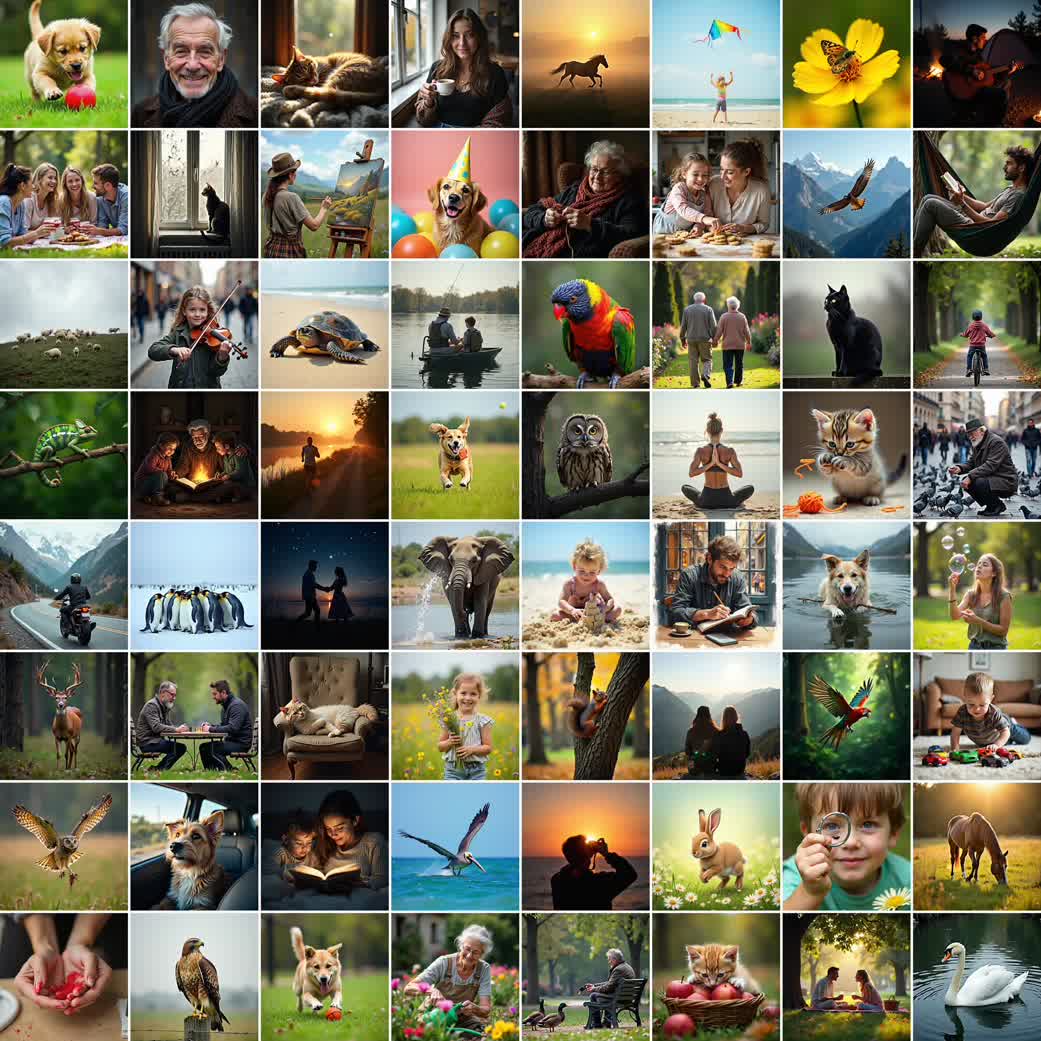} &
        \includegraphics[valign=c, width=\ww, frame]{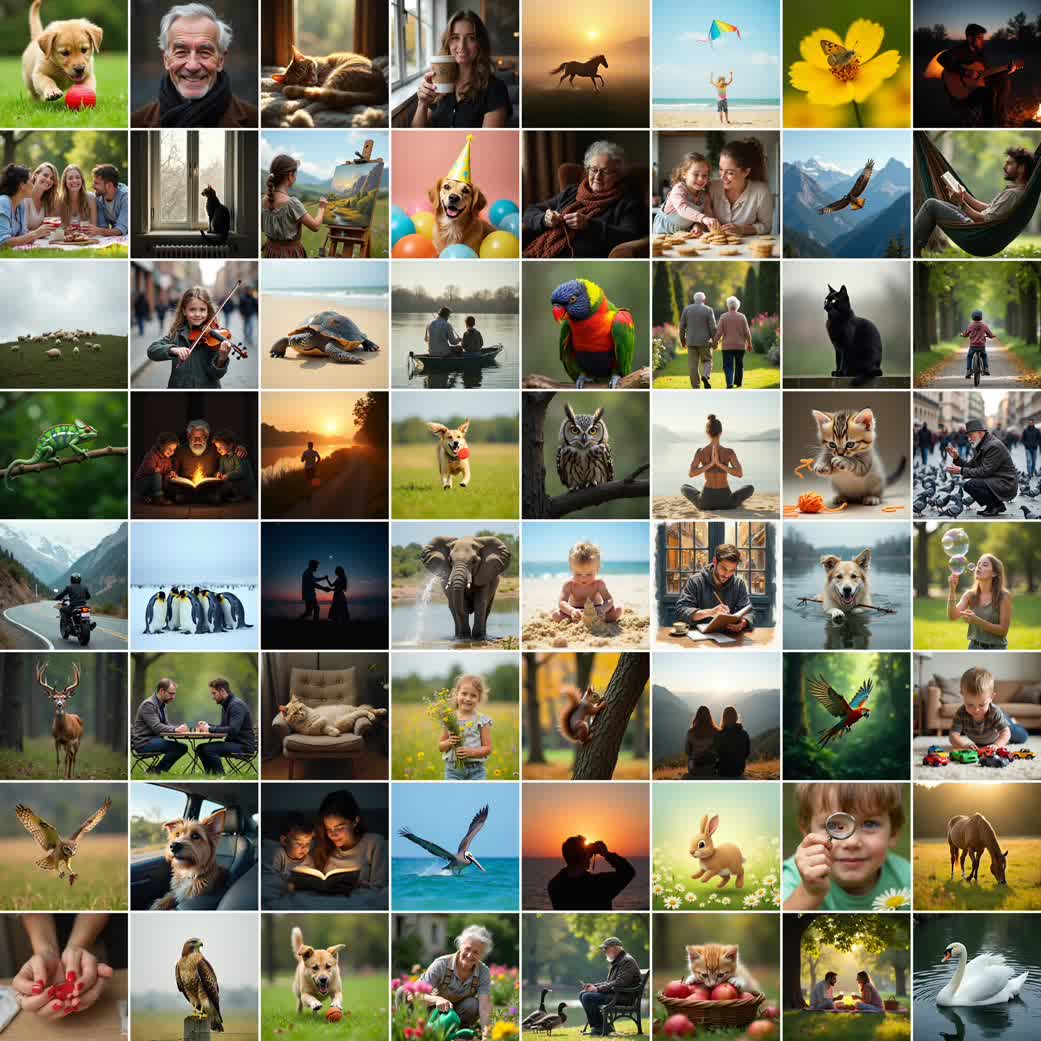}
        \vspace{3px}
        \\

        \nonvital{$G_{45}$} &
        \nonvital{$G_{46}$} &
        \nonvital{$G_{47}$}
        \vspace{3px}
        \\

    \end{tabular}
    \caption{\textbf{Full Layer Bypassing Visualization for Flux.} We visualize the individual layer bypassing study we conducted, as described in \Cref{sec:layer_bypassing_visualization}. We start by generating a set of images $G_{\textit{ref}}$ using a fixed set of seeds and prompts. Then, we bypass each layer $\ell$ by using its residual connection and generate the set of images $G_{\ell}$ using the same fixed set of prompts and seeds. In this visualization, \nonvital{$G_{40}$} -- \nonvital{$G_{47}$} are \nonvital{non-vital layers}.}
    \label{fig:full_flux_bypassing_6}
\end{figure*}

\begin{figure*}[tp]
    \centering
    \setlength{\tabcolsep}{2.5pt}
    \renewcommand{\arraystretch}{1.0}
    \setlength{\ww}{0.32\linewidth}
    \begin{tabular}{ccc}

        \includegraphics[valign=c, width=\ww, frame]{figures/full_bypassing_visualization_flux/assets/src.jpg} &
        \includegraphics[valign=c, width=\ww, frame]{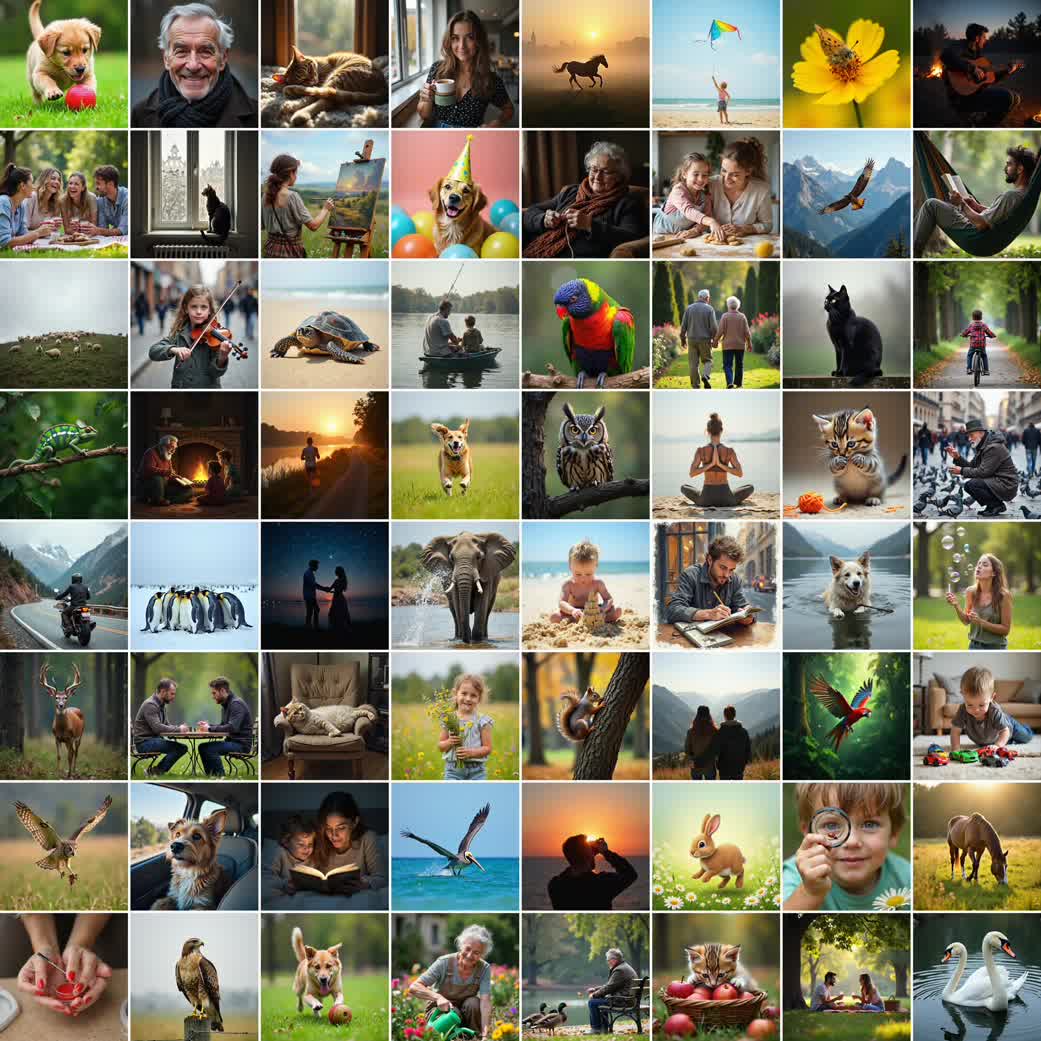} &
        \includegraphics[valign=c, width=\ww, frame]{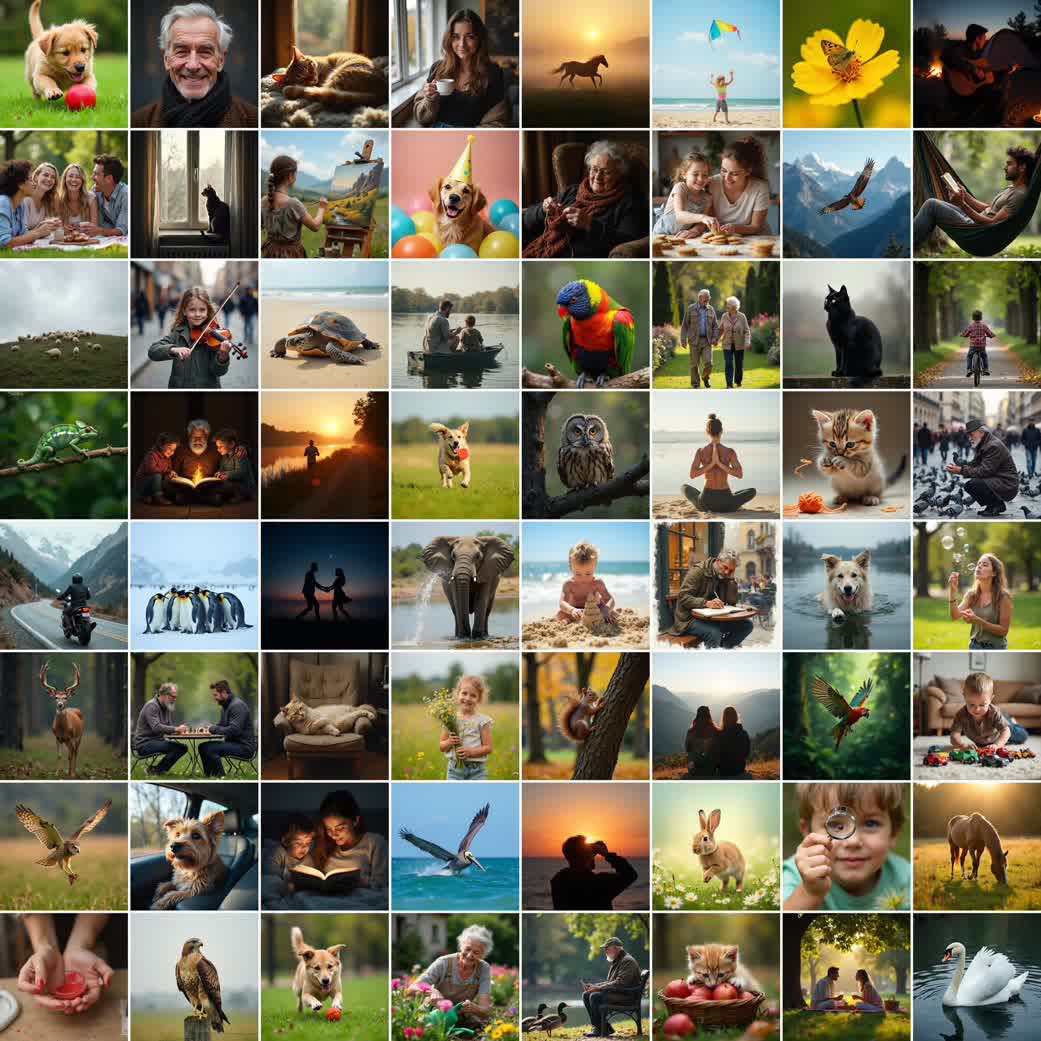}
        \vspace{3px}
        \\

        $G_{\textit{ref}}$ &
        \nonvital{$G_{48}$} &
        \nonvital{$G_{49}$}
        \vspace{15px}
        \\

        \includegraphics[valign=c, width=\ww, frame]{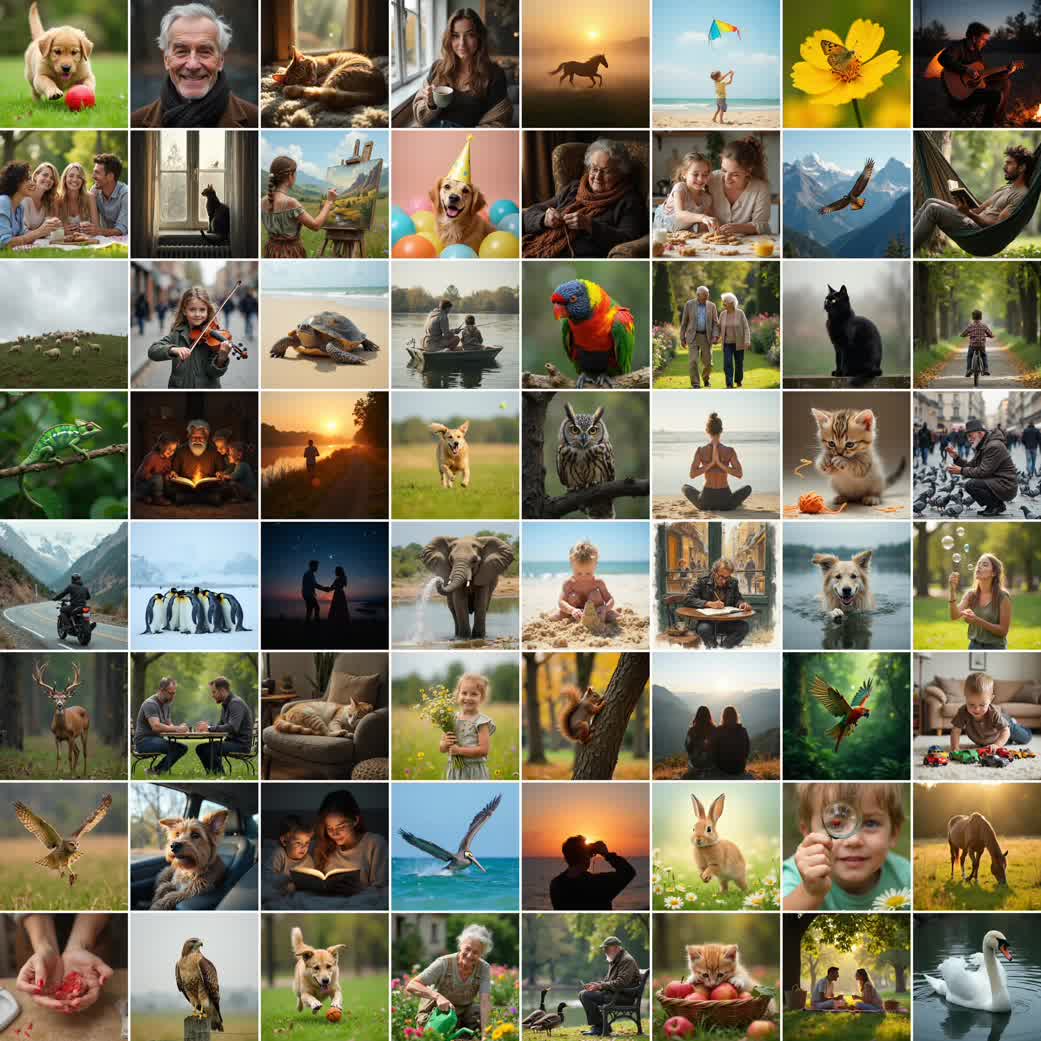} &
        \includegraphics[valign=c, width=\ww, frame]{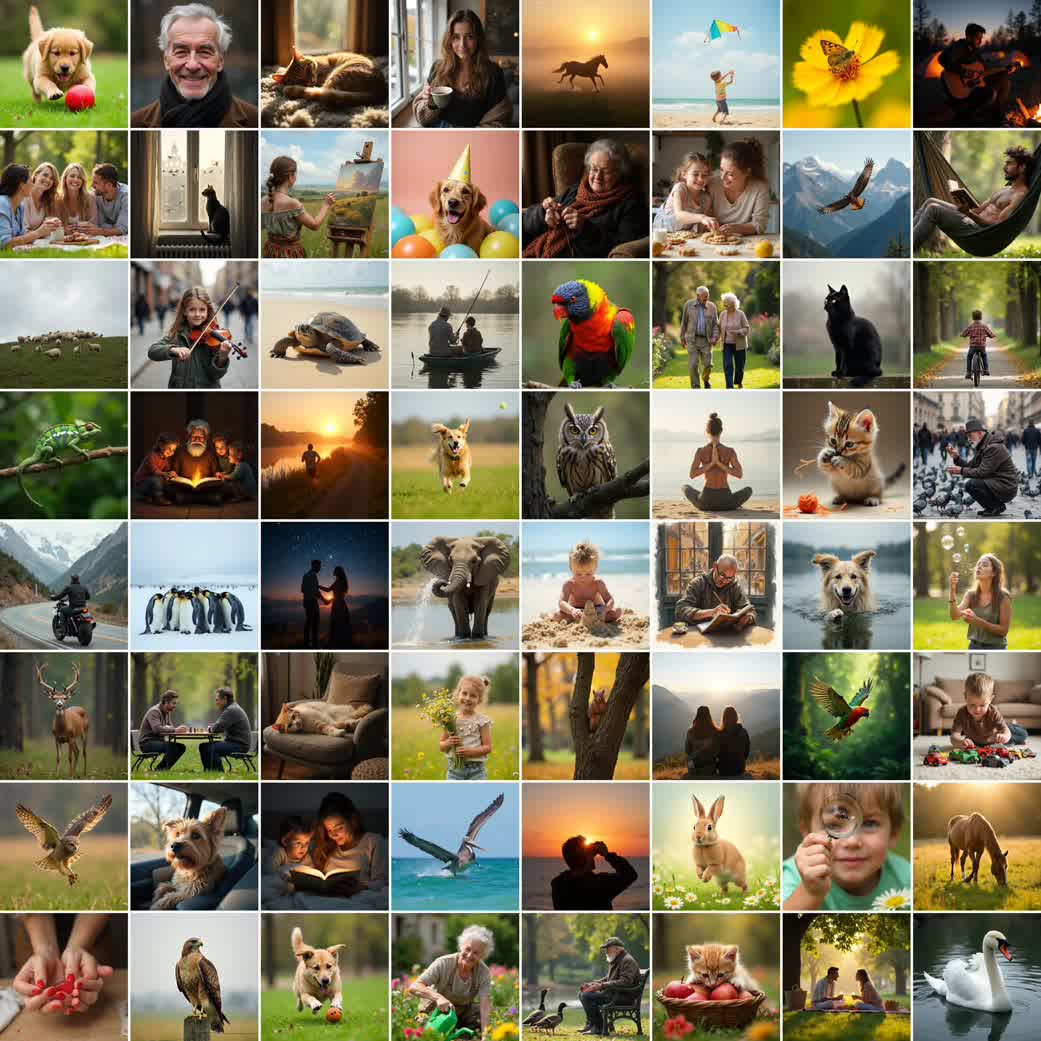} &
        \includegraphics[valign=c, width=\ww, frame]{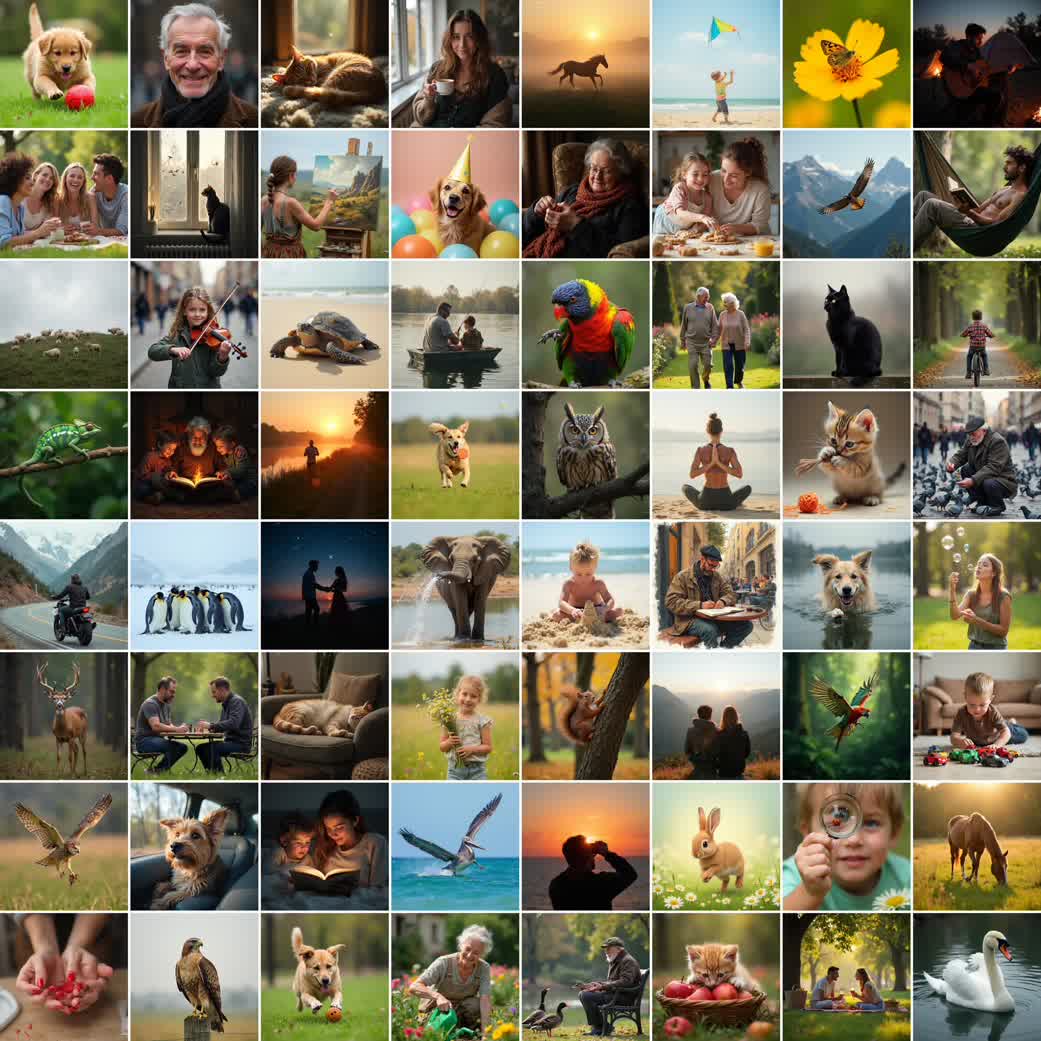}
        \vspace{3px}
        \\

        \nonvital{$G_{50}$} &
        \nonvital{$G_{51}$} &
        \nonvital{$G_{52}$}
        \vspace{15px}
        \\

        \includegraphics[valign=c, width=\ww, frame]{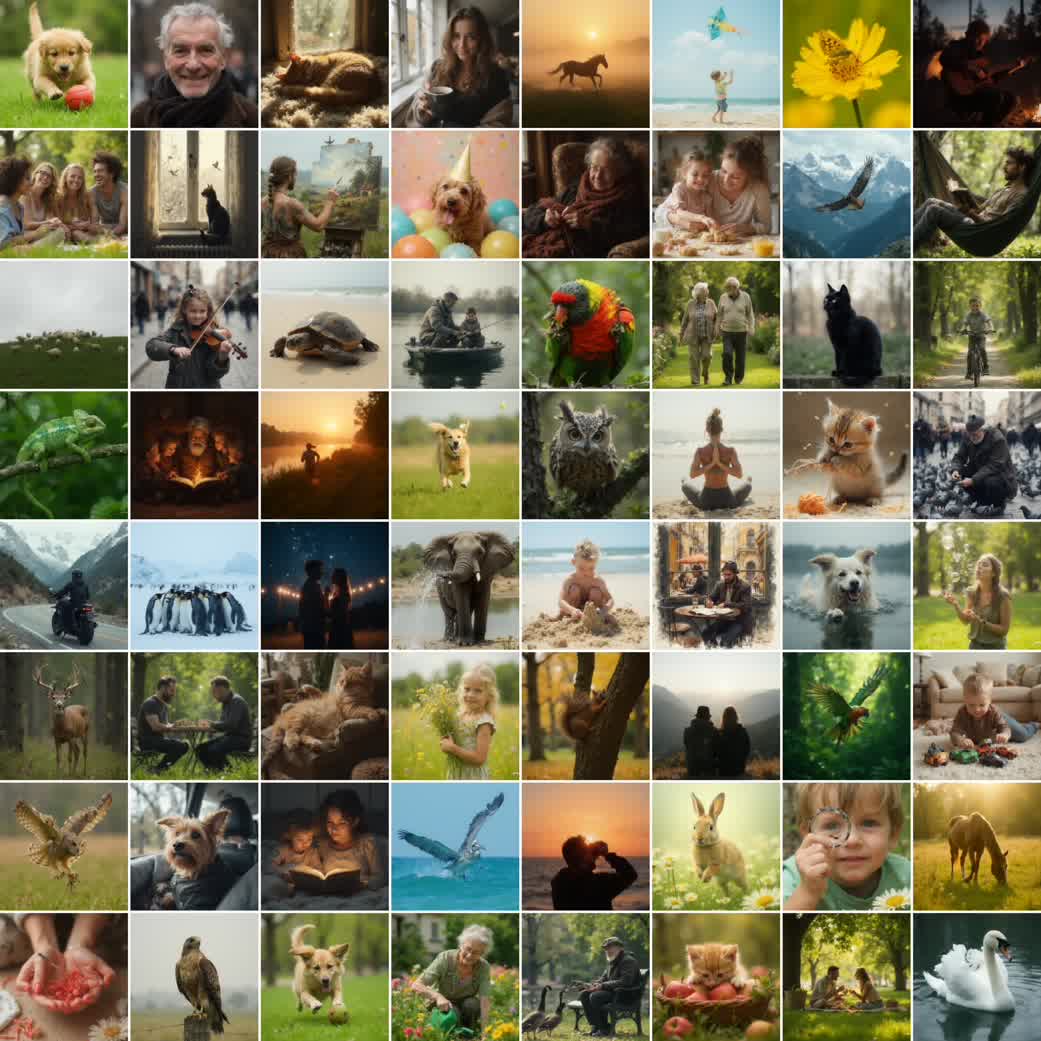} &
        \includegraphics[valign=c, width=\ww, frame]{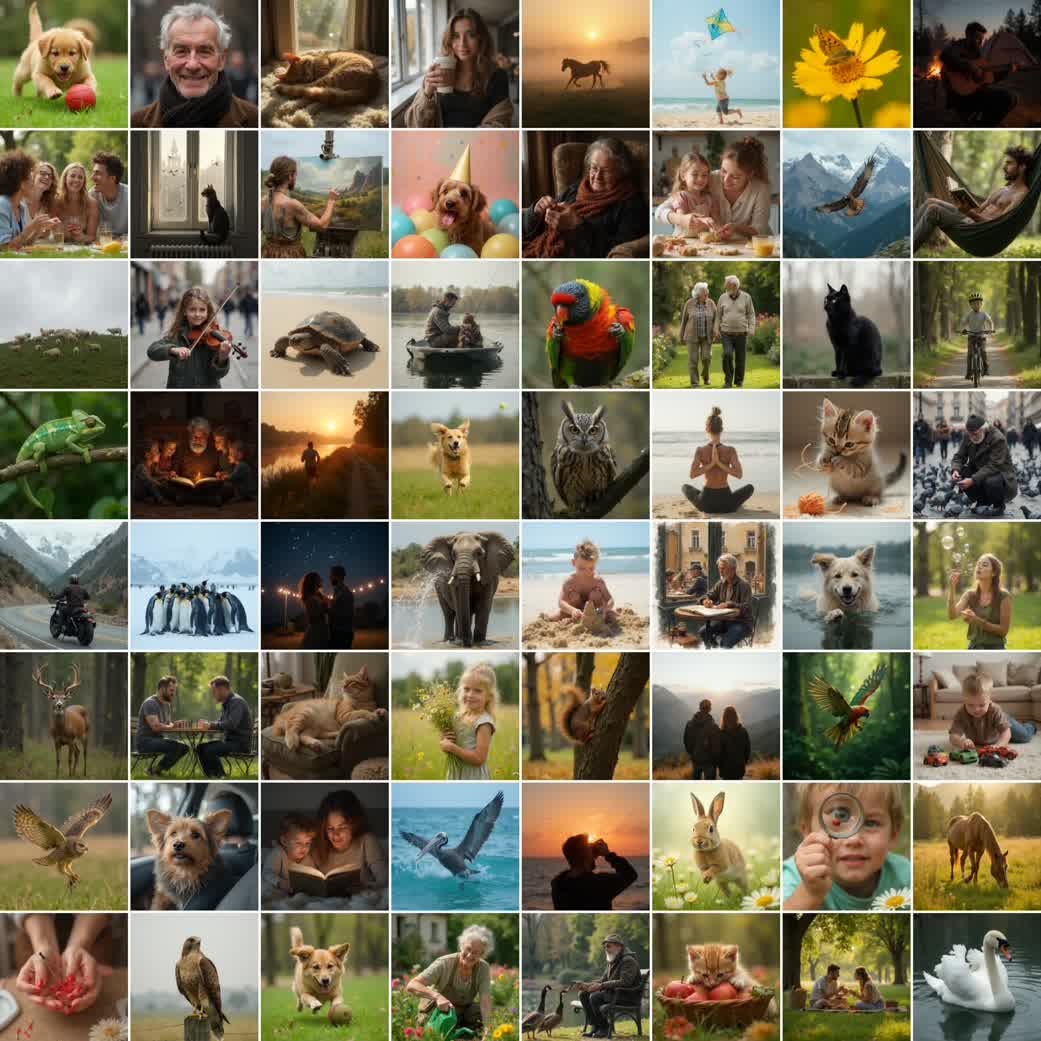} &
        \includegraphics[valign=c, width=\ww, frame]{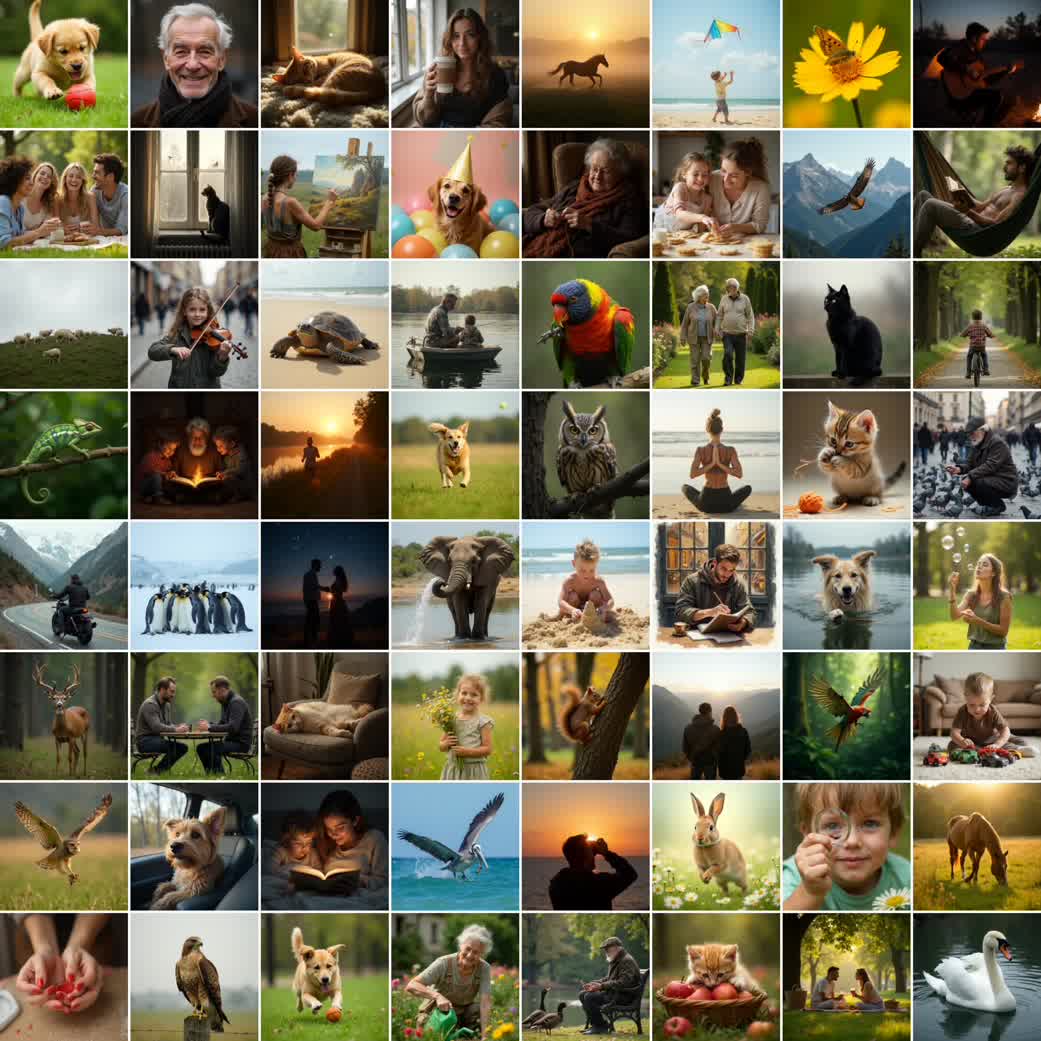}
        \vspace{3px}
        \\

        \vital{$G_{53}$} &
        \vital{$G_{54}$} &
        \nonvital{$G_{55}$}
        \vspace{3px}
        \\

    \end{tabular}
    \caption{\textbf{Full Layer Bypassing Visualization for Flux.} We visualize the individual layer bypassing study we conducted, as described in \Cref{sec:layer_bypassing_visualization}. We start by generating a set of images $G_{\textit{ref}}$ using a fixed set of seeds and prompts. Then, we bypass each layer $\ell$ by using its residual connection and generate the set of images $G_{\ell}$ using the same fixed set of prompts and seeds. In this visualization, \vital{$G_{53}$} -- \vital{$G_{54}$} are \vital{vital layers}, while \nonvital{$G_{48}$} -- \nonvital{$G_{52}$} and \nonvital{$G_{55}$} are \nonvital{non-vital layers}.}
    \label{fig:full_flux_bypassing_7}
\end{figure*}

\begin{figure*}[tp]
    \centering
    \setlength{\tabcolsep}{2.5pt}
    \renewcommand{\arraystretch}{1.0}
    \setlength{\ww}{0.32\linewidth}
    \begin{tabular}{ccc}

        \includegraphics[valign=c, width=\ww, frame]{figures/full_bypassing_visualization_flux/assets/src.jpg} &
        \includegraphics[valign=c, width=\ww, frame]{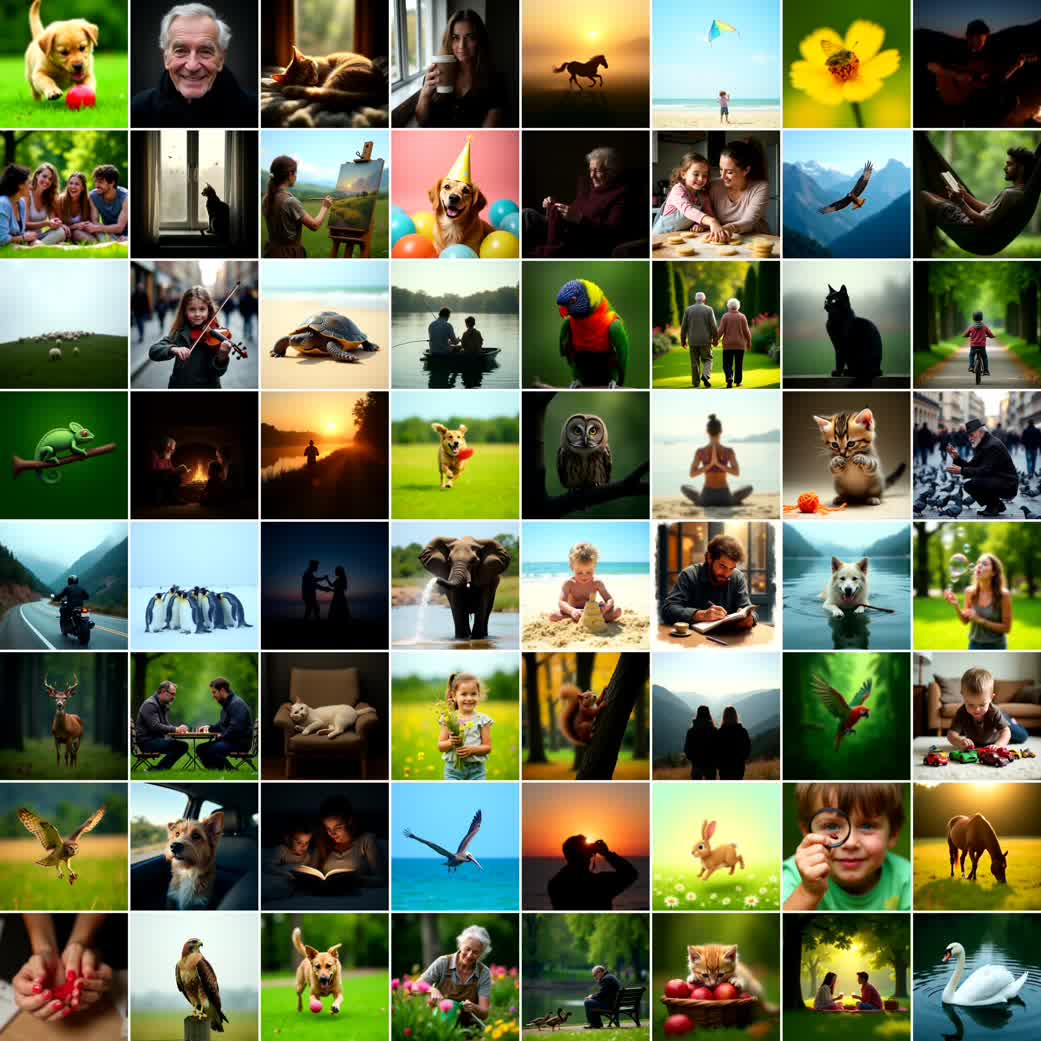}
        \vspace{3px}
        \\

        $G_{\textit{ref}}$ &
        \vital{$G_{56}$}
        \\

    \end{tabular}
    \caption{\textbf{Full Layer Bypassing Visualization for Flux.} We visualize the individual layer bypassing study we conducted, as described in \Cref{sec:layer_bypassing_visualization}. We start by generating a set of images $G_{\textit{ref}}$ using a fixed set of seeds and prompts. Then, we bypass each layer $\ell$ by using its residual connection and generate the set of images $G_{\ell}$ using the same fixed set of prompts and seeds. In this visualization, \vital{$G_{56}$} is a \vital{vital layer}.}
    \label{fig:full_flux_bypassing_8}
\end{figure*}

As explained in
\Cref{sec:layers_importance},
to quantify layer importance in FLUX model, we devised a systematic evaluation approach. Using ChatGPT~\cite{chatgpt}, we automatically generated a set $P$ of $k=64$ diverse text prompts, and draw a set $S$ of random seeds. Each of these prompts was used to generate a reference image, yielding a set $G_{\textit{ref}}$. For each DiT layer $\ell \in \mathbb{L}$, we performed an ablation by bypassing the layer using its residual connection. This process generated a set of images $G_{\ell}$ from the same prompts and seeds.

In Figures~\ref{fig:full_flux_bypassing_1}--\ref{fig:full_flux_bypassing_8}, we provide a full visualization of the reference set $G_{\textit{ref}}$ along with the generation sets $G_{0}$ -- $G_{56}$. As can be seen, removing certain layers significantly affects the generated images, while others have minimal impact. Importantly, influential layers are distributed across the transformer rather than concentrated in specific regions.

\subsection{Stable Diffusion 3 Results}
\label{sec:sd3_results}

\begin{figure}[t]
    \begin{tikzpicture} [thick,scale=0.9, every node/.style={scale=1}]
        \begin{axis}[
            xlabel={\small{Layer Index}},
            ylabel={\small{Perceptual Similarity}},
            compat=newest,
            xtick distance=5,
            xmin=-1.5,
            xmax=24,
            width=9.5cm,
            height=7.5cm,
            scatter/classes={
                a={mark=*, fill=nonvitalcolor},
                b={mark=*, fill=vitalcolor}
            }
            ]
            \addplot[scatter, only marks, scatter src=explicit symbolic]
            table[meta=label] {
                x y label
                0 0.0020854640752077103 b
                1 0.7422949075698853 a
                2 0.6501061916351318 a
                3 0.5383912324905396 a
                4 0.6440898180007935 a
                5 0.486069917678833 a
                6 0.42831894755363464 a
                7 0.23743410408496857 b
                8 0.22978109121322632 b
                9 0.22925147414207458 b
                10 0.733039140701294 a
                11 0.49151161313056946 a
                12 0.48834526538848877 a
                13 0.8173272609710693 a
                14 0.7957580089569092 a
                15 0.7881970405578613 a
                16 0.8351772427558899 a
                17 0.6954573392868042 a
                18 0.8754571676254272 a
                19 0.8159570097923279 a
                20 0.8627651929855347 a
                21 0.8981156349182129 a
                22 0.8502355813980103 a
                23 0.8472093939781189 a
            };
            \end{axis}
    \end{tikzpicture}

    \caption{\textbf{Layer Removal Quantitative Comparison Stable Diffusion 3.} As explained in \Cref{sec:sd3_results}, we measured the effect of removing each layer of the model by calculating the perceptual similarity between the generated images with and without this layer. Lower perceptual similarity indicates significant changes in the generated images. As can be seen, removing certain layers significantly affects the generated images, while others have minimal impact. For a visual comparison, please refer to \Cref{fig:layer_removal_qualitative_sd3}.}
    \label{fig:layer_removal_quantiative_sd3}
\end{figure}
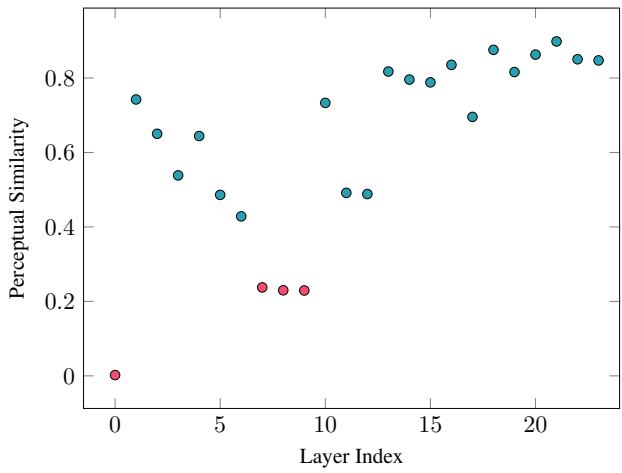

\begin{figure}[t]
    \centering
    \setlength{\tabcolsep}{0.6pt}
    \renewcommand{\arraystretch}{0.8}
    \setlength{\ww}{0.235\columnwidth}
    \begin{tabular}{ccccc}
        \rotatebox[origin=c]{90}{\footnotesize{$G_{\textit{ref}}$}} &
        \includegraphics[valign=c, width=\ww]{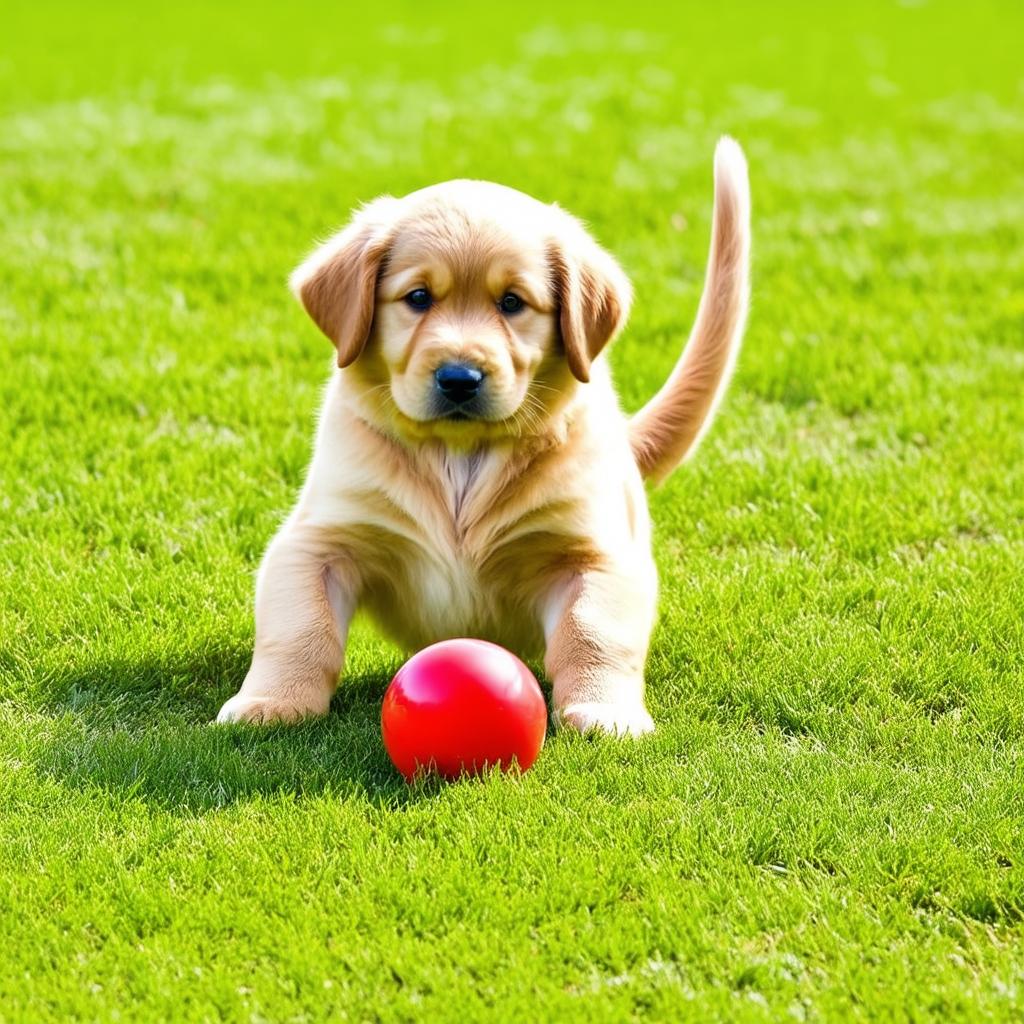} &
        \includegraphics[valign=c, width=\ww]{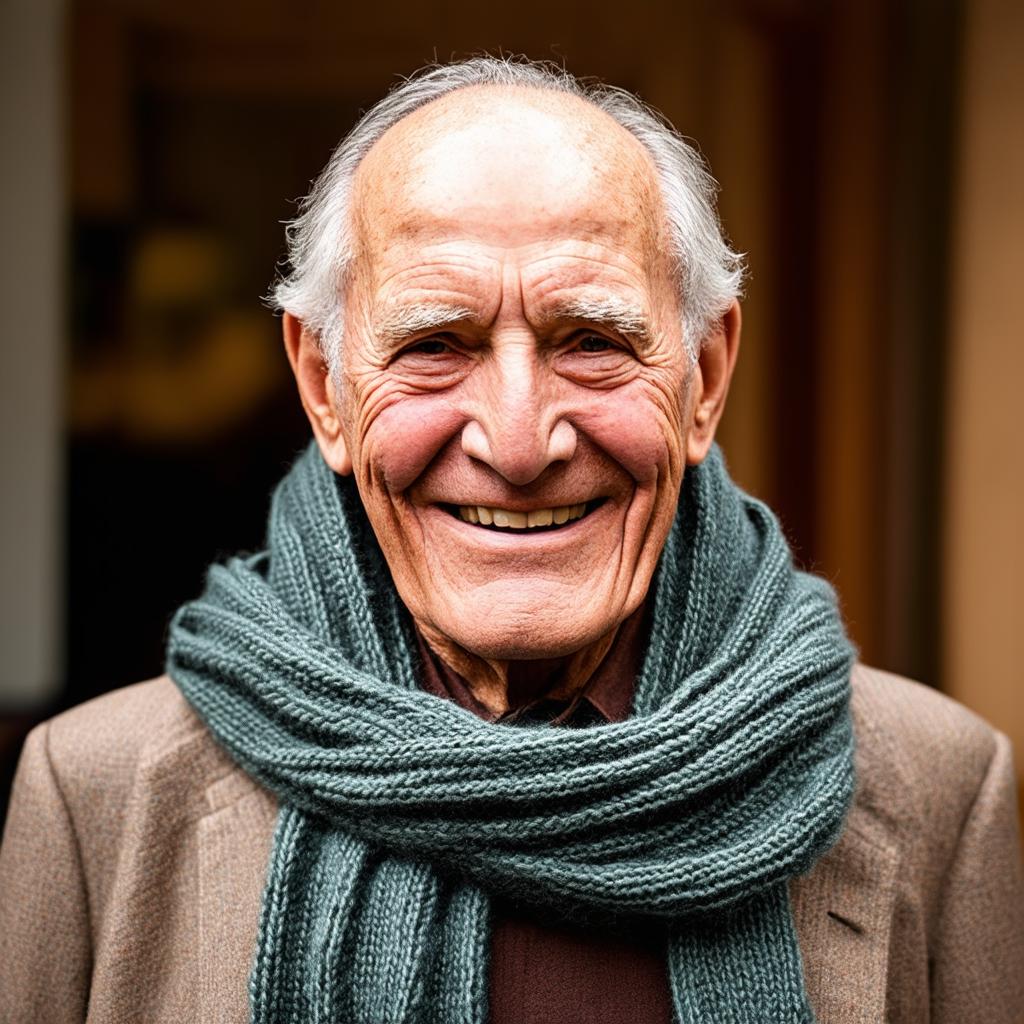} &
        \includegraphics[valign=c, width=\ww]{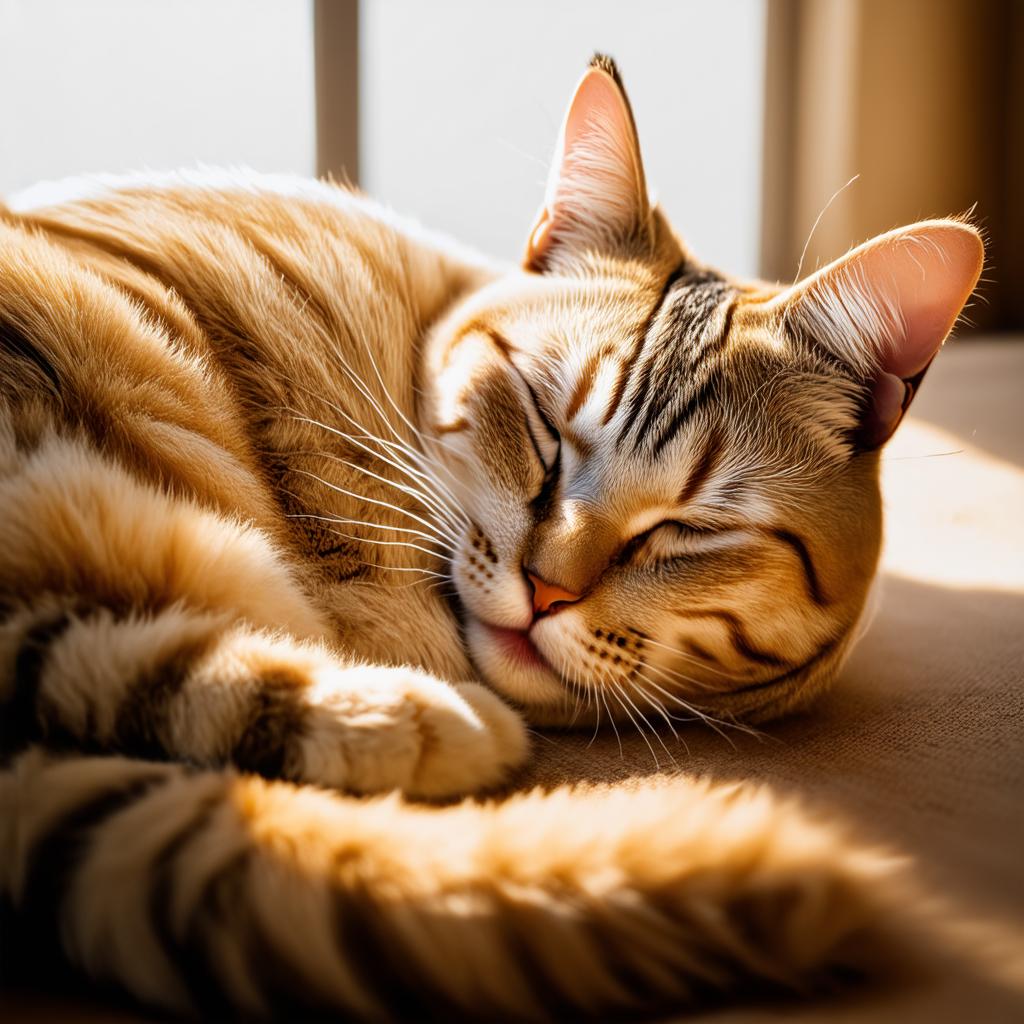} &
        \includegraphics[valign=c, width=\ww]{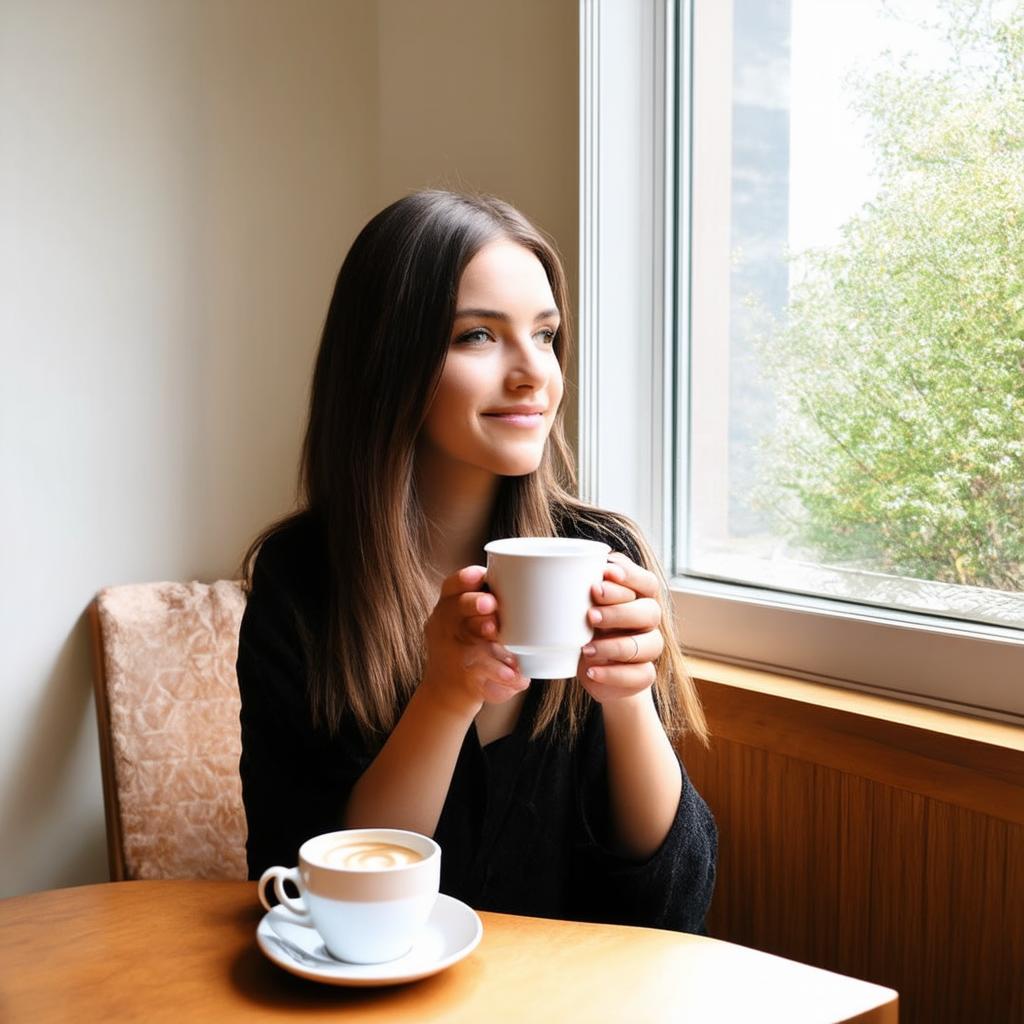}
        \vspace{5px}
        \\

        \rotatebox[origin=c]{90}{\vital{\footnotesize{$G_{0}$}}} &
        \includegraphics[valign=c, width=\ww]{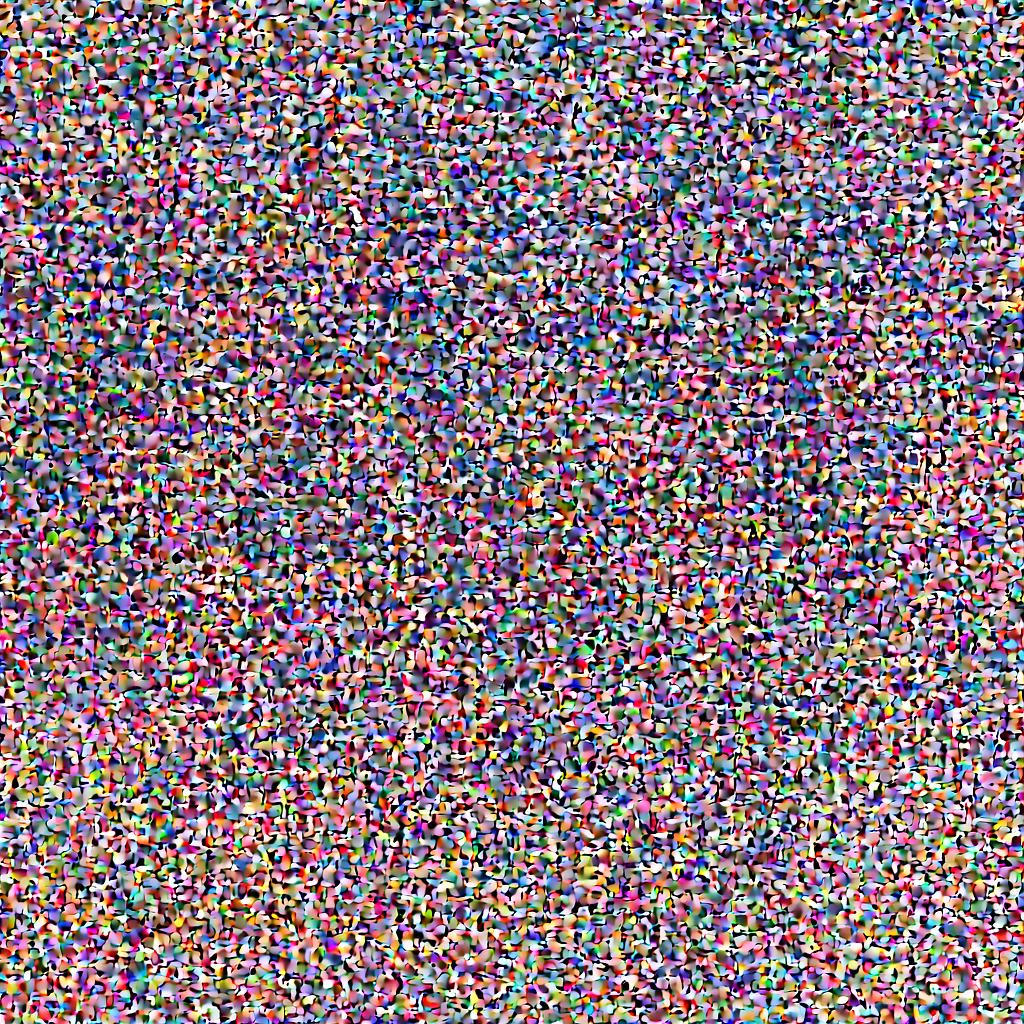} &
        \includegraphics[valign=c, width=\ww]{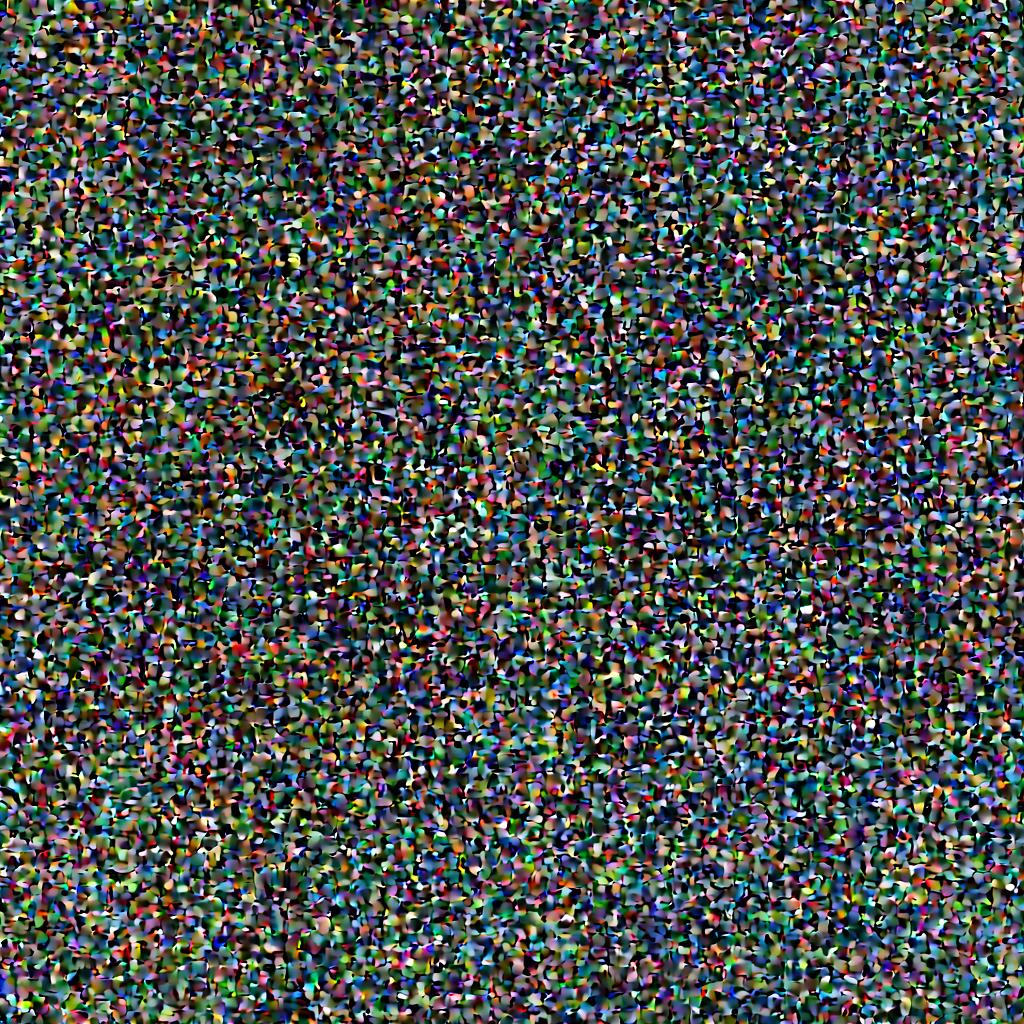} &
        \includegraphics[valign=c, width=\ww]{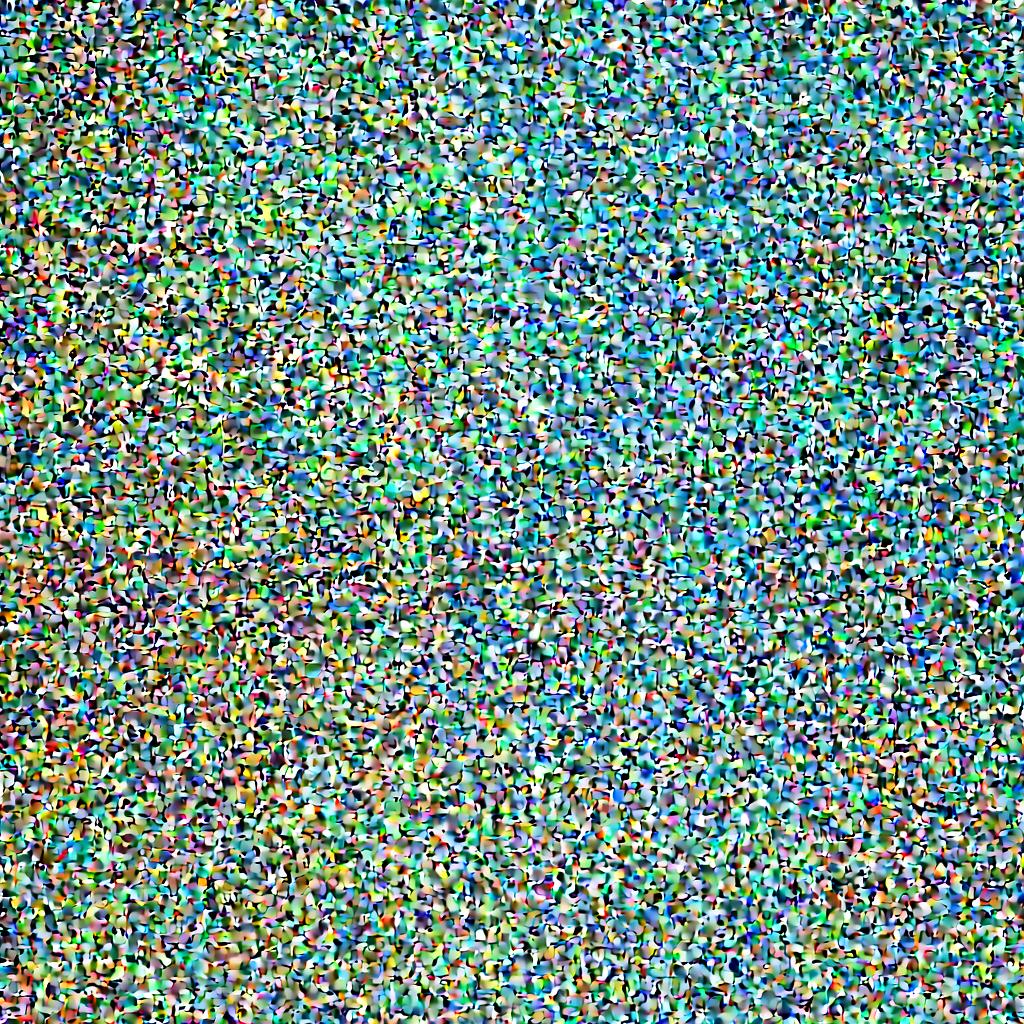} &
        \includegraphics[valign=c, width=\ww]{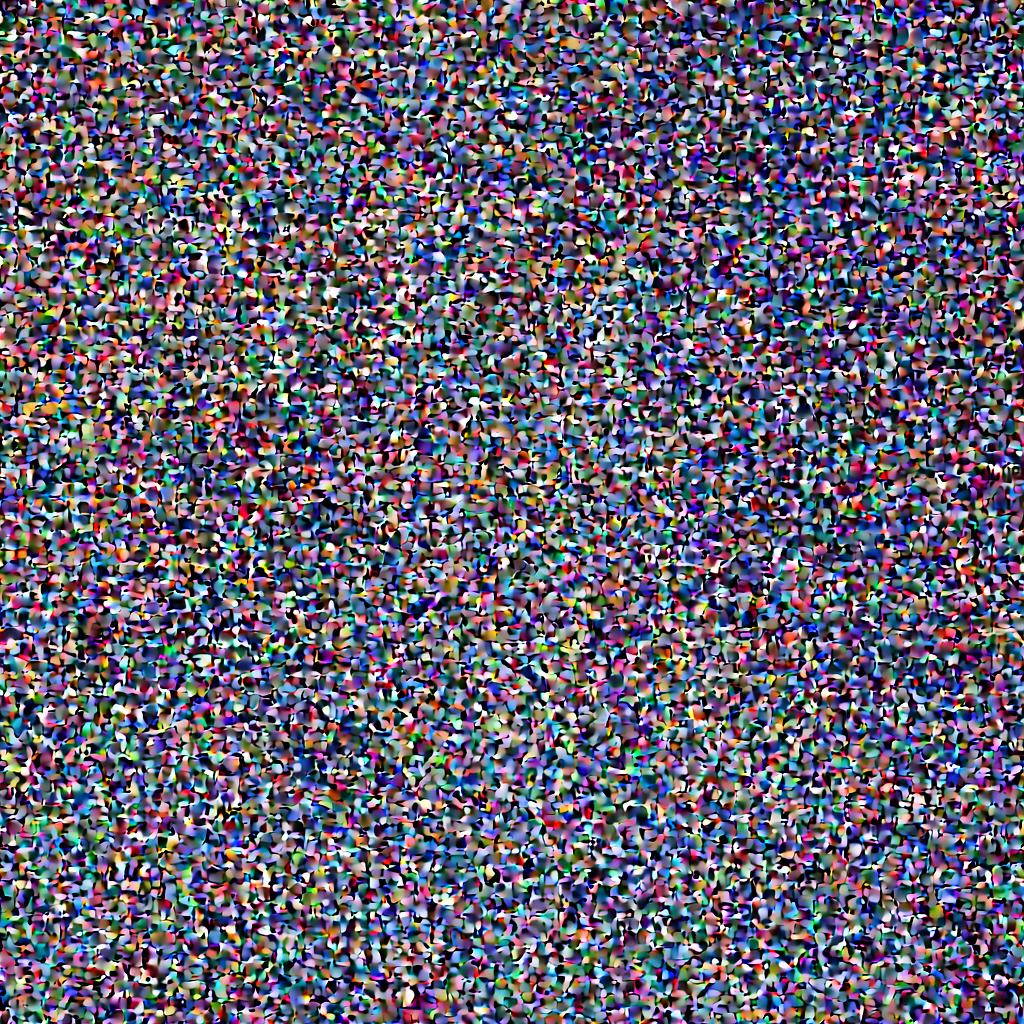}
        \vspace{1px}
        \\

        \rotatebox[origin=c]{90}{\nonvital{\footnotesize{$G_{1}$}}} &
        \includegraphics[valign=c, width=\ww]{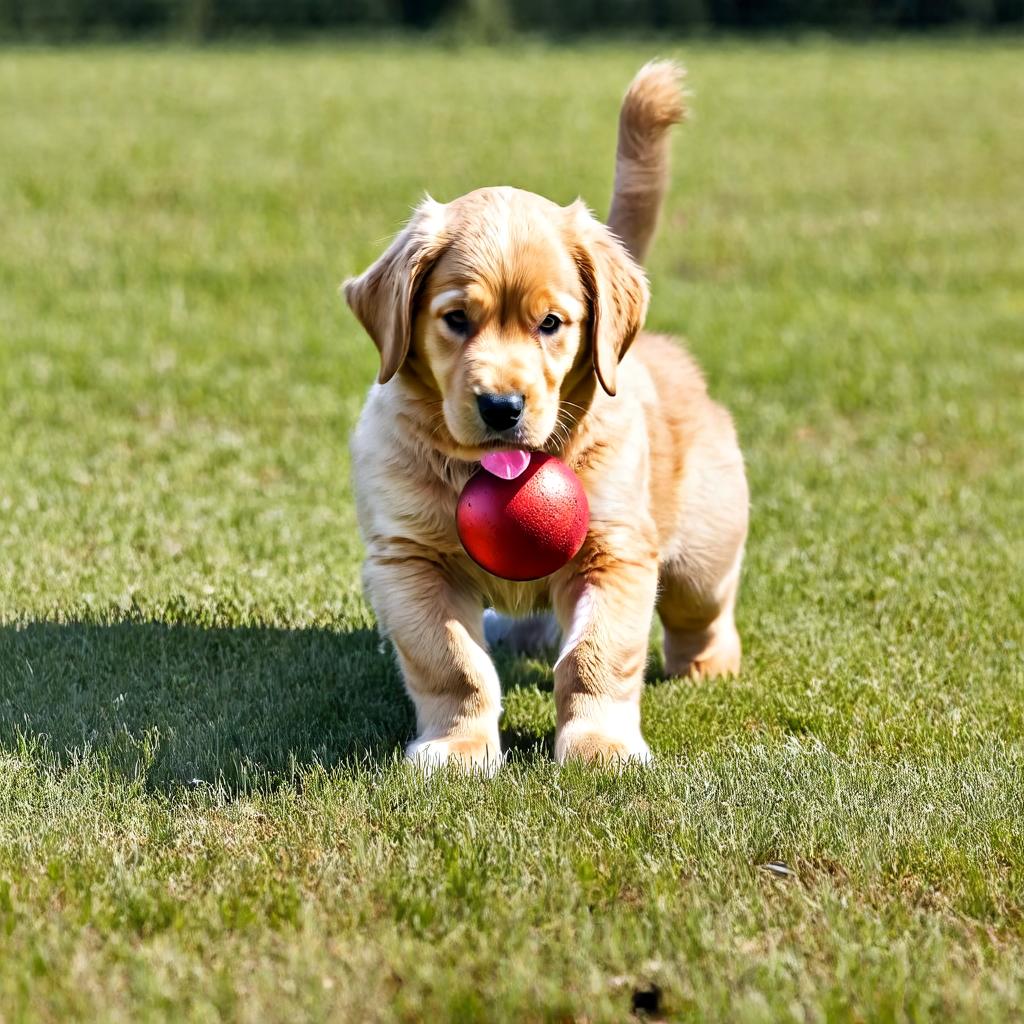} &
        \includegraphics[valign=c, width=\ww]{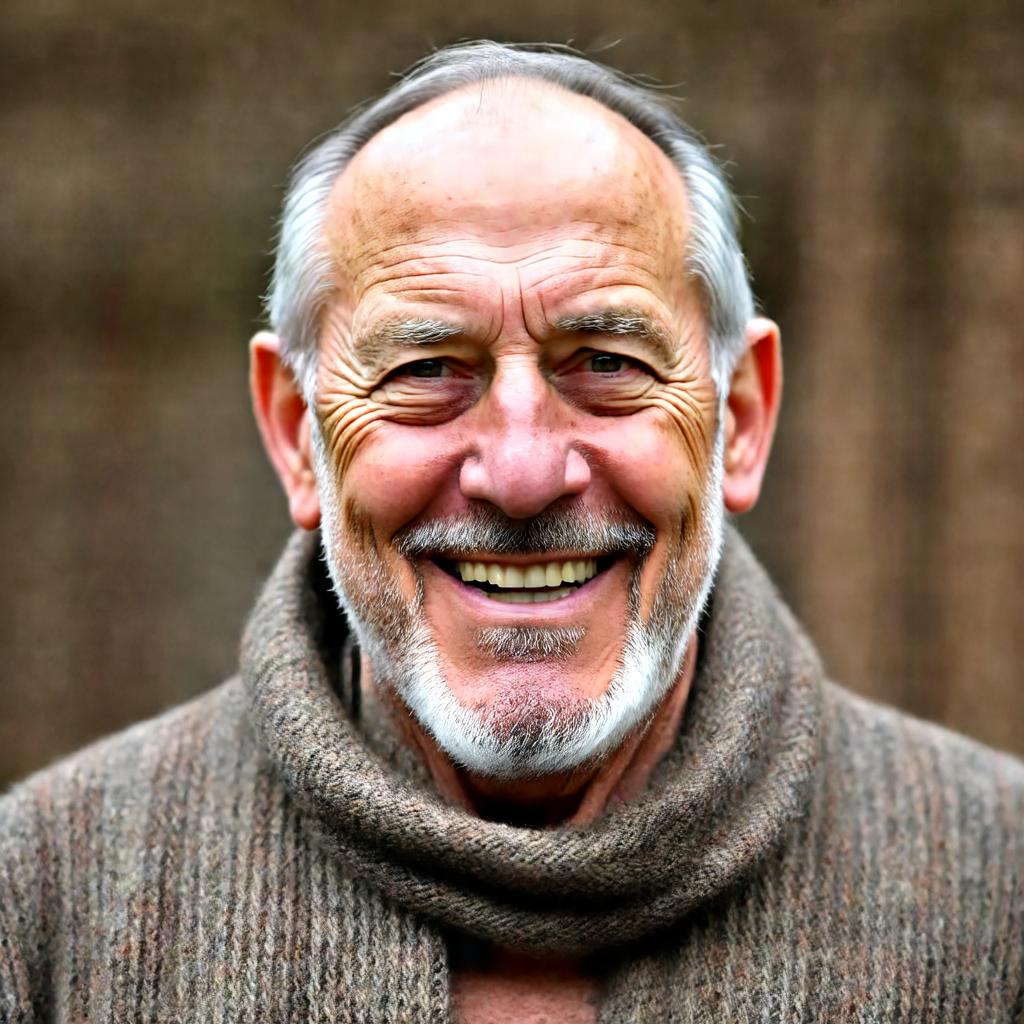} &
        \includegraphics[valign=c, width=\ww]{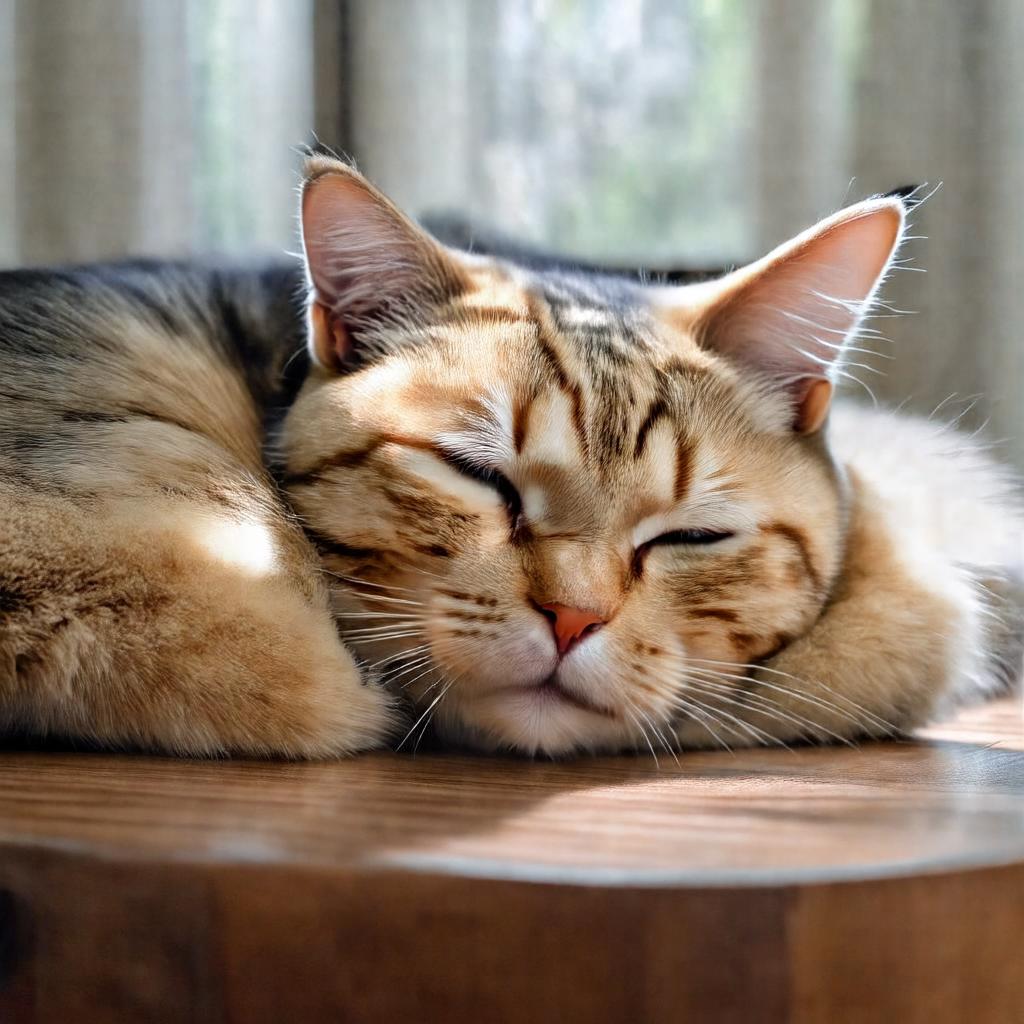} &
        \includegraphics[valign=c, width=\ww]{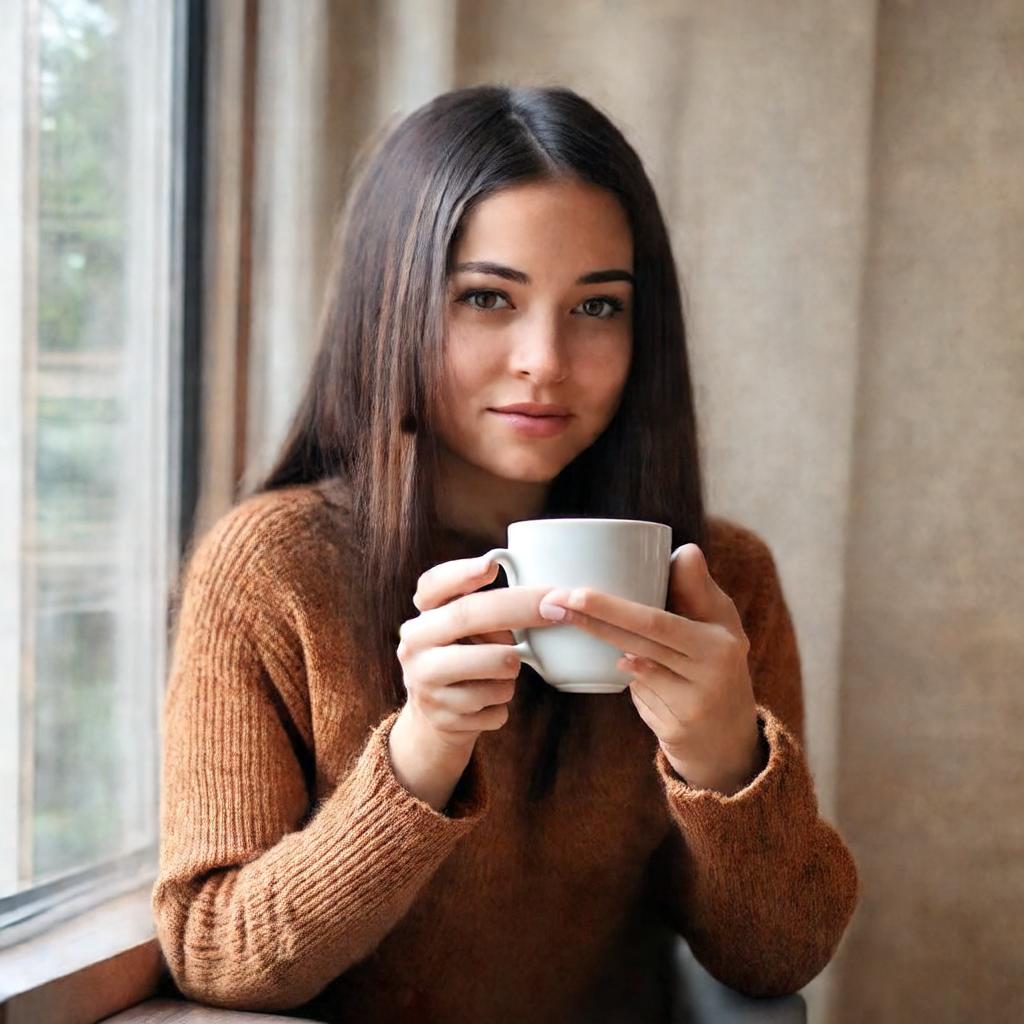}
        \vspace{1px}
        \\

        \rotatebox[origin=c]{90}{\vital{\footnotesize{$G_{7}$}}} &
        \includegraphics[valign=c, width=\ww]{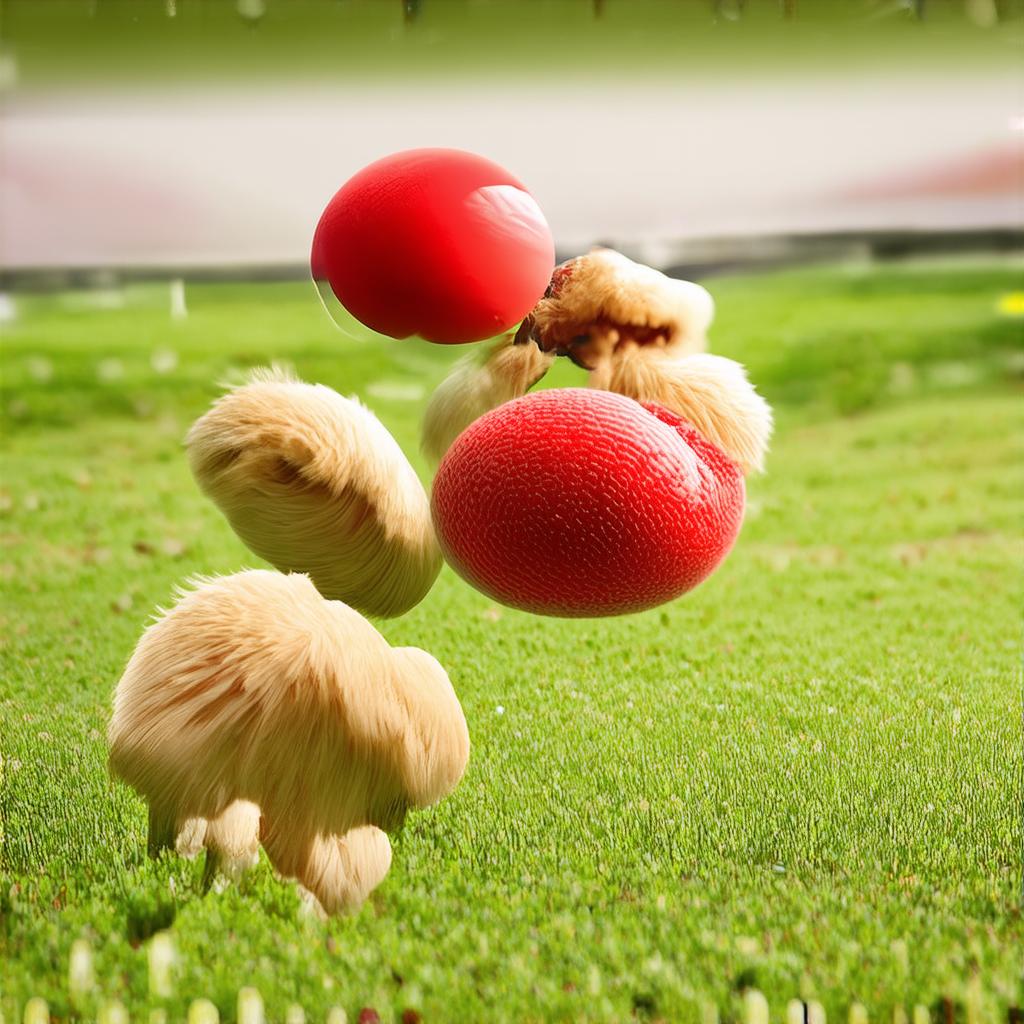} &
        \includegraphics[valign=c, width=\ww]{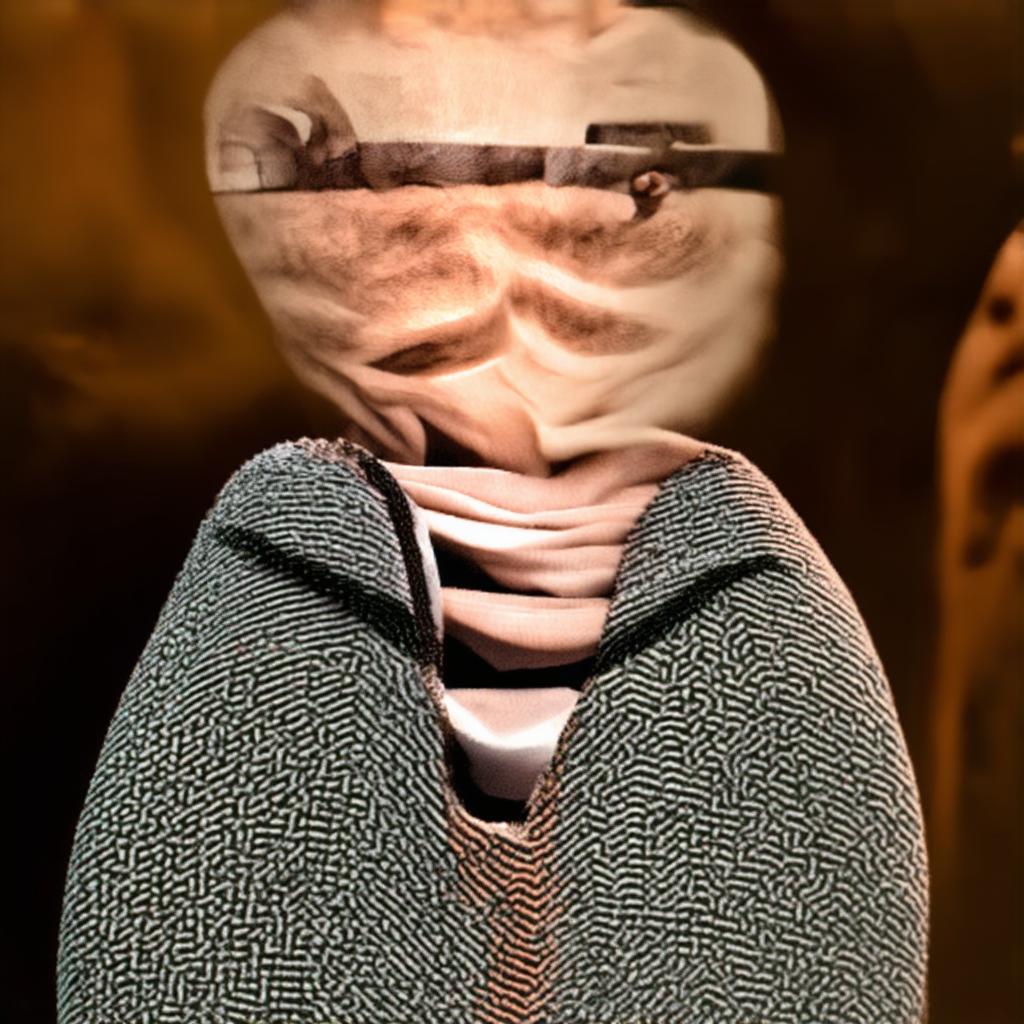} &
        \includegraphics[valign=c, width=\ww]{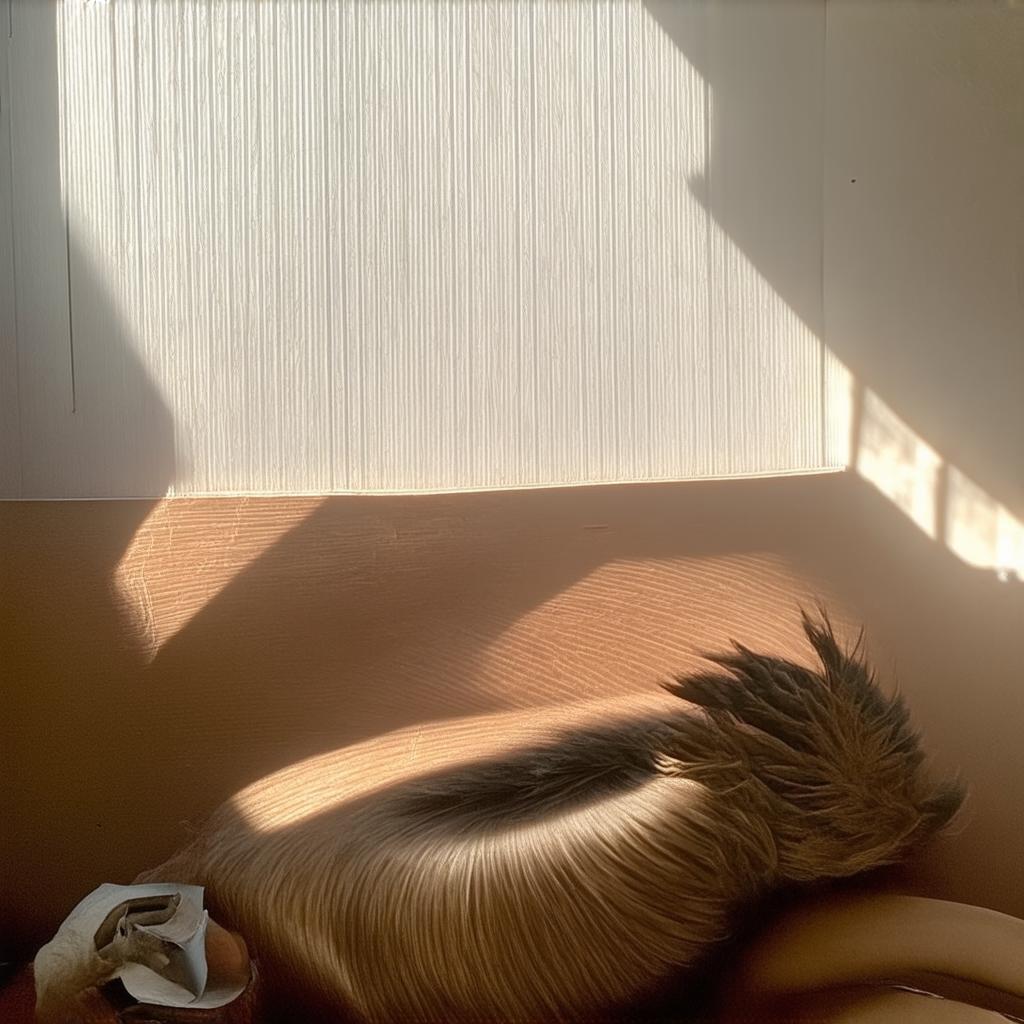} &
        \includegraphics[valign=c, width=\ww]{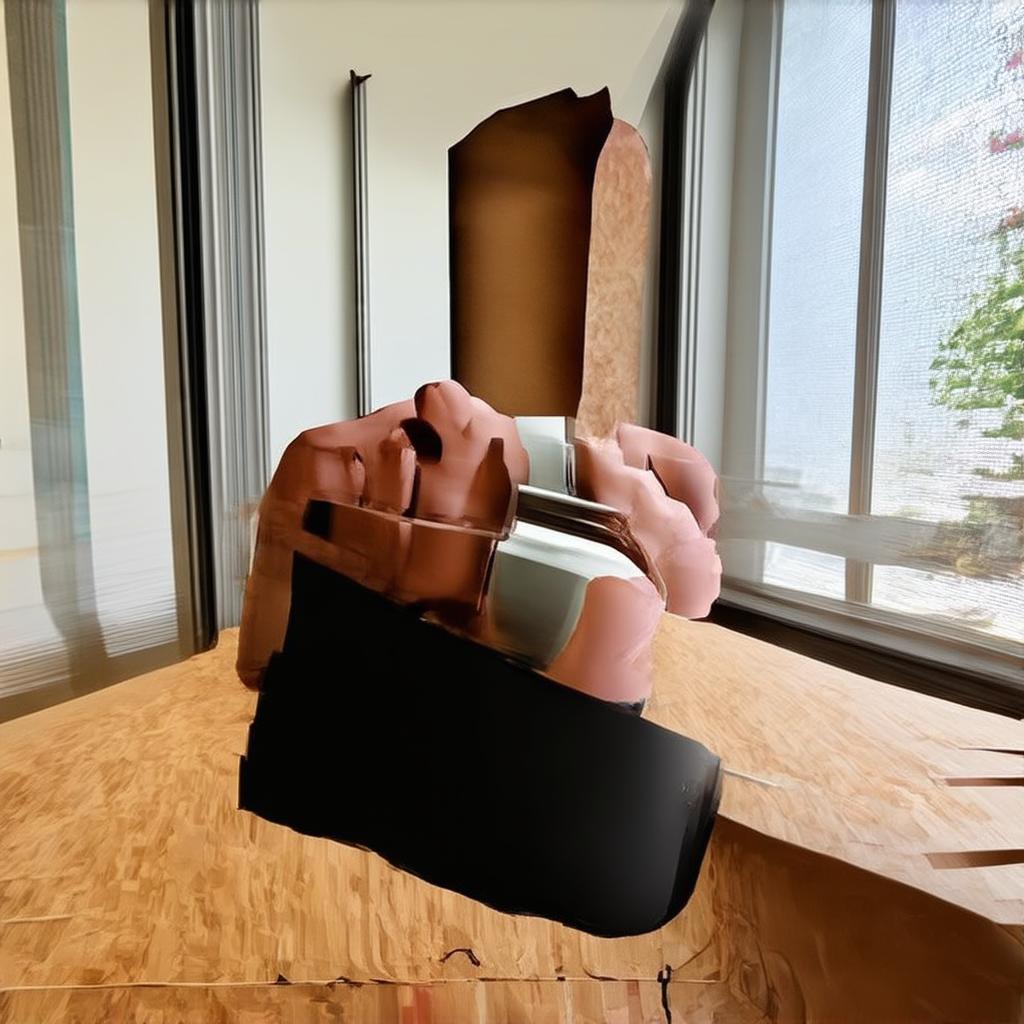}
        \vspace{1px}
        \\

        \rotatebox[origin=c]{90}{\vital{\footnotesize{$G_{8}$}}} &
        \includegraphics[valign=c, width=\ww]{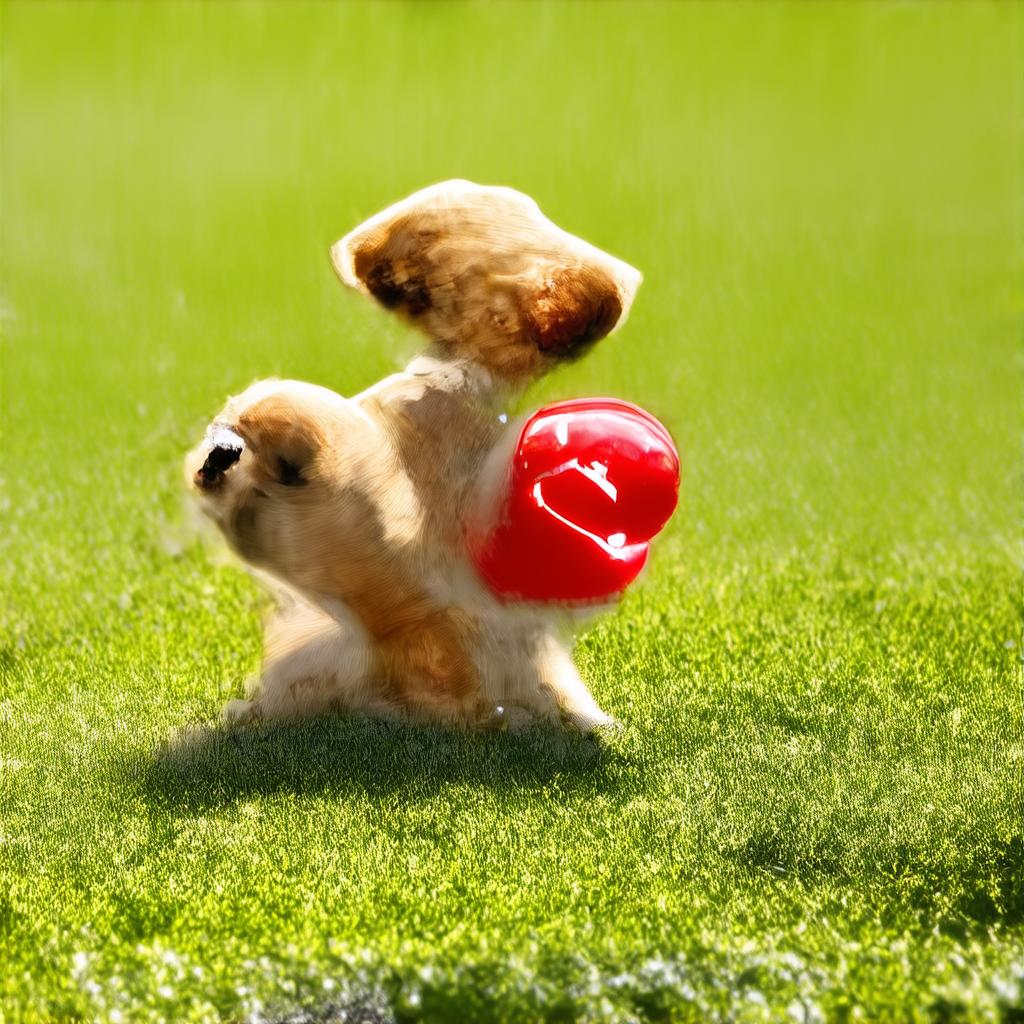} &
        \includegraphics[valign=c, width=\ww]{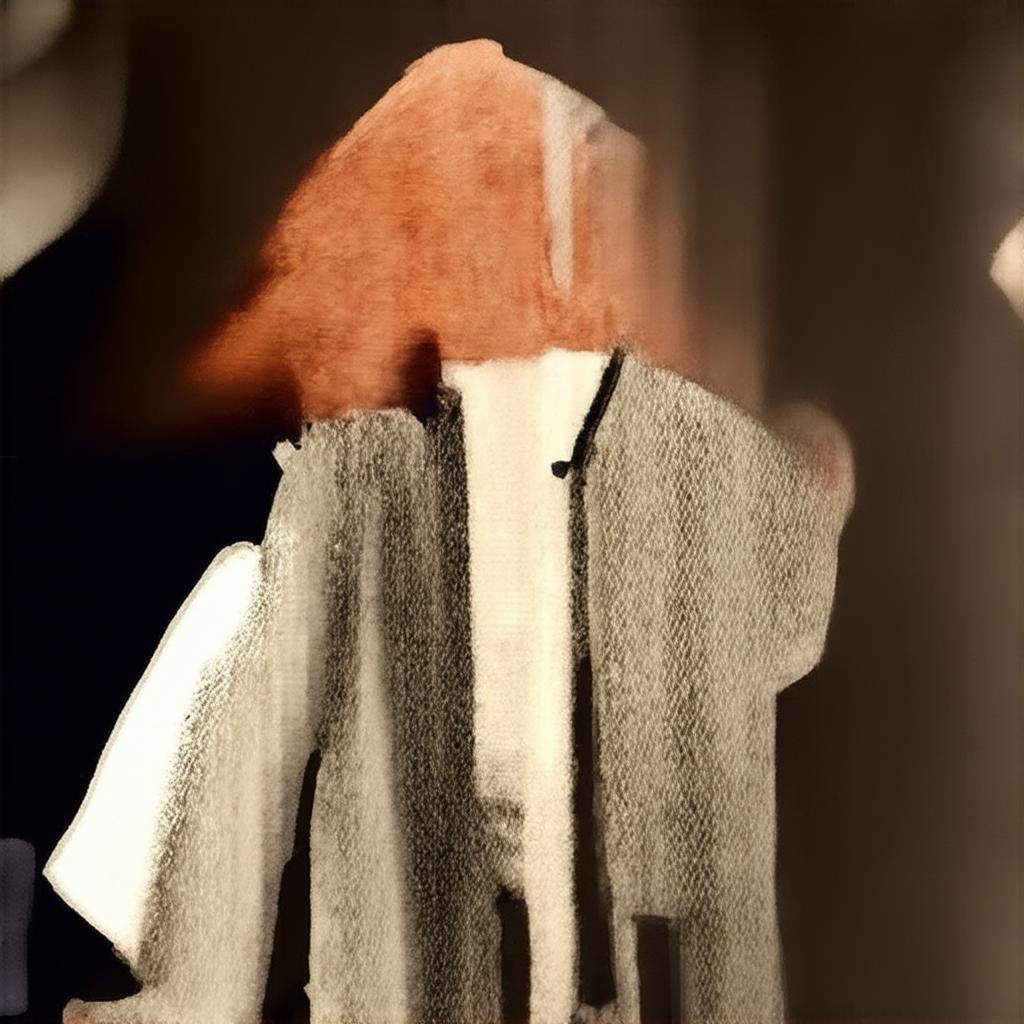} &
        \includegraphics[valign=c, width=\ww]{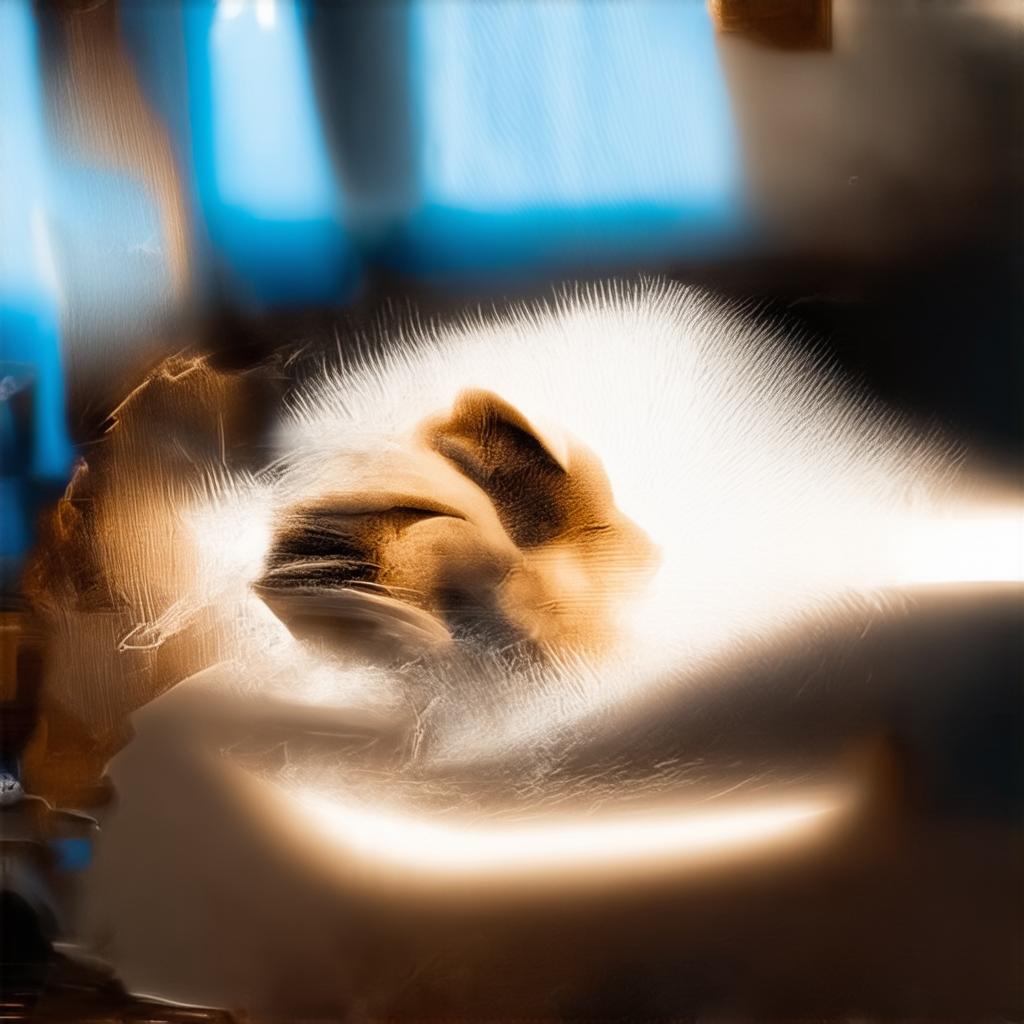} &
        \includegraphics[valign=c, width=\ww]{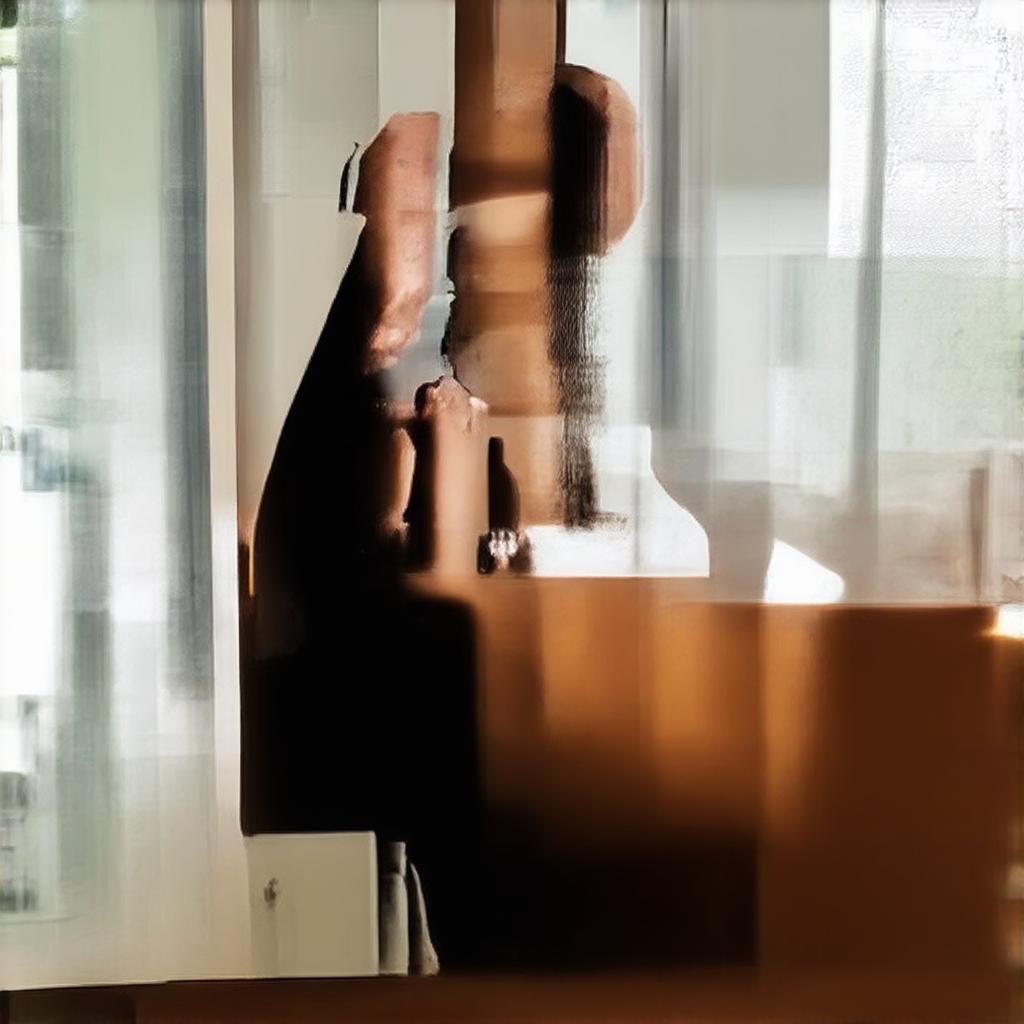}
        \vspace{1px}
        \\

        \rotatebox[origin=c]{90}{\vital{\footnotesize{$G_{9}$}}} &
        \includegraphics[valign=c, width=\ww]{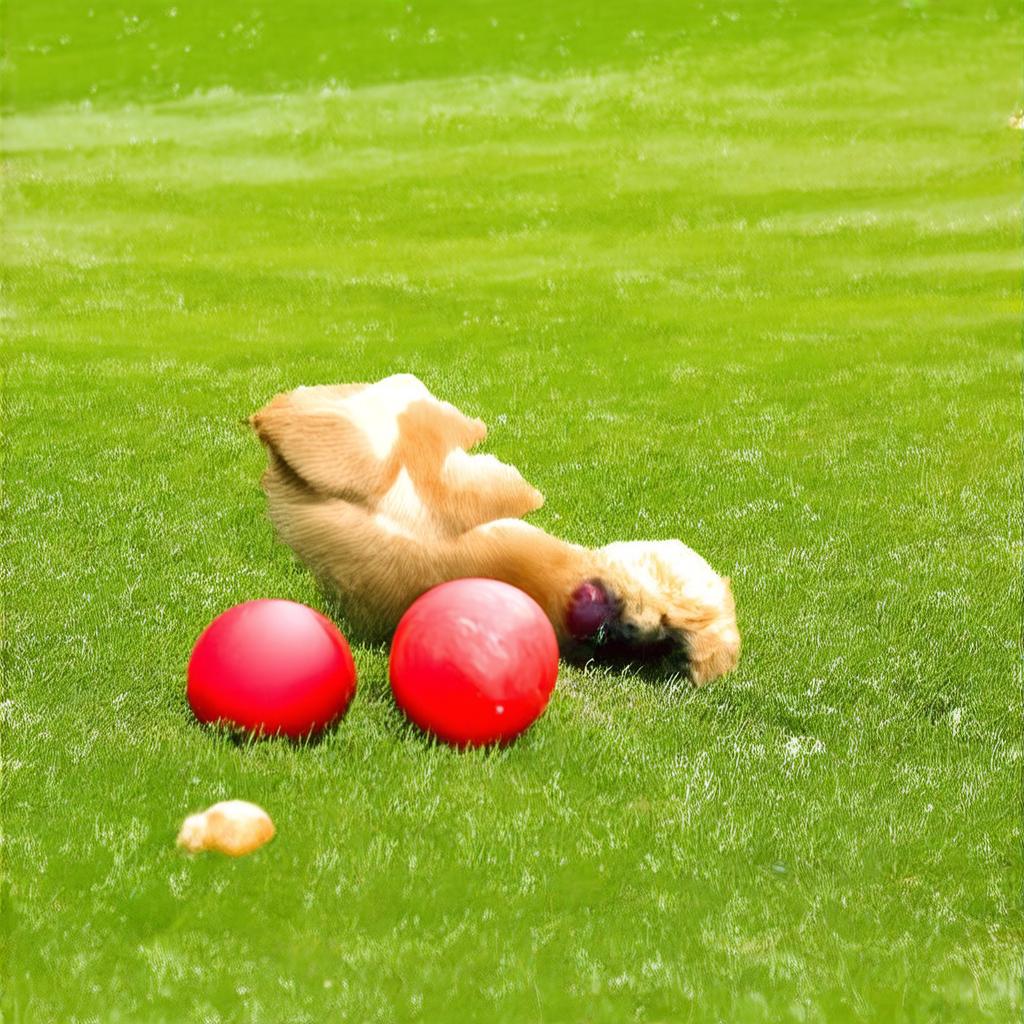} &
        \includegraphics[valign=c, width=\ww]{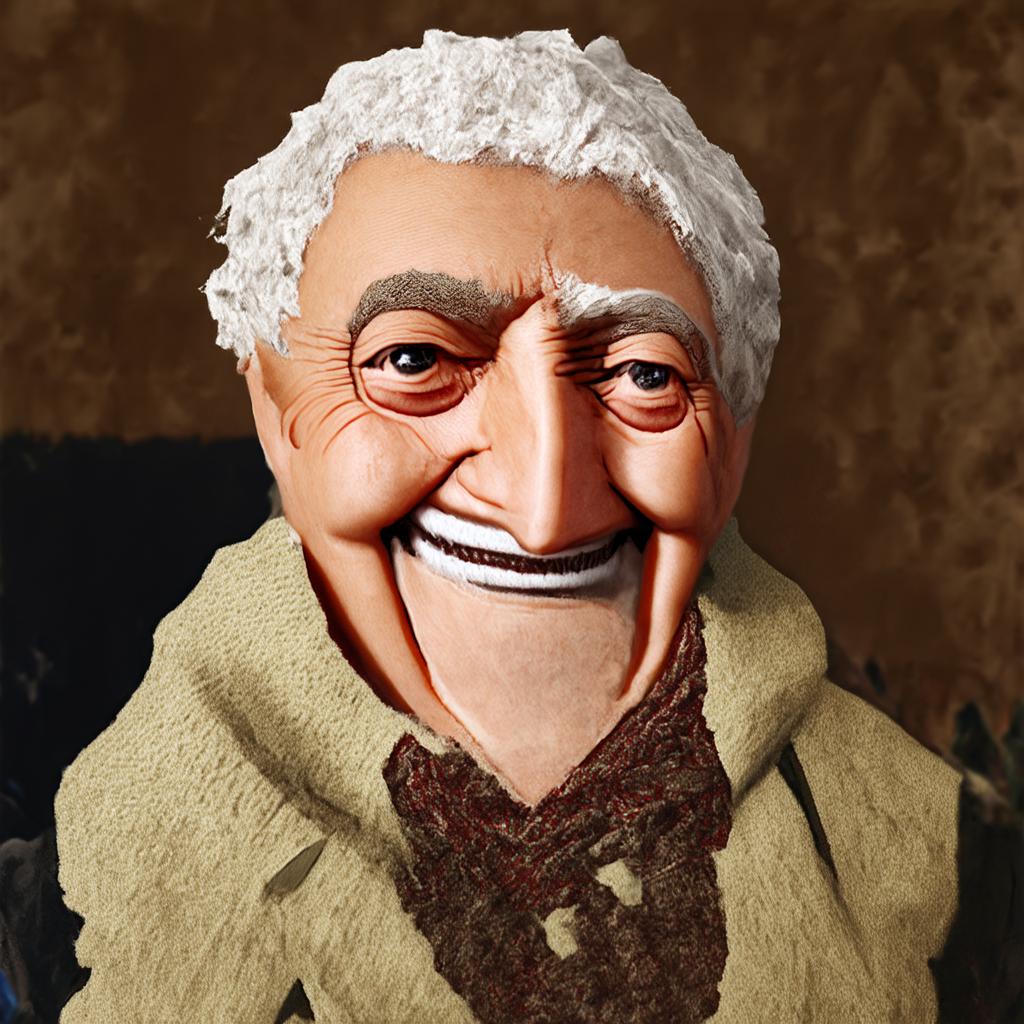} &
        \includegraphics[valign=c, width=\ww]{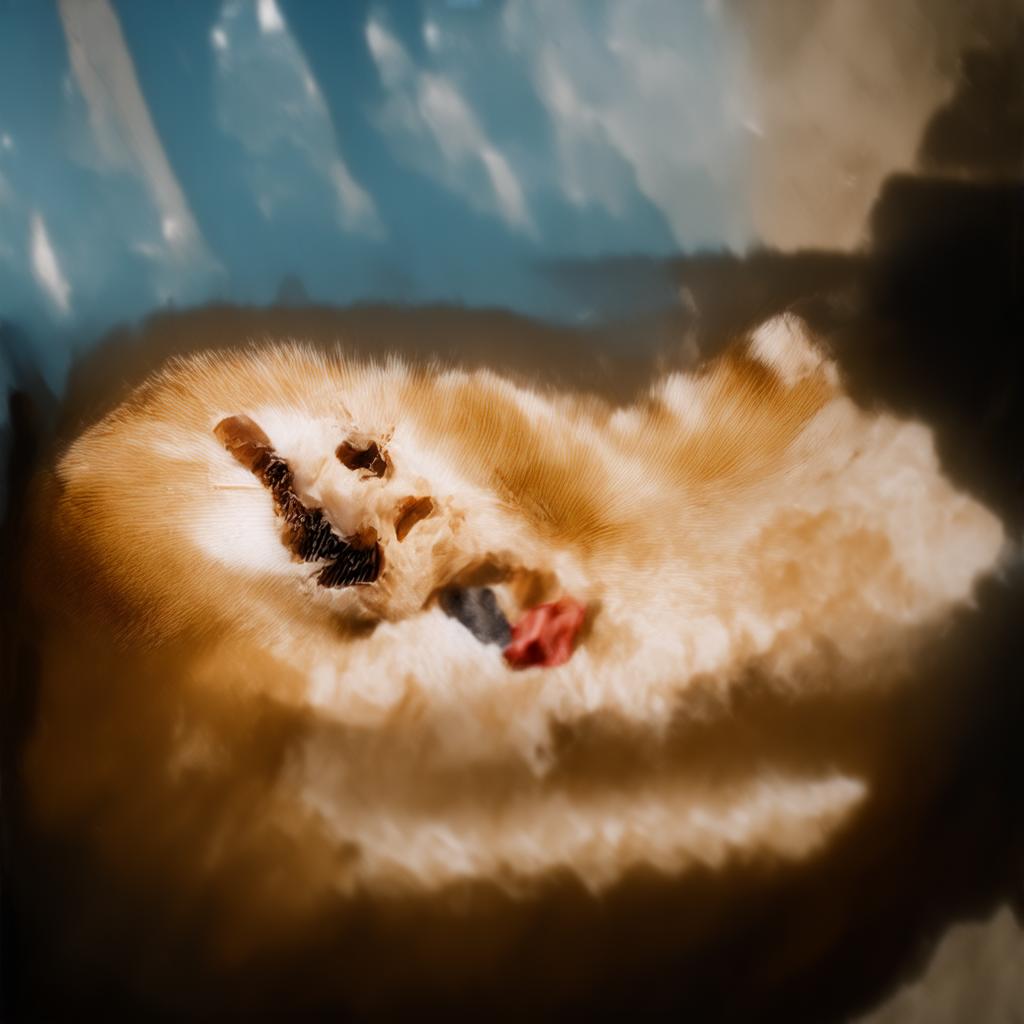} &
        \includegraphics[valign=c, width=\ww]{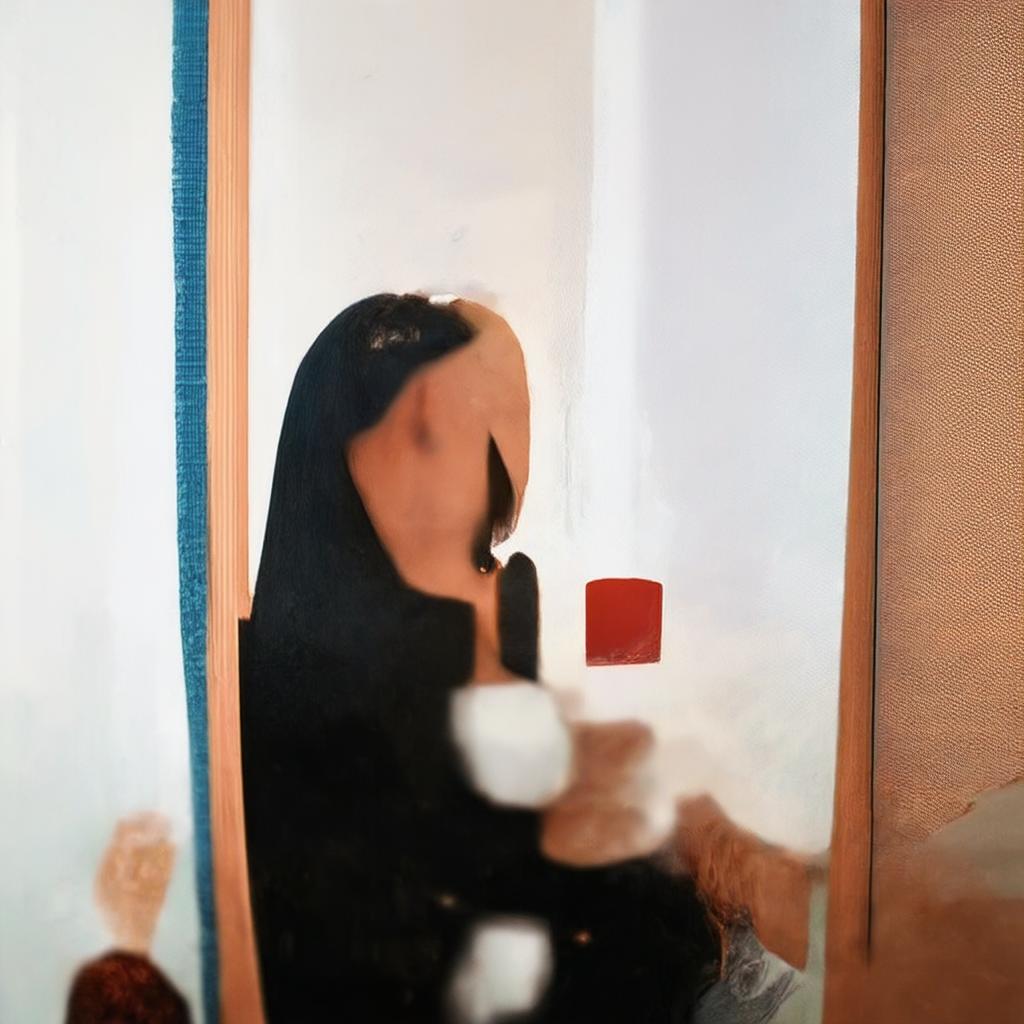}
        \vspace{1px}
        \\

        \rotatebox[origin=c]{90}{\nonvital{\footnotesize{$G_{21}$}}} &
        \includegraphics[valign=c, width=\ww]{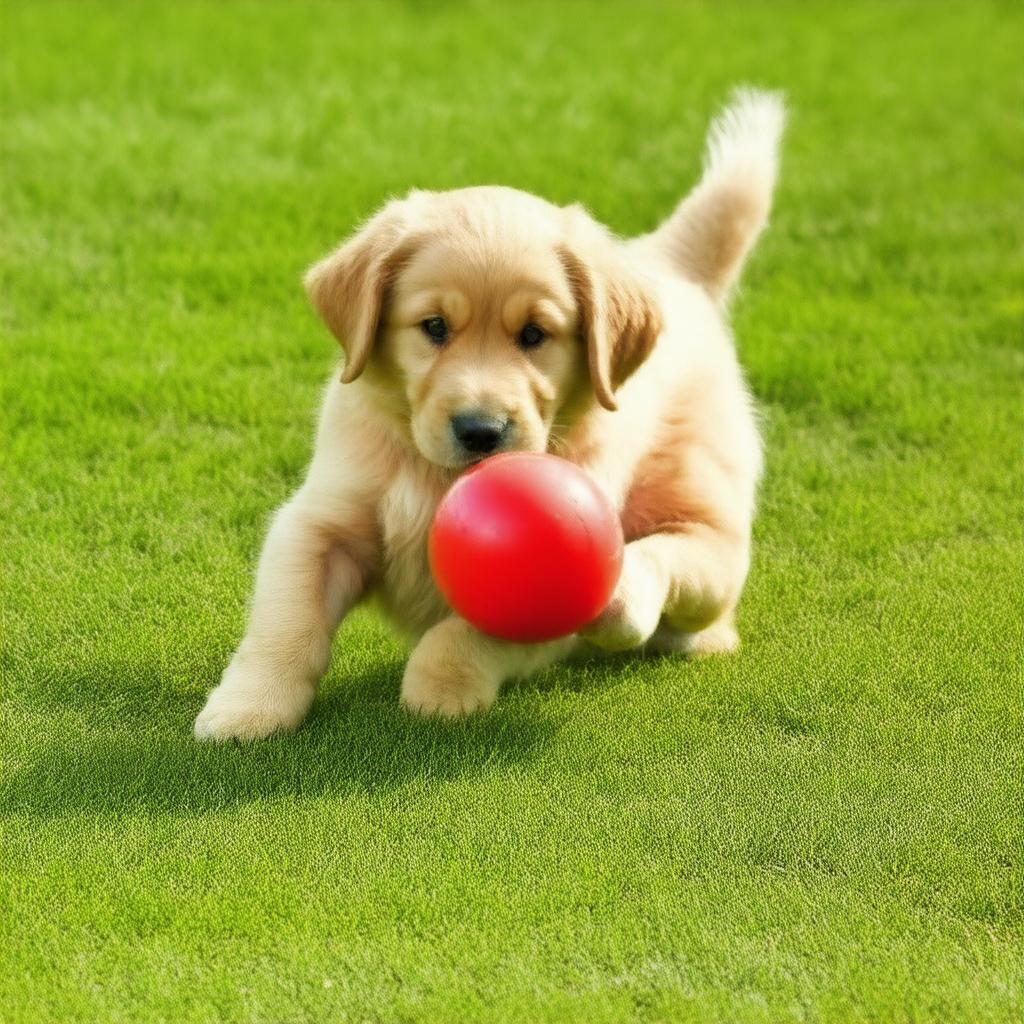} &
        \includegraphics[valign=c, width=\ww]{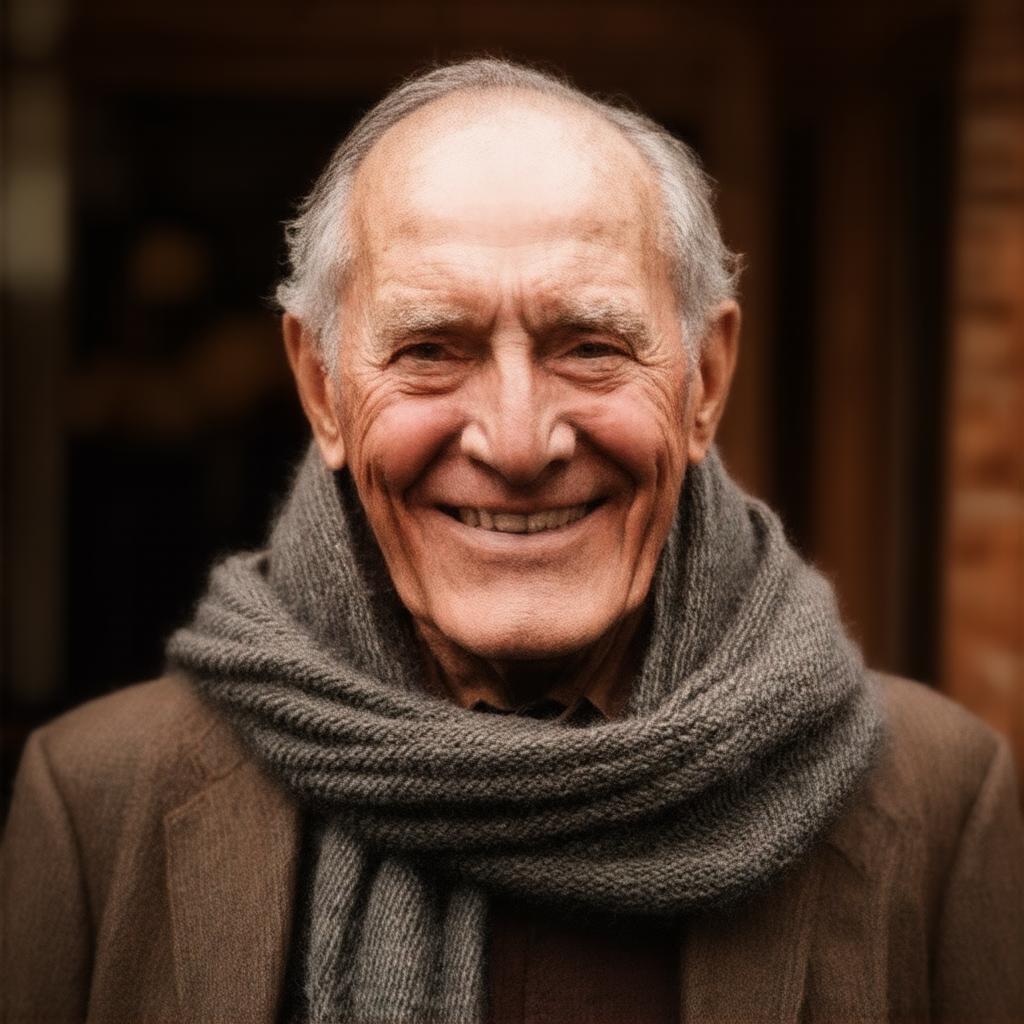} &
        \includegraphics[valign=c, width=\ww]{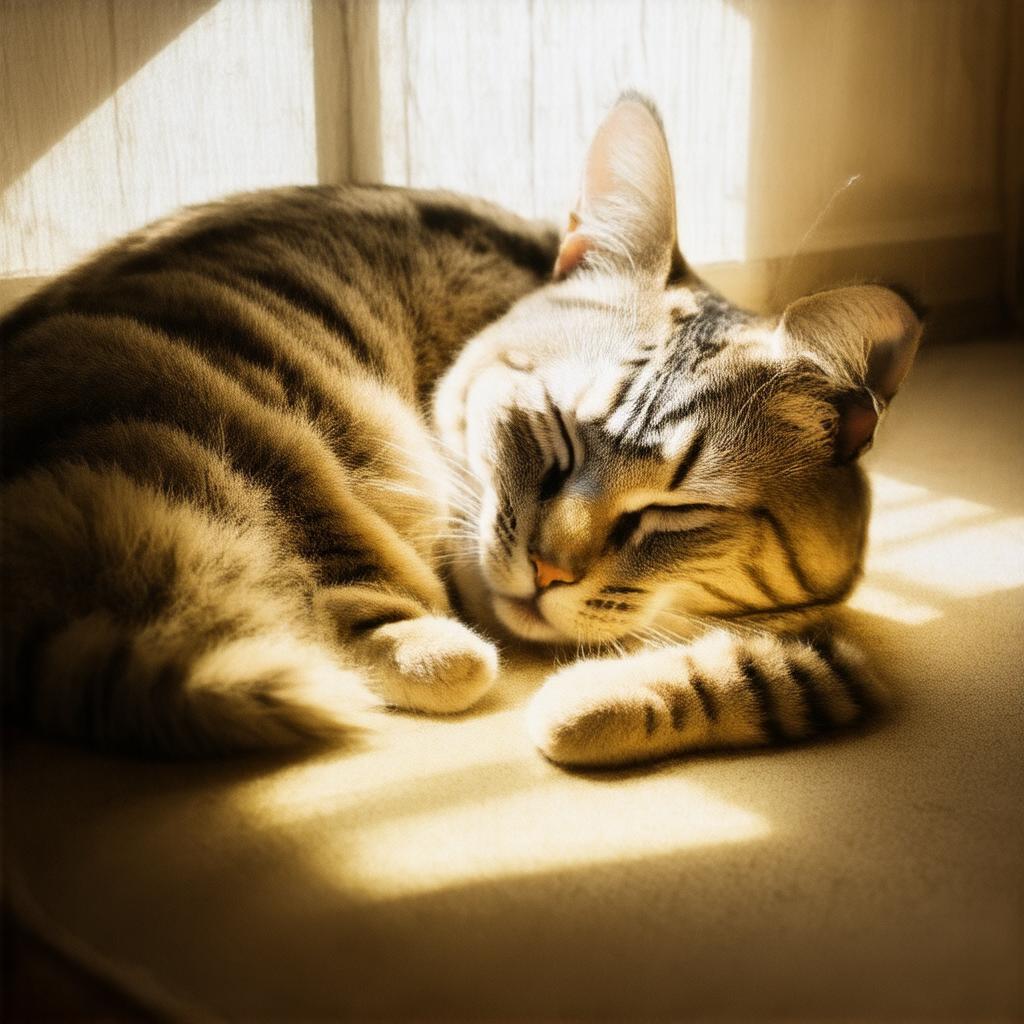} &
        \includegraphics[valign=c, width=\ww]{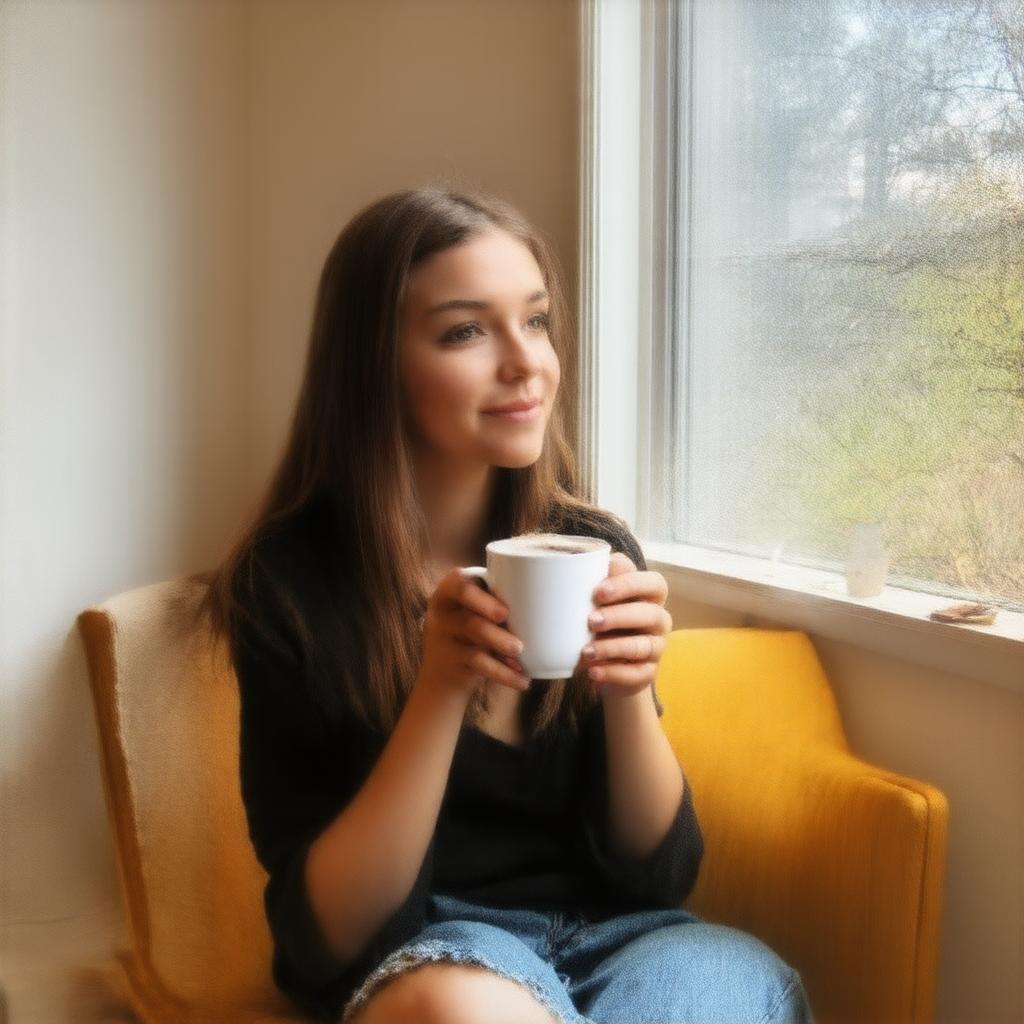}
        \vspace{1px}
        \\

    \end{tabular}
    \caption{\textbf{Layer Removal Qualitative Comparison Stable Diffusion 3.} As explained in \Cref{sec:sd3_results}, we illustrate the qualitative differences between \vital{vital} and \nonvital{non-vital} layers. While bypassing  \nonvital{non-vital} layers (\nonvital{$G_{1}$} and \nonvital{$G_{21}$}) results in modest alterations, bypassing \vital{vital} layers leads to significant changes: complete noise generation (\vital{$G_{0}$}), or severe distortions (\vital{$G_{7}$}, \vital{$G_{8}$} and \vital{$G_{9}$}). For a quantitative comparison, please refer to \Cref{fig:layer_removal_quantiative_sd3}}
    \label{fig:layer_removal_qualitative_sd3}
\end{figure}

\begin{figure*}[tp]
    \centering
    \setlength{\tabcolsep}{0.6pt}
    \renewcommand{\arraystretch}{0.8}
    \setlength{\ww}{0.19\linewidth}
    \begin{tabular}{c @{\hspace{10\tabcolsep}} cccc}

        \includegraphics[valign=c, width=\ww]{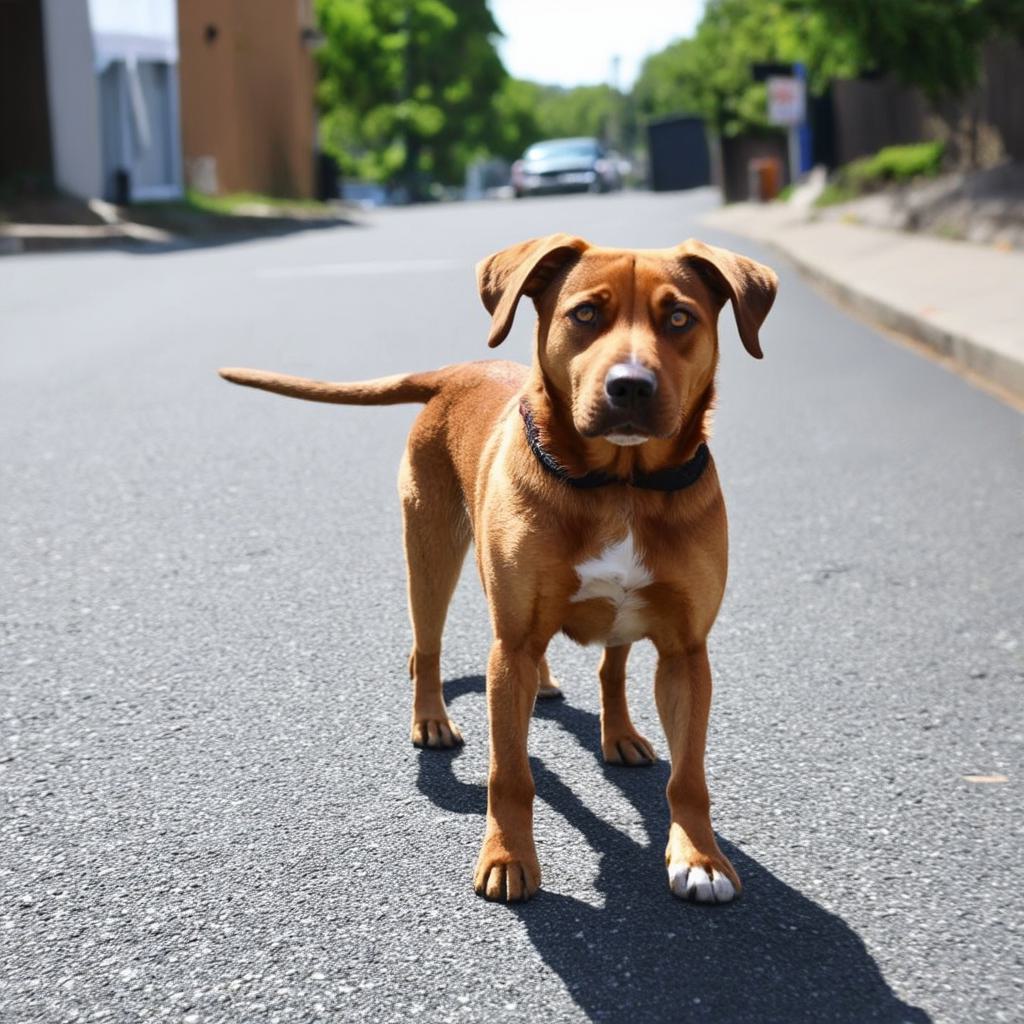} &
        \includegraphics[valign=c, width=\ww]{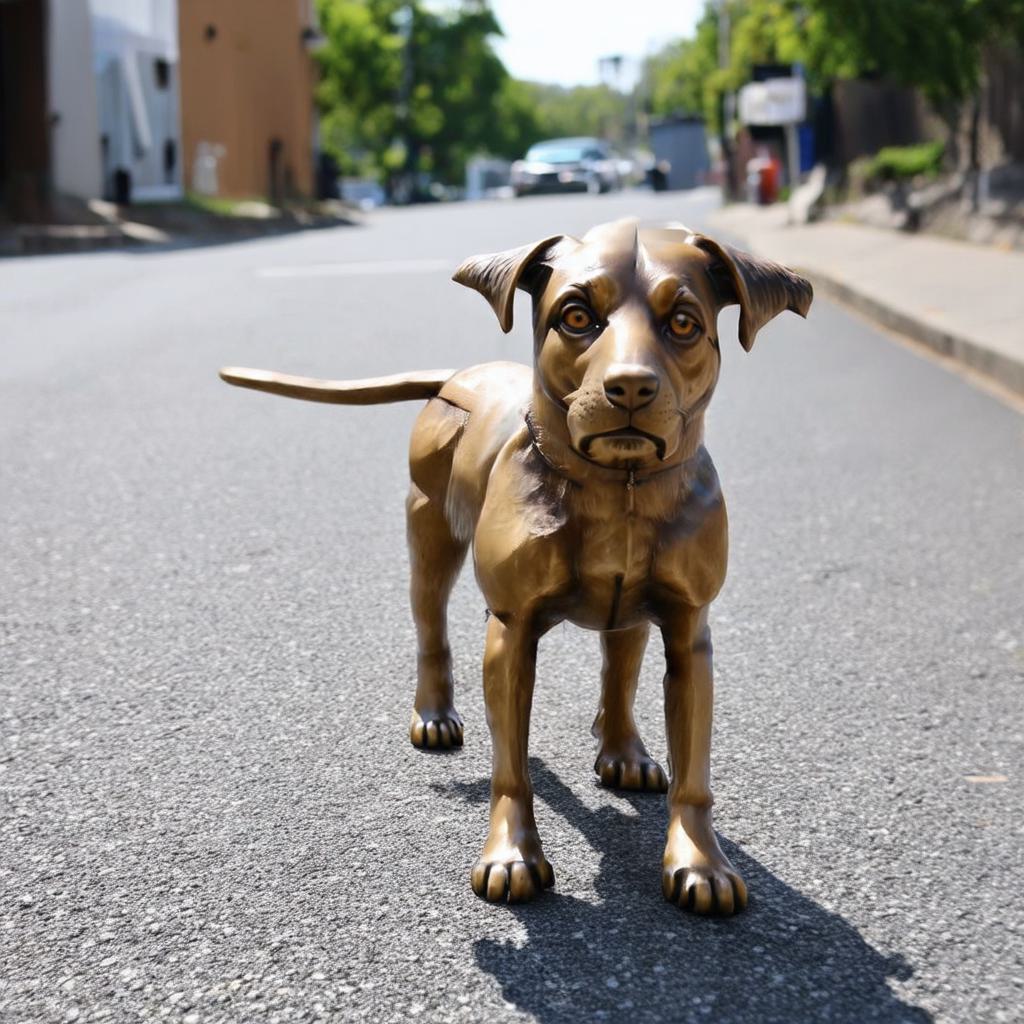} &
        \includegraphics[valign=c, width=\ww]{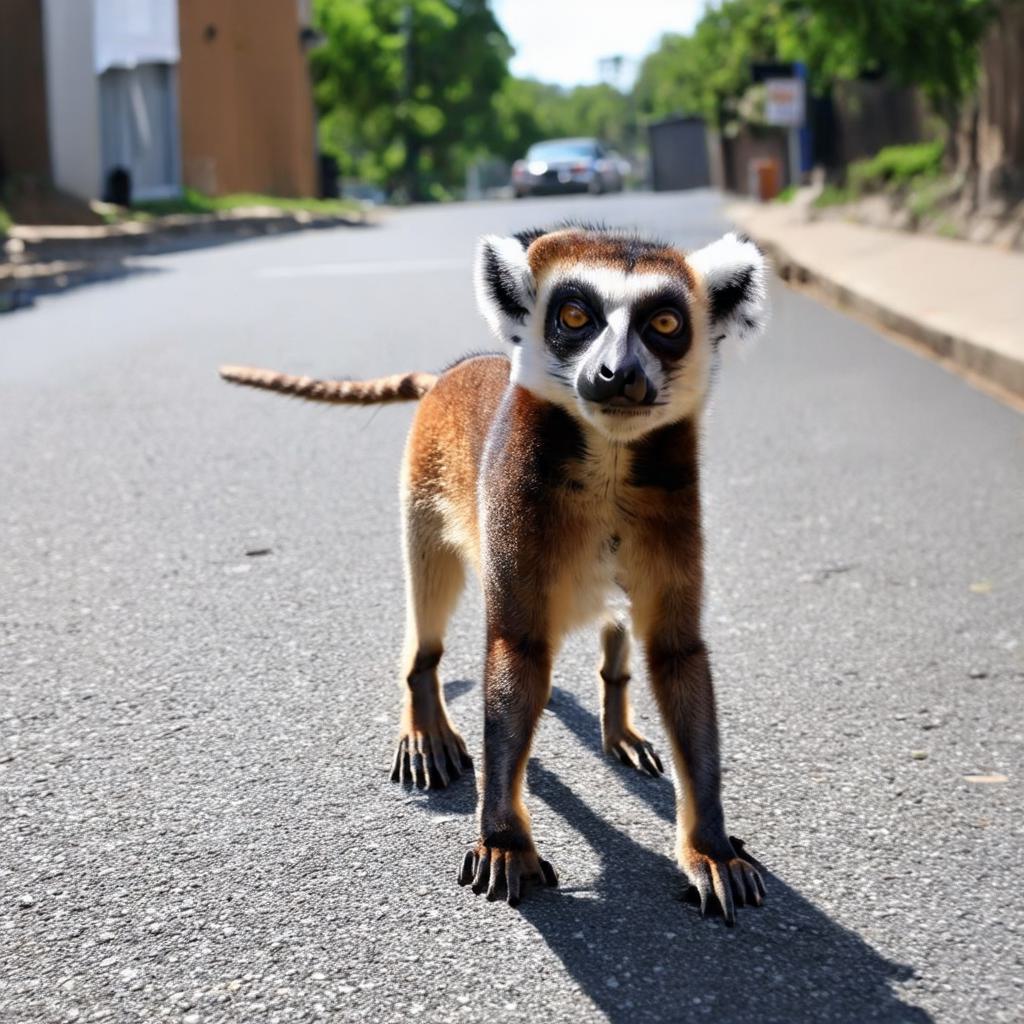} &
        \includegraphics[valign=c, width=\ww]{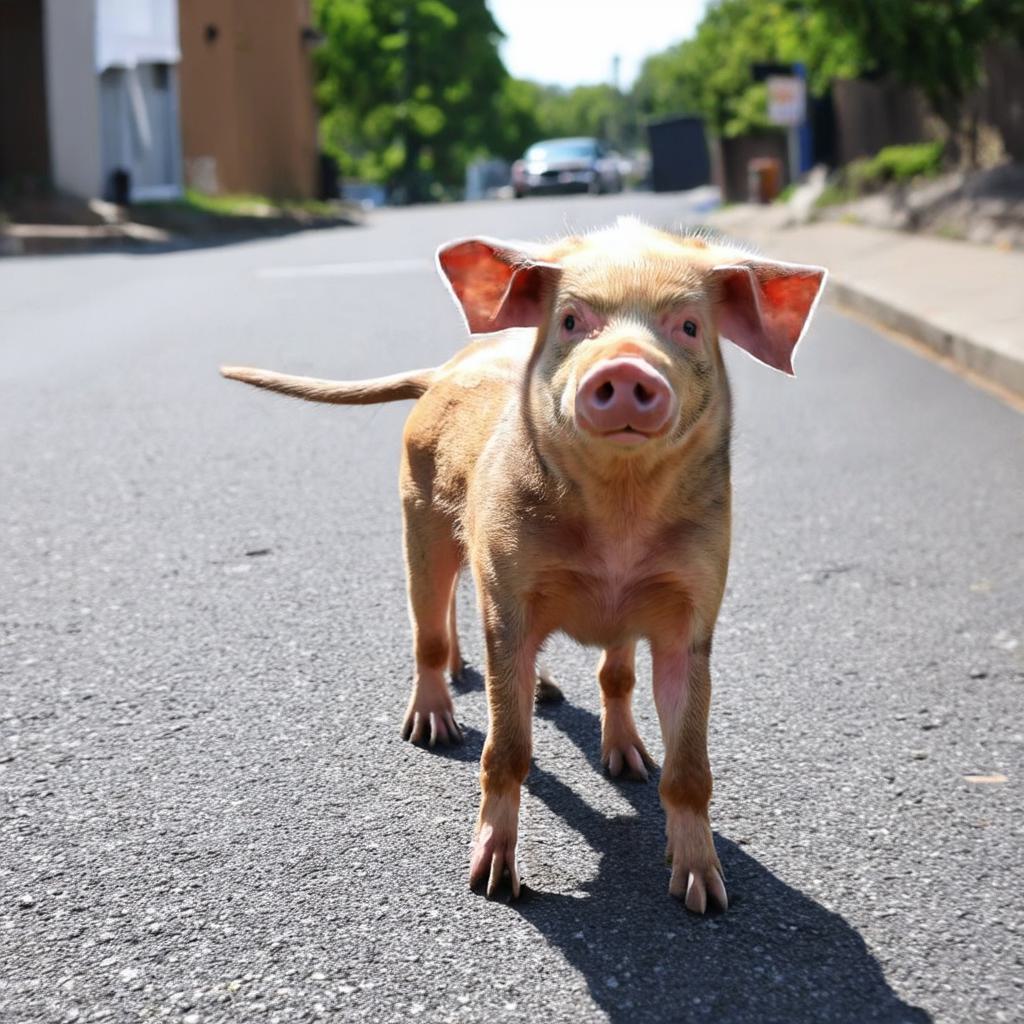} &
        \includegraphics[valign=c, width=\ww]{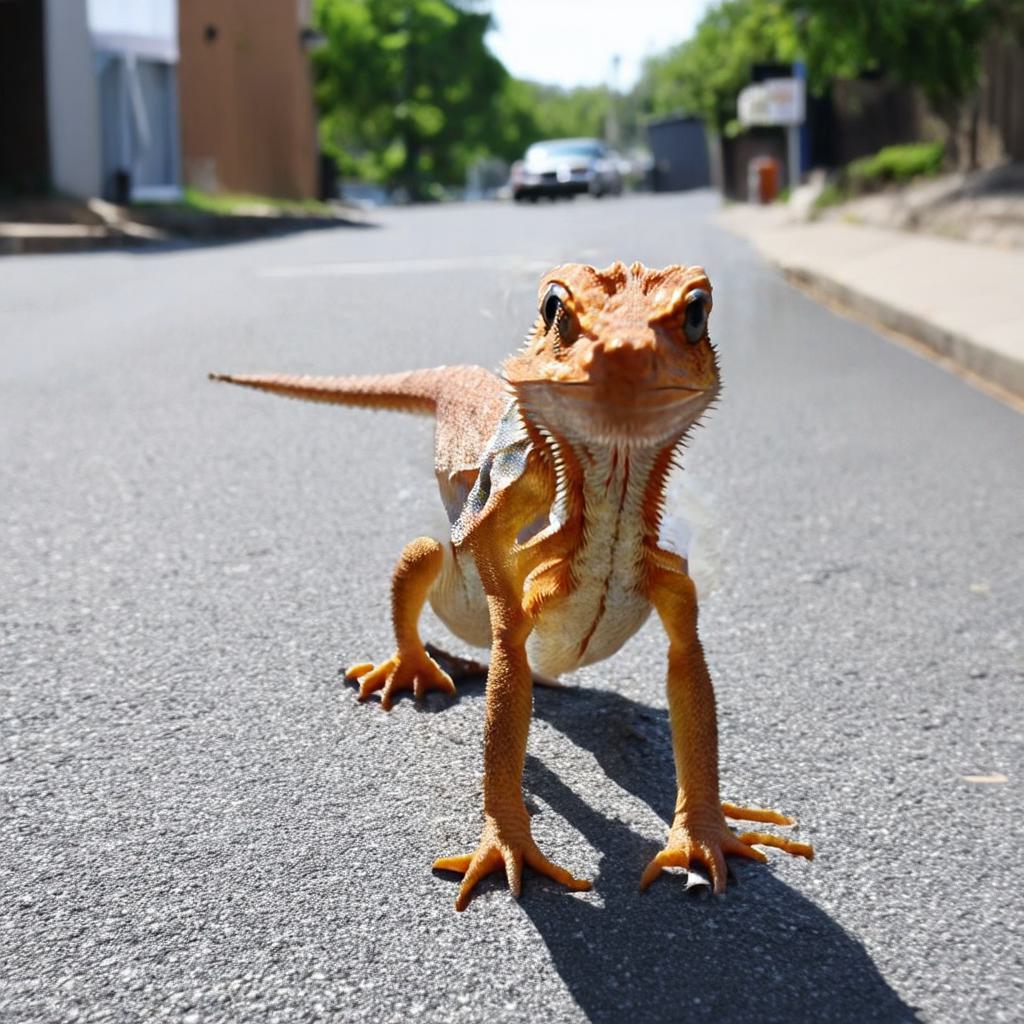} 
        \vspace{2px}
        \\

        \small{Input} &
        \small{\prompt{A dog statue}} &
        \small{\prompt{A lemur}} &
        \small{\prompt{A pig}} &
        \small{\prompt{A gecko}}
        \vspace{15px}
        \\

        \includegraphics[valign=c, width=\ww]{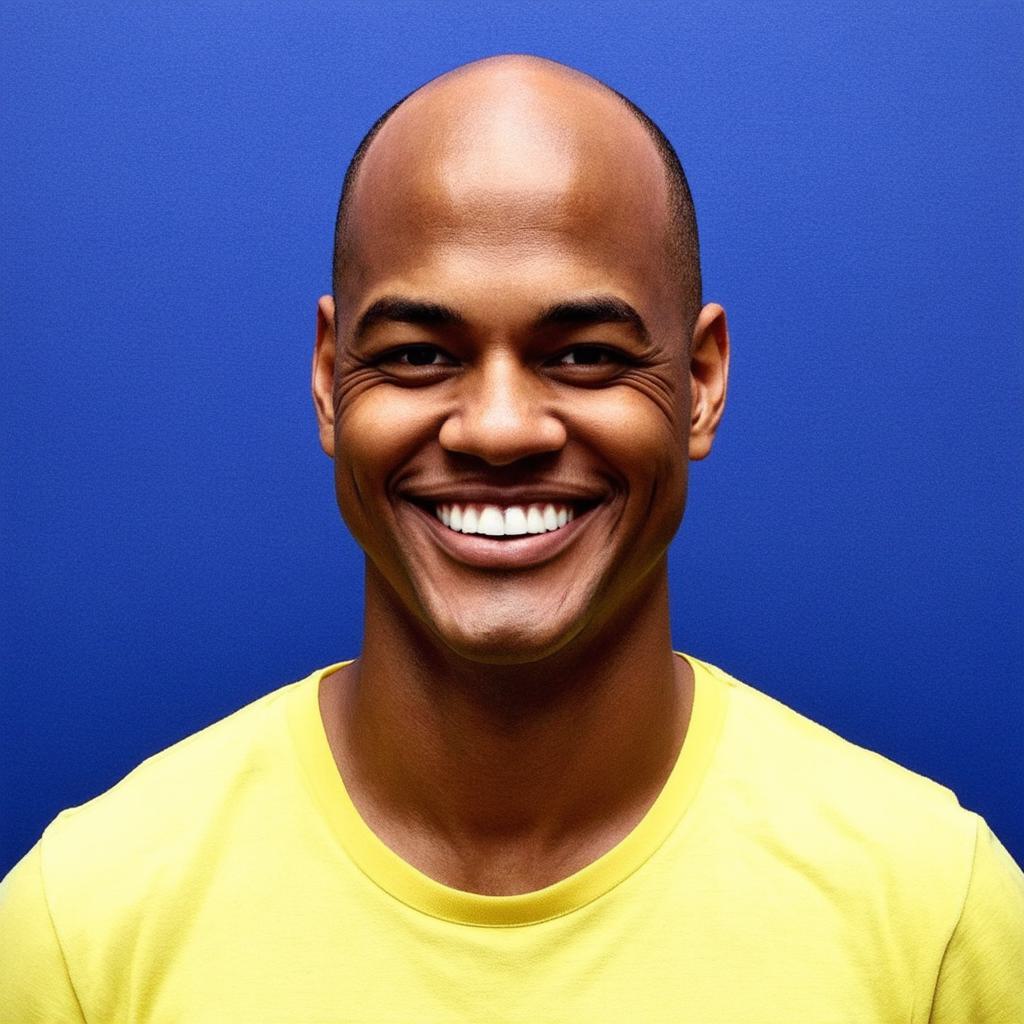} &
        \includegraphics[valign=c, width=\ww]{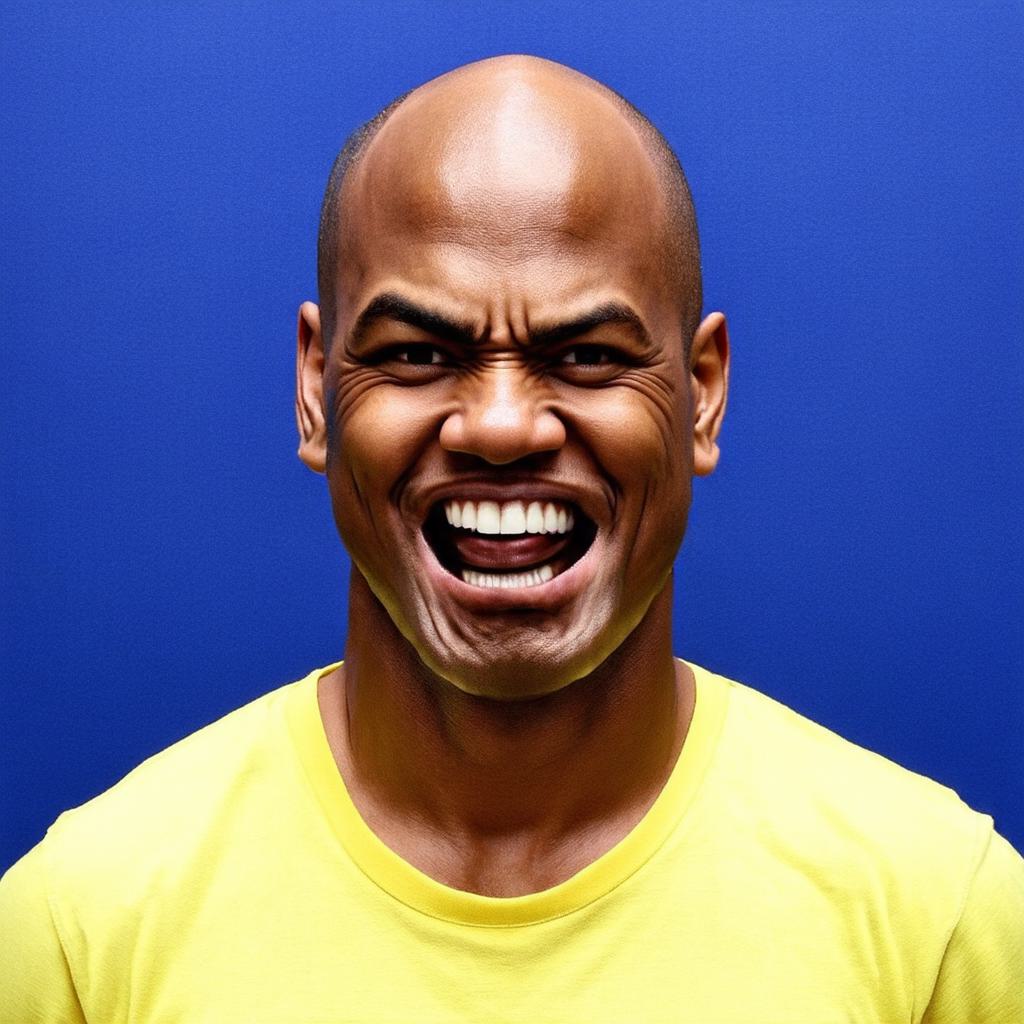} &
        \includegraphics[valign=c, width=\ww]{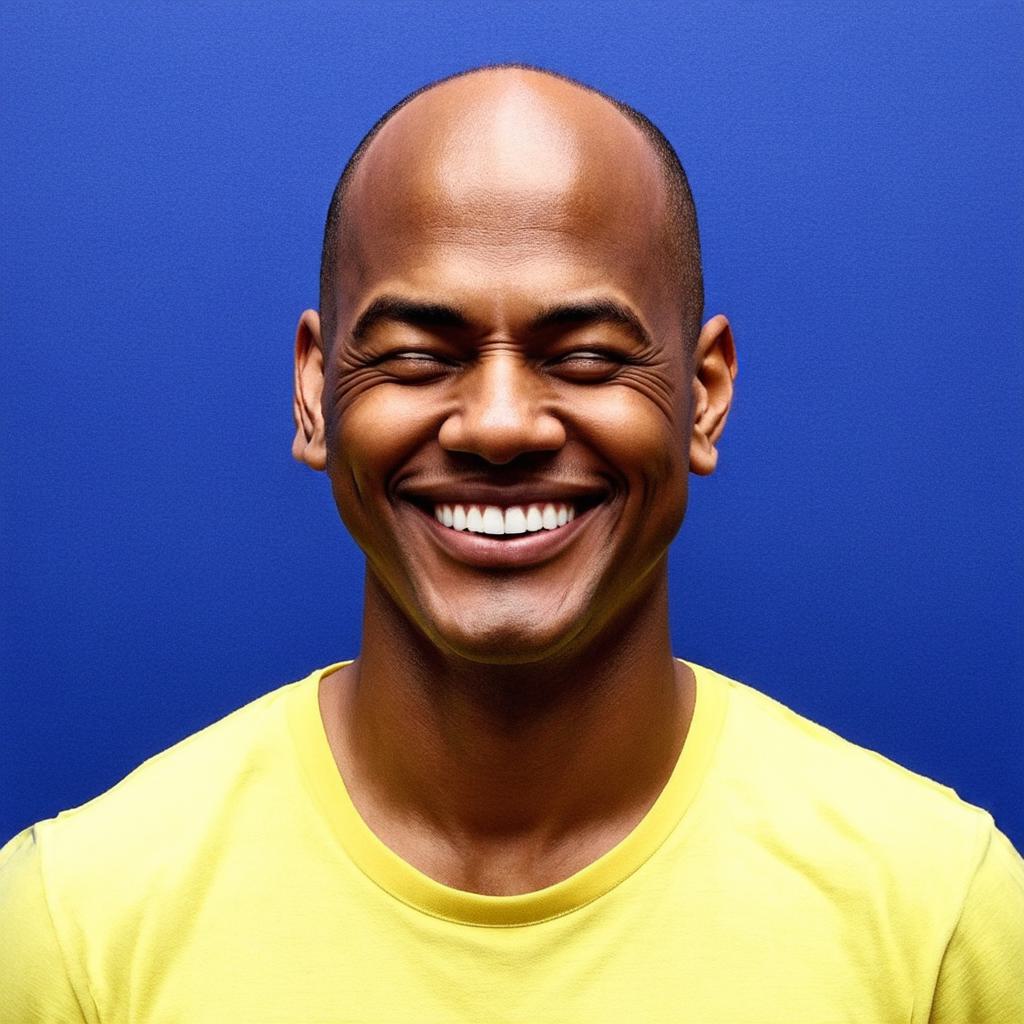} &
        \includegraphics[valign=c, width=\ww]{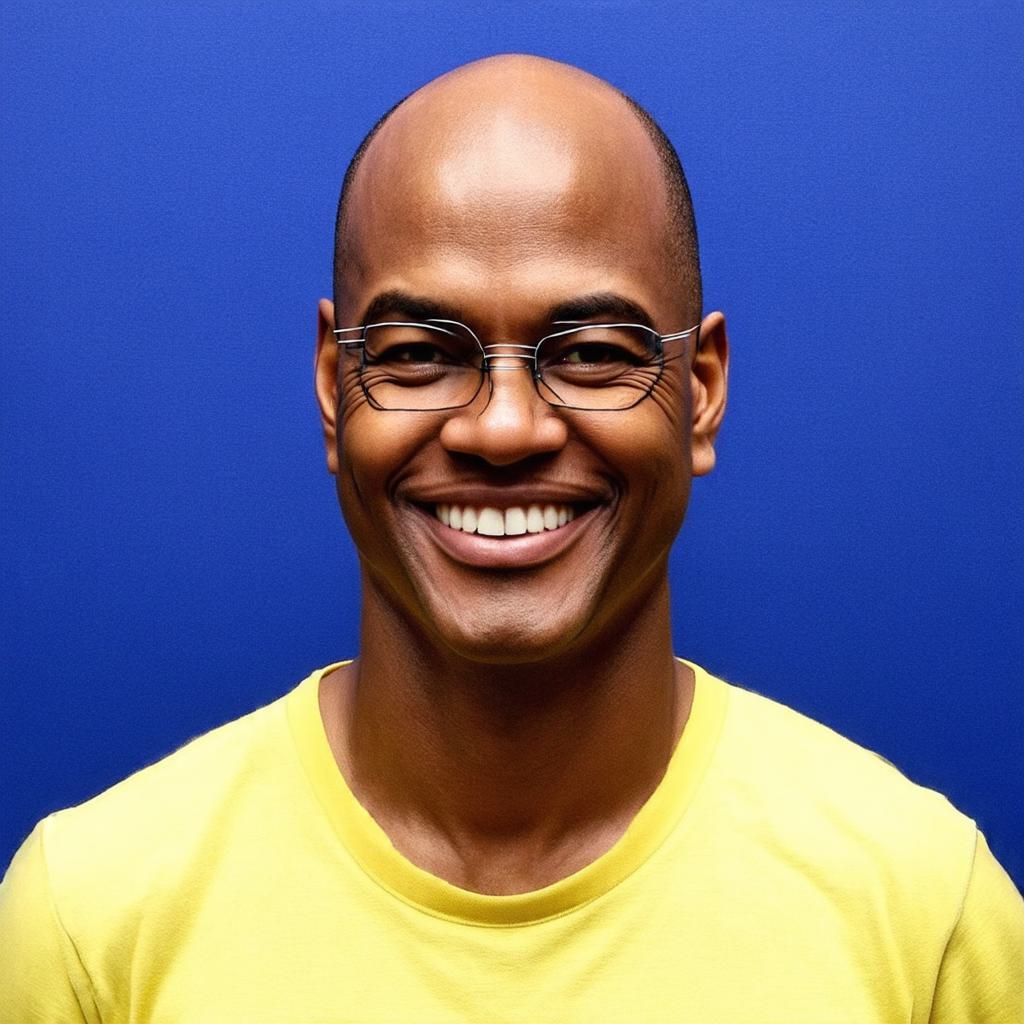} &
        \includegraphics[valign=c, width=\ww]{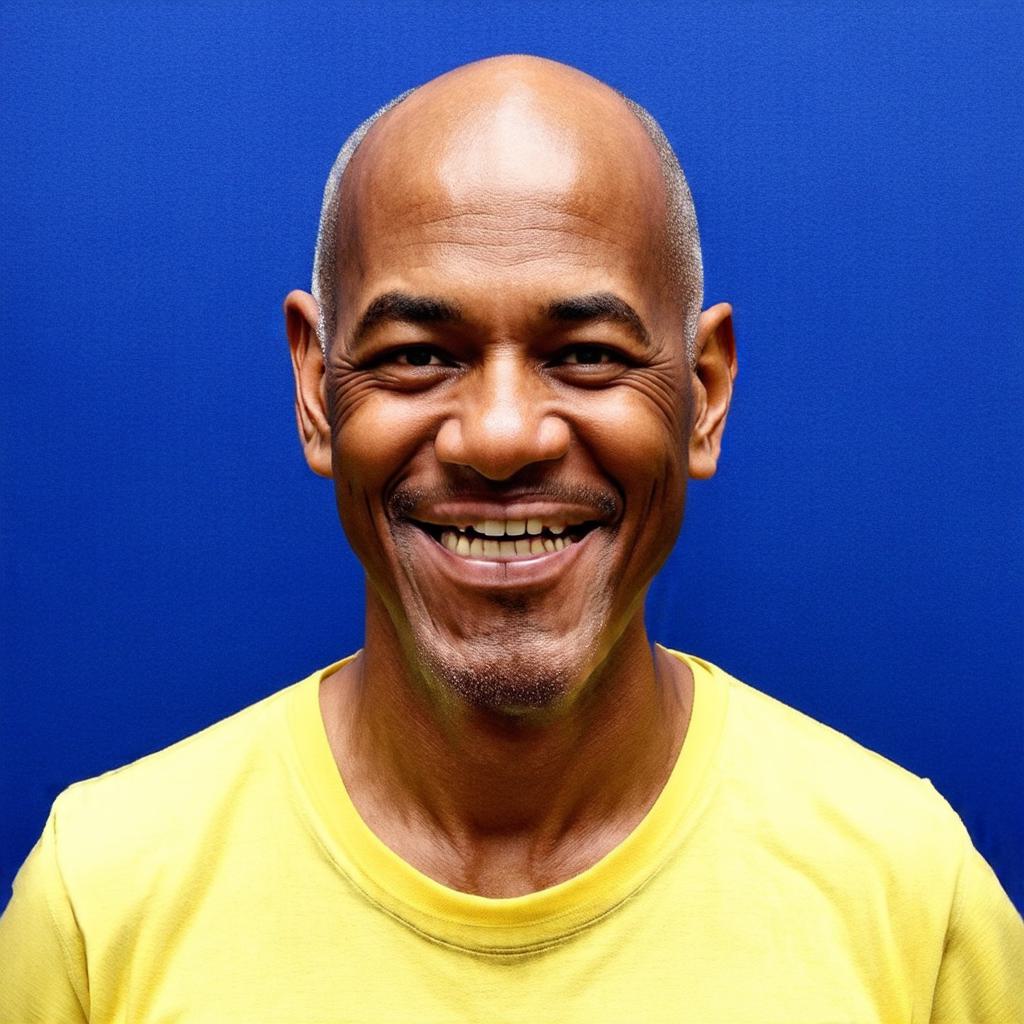} 
        \vspace{2px}
        \\

        \small{Input} &
        \small{\prompt{Angry}} &
        \small{\prompt{Closing his eyes}} &
        \small{\prompt{Wearing glasses}} &
        \small{\prompt{An old man}}
        \vspace{15px}
        \\

        \includegraphics[valign=c, width=\ww]{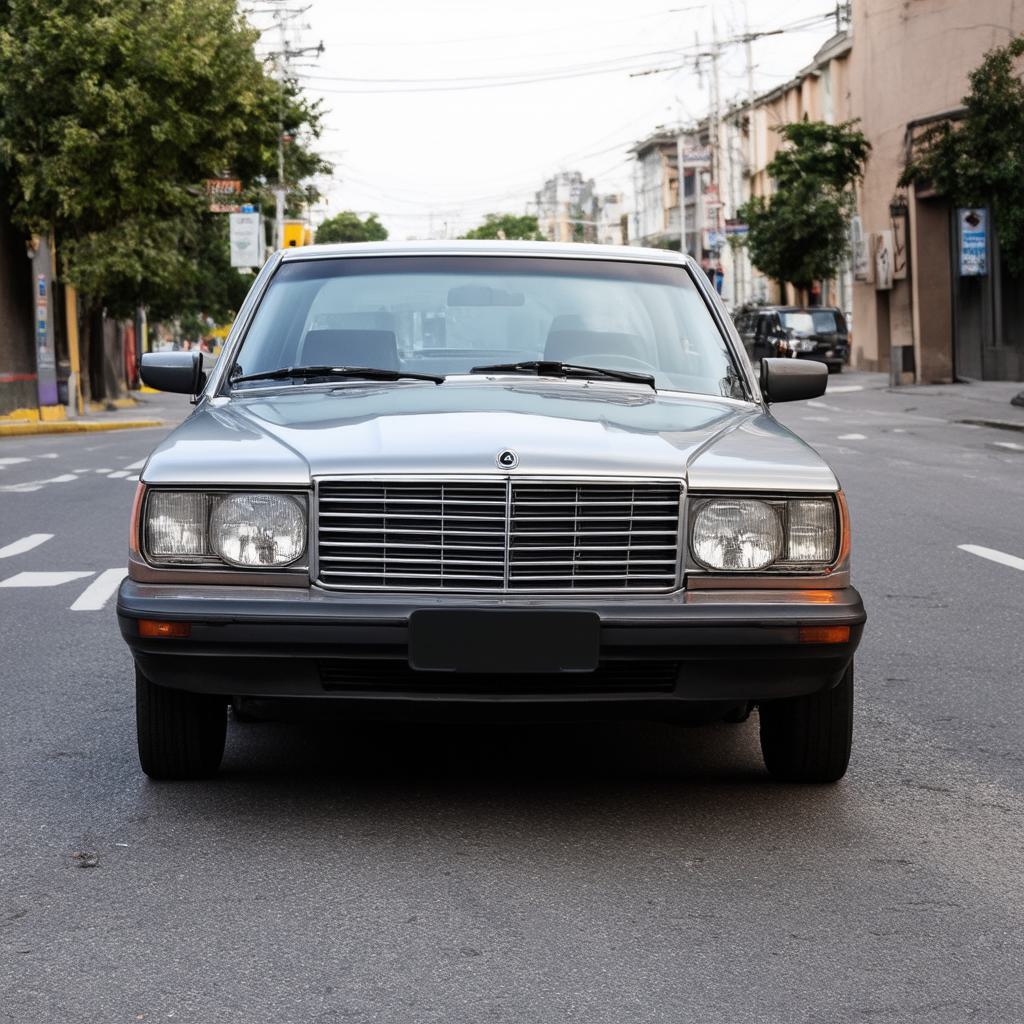} &
        \includegraphics[valign=c, width=\ww]{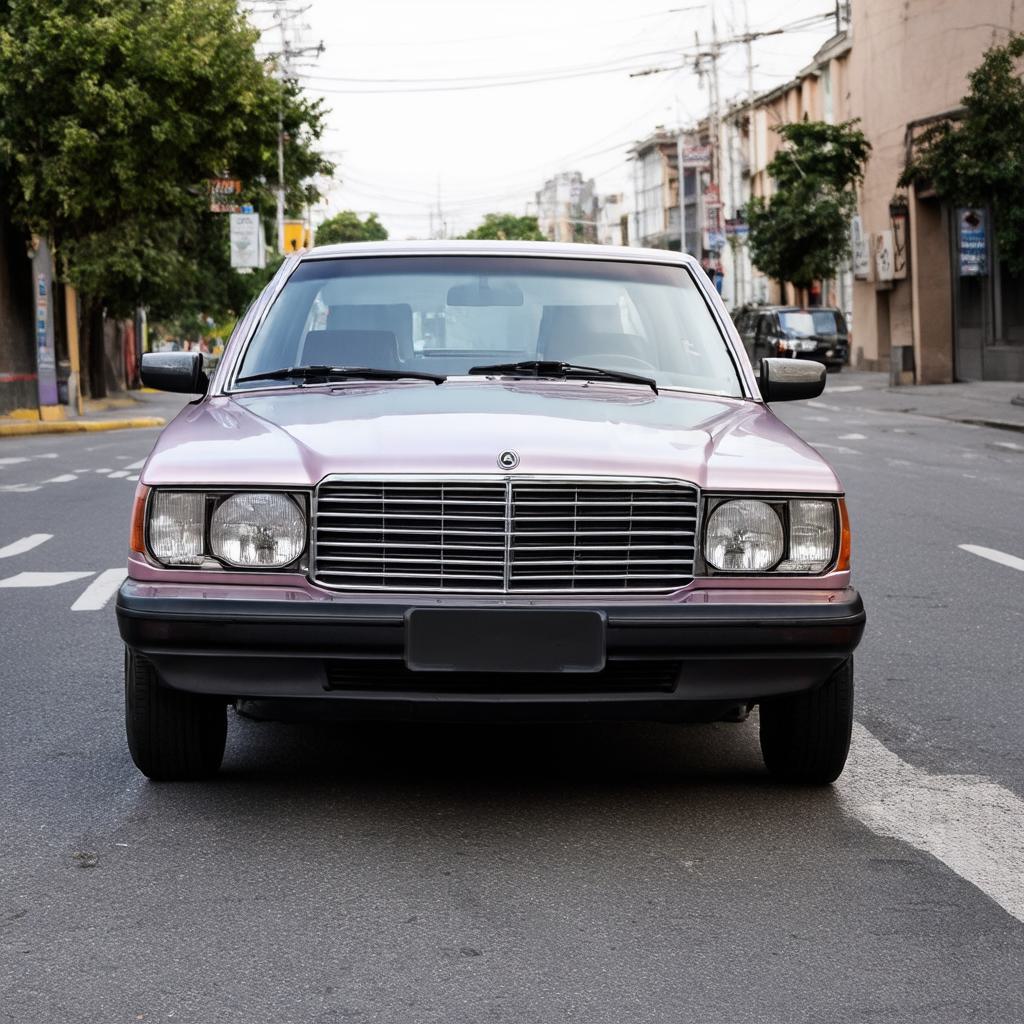} &
        \includegraphics[valign=c, width=\ww]{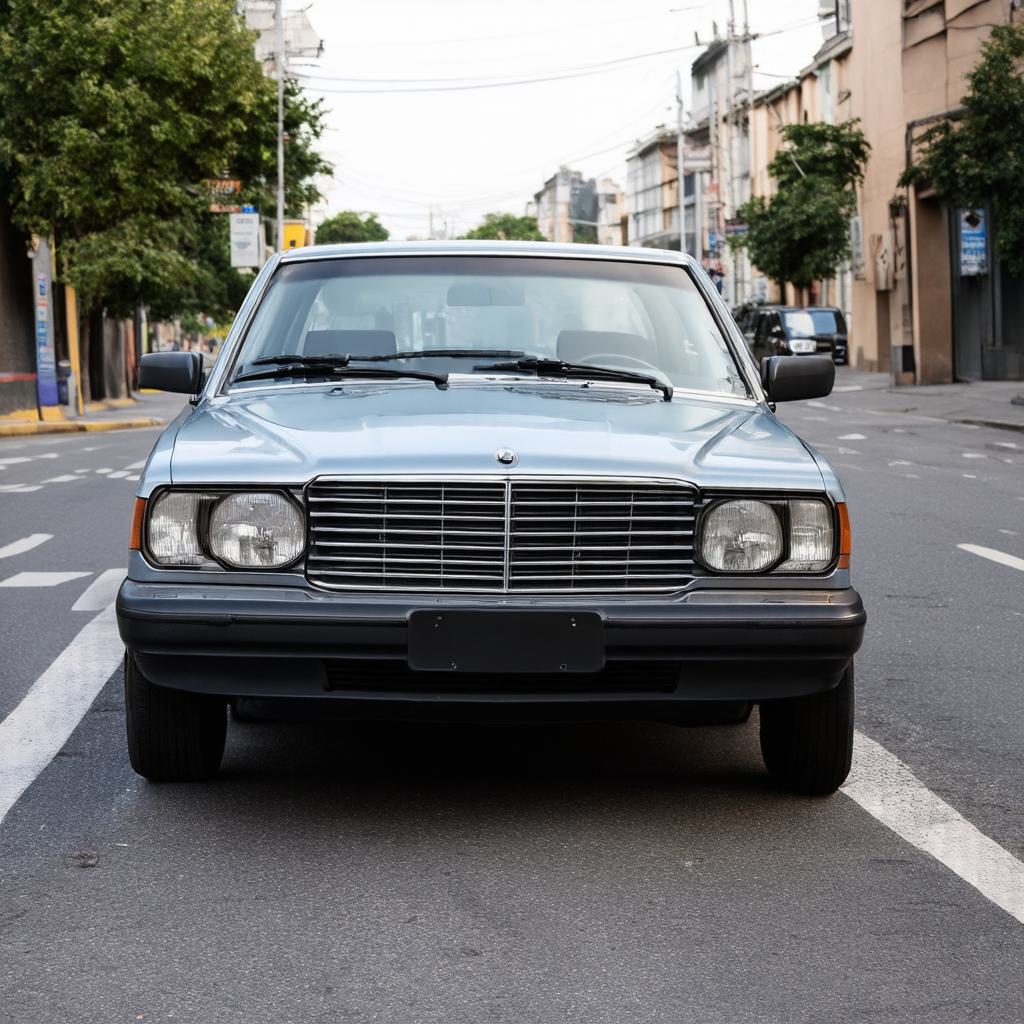} &
        \includegraphics[valign=c, width=\ww]{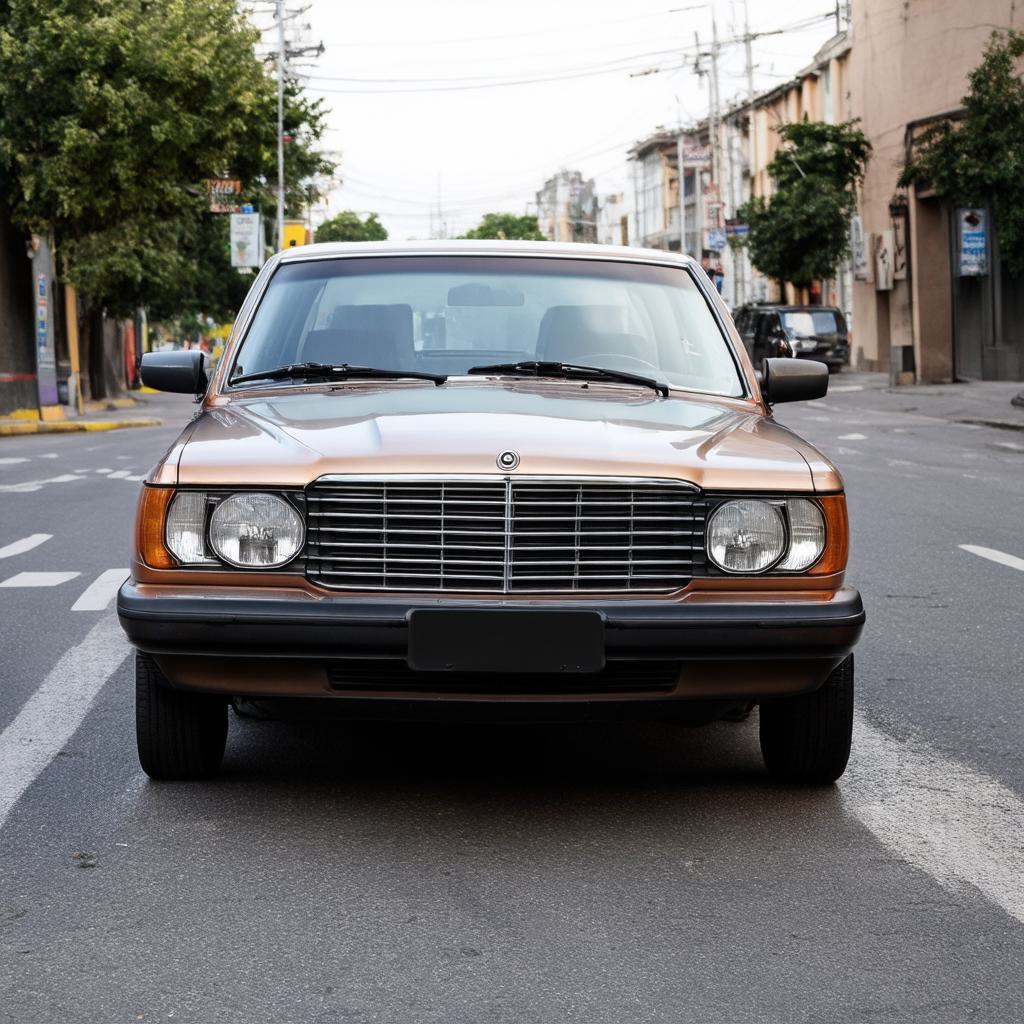} &
        \includegraphics[valign=c, width=\ww]{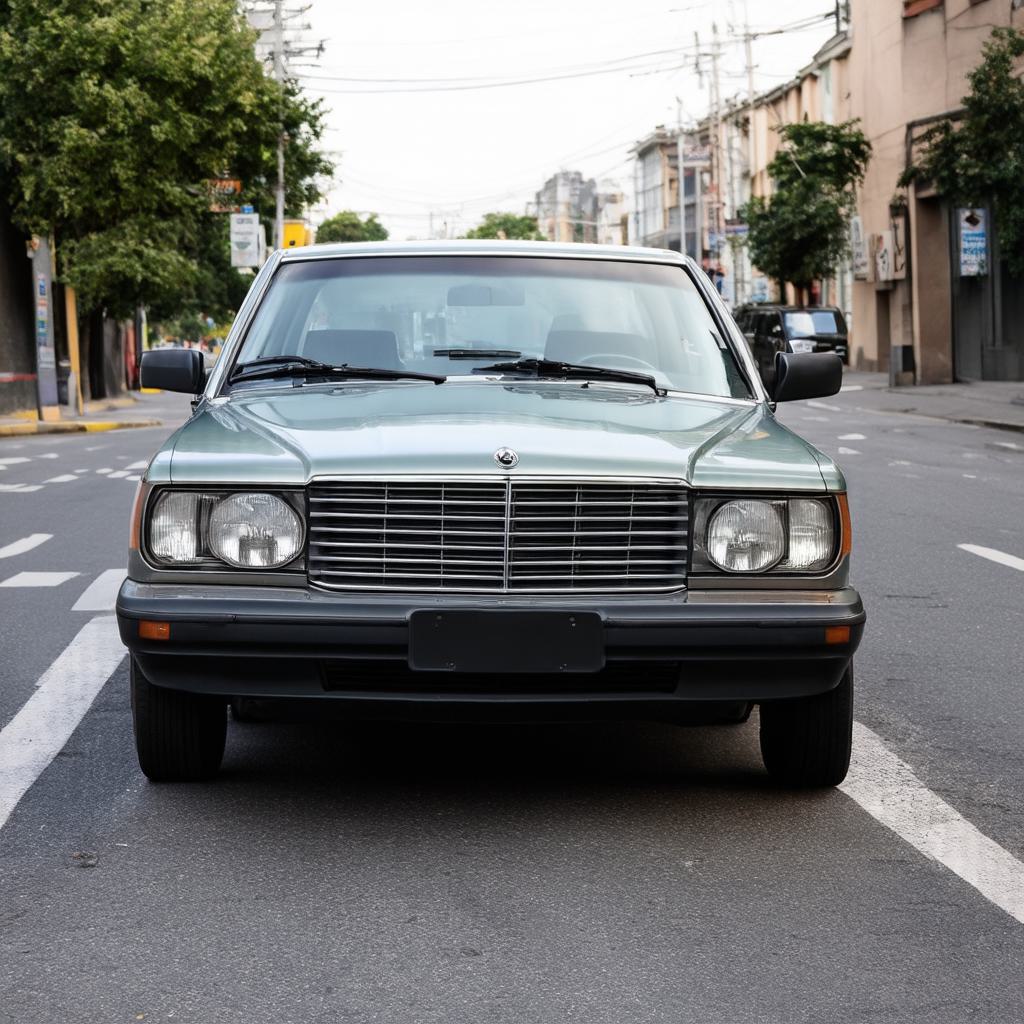}
        \vspace{2px}
        \\

        \small{Input} &
        \small{\prompt{A pink car}} &
        \small{\prompt{A blue car}} &
        \small{\prompt{An orange car}} &
        \small{\prompt{A green car}}
        \vspace{15px}
        \\

        \includegraphics[valign=c, width=\ww]{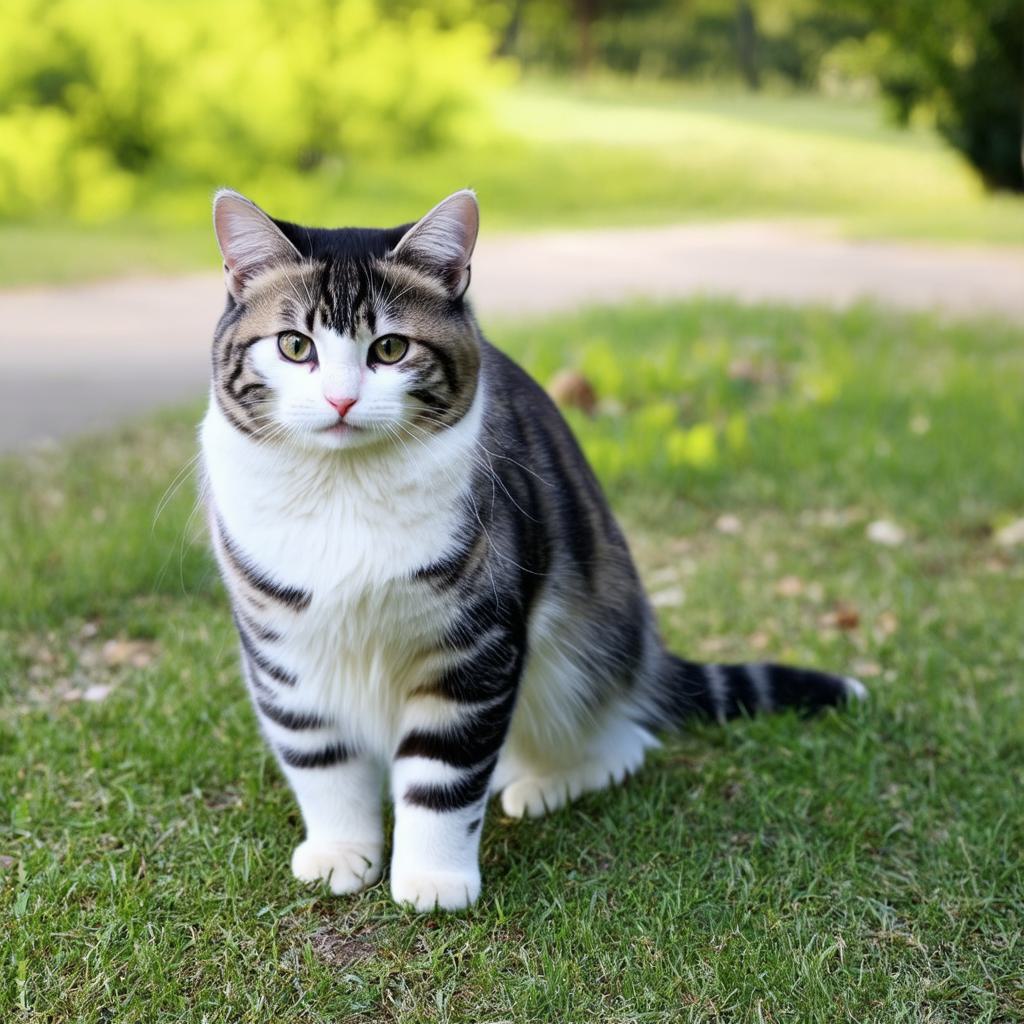} &
        \includegraphics[valign=c, width=\ww]{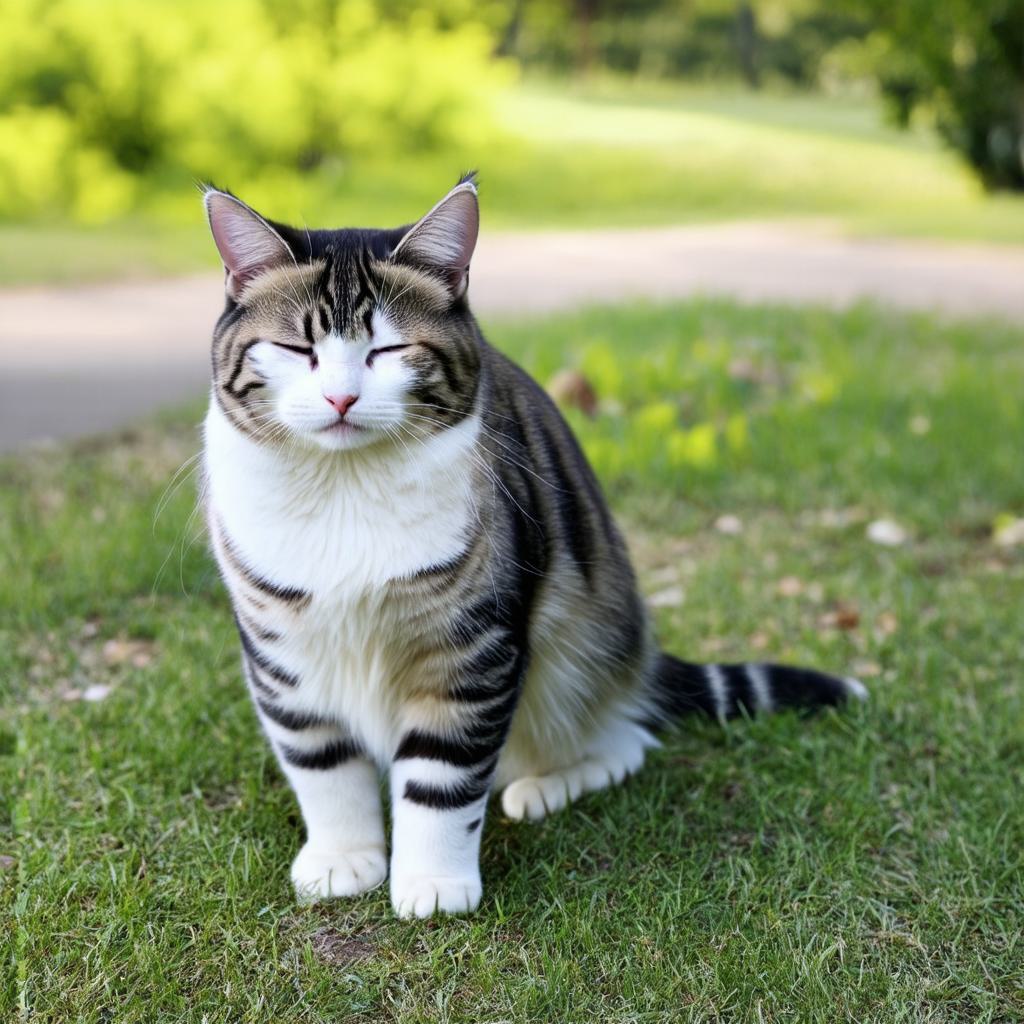} &
        \includegraphics[valign=c, width=\ww]{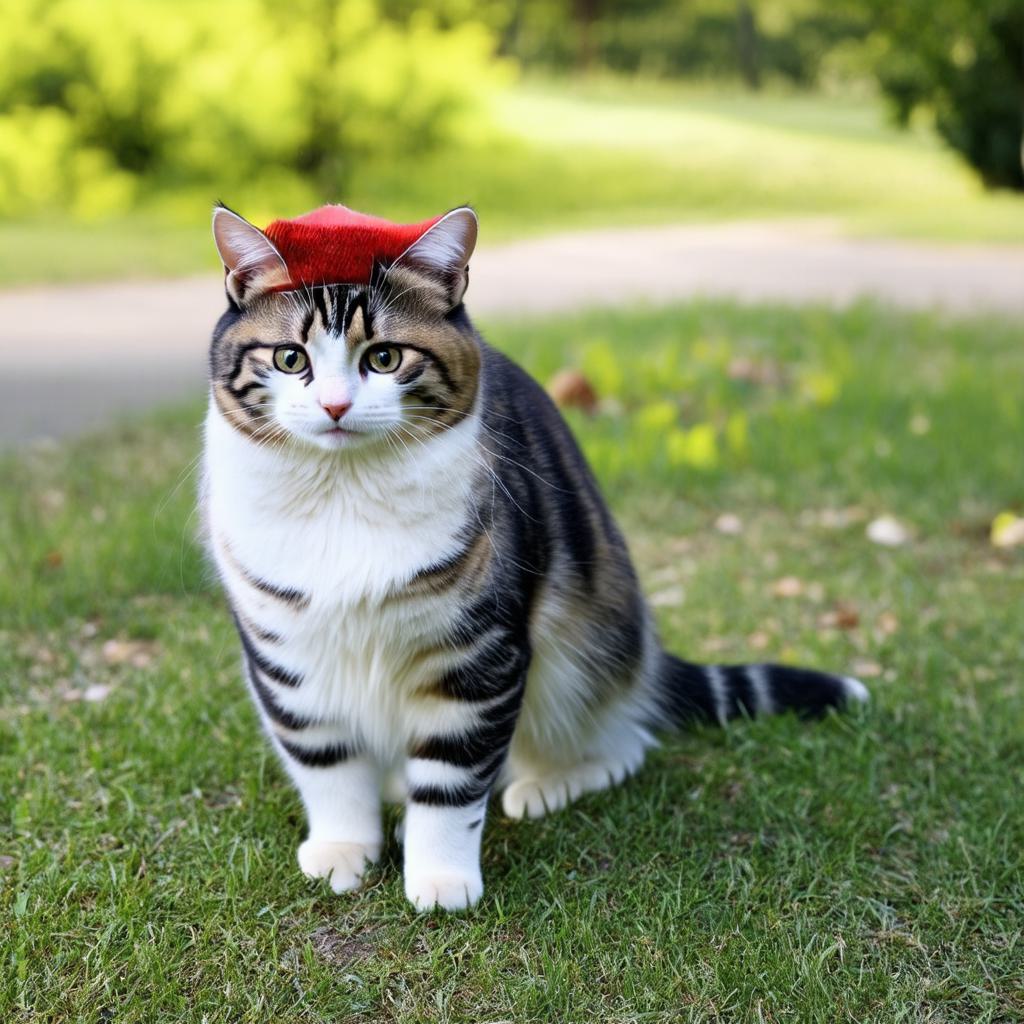} &
        \includegraphics[valign=c, width=\ww]{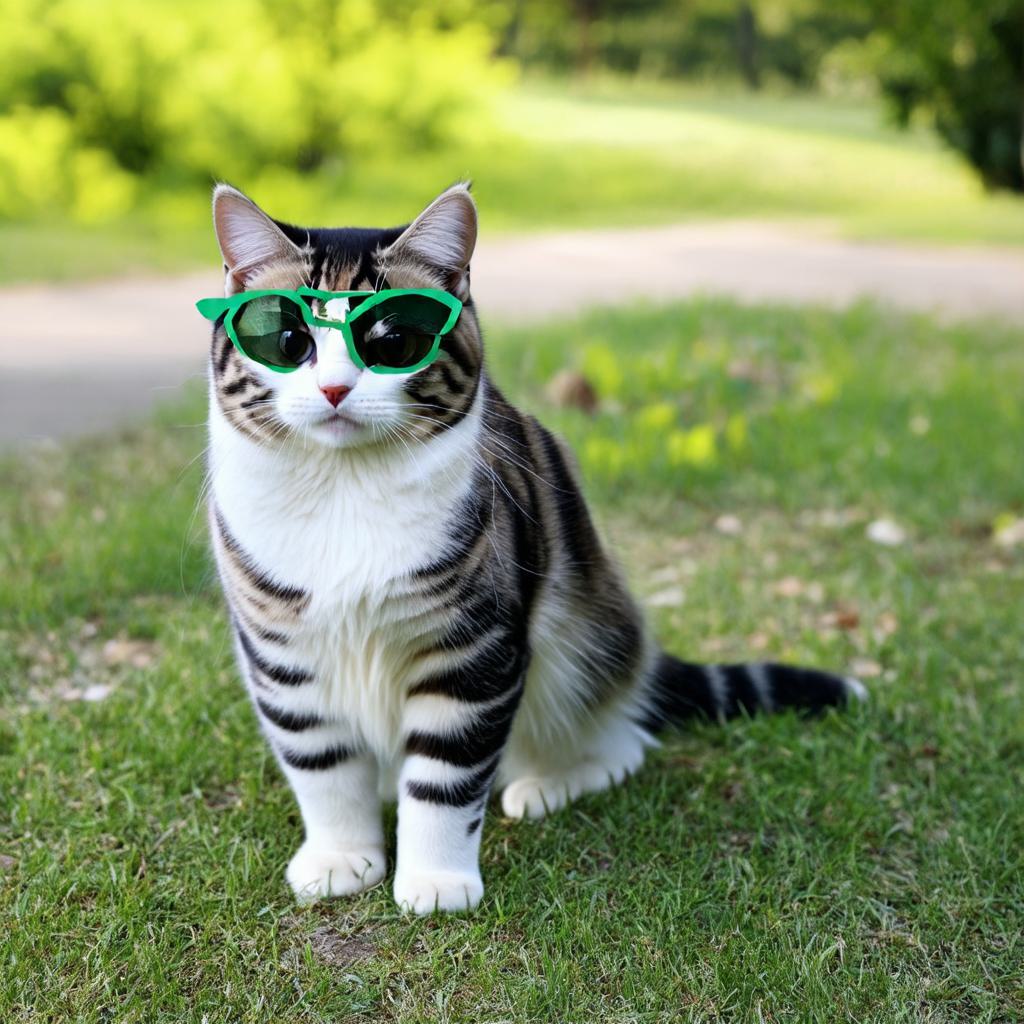} &
        \includegraphics[valign=c, width=\ww]{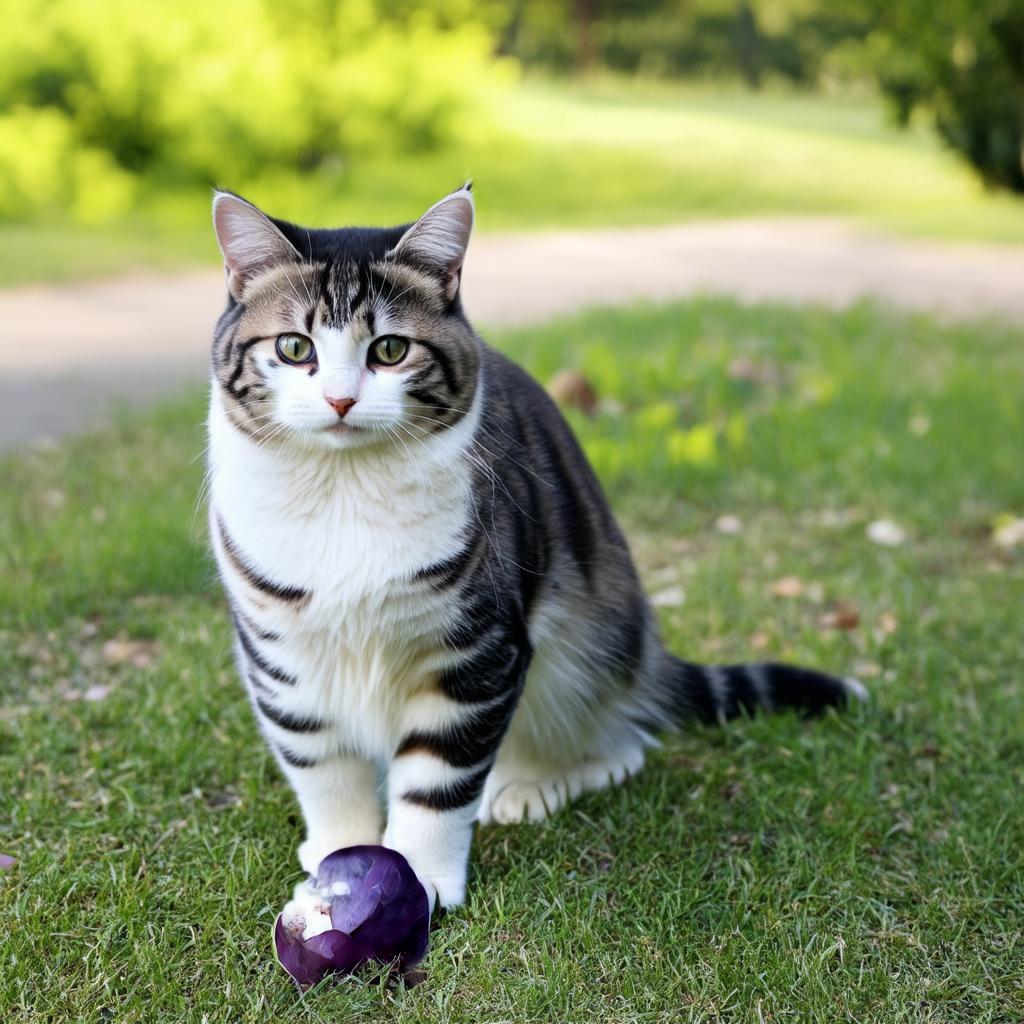} 
        \vspace{2px}
        \\

        \small{Input} &
        \small{\prompt{Closing its eyes}} &
        \small{\prompt{Wearing a red hat}} &
        \small{\promptstart{Wearing green}} &
        \small{\prompt{Next to a purple stone}}
        \\

        &
        &
        &
        \small{\promptend{glasses}} &
        \vspace{5px}
        \\

    \end{tabular}
    \caption{\textbf{Stable Diffusion 3 Editing Results.} As explained in \Cref{sec:sd3_results}, we tested our Stable Flow method on the Stable Diffusion 3 backbone~\cite{Esser2024ScalingRF}. As can be seen, we are able to perform various editing operations using the same mechanism of injecting the reference image information into the vital layers of the model.}
    \label{fig:sd3_editing_results}
\end{figure*}

All the experiments in the main paper were based on the FLUX.1-dev~\cite{flux} model. We also experimented with a different DiT text-to-image flow model named Stable Diffusion 3~\cite{Esser2024ScalingRF} based on the Diffusers~\cite{von-platen-etal-2022-diffusers} implementation of the medium model.

As described in
\Cref{sec:layers_importance},
we measured the importance of each of the layers of this model. As shown in \Cref{fig:layer_removal_quantiative_sd3}, we measured the effect of removing each layer from the model by calculating the perceptual similarity between the generated images with and without this layer. Lower perceptual similarity indicates significant changes in the generated images. As can be seen, removing certain layers significantly affects the generated images, while others have minimal impact.

Next, in \Cref{fig:layer_removal_qualitative_sd3} we illustrate the qualitative differences between vital and non-vital layers. While bypassing non-vital layers ($G_{1}$ and $G_{21}$) results in modest alterations, bypassing vital layers leads to significant changes: complete noise generation ($G_{0}$) or severe distortions ($G_{7}$, $G_{8}$, and $G_{9}$).

Finally, in \Cref{fig:sd3_editing_results}, we perform various editing operations using the same mechanism of injecting the reference image information into the vital layers of the model, as described in
\Cref{sec:image_editing}.

\end{document}